\documentclass{article}
\usepackage[T1]{fontenc}
\usepackage{lmodern}
\usepackage{epigraph}
\let\standardaddcontentsline\addcontentsline
\usepackage[preprint]{colm2025_conference}
\let\addcontentsline\standardaddcontentsline
\setcitestyle{authoryear,round,semicolon,aysep={,},yysep={;}}
\usepackage[margin=1.25in]{geometry}
\usepackage{graphicx} 
\usepackage{xcolor}
\usepackage{needspace}
\usepackage[normalem]{ulem}
\usepackage{newunicodechar}
\newunicodechar{–}{\textendash}
\newunicodechar{—}{\textemdash}
\newunicodechar{‘}{\textquoteleft}
\newunicodechar{’}{\textquoteright}
\newunicodechar{“}{\textquotedblleft}
\newunicodechar{”}{\textquotedblright}
\newunicodechar{→}{\ensuremath{\rightarrow}}
\newunicodechar{π}{\ensuremath{\pi}}

\usepackage{amsmath}
\usepackage{amsfonts}
\usepackage{amssymb}
\usepackage{mathtools}
\usepackage{amsthm}
\usepackage{eucal}

\usepackage{tcolorbox}
\tcbuselibrary{listings,breakable}

\usepackage{comment}
\usepackage[ruled,linesnumbered,noresetcount]{algorithm2e}

\usepackage{float}
\usepackage{caption}
\usepackage{booktabs}
\usepackage{array}
\usepackage{tabularx}
\usepackage{makecell}
\usepackage{adjustbox}
\usepackage{array,booktabs,longtable}
\usepackage{wrapfig}

\usepackage{flafter}

\usepackage{chngcntr}
\counterwithin{figure}{section}
\counterwithin{table}{section}
\numberwithin{equation}{section}

\usepackage{csquotes}
\usepackage{url}
\usepackage{seqsplit}

\usepackage[hidelinks]{hyperref}

\theoremstyle{definition}
\newtheorem{definition}{Definition}[section]

\DeclareMathOperator*{\argmax}{arg\,max}

\newcounter{pisectionanchor}

\title{Physical AI: Concepts, Techniques, and Challenges}
\author{Physical Intelligence Team, Shanghai AI Laboratory\\
\\
physicalai@pjlab.org.cn\\
\\
Preprint version, October 19, 2026
}

\AtBeginDocument{%
  \fancyheadoffset{0pt}%
  \setlength{\headwidth}{\textwidth}%
}

\begin{document}

\maketitle
\begin{abstract}
In recent years, physical artificial intelligence (physical AI) has experienced rapid development, capturing significant attention in academia and industry. However, the field has yet to reach consensus on its definition, technical landscape, and core challenges. To address this gap, this monograph seeks to define physical AI, present its technical landscape, and examine each key technical subfield in depth. In doing so, it establishes a unified framework for Physical AI: artificial systems that perceive, model, and act upon the physical world through continuous feedback under real-world constraints. Further, this monograph distinguishes physical AI from conventional digital AI, embodied AI, cyber-physical systems, and digital twins. It develops the foundations of physical AI across physics, control, cognition, learning, and reasoning. It reviews key methodologies, including multimodal perception, physical data and simulation, world models, embodied foundation models, motor-control systems, multi-agent coordination, and hardware-software co-design. Furthermore, it examines applications spanning robotics, manufacturing, infrastructure, scientific experimentation, and autonomous services. Particular attention is given to sim-to-real transfer, reliability, real-time execution, evaluation, safety, governance, and long-term deployment. By connecting perception, prediction, decision-making, action, and adaptation in a closed loop, this monograph identifies the technical challenges and research directions required to advance from narrow automation toward more generalizable physical AI and, ultimately, physical artificial general intelligence (Physical AGI).
\end{abstract}
\tableofcontents
\clearpage

\clearpage
\part{Overview and Foundations}

\vspace{0.2\linewidth}
\epigraph{\textit{"Pure logical thinking cannot yield us any knowledge of the empirical world; all knowledge of reality starts from experience and ends in it."}}{--- Albert Einstein}

\clearpage 

\section{Introduction to Physical AI}
\label{ch:introduction}

For decades, artificial intelligence (AI) operated inside computers primarily, processing digital data and generating symbolic outputs without directly altering the physical world. Recent advances extend AI beyond this digital boundary by enabling it to perceive and understand physical environments, reason about them, and act within them. This emerging capability has given rise to physical AI, which studies how artificial systems shall interact with the physical world. Yet, what exactly is physical AI? What does its technical landscape look like, and what core challenges does it entail? And, what are typical application scenarios of physical AI? This chapter addresses these questions by tracing AI's expansion from digital to physical space, defining physical AI and its boundaries with neighboring fields, and examining its core characteristics, system architectures, and application domains.

\begin{figure}[t]
  \centering
  \includegraphics[width=\textwidth]{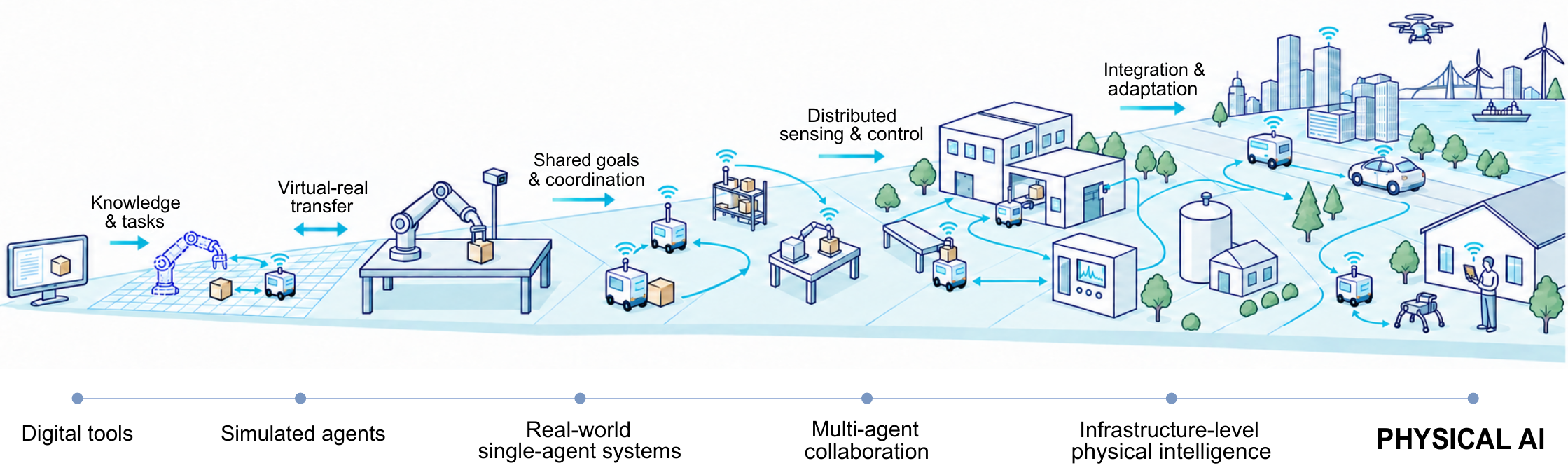}
  \caption{A conceptual expansion in the scope and organization of AI interaction with the physical world, from digital tools and simulated agents to real-world single-agent systems, multi-agent collaboration, and infrastructure-level physical AI. The real-world forms shown fall within the scope of physical AI adopted in this monograph}
  \label{fig:physical ai teaser}
\end{figure}

\subsection{Background: A Paradigm Expansion of AI}
\label{sec:physical-ai-background}
\subsubsection{Digital AI and Physical AI}

Artificial intelligence has developed through several complementary traditions, from symbolic reasoning and knowledge representation to statistical learning and deep neural networks. Deep learning substantially advances the ability of computational systems to extract useful representations from complex data, with major improvements in speech recognition, visual recognition, and related information-processing tasks. These achievements establish representation learning as a central methodological foundation of contemporary AI~\citep{lecun2015deep}.

In this chapter, \emph{Digital AI} refers to capabilities that operate directly on digital representations, such as text, images, software, and structured records. This definition emphasizes the system's operational interface over the origin of its data. For example, while an image may depict a physical scene, recognizing its contents does not require the system to actually interact with that environment. \emph{Physical AI}, in contrast, applies intelligence to perceive, influence, and adapt to physical processes through direct interaction. Definition~\ref{def:physical-ai} in Section~\ref{sec:physical-ai-definition} further clarifies this fundamental distinction.

The distinction becomes apparent when comparing the description of an action with its execution. An AI system may correctly explain how to pick up a glass while remaining unable to estimate its mass, establish a stable grasp, detect incipient slip, or adjust contact forces. Successful manipulation depends on relationships among geometry, material properties, sensing, and actuation that cannot be assessed solely through the linguistic plausibility of a response. The relevant criterion is whether the system produces the intended physical outcome under the conditions encountered.

This distinction is related to the symbol grounding problem of how symbols can acquire meaning through a system's own capacities rather than solely through interpretation by an external observer. The formulation by~\citet{harnad1990symbol} identifies the limitations of defining symbols only through other symbols and motivates their connection to nonsymbolic representations. For physical AI, the corresponding engineering problem is to connect concepts such as \emph{support}, \emph{contact}, and \emph{stability} to observations and action consequences.

Grounding should not, however, be reduced to a categorical claim that digital models possess no physical knowledge. Models trained on descriptions, images, videos, or simulated trajectories can acquire useful regularities concerning the physical world. The unresolved issue is whether these regularities support reliable intervention across changes in objects, environments, and operating conditions. Semantic competence, predictive accuracy, and physical competence are related but distinct properties.

Figure~\ref{fig:physical ai teaser} summarizes this expansion from digital tools and simulated agents to real-world single-agent systems, multi-agent collaboration, and infrastructure-level physical AI. Across these physical deployments, the unit of analysis extends from an isolated model to the integrated system of computation, sensing, actuation, communication, and environment. This perspective has important historical precedents.~\citet{brooks1991intelligence} demonstrates an approach to robotics in which intelligent behavior arises through direct perception-action coupling, challenging the assumption that autonomous behavior must depend on a centralized symbolic description of the world. Contemporary physical AI extends this longstanding concern with situated action by incorporating learned representations, scalable training, and predictive models.

\subsubsection{Core Characteristics}

Physical AI operates under constraints that cannot be removed by increasing model capacity alone. Three characteristics are particularly important, namely, consequential interaction under uncertainty, perception of evolving physical states, and the integration of empirical learning with physical structure.

\paragraph*{Consequential Interaction Under Uncertainty}
Physical systems are only partially observable. Object properties, external disturbances, and equipment conditions may be unknown or change during execution. Actions also affect the evidence available to the system. Moving a camera changes visibility, probing a surface reveals compliance, and applying force may alter an object's configuration. Perception and action consequently form an interdependent process.

Physical interaction can incur recovery costs and irreversible consequences. A simulated episode can often be restarted from a saved state, whereas a spilled chemical, damaged component, or contaminated sample may require substantial intervention or may be impossible to restore. The operational distinction is that physical systems generally lack an exact, inexpensive reset mechanism. Planning must therefore consider the recoverability of future states as well as the likelihood of immediate task success.

\paragraph*{Perception of Evolving Physical States}
Physical AI requires observations that are temporally and spatially meaningful. Vision can describe appearance and geometry, while force, touch, proprioception, temperature, pressure, and other measurements reveal complementary aspects of interaction. A manipulation system, for example, may need to distinguish visual contact from load-bearing contact and a completed motion from a successfully transferred object.

Although sensors typically produce sampled digital measurements, the processes they measure evolve in continuous time and may include discrete events such as impact, grasp release, or equipment switching. Reliable interpretation therefore depends on sampling rates, calibration, synchronization, coordinate frames, and measurement uncertainty. A correct estimate obtained too late may be unusable for control. More generally, computation and physical dynamics must be analyzed together because execution timing can affect system behavior~\citep{lee2008cyber}.

\paragraph*{Learning Informed by Physical Structure}
Data provide evidence about real operating conditions, including effects that are difficult to model analytically. Physical knowledge provides structure that can restrict implausible predictions and reduce the burden of learning. Conservation relationships, symmetries, geometric constraints, and governing equations can be incorporated through model architectures, training objectives, simulation, or constrained decision procedures. Physics-informed machine learning formalizes several ways of combining observations with mathematical descriptions of physical processes~\citep{karniadakis2021physics}.

The appropriate balance depends on the task. Explicit models may be effective when governing relationships are sufficiently understood, whereas learned components can represent uncertain interactions or compensate for model mismatch. Physical priors are themselves conditional on assumptions and operating regimes. For example, an inaccurate friction model or an inappropriate rigid-body approximation can produce systematic errors. Physical consistency therefore requires validation within the task's operating regime, regardless of the model family.

Together, these characteristics shift evaluation toward closed-loop performance. Relevant measures include task completion, robustness to disturbance, response time, resource consumption, and recovery from failure. Prediction accuracy remains valuable, but its practical significance depends on how prediction errors propagate through decisions and physical execution.

\subsubsection{Why Physical AI Is Needed and Increasingly Feasible}

Physical AI is motivated by tasks that require systems to act on the physical world.  Flexible manufacturing requires adaptation to variations in parts and processes. Transportation and energy systems require coordinated responses to changing demand and operating conditions. Scientific experimentation requires procedures to be adjusted in response to measurements. Furthermore, extreme environments, such as deep space and the deep sea, drive the need for local autonomy, as communication constraints and physical hazards severely limit continuous human intervention.

These demands do not imply that established automation is inadequate in every setting. Carefully engineered automation remains effective where tasks and operating conditions can be specified in advance. The additional research challenge is to preserve useful performance when relevant conditions change and exhaustive preprogramming becomes impractical. Physical AI addresses this challenge by integrating interactive learning with perception, inference, prediction, and adaptation.

Several technical developments support this direction. Sensing and local computation provide the practical interfaces for observing and influencing physical processes. Simulation supplies controlled environments for training and evaluation, while methods for transfer from simulation to reality reduce dependence on collecting every training example on physical hardware. For example, \citet{tobin2017domain} demonstrate that varying simulated visual conditions can support transfer of a vision model to real robotic tasks. Such results establish a useful transfer mechanism, while leaving its effectiveness dependent on the relationship between training variation and deployment conditions.

Learned physical models provide another foundation. Graph network simulators have demonstrated the ability to represent interacting particles and predict dynamics across fluids, rigid solids, and deformable materials within evaluated settings. These models illustrate how structured learning can support physical prediction beyond a single fixed configuration~\citep{sanchezgonzalez2020learning}. For autonomous action, their potential value lies in evaluating candidate interventions before execution, provided that prediction errors and operating limits are understood.

Large pretrained models also offer reusable semantic knowledge. For example, RT-2 shows that vision-language pretraining can be combined with robot demonstrations to support robotic action and improve generalization in the reported experiments~\citep{brohan2023rt2}. This provides evidence that capabilities developed through digital learning can contribute to physical interaction when connected to appropriate action data and interfaces.

These developments make it increasingly feasible to combine digital capabilities, physical models, and control mechanisms to study how systems acquire transferable physical skills while maintaining measurable performance within specified operating conditions.

\subsection{Physical AI: Definition and Boundaries}
\label{sec:physical-ai-definition}

\subsubsection{A Broad Definition of Physical AI}

At its most general, physical AI concerns the capacity of an artificial system to interact with the physical world. This monograph uses \emph{physical AI} for this artificial-system scope. The term \emph{physical intelligence} is retained for general capacities, bodily mechanisms, and source-specific usage. The following definition makes the notion precise.

\begin{definition}[Physical AI]\label{def:physical-ai}
Physical AI is the capacity of an artificial system to perceive physical environment states, maintain task-relevant representations of physical relationships and action consequences, and select and execute interventions through physical interfaces, using feedback to achieve objectives under physical, operational, and computational constraints.
\end{definition}

\begin{figure}[ht]
  \centering
  \includegraphics[width=\textwidth]{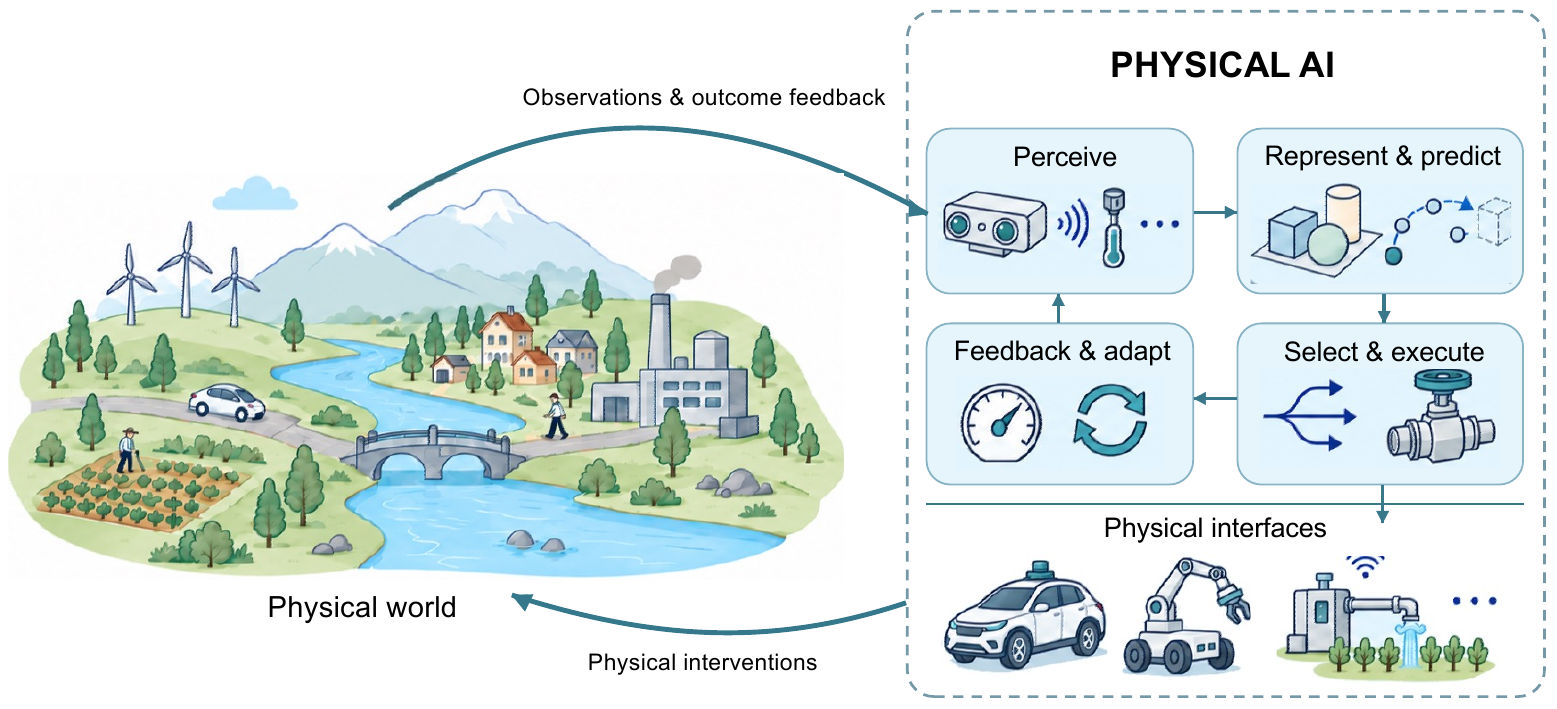}
  \caption{Physical AI as closed-loop interaction between artificial systems and the physical world. Systems observe physical states, maintain task-relevant representations of physical relationships and action consequences, and select and execute interventions through physical interfaces. Outcome feedback guides adaptation toward objectives under physical, operational, and computational constraints}
  \label{fig:physical ai difinition}
\end{figure}

This definition identifies a functional scope rather than a particular architecture. Representations may be explicit, implicit, or distributed across several components. They need not constitute a complete internal simulator, and action selection need not depend on symbolic reasoning. What matters is whether the integrated system can use information about physical conditions to guide and revise its behavior.

Four implications clarify the definition. First, physical grounding requires observations to be connected to relevant quantities, events, or conditions in the world. Second, the system must have a path for intervention, whether through direct actuation or commands to controlled equipment. Third, feedback must inform subsequent behavior, allowing discrepancies between expected and observed outcomes to affect execution. Fourth, claims of intelligent capability should be supported by evidence of task-relevant inference, prediction, planning, or adaptation, rather than inferred merely from the presence of sensors and actuators.

These criteria define a continuum of capability. A fixed feedback regulator and a system that adapts across changing tasks may share essential control principles, yet differ substantially in the breadth of conditions they can handle. The distinction between automation and Physical AI is therefore most useful when expressed through demonstrated capabilities and operating assumptions, rather than an absolute technological boundary.

Terminological care is also necessary because \emph{physical intelligence} has an established, more specific usage in mechanics and robotics.~\citet{sitti2021physical} uses the term to describe functions such as sensing, actuation, memory, and computation physically encoded in an agent's body. The broader definition of Physical AI adopted here includes such contributions while also encompassing computational and distributed systems that interact with physical processes. It should be understood as this chapter's organizing framework, rather than a claim that the literature has converged on one definition.

Finally, key components should be distinguished from a complete system. A learned simulator, perception model, or action planner may provide an essential Physical AI capability. In isolation, however, it does not satisfy the system-level requirement for closed-loop physical intervention. This distinction allows purely digital modules to be recognized as enabling technologies without attributing autonomous physical competence to them prematurely.

\subsubsection{Physical AI vs. Embodied AI}

\begin{figure}[t]
  \centering
  \includegraphics[width=\textwidth]{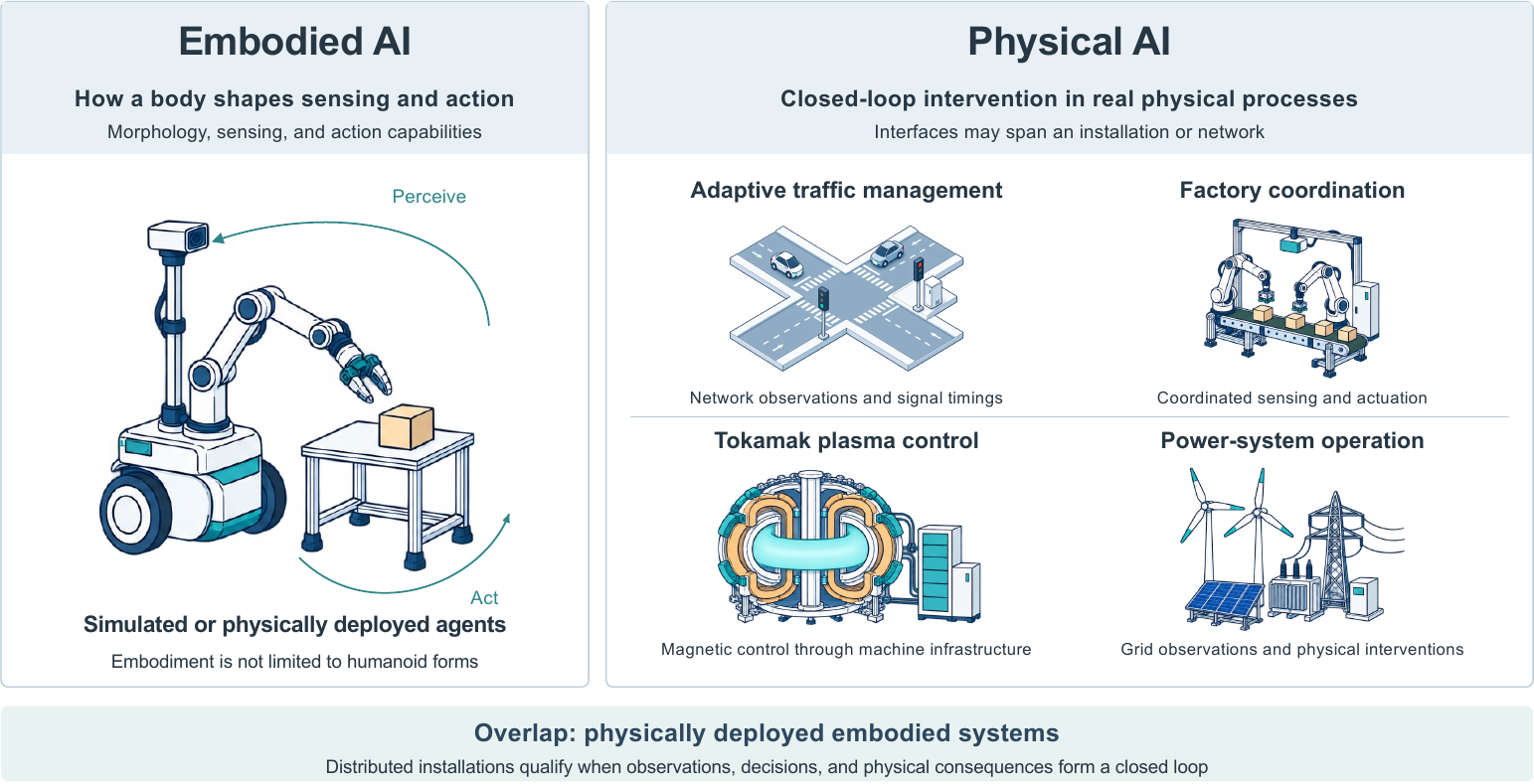}
  \caption{Embodied AI and physical AI within this monograph's framework. Embodied AI emphasizes how an agent's body shapes sensing and action. Physical AI emphasizes closed-loop intervention in real physical processes. The two overlap in physically deployed embodied systems}
  \label{fig:embodied-ai-vs-physical-ai}
\end{figure}

Embodied AI posits that artificial intelligence requires a \emph{body} to house the system and drive its active sensorimotor relationship with the environment~\citep{pfeifer2004embodied}. The concept of embodiment focuses on how this morphology, along with sensing and action capabilities, directly shapes environmental interactions and guides behavior. Importantly, embodiment is by no means limited to humanoid or biologically inspired designs. For physically deployed agents, research further highlights how mechanical structures and material properties can intrinsically support adaptive behavior, thereby reducing the control burden on a centralized controller~\citep{pfeifer2007self}.

Nor does embodied AI necessarily require a deployed physical robot. Simulated agents are widely used to study navigation, interaction, and learning under embodied constraints~\citep{shridhar2020alfworld}. Habitat, for example, was introduced as a platform for training and evaluating virtual embodied agents in three-dimensional environments~\citep{savva2019habitat}. Embodied AI therefore includes both simulated and physically deployed agents.

Within this chapter's framework, physically deployed embodied AI constitutes a major realization of physical AI. Its characteristic analytical focus is the agent and its particular sensing and action capabilities. Physical AI adopts an additional system-level perspective, encompassing arrangements in which sensing, decision-making, and actuation are distributed across infrastructure. Figure~\ref{fig:embodied-ai-vs-physical-ai} summarizes these distinctions and their overlap.

Consider an adaptive traffic-management system that estimates network conditions, changes signal timings, and evaluates the resulting traffic flows. Its interaction with the world is mediated by distributed detectors, communication links, and signal controllers rather than a single mobile body. Similar arrangements can be considered for factory coordination or power-system operation, provided that the system actually closes the loop between observations, decisions, and physical consequences. These are illustrative applications of the proposed definition, not claims that every traffic controller, factory scheduler, or grid-management platform possesses advanced intelligence.

Scientific equipment provides a concrete research example. Degrave and colleagues demonstrated deep reinforcement learning for magnetic control of plasma configurations in the TCV tokamak. The controller interacts with a physical process through the machine's measurement and actuation infrastructure~\citep{degrave2022magnetic}. Under the present framework, this qualifies as Physical AI even though the system lacks the conventional form of a mobile or manipulative robot.

Such systems are not independent of physical embodiment in the broadest sense, because sensors, actuators, and material processes remain indispensable. Their distinguishing feature is that the physical interface is distributed across an installation or network. Within this monograph's framework, embodied AI emphasizes how an agent's body shapes sensing and action, while physical AI emphasizes closed-loop intervention in real physical processes, including through distributed infrastructure. The two overlap in physically deployed embodied systems.

\subsubsection{Physical AI vs. Cyber-Physical Systems and Digital Twins}
\label{sec:physical-ai-cps-twins}

Physical AI intersects closely with cyber-physical systems (CPS), but the concepts describe different aspects of a system. CPS concerns the integration of computation with physical processes, including the consequences of timing, concurrency, and interaction between software and its physical environment~\citep{lee2008cyber}. Physical AI concerns the capabilities exercised through that integration, including interpreting conditions, anticipating consequences, choosing interventions, and adapting behavior.

A CPS may operate effectively through fixed control logic without possessing broad inference or adaptation capabilities. Conversely, a physically intelligent system depends on the engineering properties of its computational and physical infrastructure, including dependable sensing, timely communication, and executable control. Physical AI can therefore be understood as a capability layer within many CPS architectures. Describing it as a possible development toward greater autonomy is useful, provided that this does not imply that CPS research excludes advanced cognition or that learning automatically improves an engineered system.

Digital twins occupy a different position. They provide digital representations associated with identifiable physical counterparts, with connections that support updating and comparison between physical and digital states. The literature identifies multiple characteristics and implementation choices, making it important to specify how a proposed twin is connected to its physical counterpart~\citep{jones2020characterising}.

Within a Physical AI architecture, a predictive digital twin can support state estimation, diagnosis, and evaluation of alternative interventions. It may help estimate how a production line would respond to a schedule change or how equipment behavior would change under another operating condition. Such uses require more than a visually accurate replica. The representation must capture the variables and dynamics relevant to the decision.

A digital twin does not necessarily select or execute actions autonomously, and a Physical AI system need not maintain a comprehensive twin. A compact learned dynamics model may be sufficient for a particular control task. Likewise, a generic world model becomes a digital twin only when it is meaningfully associated with a specific physical counterpart and maintained through an appropriate update relationship. Counterfactual predictions remain conditional on the validity of the model and its assumptions.

These distinctions establish complementary roles. CPS describes the integration of computational and physical processes. Embodied AI examines intelligence through situated sensorimotor interaction. Digital twins provide representations tied to physical counterparts. Physical AI, in the framework developed here, identifies the capacity to combine perception, physical understanding, decision-making, and feedback into effective intervention. This system-level perspective provides the basis for the architectural elements discussed in the following sections.

\subsection{Core Elements of Physical AI}
\label{sec:core-elements}

\begin{figure}[!b]
  \centering
  \includegraphics[width=\textwidth]{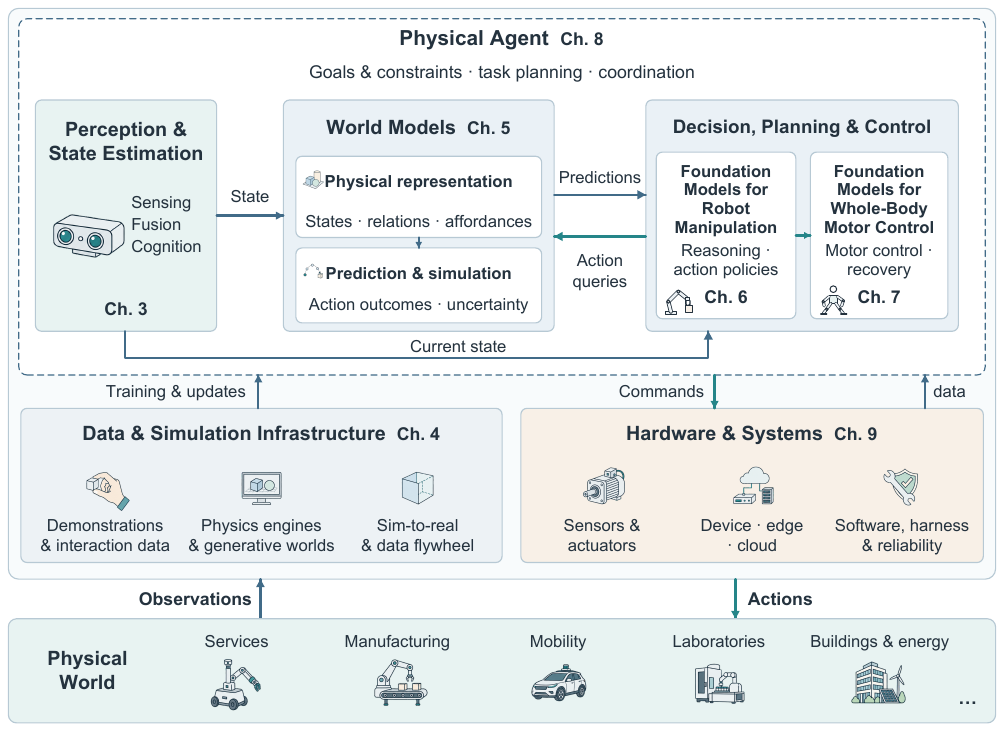}
  \caption[Key elements of Physical AI and their relationships]{Key elements of Physical AI and their relationships. Chapter labels in the diagram refer to Chapters~\ref{ch:perception} to \ref{ch:hardware-systems}}
  \label{fig:physical-ai}
\end{figure}

Definition~\ref{def:physical-ai} characterizes Physical AI as the capacity to perceive physical states, maintain task-relevant representations, and execute interventions through feedback. In practice, this capacity takes the form of a persistent closed loop. Information processing and decision-making must be translated into physical actions, and subsequent behavior must be revised according to observed outcomes. Physical AI therefore operates through a cycle of perception, modeling, prediction, decision-making, execution, and feedback, rather than through a one-shot mapping from input to output.
 
As Section~\ref{sec:physical-ai-definition} emphasizes, Physical AI is not synonymous with robotic intelligence. Whether a system exhibits it depends on its ability to close the loop between perception and action, not on its morphology. The elements discussed in this section apply equally to autonomous vehicles, industrial equipment, automated experimentation platforms, intelligent buildings, and energy infrastructures. In such systems, perception, computation, decision-making, and execution are often distributed across devices, edge nodes, cloud platforms, and physical facilities rather than concentrated within a single device.
 
Operating in a continuously changing and uncertain world also means that actions are constrained by dynamics, geometry, time, energy, materials, and safety requirements, and that their consequences are not always reversible, as Section~\ref{sec:physical-ai-background} discusses. A physically intelligent system must therefore answer not only ``What should be done?'' but also ``Can it be done under the current conditions?'', ``How can it be done safely?'', and ``Does the execution produce the intended outcome?'' Organized around these questions, Physical AI comprises four closely related elements, including physical state perception and estimation, physical world models, physical decision-making together with planning and control, and closed-loop feedback and multiscale adaptation. They correspond respectively to perception, representation, intervention, and feedback in the definition. The first three together constitute the physically intelligent agent, while the fourth connects the agent with the physical world into a continuously running loop. The relations among these elements, together with the data, hardware, and physical world on which they depend, are illustrated in Figure~\ref{fig:physical-ai}.

\subsubsection{Physical State Perception and Estimation}
\label{sec:perception}

Physical AI begins with information about the environment, task-relevant objects, and the system itself. Vision conveys identity, pose, geometry, and motion. Force and tactile sensing reveal contact, friction, and material properties~\citep{dahiya2009tactile}, while acoustic sensing detects collisions and vibrations. Proprioception measures the system's own configuration and load, and industrial systems add process variables such as temperature, pressure, and current. Chapter~\ref{ch:hardware-systems} treats the sensors themselves. The concern here is turning their signals into an estimate of physical state.

Measurements are incomplete, noisy, delayed, and often misaligned across modalities, so calibration, multimodal fusion, and state estimation are needed to obtain a temporally consistent state with explicit uncertainty~\citep{Kalman1960,baltruvsaitis2018multimodal}. Perception also supports active information acquisition. When observations are insufficient for a decision, the system can change its viewpoint, reposition sensors, or probe objects to estimate mass, stiffness, or friction, folding information acquisition into the decision loop~\citep{bajcsy1988active}.


\subsubsection{Physical World Models: Representation and Prediction}
\label{sec:world-model}

State estimates are organized into a world model, which describes the environment, the actions available to the system, and how physical states evolve over time~\citep{ha2018world,lecun2022path}. It has two interdependent aspects. Representation captures what the model holds about the current world, and prediction anticipates what the world becomes after a given action.

Physical representations cover objects, poses, geometry, and spatial relations, together with velocities, masses, contacts, forces, material properties, affordances, and task constraints. A cup is at once a shape with a pose and an object that can be grasped, holds liquid, and spills when tilted. Representations may be explicit (maps, object lists, equations of motion, relational graphs), implicit (learned from data), or hybrid, combining physical priors with learning to retain interpretability while correcting for unmodeled effects. An effective world model is updated with each new observation and distinguishes directly observed, estimated, and unobserved states.


Prediction estimates the consequences of a particular action, covering trajectories, contact outcomes, equipment conditions, task success, and safety hazards over time scales from milliseconds to hours. Predictions may come from physical equations and simulators~\citep{raissi2019physics} or from transition patterns learned from data~\citep{sanchezgonzalez2020learning}. Practical systems combine the two and revise predictions and their uncertainty online. In scenario simulation, the decision process submits candidate actions and the world model returns predicted outcomes with their uncertainty, enabling counterfactual comparison before anything is executed. This simulation runs online, unlike the physics engines and generative environments of Chapter~\ref{ch:physical-data}, which are used offline to produce training data and test policies.


\subsubsection{Physical Decision-Making, Planning, and Control}
\label{sec:decision}

Physical decision-making selects actions from task objectives, the current state, predicted outcomes, and real-world constraints, usually through a hierarchy~\citep{kaelbling2011hierarchical,sutton1999options}. At the top, the system sets goals and constraints and, when several physical agents share a task, allocates and coordinates subtasks among them. Foundation models for robot manipulation reason about the task and scene, decompose goals into executable steps, and produce action policies. Foundation models for whole-body motor control provide reusable feedback policies for executing movement goals and correcting deviations~\citep{zeng2025behavior,zeng2026scaling}. The levels differ mainly in time scale, from minutes or hours for task allocation to seconds for policies and milliseconds for motor commands. The boundary separating task-level reasoning and action-policy generation from motor control and feedback correction varies across systems.

Execution takes many forms, from joints and end effectors to steering and braking, instrument scheduling, or valves, pumps, and switches. An action is thus best understood as an observable intervention that changes a physical object, environment, or process. Such interventions must be feasible, safe, and timely. A logically valid plan can still fail because of collisions, actuator limits, or latency, so safety checks precede execution, and monitoring and emergency stops stay active throughout.


\subsubsection{Closed-Loop Feedback and Multiscale Adaptation}
\label{sec:closed-loop}

Physical AI revises behavior according to the observed consequences of action. After each command, the system perceives again, checks whether the intended result is achieved, and updates its state estimate together with its world model and plan. When outcomes deviate from predictions, it identifies the cause and compensates, retries, replans, or terminates safely.

The loop resides in no single module but in the connections among the elements. Perception supplies the state, the world model answers action queries from the decision process, commands act on the world through actuators, and sensors return the consequences to perception (Figure~\ref{fig:physical-ai}). The loop thus closes through the physical world rather than inside the agent. Fast reactive control, such as tightening a grip when an object begins to slip, may bypass the world model and respond directly to the current state.

Feedback operates at three nested scales. The fastest loop connects motor control and feedback correction with the actuators and sensors, adjusting position, velocity, or force in real time. The task-level loop connects task-level reasoning and action-policy generation with the world model and revises plans as the task progresses. The slowest passes through the offline data and simulation infrastructure, where deployment data drive continual learning~\citep{parisi2019continual} and form a data flywheel. In multi-agent and infrastructure-scale systems these loops may be distributed across devices that coordinate toward a common objective, and humans may join them by providing goals, demonstrations, supervision, and intervention.


Physical AI therefore lies not in producing a plausible action or plan, but in sustaining this cycle of observation, prediction, action, verification, and revision under real physical constraints, whether the system is a single robot or distributed infrastructure.

\subsection{The Role of Physical AI in the Evolution of AI}
\label{sec:evolution}

Section~\ref{sec:physical-ai-background} contrasts digital intelligence, whose objects of computation and evaluation are recorded content, with the Physical AI defined in Definition~\ref{def:physical-ai}. This subsection develops three claims about what that contrast means for the trajectory of AI. Physical AI is the stage at which artificial intelligence moves from understanding and generating digital information to perceiving, predicting, and changing physical processes. It builds on digital intelligence rather than replacing it and provides one of the important paths toward artificial general intelligence (AGI). 

\paragraph*{New Capabilities Beyond Digital Intelligence}
The first extension concerns what the system perceives. Digital intelligence works on symbols and pixels that come from prior collection, curation, and encoding. Physical AI deals with physical states directly (Section~\ref{sec:physical-ai-background}), including the pose and velocity of objects, forces and friction at contact surfaces, pressure and temperature in fluids, and the composition and structure of materials. These states are continuous and evolve over time. They can only be partially observed through noisy sensors, and vision, touch, force, and other modalities must be fused within a common spatiotemporal frame. The object of perception thus shifts from previously recorded content to the current state of a world that keeps changing.

The second extension concerns what the system produces. Because physical systems lack an inexpensive reset (Section~\ref{sec:physical-ai-background}), a grasp, an injection, or a turn of the steering wheel changes the subsequent state of the system and often cannot be undone. Physical AI must therefore predict, before acting, how the world will evolve if a given action is taken. This prediction is interventional in nature and depends on causal relations and dynamics, not merely on statistical correlation~\citep{scholkopf2021causal}.

\paragraph*{Building on Digital Intelligence}
These new capabilities do not make Physical AI a replacement for digital intelligence. The common-sense and domain knowledge accumulated in foundation models is the basis on which Physical AI understands and plans tasks~\citep{ichter2023saycan}. In turn, Physical AI grounds that knowledge in measurable changes of state and tests it through real interaction. A large language model can search the literature, propose a hypothesis, and write an experimental protocol, but whether the protocol succeeds depends on accurate pipetting, stable temperature control, and correct sample handling. Only when digital reasoning is closed with physical execution can AI take part in the full process from posing a question and running an experiment to drawing a conclusion.

Section~\ref{sec:physical-ai-background} explains that Physical AI enlarges the unit of analysis from the model to the system, and this enlargement has concrete consequences for how capability is attributed. The same policy may perform very differently on robot arms of different precision or in workshops with different lighting, and task success also depends on engineering conditions such as sensor calibration, fixture design, and safety interlocks. The shift also changes how general capability should be understood and measured, which leads to the third claim.

\paragraph*{A Path Toward AGI}
Benchmarks in digital environments are prone to overfitting, and it is hard to rule out memorization caused by contamination of the training data. The physical world is open, continuous, and impossible to enumerate in advance, and whether a task has been completed is decided directly by physical outcomes. Moravec's paradox points in the same direction, highlighting that perceptual and motor skills that humans find effortless are among the hardest for machines to acquire~\citep{moravec1988mind}. Physical AI therefore provides a demanding test of generality along four axes, including generalization to new objects and scenes as well as new task combinations, transfer across embodiments and tasks as well as from simulation to reality~\citep{tobin2017domain,openai2018dexterous}, robustness to disturbances and sensor degradation as well as environmental drift, and long-term autonomy over days or months with little human intervention. Because mistakes in the physical world are often irreversible, evaluation cannot rest on task success rate alone. Real-time performance, sample efficiency, energy consumption, risk controllability, and recoverability after failure must be considered together, and they frequently trade off against one another. Physical AI is not the only such path, as mathematical reasoning, scientific discovery, and social interaction advance general capability from other directions, but it is the one in which capability is checked against physical reality.

\paragraph*{Locating Physical AI on the Trajectory of AI}
Taking the degree of coupling between AI and the physical world as the axis, the position of Physical AI can be stated more concretely. At the loosely coupled end, AI acts as a \emph{digital tool} that assists humans with design, analysis, and decision-making but leaves execution entirely to them. Next, \emph{simulated agents} perceive and act in simulators or digital twins~\citep{jones2020characterising}, where trial and error is cheap and safe but limited by the gap between simulation and reality. Physical AI proper begins when AI closes the loop in the real world. It starts with \emph{single-agent systems}, such as a robot, vehicle, drone, or autonomous laboratory platform, where most current embodied intelligence research is concentrated. It then extends to \emph{multi-agent collaboration}, in which groups of agents accomplish tasks that no single agent can, and finally to \emph{infrastructure-level Physical AI}, in which distributed sensing, computing, and actuation resources embedded in cities, power grids, factories, and research facilities are scheduled and governed through common protocols. Section~\ref{sec:archetypes} returns to these three stages as the recurring physical compositions of deployed systems. In a network of autonomous laboratories spread across institutions, for example, the question is no longer only whether an experiment can be completed, but also how resources are allocated, how processes are audited, and how data ownership and reproducibility are established. These stages coexist and support one another rather than replacing each other. Digital tools and simulation continue to supply knowledge, data, and safety validation to physical systems. Physical AI thus occupies the tightly coupled end of this axis, where the value of AI is realized as a change in physical processes rather than only as the production of information. The application landscape in Section~\ref{sec:application-landscape} makes this coupling concrete across industries, and Chapter~\ref{ch:general-physical-ai} examines Physical AGI as a long-term research goal.

\subsection{The Application Landscape of Physical AI}
\label{sec:application-landscape}

Physical AI connects the interpretation of physical state with decisions that change it, and its applications span individual robots, autonomous vehicles, instrument networks, factories, cities, and scientific facilities. These settings differ in scale by orders of magnitude, yet the questions that determine whether a system is useful remain largely the same, concerning what the task requires, what the environment allows, what resources are available, and which decisions may be taken without a human in the loop. Whether a particular configuration amounts to a deployable system is ultimately reflected in the cost of the valid work it produces. The subsections that follow first state the closed loop that every application instantiates, together with the four dimensions along which applications are described, namely \emph{task}, \emph{environment}, \emph{resources}, and \emph{autonomy}. They then survey four domains, from industrial operation through everyday services and public infrastructure to scientific inquiry, and finally extract the system archetypes that recur across those domains.

\begin{figure}[t]
  \centering
  \includegraphics[width=\textwidth]{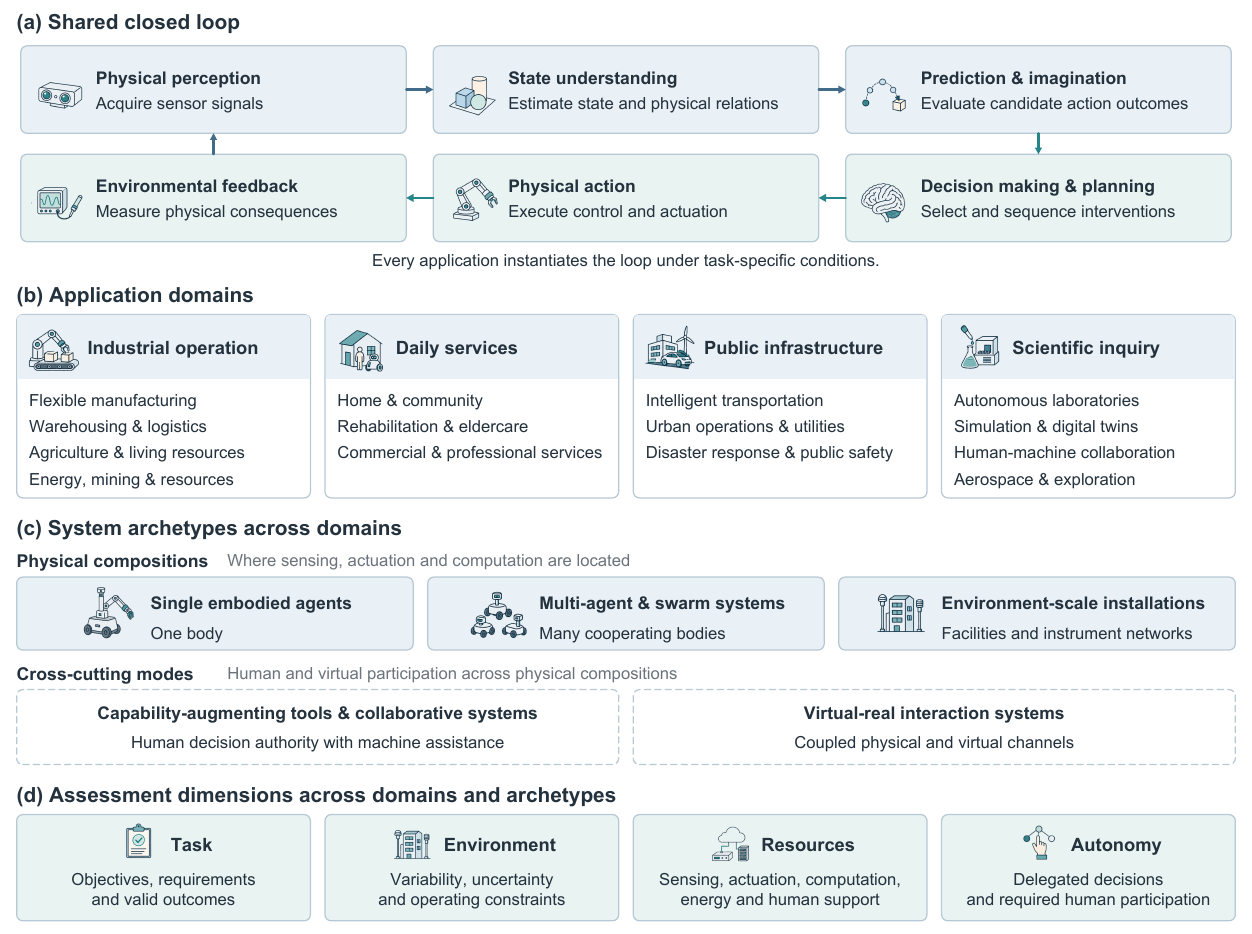}
  \caption[The application landscape of Physical AI]
    {The application landscape of Physical AI. All applications instantiate the same closed loop. The four domains illustrate its operation under different conditions. Five system archetypes (three physical compositions and two cross-cutting modes) and four assessment dimensions apply across domains}
  \label{fig:app-landscape}
\end{figure}

\subsubsection{The Closed Loop and the Distinction from Conventional Automation}
\label{sec:app-loop}

Every application of Physical AI instantiates one closed loop as shown in Figure~\ref{fig:app-landscape}. \emph{Physical perception} acquires signals from the environment through sensors. \emph{State understanding} assembles these signals into estimates of the objects, agents, and processes present, together with the representations and world models of how they evolve. \emph{Prediction and imagination} extrapolate the estimated state forward and evaluate candidate actions, whether through learned dynamics or explicit physical simulation. \emph{Decision making and planning} select and sequence actions under task and process constraints. \emph{Physical action} executes the selected behavior through control and actuation, and the resulting \emph{environmental feedback} becomes the input to the next cycle of perception. These stages give Definition~\ref{def:physical-ai} its operational form. Section~\ref{sec:core-elements} groups them into four elements comprising physical state perception and estimation, world modeling for representation and prediction, decision-making with planning and control, and closed-loop feedback with multiscale adaptation. The loop as a whole, rather than any single stage, defines an application. A system that perceives without acting merely observes its environment, and one that acts without state understanding merely executes a prescribed motion.

Conventional automation already incorporates feedback control, so the loop itself is not new~\citep{AstromMurray2008}. What distinguishes Physical AI is how the loop is sustained when conditions depart from those anticipated at design time, through adapting to changing conditions, interpreting unfamiliar tasks, and transferring knowledge across operations. Classical control holds a process near a setpoint under modeled disturbances. Physical AI re-estimates state, re-plans, and recovers when the task or the environment changes. The distinction is graded rather than binary. Applications occupy a spectrum from fixed procedures with local adaptation, such as an industrial manipulator re-grasping a shifted part, to open-ended task interpretation, in which a system must infer what is wanted from instruction, demonstration, or context~\citep{ichter2023saycan}.

The position of an application on this spectrum is determined by the task rather than by the industry in which it appears. An application can be characterized along four dimensions, namely task, environment, resources, and autonomy. Resources include sensing, actuation, computation, energy, and human support. Autonomy specifies which decisions are delegated and when human participation is required~\citep{Parasuraman2000}. These dimensions distinguish requirements that industry categories tend to merge. Within one factory, moving sealed containers and manipulating deformable materials impose different demands on contact control and recovery, and a laboratory may coordinate fixed instruments without a single mobile body. System composition and decision authority therefore require separate descriptions.

\subsubsection{From Automation to Autonomous Industrial Operation}
\label{sec:industry}

Industrial production, logistics, and agriculture involve transportation, manipulation, inspection, and process regulation. Physical AI offers value when variability makes fixed procedures costly to maintain. Parts arrive in unfamiliar configurations, crops vary in maturity, or equipment behavior changes with wear. The underlying pattern is common to these sectors. Fixed, engineered procedures are replaced by state estimation and adapted planning, while the process constraints within which the adaptation must operate remain fixed.

\paragraph*{Flexible Manufacturing and Lights-Out Production}
In an adaptive production system, visual and tactile observations update estimates of object and process state, while planning and control adjust operations within process constraints~\citep{lee2015,monostori2016}. Variability enters as small batches, frequent product changeovers, and parts presented in unfixtured configurations, and the value is realized when changeover no longer requires re-engineering~\citep{vanbrussel1998}. The benefit is measured by product quality, changeover time, resource consumption, and the effort of setting up a new task, while task variability and the resources required for changeover are the binding constraints. The limit of this development is the \emph{lights-out factory}, in which no standing human presence remains on the floor. Lights-out operation does not eliminate human work. It moves that work to exception handling, supervision, and maintenance, and the cost of these activities becomes a primary economic variable rather than an overhead~\citep{bainbridge1983ironies}.

\paragraph*{Warehousing, Logistics, and Supply-Chain Orchestration}
Warehouses and distribution networks concentrate transportation and manipulation at high repetition rates, including automated storage and retrieval, mobile-robot fleets, sortation, palletizing and depalletizing, and parcel handling~\citep{wurman2008}. The distinguishing problem is not the execution of any single motion but \emph{orchestration}, which involves allocating physical work across many machines as demand fluctuates, replanning after disruption, and keeping physical flows consistent with the inventory records and delivery commitments carried by information systems~\citep{gerkey2004allocation}. Local motion competence and network-level coordination are separate achievements. An operation can automate the former while the latter remains manual, and the gap between the two is closed in practice either by consistent state estimates or by buffer stock.

\paragraph*{Agriculture, Forestry, Fishery, and Animal Husbandry}
Agricultural settings concentrate the constraints of Physical AI, including biological variability, unstructured outdoor terrain, weather-dependent timing, and long deployment cycles. Operations under development span the production sequence, including inspection and patrol, seeding and planting, mechanical weeding, fertilization, plant protection, harvesting, and grading and sorting of produce~\citep{bac2014harvesting}, together with livestock monitoring and aquaculture operations~\citep{berckmans2017precision,fore2018precision}. Value appears in two forms, namely precision, in which seed, chemical, and water inputs are reduced by acting at the resolution of individual plants or animals, and timeliness, in which operations are performed within short agronomic windows. The acceptance criteria are correspondingly biological and seasonal rather than volumetric, such as fruit harvested without bruising at the correct maturity, weeding achieved with a specified reduction in chemical use, or illness detected early enough for intervention.

\paragraph*{Energy, Mining, and Resource Systems}
Energy and mineral systems are high-risk and large-scale, including generation equipment, transmission corridors, pipelines, refineries, and open-pit and underground mines. Applications combine inspection of assets (such as turbine blades, transmission lines, and pipe corridors)~\citep{nguyen2018powerline}, autonomous haulage and drilling~\citep{wang2017,marshall2008}, and process regulation distributed across many devices. Equipment behavior changes with wear, which motivates continuously updated estimates of degradation in place of fixed maintenance schedules~\citep{jardine2006review}. Limited access and the high consequence of failure shift the economics toward predictive operation and remote supervision. In these systems, safety cases rather than throughput determine which functions may be automated, and the allocation of authority between site automation and remote operation centers is itself a design decision.

\subsubsection{Daily Services: From Dedicated Equipment to Embodied Companions}
\label{sec:services}

Service applications place Physical AI in environments that are built for people rather than for machines. At one end are dedicated devices (single-purpose appliances that perform one task under human setup), and at the other are personalized embodied companions intended to share unstructured space with untrained users. Consumer-grade embodied products, from desktop companions to inexpensive legged and humanoid platforms, make this trajectory visible at retail prices, although the distance between demonstrations of interactive behavior and dependable household service remains substantial~\citep{srinivasa2012herb}.

Human participation is an explicit design variable in these applications. Supervision, shared control, and selective delegation assign different responsibilities to users, operators, and machines. In human-facing services, decision authority and timely assistance matter alongside completion rates, so evaluation includes service quality, accessibility, and the burden placed on users and caregivers~\citep{hoffman2019fluency,nikolaidis2016mutual}.

\paragraph*{Home and Community Services}
A household assistant must accommodate changing layouts and object locations, because neither residents nor their belongings remain arranged to suit a stored map~\citep{wisspeintner2009robocup}. Cleaning, tidying, laundry, and meal preparation combine manipulation with continual re-estimation of where things are and what state they are in. Community services extend the same capabilities into semi-public space, including delivery to doors and lockers, patrol and monitoring, and check-ins on residents who live alone. In these settings, acceptance is determined not by peak capability but by the frequency and manner in which the system requests help. If a robot completes 90\% of its tasks unaided but leaves the remaining 10\% unfinished in mid-task, it shifts the burden to its users rather than removing it~\citep{forlizzi2006service}.

\paragraph*{Medical Rehabilitation and Eldercare}
A rehabilitation system must respect clinically established limits on force, range, and intensity while still challenging the patient enough for therapeutic effect. Applications include shared-control therapy devices that adjust assistance to measured performance, hospital logistics and delivery, continuous monitoring of activity, and assistance with activities of daily living~\citep{kyrarini2021survey}. Clinical deployment adds requirements that are foreign to industrial settings. Interventions must demonstrate safety and efficacy under the conventions of clinical study~\citep{lo2010robot,esquenazi2012rewalk}, responsibility for decisions must be traceable to accountable parties, and eldercare in particular must weigh privacy and dignity alongside capability~\citep{broadbent2009acceptance}. In this domain, Physical AI cannot be separated from its regulatory and evidentiary context.

\paragraph*{Commercial and Professional Services}
Retail, hospitality, professional cleaning, and food preparation occupy a middle ground, with spaces more structured than homes, because they are designed around service processes, yet occupied by the public and by staff whose cooperation cannot be assumed. Tasks include shelf monitoring and restocking, floor care, reception and guidance, and food handling to hygienic standards. Interaction with customers and employees is part of the task itself rather than a distraction from it, which places interface design and predictable, legible behavior on the critical path alongside manipulation competence. The economics are those of semi-structured spaces. Repetition is higher than in domestic service, while the tolerance for bespoke installation is lower than in industry.

\subsubsection{Public Infrastructure: From Single-Vehicle Autonomy to City-Scale Coordination}
\label{sec:infrastructure}

Transportation, energy networks, and public infrastructure extend the concerns of the preceding sections across larger spatial and temporal scales. The unit of analysis shifts from one machine to populations of devices whose local actions must combine into system-level outcomes.

\paragraph*{Intelligent Transportation}
Vehicles and drones combine local perception and control with routing and coordination~\citep{yurtsever2020survey,murray2015}. For road vehicles, the difficult problems are increasingly systemic rather than vehicular, including mixed traffic in which autonomous and human drivers share conventions, fleet scheduling together with charging and cleaning for shared robotaxi operation, and the interfaces between vehicle autonomy and traffic management~\citep{stern2018,varaiya2013}. For low-altitude aircraft, coordination is the founding problem. Airspace organization, geofencing, and conflict resolution for large numbers of small vehicles must exist before individual autonomy is useful at scale~\citep{prevot2016utm}. In both cases, the binding constraint is the coordination of many local planners toward network-level service quality.

\paragraph*{Urban Operations and Public Facilities}
Buildings, power systems, and water networks distribute sensing and actuation across many devices with no mobile body at all, an arrangement treated as a distinct system archetype in Section~\ref{sec:archetypes}. Model predictive control of building climate, demand response across a power network, and leakage detection in a water distribution network illustrate the range~\citep{afram2014theory,siano2014demand,chan2018review}. Communication capacity and consistent state estimates constrain coordination between local actions and system-wide outcomes. Value is measured by the reliability and continuity of service, energy and resource efficiency, and responsiveness to faults, observed over the scales of cities and seasons, which places a premium on sustained and validated operation rather than on peak performance.

\paragraph*{Disaster Response and Public Safety}
Extreme environments impose additional restrictions. Disaster response combines difficult access, limited energy, and delayed or unavailable human intervention, making recovery mechanisms and bounded local autonomy particularly important. Search and rescue in collapsed structures, firefighting and hazardous-material intervention, and post-disaster inspection all require systems that decide and recover locally within a pre-authorized envelope and degrade safely when communication fails~\citep{murphy2014disaster,delmerico2019current,tranzatto2022cerberus}. Public-safety applications sharpen the question of authority more than any other setting in this chapter. The conditions under which a machine may act, and the parties who are answerable for its actions, must be specified before capability is deployed rather than after.

\subsubsection{Scientific Inquiry: From Automated Instruments to Autonomous Laboratories}
\label{sec:science}

Scientific applications use physical action to acquire evidence and guide inquiry. An autonomous laboratory selects experiments, operates instruments, measures outcomes, and revises subsequent choices~\citep{king2009,burger2020,macleod2020,dai2024}. Scientific value depends on measurement validity, sample traceability, and reproducible interpretation, alongside experimental throughput.

\paragraph*{Automated and Self-Driven Laboratories}
The A-Lab illustrates this organization within a defined inorganic-materials synthesis workflow, integrating robotic operations, computational information, automated characterization, and active learning~\citep{szymanski2023alab}. Its subsequent correction underscores the need to validate material identification and novelty claims separately from successful workflow execution~\citep{szymanski2026correction}. Workflow autonomy (the reliable execution of planned operations) and epistemic autonomy (the production of claims that survive scrutiny) are different achievements, and evaluations in this field tend to measure the former while the value lies chiefly in the latter. Self-driven laboratory programs accordingly treat the validity chain from sample to published claim as part of the system to be engineered.

\paragraph*{Physical Simulation and Digital Twins}
In scientific inquiry, simulation and digital twins, which Section~\ref{sec:physical-ai-cps-twins} distinguishes from generic world models, serve as instruments for experiment design, parameter estimation, and controlled testing. Validation takes two forms. For policy training, it concerns whether simulated variation supports performance in reality. For an operational digital twin, it concerns the accuracy, timeliness, and uncertainty of the estimates used in decisions. The two uses impose different obligations. A training environment must cover the variations a policy will meet, while an operational twin must declare how far ahead, and with what confidence, its estimates may be trusted.

\paragraph*{Human-Machine Collaborative Scientific Cognition}
Autonomous laboratories change the division of cognitive labor in science rather than removing the scientist. In deployed systems, objective setting, prioritization among hypotheses, and the acceptance of evidence remain human responsibilities, while machines contribute breadth of search, uniformity of execution, and complete procedural memory. The interpretability of proposed experiments conditions trust. A suggestion whose rationale can be examined can also be disputed, and such dispute is one of the mechanisms by which science corrects itself. The productive design question, familiar from Section~\ref{sec:app-loop}, is which parts of inquiry to delegate. Within a research program, the answer to this question is itself something to be established experimentally.

\paragraph*{Aerospace and Space Exploration}
Space exploration combines scientific objectives with the most constrained communication, energy, and mass budgets in this chapter. Planetary rovers schedule a sol of activity within command latency~\citep{AutoNav}, on-orbit systems service and assemble spacecraft, and in-situ instruments analyze samples that cannot be returned. Underwater operations share the pattern of difficult access, limited energy, and delayed or absent intervention~\citep{wynn2014autonomous}. Both settings inherit bounded local autonomy as their organizing principle of deciding locally within a pre-authorized envelope, recording faithfully, and deferring the remainder to later analysis. This is the same principle identified for disaster response in Section~\ref{sec:infrastructure}, applied where the return of data is slower still.

\subsubsection{System Archetypes across Industries}
\label{sec:archetypes}

Beneath the domain labels above, deployed systems assume a small number of recurring physical compositions, together with two modes that cut across them. Three compositions locate sensing and actuation together with computation in one body, in many bodies, or in the facility itself, while the two modes place either a human operator or a simulated world inside the loop. These compositions retrace the progression from single-agent systems through multi-agent collaboration to infrastructure-level Physical AI described in Section~\ref{sec:evolution}. Taken together, the five archetypes span the domains surveyed above, each of which can be read through the four dimensions of Section~\ref{sec:app-loop}. Figure~\ref{fig:app-landscape} shows both views together.

\paragraph*{Single Embodied Agents}
The first composition concentrates sensing, actuation, computation, and energy in a single mobile body, as in the household assistant~\citep{fu2025}, the inspection quadruped~\citep{miki2022}, the planetary rover, the delivery drone~\citep{loquercio2021}. Its defining property is that resources and autonomy are co-located and bounded by one morphology, which in turn fixes the reachable task set. Most public attention attaches to this archetype, although the surveys above show it to be one composition among three.

\paragraph*{Multi-Agent and Swarm Systems}
The second composition distributes work across many cooperating bodies, as in warehouse fleets, agricultural machine teams, drone formations, cooperative transport~\citep{dorigo2021swarm}. Capability emerges from task allocation, communication, and consensus rather than from any single member, and the fleet thereby acquires properties that its individual members lack, such as tolerance of the failure of one member~\citep{werfel2014collective,tolstaya2020decentralized,zhu2024self}. Assessment correspondingly shifts to fleet-level quantities such as throughput, availability, and the reliability of the coordination mechanism itself.

\paragraph*{Environment-Scale Installations}
The third composition embeds sensing and actuation in the facility itself, as in automated warehouses, lights-out factories, autonomous laboratories~\citep{gao2025unilabos}, intelligent buildings, instrumented grids. No mobile body need exist. The laboratory described in Section~\ref{sec:app-loop}, which coordinates fixed instruments without any mobile body, belongs to this composition. Engineering attention shifts from the agility of a body to the interfaces of a facility, including how materials, information, and people enter and leave, and the system boundary becomes an explicit design decision rather than the surface of a robot.

\paragraph*{Capability-Augmenting Tools and Collaborative Systems}
The first of the two cross-cutting modes extends a single person's capabilities through intelligent hand tools, exoskeletons, augmented-reality task guidance, and collaborative robot cells~\citep{delooze2016exoskeletons,matheson2019human}. The human retains decision making, while Physical AI supplies perception, guidance, and force at the point of work. Autonomy is deliberately low, and integration with the human is the design center rather than a constraint. Because responsibility remains with the operator, this archetype provides the economic baseline against which more autonomous alternatives are judged~\citep{iso10218_2025}.

\paragraph*{Virtual-Real Interaction Systems}
The other cross-cutting mode spans physical and simulated environments through sim-to-real policy training~\citep{tobin2017domain}, operational digital twins, teleoperation with haptic and mixed-reality interfaces~\citep{hokayem2006bilateral}, and hardware-in-the-loop test benches. The virtual channel may support development, remain part of operation, or serve both roles. Action may originate from a human operator, a learned policy, or a simulator, and its physical effect depends on how the resulting command is executed. The validation questions of Section~\ref{sec:science}, concerning whether simulated variation covers reality and how far estimates may be trusted, thereby become structural properties of the system rather than external checks.

The archetypes compose rather than compete with one another. An autonomous laboratory, for example, is an environment-scale installation that contains single agents, is trained through virtual-real interaction, is maintained by humans with augmenting tools, and is coordinated as a multi-agent system when replicated across sites. The physical composition fixes much of the resource profile of an application before any algorithm is selected, and the two modes then determine where authority resides and how much of the loop runs in simulation. For this reason, the four dimensions of Section~\ref{sec:app-loop} are best assessed per archetype rather than per industry.

\subsection{Summary and Organization of the Monograph}

This chapter introduces Physical AI as a system-level capacity to perceive physical states, maintain task-relevant representations of physical relationships and action consequences, and select and execute interventions through feedback under physical, operational, and computational constraints. It contrasts this capacity with digital intelligence, makes the contrast precise in Definition~\ref{def:physical-ai}, and marks its boundaries against embodied AI, cyber-physical systems, and digital twins. It then decomposes the capacity into four interdependent elements (Section~\ref{sec:core-elements}), locates it on the axis of coupling between AI and the physical world (Section~\ref{sec:evolution}), and shows how deployed applications instantiate one closed loop under four assessment dimensions, across three physical compositions and two cross-cutting modes (Section~\ref{sec:application-landscape}). A single criterion runs through these perspectives. What a system can do is determined less by any individual model than by the integrated loop of perception, prediction, decision, action, and feedback, and by the physical and operational conditions under which the loop is sustained.

The remainder of this monograph develops that loop. Chapter~\ref{sec:technical-foundations} assembles the technical foundations shared by all four elements, including mathematical descriptions of the physical world, control theory and autonomous systems, embodied cognition, and learning and reasoning. The chapters that follow then take up the elements in order. Chapter~\ref{ch:perception} covers perception and state estimation. Chapters~\ref{ch:physical-data} and~\ref{ch:wm} cover physical data, simulation environments, and world models as resources that support prediction. Chapters~\ref{ch:manipulation-models} and~\ref{ch:whole-body-motor-control} cover the foundation models and motor-control systems that realize decision-making and execution. Chapter~\ref{ch:physical-agent} develops the definition, organization, task execution, and coordination of physical agents. Chapter~\ref{ch:hardware-systems} covers the hardware and systems that must sustain the loop reliably over long-term operation. The final chapters return to the questions left open here concerning how far the path toward general Physical AI extends (Chapter~\ref{ch:general-physical-ai}), how capabilities are evaluated and governed (Chapter~\ref{ch:evaluation}), and what development strategies and longer-term outlook follow from them (Chapter~\ref{ch:strategy-outlook}).

The definitions and distinctions established in this chapter provide the vocabulary for all that follows. Whether a system amounts to Physical AI, where it sits between fixed automation and open-ended autonomy, and how much of the loop runs in simulation or under human authority are judged not by stated architecture but by demonstrated capability under specified conditions. 

\clearpage

\section{Technical Foundations}
\label{sec:technical-foundations}

Building on the framework introduced in Chapter~\ref{ch:introduction}, this chapter first establishes foundations of Physical AI, including the mathematical modeling, control theory, cognitive AI, and usage of machine learning. These four points are organized around the closed physical interaction loop: observations inform state estimates, models and goals guide action, physical contact generates feedback, and learning updates behavior. Particularly, Section~\ref{sec:math-physical-world} introduces mathematical descriptions of geometry, dynamics, state, and uncertainty, which provide a shared language for later discussion. Section~\ref{sec:control-autonomy} explains how the core feedback control is used to build a robust and safe embodied system. On top of the control theory, Section~\ref{sec:embodied-cognition-development} examines how to enable embodiment with perception, skill, and physical grounding of language and multimodal representations. Finally, Section~\ref{sec:learning-physical-world} moves towards the modern machine learning-based methods. This closed-loop formulation provides the shared theoretical backbone for the perception, modeling, control, hardware, and safety chapters that follow.

\subsection{Mathematical Descriptions of the Physical World}
\label{sec:math-physical-world}

Physical AI requires an internal representation of the physical world that enables an agent to perceive, predict, and interact with its environment. Unlike purely symbolic systems, embodied agents must reason about continuous geometry, physical dynamics, uncertainty, and interaction constraints. Mathematical descriptions provide the foundation for this process by defining how physical entities are represented, how their states evolve over time, and how observations relate to underlying physical reality.

Classical robotics has developed rigorous mathematical formulations for geometry, motion, and dynamics, while modern embodied intelligence extends these concepts toward learned representations of complex environments. Together, geometric representations, physical dynamics, and predictive state descriptions form the foundation for physical world modeling.

\subsubsection{Geometry and Spatial Representation}

\paragraph*{Euclidean Space and Rigid-Body Transformations}
The physical world is naturally described in three-dimensional (3D) Euclidean space. The position of an object can be represented as a point $\mathbf{x}_{\mathrm{pos}}\in\mathbb{R}^{3}$, while the complete pose of an object requires both translation and rotation.

A rigid transformation is represented as
\begin{equation}
    \mathbf{x}_{\mathrm{pos}}^{\prime}
    =
    \mathbf{R}\mathbf{x}_{\mathrm{pos}}
    +
    \mathbf{t},
\end{equation}
where $\mathbf{R}\in \mathrm{SO}(3)$ is a rotation matrix in the special orthogonal group in three dimensions and $\mathbf{t}\in\mathbb{R}^{3}$ represents translation. The rotation matrix satisfies
\begin{equation}
    \mathbf{R}^\top\mathbf{R}
    =
    \mathbf{I},
    \qquad
    \det(\mathbf{R})=1 .
\end{equation}

Although rotation matrices provide an intuitive geometric interpretation, they contain redundant parameters. Unit quaternions provide a compact representation of orientation and avoid singularities associated with Euler angles. 

For robotics and embodied systems, rigid transformations in the special Euclidean group $\mathrm{SE}(3)$ are commonly represented as
\begin{equation}
\mathbf{T}
=
\begin{bmatrix}
\mathbf{R}&\mathbf{t}\\
\mathbf{0}^\top&1
\end{bmatrix}.
\end{equation}

The corresponding Lie algebra $se(3)$ provides a local representation of transformations and is widely used in state estimation, simultaneous localization and mapping (SLAM), and motion optimization~\citep{sola2017,forster2017}.

\paragraph*{Explicit Geometric Representations}
A physical environment can be represented using explicit geometric structures that directly describe spatial entities. Point clouds represent surfaces through discrete sample sets of the form
\begin{equation}
    \mathcal{P}_{\mathrm{cloud}}
    =
    \{
    \mathbf{p}_1,\mathbf{p}_2,...,\mathbf{p}_{N_{\mathrm{point}}}
    \},
    \qquad
    \mathbf{p}_i\in\mathbb{R}^{3}.
\end{equation}

Point clouds preserve metric geometry and naturally arise from sensors such as light detection and ranging (LiDAR) devices and depth cameras. They are widely used for localization, mapping, manipulation, and 3D scene understanding.

Mesh representations describe surfaces using vertices, edges, and faces. Compared with point clouds, meshes explicitly encode surface connectivity and are therefore important for simulation, collision detection, and physical interaction.

\paragraph*{Implicit Representations of Physical Environments}
Instead of explicitly storing geometric elements, implicit representations describe the environment through continuous functions.

A signed distance function (SDF) represents geometry as the zero-level set of a continuous function:
\begin{equation}
    \mathcal{S}_{\mathrm{surf}}
    =
    \{
    \mathbf{x}_{\mathrm{pos}}
    |
    f(\mathbf{x}_{\mathrm{pos}})=0
    \}.
\end{equation}

SDFs provide useful geometric properties such as surface distance and occupancy, making them suitable for collision checking, motion planning, and reconstruction.

Neural implicit representations extend this idea by parameterizing continuous scene functions with neural networks. Neural radiance fields (NeRFs) represent a scene through a continuous mapping from spatial coordinates and viewing directions to density and appearance as
\begin{equation}
    (\sigma_{\mathrm{rad}},\mathbf{c}_{\mathrm{rgb}})
    =
    F_{\theta}^{\mathrm{NeRF}}(\mathbf{x}_{\mathrm{pos}},\mathbf{d}),
\end{equation}
where $\sigma_{\mathrm{rad}}$ represents density and $\mathbf{c}_{\mathrm{rgb}}$ represents color. Such representations enable high-quality reconstruction and novel-view synthesis, providing a foundation for simulation-ready environments~\citep{mildenhall2020nerf}.

\paragraph*{Topology and Connectivity}
Geometric representations describe where objects exist, but embodied agents must additionally understand how spaces and objects are connected. Topological representations abstract environments into nodes and edges, where nodes represent meaningful states or locations and edges represent feasible transitions.

For navigation, topology captures reachability and connectivity of environments. For manipulation, object topology describes structural relationships among parts and possible interaction paths. Combining geometry and topology allows agents to reason not only about spatial configuration but also about possible actions.

\subsubsection{Physical Dynamics and Interaction Modeling}
\label{subsec:physical-dynamics-interaction}

\paragraph*{Kinematics}
Kinematics describes how the configuration of a physical system changes without considering the forces causing the motion. For articulated robots, forward kinematics describes the relationship between joint space and task space as
\begin{equation}
    \mathbf{T}
    =
    f(\mathbf{q}),
\end{equation}
where $\mathbf{q}$ represents joint variables and $\mathbf{T}\in \mathrm{SE}(3)$ represents the end-effector pose.

The inverse kinematics problem estimates joint configurations required to achieve a desired pose:
\begin{equation}
    \mathbf{q}\in f^{-1}(\{\mathbf{T}\}),
\end{equation}
where $f^{-1}(\{\mathbf T\})$ denotes the set of joint configurations that reach the desired pose.

Inverse kinematics is fundamental for manipulation, locomotion, and motion planning, but can be challenging due to redundancy, constraints, and multiple feasible solutions~\citep{khalil2002}.

\paragraph*{Dynamics and Equations of Motion}
Dynamics describes how physical states evolve under forces and control inputs. For articulated systems, robot dynamics can be expressed as
\begin{equation}
\mathbf{M}(\mathbf{q})\ddot{\mathbf{q}}
+
\mathbf{C}(\mathbf{q},\dot{\mathbf{q}})
\dot{\mathbf{q}}
+
\mathbf{g}(\mathbf{q})
=
\boldsymbol{\tau}
+
\mathbf{J}(\mathbf q)^\top\mathbf w_{\mathrm{ext}},
\end{equation}
where $\mathbf{M}$ represents inertia, $\mathbf{C}$ represents velocity-dependent effects, $\mathbf{g}$ represents gravity, and $\boldsymbol{\tau}$ represents actuator torque. The external wrench $\mathbf w_{\mathrm{ext}}\in\mathbb R^6$ stacks force and moment. The geometric Jacobian $\mathbf J(\mathbf q)\in\mathbb R^{6\times n}$ maps the $n$ joint rates to linear and angular velocity at the wrench reference point, expressed in the same coordinate frame.

The Newton-Euler formulation provides an efficient recursive computation of rigid-body dynamics, while the Lagrangian formulation derives motion equations from energy functions:
\begin{equation}
L_{\mathrm{mech}}(\mathbf{q},\dot{\mathbf{q}})
=
E_{\mathrm{kin}}(\mathbf{q},\dot{\mathbf{q}})-E_{\mathrm{pot}}(\mathbf{q}).
\end{equation}

These formulations provide the mathematical basis for simulation, control, and physical reasoning~\citep{featherstone2008}.

\paragraph*{Contact Mechanics and Interaction}
Physical AI requires reasoning about interaction between objects. Unlike free-space motion, contact introduces constraints involving collision, friction, and force transmission.

The Coulomb friction model relates tangential and normal forces through
\begin{equation}
    \|\mathbf{f}_{\mathrm{t}}\|
    \leq
    \mu \|\mathbf{f}_{\mathrm{n}}\| ,
\end{equation}
where $\mu$ denotes the friction coefficient.

Contact dynamics are often formulated as constrained optimization problems involving non-penetration constraints and friction cones. Accurate contact modeling is essential for grasping, manipulation, locomotion, and tool use~\citep{posa2014,stewart1996}.

\subsubsection{State Representation and Predictive Modeling}
\label{sec:state-predictive-modeling}

\paragraph*{State-Space Representation}
A physical system is commonly described through a state variable $\mathbf{x}_t$ that summarizes the information necessary to predict future evolution. The state transition process can be represented as
\begin{equation}
    \mathbf{x}_{t+1}
    =
    F(\mathbf{x}_t,\mathbf{u}_t)
    +
    \mathbf{w}_t,
\end{equation}
where $\mathbf{u}_t$ represents actions and $\mathbf{w}_t$ represents disturbances.

The observation process connects physical states with sensory measurements through
\begin{equation}
    \mathbf{y}_t
    =
    h_{\mathrm{obs}}(\mathbf{x}_t)
    +
    \boldsymbol{\epsilon}_t.
\end{equation}

This formulation provides the mathematical foundation for perception, estimation, planning, and control.

\paragraph*{Latent States and Compact Physical Representations}
In realistic environments, the complete physical state is rarely directly observable. Sensors provide incomplete and noisy measurements, while variables such as object properties, hidden geometry, and contact states may remain unknown.

Therefore, modern Physical AI systems often learn latent state representations of the form
\begin{equation}
    \mathbf{z}_t
    =
    f_\theta(o_t),
\end{equation}
where $o_t$ represents observations and $\mathbf{z}_t$ represents a compact latent description of the environment.

A useful latent representation should preserve task-relevant physical information while discarding irrelevant variations. Such representations provide the foundation for predictive learning and world models discussed in Section~\ref{sec:learning-physical-world} and Chapter~\ref{ch:wm}.

\paragraph*{Uncertainty and Predictive Representations}
Physical environments are inherently uncertain due to incomplete observations, stochastic dynamics, and unknown parameters. Therefore, physical state representations should incorporate uncertainty:
\begin{equation}
    p(\mathbf{x}_t\mid\mathbf{y}_{1:t})
\end{equation}
rather than assuming exact knowledge of the state.

Probabilistic state estimation methods, including Kalman filtering and Bayesian inference, provide classical solutions for uncertainty estimation. Modern learning-based approaches extend these ideas by learning predictive representations directly from large-scale sensory experience.

These mathematical descriptions provide the bridge between explicit physical models and learned world models. Classical representations define the structure of the physical world, while learned representations enable scalable prediction and reasoning in complex environments.

\subsection{Control Theory and Autonomous Systems}
\label{sec:control-autonomy}

The mathematical descriptions in the previous section provide models of the physical world, and control theory in this section explains how an embodied agent can use measurements and actions to regulate its interaction with that world. Embodied intelligence requires continuous exchange of energy and information with the physical world. A robot must therefore not only infer an appropriate action but also realize it through actuators subject to finite torque, uncertain contact, delayed sensing, and disturbances that cannot be enumerated in advance. Control theory provides the formal language for this interaction. It asks how measurements should modify action, how models may be used without being trusted unconditionally, and under what conditions the resulting closed loop remains stable, accurate, and safe.

This section examines closed-loop autonomous operation, in which sensing, planning, and control repeatedly respond to the consequences of action. Motor-current and joint controllers react within microseconds to milliseconds. Whole-body controllers coordinate forces over tens of milliseconds, and task planners and behavior executives operate over seconds or minutes. Learning systems may update models over an entire deployment. Although these layers differ in representation and method, each compares an intended condition with an observed one and acts on the discrepancy. Feedback therefore connects classical control, optimization, planning, and learning-based autonomy rather than belonging to any one of them. The following subsections introduce feedback, control methods, and hierarchical autonomy. They also analyze stability, robustness, adaptation, and safety.

\subsubsection{Foundations of Cybernetics: Feedback and Steady State}

A feedback control system uses measurements of a physical system’s behavior to adjust its control inputs toward a desired target, which forms a feedback loop of measurement and correction. This section first introduces this feedback loop, and then discusses its transient and steady-state behavior, to assess how effectively a system approaches and maintains that target.

\paragraph*{Negative Feedback Principle}
\citet{Wiener1948} introduces cybernetics as a framework for studying control and communication in systems whose actions affect the conditions they subsequently observe. In this framework, a plant receives a control input $u(t)$, evolves under its own dynamics and external disturbances, and produces an output $y(t)$. For output tracking, a controller compares the measured output with a reference $r(t)$ and uses the tracking error to determine the next control command. The error, and the control input in the simplest case of proportional control with gain $K_{\mathrm{P}}$, are defined as
\begin{equation}
 e_{\mathrm{track}}(t) = r(t) - y(t),
 \qquad
 u(t) = K_{\mathrm{P}} e_{\mathrm{track}}(t),
\end{equation}
where the proportional law is the simplest instance of feedback. Integral and derivative action extend this relation in the proportional-integral-derivative controller introduced below.

Controllers may also use an estimate of the underlying state when the relevant variables are not directly measured. The loop is called negative feedback when the controller acts so as to reduce the measured deviation, as when a thermostat increases heating below its set point or a balancing vehicle generates torque opposing body tilt. Feedback is most valuable when the model is incomplete. If an actuator delivers less torque than commanded, or a manipulator grasps an object of unknown mass, a precomputed input cannot detect the resulting error. Feedback uses observations of the changed motion to compensate within the limits of sensing, actuation, and closed-loop bandwidth. Feedback thereby converts certain forms of uncertainty into observable error and permits repeatable behavior from imperfect components.

Negative feedback is not synonymous with stability. A correction whose sign is appropriate at low frequency may destabilize the loop once sensing, computation, or actuation introduces sufficient delay. Excessive gain has the same effect, and strong reaction to high-frequency measurement noise may excite unmodeled flexible modes. Feedback design must therefore shape the magnitude and phase of the loop response, not merely reverse the sign of the error~\citep{AstromMurray2008}. The signals entering the loop must also be distinguished. A reference describes intended behavior, a disturbance acts on the plant from outside, measurement noise corrupts the information available to the controller, and model uncertainty denotes the discrepancy between representation and mechanism. These signals can produce similar visible effects yet require different remedies, so an embodied controller is a negotiated balance among tracking, disturbance rejection, noise attenuation, actuator effort, and robustness. Feedback further depends on what is measured. Because the full state is rarely available, practical loops incorporate an estimator that fuses a dynamic model with sensor streams. Estimation and control become coupled processes in which actions determine which future observations become available.

\begin{figure}[t]
    \centering
    \includegraphics[width=0.95\linewidth]{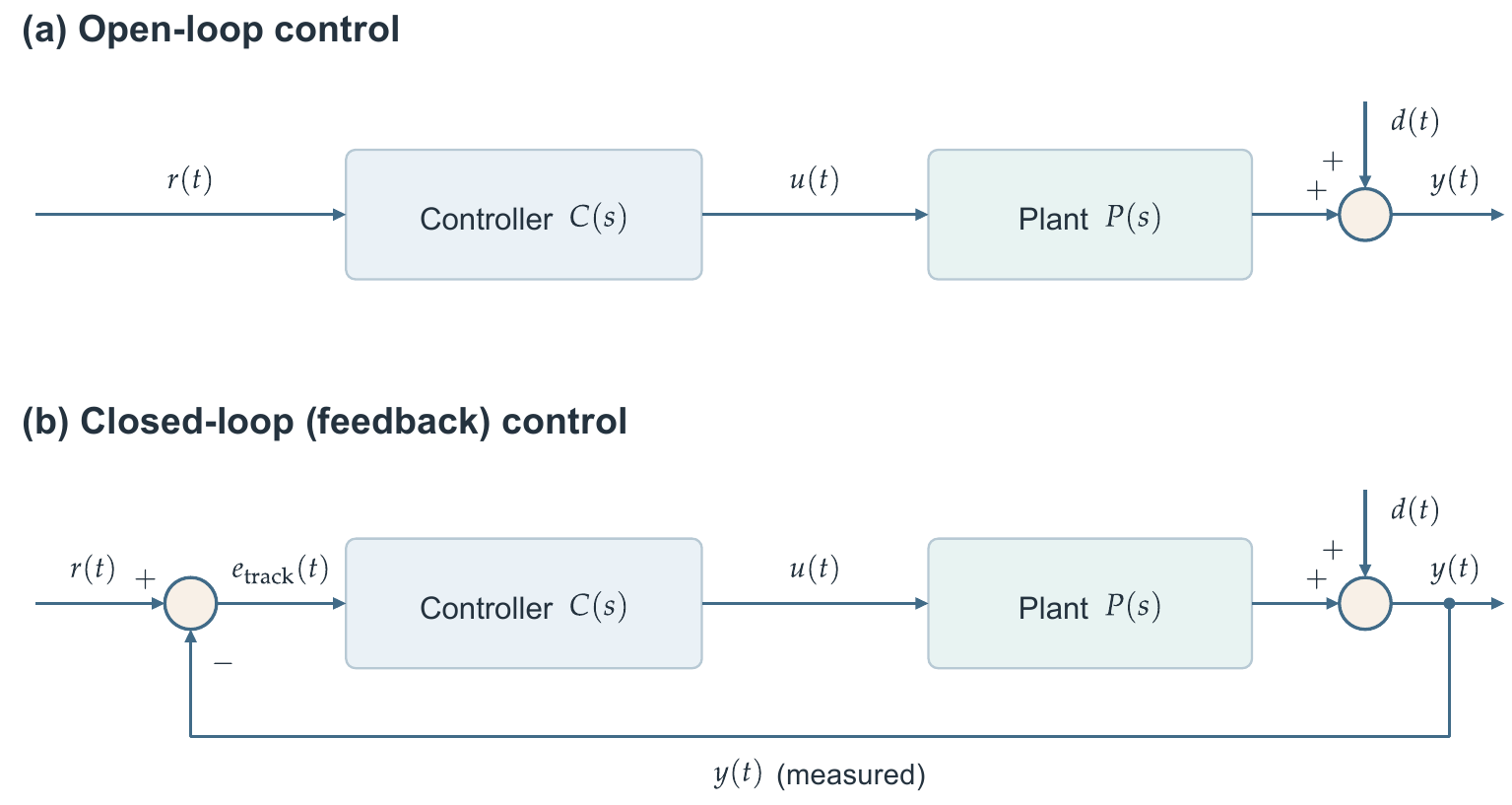}
    \caption{Open-loop and closed-loop control. (a)~In open-loop control, the controller maps the reference $r(t)$ directly to the plant input $u(t)$, so an output disturbance $d(t)$ passes to the output $y(t)$ uncorrected. (b)~In closed-loop control, the measured output is fed back and compared with the reference, and the controller acts on the error $e_{\mathrm{track}}(t)=r(t)-y(t)$. For a stable closed loop, the feedback attenuates the output-disturbance response over frequencies at which the loop-gain magnitude $|L(\mathrm j\omega)|$ is sufficiently large. $C(s)$~and $P(s)$ denote the controller and plant transfer functions, respectively}
    \label{fig:openloop-closedloop}
\end{figure}

\paragraph*{Open-Loop and Closed-Loop Control}
Two basic control arrangements are open-loop and closed-loop control, illustrated in Figure~\ref{fig:openloop-closedloop}~\citep{FranklinPowellEmamiNaeini2019,DorfBishop2017,AstromMurray2008}. An open-loop controller selects its input without measurements of the controlled outcome during execution. Such control is appropriate when the plant and environment are sufficiently repeatable, when feedback is unavailable, or when the maneuver is too fast for a sensor-to-actuator loop, but it cannot correct errors that become observable only after execution begins. Closed-loop control uses observations to revise action, and its advantage is made precise by the sensitivity functions. For a single-input, single-output linear system in the standard negative unity-feedback configuration with plant $P(s)$ and controller $C(s)$, the loop transfer function $L(s)=P(s)C(s)$ yields
\begin{equation}
  S(s)=\frac{1}{1+L(s)},
  \qquad T(s)=\frac{L(s)}{1+L(s)},
  \qquad S(s)+T(s)=1,
\end{equation}
where $T$ maps the reference to the output, $S$ maps an additive output disturbance to the output, and measurement noise reaches the output through $-T$. Where $|L|$ is large and the loop is stable, $|S|$ is generally small, whereas noise is transmitted through $T$. The multivariable case is analogous, with multiplication order significant. Feedback thus does not erase uncertainty but redistributes sensitivity across frequencies, subject to stability margins, delay, and unmodeled dynamics~\citep{SkogestadPostlethwaite2005}.

Capable robotic systems combine feedback with feedforward. Inverse dynamics, for example, predicts the torques required for a desired motion, while feedback corrects residual errors caused by inaccurate parameters, friction, contact, or estimation error. This two-degree-of-freedom structure also clarifies the role of learned models, which improve the nominal action while the conventional loop rejects deviations within a known operating envelope. The open-loop/closed-loop distinction is therefore a design spectrum rather than a dichotomy. Feedforward motion may be embedded within a slower feedback loop, and symbolic goals may be issued to feedback controllers with local termination conditions. Physical feedback also carries costs. Sensors consume power and bandwidth, estimation and communication add latency, and a loop may regulate the wrong variable with high precision, as when position control damages a fragile object because contact force is unmonitored~\citep{RaibertCraig1981,Hogan1985}.

\paragraph*{Steady-State Error and Integral Action}
Transient performance describes behavior while a system approaches its target. For a system that converges to a steady state, steady-state performance characterizes the residual behavior after transients have decayed~\citep{Ogata2010,FranklinPowellEmamiNaeini2019}. For a constant reference $r(t)$, with the tracking error $e_{\mathrm{track}}(t)=r(t)-y(t)$ as defined above, the steady-state error is
\begin{equation}
    e_{\mathrm{ss}}=\lim_{t\rightarrow\infty} e_{\mathrm{track}}(t),
\end{equation}
provided that the limit exists and the closed loop is stable. A proportional controller may fail to eliminate this error when a constant input is needed to balance a persistent load. For instance, a mobile robot ascending a slope requires a constant drive force to balance gravity, and because proportional action generates its command solely in proportion to the present error, such a nonzero force can be produced only while a nonzero error persists, so the vehicle settles into a steady climb at a speed below the reference. Integral action instead accumulates the error in an integrator state $\eta_{\mathrm{I}}(t)$ and adds an integral term $u_{\mathrm{I}}(t)$ to the control input,
\begin{equation}
    \eta_{\mathrm{I}}(t)=\int_{0}^{t} e_{\mathrm{track}}(\tau)\,\mathrm{d}\tau,
    \qquad u_{\mathrm{I}}(t)=K_{\mathrm{I}} \eta_{\mathrm{I}}(t),
\end{equation}
where $K_{\mathrm{I}}$ is the integral gain. A persistent deviation therefore continues to modify the command and, subject to closed-loop stability and sufficient control authority, is driven to zero. The internal model principle formalizes this mechanism. Under closed-loop stability and the solvability conditions of output regulation, exact asymptotic tracking or rejection that is robust to small plant-parameter variations requires a model capable of generating the reference or disturbance signal within the feedback path~\citep{FrancisWonham1976}. The integral action discussed above is the simplest instance of this principle. An integrator generates constant signals, and a loop containing an integrator can therefore track a constant reference or reject a constant disturbance exactly. Robust tracking of sinusoidal references likewise requires the controller to contain a corresponding oscillatory model at the reference frequency.

Integral action has its own failure mode under actuator saturation. An actuator, such as an electric motor or a hydraulic drive, delivers only bounded torque or force, so the applied control input $u(t)$ stops following the command whenever the requested value exceeds this physical limit. During saturation, the integrator may continue accumulating error although additional command cannot affect the plant, and the stored term produces a large overshoot when the system re-enters the controllable range. Anti-windup schemes limit or discharge this accumulation by clamping the integrator, feeding back the difference between requested and applied input, or conditioning integration on the actuator state \citep{KothareEtAl1994,TarbouriechTurner2009}. For embodied systems, saturation must be treated as part of the model rather than as a rare implementation detail. Note that the significance of zero steady-state error is task-dependent. Contact tasks often tolerate deliberate pose error in exchange for regulated force, and legged robots accept bounded periodic error in exchange for lower impact and energy consumption.

\subsubsection{Classical and Modern Control Methods}
\label{sec:control-methods}

One of the most widely used control systems is proportional-integral-derivative (PID) control. It maps three interpretable features of the error into control action. In its ideal continuous-time form,
\begin{equation}
    u(t)=K_{\mathrm{P}} e_{\mathrm{track}}(t)+K_{\mathrm{I}}\int_{0}^{t}e_{\mathrm{track}}(\tau)\,\mathrm{d}\tau+K_{\mathrm{D}}\frac{\mathrm{d}e_{\mathrm{track}}(t)}{\mathrm{d}t},
\end{equation}
where $K_{\mathrm{P}}$, $K_{\mathrm{I}}$, and $K_{\mathrm{D}}$ are the proportional, integral, and derivative gains, and $e_{\mathrm{track}}(t)$ is the tracking error defined above. The proportional term $K_{\mathrm{P}} e_{\mathrm{track}}(t)$ responds to the present error, the integral term $K_{\mathrm{I}}\int_0^t e_{\mathrm{track}}(\tau)\,\mathrm{d}\tau$ to its accumulated history, and the derivative term $K_{\mathrm{D}}\,\mathrm{d}e_{\mathrm{track}}/\mathrm{d}t$ to its local trend. For position or angle error with force or torque output, proportional action resembles a virtual spring, derivative action resembles a virtual damper, and integral action resembles a slowly adjusting bias that offsets gravity or friction. In a torque-controlled robot this analogy connects PID directly to impedance design. It also exposes the hazard of aggressive tuning. A stiff virtual spring improves free-space positioning while generating unsafe forces after unexpected contact, so that gains encode a physical interaction policy rather than a numerical convergence rate alone.

In practice, PID implementations modify the ideal equation in several ways. First, the derivative term is filtered against measurement noise and applied to the output rather than the reference to avoid derivative kick. Integral action also requires anti-windup. Additionally, actuator dead zones, quantization, rate limits, and computation delay can dominate performance if ignored~\citep{AstromHagglund2005}. Tuning rules such as Ziegler-Nichols \citep{ZieglerNichols1942} provide useful initial gains, but they do not replace validation on hardware. Practical implementations typically nest current, velocity, and position loops of decreasing bandwidth. When this bandwidth separation is weak, the independently tuned loops interact and may destabilize the mechanism. PID control is best suited to regulation tasks with few variables and nearly constant dynamics. Multivariable coupling, nonminimum-phase behavior, hard constraints, and large changes in the operating range call for model-based methods instead. Even then, PID often persists at the actuator interface, because its failure modes are familiar and its behavior is inspectable.

\paragraph*{State Space and Optimal Control}
Another model is the state-space model, which represents the internal configuration required to predict future evolution. A linear time-invariant model has the form
\begin{equation}
    \dot{\mathbf{x}}(t)=\mathbf{A}\mathbf{x}(t)+\mathbf{B}\mathbf{u}(t)+\mathbf{E}\mathbf{w}(t), \qquad
    \mathbf{y}(t)=\mathbf{C}_{\mathrm{obs}}\mathbf{x}(t)+\mathbf{v}^{\mathrm{noise}}(t),
\end{equation}
where $\mathbf{x}\in\mathbb{R}^n$ is the state, $\mathbf{u}\in\mathbb{R}^m$ the control input, $\mathbf{w}$ a process disturbance, $\mathbf{y}$ the measurement, $\mathbf{v}^{\mathrm{noise}}$ measurement noise, and $\mathbf{A}$, $\mathbf{B}$, $\mathbf{E}$, $\mathbf{C}_{\mathrm{obs}}$ are matrices of compatible dimensions. A nonlinear system $\dot{\mathbf{x}}=f_{\mathrm{c}}(\mathbf{x},\mathbf{u},\mathbf{w})$ can be linearized about an equilibrium or nominal trajectory. Controllability describes whether the input can move every state direction over a finite interval, and observability describes whether the initial state can be recovered from the input-output history. The weaker properties of stabilizability and detectability suffice for stabilization and estimator convergence, requiring only that every uncontrollable or unobservable mode already be stable. As a result, no controller compensates for an actuator lacking authority over an unstable direction, and no model complexity recovers a contact state invisible to the measurements~\citep{AstromMurray2008}. When the full state is not measured, an observer estimates it. For a linear model with zero-mean white process and measurement noises of known covariance, the Kalman filter is the minimum-mean-square-error estimator among linear estimators, and it is the optimal estimator without that restriction when the initial state and the noises are jointly Gaussian~\citep{Kalman1960}. Because estimation error enters the loop as a dynamic signal, estimator bandwidth, outlier handling, and covariance calibration are part of controller design rather than preprocessing details.

The linear-quadratic regulator (LQR) \citep{Kalman1960LQR} renders the trade-off between regulation and control effort explicit. For the system $\dot{\mathbf{x}}=\mathbf{A}\mathbf{x}+\mathbf{B}\mathbf{u}$, it minimizes
\begin{equation}
    J_{\mathrm{cost}}=\int_{0}^{\infty}
    \left(\mathbf{x}^\top\mathbf{Q}\mathbf{x}+\mathbf{u}^\top\mathbf{R}_{\mathrm{u}}\mathbf{u}\right)\,\mathrm{d}t,
\end{equation}
with $\mathbf{Q}\succeq 0$ and $\mathbf{R}_{\mathrm{u}}\succ 0$. Assuming $(\mathbf{A},\mathbf{B})$ stabilizable\footnote{A pair $(\mathbf{A},\mathbf{B})$ is stabilizable when every eigenmode of $\mathbf{A}$ that cannot be influenced by the input is already asymptotically stable. Likewise, $(\mathbf{A},\mathbf{Q}^{1/2})$ is detectable when every eigenmode of $\mathbf{A}$ that is unobservable through $\mathbf{Q}^{1/2}\mathbf{x}$ is already asymptotically stable. These are necessary and sufficient conditions for the existence of stabilizing linear state feedback and a linear observer with asymptotically stable error dynamics, respectively. Here detectability refers to the cost output $\mathbf{Q}^{1/2}\mathbf{x}$.} and $(\mathbf{A},\mathbf{Q}^{1/2})$ detectable, the unique stabilizing positive-semidefinite solution $\boldsymbol{\Pi}$ of
\begin{equation}
    \mathbf{A}^\top\boldsymbol{\Pi}+\boldsymbol{\Pi} \mathbf{A}-\boldsymbol{\Pi} \mathbf{B} \mathbf{R}_{\mathrm{u}}^{-1}\mathbf{B}^\top\boldsymbol{\Pi}+\mathbf{Q}=0,
\end{equation}
yields $\mathbf{K}=\mathbf{R}_{\mathrm{u}}^{-1}\mathbf{B}^\top\boldsymbol{\Pi}$ and the state-feedback law $\mathbf{u}=-\mathbf{K}\mathbf{x}$~\citep{LewisVrabieSyrmos2012}. The optimum is conditional on the specified model, cost, horizon, and information pattern. $\mathbf{u}^\top\mathbf{R}_{\mathrm{u}}\mathbf{u}$ is an effort surrogate that equals physical energy consumption only under appropriate modeling. Trajectory tracking applies time-varying LQR to deviations linearized along a nominal trajectory, and applying the gain to a Kalman estimate yields the linear-quadratic-Gaussian controller under the separation principle. The principal limitation is constraint handling. LQR does not natively enforce joint limits, contact, collision avoidance, or torque saturation, and clipping an optimal command after computation alters the closed-loop dynamics and can invalidate the nominal analysis. Constrained quadratic programs impose instantaneous limits, while model predictive control extends the optimization across future steps.

\paragraph*{Model Predictive Control}
Model predictive control (MPC) is also commonly used and repeatedly solves a finite-horizon optimal control problem using the latest state estimate. A representative nominal discrete-time formulation is
\begin{equation}
\begin{aligned}
    \min_{\mathbf{u}_0,\ldots,\mathbf{u}_{H_{\mathrm{MPC}}-1}}\quad &
    \sum_{k=0}^{H_{\mathrm{MPC}}-1}\ell(\mathbf{x}_k,\mathbf{u}_k)+V_{\mathrm{f}}(\mathbf{x}_{H_{\mathrm{MPC}}}) \\
    \text{subject to}\quad & \mathbf{x}_0=\hat{\mathbf{x}}_{\mathrm{now}}, \\
    & \mathbf{x}_{k+1}=f_{\mathrm{d}}(\mathbf{x}_k,\mathbf{u}_k), && k=0,\ldots,H_{\mathrm{MPC}}-1, \\
    & \mathbf{x}_k\in\mathcal{X}_{\mathrm{adm}},\quad \mathbf{u}_k\in\mathcal{U}, && k=0,\ldots,H_{\mathrm{MPC}}-1, \\
    & \mathbf{x}_{H_{\mathrm{MPC}}}\in\mathcal{X}_{\mathrm{term}}\subseteq\mathcal{X}_{\mathrm{adm}},
\end{aligned}
\end{equation}
where $\ell(\mathbf{x}_k,\mathbf{u}_k)$ and $V_{\mathrm{f}}(\mathbf{x}_{H_{\mathrm{MPC}}})$ are the stage and terminal costs, $f_{\mathrm{d}}$ is the discrete-time plant dynamics, $\hat{\mathbf{x}}_{\mathrm{now}}$ is the most recent state estimate, $\mathcal{X}_{\mathrm{adm}}$ and $\mathcal{U}$ are the admissible state and input sets, and $\mathcal{X}_{\mathrm{term}}$ is a terminal constraint set. Only the first optimized input is applied. The horizon then shifts forward, a new estimate is obtained, and the problem is solved again. When the optimization remains feasible and is solved as specified, this receding-horizon procedure induces a feedback law on the current state estimate.

MPC is attractive for embodied systems because constraints are part of the decision problem. A manipulator can bound joint angles, velocities, torques, or predicted collision distances, and a legged robot can also consider friction and feasible contact forces. For nominal regulation, recursive feasibility and asymptotic stability can be established through a positive-definite stage cost, a terminal controller that respects the constraints, a positively invariant terminal set, and a terminal cost decrease condition, provided that the initial problem is feasible~\citep{mayne2000constrained}. Design choices are consequential. Short horizons reduce computation but limit how far ahead the controller predicts. Hard constraints declare violations inadmissible, whereas selected constraints may be softened with slack variables when requirements conflict. With linear dynamics, linear constraints, and convex quadratic costs, MPC is a convex quadratic program. Nonlinear MPC typically uses iterative numerical optimization, often with warm starts and sparse solvers.

Prediction does not eliminate the need for feedback or robustness. Model mismatch grows along the horizon, particularly during impacts, slip, or deformable contact. Tube MPC, stochastic MPC, chance constraints, or min-max formulations reserve margins against possible deviations. The system must define the response to infeasibility or solver timeout, whether retaining the previous command, switching to a stabilizing backup, or braking. Controller selection should follow the structure of the task, and the methods are frequently composed. MPC supplies whole-body references, an intermediate controller stabilizes deviations, and embedded PID loops realize actuator commands.

\subsubsection{Hierarchical Architectures of Autonomous Systems}

\paragraph*{The Classical Sense-Plan-Act Pipeline}
Sense-Plan-Act decomposes autonomous behavior into perception, planning, and control. Perception converts sensor data into estimates of robot and environment state. Planning selects goals, actions, or trajectories. Control converts plans into actuator commands while correcting deviations~\citep{Gat1998}. The decomposition is misleading when read as a one-shot sequence. By the end of a long perception-and-planning cycle, the environment may have changed and localization uncertainty may have grown. Autonomous operation therefore requires repeated sensing and replanning, with components running concurrently under explicit latency budgets, because an accurate estimate of an old state can be less useful than a coarser estimate of the present one.

One design requirement for this pipeline is that information passed between the perception, planning, and control stages should include uncertainty and validity, not only nominal values \citep{SicilianoKhatib2016}. For example, a pose estimate without covariance encourages downstream modules to treat perception as certain, and a path without timing assumptions leaves control unable to distinguish acceptable tracking error from an obsolete plan. Interfaces should include reference frames, units, update rates, tolerances, and failure conditions, particularly when learned components are integrated with model-based control \citep{BrugaliScandurra2009}. Feedback paths should also cross stage boundaries. Control influences perception through active sensing, and a geometrically valid path may be dynamically infeasible. Finally, real-time performance is governed by worst-case timing rather than average throughput. Occasionally, a planner that requires 2 seconds may be unacceptable for collision avoidance. A predesigned degradation behavior provides a backup solution. Also, scheduling priorities must follow physical urgency, because motor stabilization and contact protection cannot wait for task-level planning.

\paragraph*{Behavior Trees and Finite-State Machines}
Finite-state machines (FSMs) can also formalize task execution as discrete modes and transitions triggered by events or conditions, with each state activating corresponding controllers or planners. Hybrid automata extend the representation by associating continuous dynamics with each mode \citep{Henzinger1996}, a natural description of systems that alternate among free motion, contact, and impact. However, a limitation of a flat FSM is combinatorial growth. As recovery cases and task variants accumulate, transitions proliferate and behavior becomes difficult to inspect. Hierarchical state machines mitigate this growth, but their interrupt, priority, and concurrency semantics must be specified explicitly. A logically valid transition may be physically unsafe if it switches references discontinuously or transfers control without resetting estimator and integrator states.

Behavior trees organize task logic through a rooted tree that is evaluated, or ``ticked,'' repeatedly. Internal control-flow nodes implement sequences, fallbacks, and parallel combinations. Leaf nodes evaluate conditions or execute actions. Every node returns one of the three statuses \emph{Success}, \emph{Failure}, or \emph{Running} to its parent. A sequence ticks its children until one returns \emph{Running} or \emph{Failure}. A fallback ticks its children until one returns \emph{Running} or \emph{Success}. Because a subtree exposes a small status interface, it can be reused across tasks, and repeated evaluation supports reactive switching~\citep{ColledanchiseOgren2018,IovinoEtAl2022}. Behavior trees do not themselves solve planning or control. A leaf labeled \emph{grasp object} still requires perception, control, and recovery logic, and apparent modularity can conceal resource conflicts between branches. The choice between the representations should follow execution semantics. FSMs suit a small number of modes with well-defined transitions and admit formal verification, whereas behavior trees favor hierarchical composition and reusable recovery strategies, with an independent monitor retaining override authority for safety-related functions. Task-level switching also interacts with continuous control. Stability of each mode does not imply stability under arbitrary switching. Common and multiple Lyapunov functions, dwell-time conditions, guards, and reset maps address different classes of switched and hybrid systems~\citep{Liberzon2003}, and the task executive and the low-level controller must share explicit entry, exit, and state-transfer rules.

\paragraph*{Coordinating Reactive and Deliberative Systems}
Reactive control maps current observations directly to actions, with little or no explicit look-ahead, whereas deliberative control constructs and evaluates possible futures. Neither is sufficient alone. Pure deliberation becomes slow or brittle once its world model is stale. Pure reaction may cycle, become trapped, or sacrifice a long-term goal. Layered architectures therefore emerged to combine the two~\citep{Brooks1986,Gat1998}. A useful architecture separates time scales while preserving channels for intervention. Its layers range from actuator limits, balance, and contact stability through short-horizon motion generation and skill selection to mission-level reasoning over goals and resources. Each layer has a characteristic update rate, abstraction, authority, and failure response. Higher layers parameterize lower layers, and lower layers report feasibility, progress, and exceptions rather than merely accepting commands.

Coordination can be expressed through references and invariant sets. A local controller accepts a high-level goal only if it lies within a verified operating envelope, and it modifies or rejects commands that approach limits of balance, collision distance, or actuator temperature, reporting the reason. Learned policies fit naturally into this view. A policy may propose task choices, waypoints, residual torques, or complete actions, and its placement determines the consequences of error, because a high-level proposal can be checked by planning and constrained control whereas an end-to-end torque policy admits little downstream correction. When layers disagree, summing commands can produce an action intended by neither. Priority arbitration, constrained optimization, and supervisory switching provide explicit alternatives, and hysteresis and minimum dwell times prevent chattering. The resulting system is a hierarchy of closed loops rather than a linear pipeline. Each layer monitors variables appropriate to its horizon and receives feedback from below and from the environment, and autonomy arises when these loops maintain short-term viability while advancing long-term objectives.

\subsubsection{Stability and Robustness Theory}

Feedback enables a controller to respond to deviations from a reference, while stability and robustness analysis establishes the closed-loop guarantees under specified conditions. This subsection reviews Lyapunov methods, barrier functions, and robust and adaptive control as formal tools for making such guarantees explicit.

\paragraph*{Lyapunov Stability}
The stability of a closed-loop system depends on the specified equilibrium, trajectory, set, or orbit. For an equilibrium $\mathbf{x}=\mathbf{0}$, Lyapunov stability requires that trajectories starting sufficiently close remain close, asymptotic stability additionally requires convergence, and exponential stability specifies a convergence-rate bound \citep{khalil2002nonlinear}. These statements are not interchangeable, and a local certificate asserts nothing about initial conditions outside its certified region of attraction. Lyapunov's direct method establishes stability without solving the differential equation. Let $f$ be locally Lipschitz on a neighborhood $\mathcal{D}_{\mathrm{Lyap}}$ of the equilibrium, with $f(\mathbf{0})=\mathbf{0}$. If a continuously differentiable function $V:\mathcal{D}_{\mathrm{Lyap}}\rightarrow\mathbb{R}$, called a Lyapunov function candidate and distinct from the terminal cost $V_{\mathrm{f}}$ used in model predictive control above, satisfies
\begin{equation}
    V(\mathbf{0})=0, \qquad V(\mathbf{x})>0, \qquad
    \dot{V}(\mathbf{x})=\nabla V(\mathbf{x})^\top f(\mathbf{x})<0,
    \quad \mathbf{x}\in\mathcal{D}_{\mathrm{Lyap}}\setminus\{\mathbf{0}\},
\end{equation}
then the equilibrium is locally asymptotically stable. If these conditions hold on all of $\mathbb R^n$ and $V$ is proper and radially unbounded, the equilibrium is globally asymptotically stable. If $\dot V$ is only negative semidefinite, stability follows, but asymptotic convergence requires additional reasoning such as LaSalle's invariance principle~\citep{khalil2002nonlinear}. Every stability claim should be paired with its domain, assumptions, and convergence notion.

For control-affine dynamics $\dot{\mathbf{x}}=f(\mathbf{x})+g_{\mathrm{in}}(\mathbf{x})\mathbf{u}$, a control Lyapunov function identifies inputs that render the derivative sufficiently negative and can be imposed as a constraint in an online optimization that remains close to a nominal command. This construction bridges nonlinear control and learning. A neural network may propose a command that a model-based filter modifies when the constraint would otherwise be violated. Claims that a learned certificate renders a neural controller unconditionally stable are nevertheless too broad. The positivity and decrease conditions must hold over a stated domain, not merely on sampled training points, and verification may require analytic bounds, interval methods, mixed-integer formulations, sum-of-squares relaxations, or reachability analysis. Neural representation changes the parameterization of the candidate, not the logical requirements of the theorem.

Stability also differs from safety. An asymptotically stable robot may pass through an obstacle on its way to the goal. Safety is commonly formulated as forward invariance of a set $\mathcal{C}_{\mathrm{safe}}=\{\mathbf{x}:h_{\mathrm{safe}}(\mathbf{x})\geq 0\}$. For a differentiable relative-degree-one constraint and control-affine dynamics, a zeroing control barrier function requires an admissible input $\mathbf{u}\in\mathcal{U}$ satisfying
\begin{equation}
    L_fh_{\mathrm{safe}}(\mathbf{x})+L_{g_{\mathrm{in}}}h_{\mathrm{safe}}(\mathbf{x})\mathbf{u}+\alpha_{\mathrm{safe}}(h_{\mathrm{safe}}(\mathbf{x}))\geq 0,
\end{equation}
where $L_fh_{\mathrm{safe}}=\nabla h_{\mathrm{safe}}^{\top} f$, $L_{g_{\mathrm{in}}}h_{\mathrm{safe}}=\nabla h_{\mathrm{safe}}^{\top} g_{\mathrm{in}}$, and $\alpha_{\mathrm{safe}}$ is a locally Lipschitz extended class-$\mathcal{K}$ function. If a well-posed feedback law selects such an input and the regularity assumptions of the barrier theorem hold, trajectories starting in $\mathcal{C}_{\mathrm{safe}}$ remain in $\mathcal{C}_{\mathrm{safe}}$~\citep{AmesEtAl2017}. Feasibility can fail under input limits or conflicting barriers, so the inequality is not an unconditional safety guarantee. Lyapunov and barrier constraints can be combined in a quadratic program, with safety taking priority over convergence when relaxation is necessary. Formal claims must finally encompass computation and implementation. A continuous-time proof may not survive slow sampling, solver timeout, communication delay, or state-estimation error. A credible assurance case therefore connects the theorem to timing tests, uncertainty bounds, fault handling, and a validated operating domain, with formal analysis and empirical stress testing supporting one another.

\paragraph*{Robust Control}
Robust control studies whether stability and performance persist across a specified family of plants and disturbances. The uncertainty set is the central object. It may represent bounded parameter variation, neglected high-frequency dynamics, or uncertain contact stiffness, and no controller is robust to unspecified uncertainty of arbitrary magnitude. A common objective bounds the induced gain from a disturbance $\mathbf{w}$ to a performance output $\mathbf{z}_{\mathrm{perf}}$. In $H_\infty$ control, the closed-loop transfer matrix $\mathbf{T}_{zw}$ from the disturbance $\mathbf{w}$ to the performance output $\mathbf{z}_{\mathrm{perf}}$ is designed so that
\begin{equation}
    \lVert \mathbf{T}_{zw}\rVert_\infty
    =\sup_{\omega\in\mathbb{R}}\bar{\sigma}\!\left(\mathbf{T}_{zw}(\mathrm{j}\omega)\right)
    < \gamma_{\mathrm{gain}},
\end{equation}
where $\bar{\sigma}$ denotes maximum singular value. For a stable proper linear time-invariant interconnection with zero initial condition, this is the induced $L_2$-gain bound $\lVert \mathbf{z}_{\mathrm{perf}}\rVert_2<\gamma_{\mathrm{gain}}\lVert \mathbf{w}\rVert_2$ for nonzero $\mathbf{w}\in L_2$. Weighting functions specify which signals and frequency ranges matter, and $\gamma_{\mathrm{gain}}$ is a certified performance bound for the chosen interconnection, not a probability of success in an arbitrary environment~\citep{ZhouDoyleGlover1996}.

Frequency-domain concepts clarify why high nominal performance can erode tolerance to uncertainty. Gain and phase margins describe only particular scalar perturbations in a single loop, singular values generalize sensitivity analysis to multivariable systems, and structured singular-value methods distinguish repeated, real, complex, and dynamic uncertainties. Finer uncertainty models reduce conservatism but demand stronger plant knowledge. The guarantee covers only the uncertainty that analysis has identified and bounded. Robust performance also trades against nominal agility, and excessive conservatism is itself hazardous when a vehicle cannot maneuver quickly enough to avoid a collision. Learning reduces model error but introduces new uncertainty. A residual dynamics model may improve tracking in familiar conditions while extrapolating poorly after a payload or surface change. A practical pattern lets the learned component improve performance inside a confidence region while a robust backup or safety filter limits consequences outside it, preserving the separation between empirical performance and a separately justified safety mechanism~\citep{Brunke2022}.

\paragraph*{Adaptive Control}
Adaptive control changes controller parameters online in response to observed behavior \citep{AstromWittenmark1995}. It is useful when the dynamics vary beyond what fixed gains can tolerate, as when a manipulator grasps an unknown payload or tire forces change with road conditions. A canonical analysis considers dynamics linearly parameterized by an unknown constant vector $\boldsymbol{\theta}$. The controller maintains an estimate $\hat{\boldsymbol{\theta}}$, updated from tracking errors and measured regressors. The control and update laws are designed together so that a Lyapunov argument establishes boundedness and, under stated assumptions, tracking convergence. This structure is more specific than unconstrained online learning, because the form of uncertainty, the update signal, and the stability objective are stated in advance~\citep{AstromWittenmark1995}. Gain scheduling selects among precomputed controllers, direct adaptive control updates controller parameters, and indirect adaptive control estimates plant parameters and redesigns the controller. Guarantees proved for constant $\boldsymbol{\theta}$ do not transfer unchanged to time-varying parameters, where bounded rates of variation yield ultimate bounds rather than exact asymptotic convergence.

Accurate tracking does not imply parameter convergence. For a regressor $\boldsymbol{\phi}_{\mathrm{PE}}(t)\in\mathbb{R}^{d_{\mathrm{reg}}}$, a standard persistent-excitation condition requires constants $T_0>0$ and $\mu_{\mathrm{PE}}>0$ such that
\begin{equation}
    \int_t^{t+T_0}\boldsymbol{\phi}_{\mathrm{PE}}(\tau)\boldsymbol{\phi}_{\mathrm{PE}}(\tau)^\top\,\mathrm{d}\tau
    \succeq \mu_{\mathrm{PE}} \mathbf{I}_{d_{\mathrm{reg}}}
    \qquad \text{for all } t\geq 0.
\end{equation}
Under the assumptions of a given adaptive law, such excitation supports convergence because the data distinguish all parameter directions. A robot holding a constant pose may regulate successfully while learning almost nothing about inertial coupling, whereas deliberate excitation improves identification but may conflict with task performance, comfort, or safety. Actions in adaptive control therefore influence both immediate regulation and future knowledge \citep{BarShalomTse1974}.

Classical adaptive laws can drift when disturbances, unmodeled dynamics, or sensor bias violate their assumptions. Projection operators confine estimates to plausible bounds, dead zones prevent adaptation to noise-dominated errors, and leakage or $\sigma$-modification discourages unbounded parameter growth, typically replacing exact asymptotic tracking with uniform ultimate boundedness. Modern learned adaptation extends beyond low-dimensional parameter estimation to latent-context identification, dynamics-model updating, and policy adaptation, but stability does not follow from low prediction loss alone. A conservative architecture confines adaptation to a nominal model or policy inside an envelope maintained by independent constraints, monitoring, and fallback behavior. Stability, robustness, adaptation, and safety answer distinct questions about what trajectories do, whether that behavior persists over a bounded uncertainty set, how the controller changes as information is acquired, and whether forbidden states are avoided. Dependable autonomy requires an explicit argument for each property, and it requires every feedback loop to declare what it regulates, what uncertainty it tolerates, and where its guarantees end.

\subsection{Embodied Cognition and the Development of Intelligence}
\label{sec:embodied-cognition-development}

Embodied cognition examines how intelligence depends on the interaction between an agent’s body and its environment. Perception and action shape one another, and concepts gain meaning through sensory and motor experience. For Physical AI, this view calls for models and representations that are developed and tested through physical interaction.

\subsubsection{Core Claims of Embodied Cognition}

Embodied cognition is not a single theory but a family of claims about the unit of analysis for intelligence. Classical accounts often separate sensing, internal reasoning, and actuation. Embodied accounts emphasize that perception depends on possible movement, bodily dynamics change the control problem, and action alters the observations available next. Embodied, embedded, enactive, and extended cognition make distinct claims about bodily realization, real-time environmental dependence, autonomous sense-making, and the possible inclusion of external resources in a cognitive system~\citep{wilson2002sixviews,clark1997beingthere,varela1991embodied,clark1998extended}. For Physical AI, the common methodological lesson is to analyze the coupled agent-body-environment system. Merely attaching sensors and actuators is not enough. The body and interaction history must materially organize the capability being explained.

Gibson's ecological approach shifts perception from recovering detached object properties to detecting information relevant to an acting organism~\citep{gibson1979ecological}. Its central concept, the \emph{affordance}, is relational rather than a fixed label. The same ledge may afford stepping, climbing, or no safe action for bodies with different abilities~\citep{chemero2003affordances}. An affordance-centered robot must therefore ask not only ``What is this?'' but also ``What can this body do here under its current kinematic, dynamic, and safety constraints?'' Sensorimotor theories complement this view by treating perception as mastery of action-dependent sensory regularities~\citep{oregan2001sensorimotor}. Camera motion can reveal depth, wrist rotation can disambiguate shape, and tactile exploration can reveal compliance. Brooks's situated robotics work in 1991 and subsequent dynamical analyses show how robust behavior can arise from such tight real-time coupling without an exhaustive central reconstruction~\citep{brooks1991intelligence,beer2000dynamical}. These results challenge mandatory complete world models, but they do not show that long-horizon planning or reasoning about absent events can dispense with internal state.

Intelligence also requires the ability to combine learned concepts and apply them to new situations~\citep{fodor1988connectionism}. However, these concepts must be connected to real-world experience. Harnad's symbol-grounding problem highlights that symbols cannot obtain meaning only from relations to other symbols~\citep{harnad1990symbol}. Barsalou's perceptual-symbol theory suggests that concepts are grounded in perceptual and motor experiences~\citep{barsalou1999perceptual,barsalou2008grounded}. For example, the concept of ``heavy'' can be associated with the effort required to lift an object and the adjustments needed to maintain a stable grip. Thus, intelligent systems need representations that support reasoning and generalization while remaining connected to perception and action.

A robot's physical structure can simplify certain movements, while interaction with the environment provides information that helps it make better decisions~\citep{PfeiferBongard2006,wilson2013embodied}. Rather than replacing internal models, physical interaction can complement them by providing feedback and helping robots adapt their behavior. This raises the question of how robots acquire and improve their knowledge and skills through repeated interactions with the physical world.

\subsubsection{Perception-Action Coupling and Development}

Action does more than follow perception. It can also improve perception. Action can help to select viewpoints, create contacts, and determine which perceptual signals may covary. \citet{bajcsy1988active} formalizes active perception by proposing that sensing strategies be chosen jointly with task goals. Held and Hein's 1963 yoked-control experiment further distinguishes receiving visual stimulation from causing it, supporting a developmental role for action-contingent experience under its controlled conditions~\citep{held1963movement}. In robotics, an observation paired with the agent's command, proprioception, and resulting state transition contains information absent from passive video. New motor skills also create developmental cascades. Reaching, locomotion, and tool use change what can be observed and manipulated, thereby changing the data distribution for later learning~\citep{adolph2019motor}.

One aspect of development is learning sensorimotor coordinates and a body schema. Hand-eye coordination requires image displacement, head pose, arm configuration, joint motion, and contact to become mutually predictive. \citet{metta1999reaching} show that an artificial system can refine a visuomotor reaching map through experience without a fully specified prior kinematic calibration. Body-schema research generalizes this result to revisable relations among commands, configurations, contacts, and sensory consequences~\citep{hoffmann2010bodyschema}. The exploration policy matters. Uniform motor babbling may oversample redundant regions, whereas goal babbling selects desired outcomes and learns the inverse mapping needed to reach them~\citep{rolf2010goalbabbling}. Internal models and sensorimotor coupling are therefore complementary. A model compactly predicts bodily contingencies, while intervention and feedback recalibrate it after tool use, wear, damage, or payload change.

The morphology of the physical body can also serve as a dynamical prior for learning and control. Morphological computation describes how geometry, inertia, and compliance reduce or transform what a controller must compute~\citep{paul2006morphological}. McGeer's passive walkers generate gait structure from gravity and mechanics under restricted conditions, and minimally actuated descendants in later work extend the principle to level-ground walking~\citep{mcgeer1990,collins2005efficient}. Compliant bodies can likewise absorb impacts and provide fading-memory dynamics, although formal results apply only under specific assumptions~\citep{hauser2011morphological}. These examples show how a task-matched body supplies a dynamical prior that learning can exploit. The gain is also task-specific, so morphology and learned control must be co-designed rather than treated as alternatives.

Lastly, self-generated experience creates a final selection problem concerning what the agent should practice next. Early curiosity systems reward expected improvement of a predictive model, while learning-progress methods direct attention toward activities that are neither mastered nor irreducibly unpredictable~\citep{schmidhuber1991curious,oudeyer2007intrinsic}. Deep curiosity methods extend action-conditioned prediction error to high-dimensional observations, but evidence from video games does not resolve physical safety, wear, or stochastic ``noisy-TV'' effects~\citep{pathak2017curiosity}. Taken together, active sensing determines what to observe, the body determines how action unfolds, body schemas relate commands to consequences, and intrinsic motivation allocates practice. These mechanisms explain how a learning trajectory can become structured. Developmental robotics asks how they can be integrated across an embodied lifetime.

\subsubsection{Developmental Robotics}

Developmental robotics studies how robots acquire and reorganize their abilities through experience over time. This subsection examines developmental constraints, lifelong and open-ended learning, imitation, and social interaction as complementary ways to guide that process.

Developmental robotics views intelligence as a gradual learning process rather than a collection of independently designed abilities. Autonomous mental development and cognitive developmental robotics study how robots are used both as constructive tests of developmental hypotheses and as systems that may benefit from developmental constraints~\citep{weng2001autonomous,asada2001cognitive,lungarella2003survey}. Development entails more than ordering examples by difficulty. The learner's body, reachable goals, representations, and social relations can all change. Morphological growth or progressively released degrees of freedom can therefore structure exploration by temporarily reducing the search space~\citep{lungarella2002interplay}. Infant development is valuable here as a source of mechanisms such as self-calibration, staged constraints, and skill reuse, rather than as a schedule robots must literally reproduce.

Another important research direction is lifelong robot learning, which focuses on how robots continuously learn new tasks while retaining and reusing previously acquired knowledge and skills~\citep{thrun1995lifelong,parisi2019continual}. Robotics adds active data collection, temporally correlated experience, and simultaneous change in platform and environment, which static-dataset protocols often omit~\citep{lesort2020continual}. Also, retention is only one component of development. An agent may preserve old task performance without inventing goals, composing skills, or changing its state and action representations. Open-ended learning therefore adds representational redescription and goal discovery~\citep{doncieux2018openended}. Progress-based goal exploration offers one operational mechanism. Agents generate and order goals according to current learning progress, allowing acquired skills to become stepping stones for more complex behavior~\citep{oudeyer2007intrinsic,forestier2022imgep}. A developmental architecture must consequently balance plasticity, consolidation, and expansion rather than optimize only final reward.

Additionally, individual exploration is expensive in the high-dimensional spaces of humanoid manipulators. Imitation learning is introduced to reduce this cost by directing attention toward actions another agent has made relevant. Still, it introduces a correspondence problem between different bodies, viewpoints, and intentions~\citep{schaal1999imitation}. Demonstrations should therefore be treated as priors over goals and action structure, not as trajectories to copy blindly, and embodiment-specific practice remains necessary.

Lastly, social learning also requires shared reference. Constructive models show how saliency and sensorimotor regularities can scaffold joint attention~\citep{nagai2003jointattention}, while sociable-robot studies show that timing, gaze, and expression shape the feedback humans provide~\citep{breazeal2003sociable}. Language games with AIBO further demonstrate that social feedback can participate in forming shared perceptual categories and words~\citep{steels2001aibo}. The resulting relation is reciprocal. Language directs attention and decomposes goals, while action constrains linguistic meaning~\citep{cangelosi2010roadmap}. Developmental robotics thus integrates autonomous exploration, continual retention, and social scaffolding over one trajectory. In the studies reviewed here, however, these mechanisms are typically demonstrated separately and over short periods. Evaluation should measure skill diversity, retention, compositional reuse, adaptation to bodily change, safety cost, and tutor effort. This social-developmental account also provides the transition to symbols. Language becomes meaningful when shared expressions are connected to the learner's evolving perceptual and action repertoire.

\begin{figure}[t]
    \centering
    \includegraphics[width=0.85\linewidth]{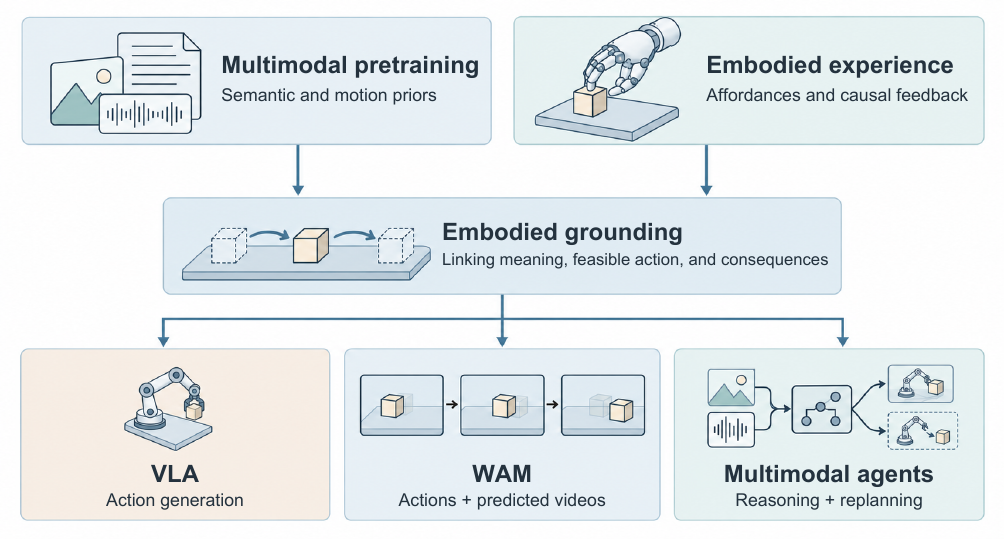}
    \caption{Grounding and multimodal pretraining}
    \label{fig:embodied_grounding_pretraining}
\end{figure}

\subsubsection{Embodied Grounding and Multimodal Pretraining}

For robots to operate effectively in the physical world, they need both broad knowledge of the world and the ability to perform physical actions. Multimodal pretraining on images, text, and videos provides useful semantic knowledge and motion priors, but this knowledge alone is not sufficient for reliable robot control~\citep{bisk2020experience}. Robots also need embodied experience to understand which actions are feasible and what consequences they may produce. Embodied grounding aims to connect these two sources of knowledge, linking what robots perceive and understand to what they can actually do. As illustrated in Figure~\ref{fig:embodied_grounding_pretraining}, this perspective helps explain the roles of different embodied AI approaches, including vision-language-action (VLA) models, world-action models (WAMs), and multimodal agents.

Multimodal pretraining provides a scalable source of prior knowledge. Early systems explicitly connect speech to visual objects and compositional instructions to executable robot plans~\citep{roy2002words,tellex2011commands}. At greater scale, CLIP learns broad image-language correspondences, while R3M extracts temporally informative representations from human video for downstream robot learning~\citep{radford2021clip,nair2023r3m}. Image-text data primarily contribute semantic and referential knowledge. Human video additionally conveys motion patterns and likely object interactions. However, neither directly specifies a particular robot's reachability, actuation limits, contact dynamics, or action costs. Such data therefore provide valuable \emph{semantic and motion priors}, rather than sufficient evidence of executable competence.

Embodied experience supplies complementary evidence by connecting an agent's actions with observed outcomes. Robot demonstrations illustrate how tasks can be accomplished with a particular embodiment, while self-generated interactions reveal feasible actions and their consequences. Work on intuitive physical prediction, robotic poking, and action-conditioned visual foresight exemplifies the transition from observation toward intervention-based learning~\citep{battaglia2013simulation,agrawal2016poking,finn2017foresight}. Modular systems also make the division between meaning and feasibility explicit. CLIPort combines pretrained visual-language semantics with spatial manipulation, and SayCan evaluates language-model skill proposals using embodiment-specific value functions~\citep{shridhar2022cliport,ichter2023saycan}. Yet successful skill selection is not equivalent to learning new affordances or revising semantic beliefs after failure. Robust grounding requires both physically informed action and feedback that can improve subsequent decisions.

Recent embodied models increasingly integrate these complementary sources. VLA models learn language-conditioned control, progressing from RT-1 and RT-2 through heterogeneous cross-robot training to systems such as $\pi_{0.5}$~\citep{pi2025pi05}, which co-trains on robot actions, web data, and semantic subtasks for open-world generalization. EgoScale further explores human egocentric video as scalable motor supervision~\citep{brohan2023rt1,brohan2023rt2,openx2024,pi2025pi05,zheng2026egoscale}. WAMs, including DreamZero and Cosmos Policy, extend action learning by predicting future observations alongside actions, potentially supporting consequence-aware planning~\citep{wang2026wamsurvey,ye2026dreamzero,kim2026cosmospolicy}. General-purpose multimodal agents, as explored in preliminary evaluations of GPT-6 Astra, may instead contribute task-level reasoning, failure diagnosis, and replanning, but are not established substitutes for reliable low-level control~\citep{openai2026astra,su2026astra,zhang2026unexpected}. These are complementary modeling emphases rather than successive stages. The shared challenge is to turn broad multimodal priors and embodiment-specific experience into reliable, adaptive behavior.

\subsection{Learning and Reasoning in the Physical World}
\label{sec:learning-physical-world}

Physical AI requires more than representing the physical world. A learning-based physical AI agent must learn the underlying regularities of physical processes, reason about possible future states, and acquire effective behaviors through interaction. Unlike learning methods that primarily exploit statistical correlations in static data, Physical AI requires models that respect geometric constraints, dynamical laws, and causal consequences of actions.

Modern approaches therefore combine three complementary principles. Physical knowledge provides inductive biases, machine learning extracts unknown structures from experience, and reinforcement learning enables agents to improve through interaction. Together, these components provide the foundation for learning and reasoning in physical environments.

\subsubsection{Physics-Informed Learning Architectures}
\label{sec:physics-informed-learning}

\paragraph*{Physical Inductive Biases}
Physical environments contain strong structural regularities, including symmetry, conservation laws, locality, and relational interactions. Incorporating these properties into machine learning models reduces the hypothesis space and improves data efficiency.

A fundamental example is equivariant learning. Physical predictions should transform consistently under changes of coordinate systems. For a transformation $g_{\mathrm{group}}$ from a symmetry group, an equivariant model satisfies
\begin{equation}
    f(T_{g_{\mathrm{group}}}(x))
    =
    T'_{g_{\mathrm{group}}}(f(x)).
\end{equation}

For 3D physical systems, SE(3)-equivariant networks preserve rotational and translational consistency and have been widely applied to molecular modeling, geometric reasoning, and robotics~\citep{thomas2018tensor,satorras2021e3nn}.

\paragraph*{Graph Neural Networks for Physical Systems}
Many physical environments naturally consist of interacting entities. Particles, objects, robot links, and articulated components can be represented as graphs of the form
\begin{equation}
    G=(\mathcal{V},\mathcal{E}),
\end{equation}
where nodes describe physical entities and edges represent interactions.

Graph neural networks learn system evolution through message passing of the form
\begin{equation}
\mathbf h_i^{\mathrm{node},t+1}
=
\phi_{\mathrm{node}}
\left(
\mathbf h_i^{\mathrm{node},t},
\sum_{j\in \mathcal{N}(i)}
\psi_{\mathrm{msg}}(\mathbf h_i^{\mathrm{node},t},\mathbf h_j^{\mathrm{node},t},\mathbf e_{ij}^{\mathrm{edge}})
\right),
\end{equation}
where $\mathbf h_i^{\mathrm{node},t}\in\mathbb R^{d_{\mathrm{node}}}$ is a node-feature vector and $\mathbf e_{ij}^{\mathrm{edge}}\in\mathbb R^{d_{\mathrm{edge}}}$ is an edge-feature vector. The message function is $\psi_{\mathrm{msg}}:\mathbb R^{d_{\mathrm{node}}}\times\mathbb R^{d_{\mathrm{node}}}\times\mathbb R^{d_{\mathrm{edge}}}\to\mathbb R^{d_{\mathrm{msg}}}$, and the update function is $\phi_{\mathrm{node}}:\mathbb R^{d_{\mathrm{node}}}\times\mathbb R^{d_{\mathrm{msg}}}\to\mathbb R^{d_{\mathrm{node}}}$. The positive integer dimensions $d_{\mathrm{node}}$, $d_{\mathrm{edge}}$, and $d_{\mathrm{msg}}$ need not be equal.

This representation preserves relational structures and enables learning of complex physical dynamics, including particle simulation and object interaction modeling~\citep{sanchezgonzalez2020learning}.

\paragraph*{Physics-Informed Neural Networks}
When governing equations are partially known, physical laws can be directly incorporated into the learning objective. Given a differential equation
\begin{equation}
    \mathcal{N}_{\mathrm{PDE}}[u^{\mathrm{field}}](x,t)=0,
\end{equation}
a neural approximation $u_{\theta}^{\mathrm{field}}(x,t)$ can be trained by minimizing both data error and physical inconsistency:
\begin{equation}
    \mathcal{L}
    =
    \mathcal{L}_{\mathrm{data}}
    +
    \lambda
    \mathcal{L}_{\mathrm{physics}}.
\end{equation}

The physical constraint is enforced through
\begin{equation}
    \mathcal{L}_{\mathrm{physics}}
    =
    \sum_i
    \|
    \mathcal{N}_{\mathrm{PDE}}[u_{\theta}^{\mathrm{field}}](x_i,t_i)
    \|^2 .
\end{equation}

Physics-informed neural networks (PINNs) provide a framework for combining sparse observations with known physical principles and have been widely used for solving and learning differential equations~\citep{raissi2019physics}.

\paragraph*{Neural Differential Equations}
Instead of manually specifying analytical dynamics, neural differential equations use neural networks to parameterize continuous-time evolution as
\begin{equation}
    \frac{\mathrm{d}\mathbf x}{\mathrm{d}t}
    =
    \mathbf f_\theta^{\mathrm{drift}}(\mathbf x,t).
\end{equation}

Neural ordinary differential equations (Neural ODEs) provide a flexible framework for learning continuous latent dynamics~\citep{chen2018neural}. Stochastic extensions further model uncertainty through
\begin{equation}
    \mathrm{d}\mathbf x_t
    =
    \mathbf f_\theta^{\mathrm{drift}}(\mathbf x_t,t)\,\mathrm{d}t
    +
    \mathbf G_\theta^{\mathrm{diff}}(\mathbf x_t,t)\,\mathrm{d}\mathbf W_t^{\mathrm{Wiener}},
\end{equation}
where $\mathbf x_t\in\mathbb R^{d_{\mathrm{state}}}$ is the state and $\mathbf W_t^{\mathrm{Wiener}}\in\mathbb R^{d_{\mathrm{noise}}}$ is a standard Wiener process. For the deterministic and stochastic dynamics above, $\mathbf f_\theta^{\mathrm{drift}}:\mathbb R^{d_{\mathrm{state}}}\times[0,\infty)\to\mathbb R^{d_{\mathrm{state}}}$ is the drift mapping. The diffusion mapping is $\mathbf G_\theta^{\mathrm{diff}}:\mathbb R^{d_{\mathrm{state}}}\times[0,\infty)\to\mathbb R^{d_{\mathrm{state}}\times d_{\mathrm{noise}}}$. The positive integers $d_{\mathrm{state}}$ and $d_{\mathrm{noise}}$ need not be equal, and $\mathrm{d}\mathbf W_t^{\mathrm{Wiener}}$ denotes the Wiener increment.

These approaches bridge classical dynamical systems and modern representation learning by allowing models to learn unknown physical processes while preserving temporal structure.

\subsubsection{Reinforcement Learning for Physical Interaction}
\label{sec:rl-physical-interaction}

\paragraph*{Markov Decision Processes and Physical Interaction}
Learning Physical AI requires not only predicting the world but also selecting actions that influence future states. Reinforcement learning formulates this process as sequential decision-making.

A Markov decision process (MDP) is defined as
\begin{equation}
    (\mathcal{S},\mathcal{A},P^{\mathrm{tr}},R,\gamma),
\end{equation}
where $\mathcal{S}$ represents states, $\mathcal{A}$ represents actions, $P^{\mathrm{tr}}$ represents transition dynamics, and $R$ represents rewards.

The goal of an agent is to learn a policy
\begin{equation}
    \pi(a|s)
\end{equation}
that maximizes cumulative reward:
\begin{equation}
J(\pi)
=
\mathbb{E}_{\pi}
[
\sum_t
\gamma^t r_t
].
\end{equation}

However, physical agents rarely observe complete states. A robot may observe an object's appearance and motion without directly knowing its mass, friction coefficient, or contact condition. Sensor noise, occlusion, and delayed measurements further limit the information available for decision-making.

A partially observable Markov decision process (POMDP) extends the MDP with an observation space $\mathcal{O}$ and an observation model:
\begin{equation}
    (\mathcal{S},\mathcal{A},P^{\mathrm{tr}},R,\mathcal{O},O,\gamma),
\end{equation}
where $O(o_{t+1}\mid s_{t+1},a_t)$ specifies the probability of receiving observation $o_{t+1}$ after action $a_t$ leads to state $s_{t+1}$, and $\gamma\in[0,1)$ is the discount factor. The physical state remains Markovian, but an individual observation need not contain enough information to predict future outcomes.

Decisions therefore depend on both past observations and past actions:
\begin{equation}
    h_t=(o_0,a_0,o_1,\ldots,a_{t-1},o_t),\qquad h_0=(o_0),
    \qquad
    \pi(a_t\mid h_t).
\end{equation}
Past actions matter because they help explain subsequent measurements. A change in an object's position may result from the robot's own push or from an external disturbance.

The history can be summarized by a belief state
\begin{equation}
    b_t(s)=P(s_t=s\mid h_t),
\end{equation}
which represents uncertainty about the current physical state. With a known transition and observation model, the belief is a sufficient statistic of the history for decision-making, allowing a policy to be written as $\pi(a_t\mid b_t)$. Belief updates combine predictions from the previous action with evidence from new observations.

Practical systems may approximate this inference using state estimators, recurrent networks, or policies conditioned on a recent interaction history. For example, lifting an object provides force measurements that help infer its weight and guide subsequent control. POMDPs thus make explicit how uncertain perception, memory, and action jointly shape physical decision-making~\citep{sutton2018reinforcement}.

\paragraph*{Reward Design and Task Specification}
The reward function defines the objective of an embodied agent. Designing appropriate rewards is challenging because physical tasks often involve long horizons and complex constraints.

Sparse rewards provide feedback at infrequent events, such as successful grasping or completion of a navigation task. Although simple, they create exploration difficulties.

Dense rewards introduce intermediate objectives, such as distance reduction, energy efficiency, or trajectory smoothness. However, poorly designed dense rewards may lead to unintended behaviors.

Curriculum learning gradually increases task complexity and allows agents to acquire complex physical skills from simpler subtasks.

\paragraph*{Policy Optimization}
Policy-gradient methods improve a parameterized policy using the outcomes of its actions. With the history-conditioned policy introduced above, a trajectory-based expression is
\begin{equation}
    \nabla_\theta J(\theta)
    =
    \mathbb{E}_{\tau_{\mathrm{ep}}\sim\pi_\theta}
    \left[
        \sum_{t=0}^{\infty}\gamma^t
        \nabla_\theta\log\pi_\theta(a_t\mid h_t)
        A^{\mathrm{adv},\pi_\theta}(h_t,a_t)
    \right],
\end{equation}
where $\tau_{\mathrm{ep}}$ is a sampled interaction trajectory and $A^{\mathrm{adv},\pi_\theta}(h_t,a_t)=Q^{\pi_\theta}(h_t,a_t)-V^{\pi_\theta}(h_t)$ is the advantage, defined as the expected return of the selected action relative to the policy's average return at that history. For fully observed systems, the current state $s_t$ can replace $h_t$. Actor-critic methods use a learned value estimator to approximate the advantage while updating the policy.

Proximal Policy Optimization (PPO) uses a clipped surrogate objective to limit the incentive for large policy changes:
\begin{equation}
\begin{aligned}
    L^{\mathrm{CLIP}}(\theta)
    &=\mathbb{E}_t\left[
        \min\left(
            \rho_t^{\mathrm{PPO}}(\theta)\widehat A_t^{\mathrm{adv}},
            \operatorname{clip}\bigl(\rho_t^{\mathrm{PPO}}(\theta),1-\epsilon_{\mathrm{clip}},1+\epsilon_{\mathrm{clip}}\bigr)\widehat A_t^{\mathrm{adv}}
        \right)
    \right],\\
    \rho_t^{\mathrm{PPO}}(\theta)
    &=\frac{\pi_\theta(a_t\mid h_t)}{\pi_{\theta_{\mathrm{old}}}(a_t\mid h_t)},
\end{aligned}
\end{equation}
where $\theta_{\mathrm{old}}$ denotes the policy parameters used to collect the batch, $\widehat A_t^{\mathrm{adv}}$ is an estimated advantage, $\epsilon_{\mathrm{clip}}>0$ controls clipping, and $\mathbb{E}_t$ denotes averaging over sampled steps. The symbol $\rho_t^{\mathrm{PPO}}$ distinguishes the policy probability ratio from the reward $r_t$. These methods are widely used for locomotion and manipulation~\citep{schulman2017ppo}. Optimization stability and the physical closed-loop guarantees discussed in Section~\ref{sec:control-autonomy} remain separate questions.

\paragraph*{Offline Reinforcement Learning}
Unlike simulation environments, real physical systems cannot collect unlimited interaction data. Offline reinforcement learning addresses this limitation by learning policies from previously collected datasets, including demonstrations and robot trajectories.

This paradigm is particularly important for Physical AI because large-scale datasets can provide diverse physical experiences without requiring unsafe exploration on real hardware.

\subsubsection{Predictive Learning and World Models}

This subsection develops the predictive-learning mechanisms underlying the action-consequence relation discussed in Section~\ref{sec:embodied-cognition-development}. The focus is on constructing internal states, learning how they evolve under actions, and using predicted outcomes for decision-making. The mathematical notion of state and the motivation for latent representations are introduced in Section~\ref{sec:state-predictive-modeling}.

\paragraph*{Learning Latent Physical Dynamics}
Under partial observability, an observation feature and an internal predictive state serve different purposes. An encoder extracts features from the current measurement, while a state-update model integrates them with past actions and the previous internal state:
\begin{equation}
    e_{t}^{\mathrm{enc}}=f_\theta(o_t),
    \qquad
    \mathbf{z}_t=H_\theta(\mathbf{z}_{t-1},a_{t-1},e_{t}^{\mathrm{enc}}),
\end{equation}
where $e_{t}^{\mathrm{enc}}$ denotes an observation feature, $\mathbf{z}_t$ an internal state, and $H_\theta$ a learned update function. This recurrent form is one way to summarize the interaction history $h_t$ used in the POMDP formulation of Section~\ref{sec:rl-physical-interaction}. The internal state should retain task-relevant information about hidden properties and contact conditions. A learned latent vector does not automatically constitute a calibrated belief distribution.

An action-conditioned dynamics model then predicts the next internal state as
\begin{equation}
    \mathbf{z}_{t+1}\sim p_\theta(\cdot\mid \mathbf{z}_t,a_t),
\end{equation}
where $p_\theta$ is a learned transition distribution. A deterministic predictor, $\hat{\mathbf{z}}_{t+1}=F_\theta(\mathbf{z}_t,a_t)$, is a simplified alternative. Training uses recorded observations, actions, and outcomes to learn useful state updates and transitions. Known physical structure can constrain these models through the architectural and loss-based approaches in Section~\ref{sec:physics-informed-learning}. Their adequacy depends on preserving the distinctions that affect future outcomes, including uncertainty that remains after observing the interaction history.

\paragraph*{Joint Embedding Predictive Architectures}
The Joint-Embedding Predictive Architecture (JEPA) approach learns representations by predicting target features from context features. I-JEPA predicts representations of masked image regions from visible context~\citep{assran2023ijepa}, while V-JEPA extends feature prediction to masked spatiotemporal regions in video~\citep{bardes2024vjepa}. A schematic prediction is
\begin{equation}
    \hat{\mathbf{z}}_{\mathrm{target}}
    =
    g_\phi(\mathbf{z}_{\mathrm{context}},m_{\mathrm{target}}),
\end{equation}
where $\mathbf{z}_{\mathrm{context}}$ represents the visible context, $m_{\mathrm{target}}$ specifies target positions, and $g_\phi$ is the predictor. Predictions are matched to features produced by a target encoder. In these implementations, a stop-gradient target and an exponentially averaged target encoder help prevent representation collapse.

Such representations can supply observation features for a predictive world model. Learning from passive images or videos, however, does not by itself identify the effects of executable robot actions. An action-conditioned dynamics model additionally requires experience linking the agent's commands to resulting changes. This distinction separates predictive representation learning from the action-grounded models needed for physical planning and policy improvement.

\paragraph*{World Models, Imagination, and Decision-Making}
Given a current internal state, a dynamics model can roll forward candidate actions. For a deterministic model and an action sequence $a_{t:t+H_{\mathrm{roll}}-1}$ of length $H_{\mathrm{roll}}$, the rollout is
\begin{equation}
    \hat{\mathbf{z}}_t=\mathbf{z}_t,
    \qquad
    \hat{\mathbf{z}}_{t+i+1}=F_\theta(\hat{\mathbf{z}}_{t+i},a_{t+i}),
    \quad i=0,\ldots,H_{\mathrm{roll}}-1.
\end{equation}
A stochastic model instead produces a distribution of possible rollouts. A planner evaluates predicted outcomes against the task objective and constraints, executes a selected action, and updates the internal state from new observations. The receding-horizon control principle is developed in Section~\ref{sec:control-methods}.

Model-based reinforcement learning can also use imagined trajectories to improve a policy. Dreamer learns behavior through rollouts in a learned latent world model~\citep{hafner2019dreamer}, and DayDreamer applies world-model-based learning to physical robots~\citep{wu2023daydreamer}. Prediction thus supports both selecting actions and learning better action policies. Model errors can accumulate over long rollouts, and a policy may exploit inaccurate predictions, so useful imagination requires validation against observed physical outcomes.

These mechanisms connect state representation, physical priors, and reinforcement learning through prediction of action consequences. Chapter~\ref{ch:wm} develops world-model architectures, training objectives, uncertainty, and validation in detail. Chapter~\ref{ch:manipulation-models} and Chapter~\ref{ch:whole-body-motor-control} examine their relationship to learned manipulation and motor capabilities.

\clearpage
\part{Methodology}
\vspace{0.2\linewidth}\epigraph{\textit{"The world is its own best model."}}{--- Rodney Brooks}
\clearpage 

\section{Perception and Cognition}
\label{ch:perception}

This chapter dives deeply into perception and cognition. One of the primary differences between normal AI systems and physical AI systems is that physical AI uses perception and cognition to bridge the physical environment with the actual AI system. The key part for that bridge is the multimodal sensing, which brings information about this complex and diverse world into digital form, and produces a compressed representation.

Therefore, the perception and cognition module aims to obtain a coherent and comprehensive understanding of the physical environment from raw sensor observations. This understanding grounds world models in sensory evidence and supports prediction, planning, and action. This chapter traces that pipeline from low-level multimodal sensing and sensor fusion, 3D spatial and four-dimensional (4D) spatiotemporal perception, to high-level semantic and interaction reasoning.

\subsection{Tasks and Characteristics of Physical Perception}

\begin{figure}[h]
  \centering
  \includegraphics[width=\textwidth]{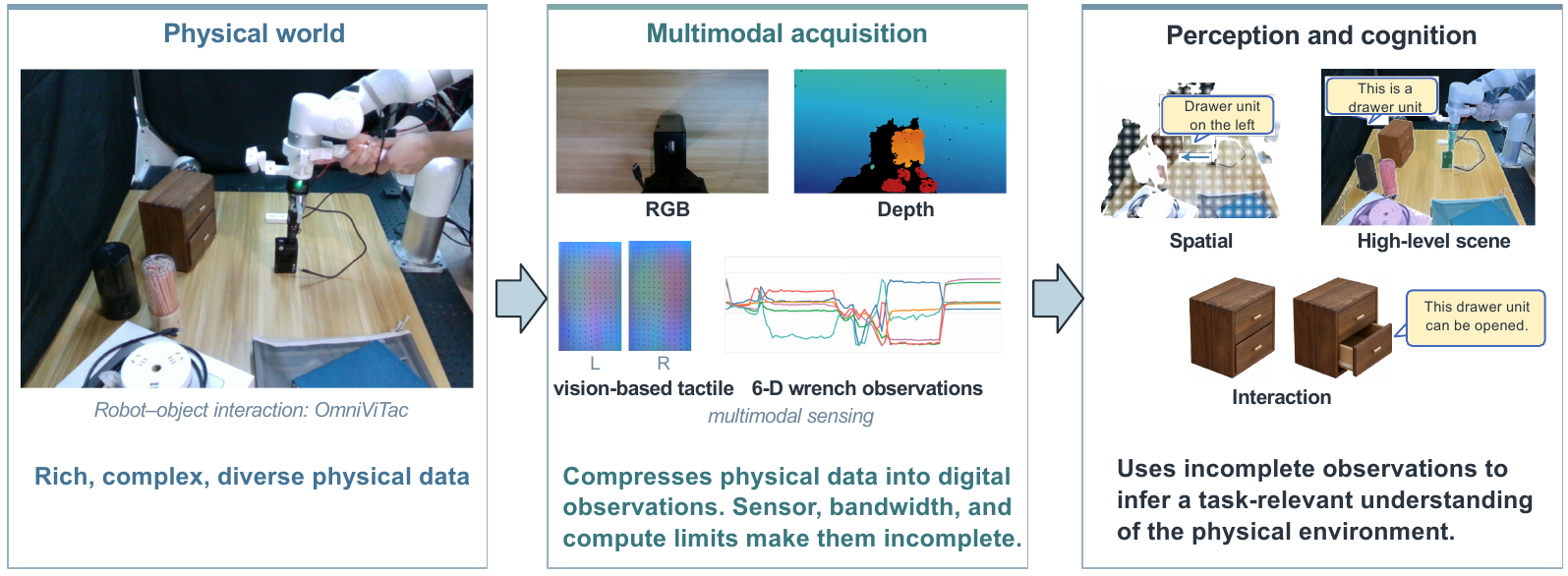}
  \caption{Embodied perception from multimodal observations. A complex physical environment is sampled through multiple sensors, producing a compressed and incomplete digital representation. Perception and cognition integrate these observations to infer the environment state relevant to an embodied agent’s task}
  \label{fig:ch03-overall}
\end{figure}

Perception provides the fundamental interface between the physical world and embodied intelligence. Unlike symbolic or language-based environments, where information is already presented to a model in discrete and machine-readable forms, the physical world does not directly expose its underlying state. It is continuous, high-dimensional, dynamic, and only partially observable. As illustrated in Fig.~\ref{fig:ch03-overall}, an embodied agent can access this world only through sensors that transform physical signals into digital observations. Before reasoning, prediction, planning, or action can take place, these observations must therefore be organized into representations that a computational model can interpret and use.

This transformation is inherently incomplete. Cameras have finite resolution and fields of view, depth sensors contain missing or uncertain measurements, force and tactile sensors observe only local contacts, and all sensing modalities operate with finite sampling rates and measurement noise. Occlusion further prevents parts of the environment from being observed at a given time. Consequently, the digital observations available to an embodied agent constitute only a partial and compressed representation of the underlying physical world. Perception must infer task-relevant structure from this incomplete evidence rather than assuming direct access to the true physical state.

This distinction makes perception particularly important for Physical AI. Downstream models reason and act on the basis of what has been observed and represented, rather than on the physical world itself. Errors or missing information in perception can therefore propagate to state estimation, prediction, reasoning, planning, and ultimately physical action. A missed obstacle may lead to a collision. An incorrect depth estimate may produce an invalid grasp. An incorrect estimate of object or material state may cause an otherwise reasonable plan to fail. Reliable Physical AI thus depends not only on the capability of downstream models, but also on the accuracy, completeness, consistency, and uncertainty of the perceptual representations on which they operate.

Physical perception addresses this problem by acquiring multimodal observations and progressively transforming them into representations of the environment that are useful for embodied tasks. Beyond recognizing \emph{what} is present, a physical perception system must determine \emph{where} entities are, \emph{how} they move and change over time, \emph{what} physical and semantic states they are in, and \emph{how} they can be interacted with. Perception is therefore not merely a preprocessing stage for downstream intelligence, but a continuous process of constructing and updating an internal representation of the physical world.

Physical perception can be broadly organized into three levels. \emph{Low-level perception} acquires observations through sensory modalities. These include color cameras using red, green, and blue (RGB) channels, together with depth sensors, LiDAR, infrared and event cameras, tactile and force sensors, and proprioception. These modalities capture complementary aspects of appearance, geometry, motion, temperature, contact, and the agent's own physical state. \emph{Mid-level perception} organizes these observations into spatially and temporally consistent representations, including 3D geometry, motion, and dynamic 3D/4D scene structure. \emph{High-level perception and cognition} further associates these representations with semantic entities, physical properties and states, functional relationships, affordances, and interaction possibilities, enabling task-dependent interpretation and reasoning.

Across these levels, physical perception is characterized by \emph{multimodality, spatiotemporal consistency, physical grounding, uncertainty, and interaction-oriented understanding}. Multimodal observations compensate for the limitations of individual sensors, while spatiotemporal modeling connects partial observations across viewpoints and time. Physical grounding associates semantic interpretation with measurable geometry, state, and physical evidence, and explicit treatment of uncertainty reflects the unavoidable incompleteness of sensing. Interaction-oriented understanding further focuses perception on the information that matters for action, including contact regions, object states, material conditions, and possible state transitions.

These requirements drive physical perception beyond isolated single-modality recognition toward integrated systems that continuously combine multimodal sensing, 3D/4D spatial modeling, semantic understanding, physical measurement, and interaction feedback. From this perspective, perception extends beyond recognition to make the physical world observable, representable, and actionable to an embodied intelligent system.

\subsection{Multimodal Perception and Fusion}
\label{sec:multimodal-perception}

\subsubsection{Multimodal Sensing and Data Acquisition}
\label{subsec:multimodal-sensing-acquisition}

Multimodal sensing and data acquisition refer to the processes by which an embodied system measures the physical state of itself and its environment and converts heterogeneous measurements into digital observations. Perceiving and interacting with a physical environment require information about several complementary properties, including appearance, geometry, motion, temperature, and contact state. No single sensing modality can reliably and completely observe all of these properties. Embodied systems therefore commonly combine RGB cameras, depth cameras, LiDAR, force and tactile sensors, and proprioceptive sensors. Exteroceptive sensors observe the external environment, proprioceptive sensors measure the internal configuration and motion of the robot, and contact sensors characterize physical interaction at the robot-environment interface. Together, these measurements form the observational basis for subsequent perception, state estimation, reasoning, planning, and control. Section~\ref{sec:c9:sensing-actuation} provides additional details.

\paragraph*{RGB Cameras}
RGB cameras are the most widely used exteroceptive sensors in embodied intelligence. They measure visible optical radiation and provide dense observations of scene color, texture, boundaries, and appearance. These observations contain rich semantic cues but do not directly encode metric depth or contact state.

To describe how an RGB camera converts incident light into image measurements, this chapter first considers its radiometric image-formation model~\citep{healey1994radiometric,bhuyan2019computer,holst2007cmos}. During an exposure, each photosite integrates the incident light according to its wavelength-dependent response. For a photosite associated with color channel $i_{\mathrm{rgb}}\in\{\mathrm{R},\mathrm{G},\mathrm{B}\}$ and spatial location $\mathbf{u}_{\mathrm{pix}}=(u_{\mathrm{pix}},v_{\mathrm{pix}})$, the resulting raw measurement can be expressed as
\begin{equation}
    y_{i_{\mathrm{rgb}}}^{\mathrm{raw}}(\mathbf{u}_{\mathrm{pix}},t)
    = Q_{i_{\mathrm{rgb}}}^{\mathrm{quant}}\!\left(
      \int_{t}^{t+\Delta t}
      \int_{\Lambda_{\mathrm{vis}}}
      S_{i_{\mathrm{rgb}}}(\lambda)E_{\lambda}(\mathbf{u}_{\mathrm{pix}},\tau)
      \,\mathrm{d}\lambda\,\mathrm{d}\tau
      + \varepsilon_{i_{\mathrm{rgb}}}(\mathbf{u}_{\mathrm{pix}},t)
    \right),
    \label{eq:rgb-formation}
\end{equation}
where $y_{i_{\mathrm{rgb}}}^{\mathrm{raw}}(\mathbf{u}_{\mathrm{pix}},t)$ is the raw measurement of an individual photosite, $E_{\lambda}(\mathbf{u}_{\mathrm{pix}},\tau)$ denotes the spectral irradiance incident on the sensor, $\Delta t$ is the exposure duration, and $S_{i_{\mathrm{rgb}}}(\lambda)$ is the effective spectral responsivity of channel $i_{\mathrm{rgb}}$, incorporating wavelength-dependent optical transmission, color-filter response, quantum efficiency, and photoelectric conversion. The term $\varepsilon_{i_{\mathrm{rgb}}}(\mathbf{u}_{\mathrm{pix}},t)$ represents sensor noise, while $Q_{i_{\mathrm{rgb}}}^{\mathrm{quant}}(\cdot)$ summarizes electronic amplification, analog-to-digital conversion, and quantization. A camera image is therefore an exposure-time integral rather than an instantaneous sample.

The collection of photosite measurements forms a raw mosaic image $Y_{\mathrm{C}}^{\mathrm{raw}}$, which is subsequently transformed by an image signal processor (ISP) into an RGB image
\begin{equation}
    I^{\mathrm{RGB}}
    = F_{\mathrm{ISP}}\!\left(Y_{\mathrm{C}}^{\mathrm{raw}}\right),
    \label{eq:isp-pipeline}
\end{equation}
where $I^{\mathrm{RGB}}$ denotes the output RGB image and $F_{\mathrm{ISP}}(\cdot)$ denotes the camera-specific processing pipeline, which may include demosaicing, white balancing, denoising, color correction, tone mapping, gamma encoding, and compression. Consequently, the final RGB values are not generally proportional to scene radiance, and their photometric interpretation depends on both the camera response and the processing pipeline~\citep{debevec2023recovering,brooks2019unprocessing}.

Complementary to the radiometric model, which describes how pixel values are formed, a geometric image-formation model specifies how 3D scene points are projected onto the two-dimensional (2D) image plane~\citep{hartley2003multiple,szeliski2011computer,torralba2024foundations}. Under the pinhole camera model, the projection of a point $\mathbf{p}_{\mathrm{C}}=(X_{\mathrm{C}},Y_{\mathrm{C}},Z_{\mathrm{C}})^\top$ expressed in the camera coordinate frame is represented in homogeneous coordinates as
\begin{equation}
    \tilde{\mathbf u}_{\mathrm{pix}} \propto \mathbf{K}_{\mathrm{cam}}\mathbf{p}_{\mathrm{C}},
    \qquad
    \tilde{\mathbf u}_{\mathrm{pix}}=(u_{\mathrm{pix}},v_{\mathrm{pix}},1)^\top,
    \label{eq:pinhole-camera}
\end{equation}
where $\propto$ denotes equality up to a nonzero scale and $\mathbf{K}_{\mathrm{cam}}$ is the camera intrinsic matrix, conventionally written as
\begin{equation}
    \mathbf{K}_{\mathrm{cam}}
    =
    \begin{bmatrix}
        f_{\mathrm{x}} & \gamma_{\mathrm{cam}} & c_{\mathrm{x}} \\
        0   & f_{\mathrm{y}}    & c_{\mathrm{y}} \\
        0   & 0      & 1
    \end{bmatrix},
    \label{eq:camera-intrinsics}
\end{equation}
where $f_{\mathrm{x}}$ and $f_{\mathrm{y}}$ are the effective focal lengths measured in pixels, $(c_{\mathrm{x}},c_{\mathrm{y}})$ is the principal point, and $\gamma_{\mathrm{cam}}$ is the skew parameter between the image axes, which is commonly assumed to be zero in modern digital cameras. The intrinsic matrix maps normalized camera coordinates to pixel coordinates. Lens distortion is modeled separately. Its parameters are estimated during camera calibration and subsequently used to undistort or rectify the captured images~\citep{zhang2000flexible}.

Within embodied systems, RGB cameras are commonly mounted on the head or torso to provide egocentric observations, on the wrist or end effector to capture close-range interactions, or in the environment to provide a global view of the robot and workspace. These visual streams constitute a primary source of exteroceptive information for semantic scene understanding, object localization, visual navigation, and visuomotor control. Visual navigation models such as ViNT use camera observations to learn navigational affordances under a general goal-reaching objective~\citep{shah2023vint}, while large-scale manipulation datasets and policies increasingly combine synchronized multi-view imagery with language and proprioceptive state. DROID, for example, records three synchronized stereo RGB camera streams, together with robot states and actions, exterior-camera calibration, and natural-language instructions across 86 manipulation tasks~\citep{khazatsky2024droid}. Generalist VLA models such as $\pi_0$ and $\pi_{0.5}$ similarly combine camera observations with language instructions and robot state to generate actions across diverse tasks and robot embodiments~\citep{black2024pi0,pi2025pi05}. Despite their low cost and semantic richness, RGB cameras do not directly measure metric geometry or physical contact, and their reliability remains sensitive to illumination changes, saturation, specular reflections, motion blur, and occlusion.

\paragraph*{Depth Cameras and LiDAR}
Depth cameras and LiDAR provide metric observations of scene geometry, but their measurement principles and sampling patterns are not determined by the device category alone. Some depth cameras employ stereo triangulation, whereas others use time-of-flight principles closely related to LiDAR. A useful distinction is therefore made between \emph{axial depth} $Z_{\mathrm{depth}}$, defined as the coordinate of a scene point along the camera optical axis, and \emph{ray range} $r$, defined as the Euclidean distance from the sensing origin along a viewing or emission ray. The systems considered here are organized by how distance is inferred, distinguishing \emph{triangulation-based depth sensing} from \emph{propagation-based optical ranging}.

Triangulation-based systems infer depth from correspondences between calibrated rays originating from spatially separated camera or projector centers. For a rectified stereo pair with baseline $B_{\mathrm{stereo}}$, common horizontal focal length $f_{\mathrm{x}}$, and aligned principal points, the horizontal disparity $d_{\mathrm{stereo}}(\mathbf{u}_{\mathrm{pix}})=u_{\mathrm{pix},\mathrm{L}}-u_{\mathrm{pix},\mathrm{R}}$ determines the axial depth. The resulting depth can then be back-projected into the reference-camera coordinate frame using
\begin{equation}
\begin{aligned}
    Z_{\mathrm{depth}}(\mathbf{u}_{\mathrm{pix}})
    &= \frac{f_{\mathrm{x}} B_{\mathrm{stereo}}}{d_{\mathrm{stereo}}(\mathbf{u}_{\mathrm{pix}})}, \\
    \mathbf{p}_{\mathrm{C}}(\mathbf{u}_{\mathrm{pix}})
    &= Z_{\mathrm{depth}}(\mathbf{u}_{\mathrm{pix}})\mathbf{K}_{\mathrm{cam}}^{-1}\tilde{\mathbf u}_{\mathrm{pix}},
\end{aligned}
    \label{eq:stereo-geometry}
\end{equation}
where $f_{\mathrm{x}}$ and $d_{\mathrm{stereo}}(\mathbf{u}_{\mathrm{pix}})$ are expressed in pixels, $\mathbf{K}_{\mathrm{cam}}$ is the intrinsic matrix of the reference camera, and $\tilde{\mathbf u}_{\mathrm{pix}}=(u_{\mathrm{pix}},v_{\mathrm{pix}},1)^\top$ is the homogeneous image coordinate~\citep{hartley2003multiple}. The second relation assumes that the depth image stores axial depth rather than distance along the viewing ray. Passive stereo relies on naturally occurring scene texture to establish correspondence. Active-stereo systems retain the same binocular geometry but project an unstructured infrared pattern to add texture in weakly textured regions. The pattern assists left-right image matching and need not be decoded into projector coordinates. The Intel RealSense D400 series is a representative example~\citep{keselman2017intel}. Structured-light systems also recover depth by triangulation, but use a calibrated camera-projector pair and establish correspondences between observed image locations and encoded projector coordinates~\citep{zanuttigh2016time}. Thus, although active stereo and structured light employ active illumination, both remain triangulation-based because depth is inferred from spatial correspondence rather than optical propagation time.

Propagation-based sensors emit optical signals and infer distance from temporal or phase properties of the returned signal rather than from spatial disparity. For a pulsed time-of-flight LiDAR, the measured round-trip propagation delay $\Delta t_i$ determines the range of the $i$th return. Combining this range with the calibrated beam direction gives its three-dimensional position as
\begin{equation}
\begin{aligned}
    r_i
    &= \frac{c_{\mathrm{light}}\,\Delta t_i}{2}, \\
    \mathbf{p}_i
    &= \mathbf{o}+r_i\mathbf{d}_i,
\end{aligned}
    \label{eq:lidar-geometry}
\end{equation}
where $c_{\mathrm{light}}$ is the speed of light, $\mathbf{o}$ is the sensing origin, and $\mathbf{d}_i$ is a unit vector specifying the calibrated beam direction~\citep{royo2019overview}. A return may additionally include a measure of return strength or intensity $a_{i}^{\mathrm{lidar}}$ and an acquisition timestamp $t_i$. Per-return timing is important for compensating motion distortion, or \emph{deskewing}, when a scanning LiDAR assembles a point cloud over a nonzero time interval. Time-of-flight depth cameras apply a related propagation-based principle at many pixels over a two-dimensional detector array. Depending on the sensor design, they estimate distance from the direct time of flight of optical pulses or from the phase shift of modulated illumination~\citep{horaud2016overview}. The stored value may represent ray range or may be converted into axial depth according to the sensor's calibration and output convention.

The native measurements of these sensors induce different spatial structures. A camera depth image can be back-projected into an organized point cloud whose samples preserve the image lattice, whereas LiDAR measurements generally form an irregular, sparse point cloud indexed by beam direction and acquisition time. After calibration and coordinate transformation, either representation may be converted into voxels, signed-distance fields, or continuous neural implicit fields for downstream perception and planning.

In embodied systems, depth observations are particularly useful for close-range geometry, including object-pose estimation, grasping, and workspace reconstruction. PerAct, RVT, and Act3D, for example, demonstrate how calibrated RGB and depth (RGB-D) observations can be transformed into volumetric, multi-view, or continuous three-dimensional representations for robotic manipulation~\citep{shridhar2023perceiver,goyal2023rvt,gervet2023act3d}. LiDAR is more commonly employed on mobile and field robots, where its typically longer sensing range and direct metric measurements support localization, mapping, obstacle avoidance, and exploration. CERBERUS provides a representative example in subterranean environments~\citep{tranzatto2022cerberus}. These modalities nevertheless exhibit different failure modes. Passive stereo correspondence becomes unreliable in textureless, repetitive, or occluded regions. Active stereo and structured light reduce texture ambiguity but remain limited by projector range, strong ambient illumination, interference, and transparent or specular surfaces. Propagation-based ranging can likewise be degraded by low-return surfaces, strong ambient illumination, multipath interference, and adverse atmospheric conditions.

\paragraph*{Force and Tactile Sensors}
Force and tactile sensors provide measurements of physical interaction that cannot be obtained directly from conventional vision. Force and torque sensors measure the resultant loads transmitted through a robot component, whereas tactile sensors measure local contact conditions over a sensing surface. This distinction concerns the spatial support and physical interpretation of the measurements rather than the transduction material alone. Resistive, capacitive, piezoelectric, optical, and magnetic mechanisms may be used in different sensor configurations.

A multi-axis force/torque sensor measures the resultant force and moment acting on a robot component, commonly at the wrist, ankle, or end effector. A six-axis sensor reports the wrench, expressed in a declared sensor coordinate frame, as
\begin{equation}
\begin{aligned}
    \mathbf{w}_{\mathrm{S}}
    &= [\,\mathbf{f}_{\mathrm{S}}^\top,\boldsymbol{\tau}_{\mathrm{S}}^\top\,]^\top, \\
    \mathbf{w}_{\mathrm{S}}
    &= \mathbf{C}_{\mathrm{cal}}(\mathbf{v}_{\mathrm{sens}}-\mathbf{v}_{\mathrm{sens},0}),
\end{aligned}
    \label{eq:ft-sensing}
\end{equation}
where $\mathbf{f}_{\mathrm{S}}\in\mathbb{R}^3$ and $\boldsymbol{\tau}_{\mathrm{S}}\in\mathbb{R}^3$ are the measured force and torque, $\mathbf{v}_{\mathrm{sens}}$ denotes the raw electrical outputs, $\mathbf{v}_{\mathrm{sens},0}$ is the zero-load offset, and $\mathbf{C}_{\mathrm{cal}}$ is a calibration and decoupling matrix. The second relation is a common linear approximation for strain-based sensors, which infer applied loads from the deformation of an elastic structure~\citep{cao2021six}. Such sensors provide accurate aggregate interaction loads but do not resolve how those loads are distributed across multiple contact points. Joint-torque sensors and motor currents can also support external-force estimation through a robot dynamics model, but these indirect estimates should not be treated as equivalent to a directly measured endpoint wrench.

Tactile sensors measure contact locally at the robot surface. Their outputs may describe contact location, surface deformation, normal and tangential loading, or slip. According to their dominant sensing and readout mechanisms, tactile sensors commonly used in embodied systems can be broadly organized into vision-based tactile sensors and taxel-based tactile arrays.

Vision-based tactile sensors illuminate a compliant elastomer and use an internal camera to observe its deformation during contact. Their native digital observation is an image rather than a direct force measurement:
\begin{equation}
    I_{\mathrm{tac}}(\mathbf{u}_{\mathrm{pix}},t)
    = \mathcal{F}_{\mathrm{opt}}\!\left(
      \boldsymbol{\delta}(\mathbf{x}_{\mathrm{pos}},t);\boldsymbol{\theta}_{\mathrm{tac}}
      \right)+\varepsilon_I(\mathbf{u}_{\mathrm{pix}},t),
    \qquad
    \widehat{\mathbf{z}^{\mathrm{tac}}}_t=G_{\mathrm{tac}}\!\left(I_{\mathrm{tac}}\right),
    \label{eq:vision-tactile}
\end{equation}
where $\boldsymbol{\delta}$ denotes the elastomer deformation, $\mathcal{F}_{\mathrm{opt}}$ summarizes the illumination and imaging process with sensor parameters $\boldsymbol{\theta}_{\mathrm{tac}}$, and $G_{\mathrm{tac}}(\cdot)$ is a calibrated or learned estimator of contact quantities $\widehat{\mathbf{z}^{\mathrm{tac}}}_t$, such as contact geometry, force, shear, or slip. The GelSight study establishes high-resolution optical measurement of contact geometry and force-related deformation~\citep{yuan2017gelsight}, while the DIGIT study demonstrates a compact, lower-cost design suitable for robotic hands~\citep{lambeta2020digit}. Because their output is image-like, these sensors also support visual representation learning. Sparsh, for example, learns transferable representations from unlabeled tactile images~\citep{higuera2024sparsh}.

Taxel-based sensors use arrays of discrete tactile elements distributed over a compliant surface. A pressure-sensitive taxel primarily measures normal contact stress, whereas more general taxels may also respond to tangential stress or local deformation. For taxel $i$ with sensitive region $\mathcal{A}_i$ and area $A_i=\int_{\mathcal{A}_i}\mathrm{d}A$, a generic calibrated measurement model is
\begin{equation}
\begin{aligned}
    \overline{\boldsymbol{\sigma}}_i(t)
    &= \frac{1}{A_i}\int_{\mathcal{A}_i}
       \boldsymbol{\sigma}_{\mathrm{c}}(\mathbf{x}_{\mathrm{pos}},t)\,\mathrm{d}A, \\
    \mathbf{y}_i(t)
    &= g_{i}^{\mathrm{taxel}}\!\left(\overline{\boldsymbol{\sigma}}_i(t),T_{\mathrm{temp}}(t)\right)
       + \boldsymbol{\varepsilon}_i(t),
\end{aligned}
    \label{eq:taxel-measurement}
\end{equation}
where $\boldsymbol{\sigma}_{\mathrm{c}}=[p_{\mathrm{press}},s_{1}^{\mathrm{shear}},s_{2}^{\mathrm{shear}}]^\top$ contains normal pressure $p_{\mathrm{press}}$ and two tangential stress components, and $g_{i}^{\mathrm{taxel}}(\cdot)$ represents the sensor-specific transduction and calibration. Resistive, piezoresistive, and capacitive arrays are widely used for compact quasi-static contact sensing. Piezoelectric elements are particularly responsive to changes in load and vibration. Magnetic or inductive designs can provide compliant multi-axis measurements. These mechanisms differ in sensitivity, bandwidth, hysteresis, drift, wiring complexity, and spatial resolution~\citep{dahiya2009tactile,meribout2024tactile}.

Some specialized tactile devices additionally measure vibration, acoustic emission, or temperature. High-bandwidth vibration and acoustic signals can reveal impact, sliding, texture, and incipient slip, while thermal measurements can contribute to material recognition. These signals are less commonly used as standalone tactile modalities in embodied systems and are often incorporated as complementary channels in multimodal tactile sensors.

The appropriate sensor depends on the spatial scale and information required by the task. Six-axis force/torque sensors are mature choices for aggregate load feedback in compliant control, insertion, assembly, and human-robot interaction. Vision-based tactile sensors are particularly attractive in current robot-learning research because they provide dense contact geometry and can be processed with image-based models, whereas electrical or magnetic taxel arrays can be integrated into compact fingertips, finger links, palms, and larger-area robotic skin. Force and tactile measurements support contact detection, grasp stabilization, slip detection, texture recognition, object reorientation, and dexterous manipulation. RH20T illustrates their role in multimodal robot learning by recording force together with visual, audio, robot-state, action, and language information across diverse manipulation skills~\citep{fang2024rh20t}.

These sensing families also have distinct limitations. Force/torque sensors do not localize individual contacts without additional geometric assumptions. Taxel arrays may exhibit hysteresis, drift, cross-talk, and limited spatial resolution. Vision-based tactile sensors are sensitive to illumination and camera-calibration errors, as well as elastomer wear and viscoelastic effects such as hysteresis and stress relaxation. Their physical estimates additionally depend on calibrated or learned mappings from deformation to contact quantities. All tactile modalities have a local field of view and become informative primarily after contact. They are therefore commonly combined with vision and proprioception. Vision estimates where interaction should occur, proprioception supplies the robot configuration, and force or tactile feedback verifies the resulting physical interaction.

\paragraph*{Proprioceptive Sensors} 
Proprioceptive sensors measure the internal state and motion of the embodied platform. For an articulated robot, a representative proprioceptive observation is
\begin{equation}
    \mathbf{z}_{\mathrm{prop}}(t)
    = [\,
      \mathbf{q}^\top,
      \dot{\mathbf{q}}^\top,
      \boldsymbol{\tau}^\top,
      \mathbf{i}^\top,
      \mathbf{a}_{\mathrm{imu}}^\top,
      \boldsymbol{\omega}_{\mathrm{imu}}^\top
      \,]^\top,
    \label{eq:proprioception}
\end{equation}
where $\mathbf{q}$ and $\dot{\mathbf{q}}$ are joint positions and velocities, $\boldsymbol{\tau}$ denotes measured or estimated joint torque, $\mathbf{i}$ is motor current, and $\mathbf{a}_{\mathrm{imu}}$ and $\boldsymbol{\omega}_{\mathrm{imu}}$ are specific force and angular velocity from an inertial measurement unit. Depending on the platform, proprioception may additionally include wheel odometry, base orientation, gripper aperture, actuator temperature, or foot-contact switches.

These signals support forward kinematics, state estimation, trajectory tracking, balance, and body-aware control. They also provide the configuration needed to transform head-, wrist-, and body-mounted sensor data into a common coordinate system. Joint torque and motor current can support external-force estimation through an explicit robot dynamics model, but the resulting estimate remains sensitive to transmission, friction, and model error. In learning-based embodied systems, proprioceptive state is commonly combined with visual and language inputs so that action generation is conditioned on both the external scene and the current robot configuration~\citep{black2024pi0,khazatsky2024droid}.

\paragraph*{Multisensor Acquisition, Calibration, and Time Synchronization}
Combining sensors requires more than recording their outputs. Different devices use different coordinate frames, sampling rates, exposure intervals, clock domains, communication paths, and latency conventions. Reliable fusion therefore depends on calibrated spatial and temporal relationships between measurements.

Spatial calibration comprises intrinsic calibration of each imaging or ranging sensor and extrinsic calibration between sensor frames. If a point measured in sensor frame $\mathrm{S}_m$ is to be expressed in a reference frame $\mathrm{S}_0$, the homogeneous-coordinate transformation is
\begin{equation}
    \widetilde{\mathbf{p}}_{\mathrm{S}_0}
    = {}^{\mathrm{S}_0}\mathbf{T}_{\mathrm{S}_m}\widetilde{\mathbf{p}}_{\mathrm{S}_m},
    \qquad
    {}^{\mathrm{S}_0}\mathbf{T}_{\mathrm{S}_m}\in \mathrm{SE}(3).
    \label{eq:extrinsic-transform}
\end{equation}
For moving sensors, this transform is composed with the robot kinematic state at the measurement time. Calibration may also include image rectification, depth-to-color registration, tactile-frame localization, force-sensor bias estimation, and calibration of the transformation from sensor output to physical units.

Temporal alignment requires a model of each sensor clock. A common affine approximation relates sensor time $t_m$ to a reference time $t_0$ as
\begin{equation}
    t_m = \alpha_{m}^{\mathrm{clk}} t_0+\beta_{m}^{\mathrm{clk}},
    \label{eq:clock-model}
\end{equation}
where $\beta_{m}^{\mathrm{clk}}$ is the clock offset and $\alpha_{m}^{\mathrm{clk}}$ represents the relative clock rate. A timestamp must also be associated with a clearly defined event. For a camera, that event may be the beginning, midpoint, or end of exposure. For a ranging pulse, it may be emission or reception. A timestamp may also mark the arrival of a sample at the host computer. Joint estimation of spatial and temporal calibration is often necessary because an unmodeled time offset during robot motion can be mistaken for an extrinsic-calibration error~\citep{furgale2013unified}.

Software synchronization aligns independently acquired streams through timestamps, buffering, interpolation, and nearest-neighbor association. It is flexible but cannot recover information lost through inaccurate timestamps or variable transport delay. Hardware synchronization distributes a common clock, trigger, or pulse-per-second signal and is preferable when the platform undergoes rapid motion or contact transitions. Even with hardware triggering, an acquisition system should account for exposure duration, rolling-shutter readout, sensor processing delay, network latency, and the time required to assemble scanning measurements.

The final synchronized record should preserve the raw measurements, calibrated measurements, timestamps, coordinate-frame definitions, calibration parameters, and validity indicators. An RGB image, depth frame, LiDAR return, tactile image, wrench sample, and joint state should either refer to the same physical instant or carry sufficient timing information to be transformed or interpolated to a common instant. This requirement is particularly important for visual-tactile learning, force-guided manipulation, and high-speed navigation, where small temporal errors can associate observations with different robot poses or contact regimes. Careful acquisition and calibration are thus integral to perception and determine the reliability of all subsequent multimodal fusion.

\subsubsection{Multimodal Fusion for Joint Perception}

\label{sec:multimodal-information-fusion}

Multimodal fusion combines complementary observations to estimate the state of the robot, its surroundings, and their interactions. RGB images describe appearance. Depth and LiDAR provide geometric measurements. Force, tactile, and proprioceptive signals supply information about contact and robot motion. A useful organizing principle is the stage at which information is combined, whether at the level of observations, intermediate features, or task-level estimates. These levels describe possible operations rather than mutually exclusive architectures~\citep{hall1997introduction,baltruvsaitis2018multimodal}.

\paragraph{Observation-Level Integration and Cross-Modal Enhancement} When measurements have been spatially registered and temporally aligned, they can be associated in a common coordinate frame. For example, a registered RGB image and depth map associate appearance with measured distance at corresponding image locations. RGB-guided depth completion additionally uses image information to infer depth where measurements are sparse or missing~\citep{ma2018sparse}. Such estimates must remain distinguishable from measured depth. An appearance edge need not coincide with a geometric boundary. Validity masks, calibration, and timestamps are therefore part of the fused observation, especially near occlusions or when the sensors move.

\paragraph{Feature-Level Fusion for Joint Perception} Modality-specific encoders can extract representations that are then combined through concatenation, learned weighting, or attention. The features may be aligned on an image grid or mapped into a shared three-dimensional or bird's-eye-view space. FuseNet combines color and depth features for semantic segmentation~\citep{hazirbas2016fusenet}. BEVFusion combines camera and LiDAR features in a bird's-eye-view representation for three-dimensional perception~\citep{liu2023bevfusion}. In embodied systems, visual, tactile, and proprioceptive features can likewise be related through robot kinematics and contact geometry, even when pixelwise correspondence is unavailable. Feature fusion allows task-dependent interactions but can be sensitive to missing modalities and errors in spatial or temporal alignment.

\paragraph{Estimate-Level Fusion} Alternatively, each sensor stream can produce an estimate, such as object pose, occupancy, or contact state, before the estimates are combined. This approach permits modality-specific processing and can accommodate different sampling rates, provided that the estimates refer to compatible states and their uncertainties and dependencies are accounted for~\citep{hall1997introduction}. Because high-level estimates discard some measurement detail, they may offer less opportunity to exploit fine cross-modal correspondences. In practice, the fusion level is chosen according to the perceptual task, available calibration, measurement reliability, and computational constraints. A system may combine several levels.

\subsection{3D/4D Spatial Perception}
\label{sec:spatial-perception}

\begin{figure}[tbp]
  \centering
  \includegraphics[width=\textwidth]{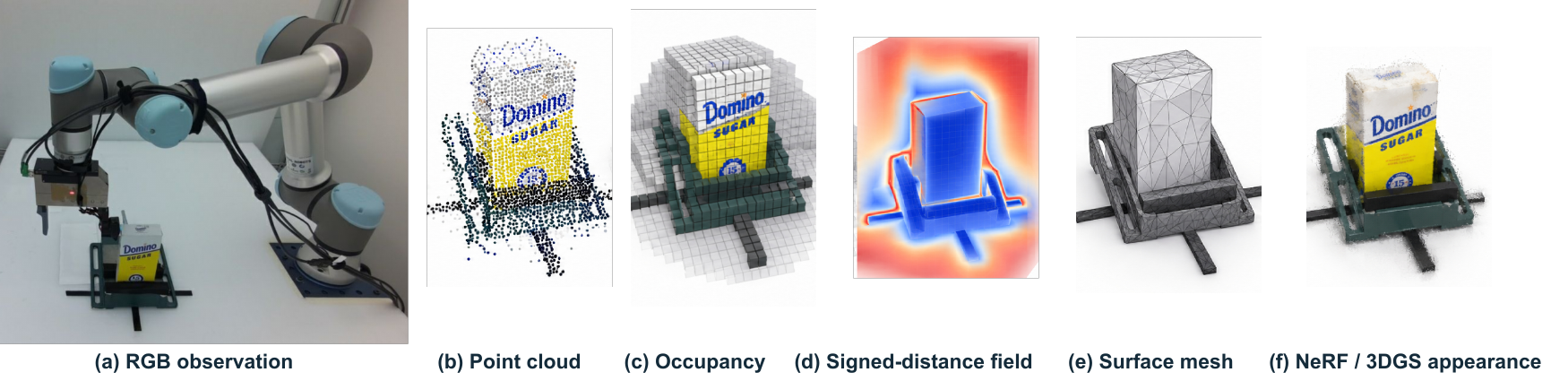}
  \caption{Schematic illustration of different 3D representations of the same object. The 3D visualizations are illustrative and are not ground-truth reconstructions or outputs computed from the source data. The scene image is from~\citet{suresh2022shapemap}}
  \label{fig:ch03-3D-visual}
\end{figure}

Three-dimensional spatial perception estimates scene geometry, object locations, and spatial relationships. 4D perception additionally represents changes over time, including object motion and deformation. Figure~\ref{fig:ch03-3D-visual} illustrates alternative 3D representations of the same object. For an embodied agent, these estimates should support interaction with the physical environment, not only recognition or novel-view rendering. Because sensors observe only limited, noisy projections of the world, spatial perception maintains a task-relevant estimate rather than a complete copy of physical reality.

\subsubsection{3D/4D Scene Representation and Modeling}
Scene representations differ in the properties they encode and the queries they support. In a calibrated pinhole camera, a world point $\mathbf{p}_{\mathrm{W}}$ projects to image coordinates according to
\begin{equation}
    \mathbf{p}_{\mathrm{C}}=\mathbf{R}_{\mathrm{CW}}\mathbf{p}_{\mathrm{W}}+\mathbf{t}_{\mathrm{CW}},
    \qquad \tilde{\mathbf u}_{\mathrm{pix}}\propto\mathbf{K}_{\mathrm{cam}}\mathbf{p}_{\mathrm{C}},
    \label{eq:spatial-camera-projection}
\end{equation}
where $\mathbf{R}_{\mathrm{CW}}$ and $\mathbf{t}_{\mathrm{CW}}$ define the transform from world to camera coordinates, and $\mathbf{K}_{\mathrm{cam}}$ is the intrinsic matrix~\citep{hartley2003multiple}. Given axial depth $Z_{\mathrm{depth}}(u_{\mathrm{pix}},v_{\mathrm{pix}})$, the corresponding camera-frame point is $\mathbf{p}_{\mathrm{C}}=Z_{\mathrm{depth}}(u_{\mathrm{pix}},v_{\mathrm{pix}})\mathbf{K}_{\mathrm{cam}}^{-1}\tilde{\mathbf u}_{\mathrm{pix}}$. Thus, RGB-D data can be lifted into 3D and registered to robot or world coordinates, subject to calibration, depth error, and temporal alignment.

Point clouds retain measured or predicted surface samples, but do not by themselves define continuous surfaces, object interiors, or free space. Occupancy grids represent cells as occupied, free, or unknown, often with probabilistic occupancy. OctoMap is a probabilistic octree-based example~\citep{hornung2013octomap}. Signed-distance fields encode distance to a surface, with the zero level set representing that surface under a chosen sign convention. They support geometric queries such as clearance and collision checking. Voxblox incrementally constructs Euclidean SDFs for planning~\citep{oleynikova2017voxblox}. Meshes explicitly represent surfaces and can support rendering or collision queries when constructed for that purpose. None of these representations alone supplies all physical properties required for interaction, such as friction, mass, articulation, or contact dynamics. Their coverage is also limited by sensor visibility and spatial resolution.

NeRFs represent a scene by mapping 3D position and viewing direction to volume density and view-dependent radiance, $(\sigma_{\mathrm{rad}},\mathbf{c}_{\mathrm{rgb}})=F_{\theta}^{\mathrm{NeRF}}(\mathbf{x}_{\mathrm{pos}},\mathbf{d})$~\citep{mildenhall2020nerf}. 3D Gaussian Splatting (3DGS) represents a scene using anisotropic Gaussian primitives optimized for efficient rendering~\citep{kerbl20233d}. These appearance-oriented representations can support high-quality novel-view synthesis, but rendering quality alone does not establish accurate metric geometry or collision-ready surfaces.

For dynamic scenes, the representation must also describe temporal change. Time-dependent radiance fields and Gaussian models can reconstruct or render scenes at different times. 4D Gaussian Splatting is one example that models dynamic scenes with Gaussian primitives and learned deformations~\citep{wu20244d}. When motion estimation or interaction prediction is required, the system must additionally maintain quantities such as object identity, pose or shape trajectories, and correspondences over time. A temporally plausible rendering does not by itself guarantee physically valid motion.

Learning-based reconstruction has reduced the need to optimize a separate model from scratch for every scene. For example, VGGT directly predicts camera parameters as well as geometric attributes such as depth maps, point maps, and point tracks from one or more images~\citep{wang2025vggt}. pixelSplat predicts a Gaussian radiance-field representation from image pairs~\citep{charatan2024pixelsplat}. These predictions can complement geometric reconstruction, but scale, coordinate-frame alignment, calibration, and task-specific accuracy still require attention. Geometry in regions not constrained by observations is inferred rather than measured.

World models can combine scene representation with prediction or generation, but these functions should be distinguished. A renderer that generates plausible views is not necessarily a model that predicts physically valid outcomes under robot actions. Atlas, for example, is described by its developers as producing point-cloud or Gaussian-splat outputs from images and generating novel views. Its developers also state that unseen regions may be inferred to fill gaps~\citep{worldlabs2026atlas}. A related but distinct capability is tool-assisted 3D authoring. OpenAI reports that GPT-6 Astra can generate computer-aided design (CAD) models from multi-view renders with tools and, in the reported demonstration, is used to create an editable house scene in Blender and a walk-through version in Unreal Engine~\citep{openai2026astra}.\footnote{Thomas Ricouard, OpenAI Developers Blog, \href{https://developers.openai.com/blog/architectural-visualization-with-astra}{\emph{Architectural Visualization with Astra}}, September 4, 2026.} These developer-reported examples demonstrate 3D generation and software-mediated modeling. Their use for robot interaction requires distinguishing generated content from sensor-supported estimates and validating geometry and motion against physical constraints.

\subsubsection{Model-Driven Spatial Perception and Reasoning}
Learned models can associate geometric representations with semantic concepts and use these associations to answer spatial queries or guide actions. The approaches below differ mainly in where spatial information is retained and how it is grounded in observations.

\paragraph{Models Using Explicit Spatial Representations} An explicit map can associate semantic features or object labels with points, surfaces, or regions in a common coordinate frame. OpenScene, for instance, co-embeds features of 3D scene points with image and text features, enabling open-vocabulary queries about observed space~\citep{peng2023openscene}. Such maps allow geometric grounding and repeated queries, although their coverage and semantic accuracy depend on observation quality, registration, and the underlying visual features.

\paragraph{Spatial Perception with Vision-Language Models} Vision-language models can infer object relations and answer spatial questions directly from images and text without necessarily maintaining a persistent 3D map. SpatialVLM shows that spatially grounded training can improve qualitative and quantitative spatial question answering~\citep{chen2024spatialvlm}. These outputs can help interpret instructions and propose likely relationships, but a linguistic answer should not be treated as a verified metric measurement or reachability judgment, particularly when viewpoint, occlusion, or unknown scale creates ambiguity.

\paragraph{Agent-Based Spatial Perception and Reasoning} An agent can couple perception to action. It maintains a spatial state, chooses where to look or move, and revises its estimates as new observations arrive. SayNav incrementally builds a 3D scene graph to ground language-model plans for navigation in unfamiliar environments~\citep{rajvanshi2024saynav}. VoxPoser grounds language instructions in 3D value maps and uses model-based planning to generate closed-loop manipulation trajectories~\citep{huang2023voxposer}. These examples illustrate how spatial representations and learned knowledge can inform action. Reliable operation still requires checking proposed plans against current geometry, robot configuration, collision constraints, and physical limits, and updating the estimate from feedback during execution.

\subsection{High-Level Perception and Cognition}

High-level perception organizes sensory observations and spatial representations into entities, attributes, states, and relations that are relevant to a task. Cognition builds on these representations by integrating prior knowledge, retaining relevant history, reasoning about spatial and physical relationships, and identifying what additional evidence is needed when current observations are insufficient.

For example, picking up a cup requires more than detecting and localizing it. The system may need to estimate its contents and physical state, identify an appropriate contact region, and determine whether the intended interaction is feasible. Similarly, monitoring an automated experiment requires associating a vessel with the correct sample, interpreting changes in material state, and relating visual observations to instrument measurements and experimental history. Both cases require representations that connect semantic understanding with physical evidence and persist beyond a single observation. This section develops a progression from object- and scene-level semantic perception to the understanding of physical properties, states, and interaction possibilities. It finally addresses grounded reasoning with vision-language models, memory, measurements, and active perception.

\subsubsection{Object Recognition and Scene Semantic Understanding}

Object- and scene-level semantic perception provides the foundation for high-level physical understanding. It identifies the entities present in an environment, localizes them in observations, maintains their identities over time, and organizes their semantic and spatial relationships. These representations provide the targets and contextual structure required for subsequent state estimation, interaction perception, and task-level reasoning.

\begin{figure}[h]
  \centering
  \includegraphics[width=\textwidth]{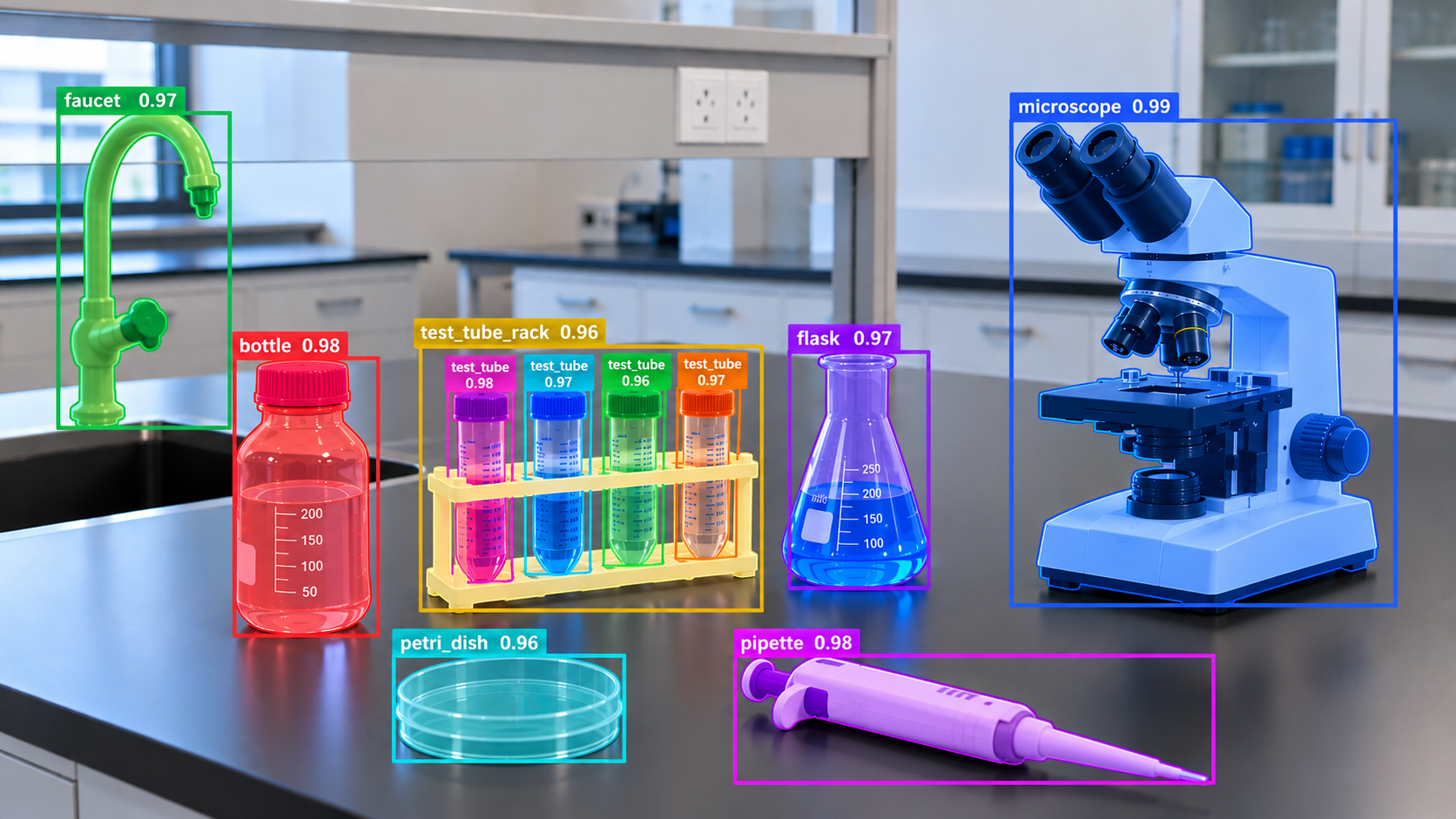}
  \caption{Illustration of object detection and instance segmentation. 
  Bounding boxes indicate the locations and categories of detected objects, 
  while colored masks delineate individual object instances at the pixel level}
  \label{fig:ch03-detection-segmentation}
\end{figure}

\paragraph{Object Detection, Segmentation, and Tracking} Object detection predicts object categories and locations, typically represented as bounding boxes. Semantic segmentation assigns category labels at the pixel level, whereas instance segmentation distinguishes individual object instances within each category, as illustrated in Fig.~\ref{fig:ch03-detection-segmentation}. Mask R-CNN jointly predicts object categories, bounding boxes, and instance masks~\citep{he2017mask}, while DETR formulates object detection as Transformer-based set prediction~\citep{carion2020end}. Tracking extends recognition across time by maintaining object identities using appearance, motion, and geometric information. ByteTrack, for example, incorporates lower-confidence detections during association to recover partially occluded objects~\citep{zhang2022bytetrack}. Together, detection, segmentation, and tracking provide object categories, spatial regions, and persistent identities that support target selection and state estimation. A cup temporarily occluded by a gripper, for example, should remain associated with the same contents, physical state, and interaction history after it becomes visible again.

Open-vocabulary recognition extends this capability beyond a predefined set of object categories. Grounding DINO localizes objects specified through textual descriptions~\citep{liu2024grounding}, while the Segment Anything Model (SAM) generates segmentation masks from spatial prompts such as points and bounding boxes ~\citep{kirillov2023segment}. Combining language-grounded detection with promptable segmentation connects task descriptions to observable image regions, allowing a system to identify and segment relevant targets without requiring an exhaustive task-specific label set.

\paragraph{Scene-Level Semantic Perception and Understanding} Recognizing individual objects is only the first step toward understanding a physical environment. Scene-level perception organizes entities according to their spatial, semantic, and functional relationships. Scene graphs, for example, represent objects or regions as nodes and relationships such as support, containment, proximity, or attachment as edges. These structured relationships provide context for interpreting both individual objects and the overall scene. For example, cups and a kettle located on the same countertop may jointly indicate a beverage-preparation area. In a laboratory, scene relationships may associate a sample with its vessel, a vessel with an instrument, and an observation with the experimental stage at which acquisition occurs. Such associations make isolated sensor observations interpretable within the context of an ongoing task or physical process.

Context can also help resolve ambiguous observations. An object that appears partially behind a drawer may be contained inside it or merely occluded by it, while a visible liquid region may belong to one of several overlapping transparent vessels. Scene representations should therefore retain not only predicted relationships but also the evidence and uncertainty associated with them. New observations can then confirm, reject, or revise previously inferred relationships.

These semantic representations answer questions about \emph{what} entities are present, \emph{where} they are, and \emph{how} they are related. Physical interaction, however, requires additional information about \emph{what properties they have}, \emph{what states they are in}, and \emph{what actions they afford}. These aspects are considered next.

\subsubsection{Object Properties, States, and Interaction Perception}

High-level physical perception extends object recognition beyond identifying \emph{what an object is} toward understanding \emph{what properties it has}, \emph{what state it is in}, and \emph{how it can be interacted with}. Such understanding requires connecting semantic observations with physical properties, functional structures, interaction possibilities, and feedback from the environment.

\paragraph{Observable Attributes and Latent Physical Properties} Objects exhibit directly observable attributes, such as color, texture, shape, and visible geometry, as well as physical properties that are only indirectly observable, including mass, friction, stiffness, compliance, and fragility. Visual appearance provides useful cues for these properties but is often insufficient for reliable estimation. Visually similar objects, for example, may differ substantially in mass because of their contents or internal structure.

Multimodal sensing and physical interaction provide complementary evidence for resolving such ambiguities. Deformation under load can reveal compliance, object motion during controlled contact can constrain friction estimates, and force measurements during lifting can provide evidence about mass. These estimates remain dependent on contact conditions, sensing quality, and dynamical assumptions, and should therefore retain uncertainty. Their primary role in physical perception is not necessarily to recover exact physical parameters, but to provide sufficient information for selecting appropriate contact regions, forces, and interaction strategies.

\paragraph{States, Functional Parts, and Affordances} Recognizing an object category alone does not determine how the object should be used or manipulated. Object function describes its typical role, while functional parts identify regions that support particular interactions. A cup, for example, may function as a container, while its handle provides a functional region for grasping.

Affordances further describe the actions that are possible for a particular agent under the current physical and task conditions. Unlike object categories or intrinsic properties, affordances are relational. They depend jointly on object geometry, physical state, environmental context, agent capabilities, and task requirements. A handle may be graspable by one gripper but inaccessible to another, and an otherwise graspable cup may be unsuitable for lifting when its contents are likely to spill. In an automated laboratory, relevant affordances may instead concern access to a vessel opening, available capacity, or whether a valve can be operated in its current state.

Interaction perception connects these semantic and physical descriptions to possible actions and state transitions. It estimates discrete conditions such as open, closed, grasped, or slipping, as well as continuous quantities such as deformation and contact force. It also associates functional parts with candidate interactions and evaluates whether an observed response is consistent with the intended action. Methods such as Where2Act demonstrate how actionable regions and interaction proposals can be inferred for articulated objects ~\citep{mo2021where2act}. Such proposals nevertheless remain candidates rather than executable actions and must be checked against reachability, collision, object state, and task constraints.

Physical interaction also reveals information that may be unavailable from external vision. During contact, relevant regions can become occluded, while incipient slip, local deformation, and contact loading may be difficult to infer visually. Tactile and force sensing provide complementary observations of these conditions. GelSight, for example, demonstrates how vision-based tactile sensing can capture local contact geometry and deformation ~\citep{yuan2017gelsight}. Observed slip, deformation, or unexpected resistance can then update the estimated interaction state and support adjustments to contact location, grasp force, or subsequent actions.

\paragraph{Material and Process-State Perception} High-level physical perception is not limited to robot-object contact. In autonomous laboratories and other process-oriented environments, the target of perception may instead be a material state or a process variable. Relevant observations include liquid levels, phase boundaries, precipitates, vessel contents, and changes occurring during an experiment. Distinguishing the container from its contents is particularly important because material regions may lie within, overlap with, or be partially occluded by the surrounding vessel. Reflections, refraction, foam, transparency, and occlusion can further make visible boundaries ambiguous.

Visual observations should therefore be connected to calibrated measurements and process context. Estimating a liquid level may require vessel geometry, camera calibration, or a visible scale, while determining material identity may require sample metadata or analytical measurements. Other quantities, including temperature, pressure, mass, and spectral measurements, may be obtained directly from instruments or extracted from their displays. Regardless of the acquisition mechanism, observations should retain their units, timestamps, sample associations, and measurement quality. Maintaining these relationships allows semantic interpretation to remain grounded in the physical process being observed.

\subsubsection{Grounded Perception and Cognition with Vision-Language Models}

Vision-language models (VLMs) provide a flexible semantic interface between visual observations, task instructions, and prior knowledge. In Physical AI, however, semantic interpretation alone is insufficient. Model predictions must ultimately be grounded in specific entities, physical states, spatial geometry, measurements, and interaction history. VLMs should therefore be viewed as one component of a broader perception and cognition system rather than as a replacement for geometric estimation, physical sensing, memory, or measurement.

\paragraph{Semantic, Property, and Affordance Understanding} VLMs can associate visual appearance with semantic concepts, material cues, object functions, and typical interactions. This capability is particularly useful when a system encounters unfamiliar objects or must interpret them in the context of a task. Given an image and an instruction to move a cup, for example, a ceramic appearance may suggest fragility and careful handling, while a visible handle may indicate a candidate interaction region. PhysObjects illustrates the use of physical-concept annotations to improve the understanding of properties relevant to manipulation~\citep{gao2024physically}.

Such predictions should be interpreted as hypotheses rather than direct physical measurements. A visual model may suggest that an object is heavy, fragile, or slippery, but quantitative or safety-critical judgments generally require additional sensing or interaction evidence. Semantic knowledge therefore provides useful priors that can guide subsequent observation and interaction without replacing physical verification.

Semantic knowledge must also be grounded in specific locations in the current environment. RoboPoint, for example, predicts image-space affordance points from visual observations and language instructions, linking semantic requests to candidate interaction regions~\citep{yuan2024robopoint}. These predictions help identify \emph{where} an interaction may occur but do not by themselves specify a feasible physical action. Depth, geometry, object state, collision constraints, and robot kinematics are still required to convert semantic interaction targets into executable contact poses and motions.

\paragraph{Reasoning with Geometry, Memory, and Measurements} High-level physical reasoning frequently requires information beyond the current image. Spatial questions may require calibrated geometry, questions about change require previous observations, and process-state interpretation may depend on event history or instrument measurements. A grounded cognition system therefore combines semantic reasoning with explicit geometric representations, persistent memory, and quantitative measurements.

Relevant information can be retrieved through object, sample, or task identity and associated with the current observation. Geometry provides spatial structure, memory provides temporal context, and sensors or instruments provide quantitative physical evidence. VLMs can integrate these sources to interpret a request, relate observations to the intended task, and identify missing or inconsistent evidence. Memory management, geometric computation, and numerical measurement remain explicit system functions even when their outputs are consumed by a multimodal model.

PaLM-E demonstrates one architecture for incorporating visual observations, continuous state estimates, and textual representations into an embodied language model~\citep{driess2023palme}. More generally, physical information may be provided to a reasoning model through structured records, tool outputs, or textual representations. Regardless of the interface, physical semantics should be preserved, including units, timestamps, coordinate frames, object identities, and measurement uncertainty.

Consider determining whether a vessel has reached a target fill level. Visual perception first identifies the vessel and candidate liquid boundary, while calibration or vessel geometry converts image evidence into a physical estimate. Memory provides the target level and relevant filling history, and complementary measurements can be consulted when the visual evidence is ambiguous. The VLM helps interpret the task and associate the available evidence with the correct sample, while geometric and measurement tools support the quantitative comparison. This division of roles prevents semantic plausibility from being mistaken for physical certainty.

\paragraph{Active Perception and Feedback-Based Updating} Physical cognition becomes particularly important when the available observations do not uniquely determine the current state. Rather than forcing a decision from incomplete evidence, a high-level reasoning module, potentially implemented with a VLM, can maintain competing hypotheses, identify unresolved variables, and propose additional observations or controlled interactions that reduce uncertainty.

The resulting process forms a perception-cognition loop. The system observes the environment, interprets the available evidence, identifies relevant uncertainty, acquires additional evidence, and updates its internal state. Additional evidence may come from a new camera viewpoint, a different sensing modality, an instrument measurement, or a controlled physical probe. The choice of observation should consider expected information gain, acquisition cost, execution risk, and potential disturbance to the physical process.

For example, when observing unfamiliar storage furniture, a change in viewpoint may reveal hinges or rails and help distinguish rotational from translational motion. If ambiguity remains, a gentle probe with measured motion and force can provide further evidence about the underlying mechanism. In a laboratory, an apparent liquid boundary may instead correspond to a reflection or foam edge. A different viewpoint or illumination condition can test the visual hypothesis, while a mass measurement may provide independent evidence. In both cases, active perception should target the specific uncertainty rather than simply acquire more observations.

Feedback from execution provides another source of evidence. A failed grasp, unexpected resistance, observed slip, or inconsistent measurement can revise the estimated object state, interaction hypothesis, or process interpretation. Perception and cognition are therefore not restricted to a single observation-decision sequence. They continuously update their representation of the physical environment as new evidence becomes available.

\paragraph{Limitations, Multiple Time Scales, and Outlook} High-level semantic reasoning remains vulnerable when model outputs become detached from geometry, measurement, or physical history. Plausible semantic predictions do not guarantee correct localization, physical feasibility, or accurate state estimation. Evaluation should therefore extend beyond language quality to include entity grounding, spatial relations, physical-property judgments, state estimation, affordance prediction, and the appropriate use of new evidence. Occlusion, viewpoint changes, ambiguous material boundaries, and conflicting instrument readings provide useful settings for testing whether a system can recognize uncertainty and request additional evidence rather than produce an unsupported conclusion.

Another challenge arises from the multiple time scales of physical perception. Contact and force measurements may change rapidly, visual observations arrive at another rate, and material, thermal, or experimental processes may evolve over much longer intervals. A shared physical state should therefore preserve acquisition time, validity, uncertainty, and evidence provenance. High-level semantic reasoning can query this state when needed without requiring every sensor update to pass through a language model. This separation enables high-frequency physical sensing and lower-frequency semantic reasoning to coexist within the same perception architecture.

Looking forward, a research direction is the development of persistent and grounded physical cognition that combines multimodal memory, 3D spatial reasoning, quantitative measurements, and active perception. Such systems should not only recognize objects and describe scenes, but also maintain up-to-date estimates of physical states, identify missing evidence, and select informative observations or interactions when uncertainty matters. For autonomous laboratories, this requires connecting sample identity, material observations, instrument measurements, and experimental history. For robotic manipulation, it requires linking semantic targets to geometry, physical properties, interaction states, and execution feedback.

Progress should ultimately be assessed by whether these capabilities improve the interpretation and control of physical systems. The assessment should determine whether a system can correctly identify relevant entities, maintain their evolving states, recognize when its evidence is insufficient, acquire information that resolves the ambiguity, and revise its interpretation under real feedback. From this perspective, VLMs provide a flexible semantic and reasoning interface within a broader physical perception and cognition system, while grounded sensing, geometry, memory, and interaction provide the evidence that keeps such reasoning connected to the physical world.

\clearpage 

\section{Data for Physical AI}
\label{ch:physical-data}

Perception and cognition are only passive to understand our physical world. Still, to build an active embodiment, interaction data in the real world is critical, as it provides the empirical foundation for AI systems to understand the actual physical dynamic environment and how to interact with it. Since this data records how actions alter physical states under known operating conditions, one can learn a control policy, calibrate a simulation pipeline, or build a validation benchmark. This chapter focuses on how to build a robust data collection pipeline for these purposes. It first introduces the data formats, curation pipelines, and real-world collection strategies spanning human teleoperation, autonomous exploration, and interactive feedback. It then examines how physics engines and generative world simulators expand data coverage, and how sim-to-real transfer aligns synthetic trajectories with physical hardware. Finally, it analyzes physical data scaling laws, the role of heterogeneous cross-embodiment datasets, and closed-loop data flywheels driven by deployment feedback.

\subsection{Data Foundations for Physical AI}
\label{sec:physical-data-foundations}

Physical data link observations, actions, and their consequences under specified operating conditions. A drawer image may support recognition. Learning to open it also requires the applied commands, contact events, and resulting motion. The same distinction applies to industrial processes, traffic control, and automated experiments. This section defines the evidence and record structure needed to interpret such data. Subsequent sections address its collection, generation, alignment, and growth.

\subsubsection{The Spectrum of Physical Data}

Data sources differ in how directly they support physical execution. The \emph{Data Pyramid for Embodied Manipulation} organizes real-robot data, Universal Manipulation Interface (UMI)-style demonstrations, egocentric and exocentric observations, simulation data, and general vision-language data around the tension between scalability and robot alignment~\citep{ye2026pyramid}. Here, the complementary question is what each record actually establishes. Video can reveal an object's motion without identifying the force or command that caused it. A simulated trajectory can include exact action labels without establishing the corresponding real-world response.

Table~\ref{tab:physical-data-evidence} distinguishes these forms of evidence. They can coexist within one dataset and do not constitute a ranking of quality. In particular, reconstructed poses and latent actions are \emph{action proxies}. Converting them into executable supervision requires a mapping to the target embodiment and controller. Representative released datasets spanning robot execution, portable demonstrations, and human observation are cataloged in Table~\ref{tab:representative-physical-datasets} in Section~\ref{sec:physical-data-collection}.

\begin{table}[tbp]
    \centering
    \caption{Evidence levels in physical data and their limits}
    \label{tab:physical-data-evidence}
    \small
    \renewcommand{\arraystretch}{1.15}
    \setlength{\tabcolsep}{4pt}
    \begin{tabularx}{\textwidth}{
        @{}
        >{\hsize=0.7\hsize\linewidth=\hsize\raggedright\arraybackslash}X
        >{\hsize=1.15\hsize\linewidth=\hsize\raggedright\arraybackslash}X
        >{\hsize=1.15\hsize\linewidth=\hsize\raggedright\arraybackslash}X
        @{}
    }
        \toprule
        \textbf{Evidence}
        & \textbf{Provides}
        & \textbf{Does not establish} \\
        \midrule

        Observation
        & Appearance, scene context, visible motion
        & Executable commands or contact forces \\
        \addlinespace

        Action proxies
        & Reconstructed poses, trajectories, or latent actions
        & Feasibility on the target embodiment \\
        \addlinespace

        Execution
        & Recorded commands and measured responses
        & Coverage of unfamiliar states or failures \\
        \addlinespace

        Closed-loop experience
        & Policy-induced states, interventions, and outcomes
        & Correct behavior or reliable annotations \\
        \addlinespace

        Synthetic evidence
        & Controlled variation, scalable trajectories, dense model-derived labels
        & Real-world dynamics or sensor behavior \\

        \bottomrule
    \end{tabularx}
\end{table}

For interaction data, let $s_t$ denote the underlying physical state, $o_t$ the observation, and $a_t$ the action. A useful conceptual record is
\begin{equation}
\tau_{\mathrm{ep}} = \left(
    c,g,
    \left\{
        \left(
            o_t,\zeta_t^{\mathrm{state}},
            a_t^{\mathrm{cmd}},a_t^{\mathrm{exec}},
            e_{t}^{\mathrm{event}},\Delta t_t
        \right)
    \right\}_{t=0}^{T},y
\right),
\end{equation}
where $c$ records the embodiment, environment, and collection conditions, and $g$ is the goal. The optional field $\zeta_t^{\mathrm{state}}$ is a state estimate or belief, while $e_{t}^{\mathrm{event}}$ records events such as contact or intervention. The field $\Delta t_t$ gives elapsed time, and $y$ records the outcome. The command $a_t^{\mathrm{cmd}}$ is distinguished from $a_t^{\mathrm{exec}}$, which denotes the measured execution response rather than a second command. The available fields vary by source. Missing measurements require validity masks, and inferred values require provenance.

\subsubsection{Unified Representation}
\label{subsec:physical-data-representation}

A common schema must preserve the meaning of its fields. In the RT-X experiments, Open X-Embodiment converts actions into a common seven-dimensional end-effector representation, but retains differences in coordinate frames and in absolute, relative, or velocity control. The same vector can therefore induce different motions on different robots~\citep{openx2024}. Matching dimensions enables shared processing but does not establish physical equivalence.

Each record should retain the embodiment, tool geometry, control interface, and calibration needed to interpret it. Coordinate transforms must specify their direction and reference frames. Sensor acquisition, command issue, and physical response times must also remain distinguishable. Pairing an image with a command issued before its exposure can change the supervision target. Sampling rate, action duration, and controller update rate should be recorded separately.

Measured, estimated, and generated quantities require distinct provenance. A reconstructed wrist pose is an estimate, and a simulator's contact force is a model output. Neither should silently replace a physical measurement. Derived fields should identify their source records, generating method and version, and any available uncertainty or validity bounds. A shared coordinate frame resolves geometric ambiguity but leaves embodiment-dependent limits and dynamics to be accounted for.

\paragraph{Minimum Action Specification} An action field should identify its issuer and target, control mode, reference frame and tool point, units, duration and effective time, applicable limits or saturation, controller gains where relevant, and the associated execution feedback.

\subsubsection{Data Organization}

Physical records are organized into steps, transitions, segments, and episodes. These boundaries carry learning semantics. Reinforcement Learning Datasets (RLDS), for example, distinguishes the last recorded step from an environment terminal state. A truncated episode does not establish that the task terminated, and its final observation need not have a valid outgoing action. Training views must preserve these distinctions instead of treating every stored row as a complete transition.

Raw records, standardized records, and training views serve different purposes. Raw records preserve the original signals and metadata. Standardized records make units, timestamps, coordinates, and episode boundaries explicit. Training views specify selection, segmentation, annotation, and augmentation. Keeping these stages traceable allows a calibration or labeling error to be corrected without repeating collection.

Every episode should retain a stable identifier and links to its source, embodiment, operator or policy version, calibration, and processing history. Segments and generated variants should retain their parent identifiers. Dataset splits can then keep related records together and hold out the task, scene, object, or embodiment relevant to the generalization claim. Splitting adjacent frames at random cannot establish transfer to unseen physical conditions.

\subsubsection{Annotation Framework and Workflow}

A recording may show an object moving, a temperature rising, or a machine stopping. The signals are recorded, but what happened? Annotations identify the objects and materials involved, the operating conditions, the actions taken, and their consequences. They may mark an object in an image, contact during a motion, stages of a production process, or the outcome of an experiment. The same record can then support different learning tasks, including recognizing a material, predicting its response, choosing an action, or judging whether the outcome meets a goal.

Annotation starts with rules that different people and tools can apply consistently. A small trial reveals unclear definitions and disputed boundaries before labeling expands. Automated tools can propose labels, while reviewers examine uncertain cases and spot-check the rest against the recordings. Each label keeps a link to its evidence and records whether it comes from measurement, human judgment, inference, or simulation. Uncertainty remains visible, and corrections update the labels without rewriting the underlying evidence.

\subsubsection{Quality Governance}

Quality checks should distinguish recording defects from difficult but valid experience. A force spike may indicate contact. A stalled gripper may reveal an execution failure. Neither warrants deletion solely because the signal is irregular. Checks should cover signal integrity, consistency across timestamps and coordinate frames, correspondence between commands and responses, and agreement between event annotations and outcomes.

Learning utility also depends on the learner and the intended supervision. The robomimic study systematically varies demonstration quality, observations, and learning algorithms, showing that these choices materially affect policy performance~\citep{mandlekar2022matters}. A valid failure trajectory can support outcome prediction while providing unsuitable actions for nominal behavior cloning. Data validity, task outcome, and suitability as an imitation target should therefore be recorded separately.

Dataset documentation makes these judgments inspectable. Following \emph{Datasheets for Datasets}, a release should describe its composition, collection process, intended uses, and limitations~\citep{pi12_gebru2021}. For physical data, this should include embodiment and control specifications, calibration history, missing modalities, outcome definitions, and known recording defects. A versioned manifest should link released samples to their processing code and access or reuse conditions. These records provide the basis for the collection procedures in Section~\ref{sec:physical-data-collection} and the synthetic-data validation developed in Section~\ref{sec:physics-engines-generative-worlds} and Section~\ref{sec:sim-to-real-data-alignment}.

\subsection{Data Collection and Interaction}
\label{sec:physical-data-collection}

Data collection and interaction proceed through collection design, human demonstrations, autonomous collection, and interactive feedback in four linked steps. In this view, producing data means actively sampling the physical world. The emphasis is on distributions, supervision, efficiency, and safety. Implementation details of individual devices are addressed only where they constrain the evidence that can be collected.

\subsubsection{Collection Design}

Collecting physical data is not indiscriminate recording of real-world processes. It is an active sampling of the space of reachable states, possible actions, and actual outcomes. Initial states, task goals, acting agents, control constraints, disturbances, and termination conditions all shape the resulting distribution. The collection policy is therefore part of the data-generating mechanism. A sound design first identifies the capability to be developed or the hypothesis to be tested, and then determines what to collect.

Design typically starts with the intended deployment distribution and works backward to define task families. A coverage matrix can span objects, scenes, initial states, goals, disturbances, embodiments, operators or policies, outcomes, and risks. Frequent states require sufficient density to establish basic competence. Rare but consequential states require targeted collection. Matched comparison cases can weaken spurious correlations with backgrounds, positions, or operator styles. The number of task names is not a measure of coverage. ``Tidy the desk'' can probe very different capabilities as object combinations, occlusion, starting poses, and time pressure change.

The variable specification should distinguish controlled factors, observable quantities, and potential confounders. In robotic manipulation, one can vary object positions and materials, container states, illumination, and disturbances. Traffic and industrial systems also require records of topology, load, shifts, operating rules, and decisions by multiple actors. Intelligent laboratory systems should retain reagent batches, instrument states, step dependencies, and measurement uncertainty. These conditions must be represented in task templates and metadata if performance differences are to be attributed to plausible causes rather than described merely as isolated successes or failures.

Sensor configuration further bounds what can be observed. Measurement principles, mounting locations, and coverage determine which quantities can be measured directly. Range, noise, and calibration determine how trustworthy those measurements are. An uninstalled or uncovered channel should be marked as missing, rather than represented by a zero reading. Contact forces inferred from images or a dynamics model must likewise be distinguished from directly measured forces. Temporal resolution depends jointly on sensor bandwidth, sampling rate, filtering, and the transmission path. A higher logging rate does not necessarily provide more usable information.

Force and tactile sensing illustrate these constraints. A six-axis wrist force/torque sensor measures the net force and moment, not the distribution of fingertip contacts. A tactile array covers only its sensitive surface. Some force sensors support sampling at kilohertz rates~\citep{atiAxiaEtherCAT}, whereas vision-based tactile sensors are limited by camera frame rate. The DIGIT prototype, for example, operates at 60~frames/s~\citep{lambeta2020digit}. Directly reducing a high-rate force stream to the video frame rate can hide collision peaks and transient slip. The raw time series should therefore be retained, with a separately aligned stream generated after antialiasing. Collection also requires calibration and zeroing, together with compensation of wrist readings for tool weight and inertial effects. Quality checks are also required for thermal drift, hysteresis, saturation, and dropped frames. Calibration versions, filter settings, and anomaly flags should accompany the data. The confidence assigned to contact labels should account for both synchronization error and measurement uncertainty.

Success criteria, failure categories, safety limits, and reset rules must be defined before collection. Success should be expressed in verifiable object states, process conditions, or operational metrics rather than left to an operator's impression. Failures should distinguish perception errors, planning errors, control deviations, unstable contact, timeouts, safety aborts, and external disturbances. An incomplete reset, or physical traces left by a previous trial, can also compromise the independence of subsequent data. When an outcome is difficult to determine automatically, sensor-based rules can be combined with visual checks and human review in a staged assessment.

Temporal segmentation is likewise part of collection design. An episode should start at a verifiable initialization or reset and end in success, failure, timeout, or an abort. Skill segments and action chunks should have explicit temporal meanings. The acquisition system must distinguish the rates of sensor sampling, model decisions, action updates, and low-level servo control. It should record the issue, delivery, and actual effective time of each command. Source timestamps, coordinate-frame calibration, latency measurement, and completeness monitoring are minimum requirements. Detailed hardware and software implementation belongs in Chapter~\ref{ch:hardware-systems}.

Recent large-scale real-world datasets illustrate different organizational approaches. DROID shows how relatively uniform hardware and protocols can support distributed collection across sites. RoboMIND brings multiple embodiments, task protocols, and demonstrations of failure with recorded causes into one platform. AgiBot World scales trajectories through centralized facilities, standard procedures, and human-in-the-loop verification. Humanoid Everyday further includes whole-body locomotion, manipulation, and human-robot interaction~\citep{khazatsky2024droid,wu2025robomind,agibot2025colosseo,zhao2025humanoideveryday}. Together, these efforts show that scaling requires reproducible task definitions, staffing procedures, equipment states, and quality judgments, as well as more devices.

Collection performance should be measured by \emph{effective data yield}. Beyond raw hours and trajectory counts, useful measures include minutes of meaningful interaction, the share of recordings that pass training quality checks, incremental coverage after deduplication, the yield of rare events, equipment availability, human and machine time, time spent resetting, and the final retention rate from raw records to the training release. Consecutive stationary frames, repeated successes, and multiple derived versions of the same behavior inflate volume without necessarily adding new experience. Efficient collection uses limited physical interaction to close identifiable capability gaps.

The output of collection design is thus an executable data-collection protocol, not merely a task list. It specifies the target distribution, variable combinations, episode boundaries, success and failure definitions, safety constraints, data schema, quality gates, and cost metrics. Human demonstrations, autonomous execution, and interactive corrections are different ways of gathering evidence under this plan.

\subsubsection{Human Demonstrations}
\label{subsec:human-demonstrations}

\begin{figure}[h]
  \centering
  \includegraphics[width=\textwidth]{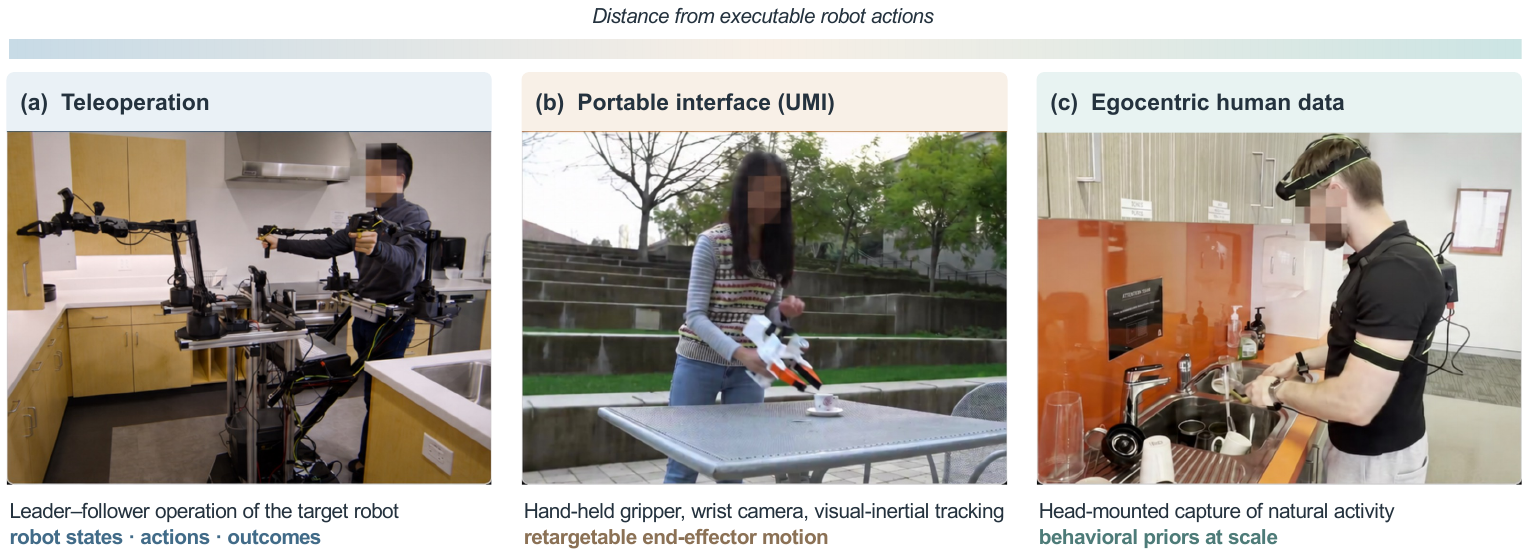}
  \caption{Human-demonstration data collection paradigms, arranged by their
  \emph{distance from executable robot actions}.
  (a)~Teleoperation drives the target robot through leader-follower
  mechanisms and records robot observations, actions, and outcomes directly
  (Mobile ALOHA, photo from Stanford News).
  (b)~Portable interfaces such as UMI record retargetable end-effector motion
  in natural environments using a hand-held gripper, a wrist-mounted camera,
  and visual-inertial tracking (photo from the UMI project).
  (c)~Egocentric human data capture natural activity at scale with
  head-mounted devices, providing broad behavioral priors without native
  robot states (photo from ABC News)}
  \label{fig:human-demo-collection}
\end{figure}

Human demonstration converts human intent into behavioral evidence that a machine can learn from and, ideally, execute on the target embodiment. Methods can be arranged by their distance from executable robot actions (Figure~\ref{fig:human-demo-collection}). Teleoperation and kinesthetic teaching with the target robot in the loop directly produce embodied actions. Portable tools, motion capture, and wearable systems record motions that can be retargeted. Egocentric and Internet videos provide broad observations and behavioral priors. These sources complement one another in action fidelity, coverage of natural settings, equipment occupancy, and processing cost.

Master-slave robot arms, spatial controllers, and immersive teleoperation operating on the target embodiment can record robot observations, control commands, embodiment states, contacts, and outcomes together. ALOHA uses matched leader and follower mechanisms for fine-grained bimanual demonstrations~\citep{zhao2023}. GELLO provides a low-cost, kinematically matched interface that helps an operator perceive joint limits~\citep{wu2023gello}. Open-TeleVision incorporates stereoscopic views and active head motion into the teleoperation loop~\citep{cheng2024tv}. The collection interface is therefore not a neutral accessory. Its mapping, feedback delay, viewpoint strategy, and operator burden systematically shape trajectories.

Portable, robot-independent interfaces bring demonstrations into natural environments. UMI records relative end-effector motion with a handheld gripper, wrist-mounted cameras, and visual-inertial tracking, then accounts for target-robot reachability and latency at deployment~\citep{chi2024umi_ch6}. DexWild uses a portable, human-centered system to extend dexterous demonstrations in everyday settings~\citep{tao2025dexwild}. FreeTacMan synchronizes visual and tactile collection, so that contact feedback can support both the operator and supervision of contact phases~\citep{wu2025freetacman}. These methods reduce occupancy of the target robot and expand scene coverage, but still have to address tracking errors, scale, action retargeting, and the target embodiment's dynamics.

Humanoid and mobile manipulation extend demonstration beyond tabletop end-effector motion to whole-body coordination. TWIST2 uses portable full-body teleoperation without external motion capture to bring legged locomotion, torso, both arms, hands, and head-mounted views into one loop~\citep{ze2025twist2}. The central requirement is not simply more degrees of freedom. Navigation, support, manipulation, and active vision must remain temporally consistent. A recording should also distinguish actions supplied by the operator from compensation supplied by a stabilizer or whole-body controller. Otherwise, high-level intent and low-level control corrections become conflated.

Egocentric human data offer natural behavior at greater scale. EgoDex synchronizes egocentric images with three-dimensional head, upper-body, and hand poses, turning video into a more calibratable behavioral record~\citep{hoque2025egodex}. AoE explores sustained collection of natural egocentric experience and joint training with a smaller amount of robot teleoperation data~\citep{yang2026aoe}. ActiveMimic jointly recovers camera and wrist trajectories from single-camera in-the-wild egocentric video, emphasizing the relationship between active viewpoints and manipulation motion~\citep{lin2026activemimic}. These approaches broaden the range of objects, scenes, behavioral phases, and viewpoint priors. Human-video records themselves, however, do not natively supply robot joint states, gripper signals, tactile measurements, or feedback from actual robot execution.

Turning human or Internet videos into robot supervision requires alignment of camera frames, recovery of scale, mapping from hands or bodies to tools, correction for kinematic reachability, temporal alignment, contact assumptions, and confidence assessment. Point tracks, wrist poses, body poses, and latent actions are action proxies with different levels of reliability. They should not all be labeled as ground-truth robot actions. Table~\ref{tab:data-interaction-supervision} compares the supervision, executability, and limitations of interaction data sources. More defensible training strategies include learning visual and temporal representations first and then post-training on robot data, using separate loss heads for human and robot data, weighting examples by confidence, or restricting explicit action supervision to reliable channels.

\begin{table}[htbp]
    \centering
    \caption{Supervisory value and principal limitations of interaction data sources}
    \label{tab:data-interaction-supervision}
    \small
    \renewcommand{\arraystretch}{1.15}
    \setlength{\tabcolsep}{4pt}
    \begin{tabularx}{\textwidth}{
        @{}
        >{\hsize=0.85\hsize\linewidth=\hsize\raggedright\arraybackslash}X
        >{\hsize=1.05\hsize\linewidth=\hsize\raggedright\arraybackslash}X
        >{\hsize=0.75\hsize\linewidth=\hsize\raggedright\arraybackslash}X
        >{\hsize=1.35\hsize\linewidth=\hsize\raggedright\arraybackslash}X
        @{}
    }
        \toprule
        \textbf{Source} & \textbf{Main supervision} & \textbf{Executability} & \textbf{Main bias and suitable use} \\
        \midrule
        Target-embodiment teleoperation & Robot observations, commands, executed motions, contacts, and outcomes & High & Affected by interfaces, delays, and operator styles. Suitable for behavior cloning and skill fine-tuning \\
        \addlinespace
        Portable human demonstration & Camera, tool, and body trajectories in natural scenes & Medium & Requires calibration, retargeting, and reachability filtering. Expands object and contact coverage \\
        \addlinespace
        Egocentric or public video & Visual content, semantics, task steps, and behavioral timing & Low & Lacks native actions and physical feedback. Supports representations, latent actions, and high-level priors \\
        \addlinespace
        Autonomous policy rollout & Policy-induced states, actions, events, and outcomes & High & May repeat behavior or amplify policy bias. Supports learning from boundaries, failures, and online experience \\
        \addlinespace
        Human intervention and recovery & Warning signs, takeovers, corrections, recovery, and consequences & High & Selected by intervention triggers. Supports correction, recovery, and safety-oriented post-training \\
        \bottomrule
    \end{tabularx}
\end{table}

Demonstration quality also depends on how people are organized. A collection plan should retain anonymized operator identifiers, levels of training, interface configurations, and work periods. Task rotation, fatigue monitoring, spot checks of replays, and coverage across operators can reduce style bias. Expert demonstrations may not visit the error states encountered by novices, while natural human movements may violate a robot's safety constraints. The value of a demonstration lies in how clearly it conveys intent, constraints, and an executable path.

Human data thus form a gradient from rich semantics to reliable executable actions. A sound combination uses observational data to broaden tasks and scenes, wearable and tool-based data to reduce the gap in action semantics, and interactions on the target embodiment to calibrate execution. Whether cross-domain data actually help must ultimately be established through closed-loop evaluation on robots or other real systems. {Section~\ref{sec:c9:dex-data} follows these collection choices into a dexterous-hand system, where wrist tracking, finger capture, and contact sensing supply different parts of the demonstration.}

Table~\ref{tab:representative-physical-datasets} makes this spectrum concrete through representative publicly available datasets. Its focus is real-world manipulation and human activity relevant to the collection methods above. Reading the scale together with the supervision distinguishes robot execution records from tool-mediated demonstrations and human behavioral observations. The official resource links provide entry points for inspecting each release.

\begingroup
\small
\renewcommand{\arraystretch}{1.13}
\setlength{\tabcolsep}{4pt}
\setlength{\LTleft}{0pt}
\setlength{\LTright}{0pt}
\setlength{\LTcapwidth}{\textwidth}
\begin{longtable}{@{}
    >{\raggedright\arraybackslash}p{\dimexpr0.19\textwidth-6pt\relax}
    >{\raggedright\arraybackslash}p{\dimexpr0.19\textwidth-6pt\relax}
    >{\raggedright\arraybackslash}p{\dimexpr0.47\textwidth-6pt\relax}
    >{\raggedright\arraybackslash}p{\dimexpr0.15\textwidth-6pt\relax}
@{}}
\caption{Representative real-world datasets for Physical AI. Each entry identifies a release or collection, its reported scale, the available supervision, and official access points}\label{tab:representative-physical-datasets}\\
\toprule
\textbf{Dataset / release} & \textbf{Scale and coverage} & \textbf{Overview and supervision} & \textbf{Official resources} \\
\midrule
\endfirsthead
\caption[]{Representative real-world datasets for Physical AI (continued)}\\
\toprule
\textbf{Dataset / release} & \textbf{Scale and coverage} & \textbf{Overview and supervision} & \textbf{Official resources} \\
\midrule
\endhead
\midrule
\multicolumn{4}{r@{}}{\footnotesize Continued on the next page.}\\
\endfoot
\bottomrule
\endlastfoot

\multicolumn{4}{@{}l}{\textbf{A. Real-robot interaction datasets}}\\*
\addlinespace[3pt]
Open X-Embodiment (OXE)
& $>$1,000,000 trajectories, 22 embodiments (reported collection)
& Aggregates heterogeneous robot datasets in a common format for learning across embodiments.\par\emph{Signals:} images, robot states/actions, and language where available. Modalities vary by source.
& \href{https://robotics-transformer-x.github.io/}{Project}\par\href{https://github.com/google-deepmind/open_x_embodiment}{Data / code}\par\emph{Source-specific terms} \\
\addlinespace[5pt]

BridgeData V2
& 60,096 trajectories, 24 environments, 13 skills
& Combines teleoperated and scripted WidowX manipulation across varied household-style settings.\par\emph{Signals:} RGB, robot states/actions, and language, with depth and additional views in subsets.
& \href{https://rail-berkeley.github.io/bridgedata/}{Project / code}\par\href{https://rail.eecs.berkeley.edu/datasets/bridge_release/data/}{Data} \\
\addlinespace[5pt]

DROID
& 76,000 trajectories, 350~h, 564 scenes
& Uses a shared Franka platform for distributed collection of manipulation in varied real environments.\par\emph{Signals:} wrist/external stereo video, robot states/actions, calibration, and language.
& \href{https://droid-dataset.github.io/}{Project / code}\par\href{https://droid-dataset.github.io/droid/the-droid-dataset}{Data} \\
\addlinespace[5pt]

RH20T
& $>$110,000 sequences, 147 tasks
& Pairs contact-rich robot manipulation with human demonstrations to study multimodal skill transfer.\par\emph{Signals:} RGB-D, robot states/actions, force/torque, audio, and human video, with tactile data in one configuration.
& \href{https://rh20t.github.io/}{Project / data / code} \\
\addlinespace[5pt]

RoboMIND\par Original release
& 107,000 trajectories, 4 embodiments, 479 tasks
& Applies a unified teleoperation protocol across single-arm, dual-arm, and humanoid platforms.\par\emph{Signals:} multi-view images, robot states/actions, language, and failure demonstrations with causes.
& \href{https://x-humanoid-robomind.github.io/}{Project / code}\par\href{https://huggingface.co/datasets/x-humanoid-robomind/RoboMIND}{Data} \\
\addlinespace[5pt]

AgiBot World\par Beta (2025)
& 1,003,672 trajectories, 100 robots
& Scales bimanual manipulation through coordinated collection across replicated real-world scenarios.\par\emph{Signals:} multi-camera video, robot states and issued actions, task descriptions, and segment annotations.
& \href{https://github.com/OpenDriveLab/AgiBot-World}{Project / code}\par\href{https://huggingface.co/datasets/agibot-world/AgiBotWorld-Beta}{Data} \\
\addlinespace[7pt]

\multicolumn{4}{@{}l}{\textbf{B. UMI-style tool demonstrations and wearable hand capture}}\\*
\addlinespace[3pt]
UMI\par In-the-wild cup collection
& 1,400 demonstrations, 30 locations
& Captures cup arrangement with a handheld gripper for transfer to compatible robot end effectors.\par\emph{Signals:} wrist RGB, visual-inertial tool trajectories, and gripper width, provided in a processed Zarr release.
& \href{https://umi-gripper.github.io/}{Project}\par\href{https://github.com/real-stanford/universal_manipulation_interface}{Data / code} \\
\addlinespace[5pt]

FastUMI-100K
& $>$100,000 trajectories, 54 tasks
& Extends UMI-style collection to a large set of single-arm and bimanual household demonstrations.\par\emph{Signals:} wrist fisheye RGB, relative tool poses, gripper states, and text, provided in a LeRobot-format release.
& \href{https://github.com/MrKeee/FastUMI-100K}{Project / code}\par\href{https://huggingface.co/datasets/IPEC-COMMUNITY/FastUMI_100k_lerobot}{Data} \\
\addlinespace[5pt]

DexWild\par Released collection
& 9,505 episodes (human + robot), 93 environments
& Combines portable human-hand demonstrations with robot data for dexterous policy transfer.\par\emph{Signals:} hand-mounted images, tracked wrist motion, and hand articulation, with human and robot subsets.
& \href{https://dexwild.github.io/}{Project / code}\par\href{https://huggingface.co/datasets/boardd/dexwild-dataset}{Data} \\
\addlinespace[7pt]

\multicolumn{4}{@{}l}{\textbf{C. Human egocentric and synchronized ego-exo datasets}}\\*
\addlinespace[3pt]
EPIC-KITCHENS-100
& 100~h, 45 kitchens, approximately 90,000 action segments
& Records unscripted kitchen activities for learning object-centric actions and their temporal structure.\par\emph{Signals:} first-person video/audio, narrations, and verb-noun action segments.
& \href{https://epic-kitchens.github.io/2024}{Project / data / code} \\
\addlinespace[5pt]

Ego4D\par Original collection
& 3,670~h, 74 locations, 9 countries
& Covers diverse daily activities to support first-person perception, episodic memory, and anticipation.\par\emph{Signals:} egocentric video, narrations, and benchmark annotations, with extra sensors in subsets.
& \href{https://ego4d-data.org/}{Project}\par\href{https://ego4d-data.org/docs/start-here/}{Data access}\par\emph{Agreement required} \\
\addlinespace[5pt]

Ego-Exo4D\par V2 release
& 5,035 takes, 1,286.30 video~h, including 221.26 ego~h
& Synchronizes first- and third-person views of skilled activities for linking observed motion and behavior.\par\emph{Signals:} ego/exo video, audio, gaze, camera trajectories, and task-specific body/hand pose annotations.
& \href{https://docs.ego-exo4d-data.org/}{Project / docs}\par\href{https://docs.ego-exo4d-data.org/getting-started/}{Data access}\par\emph{Agreement required} \\
\addlinespace[5pt]

EgoDex
& 829~h, 194 tabletop tasks
& Couples egocentric manipulation video with tracked human motion for learning dexterous behavior.\par\emph{Signals:} RGB, 3D head/upper-body/hand poses, and language captured using Apple Vision Pro.
& \href{https://github.com/apple-aiml-research/ml-egodex}{Project / data / code} \\
\addlinespace[5pt]

HOT3D
& 833~min, 19 participants, 33 rigid objects
& Provides calibrated egocentric hand-object interactions for evaluating 3D tracking and pose estimation.\par\emph{Signals:} synchronized head-mounted views, motion-capture hand/object poses, and scanned object models.
& \href{https://facebookresearch.github.io/hot3d/}{Project / data / code} \\

\end{longtable}
\noindent\begin{minipage}{\textwidth}
\footnotesize
\textit{Reading the Table} Scales follow the named releases and official documentation checked on October 9, 2026. Trajectories, video hours, and synchronized takes are different units. OXE aggregates other datasets, so row totals are not additive. BridgeData V2 includes 50,365 teleoperated trajectories and 9,731 scripted rollouts. DexWild uses the released data-card count, which differs from the paper's 9,290 human demonstrations. Tool and human poses are action proxies requiring embodiment alignment, not native robot commands. Resource links lead to official data, code, or download instructions. Data-access and reuse terms remain specific to each release.
\end{minipage}
\endgroup

\subsubsection{Autonomous Data Collection}

Autonomous data collection should be understood as an agent actively sampling the space of possible experiences, rather than a robot repeating its current policy indefinitely. Passive production logs reflect the natural frequency of events but are often dominated by routine states. An active system can instead select tasks and initial conditions according to novelty, uncertainty, prediction error, failure density, gaps in coverage, or risk priority. If a collection policy repeatedly succeeds only in familiar regions, the sample count may grow while behavioral diversity and the boundaries of competence remain unchanged.

AutoRT illustrates an early use of foundation models to orchestrate collection. A vision-language model describes the scene, a language model proposes and filters tasks, and the system schedules their execution using existing policies or human operators~\citep{ahn2024autort}. Its contribution in this context is not to replace physical control with a language model, but to bring scene understanding, task discovery, safety rules, and available policy resources into a common orchestration process. A high-level model may likewise identify gaps in coverage for factories, traffic systems, mines, or laboratories, while concrete actions remain subject to control authority, real-time constraints, and safety interlocks.

A systematic study of scaling autonomous robot data collection finds that environment instrumentation, reliable object resets, success detection, and maintaining stable conditions remain major costs. In some tasks, adding a small number of human demonstrations is more effective than adding many autonomous episodes~\citep{mirchandani2025autonomous}. These findings show how autonomous collection shifts some labor from episode-by-episode teleoperation to environment interfaces, online assessment, and risk control. If that infrastructure is unreliable, scaling up may merely accumulate repetitive, mislabeled, or dangerous data.

The distinctive value of autonomous interaction is its access to states rarely covered by successful human demonstrations. PlayWorld uses autonomous play to collect contact-rich and long-tail robot-object interactions for learning action-conditioned changes in the world~\citep{yin2026playworld}. The lesson for the present section concerns data production. Open-ended play can complement demonstrations with failures, slips, precursors to collisions, and nonexpert trajectories. The use of such data to train generative world models is discussed in Section~\ref{sec:physics-engines-generative-worlds} and Chapter~\ref{ch:wm}.

Effective active sampling must balance information gain with representativeness. A collection budget can be divided into three parts. One maintains a baseline density for states common in deployment, another targets regions where the current model is uncertain or incurs high loss, and the third explores rare risks and new tasks. A task selector should also suppress near-duplicate trajectories and avoid favoring short tasks merely because success is easy to detect. For high-risk states, constrained environments, low-speed operation, safe proxy tasks, or simulation-based prescreening should take priority over deliberately inducing accidents.

A complete autonomous episode should retain the policy version, model inputs, raw outputs, commands after safety filtering, actual execution, random seed, initial state, reset method, termination reason, success-detector version, and any human involvement. In particular, a policy-initiated abort, a safety-system intervention, a human emergency stop, and an equipment failure must be distinguished because they carry different training semantics. Continuous operation also requires monitoring for clock drift, frozen sensors, saturated actions, repeated tasks, degradation of the scene, and failed resets, so that the pipeline does not continue blindly accumulating data after the physical system has ceased to behave as intended.

Multiple robots can collect data in parallel, but the number of concurrent robots is not the only measure of efficiency. As fleets grow, task scheduling, spare parts, networks, storage, calibration, and on-site supervision become more complex. Drift and maintenance history across devices may also introduce hidden domain differences. A more mature system coordinates equipment health, task queues, risk levels, gains in coverage, and human workload, and admits each new data batch into training only after it passes a quality gate rather than adding every operating log directly to the model.

The scope of autonomous collection should be clear. This section concerns the selection and execution of interactions in the real world. Scaling simulated agents and generative scenes belongs to Section~\ref{sec:physics-engines-generative-worlds}, the simulation-reality gap to Section~\ref{sec:sim-to-real-data-alignment}, and the longer-term deployment and retraining flywheel to Section~\ref{sec:physical-data-scaling-flywheel}. Autonomy is, first, a means of collecting informative experience and, second, a means of reducing the human cost of each individual data point.

\subsubsection{The Interactive Feedback Loop}
\label{subsec:interactive-data-feedback}

The most valuable supervision often lies at the boundary of a policy's competence. Complete, successful demonstrations describe a manifold of nominal behavior, whereas autonomous deployment exposes the distribution of states induced by the current policy itself. Interactive supervision connects the two. A failed episode, the warning signs before failure, the instant of human takeover, corrective actions after takeover, and the recovery segment that returns the system to a feasible state have different learning semantics. They should not be collapsed into a single ``failure'' label.

The collection system should retain the policy's original actions, commands after safety filtering, the human's modifications, and the actual execution results, together with the reasons for intervention, its start and end times, and the subsequent outcome. If only the smooth trajectory after takeover is saved, a model may incorrectly attribute the human's behavior to the policy. If only the segment before failure is retained, it cannot learn how to recover. Linking warning, trigger, takeover, correction, recovery, and outcome within one event chain supports risk prediction, corrective policies, and recovery skills.

Human-in-the-Loop Sample-Efficient Robotic Reinforcement Learning (HIL-SERL) connects initial demonstrations and online human interventions with reinforcement learning on real robots, showing that human-in-the-loop data can provide continued supervision on the policy's execution distribution, beyond offline imitation~\citep{luo2025hilserl}. Hi-ORS further combines online rejection sampling with flexible human correction, exploiting recovery segments under comparatively stable supervision targets~\citep{lu2025hiors}. Together these works point to a shift from a single sequence of ``collect, then train'' to a short-cycle loop among execution, evaluation, and updating.

HELP further organizes human-in-the-loop involvement as a scarce supervision resource. Different staff members undertake multi-robot monitoring, physical resets, and high-value remote takeovers, while rollout segmentation distinguishes progress-making, idle, failure-inducing, and recovery segments~\citep{zhai2026help}. This exploration suggests that a scalable loop need not assign one person to continuously control each robot. One person can monitor several robots, intervene when risk warrants it, and concentrate limited human effort on states with the greatest learning value. The results still depend on the platform and task. No particular ratio of people to robots should be treated as a universal rule.

The quality gate is an intermediate layer of the loop. Before each round of data enters training, it should undergo completeness checks, physical-consistency checks, event segmentation, review of successes and failures, near-duplicate suppression, and provenance audits. After training, improvements must also be verified on held-out scenes, objects, operators, and embodiments. A new version that improves only on the collection distribution, or raises average success at the cost of safety and recovery, does not establish an effective feedback loop.

Safety is a system variable that determines what may be collected, when to intervene, and how the resulting data may be used. Task rules, policy uncertainty, kinematic and dynamic constraints, low-level monitoring, software and hardware interlocks, and human oversight should provide protection together. New policies should progress from offline checks through constrained environments and low-speed operation without a payload before normal operation. In settings involving people, hazardous materials, public roads, or critical infrastructure, near misses, controlled perturbations, and safe proxy tasks should take priority. Long-tail data should not be sought by causing real accidents.

Interactive feedback loops also apply to Physical AI systems that do not have a single robot body. Human takeover in a traffic system may mean reverting to a previous signal plan. Recovery in a factory may mean degrading process parameters and isolating equipment. Correction in a laboratory may mean stopping a procedure, recalibrating, or changing experimental conditions. Whatever the system, its recommendations, authorization decisions, issued commands, actual process, alarms, human interventions, and final results should remain linked in the same event chain.

The efficiency of the loop should be measured by capability gains per human-hour or robot-hour, alongside effective minutes of interaction, the fraction of usable episodes, rare-event yield, takeover frequency, recovery success, gains in coverage, and safety incidents. Raw operating time is merely a proxy for cost and cannot stand in for data value. The output of a short-cycle loop consists of training batches that pass quality gates and traceable model versions. The longer-cycle flywheel of deployment, failure discovery, renewed collection, and retraining is developed in Section~\ref{sec:physical-data-scaling-flywheel}.

\subsection{Physics Engines and Generative Worlds}
\label{sec:physics-engines-generative-worlds}
Section~\ref{sec:physical-data-foundations} defines physical data as records that link observations and actions to their consequences. Among these records, failures and the recoveries that follow them are the most informative and also the hardest to obtain (Section~\ref{sec:physical-data-collection}). A policy that must recover when a grasped object starts to slip needs to experience many slips, and a legged robot that must get up after a fall needs to fall, yet physical systems cannot be reset cheaply (Section~\ref{sec:physical-ai-background}). In simulation, a slip can be staged 1,000 times, and a simulated fall causes no physical damage to robot hardware. A model computes the recorded consequences, so their physical consistency has to be tested before the data can be trusted.

\begin{table}[tbp]
    \centering
    \caption{Sources of synthetic experience and what must be checked before use}
    \label{tab:synthetic-sources}
    \small
    \renewcommand{\arraystretch}{1.15}
    \setlength{\tabcolsep}{4pt}
    \begin{tabularx}{\textwidth}{
        @{}
        >{\hsize=0.8\hsize\linewidth=\hsize\raggedright\arraybackslash}X
        >{\hsize=1.15\hsize\linewidth=\hsize\raggedright\arraybackslash}X
        >{\hsize=0.95\hsize\linewidth=\hsize\raggedright\arraybackslash}X
        >{\hsize=1.1\hsize\linewidth=\hsize\raggedright\arraybackslash}X
        @{}
    }
        \toprule
        \textbf{Source}
        & \textbf{Role}
        & \textbf{Main benefit}
        & \textbf{Must be checked} \\
        \midrule

        Reconstructed scenes and digital cousins
        & The world: geometry and appearance taken from real sites, or variations of them
        & Deployment-like layouts and appearance
        & Collision geometry, articulation, physical parameters \\
        \addlinespace

        Generated assets and tasks
        & The world and its goals
        & Object, scene, and task diversity
        & Scale, mass and inertia, joint axes, task feasibility \\
        \addlinespace

        Physics engine
        & Consequences, from hand-written dynamics and contact
        & Privileged state, dense labels, counterfactual replay
        & Parameter values, missing mechanisms, solver artifacts \\
        \addlinespace

        Learned simulator
        & Consequences, from a network trained on solver or real trajectories
        & Fast rollouts, materials that are hard to model
        & Extrapolation, conservation, long-horizon drift \\
        \addlinespace

        Observation generator
        & Appearance only, with physics computed by an engine
        & Visual diversity over fixed interactions
        & Label preservation near contacts, thin parts, and occlusions \\
        \addlinespace

        Video world model
        & Consequences and observations, learned from video
        & Broad visual and behavioral priors
        & Action grounding, physical consistency, exploitation by the policy \\
        \addlinespace

        Demonstration generator
        & Actions, adapted from human seed demonstrations
        & Many demonstrations from few
        & Coverage of strategies, missing failures, controller match \\

        \bottomrule
    \end{tabularx}
\end{table}

Every synthetic trajectory depends on three choices. The world in which it unfolds may be authored by hand, reconstructed from scans, or generated. The consequences of each action may be computed by a physics engine or by a network trained on data. The actions may come from planners, reinforcement-learning experts, or adapted human demonstrations. The section first considers what simulation adds to a dataset and then takes the three choices in turn. Table~\ref{tab:synthetic-sources} summarizes the sources involved and what must be checked before their output is used. In what follows, synthetic data are any data produced by a model. Simulated data come from a physics engine, and generated data from a learned model. Most examples are drawn from manipulation and locomotion, where the evidence is richest, but the same questions arise for the simulators and digital twins of process plants, traffic systems, and laboratories.

\subsubsection{What Simulation Adds to a Dataset}

In the notation of Section~\ref{sec:physical-data-foundations}, a simulator turns chosen actions into trajectories through two conditional distributions,
\begin{equation}
    s_{t+1}\sim P_{\xi_{\mathrm{sim}}}^{\mathrm{tr}}(\cdot\mid s_t,a_t),
    \qquad
    o_t\sim O_{\psi_{\mathrm{render}}}(\cdot\mid s_t),
    \label{eq:piw:interface}
\end{equation}
where the physical and numerical parameters $\xi_{\mathrm{sim}}$ (masses, friction coefficients, time step, solver tolerance) govern the dynamics and the observation parameters $\psi_{\mathrm{render}}$ (camera models, lighting, sensor noise) govern what is sensed. Delays in the real system, such as actuator response time or sensor latency, have to be represented in $s_t$ or in an explicit history buffer. Otherwise the simulator assumes that commands act instantly and that sensors report the present. Two further ingredients complete a simulated dataset, namely the distribution of initial states and tasks from which episodes start and the predicate that decides whether an episode meets the success criterion. The dynamics, the observation model, the initial conditions, and the success predicate can each differ from reality, and Section~\ref{sec:sim-to-real-data-alignment} classifies the reality gap along these four lines.

The most visible gain is access to the full state. Object poses, contact forces, and material parameters can be recorded at every step, even though a deployed policy cannot observe them directly. This privileged information is valuable for training teachers, critics, and state estimators (Section~\ref{sec:dynamics-estimation} and Section~\ref{sec:motion-control}), provided it never leaks into the deployed policy's inputs. Each synthetic field should therefore be tagged as simulator state, rendered observation, or inferred quantity.

A deeper gain is the exact and inexpensive reset, which physical systems lack. Episodes restart from saved states, thousands of environments run in parallel, and the same initial state can be replayed under different actions, friction coefficients, or disturbances to see what changes the outcome. Counterfactual replay of this kind supports curricula, failure diagnosis, and targeted generation. Its findings concern the model, however. A policy's sensitivity to a simulated friction coefficient says something about a real material only after the simulator has been checked against measurements.

Simulated episodes need the same conventions as real trajectories and the same care in task design. An action vector means different things under torque, position, and Cartesian control, so the action specification of Section~\ref{sec:physical-data-foundations} belongs with every episode. Favorable starting poses hide difficult approaches, and a reset triggered as soon as an object slips ends the episode before the recovery that a deployed policy will need to learn. Success is best defined by the physical outcome of interest, such as a plug that is fully seated or a lid that stays closed, instead of by whatever the simulator makes easy to check.

\subsubsection{Scenes from Scans and from Generators}

Hand-built scenes are slow to author and tend to be few and tidy, so the training data available to policies cover a narrow range of environments. Reconstruction from real scans and automatic generation both widen that slice. Reconstruction has considerably reduced the appearance gap for visual policies. Neural radiance fields and Gaussian splatting (Section~\ref{sec:spatial-perception}) recover geometry and appearance, although collision geometry, articulation, and physical parameters must still be added before a scene can be simulated. With these additions, reconstructed scenes can largely stand in for real ones. RGB manipulation policies trained entirely on data from SplatSim, which renders a simulator's scenes with Gaussian splats, achieve 86.25\% average success in the reported SplatSim experiments zero-shot on real hardware, against 97.5\% for policies trained on real data~\citep{qureshi2024splatsim}. RialTo fine-tunes policies with reinforcement learning inside a reconstructed replica of the deployment scene itself, reporting an average improvement of 67\% in success over baselines under perturbed object poses, distractors, and disturbances~\citep{torne2024rialto}.

Some pipelines use reconstruction without simulating dynamics at all. Real2Render2Real turns a phone scan of an object and a single video of a person manipulating it into thousands of rendered robot demonstrations~\citep{yu2025r2r2r}. The object's motion is copied from the human and collision modeling is switched off, so no model computes the consequences of the robot's actions. Even so, in the reported Real2Render2Real experiments, policies trained on data rendered from one human video match policies trained on 150 teleoperated demonstrations. The approach works when the object's path alone defines the task, and it cannot handle tasks that depend on force.

Reconstruction also raises the question of how faithful a training scene needs to be. A digital twin is tied to one physical site and kept up to date with it. A digital cousin, by contrast, is a generated environment that preserves what matters for interaction in a real scene, such as where objects are and how they can be grasped or opened, without copying it exactly. In the reported digital-cousin experiments, policies trained across a set of cousins achieve 90\% zero-shot success in the original real scene, whereas policies trained on a single faithful twin achieve 25\%~\citep{dai2024acdc}.

Generating scenes at scale requires assets, and a training asset must be a physical specification as well as a shape. It needs consistent scale, collision geometry, mass and inertia, joint axes and limits, and material parameters. A photorealistic cabinet with a misplaced hinge axis adds visual variety but nothing a policy can practice on. Articulation is particularly hard to produce automatically. Articulate-Anything has a vision-language model write articulation code from text, images, or video and revise it after inspecting the result. On the PartNet-Mobility benchmark, the reported Articulate-Anything evaluation shows that this raises the success rate of reconstructed articulations from 8.7 to 11.6\% for earlier methods to 75\%~\citep{le2025articulate}.

Generative models can supply tasks as well as assets. GenSim and RoboGen prompt a language model to generate tasks, write the code that builds their scenes, and produce the supervision needed to learn them~\citep{wang2024gensim,wang2024robogen}. RoboCasa populates kitchens with assets from text-to-3D and text-to-image models, and RoboCasa365 scales the approach to 365 tasks across 2,500 kitchen layouts~\citep{nasiriany2024robocasa,nasiriany2026robocasa365}. A generated task is usable only if it resolves to identifiable objects, admissible actions, and a success predicate that can be checked, so generation pipelines usually execute each proposed task and discard those that a solver cannot complete.

\subsubsection{Agent-Built Scenes and Simulations}

The methods above use a language model for one part of a scene at a time. Agents built on language and vision-language models now assemble whole scenes, and increasingly whole simulations, by writing code, running it, and inspecting the result. Early systems focus on layout. Holodeck asks GPT-4 which objects a described room should contain and how they relate spatially, then places Objaverse assets with a constraint solver~\citep{yang2024holodeck}. Fine-tuning on Holodeck scenes raises a navigation agent's success in unfamiliar room types, such as music rooms and gyms, from 4.1\% to 20.4\%, against 17.0\% for scenes with the same assets in procedural layouts.

Later systems also supply physical properties. In SceneSmith, designer and critic agents build a room in stages, from the floor plan to furniture to small objects, while a vision-language model estimates each object's mass and material from rendered views and articulated furniture is drawn from a library~\citep{pfaff2026scenesmith}. The generated rooms contain about 71 objects, against 11 to 23 for earlier generators, and 95.6\% of the objects stay in place under simulation. SAGE uses a similar generate-and-critique loop to build scenes for a given task~\citep{xia2026sage}. Parameters inferred from appearance are enough to keep a scene standing, but contact-sensitive tasks still need them calibrated.

General-purpose models released in 2026 can carry out the whole real-to-sim process. In its 2026 release materials, OpenAI presents 3D world and asset generation as a main capability of GPT-6 Astra~\citep{openai2026astra}, and reports published within weeks of release describe several groups using it to build simulation scenes~\citep{li2026agenticrsr}.\footnote{Max Bickley and Ashley Goldstein, NVIDIA Technical Blog, \href{https://developer.nvidia.com/blog/how-to-use-ai-agents-to-prepare-3d-scenes-for-simulation/}{\emph{How to Use AI Agents to Prepare 3D Scenes for Simulation}}, September 16, 2026.} In Agentic RSR, a Codex agent running GPT-6 Astra turns an RGB video of a robot workspace into a MuJoCo scene with collision geometry and joints, then develops a policy program by running it there~\citep{li2026agenticrsr}. Across 18 tasks on 2 robot arms, 10 succeed in simulation and 8 of those also succeed on the real robots. The authors find that the agent behaves differently across interfaces and runs, so the model version belongs in the record of any data it generates.

Benchmarks show where such agents still fall short. In Video2World, agents rebuild a simulated world and the demonstrated behavior from a video of a robot or a person performing a task~\citep{tang2026video2world}. The best of 9 agent systems achieves 25.5\% task success over 222 instances, against 58.8\% for human-assisted reconstructions, and GPT-6 Astra achieves only 10.7\% although it recovers object shapes more accurately than any other agent. On indoor scans, the reported LiteReality-Agent evaluation shows 24 interpenetrating furniture pairs for GPT-6 Astra and none for the specialized LiteReality-Agent pipeline~\citep{huang2026literealityagent}, and on 4DCodeBench good static reconstruction does not carry over to deformation, fluids, or fracture~\citep{shen2026fourdcodebench}. Accurate geometry does not guarantee a scene in which the task can be performed.

Two recent systems close part of this gap by adding structure around the model. Text2Sim splits the writing of a Genesis simulation among planner, writer, and critic agents, and the writers consult Debug Cards, short notes distilled from 99 demonstrations in graphics papers on how simulations of a given kind fail and how to repair them~\citep{xiong2026text2sim}. On 42 held-out prompts, its automatic physical-quality score is 95.3, against 86.6 for direct use of its own underlying model (GPT-5.6-Sol). Without the cards, the score falls to 91.3. WorldAgent instead verifies every construction and simulation step in Blender, checking interpenetration, support, numerical stability, and causal order against scene geometry and simulator states, so it also catches overlaps hidden from the camera~\citep{wang2026worldagent}. With GPT-6 Astra as the underlying model, WorldAgent reduces the worst interpenetration during simulation to about 1/5 of that from direct prompting. In these evaluations, neither system is checked against measured behavior, and both infer physical parameters instead of measuring them.

World generators skip the agent altogether. Marble turns text, images, or video into Gaussian splats with a collision mesh, and such worlds have been imported into MuJoCo and Isaac Sim. The mesh is a single static surface, so the robot and the objects it manipulated were added from asset libraries~\citep{wmWorldLabs2025Marble}.\footnote{World Labs, \href{https://www.worldlabs.ai/case-studies/1-robotics}{\emph{Scaling Robotic Simulation with Marble}}, 2025.} Genie 3 generates an interactive environment from a prompt in real time and keeps it consistent for several minutes, and DeepMind's SIMA agent has pursued goals inside such worlds.\footnote{Jack Parker-Holder and Shlomi Fruchter, Google DeepMind, \href{https://deepmind.google/discover/blog/genie-3-a-new-frontier-for-world-models/}{\emph{Genie 3: A New Frontier for World Models}}, August 5, 2025.} A world of this kind has no explicit state or parameters to inspect, so, like the video world models discussed later in this section, it can be judged only by the consequences it predicts.

\subsubsection{Contact, Fidelity, and Throughput}

Once a world is built, a model must compute what happens in it, and the accuracy required depends on the task. Articulated rigid-body dynamics suffice for most legged locomotion and for pick-and-place with rigid objects, whereas cloth, cables, food, and soft tissue need deformable or continuum models. A useful test is whether the model reproduces the events that change the correct action, such as a grasp that slips, a peg that jams, or a support that gives way.

Most such events happen at contact. Section~\ref{sec:math-physical-world} describes contact through non-penetration and Coulomb friction, and engines differ mainly in how they enforce these two conditions. Classical time-stepping schemes pose them as a complementarity problem, in which two bodies at a contact either separate or push on each other, never both at once~\citep{stewart1996}. MuJoCo relaxes them into soft constraints and solves a convex optimization problem at each step, obtaining smooth, well-posed dynamics at the price of small interpenetrations and some compliance~\citep{todorov2012mujoco}. For many tasks the relaxation does no harm, and a calibrated compliance can even describe a rubber fingertip better than a perfectly rigid model. In tight-tolerance insertion or stacking, however, a soft contact model can let a part slide into place where the corresponding real part would bind.

Deformable objects make non-penetration harder to enforce. Incremental Potential Contact guarantees, under its algorithmic assumptions, that simulated deformable bodies never pass through each other and that mesh elements never turn inside out~\citep{li2020ipc}. Trajectories computed with it are free of interpenetration artifacts, but the method says nothing about the stiffness, damping, or friction of a particular sponge or garment, and these values must be measured before the recorded forces and deformations can be trusted.

Numerical artifacts are a separate hazard, because reinforcement learning can favor policies that exploit them. A policy can learn to exploit interpenetration, spurious impulses from numerical jitter, or contacts that stick when they should slide, and the time step, collision geometry, solver tolerance, and constraint softness determine which of these artifacts the optimizer can find. A cheap safeguard is to replay a sample of accepted trajectories with a smaller time step, tighter tolerances, or perturbed contact parameters, and to discard behaviors whose success does not survive. The check is best run before a large dataset is produced, since an artifact that goes unnoticed is copied into thousands of examples.

Throughput determines how much variation can be explored. Isaac Gym keeps physics simulation and policy training on the same graphics processing unit (GPU) and reports speedups of two to three orders of magnitude over pipelines that simulate on the central processing unit (CPU)~\citep{makoviychuk2021isaac}, which makes it possible to train quadruped walking policies in minutes~\citep{rudin2022learning}. For policies that use images, the bottleneck moves to rendering. ManiSkill3 parallelizes rendering on the GPU along with physics and reports that, with rendering enabled, it runs 10 to 1,000 times faster than other platforms on its benchmarks~\citep{tao2024maniskill3}. Frame rates are nonetheless a poor guide to the cost of data. They vary with scene complexity and contact density, and they leave out resets, policy inference, filtering, and storage. A better measure is the simulated counterpart of the effective data yield of Section~\ref{sec:physical-data-collection}, which counts only trajectories that pass the quality gates and weighs their cost against how much they improve the policy.

\subsubsection{Learned Simulators and Video World Models}

Hand-written engines are limited to what their authors can model. Learned models trade explicit physical structure for coverage of phenomena that are hard to write down, and they can replace the solver, the renderer, or both. The price is that their errors are harder to diagnose. An error in an engine can usually be traced to a parameter, a missing mechanism, or a solver setting, whereas an error in a learned model is spread across its weights. Measurements on the physical system then become the main check on its physical consistency.

Learned simulators replace the solver but keep the state-based interface of Eq.~\eqref{eq:piw:interface}. Graph Network-based Simulators and MeshGraphNets, for example, learn how particles and meshes evolve for fluids, granular media, cloth, and deformable solids~\citep{sanchezgonzalez2020learning,wmPfaff2021}. In its evaluation, MeshGraphNets runs 1 to 2 orders of magnitude faster than the solvers used to produce its training data. Such models are reliable mainly within their training regime. Over long rollouts their errors accumulate, and conservation laws hold only approximately unless they are built into the architecture (Section~\ref{sec:learning-physical-world}).

Observation generators leave the physics to an engine and replace only the renderer, changing the lighting, textures, backgrounds, and clutter that the camera sees. Cosmos-Transfer, for instance, generates realistic video conditioned on segmentation, depth, and edge maps rendered by a simulator~\citep{nvidia2025cosmostransfer}. The risk is that the generator alters something the labels depend on. If it shifts a handle by a few centimeters while restyling the scene, and the simulator keeps the original pose, a correct grasp label becomes wrong. Agreement between image and simulator state therefore needs checking near contacts, around thin structures, and under occlusion, where such errors are most likely.

Video world models replace both, learning the consequences of actions directly from recordings (Chapter~\ref{ch:wm})~\citep{bruce2024genie,yang2024unisim,nvidia2025cosmos}. UniSim, an action-conditioned video simulator, learns from image, robot, and navigation data and is used to train policies that transfer to real robots. For generating data, the first question is which training data the model has received. A model trained on demonstrations rarely observes failures or unusual contact, so it tends to predict that actions succeed. Training on autonomous play, as in PlayWorld (Section~\ref{sec:physical-data-collection}), exposes a world model to the slips, collisions, and near misses that demonstrations omit~\citep{yin2026playworld}.

Because video models produce pixels, actions have to be recovered before their output can train a policy. DreamGen fine-tunes an image-to-video model on the target embodiment and uses it to generate videos of familiar and new tasks~\citep{jang2025dreamgen}. The videos are then labeled with actions, either by an inverse-dynamics model, which infers the action taken between consecutive frames, or by a latent-action model, which infers an abstract action code. Starting from teleoperation data for a single pick-and-place task in one environment, a humanoid trained on these generated trajectories acquires 22 new behaviors in seen and unseen environments. The recovered actions are action proxies in the sense of Section~\ref{sec:physical-data-foundations}, and they can be screened by replaying them in a physics simulator and keeping those whose simulated motion matches the generated video~\citep{kim2026robocurate}. A policy can also be fine-tuned inside a world model, on imagined successes or on rewards that a vision-language model assigns to imagined rollouts~\citep{guo2026ctrlworld,sharma2026worldgymnast}, at the risk of learning to exploit the model's errors (Section~\ref{sec:wm-validation}).

\subsubsection{Synthetic Demonstrations and Pretraining Corpora}

The third choice is the agent. Behavior can come from motion planners that use the simulator's exact state, from reinforcement-learning experts trained with privileged observations and then distilled into policies that use deployable sensors (Section~\ref{sec:dynamics-estimation}), from programs written by language models, or from a few human demonstrations adapted to new situations. Each source leaves its mark on the data. Planners produce smoother and more direct motions than an operator would produce, reinforcement learning may find strategies that a person would not demonstrate, and adapted demonstrations carry human habits along with human skill.

MimicGen shows how far a few demonstrations can be stretched. It splits each demonstration into segments defined relative to the objects involved, transforms the segments to new object poses, stitches them together, and keeps the results that succeed when executed in simulation. In MimicGen, about 200 human demonstrations yield more than 50,000 demonstrations across 18 tasks with varied scenes, objects, and robot arms. Policies trained on the synthesized demonstrations perform comparably to policies trained on additional human demonstrations~\citep{mandlekar2023mimicgen}. Later systems extend the method to long-horizon skill sequences and to bimanual dexterous hands~\citep{garrett2024skillmimicgen,jiang2025dexmimicgen}. MimicGen's output, however, recombines the strategies in the seed demonstrations without adding new ones, and because failed executions are discarded, the resulting training data provide few examples of failure or recovery.

Synthetic demonstrations should pass the same quality gates that Section~\ref{sec:physical-data-collection} applies to real batches, along with checks specific to synthesis. These checks cover action semantics that match the deployed controller, privileged fields that are marked, and outcomes that survive the solver checks described earlier. Rejected samples are worth keeping, sorted by the reason for rejection, because different reasons call for different responses. Invalid geometry, a mismatch between the rendered observation and the simulator state, and an infeasible task all point to defects in the generator. A valid execution that ends in failure is ordinary failure data, unsuitable as an imitation target but useful for learning to predict outcomes and to recover. If generator versions, parameter distributions, solver and renderer settings, and random seeds are kept on record, a problem found later can be fixed at its source before more data are produced from it.

Synthetic data have also moved beyond narrow skills into pretraining. Simulated trajectories make up a large part of the training mixture of the GR00T~N1 humanoid model, and GraspVLA, pretrained on 1 billion frames of simulated grasping, transfers zero-shot to real objects~\citep{bjorck2025gr00t,deng2025graspvla}. The most direct comparison comes from InternData-A1. In InternData-A1, a model with the $\pi_0$ architecture, pretrained only on its 630,000 simulated trajectories, matches the official $\pi_0$ on simulated, real, and long-horizon dexterous tasks. The official $\pi_0$ had been pretrained largely on real robot data~\citep{tian2025interndata}. Results of this kind are strongest where outcomes depend mainly on geometry, as in grasping and pick-and-place. Where outcomes depend on contact dynamics, a synthetic corpus is only as useful as the calibration of the engine that produced it, and calibration requires real data.

\subsection{Sim-to-Real Transfer and Real-World Alignment}
\label{sec:sim-to-real-data-alignment}

A policy trained on synthetic data inherits the errors of its source, and how much those errors matter depends on the task. In legged locomotion, policies trained entirely in simulation with randomized dynamics and learned actuator models now routinely transfer without real-world fine-tuning. Manipulation is harder. Contact, deformation, and vision all differ between simulation and reality, and transfer there typically involves some real data. This section begins with what simulated and real data each contribute. It then follows the order in which the reality gap is commonly addressed. This sequence consists of finding where simulation and reality disagree, correcting the simulator with measurements, training policies that are robust to the remaining error, and checking the result in closed loop on the physical system.

\subsubsection{What Simulated and Real Data Each Contribute}

\begin{table}[tbp]
    \centering
    \caption{Complementary strengths of simulated and real data}
    \label{tab:sim-vs-real}
    \small
    \renewcommand{\arraystretch}{1.15}
    \setlength{\tabcolsep}{4pt}
    \begin{tabularx}{\textwidth}{
        @{}
        >{\hsize=0.7\hsize\linewidth=\hsize\raggedright\arraybackslash}X
        >{\hsize=1.15\hsize\linewidth=\hsize\raggedright\arraybackslash}X
        >{\hsize=1.15\hsize\linewidth=\hsize\raggedright\arraybackslash}X
        @{}
    }
        \toprule
        \textbf{Aspect}
        & \textbf{Simulated data}
        & \textbf{Real data} \\
        \midrule

        Scale and cost
        & Millions of episodes, with cost dominated by building scenes and by compute
        & Limited by robots, operators, and resets, with cost growing with every episode \\
        \addlinespace

        Perception and labels
        & Exact poses, contacts, masks, and depth, although rendering may look unrealistic
        & Sensor noise, calibration error, missing depth, with labels that must be estimated or annotated \\
        \addlinespace

        Dynamics
        & Only as accurate as the contact, material, and actuator models
        & Real consequences, including effects that no model captures, though their causes may go unmeasured \\
        \addlinespace

        Rare and dangerous states
        & Staged and replayed on demand
        & Rare, costly, and sometimes unsafe to collect \\
        \addlinespace

        Main role
        & Breadth, privileged supervision, pretraining, stress tests
        & Training where dynamics matter, calibration of simulators and world models, validation \\

        \bottomrule
    \end{tabularx}
\end{table}

Choosing between simulated and real data means trading scale and label accuracy against the reliability of the dynamics (Table~\ref{tab:sim-vs-real}). Simulation offers scale, exact perception labels, and free resets (Section~\ref{sec:physics-engines-generative-worlds}), but its dynamics, and therefore the consequences it computes, are only as accurate as its contact, material, and actuator models. Real data have the reverse profile. They are limited by robot time, operators, and the cost of resetting the world, and their observations are imperfect. Cameras add noise and blur, depth sensors drop out on shiny surfaces, calibration drifts, and object poses or contact forces must be estimated instead of read off. Their dynamics, on the other hand, are the real ones. A real trajectory reflects the actual friction, compliance, backlash, and delay of the system, including effects that no simulator models, even without direct measurements of those effects. Neither source dominates. Simulation is most useful where exact labels and broad coverage matter and the dynamics are easy to model, while real data are hard to replace where the outcome depends on contact, deformation, or actuation that simulators approximate poorly.

Because real data are scarce, how they are spent matters as much as how much is collected. They serve three purposes, namely training policies directly, calibrating the simulators and world models that generate synthetic data, and validating both the generators and the policies trained on their output. The three purposes compete for the same evidence, and evidence spent on one cannot serve as an independent check on another. Trajectories used to fit a simulator cannot show that the fit is accurate, and trials used to select a policy cannot certify it.

\subsubsection{Where the Reality Gap Comes from}

The four ingredients of a simulated dataset listed in Section~\ref{sec:physics-engines-generative-worlds} also give a natural way to classify the reality gap (Table~\ref{tab:sim2real-gaps}). A dynamics gap means that $P_{\xi_{\mathrm{sim}}}^{\mathrm{tr}}$ in Eq.~\eqref{eq:piw:interface} is wrong, either because its parameters are off or because it lacks a mechanism. Errors in mass, inertia, friction, and compliance alter accelerations and contact outcomes, and actuators add discrepancies of their own through gains, torque limits, backlash, and delay. An observation gap means that $O_{\psi_{\mathrm{render}}}$ misrepresents what the sensors report, whether in appearance, camera calibration, depth dropouts on shiny or transparent surfaces, proprioceptive bias, or timing. Touch is a particular weak point, since many tactile sensors respond through elastomer deformation and internal optics that simulators capture only roughly. The remaining two gaps are easier to miss. A distribution gap arises when training does not cover the initial states, objects, and tasks the robot actually meets. An objective gap arises when simulated success is not real success, as when a success predicate credits an insertion that reaches the right depth but would not hold in a real fixture.

\begin{table}[tbp]
    \centering
    \caption{Components of the reality gap and how they are diagnosed and addressed}
    \label{tab:sim2real-gaps}
    \small
    \renewcommand{\arraystretch}{1.15}
    \setlength{\tabcolsep}{4pt}
    \begin{tabularx}{\textwidth}{
        @{}
        >{\hsize=0.75\hsize\linewidth=\hsize\raggedright\arraybackslash}X
        >{\hsize=1.05\hsize\linewidth=\hsize\raggedright\arraybackslash}X
        >{\hsize=1.15\hsize\linewidth=\hsize\raggedright\arraybackslash}X
        >{\hsize=1.05\hsize\linewidth=\hsize\raggedright\arraybackslash}X
        @{}
    }
        \toprule
        \textbf{Gap}
        & \textbf{Typical symptom}
        & \textbf{Diagnostic}
        & \textbf{Main remedies} \\
        \midrule

        Dynamics ($P_{\xi_{\mathrm{sim}}}^{\mathrm{tr}}$)
        & Overshoot, sluggish tracking, unexpected slip or jamming, wrong deformation
        & Replay identical command sequences in simulation and on hardware and compare the responses
        & System identification, actuator or residual models, dynamics randomization \\
        \addlinespace

        Observation ($O_{\psi_{\mathrm{render}}}$)
        & A correct plan executed at the wrong place or time, and failures under particular lighting or materials
        & Compare sensor streams for matched states and measure end-to-end latency
        & Calibration, latency modeling, sensor noise models, visual randomization \\
        \addlinespace

        Distribution (initial states and tasks)
        & Failures concentrated on unseen objects, layouts, or initial states
        & Stratify evaluation by condition
        & Targeted real or synthetic collection, broader scene and task generation \\
        \addlinespace

        Objective (success predicate)
        & Episodes counted as successes in simulation that fail in reality
        & Audit success predicates against physical outcomes
        & Redefine success and reward and relabel data \\

        \bottomrule
    \end{tabularx}
\end{table}

Each gap calls for a different remedy, so the gap has to be identified before anything is changed. A delayed command is fixed by calibrating the interface, an unmodeled deformation needs a better model, and an action that is executed correctly but is wrong for the task needs new supervision. Telling these cases apart requires the commanded and executed actions, events, and outcomes in the trajectory record (Section~\ref{sec:physical-data-foundations}), which a binary success flag does not provide.

A gap of any kind is amplified in closed loop. An early mistake moves the robot into a state that no demonstration covered, the policy's next action there is worse, and errors snowball. For imitation learning, this intuition has a precise form. With per-step task costs bounded in $[0,1]$, a cloned policy with error probability $\epsilon_{\mathrm{BC}}$ under the expert state distribution can incur an excess expected cost of order $T^2\epsilon_{\mathrm{BC}}$ over $T$ steps~\citep{ross2010efficient}. Dataset Aggregation (DAgger) collects expert labels on states visited during rollouts that may mix expert and learner policies. Under the assumptions of its analysis and with sufficiently many iterations, some learned iterate has excess expected task cost relative to the expert bounded by $u_{\mathrm{dev}}T\epsilon_{\mathrm{DAgger},N_{\mathrm{iter}}}+O(1)$. Here $\epsilon_{\mathrm{DAgger},N_{\mathrm{iter}}}$ is the minimum average surrogate loss achievable within the policy class over the state distributions induced by the rollout policies across $N_{\mathrm{iter}}$ iterations, and $u_{\mathrm{dev}}$ bounds the increase in the expert's remaining cost after one deviating action~\citep{ross2011dagger}. The constant $u_{\mathrm{dev}}$ is a measure of recoverability, the property that Section~\ref{sec:physical-ai-background} argues planning must take into account when resets are not free.

The analysis assumes access to an expert policy that can label the states visited during data collection. A privileged teacher in simulation can supply such labels when its policy is defined on those states, and teacher-student distillation usually labels the states that the student visits. Human operators find it hard to say what the robot should have done in a state they are not controlling, which is why intervention-based collection (Section~\ref{sec:physical-data-collection}) lets them take over when the policy goes wrong. Either way, training data must include the states that the deployed policy actually reaches.

\subsubsection{Calibrating Simulators and World Models with Real Data}

Calibration adjusts a model's parameters until it reproduces real measurements. With $\mathcal D_{\mathrm r}$ denoting a set of real trajectories, system identification can be written as Bayesian inference over the simulator parameters,
\begin{equation}
    p(\xi_{\mathrm{sim}}\mid\mathcal D_{\mathrm r})
      \propto p(\mathcal D_{\mathrm r}\mid\xi_{\mathrm{sim}})\,p(\xi_{\mathrm{sim}}),
    \label{eq:pir:identification}
\end{equation}
where the prior $p(\xi_{\mathrm{sim}})$ encodes datasheet values and physical bounds and the likelihood measures how well the simulator with parameters $\xi_{\mathrm{sim}}$ reproduces the recorded observations. When states are only partially observed, the likelihood must also account for state and sensor uncertainty. A point estimate is often enough in practice, but the full posterior keeps track of ambiguities that matter later, for instance when randomization ranges are chosen.

The data used for identification must be informative. As Section~\ref{sec:control-autonomy} explains, accurate tracking alone does not imply parameter convergence. Persistent excitation of the regressor supports parameter convergence under the assumptions of the specified adaptive law. Identification likewise requires data that distinguish the relevant parameter directions. Slow motions in free space reveal little about actuator bandwidth, grasps that never slip reveal little about sliding friction, and different combinations of motor strength, payload, and gain can produce nearly identical trajectories. Calibration experiments should therefore excite the mechanisms the task depends on, for example with fast reversals for actuator response, controlled pushes for friction, and loaded lifts for inertia, while staying within safe limits.

Where first-principles models fall short, models learned from data often do better. Learned actuator networks, which map commanded positions and recent history to delivered torque, are central in the early legged-transfer study (Section~\ref{sec:dynamics-estimation})~\citep{hwangbo2019}. ASAP applies data-driven correction to a whole humanoid. A motion-tracking policy trained in simulation is run on the real robot, and the recorded rollouts train a delta-action model, a network that adds a correction to each action inside the simulator so that simulated motion matches the recorded real-robot motion. The policy is then fine-tuned in the corrected simulator. In tests of transfer between two simulators and deployment on a real Unitree G1, the reported ASAP evaluation shows that the approach tracks agile whole-body motions more accurately than system identification, domain randomization, or a learned correction to the simulator's dynamics~\citep{he2025asap}.

Learned corrections can, however, fit the data for the wrong reason. A correction trained on one set of motions may absorb a calibration error elsewhere in the system, and testing at speeds, loads, and contact points outside the fitting data exposes such errors. The same tests separate parameter errors, which calibration can fix, from structural ones, which it cannot. No value of $\xi_{\mathrm{sim}}$ repairs a model that omits cable twisting or a compliant gripper pad.

Nor is a single best-fit simulator always necessary. SimOpt instead calibrates a distribution over parameters. It adjusts the distribution so that simulated rollouts of the current policy match real ones, retrains the policy under the updated distribution, and repeats. Policies trained in this way transfer to real swing-peg-in-hole and drawer-opening tasks~\citep{chebotar2019simopt}. A simulator fitted to one policy's behavior can remain wrong in regions that policy never visits, so calibration data should also cover the behaviors that later policies are likely to try.

Learned simulators can be calibrated more directly, because their parameters can be updated by gradient descent on real data. NeRD replaces the dynamics and contact solvers of a conventional simulator with a learned model that can be fine-tuned on real trajectories~\citep{xu2025nerd}. For video world models, VLAW alternates between improving the model and improving the policy. Real rollouts of a vision-language-action policy refine the world model, and the refined model generates rollouts for training the policy~\citep{guo2026vlaw}. On a real robot, the authors report that the method raises success by 39.2 percentage points over the pretrained policy and by 11.6\% over fine-tuning on the same real rollouts filtered for success. In that evaluation, the refined world model thus adds value beyond the real data used to build it. Establishing the accuracy of the model would still require fresh trajectories from conditions outside the fitting data.

\subsubsection{Randomization, Adaptation, and Co-Training}

Calibration narrows the uncertainty that measurements can resolve, and the policy must tolerate whatever remains. Domain randomization trains it to do so across a distribution $p_{\eta_{\mathrm{sim}}}(\xi_{\mathrm{sim}})$ over simulator parameters, shaped by design parameters $\eta_{\mathrm{sim}}$,
\begin{equation}
    \max_{\theta}\;
    \mathbb E_{\xi_{\mathrm{sim}}\sim p_{\eta_{\mathrm{sim}}}}
       \bigl[J(\pi_\theta;\xi_{\mathrm{sim}})\bigr],
    \label{eq:pir:randomization}
\end{equation}
where $J(\pi_\theta;\xi_{\mathrm{sim}})$ is the expected return of the policy in the world with parameters $\xi_{\mathrm{sim}}$. Training across varied models aims to reduce reliance on any one exact model, and randomizing appearance and dynamics in this way underlies much of sim-to-real transfer in perception and control~\citep{tobin2017domain,peng2018}.

Choosing $p_{\eta_{\mathrm{sim}}}$ in Eq.~\eqref{eq:pir:randomization} is a data-design problem. Parameters should be resampled on the timescale at which they actually vary. An object's mass is fixed within an episode, sensor noise changes with every reading, and actuator response drifts slowly with temperature and wear. Dependencies should be explicit, since appearance and friction are correlated for a given material.

Wide ranges are not enough by themselves. A range that contains the true value is of little use if the distribution places almost no probability near it, and ranges that are too broad produce cautious policies without fixing the failures seen in deployment. Moderate variation around a well-chosen nominal value often helps more than pinning each parameter to a single estimate, as the digital-cousin result of Section~\ref{sec:physics-engines-generative-worlds} suggests, and Section~\ref{sec:physical-data-scaling-flywheel} reports a similar effect of diversity in real data. The posterior in Eq.~\eqref{eq:pir:identification} is a natural source for both the nominal values and their spread, and evaluation should report performance for groups of conditions as well as the average over the distribution.

Automatic domain randomization automates part of this design by widening each range as the policy meets the performance criterion within the current one, which turns randomization into a curriculum. OpenAI uses it to train a Shadow Dexterous Hand to solve a Rubik's cube. In ten reported trials, the hand solves cubes needing 15 face rotations 6 times and solves the hardest cubes (which need 26 face rotations) 2 times~\citep{akkaya2019rubik}. The result also depends on a cube instrumented with sensors, a classical solver that chooses the sequence of face rotations, and months of large-scale training.

Adaptive policies deal with the remaining uncertainty differently. Instead of performing well under every condition, they infer the current one from recent history and adjust to it, as in the teacher-student and rapid motor adaptation methods of Section~\ref{sec:dynamics-estimation}~\citep{kumar2021}. Training them requires data that contain changes and transients, such as a payload added in the middle of a task or the onset of slip, as well as conditions that are unknown but fixed. When adaptive rollouts are later reused for training, the inferred context should be stored with each example, because an action that is correct for a heavy payload becomes a wrong label once the payload information is dropped.

In vision-based manipulation, where both the visual and the physical gap are large, a common alternative is to train on simulated and real data at once and let each cover the other's weaknesses (Table~\ref{tab:sim-vs-real}). Co-training minimizes a weighted sum of losses over the two sources, the mixture objective of Eq.~\eqref{eq:c9cao:mixture} with one simulated and one real component. Across tasks on a robot arm and a humanoid, the reported co-training study shows that adding simulation data improves real-world performance by an average of 38\%, and the benefit persists when the real dataset grows to 400 demonstrations~\citep{maddukuri2025cotraining}. A controlled study of planar pushing from pixels examines when and why the benefit appears~\citep{wei2025cotraining}. In that study, simulation data raise success two- to sevenfold when real data are scarce, but the gains plateau unless real data grow too. Narrowing the physics gap matters more than improving visual fidelity. Unexpectedly, the best policies can distinguish simulated inputs from real ones, and a modest visual difference between the domains helps. A mechanistic analysis attributes this to a balance between aligning representations across domains, so that knowledge transfers, and keeping the domains distinguishable, so that the policy can apply the right dynamics to each~\citep{lei2026mechanistic}.

The same pushing study finds that mixing the sources during training outperforms pretraining in simulation and then fine-tuning on real data, and that fine-tuning a co-trained policy does not consistently help. These comparisons involve single tasks and modest datasets. Whether broad multi-task pretraining followed by fine-tuning helps is a separate question, taken up in Section~\ref{sec:physical-data-scaling-flywheel}.

\subsubsection{Validating Transfer and Closing the Loop}

Most of the evaluation that guides development happens away from hardware, yet the criterion remains closed-loop performance on the physical system (Section~\ref{sec:physical-ai-background}), so every proxy evaluator has to be checked against it. SIMPLER builds simulated counterparts of real manipulation setups, matching camera views, appearance, and control. It compares policy rankings in simulation with their rankings on real robots~\citep{li2024simpler}. Agreement is measured by the Pearson correlation between simulated and real success rates and by the mean maximum rank violation (MMRV), which penalizes the simulator for misordering two policies in proportion to the gap between their real success rates. By these measures, simulated evaluation tracks real results far better than offline metrics such as the error of predicted actions on held-out demonstrations. World models can serve the same purpose. In WorldGym, success rates measured inside an action-conditioned video model correlate with real success at $r=0.78$ on tasks from the BridgeData V2 tabletop setup~\citep{quevedo2026worldgym}.

Simulated and real trials can also be combined statistically instead of merely compared. SureSim treats the problem as prediction-powered inference. A small number of paired real and simulated trials estimate the simulator's bias, and that estimate corrects a large number of simulated trials~\citep{badithela2025suresim}. The result is a valid confidence interval on real performance, which raw simulated success rates cannot provide, and an interval of similar width needs 20 to 25\% fewer hardware trials than one estimated from hardware trials alone. Like any agreement between a proxy and reality, the correction holds for the policies and conditions for which it is measured and must be rechecked when either changes.

Consider a fleet of mobile manipulators that open drawers in office kitchens. At one site, pulls repeatedly stall after an approach and grasp that look correct in the logs. Engineers rebuild the scene in simulation from the logs and a scan, and the simulated robot reproduces the approach but opens the drawer without difficulty. Since the approach is reproduced and the stall is not, the discrepancy most likely lies in the simulated drawer rather than in perception or in the grasp. Pulls with a force-sensing gripper reveal a soft-close damper and a detent that the simulated drawer lacks. Once the damper and detent are modeled and their parameters fitted (Eq.~\eqref{eq:pir:identification}), randomization is restricted to damping and detent force around the identified values, and new scenes vary handle position and pull direction near the failure. The retrained policy is screened in simulation and then tested on held-out drawers at that site and at others, to confirm that the fix generalizes beyond the cabinet where the failure occurs.

The example also shows the many points at which real data enter a pipeline nominally trained in simulation, and claims of transfer should account for all of them. ``Zero-shot'' usually means that the policy was not fine-tuned on the target embodiment, yet real data may still have been used to identify parameters, reconstruct scenes, pretrain perception, or select checkpoints. Reporting these uses makes the contribution of simulation measurable. It also sharpens a question that co-training raises and that Section~\ref{sec:physical-data-scaling-flywheel} takes up. The question concerns how many synthetic trajectories one real trajectory is worth.

\subsection{Scaling Laws and the Data Flywheel for Physical AI}
\label{sec:physical-data-scaling-flywheel}

Language-model scaling studies show predictable improvements as data, parameters, and compute grow together, and progress in vision-language-action models now depends mainly on data (Section~\ref{sec:brain-vla}). Whether physical data will yield equally predictable gains is far from clear. Every trajectory comes from a particular body in a particular world and costs robot time and usually human time. The practical question concerns allocation. It covers which data to collect or generate and in what proportions. It also covers how to divide scarce physical interaction among training, calibration, and validation. A further aspect is when a robot system can collect what it needs through its own operation. Such a system, a data flywheel, is worth running only if the capability it adds outweighs its cost. Throughout, value is measured by closed-loop performance on the physical system, not by loss, frame counts, or simulated success.

\subsubsection{The Data Pyramid}

\begin{figure}[tbp]
    \centering
    \includegraphics[width=0.95\linewidth]{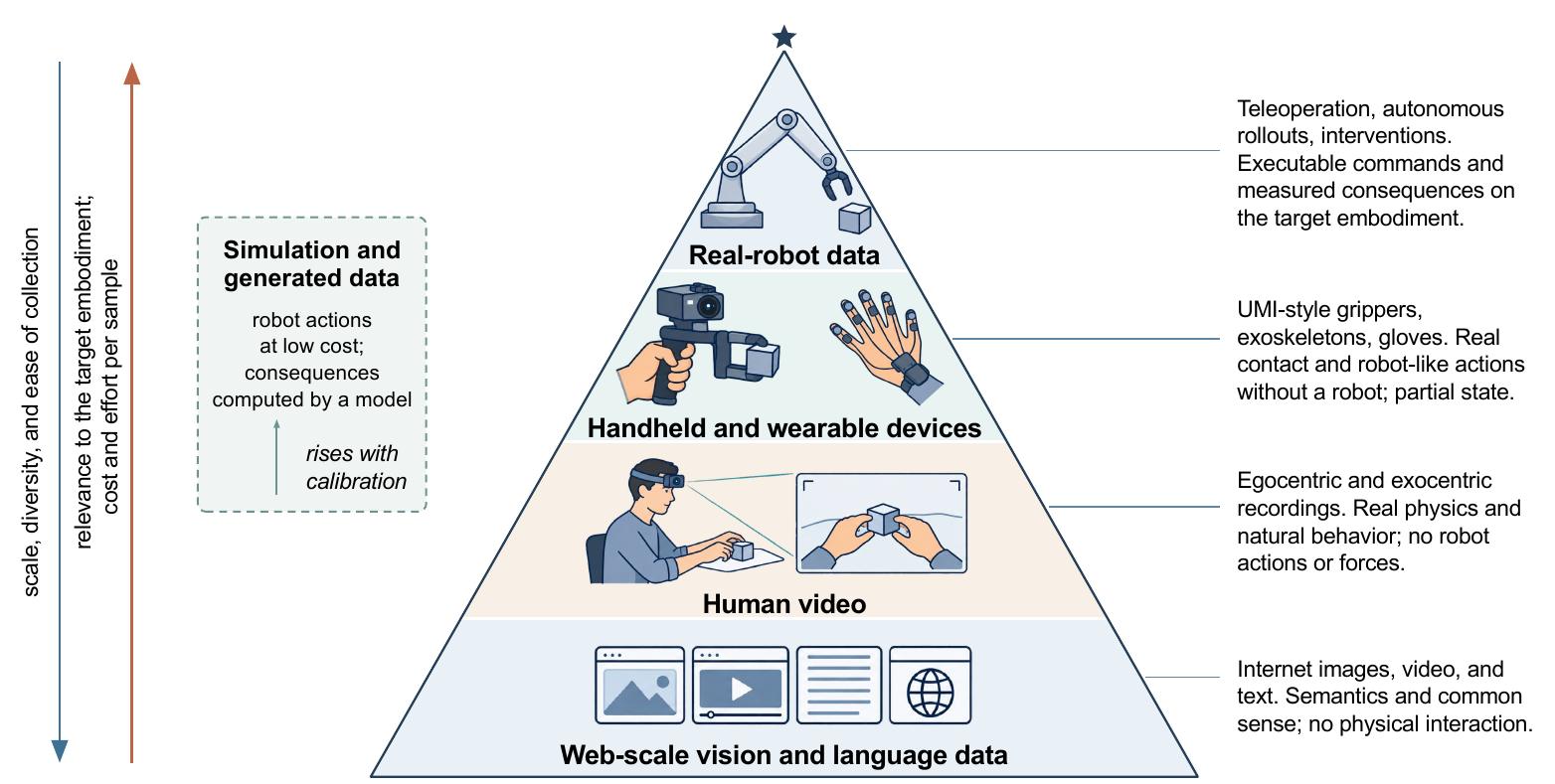}
    \caption{The data pyramid for Physical AI. For physically collected data, relevance to the target embodiment and cost per sample rise together toward the apex, where the target embodiment acts in the physical world. Lower layers are cheaper, larger, and more diverse but supervise physical execution less directly. Simulated and generated data are drawn separately because they supply robot actions at low cost while computing their consequences with a model, so their effective position depends on calibration. Adapted from~\citet{bjorck2025gr00t} and~\citet{ye2026pyramid}. The order of the layers reflects relevance and cost, not dataset quality}
    \label{fig:data-pyramid}
\end{figure}

Section~\ref{sec:physical-data-collection} describes human data as a gradient from rich semantics to executable actions. Figure~\ref{fig:data-pyramid} extends that gradient to all physically collected data and adds the dimension of scale, with layers widening toward the base as cost per sample falls. The trajectory record of Section~\ref{sec:physical-data-foundations} gives a way to judge relevance. Data are most relevant when they record the target embodiment issuing executable commands under deployment conditions, together with the measured consequences of those commands. Each layer below the apex gives up part of this record. Handheld and wearable devices capture real contact and robot-like actions without a robot, human video captures real physics without robot actions, and web data supply semantics without physical interaction. The most relevant data are also the most expensive, since every hour of them needs a robot, an operator, and resets, and they are correspondingly scarce. Goldberg estimates that the largest teleoperation datasets amount to about a year of robot experience, whereas the text and images used to train vision-language models would take a person some 100,000 years to read and view~\citep{goldberg2025datagap}.

Simulation does not fit this ordering, which is why the figure draws it separately. It supplies executable actions on a model of the target embodiment at almost no marginal cost. The reported GR00T~N1 study produces 780,000 simulated trajectories, equivalent to about 6,500 hours of human demonstration, in 11 hours~\citep{bjorck2025gr00t}. Because its consequences are computed, its relevance depends on calibration (Section~\ref{sec:sim-to-real-data-alignment}), and simulated corpora have so far done best where geometry decides the outcome (Section~\ref{sec:physics-engines-generative-worlds}). Section~\ref{sec:c9:sensing-actuation} places simulation alongside handheld devices in a related pyramid organized by the kind of evidence each source provides.

A capable system needs both ends of the pyramid. The apex provides executable actions and real consequences on the system's own body. The base provides diversity. Chen and colleagues argue that diversity sets the ceiling on how far a model can generalize and that, for now, only internet-scale data supply enough of it~\citep{wmChen2026Roadmap}. Much of this chapter concerns lowering the cost of relevant data and raising the value of cheap data through calibration, retargeting, and filtering.

Within the physically collected layers, the collection method largely determines cost and yield. Section~\ref{sec:physical-data-collection} describes how each method works. The paragraphs below compare their economics, including what an hour of collection costs, how much it yields, and what the resulting data lack.


Egocentric video is the cheapest to collect and the easiest to scale, since it needs only a head-mounted camera or glasses. Its drawback is the embodiment gap, since human hands differ from robot grippers and video records neither commands nor forces. The gap can be narrowed by design. If the robot is built to resemble the human and both are recorded with the same sensors, human data can be worth more per hour than robot data. In EgoMimic, the reported evaluation shows that an hour of egocentric recording yields about 1,400 demonstrations against 135 from teleoperation, and adding an hour of human data improves the co-trained policy more than adding an hour of robot data~\citep{kareer2025egomimic}.

A handheld gripper costs a few hundred dollars and needs no robot. Using UMI, 3 people collect 1,400 demonstrations across 30 locations in 12 person-hours, and on 1 task collection runs more than 3 times faster than teleoperation with a 3D mouse~\citep{chi2024umi_ch6}. Because the fingers match the robot's, gripper-level actions transfer directly. The arm itself goes unrecorded, so its joint states and dynamics are missing and reachability has to be checked at deployment. Forces are also absent unless tactile sensors are added.

Teleoperation yields the most relevant data and the least of it. The hardware has become cheaper. A kinematically matched leader arm costs under \$300, and a complete bimanual system, robots included, costs about \$20,000~\citep{wu2023gello,zhao2023}. A robot and an operator, however, remain in the loop for every hour. The reported DROID data-collection effort involves 50 collectors at 13 institutions gathering 76,000 trajectories over one year~\citep{khazatsky2024droid}. As Section~\ref{sec:physical-data-collection} notes, the data also carry the imprint of the interface, since latency, limited haptic feedback, and operator style shape the trajectories.

Exoskeletons and hand trackers fall between the handheld gripper and teleoperation, both in cost and in how closely they match the robot. An exoskeleton with the same kinematics as the target arm records joint-level motion directly. AirExo-2 costs about \$600 for a dual-arm rig, and in the reported user study its operators complete a task in 5.7~s on average against 17.3~s with joint-space teleoperation~\citep{fang2025airexo2}. Gloves and motion-capture systems extend joint-level recording to dexterous hands, and the DexCap study reports collection 3 times faster than teleoperation~\citep{wang2024dexcap}. In exchange, the operator has to wear the rig, the data must be retargeted whenever the human and robot hands differ, and force feedback is limited.

Across the four methods, the cheaper the source, the further its data sit from the target embodiment. How much each source is worth for a given task has to be measured. Scaling studies are the usual instrument, but in their standard form they rest on assumptions that physical data do not meet.

\subsubsection{Scaling Laws and What They Assume}
\label{subsec:physical-scaling-assumptions}

Scaling laws come from language modeling. \citet{kaplan2020scaling} find that the test loss of language models falls as a power law in model size, dataset size, and training compute, with some trends spanning more than seven orders of magnitude. A common form separates the contributions of the parameter count $N_{\mathrm{par}}$ and the amount of training data $N_{\mathrm{data}}$,
\begin{equation}
    L_{\mathrm{scale}}(N_{\mathrm{par}},N_{\mathrm{data}})=L_{\mathrm{floor}}+a_{\mathrm{scale}}N_{\mathrm{par}}^{-\alpha_{\mathrm{scale}}}+b_{\mathrm{scale}}N_{\mathrm{data}}^{-\beta_{\mathrm{scale}}},
    \qquad a_{\mathrm{scale}},b_{\mathrm{scale}},\alpha_{\mathrm{scale}},\beta_{\mathrm{scale}}>0,
    \label{eq:pis:scaling}
\end{equation}
where $L_{\mathrm{scale}}$ is a held-out loss and $L_{\mathrm{floor}}$ an irreducible floor. With compute $C_{\mathrm{train}}\approx 6N_{\mathrm{par}}N_{\mathrm{data}}$, about six floating-point operations per parameter per training token, minimizing Eq.~\eqref{eq:pis:scaling} at fixed $C_{\mathrm{train}}$ makes the optimal $N_{\mathrm{par}}$ and $N_{\mathrm{data}}$ grow as powers of $C_{\mathrm{train}}$. The exponents prove harder to pin down than the form. The original study concludes that most additional compute should go into parameters, whereas \citet{hoffmann2022compute} find that parameters and data should grow in equal proportion as compute increases. Later work traces most of the difference to how parameters and compute are counted and to optimizer settings~\citep{pearce2024reconciling,porian2024resolving}. An exponent, in other words, reflects the measurement protocol as well as the model family.

The allocation argument also rests on the three assumptions that every unit of data is equally informative, that compute is the only cost, and that lower loss means a better model. Physical data strain all three. Robot trajectories are highly correlated, so much of a robot dataset behaves like repeated data. Even in language modeling, repeating data for more than about 4 epochs yields rapidly diminishing returns~\citep{muennighoff2023scaling}. Collecting robot data consumes robot time, operators, and resets that never appear in $C_{\mathrm{train}}$. Section~\ref{sec:physical-ai-development-strategies} calls this resource physical interaction capacity and argues that it must grow along with investment in compute. Held-out loss, finally, is a weak guide to closed-loop success. In the robomimic study, the checkpoint with the lowest validation loss performs significantly worse than the checkpoint with the highest rollout success~\citep{mandlekar2022matters}.

\begin{table}[h]
    \centering
    \caption{Selected evidence on scaling in robot learning and related domains}
    \label{tab:robot-scaling-evidence}
    \small
    \renewcommand{\arraystretch}{1.15}
    \setlength{\tabcolsep}{4pt}
    \begin{tabularx}{\textwidth}{
        @{}
        >{\hsize=0.7\hsize\linewidth=\hsize\raggedright\arraybackslash}X
        >{\hsize=0.9\hsize\linewidth=\hsize\raggedright\arraybackslash}X
        >{\hsize=1.0\hsize\linewidth=\hsize\raggedright\arraybackslash}X
        >{\hsize=0.9\hsize\linewidth=\hsize\raggedright\arraybackslash}X
        >{\hsize=1.5\hsize\linewidth=\hsize\raggedright\arraybackslash}X
        @{}
    }
        \toprule
        \textbf{Study}
        & \textbf{Varied}
        & \textbf{Setting}
        & \textbf{Measured}
        & \textbf{Finding and caveat} \\
        \midrule

       ~\citet{lin2025datascaling}
        & Environments, objects, demonstrations
        & Single-task imitation with handheld-gripper data
        & Real success in new environments with new objects
        & Rough power law in environments and objects, demonstrations saturate, single tasks only \\
        \addlinespace

       ~\citet{ai2025embodiment}
        & Number of training embodiments
        & Locomotion on about 1,000 generated bodies
        & Performance on unseen bodies
        & Adding embodiments beats adding data on fixed ones, locomotion only \\
        \addlinespace

       ~\citet{pearce2025agents}
        & Model size, data, compute
        & Pretraining of imitation policies and world models
        & Pretraining loss
        & Language-like power laws, with coefficients that depend on tokenizer, task, and architecture \\
        \addlinespace

       ~\citet{baniodeh2025waymo}
        & Compute, model size, data
        & Motion forecasting and planning on 500,000 hours of driving
        & Loss, open- and closed-loop metrics
        & Power law in compute, closed-loop metrics also improve, optimal model size grows 1.5 times as fast as data \\
        \addlinespace

       ~\citet{tri2025lbm}
        & Pretraining data and tasks
        & Multitask dexterous manipulation
        & Blind real and simulated success after fine-tuning
        & 3 to 5 times less task data needed, mixed results without fine-tuning \\
        \addlinespace

       ~\citet{generalist2025gen0}
        & Pretraining data, model size
        & Over 270,000 hours of real manipulation data
        & Validation loss and action-prediction error
        & Power law in data, small models stop improving, company blog, not peer reviewed \\
        \addlinespace

       ~\citet{xu2026precision}
        & Required precision
        & Simulated high-precision manipulation
        & Demonstrations needed for a fixed success rate
        & $\log N_{\mathrm{demo}}=m_{\mathrm{fit}}/(P_{\mathrm{tol}}-c_{\mathrm{lim}})+n_{\mathrm{fit}}$ for $P_{\mathrm{tol}}>c_{\mathrm{lim}}$. The fitted limit depends on sensing and expert quality. \\

        \bottomrule
    \end{tabularx}
\end{table}

\subsubsection{What Robot-Learning Studies Show}

Evidence for scaling in robot learning accumulates, but controlled studies remain rare. The cited meta-analysis of 327 papers finds that robot performance generally improves as a power law in data, model size, and compute, and that it improves faster than performance on language tasks~\citep{sartor2024neural}. Because meta-analyses average over very different tasks and metrics, the clearest evidence comes from studies that vary one factor at a time, summarized in Table~\ref{tab:robot-scaling-evidence}. Some of these studies measure loss rather than success. Pretraining studies of imitation policies and world models, for example, find language-like power laws in loss, with coefficients that depend on the tokenizer, task, and architecture~\citep{pearce2025agents}.

The most consistent finding is that diversity matters more than volume. In a study of single-task imitation with handheld grippers, generalization to new environments and objects follows a rough power law in the number of environments and objects seen in training, while extra demonstrations in environments already covered stop helping beyond a threshold~\citep{lin2025datascaling}. Guided by this finding, the study reports that 4 people collect enough data in one afternoon to train policies that succeed about 90\% of the time on 2 tasks in new environments with unseen objects. Locomotion shows the same pattern across bodies. Adding procedurally generated embodiments to training improves generalization to unseen ones substantially more than adding data for a fixed set, and the best policy transfers zero-shot to real Unitree Go2 and H1 robots~\citep{ai2025embodiment}.

Broad pretraining pays off chiefly by making new tasks cheaper to learn. In the Toyota Research Institute's study of large behavior models, evaluated in blind real and simulated trials, fine-tuning a pretrained model matches single-task training with 3 to 5 times less task data, and the advantage grows under distribution shift~\citep{tri2025lbm}. Without fine-tuning, the pretrained model gives mixed results. Pretraining may also extend what a model can handle. In the first training phase of $\pi_{0.5}$, 97.6\% of the examples come from sources other than mobile manipulators working in homes, such as other robots and web data, yet the reported model controls mobile manipulators in homes that had not appeared in its training data~\citep{pi2025pi05}.

The split of additional compute between model size and data has been measured directly in at least one physical domain. In motion forecasting and planning for driving, closed-loop metrics improve along with the loss as compute increases, and the compute-optimal model size grows 1.5 times as fast as the dataset~\citep{baniodeh2025waymo}. Driving logs accumulate passively across a fleet and are far cheaper than manipulation data, so the result does not carry over directly, but it shows that the measurement can be made in a physical setting.

Industry now reports manipulation datasets far larger than academic ones. In a blog post, Generalist AI describes pretraining on more than 270,000 hours of real-world manipulation data~\citep{generalist2025gen0}. It reports that the post-training validation error falls as a power law in the amount of pretraining data, and that roughly billion-parameter models stop improving with more data while seven-billion-parameter and larger models keep improving. These are prediction errors on the company's own tasks rather than closed-loop success rates, and they have not been peer reviewed.

Precision imposes a different kind of limit. In the cited simulation study, the demonstration count $N_{\mathrm{demo}}$ needed for a fixed success rate is modeled by the empirical fit $\log N_{\mathrm{demo}}=m_{\mathrm{fit}}/(P_{\mathrm{tol}}-c_{\mathrm{lim}})+n_{\mathrm{fit}}$, with fitted coefficients $m_{\mathrm{fit}}>0$ and $n_{\mathrm{fit}}$. Here $P_{\mathrm{tol}}$ is the length tolerance and $c_{\mathrm{lim}}$ the fitted system limit. For $P_{\mathrm{tol}}>c_{\mathrm{lim}}$, this fit predicts that $N_{\mathrm{demo}}$ diverges as $P_{\mathrm{tol}}$ approaches $c_{\mathrm{lim}}$ from above~\citep{xu2026precision}. The limit depends on the whole system, and adding a wrist camera or using a better expert moves it toward finer precision. For such tasks, better sensing can pay off more than additional demonstrations.

All such curves are hard to measure precisely. Real-robot success rates are estimated from tens or hundreds of trials, so each point carries a wide confidence interval (Section~\ref{sec:c9:reliability}) and a fitted exponent is loosely constrained. With that caveat, the evidence supports consistent gains from diversity across environments, objects, embodiments, and data sources, and from pretraining followed by fine-tuning. It does not yet establish universal exponents, compute-optimal model sizes for robot policies, or a dependable link between loss and closed-loop success.

Better evidence requires purpose-built experiments that vary the quantity and the diversity of data independently while keeping the training procedure fixed. Such experiments would report new trajectories apart from extra passes over old ones, evaluate on held-out conditions that stay the same between runs, and check predictions made from small runs against a larger run held in reserve. A plateau in such an experiment deserves attention in its own right, since it may point to missing sensing, an inadequate model, or an unsuitable objective instead of a shortage of data.

\begin{table}[p]
    \centering
    \caption{Selected hubs and large public datasets for robot learning, with access conditions as of October 2026}
    \label{tab:data-hubs}
    \footnotesize
    \renewcommand{\arraystretch}{1.15}
    \setlength{\tabcolsep}{4pt}
    \begin{tabularx}{\textwidth}{
        @{}
        >{\hsize=0.7\hsize\linewidth=\hsize\raggedright\arraybackslash}X
        >{\hsize=1.25\hsize\linewidth=\hsize\raggedright\arraybackslash}X
        >{\hsize=0.7\hsize\linewidth=\hsize\raggedright\arraybackslash}X
        >{\hsize=1.35\hsize\linewidth=\hsize\raggedright\arraybackslash}X
        @{}
    }
        \toprule
        \textbf{Resource}
        & \textbf{Contents}
        & \textbf{Access}
        & \textbf{Link} \\
        \midrule
        \multicolumn{4}{@{}l}{\textit{Hubs and indexes of many datasets}} \\
        \addlinespace
        LeRobot hub
        & Community robot datasets in the LeRobot format, numbering in the tens of thousands
        & Licenses set per dataset, with some gated
        & \href{https://huggingface.co/datasets?other=LeRobot}{\nolinkurl{huggingface.co/datasets?other=LeRobot}} \\
        \addlinespace
        Open X-Embodiment
        & 60 datasets from 34 labs, 22 embodiments, over 1 million trajectories in RLDS format~\citep{openx2024}
        & Open
        & \href{https://robotics-transformer-x.github.io/}{\nolinkurl{robotics-transformer-x.github.io}} \\
        \addlinespace
        ARIO
        & About three million episodes from 258 series and 321,064 tasks in a unified format~\citep{wang2024ario}
        & Open, with original licenses retained
        & \href{https://imaei.github.io/project_pages/ario/}{\nolinkurl{imaei.github.io/project_pages/ario}} \\
        \addlinespace
        TensorFlow Datasets (TFDS) catalog
        & Robotics section with RLDS versions of many robot datasets
        & Open
        & \href{https://www.tensorflow.org/datasets/catalog/overview}{\nolinkurl{www.tensorflow.org/datasets/catalog/overview}} \\
        \addlinespace
        OpenDataLab
        & Dataset platform hosting AgiBot World among other collections
        & Registration
        & \href{https://opendatalab.com/}{\nolinkurl{opendatalab.com}} \\
        \addlinespace
        Awesome Embodied Data Pyramid
        & Curated list of datasets organized by pyramid layer~\citep{ye2026pyramid}
        & Open
        & \href{https://github.com/worldbench/awesome-embodied-data-pyramid}{\nolinkurl{github.com/worldbench/awesome-embodied-data-pyramid}} \\
        \addlinespace
        \multicolumn{4}{@{}l}{\textit{Large single-source datasets}} \\
        \addlinespace
        AgiBot World
        & About 1 million trajectories over 217 tasks from 100 robots~\citep{agibot2025colosseo}
        & Gated, CC BY-NC-SA 4.0
        & \href{https://agibot-world.com/}{\nolinkurl{agibot-world.com}} \\
        \addlinespace
        DROID
        & About 76,000 Franka manipulation trajectories from 564 scenes~\citep{khazatsky2024droid}
        & Open download
        & \href{https://droid-dataset.github.io/}{\nolinkurl{droid-dataset.github.io}} \\
        \addlinespace
        RoboMIND
        & 107,000 trajectories on four embodiments and 479 tasks, including failures~\citep{wu2025robomind}
        & Gated, Apache 2.0
        & \href{https://x-humanoid-robomind.github.io/}{\nolinkurl{x-humanoid-robomind.github.io}} \\
        \addlinespace
        RH20T
        & Over 110,000 contact-rich sequences with force, audio, and tactile data~\citep{fang2024rh20t}
        & CC BY-SA 4.0 or non-commercial subsets
        & \href{https://rh20t.github.io/}{\nolinkurl{rh20t.github.io}} \\
        \addlinespace
        BridgeData V2
        & 60,096 trajectories in 24 environments on a low-cost arm~\citep{walke2023bridge}
        & CC BY 4.0
        & \href{https://rail-berkeley.github.io/bridgedata/}{\nolinkurl{rail-berkeley.github.io/bridgedata}} \\
        \addlinespace
        Ego4D
        & 3,670 hours of egocentric video~\citep{grauman2022ego4d}
        & License agreement
        & \href{https://ego4d-data.org/}{\nolinkurl{ego4d-data.org}} \\
        \addlinespace
        EgoDex
        & 829 hours of egocentric video with tracked 3D hand poses~\citep{hoque2025egodex}
        & CC BY-NC-ND
        & \href{https://github.com/apple/ml-egodex}{\nolinkurl{github.com/apple/ml-egodex}} \\
        \bottomrule
    \end{tabularx}
\end{table}

\subsubsection{Valuing and Curating Heterogeneous Data}

Raw volume overstates what a dataset contains. Experience is nested, with frames inside trajectories, trajectories inside tasks and scenes, and scenes inside embodiments and sites, so counts of hours and frames treat many correlated samples as independent. This is why Section~\ref{sec:physical-data-collection} measures collection by effective data yield, and scaling studies should likewise report growth in independent interactions and their spread over tasks, scenes, objects, and embodiments.

How many samples from one layer are worth one sample from another can be measured in specific settings. In the InternData-A1 study (Section~\ref{sec:physics-engines-generative-worlds}), fewer than 1,600 simulated samples match the performance of 200 real ones on 4 tasks with closely aligned simulated and real setups, a ratio below 8:1~\citep{tian2025interndata}. EgoMimic's hour-for-hour comparison, described above, gives a corresponding rate for human data. Both rates depend on how closely the source matches the task, on how much contact is involved, and on how the sources are weighted in training, and neither should be expected to hold elsewhere.

Curation can matter as much as collection. CUPID uses influence functions to estimate how a trained policy's closed-loop return would change if a given demonstration were removed, and ranks every demonstration by its effect on performance~\citep{agia2025cupid}. On the simulated robomimic benchmark, training on less than a third of the data, selected in this way, produces diffusion policies that match the state of the art. Hardware experiments show similar gains. Related methods weight whole datasets, as Re-Mix does for Open X-Embodiment by choosing mixture weights that maximize worst-case performance across its constituent datasets~\citep{hejna2024remix}, or score demonstrations by the mutual information between states and actions~\citep{hejna2025deminf}. A scaling curve plotted against raw demonstration counts can therefore partly reflect how much harmful or redundant data a collection contains.

Knowing which kinds of diversity matter helps both collection and retrieval. A study using procedurally generated datasets with controlled composition identifies camera poses and the spatial arrangement of objects as the dimensions most relevant in that study, both for diversifying new collections and for retrieving relevant demonstrations from existing datasets. Retrieval guided by these findings raises the real-world performance of policies trained with DROID data by up to 70\%~\citep{saxena2025matters}.

Retrieval and curation presuppose that data can be found. A growing number of hubs collect robot datasets, and Table~\ref{tab:data-hubs} lists some of the largest. Aggregators such as the LeRobot hub, Open X-Embodiment, and ARIO gather datasets in common formats, making cross-dataset training and retrieval practical, while large single-source collections offer consistent protocols at scale. Licenses and access conditions range from open to gated and non-commercial, and, as the documentation practices of Section~\ref{sec:physical-data-foundations} require, they should travel with any dataset derived from these sources.

\subsubsection{Building a Data Flywheel}

Flywheels have worked best where success can be detected cheaply, because the robot can then label its own experience. Large-scale autonomous collection on real robots therefore focuses on grasping in early studies, where a robot can tell by itself whether it is holding the object. One early study reports more than 800,000 grasp attempts on up to 14 robots~\citep{levine2018grasping}, and the QT-Opt study uses more than 580,000 attempts to train a vision-based Q-function that grasps unseen objects with 96\% success~\citep{kalashnikov2018qtopt}. Where success is harder to judge, people or learned models have to supply the labels. Sirius-Fleet uses a visual world model to predict whether a robot's upcoming actions are likely to fail and asks a human to intervene when they are. As the policy improves, the failure predictor adapts and requests for help become rarer~\citep{liu2024sirius}.

Post-training has since moved to the fleet level. RECAP, used to train $\pi^{*}_{0.6}$, combines demonstrations, autonomous experience, and expert corrections in batched offline reinforcement learning (Section~\ref{sec:manipulation-training})~\citep{physicalintelligence2025b}. SOP streams on-policy experience and intervention signals from a fleet to a central learner that runs either interactive imitation, in which humans correct the robot during execution, or RECAP~\citep{pan2026sop}. In its experiments, the wall-clock time needed to reach 80\% success falls from about 174 minutes with 1 robot to 127 with 2 and 72 with 4, a 2.4-fold speedup from 4 times as many robots. The human-in-the-loop methods of Section~\ref{sec:physical-data-collection} are the short-cycle version of the same idea.

One iteration of such a loop can be written so that the three uses of real data become explicit. Let $\mathcal D_k$ be the accumulated training data at iteration $k$, $\pi_k$ the current policy, $W_{k}^{\mathrm{sim}}$ the simulator or world model, and $\mathcal E_{\mathrm{eval}}$ a held-out evaluation set collected under conditions that training never sees. Then
\begin{align}
    \mathcal B_{k}^{\mathrm{data}}
      &=\operatorname{Collect}(\pi_k,\mathcal A_{k}^{\mathrm{acq}}), \notag\\
    \mathcal D_{k+1}
      &=\mathcal D_k\cup\operatorname{Curate}(\mathcal B_{k}^{\mathrm{data}}), \notag\\
    (\tilde\pi,W_{k+1}^{\mathrm{sim}})
      &=\operatorname{TrainAndCalibrate}(\mathcal D_{k+1},W_{k}^{\mathrm{sim}}), \notag\\
    \pi_{k+1}
      &=\operatorname{Accept}(\tilde\pi,\pi_k;\mathcal E_{\mathrm{eval}}),
    \label{eq:pis:flywheel}
\end{align}
where $\mathcal A_{k}^{\mathrm{acq}}$ is the collection plan, $\operatorname{Curate}$ applies the quality gates of Section~\ref{sec:physical-data-collection}, and $\operatorname{Accept}$ keeps the candidate $\tilde\pi$ only if it outperforms $\pi_k$ in closed loop on $\mathcal E_{\mathrm{eval}}$. Each batch $\mathcal B_{k}^{\mathrm{data}}$ feeds both policy training and, through the procedures of Section~\ref{sec:sim-to-real-data-alignment}, the recalibration of $W_{k}^{\mathrm{sim}}$, while $\mathcal E_{\mathrm{eval}}$ stays outside both. Even so, repeated acceptance decisions gradually fit the policy to $\mathcal E_{\mathrm{eval}}$ itself, so the evaluation set has to be refreshed from time to time, with a final set held back for reporting.

For the drawer fleet of Section~\ref{sec:sim-to-real-data-alignment}, $\operatorname{Collect}$ in Eq.~\eqref{eq:pis:flywheel} uploads every operator takeover and every pull whose force profile leaves the expected envelope, while subsampling routine successes. $\operatorname{Curate}$ turns takeovers into recovery demonstrations attributed to the operator, $\operatorname{TrainAndCalibrate}$ retrains the policy and refits the drawer model, and $\operatorname{Accept}$ compares the result with the frozen previous version on held-out drawers at the failing site and at others. The release process that follows, including replay testing, staged deployment, and release gates, is described in Section~\ref{sec:software-infrastructure}.

\subsubsection{Choosing What to Collect Next}

Section~\ref{sec:physical-data-collection} discusses how active sampling selects tasks and states within a collection campaign. At the level of the flywheel the choice is broader, because the next unit of effort may go to a new site, a calibration experiment, corrective demonstrations, a batch of simulated scenarios, or held-out trials for validation. A natural criterion is the expected improvement per unit cost,
\begin{equation}
    x^{\star}=\argmax_{x}\;
    \frac{\mathbb E\bigl[\Delta J\mid x\bigr]}{c_{\mathrm{acq}}(x)},
    \label{eq:pis:acquisition}
\end{equation}
where $\Delta J$ is the resulting change in closed-loop performance on held-out conditions and $c_{\mathrm{acq}}(x)$ is the cost of collecting, curating, and learning from $x$. Physical options are expensive, since resets take time and some trials carry risk. Synthetic options are cheaper but less certain to help. In practice, candidates are ranked with heuristic scores such as novelty, disagreement between models, gaps in coverage, and predicted failure~\citep{agia2024sentinel,xu2025faildetect}. Each score can mislead. An unfamiliar background may look novel without adding any physical information, and persistent uncertainty may come from sensor noise instead of a gap the policy could learn to close (Section~\ref{subsec:self-evolution}).

Since Eq.~\eqref{eq:pis:acquisition} can at best be approximated before training, a collection strategy has to be judged after the fact, by retraining on the selected data and testing the result. The comparison should be made at a matched budget against reasonable alternatives, in particular stratified sampling over known conditions. It should also include a frozen baseline evaluated under the same conditions (Section~\ref{sec:physical-ai-development-strategies}), so that a better strategy is not confused with a larger investment. All costs belong in the same account, from robot occupancy and resets to interventions, annotation, simulation, and training. For repeatable tasks, total cost per accepted output is a convenient summary. Flywheels complicate the comparison further, because each policy changes the data collected next and because hardware, sites, and operators drift between iterations.

The measurements that would settle the central questions of this chapter are still largely missing. Most published scaling curves for robot learning put demonstrations, hours, frames, or tokens on the horizontal axis. Few relate closed-loop capability to the total amount of real interaction, counting the simulation and human effort that come with it. Such curves would show whether the next robot-hour is better spent on training data, on calibration, or on evaluation. They would also show whether a flywheel is actually improving the system. Chapter~\ref{ch:general-physical-ai} considers what it would take for loops of this kind to sustain open-ended growth in capability.

\clearpage 

\section{World Models}
\label{ch:wm}

\begin{figure}[hb]
  \centering
  \includegraphics[width=\linewidth]{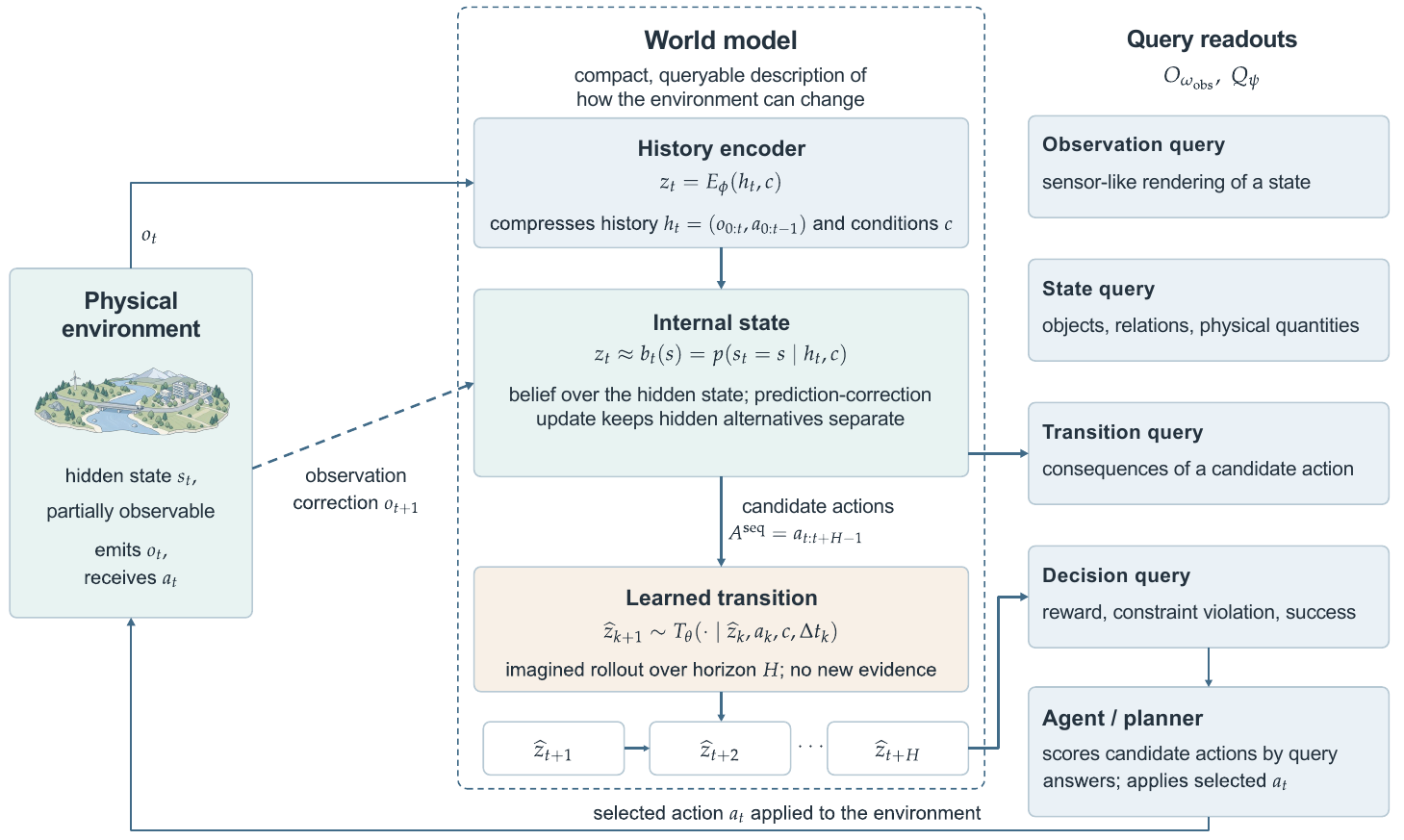}
  \caption{A world model as a compact, queryable description of how the environment can change. Solid arrows carry information available at deployment. The dashed arrow carries the observation correction that revises the internal state after each step. A particular model may expose only a subset of the four query interfaces}
  \label{fig:wm-concept}
\end{figure}

Previous chapters discuss how perception captures physical scenes at the current moment, and physical data record how it evolves over time (Chapter~\ref{ch:physical-data}). With them, we can present a more comprehensive account of world modeling, which studies how an environment evolves over time and how external actions or interventions change that evolution. This is also an important step for the physical agent that will be discussed in the later section, as it helps a physical agent to predict, before acting, how the world will evolve and how its own action would change that evolution. Figure~\ref{fig:wm-concept} illustrates this closed-loop interaction among the physical environment, the world model, and the querying agent. Within the technical stack of this book, world models build on the state-space formulations of Chapter~\ref{sec:technical-foundations}, the perception of Chapter~\ref{ch:perception}, and the data and simulation of Chapter~\ref{ch:physical-data} to provide action-conditioned predictions for the manipulation and whole-body control models in Chapters~\ref{ch:manipulation-models} to \ref{ch:whole-body-motor-control}.

Based on this, this chapter frames physical world modeling in five parts. Section~\ref{sec:wm-concepts} establishes the conceptual framework (state, belief, query, and evidence) and clarifies what a world model is and is not. Section~\ref{sec:wm-architectures} surveys how the internal state is represented, ranging from generative observation-space models to latent dynamics, 3D and object-centric structures, and unified multimodal architectures. Section~\ref{sec:wm-training} develops training strategies, from self-supervised pretraining to action-conditioned and interaction-driven learning. Section~\ref{sec:wm-validation} addresses how predictions are deployed and validated across forecast accuracy, physical consistency, and downstream decision utility. Section~\ref{sec:wm-outlook} concludes with open challenges and the long-term role of world models in Physical AI.

Additionally, \emph{A Definition and Roadmap for World Models} organizes physical world modeling around finite-resource compression, local observation, asynchronous multimodal evidence, and the relationship between understanding and prediction~\citep{wmChen2026Roadmap}. This perspective organizes the chapter's discussion of current systems and future directions. The scope of the chapter includes robots, scientific instruments, and distributed physical processes rather than only visually observed manipulation.

\subsection{Concepts and Theoretical Frameworks of World Models}
\label{sec:wm-concepts}

World models are compact, queryable descriptions of how an environment can change. They are not required to reproduce every physical detail. Their purpose is to retain the distinctions needed by the questions that an agent must answer, including what is present, what is hidden, what will happen under an action, and which action is preferable. This section fixes the common vocabulary of \emph{state}, \emph{transition}, \emph{query}, and \emph{evidence} used throughout the chapter. A model is useful when these elements are connected and its claims can be checked against new observations.

\subsubsection{Definition, Boundaries, and Basic Properties}
\label{sec:wm-definition}

Physical scenes contain objects, fields, energy, and interactions at many scales, while an agent has finite sensors, memory, and time. A world model therefore compresses observations into an internal state from which future states and task-relevant outputs can be queried. The compression is task-relative. A logo may be irrelevant when predicting whether a cup will spill, but essential when identifying a reagent. A lower-dimensional representation is not automatically a better one. It is better only when it preserves the distinctions needed by the intended queries.

The same visible image can correspond to different velocities, contact modes, temperatures, or internal masses. A stationary drawer can be unlocked or held by a latch, and the same pull then has different consequences. A world model must use history, interaction, and background conditions to maintain these hidden alternatives. Operationally, its functions include inference from evidence, conditional evolution over time, and readout of observations or physical and task quantities. These functions may be modular or share parameters. A particular model can expose only a subset of them, but its advertised interfaces should have testable meanings.

\paragraph{A Development Trajectory Organized by Underlying Problems}
Three enduring questions explain the main technical lineages. First, how can an agent infer a sufficient state from incomplete measurements? State-space estimation and filtering make a distinction between an evolving state and its noisy observations. Kalman's work provides a foundational linear formulation~\citep{Kalman1960}. Second, how can an internal model improve action rather than merely fit observations? Dyna connects real experience, model learning, planning, and acting, so the learned model supplies additional experience for decision learning~\citep{wmSutton1991}. Third, how can these functions operate on high-dimensional sensory data? Learned encoders and recurrent dynamics replace hand-specified state variables with representations shaped by prediction and control. The \emph{World Models} architecture separates visual compression, temporal memory, and a controller. PlaNet performs planning through learned latent dynamics. Dreamer trains behavior using imagined latent trajectories~\citep{ha2018world,hafner2019planet,hafner2019dreamer}.

These developments do not form a sequence in which each new architecture makes its predecessors obsolete. Explicit states remain useful when physical variables are measurable, generative models retain information for rich observation queries, and decision-oriented latent models can avoid reconstructing details that a task does not use. Recent multimodal and world-action models extend the same underlying questions to the sharing of semantic, predictive, and behavioral information. Their shared foundation is the connection among state inference, conditional prediction, and action learning. The discussion below compares implementations of these functions rather than presenting successive product generations.

\paragraph{Predictive Sufficiency Rather than a Miniature Copy of the World}
The central abstraction is a state that preserves answers, not necessarily a state that reproduces every observation. Predictive state representations make this idea explicit by representing a controlled system through predictions of future observation tests rather than through a prescribed list of hidden variables~\citep{wmLittman2002}. Learned embeddings, object states, and physical fields are different ways of approximating such predictive sufficiency. A useful idealized criterion is
\begin{equation}
 p(y_{t+k}^{q_{\mathrm{query}}}\mid h_t,A^{\mathrm{seq}},c)
 \approx p_\Theta(y_{t+k}^{q_{\mathrm{query}}}\mid E_\phi(h_t,c),A^{\mathrm{seq}},c),
 \qquad q_{\mathrm{query}}\in\mathcal Q,\;A^{\mathrm{seq}}\in\mathcal A_H,
 \label{eq:wm-predictive-sufficiency}
\end{equation}
where $\mathcal Q$ is the set of supported queries, $y_{t+k}^{q_{\mathrm{query}}}$ the outcome of query $q_{\mathrm{query}}$ at offset $k$, $\mathcal A_H$ a declared set of input sequences over horizon $H$, and $A^{\mathrm{seq}}$ one such externally supplied sequence. The history is $h_t$, $c$ records known conditions, $E_\phi$ compresses history into an internal state, and $\Theta$ parameterizes the predictive distribution. The left side describes the environment under the specified experimental input protocol. The right side describes its learned approximation. This is a design criterion, not a guarantee of identifiability from passive recordings. Its tolerance, horizon, and query set must be stated.

The criterion explains why compression and generality trade off. A representation sufficient to decide whether a drawer opens can omit texture but must preserve the latch hypothesis. A representation intended to answer new visual questions may need much more detail. A state sufficient for ranking a particular policy's actions need not be sufficient for estimating forces, synthesizing views, or optimizing a different objective. Broader query coverage therefore requires richer targets, retained evidence, or a mechanism for refining the representation on demand. Changes of latent coordinates do not change predictive capability when all interfaces transform consistently, so physically meaningful readouts matter more than requiring every latent coordinate to have a unique physical name.

\paragraph{Boundaries with Neighboring Systems}
Several neighboring systems should be distinguished by their interfaces. A video generator can reproduce motion without correctly responding to a specified intervention. A physics engine advances a supplied state but need not infer that state from open-ended observations. A digital twin is tied to a particular physical counterpart, whereas a policy outputs actions without necessarily exposing conditional futures. Table~\ref{tab:wm-neighboring-concepts} summarizes the distinctions used here. These are functional distinctions. One implementation may combine several rows, and a model's name does not establish all the listed capabilities.

\begin{table}[tbp]
  \centering
  \caption{Neighboring concepts and the capabilities their interfaces do not establish. The rows describe functions and may overlap within one system}
  \label{tab:wm-neighboring-concepts}
  \small
  \renewcommand{\arraystretch}{1.12}
  \begin{tabularx}{\textwidth}{@{}>{\raggedright\arraybackslash}p{0.17\textwidth} >{\raggedright\arraybackslash}X >{\raggedright\arraybackslash}X >{\raggedright\arraybackslash}X@{}}
    \toprule
    \textbf{System} & \textbf{Primary object} & \textbf{Typical interface} & \textbf{Not established by the interface alone} \\
    \midrule
    Video generator & Observations and temporal appearance & Images or videos under conditions & Correct interventions or physical quantities \\
    Physics engine & Explicit state and physical dynamics & State propagation, contact, and forces & State inference from open observations \\
    Digital twin & A specified physical counterpart & Monitoring, updating, and prediction & Transferable modeling across entities \\
    Policy model & Goal-conditioned behavior & Actions, skills, or control targets & Independently queryable consequences \\
    World model & Internal state and conditional evolution & Inference, prediction, and queries & A complete agent or controller \\
    \bottomrule
  \end{tabularx}
\end{table}

The state boundary must specify which variables are explicit, which are conditions, and which are unknown disturbances. Precision and horizon are part of this choice. A temperature interval can suffice for a safe-touch query but not for metrology. A short transport may treat material parameters as fixed while a long heating process cannot. Local boundaries also require attention to external inputs. A container can exchange heat and material with its surroundings. Missing measurements should remain marked as missing and should not create unjustified certainty. Claims of generality should state the tested state range, intervention interface, timescale, and query precision.

\subsubsection{Internal State and Belief Updating Under Partial Observation}
\label{sec:wm-belief}

A POMDP distinguishes the environmental state, the actions applied to it, and the measurements available to an agent~\citep{Kaelbling1998}. Let $s_t$ be the physical state at discrete index $t$, $o_t$ a possibly multimodal observation, and $a_t$ the control input applied over the subsequent interval. Let $c$ denote the known embodiment, environment, and sensor conditions. Here a discrete index orders updates. It need not imply equally spaced measurements. The history $h_t=(o_{0:t},a_{0:t-1})$ contains observations and past inputs, with their timestamps and validity metadata understood. The belief over the hidden physical state is the conditional distribution
\begin{equation}
  b_t(s)=p(s_t=s\mid h_t,c).
  \label{eq:wm-belief}
\end{equation}
The argument $s$ is a possible state value. $b_t$ represents uncertainty about which value is current. A learned internal representation $z_t$ is distinct from $s_t$. It may encode a point estimate, parameters of a distribution, or information sufficient for selected future queries. Calling a vector a belief does not by itself give it probabilistic semantics.

Belief maintenance alternates between prediction and correction. For a transition kernel $T$ and an observation likelihood $O$, the Bayesian filtering recursion is
\begin{equation}
\begin{aligned}
  \widetilde b_{t+1}(s')
    &=\int T(s'\mid s,a_t,c,\Delta t)\,b_t(s)\,\mathrm d s,\\
  b_{t+1}(s')
    &=\frac{O(o_{t+1}\mid s',c)\widetilde b_{t+1}(s')}
    {\int O(o_{t+1}\mid u,c)\widetilde b_{t+1}(u)\,\mathrm d u},
\end{aligned}
\label{eq:wm-belief-update}
\end{equation}
where $s'$ is a candidate next state, $u$ a dummy integration variable, $\Delta t$ the elapsed physical time, and $\widetilde b_{t+1}$ the predictive prior before observing $o_{t+1}$. The denominator normalizes the posterior when the observation has nonzero probability. Integrals become sums for discrete states. The recursion assumes a Markov state and an observation model conditionally independent of earlier history given that state and $c$. Delays or actuator memory that violate these assumptions must be included in the state or conditioning record. Learned filters approximate this recursion rather than inheriting its correctness merely from architectural resemblance.

The distinction is concrete during grasping. If the gripper moves upward but the object remains on the table, the observation should revise the attachment hypothesis. If the camera is occluded, lack of visible motion need not mean a failed grasp. A delayed temperature reading belongs to its acquisition time, not its arrival time. Using it may require correcting an earlier state and propagating the correction through intervening actions. Asynchronous processing therefore requires temporal bookkeeping, not compulsory resampling of every sensor to one frequency.

Keeping alternatives separate can be more meaningful than averaging them. The mean of two possible object locations may lie inside an obstacle, and a mean of sticking and sliding contact is not a realizable contact mode. Partial-observation uncertainty concerns the current hidden state. Aleatoric uncertainty concerns variability unresolved at the chosen modeling level. Epistemic uncertainty concerns limited knowledge of model parameters or mechanisms~\citep{wmKendall2017}. These categories can interact, and their decomposition depends on the state and sensor specification. Extra sensing may resolve a hidden contact, whereas collecting diverse interactions may reduce model error. Simply increasing the number of sampled futures addresses neither problem automatically.

\subsubsection{Understanding-Oriented and Prediction-Oriented Objectives}
\label{sec:wm-understanding}

Understanding is best stated as a set of testable abilities, including identifying entities and relations, recognizing operating modes, distinguishing variables that control subsequent change, and identifying what present evidence cannot determine. A fluent explanation is not sufficient. For example, a valve model must distinguish a command, the mechanical position, and the resulting flow. A grasp model must distinguish a closed gripper from a stable attachment. The understanding-prediction distinction in~\citet{wmChen2026Roadmap} provides a useful way to organize these requirements without treating them as mutually exclusive capabilities.

Prediction supplies a complementary pressure. Given a history and conditions, the model should produce a measurable future state, observation, or distribution. Copying a background and extrapolating smooth motion may solve short video prediction while ignoring contact. Longer horizons, action perturbations, and event-level targets test whether the state retains the missing distinctions. Distributional prediction matters when several qualitatively different outcomes remain possible. A single mean trajectory can represent none of them.

Reconstruction is a useful information diagnostic but not proof of understanding. DIAMOND demonstrates that visual detail can affect learned behavior in Atari, illustrating why discarding apparently minor observations may be harmful~\citep{wmAlonso2024}. Conversely, a model can render a convincing container without connecting liquid level, acceleration, and spillage. Decision-oriented abstractions are legitimate when they preserve distinctions needed to compare actions. MuZero learns quantities needed for planning without requiring reconstruction of the original observation, showing that control sufficiency and observation reconstruction are different targets~\citep{wmSchrittwieser2020}. Neither example establishes a universally preferred representation for physical modeling.

Understanding also requires separating reusable relations from accidental correlations. If every heavy training container is red, color may predict lifting success until the correlation changes. A useful representation should allow mass hypotheses to be revised after force feedback without having to change color identity. This example does not require every latent coordinate to have a human name. It requires an intervention-relevant distinction to remain available to downstream queries. Prediction and understanding can thus be mutual checks. An explanation proposes a testable consequence, while a prediction error exposes a missing variable or an inadequate transition.

\paragraph{Representation Learning, Forecasting, and Identification}
These goals differ in the evidence they require. Representation learning can exploit correlations across regions or times. Forecasting must respect the observations available at its issue time. Identifying an intervention response additionally requires the action interface and relevant confounders to be accounted for. A passive video predictor can learn that a human usually opens a drawer after reaching toward it, while still being unable to distinguish a push from a pull applied by a robot. Likewise, fitting a trajectory does not uniquely identify mass and friction when several parameter combinations produce the same motion. An informative probe is one that makes these alternatives predict measurably different outcomes. This connects world modeling to state estimation and system identification, rather than treating physical knowledge as an automatic consequence of video scale.

Learning with a JEPA offers a complementary response to the information problem by predicting selected representations of observations rather than all their low-level details. LeCun's architectural proposal places this idea within a larger system of perception, predictive modeling, memory, and action selection~\citep{lecun2022path}. The reusable principle is selective prediction. Whether the selected information supports causal forecasting, planning, or physical explanation is determined by the conditioning, training data, and readout tests developed below.

\subsubsection{Functional Roles, Query Interfaces, and System Positioning}
\label{sec:wm-position}

The role of a world model is determined by the questions it answers. An observation query produces a sensor-like output. A state query returns objects, relations, or physical quantities. A transition query predicts an action's consequences. And a decision-related query estimates reward, constraint violation, or task success. Figure~\ref{fig:wm-concept} assembles these roles into a reference architecture. A compact conceptual implementation is
\begin{equation}
\begin{aligned}
  z_t&=E_\phi(h_t,c),\qquad
  \widehat z_{t+1}\sim T_\theta(\cdot\mid z_t,a_t,c,\Delta t),\\
  \widehat o_{t+1}&\sim O_{\omega_{\mathrm{obs}}}(\cdot\mid\widehat z_{t+1},c),\qquad
  \widehat y_{t+1}=Q_\psi(\widehat z_{t+1},g,c),
\end{aligned}
\label{eq:wm-observation-prediction}
\end{equation}
where $E_\phi$ is a history encoder, $T_\theta$ a learned transition, $O_{\omega_{\mathrm{obs}}}$ an observation decoder, and $Q_\psi$ a task readout. Their subscripts denote parameter sets, which may overlap in a shared network. The goal or query specification is $g$, $y$ is the requested task quantity, a hat denotes a prediction, and $\sim$ denotes sampling. $Q_\psi$ is a generic query head, not necessarily an action-value function. Eq.~\eqref{eq:wm-observation-prediction} is an interface-level synthesis of state-space and learned-model formulations, not a requirement that these functions be separate networks~\citep{Kaelbling1998,hafner2019planet}.

In the figure, the history encoder $E_\phi$ compresses the observation-action history $h_t$ and known conditions $c$ into the internal state $z_t$, which approximates the belief $b_t(s)$ over the hidden physical state. The prediction-correction update keeps hidden alternatives separate. The learned transition $T_\theta$ then rolls out imagined latent states $\widehat z_{t+1},\ldots,\widehat z_{t+H}$ under a candidate action sequence, and the readouts $O_{\omega_{\mathrm{obs}}}$ and $Q_\psi$ answer observation, state, transition, and decision queries over the inferred state and the imagined rollout. Figure~\ref{fig:wm-concept} also makes two boundaries explicit. The first separates inference from imagination. The internal state is updated from real observations only at the correction step, whereas the rollout is advanced by the learned transition alone and accumulates model error with horizon. The second separates the model from its consumer. Readouts answer queries over the inferred state and the imagined futures, but the scoring of candidate actions and the choice of the applied $a_t$ belong to the agent or planner. A model may support transition and decision queries without an observation decoder. Evaluation should follow the query interfaces it exposes.

This decomposition helps localize failure. A wrong rendered texture with a correct measured trajectory suggests a decoding problem. A wrong contact event suggests a state or dynamics problem. A correct prediction that never influences action suggests a decision-interface problem. Such diagnoses are hypotheses to test, since errors may also propagate across modules.

The world model can sit inside a model-predictive controller, provide training-time imagination to a policy, supply predictive features to an action model, or expose a shared state service to several agents. The surrounding loop remains perception, state inference, prediction, decision, execution, and feedback. Actions change both the world and the observations available next. This connects the chapter to physical data acquisition in Chapter~\ref{ch:physical-data} and to the foundation-model and motor-control interfaces in Chapters~\ref{ch:manipulation-models} to \ref{ch:whole-body-motor-control}. World modeling is one component of Physical AI. Sensing, actuation, safety, and recovery do not follow automatically from a predictive network.

The remaining sections separate four questions concerning how information is represented, how parameters are learned, how predictions are used and verified, and how the full loop can improve. In particular, training with future observations, predicting future observations at runtime, and executing actions with real feedback are different operations. This chapter makes their information boundaries explicit throughout.

\subsection{Representations and Architectures of World Models}
\label{sec:wm-architectures}

No single representation dominates all physical queries. Observation generators retain visible detail, latent dynamics can reduce imagination cost, and structured three-dimensional or object-centric states make entities and relations explicit. These are composable design routes rather than disjoint model classes. Following the distinction between functional and architectural taxonomies in~\citet{wmChen2026Roadmap}, it is useful to ask separately \emph{what is predicted}, \emph{how dynamics are represented}, \emph{which functions share computation}, and \emph{which modules run at deployment}. A diffusion model can operate over pixels, compressed video latents, or object states. A geometric representation can support both rendering and planning. And any of these can be coupled to a policy.

\begin{figure}[tb]
    \centering
    \includegraphics[
        width=\linewidth
    ]{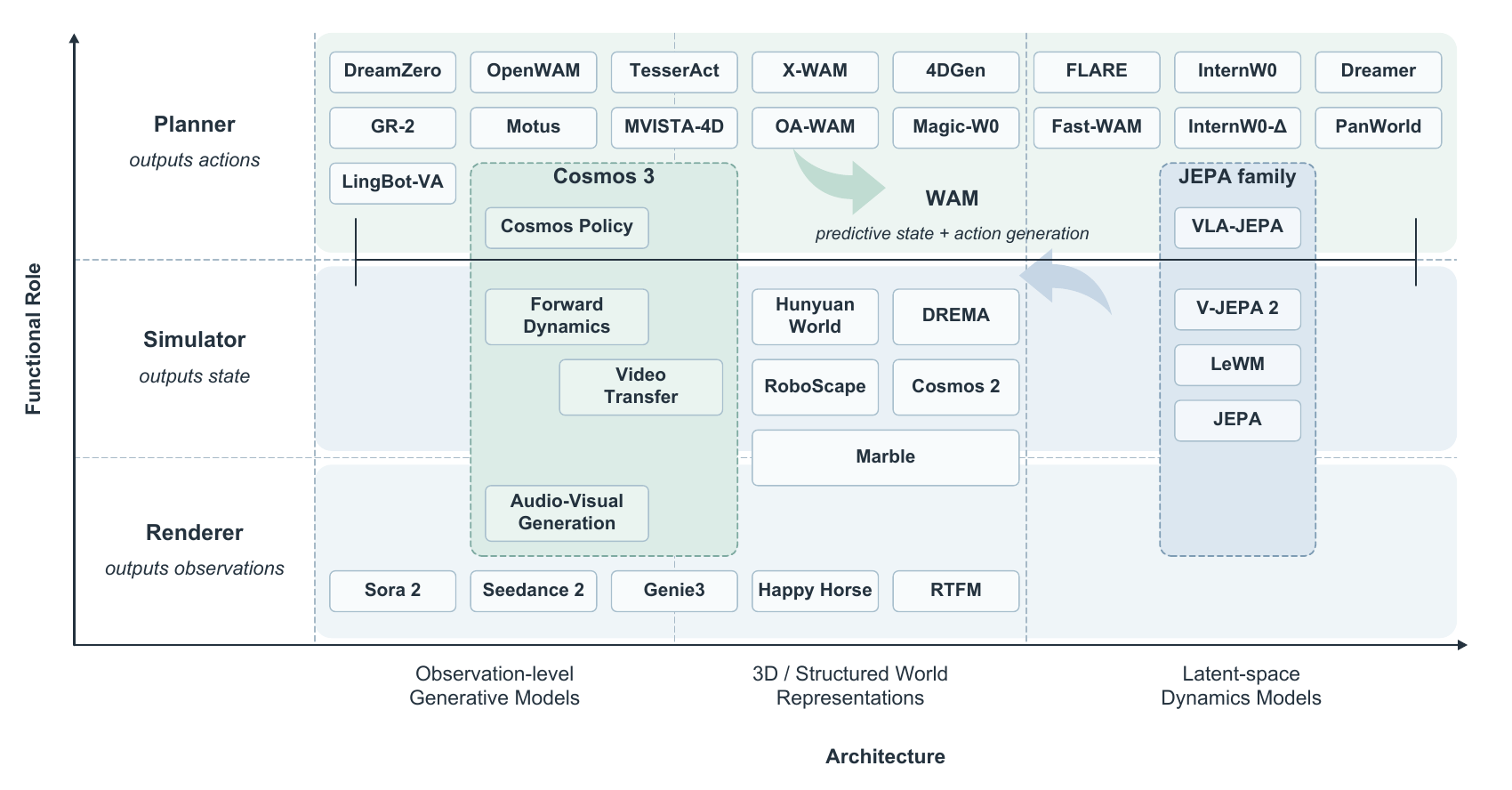}
    \caption{Two-dimensional taxonomy of representative world models. The vertical axis distinguishes functional roles. Renderers produce observations, simulators predict states, and planners produce actions. The horizontal axis distinguishes architectural routes, including observation-level generative models, 3D or structured world representations, and latent-space dynamics models. Spanning regions indicate model families or systems that support multiple roles or combine representational choices. World-action models (WAMs) couple predictive modeling with action generation. Placement is schematic rather than a performance ranking. Planning capability may depend on additional policy or search components}
    \label{fig:wm-two-dimensional-taxonomy}
\end{figure}

Figure~\ref{fig:wm-two-dimensional-taxonomy} organizes representative world models along two complementary dimensions of functional role and architectural representation. The functional axis distinguishes rendering observations, simulating state evolution, and generating actions for decision making. The architectural axis distinguishes observation-level generation, structured world representations, and latent-space dynamics. These categories are not mutually exclusive. A model family may span several functions, and a system may combine multiple representational choices.

\subsubsection{Generative Architectures in Observation Space}
\label{sec:wm-generative}

An observation-oriented model predicts images, video, audio, or other sensor streams. For an externally specified action sequence, a causal autoregressive factorization is
\begin{equation}
  p_\theta(o_{1:N_{\mathrm{step}}}\mid o_0,a_{0:N_{\mathrm{step}}-1},c)
  =\prod_{t=0}^{N_{\mathrm{step}}-1}p_\theta(o_{t+1}\mid o_{0:t},a_{0:t},c),
  \label{eq:wm-observation-factorization}
\end{equation}
where $N_{\mathrm{step}}$ is the final sequence index, $\theta$ the model parameters, and $i{:}j$ an inclusive sequence of indices. The elapsed times are included in $c$ for this expression. Each action is aligned to the interval in which it acts. The equation assumes that future prescribed actions do not affect earlier observations. With actions selected in response to future measurements, conditioning retrospectively on the entire executed action sequence can reveal additional information. Such observational conditioning is not automatically equivalent to an intervention.

\paragraph{Tokenization and Generative Mechanism}
Autoregressive models predict a sequence of discrete or continuous tokens from preceding context. Their ordering makes causal rollout natural, but generating many spatial and temporal tokens can require substantial sequential computation. Diffusion and flow-based models instead generate a frame or chunk through iterative denoising or transport, enabling parallel updates within the chunk while requiring multiple network evaluations. Chunkwise generation combines these choices. Chunks advance causally in physical time, while tokens within each chunk can interact bidirectionally. Physical rollout time and the artificial denoising time are different axes. Section~\ref{sec:wm-representation-training} defines the corresponding training objectives.

An observation-oriented generator need not perform all computation in pixel coordinates. A variational autoencoder (VAE) can compress observations, a generative model can predict the compressed variables, and a decoder can reconstruct the output. Latent diffusion demonstrates this division between perceptual compression and generative computation~\citep{wmRombach2022}. It reduces the number of generated elements but does not, by itself, turn video latents into a sufficient physical state. The architectural distinction from an abstract latent-dynamics model concerns the representation's role and supervision, not whether a tensor is called a latent.

\paragraph{Conditioning and Interaction}
Genie learns latent controls from videos and uses them for interactive generation, whereas UniSim learns conditional real-world video simulation from heterogeneous data~\citep{bruce2024genie,yang2024unisim}. These examples illustrate different intervention interfaces. A latent action inferred from visual changes is not automatically a robot command. Camera motion, actor motion, and object motion can be confounded. Executable control requires a mapping to an embodiment and verification of the resulting response. Language can specify desired behavior, but the same sentence may leave force, speed, contact, and timing unspecified.

Observation generation is useful when detail matters for perception, when imagined data will train a visual policy, or when a person must inspect a predicted future. Its limitations include high spatial cost, temporal drift, and the possibility of replacing a wrong state with a plausible texture. A video of a drawer opening should be checked for the commanded displacement, latch state, object identity, and occluded contents, not only visual sharpness. Multiple rendered views are valuable only when they describe a compatible underlying scene. These requirements motivate persistent latent states, geometric structure, and action-grounded supervision rather than ruling out generative prediction.

\paragraph{Broad Video Priors and Spatially Indexed Memory}
Cosmos illustrates how pretrained video models, tokenizers, and data processing can form a reusable platform for physical-AI adaptation~\citep{nvidia2025cosmos}. Such a prior can supply appearance and motion regularities before task-specific action data are available. Its contribution is different from learning the calibrated response of a particular actuator. Downstream conditioning and validation must establish that interface. This separation between broad pretraining and local grounding remains useful as individual video backbones change.

Persistent visual context can also be organized spatially rather than only as a recent temporal window. World Labs' RTFM uses an autoregressive diffusion frame model with camera poses and retrieval of relevant posed frames as spatial memory~\citep{wmWorldLabs2025RTFM}. It illustrates a learned rendering route that need not construct an explicit mesh at every update. Revisiting a place queries stored evidence about that place rather than forcing an ever-growing sequence through a fixed context window. Camera-conditioned view generation, however, primarily specifies how observation changes with viewpoint. Controlling a camera and predicting an object's response to force are distinct interfaces. This distinction is important when a navigable generated world is later used for interaction.

\subsubsection{Latent-Space Dynamics Architectures}
\label{sec:wm-latent}

Latent models encode observation history and propagate a compact state. A generic form is
\begin{equation}
\begin{aligned}
  z_t&=E_\phi(h_t,c),\qquad \widehat z_t=z_t,\\
  \widehat z_{t+k+1}&\sim T_\theta(\cdot\mid\widehat z_{t+k},a_{t+k},c,\Delta t_k),\\
  \widehat y_{t+k+1}&=Q_\psi(\widehat z_{t+k+1},g,c),\quad k=0,\ldots,H-1,
\end{aligned}
\label{eq:wm-latent-dynamics}
\end{equation}
where $H$ is the rollout horizon, $\Delta t_k$ the duration of its $k$th interval, $E_\phi$ the history encoder, and $Q_\psi$ the query readout. The symbols $z$ and $\widehat z$ distinguish inferred and imagined states. $y$ may be an observation, reward, physical variable, or event. The readout can be omitted during intermediate rollout steps. A useful state preserves the conditional distribution of the intended queries. Histories merged into one representation should not demand incompatible answers after the same future input. A recently heated vessel and a long-cooled one may look alike while requiring different temperature forecasts.

\paragraph{Generative State-Space Models}
PlaNet combines deterministic recurrent memory with stochastic latent variables, while Dreamer uses learned latent dynamics to train an actor and critic through imagination~\citep{hafner2019planet,hafner2019dreamer}. A recurrent component can retain predictable history, and stochastic variables can represent outcomes that remain uncertain. Training distinguishes a prior predicted without the next observation from an approximate posterior corrected by that observation. Reconstruction and auxiliary targets encourage informative states. Prior-posterior regularization makes them predictable. Overly strong regularization can suppress observation information, whereas a weakly trained prior can fail when real frames are absent. The loss and its multi-step extensions are developed in Section~\ref{sec:wm-training}.

\paragraph{JEPA: Predicting in a Learned Target Space}
A JEPA relates a context representation to a target representation through a predictor. Its defining choice is the prediction target, not a particular recurrent layer, transformer size, or policy head. I-JEPA learns image representations by predicting target-region embeddings from visible context. It is a spatial representation learner, not by itself an action-conditioned temporal simulator~\citep{assran2023ijepa}. V-JEPA extends feature prediction to masked spatiotemporal regions of video, using a predictor and a slowly updated target encoder rather than a pixel-reconstruction objective~\citep{bardes2024vjepa}. In both cases, what is hidden and what remains visible determine whether the task teaches spatial completion, temporal interpolation, or future anticipation.

V-JEPA~2 connects large-scale action-free representation learning to a separately post-trained action-conditioned latent predictor. Its planning variant uses robot observations, proprioception, and recorded actions to predict future features, then evaluates candidate actions against an image goal in representation space~\citep{assran2025vjepa2}. Recorded interactions ground the predictor in robot actions. Self-supervision constructs targets from these records without task-success annotations or a hand-designed reward for each training example.

The architectural advantage is that a predictor can avoid synthesizing every unpredictable texture while retaining information useful for recognition or control. The corresponding obligation is to show that contacts, relative motion, object identity, and material cues have survived the target representation. A deterministic embedding forecast need not describe the distribution of possible futures, and its distance to a goal need not be a calibrated measure of physical reachability. Observation reconstruction and predictive embeddings are therefore complementary choices. A model can learn from both, or attach an optional decoder without making rendering a prerequisite for each decision. Section~\ref{sec:wm-representation-training} develops the target update and loss, while Section~\ref{sec:wm-prediction-validation} discusses tests that do not privilege models with a visual decoder.

\paragraph{Decision-Oriented Abstraction and State Maintenance}
MuZero and TD-MPC-style systems emphasize reward, value, or latent consistency sufficient for decision making rather than unrestricted observation queries~\citep{wmSchrittwieser2020,wmHansen2024}. DreamerV3 retains a generative latent model while improving robustness of model and behavior learning across different domains through normalization and loss balancing~\citep{wmHafner2023DreamerV3}. These examples distinguish the choice of representation from the numerical techniques needed to train it reliably. A small state is efficient only if it preserves the distinctions required by the objective. Changing the goal or downstream query can expose previously harmless information loss.

Latent coordinates need not correspond one-to-one to physical quantities. A feature from which mass can be decoded may still entangle mass with shape, and an arbitrary coordinate edit need not realize a meaningful mass intervention. A history encoder should also distinguish an observation-corrected state from one propagated without new evidence. Its memory may be recurrent, attention-based, spatially retrieved, or hierarchically summarized. What matters is whether the representation preserves consequential hidden alternatives through occlusion and updates them when new evidence arrives. Tests should compare posterior updates, pure open-loop prediction, and feedback-interleaved prediction at the same physical horizons.

\subsubsection{3D, 4D, and Object-Centric Architectures}
\label{sec:wm-structured}

Structured states make geometry, entities, and relations directly queryable. They can use occupancy or bird's-eye grids, object slots, scene graphs, particles, meshes, or continuous fields. Their relation to world modeling depends on the associated transition. A static 3D reconstruction supports novel views but does not specify what a push will do. A 4D reconstruction describes temporal observations but may not predict a new intervention. The spatial perception methods in Chapter~\ref{ch:perception} become a world-model substrate when they are connected to evolving attributes, interactions, and conditional prediction.

Object-centric methods allocate separate representations to entities and model their interactions. Slot Attention provides a mechanism for learning object-centric representations from observations~\citep{wmLocatello2020}. Graph-network simulators and MeshGraphNets illustrate dynamics expressed through local interactions among particles or mesh elements~\citep{sanchezgonzalez2020learning,wmPfaff2021}. These structures can make interaction locality and changes in object count easier to express than an undifferentiated global vector. However, the structure is an inductive bias, not a guarantee. Learned slots may exchange identities, an object may split or merge, and contact can change the interaction graph discontinuously.

Geometry and physical attributes should be distinguished. Shape alone does not identify mass, friction, stiffness, temperature, or internal contents. A deformable state may need both configuration and a constitutive description. A liquid requires boundary conditions and appropriate conservation accounting. PhysGaussian couples Gaussian-based rendering with continuum-mechanics simulation, illustrating how visual representation and physical evolution can share a substrate within a specified regime~\citep{wmXie2024}. The same representation must still be supplied with suitable physical parameters and verified against the intended materials and interactions.

Hybrid representations can assign different resolutions to different objects or processes. A rigid container may be represented as an object, while its contents are a field or a small set of effective variables. A coarse query could ask whether spilling is likely. A fine query might require the flow near an outlet. Expanding the representation only where the query requires it is a plausible way to balance cost and detail. The levels must share coordinate frames, identity, units, boundary fluxes, and physical time so that a coarse state is compatible with a refined prediction.

The structured route is attractive for metric queries, contact reasoning, and compositional changes. It introduces costs in reconstruction, correspondence, annotation, and numerical integration, and it may inherit a simulator's modeling assumptions. A representation built around rigid objects can fail on cloth or fragmentation. A fixed field resolution can miss thin contacts. Validation should test identity through occlusion, changed topology, altered parameters, and action-dependent response. Structured models can be combined with learned latent residuals and generative observation heads, rather than being treated as alternatives to neural representation learning.

\paragraph{Hierarchies, Fields, and Physical-Prediction Structure}
The Hierarchical Relation Network (HRN), developed by Mrowca and colleagues, represents bodies with hierarchies of particles and learns their interactions through graph-based computation~\citep{wmMrowca2018}. Its lasting contribution is the separation of local interactions from larger-scale object structure, allowing rigid and deformable dynamics to share a representation without pretending that all objects are single rigid slots. Hierarchical modeling can reduce the distance over which information must propagate, but the useful grouping and coupling still depend on the physical process. The related Physion evaluation program tests future object contact across varied physical scenarios and distinguishes learning from images from prediction with privileged state~\citep{wmBear2021Physion}. It provides an empirical counterpart to the architectural question of which state distinctions are needed to forecast an event.

Continuous-field models broaden the scope beyond articulated robots and discrete objects. Fourier neural operators learn mappings between parameter or input fields and solution fields for families of partial differential equations~\citep{wmLi2021FNO}. They can serve as components for fluid, thermal, or other distributed-process predictions. A field-to-solution surrogate need not include perception, uncertainty, or an action policy. It supplies one transition or query component of a larger world model. Generalization across spatial resolution also differs from generalization across boundary conditions or physical regimes. Coupling a learned field solver to objects requires consistent interfaces for forces, fluxes, time, and units.

\paragraph{Persistent Spatial Generation and Multimodal Reconstruction}
The spatial-intelligence route associated with World Labs should be distinguished from, rather than conflated with, the physical-dynamics work above. Marble generates persistent, editable 3D environments from multimodal prompts and supports explicit scene outputs~\citep{wmWorldLabs2025Marble}. It illustrates the value of a scene that can be revisited and queried from different viewpoints, instead of a sequence whose geometry is reconstructed anew at every frame. The public Atlas description further combines text, posed images, and depth within a shared spatial context using an autoregressive diffusion architecture, with reconstruction and space-time generation interfaces~\citep{worldlabs2026atlas}. These are developer-described systems. Their available interfaces should not be expanded into unsupported claims about arbitrary contact dynamics or physical precision.

The underlying distinction is between reconstructing evidence, completing unobserved regions, and evolving a state under interventions. A model may do all three, but their outputs carry different evidential status. Adding real views can constrain a previously imagined region. Creating a visually plausible hidden room does not measure that room. An explicit 3D scene can supply geometry for a simulator while leaving collision properties, mass, and material parameters to be assigned or identified. Conversely, an implicit renderer can maintain useful spatial consistency without exposing a simulation-ready mesh. The choice of explicit or implicit representation should therefore follow the required queries, not a blanket identification of 3D output with physical understanding.

\subsubsection{Unified Multimodal and Cross-Paradigm Architectures}
\label{sec:wm-unified}

Unification concerns several different design decisions, including sharing an input interface, sharing a representation, learning multiple objectives together, and executing multiple functions in one runtime. A system may unify one level while retaining separate modules at another. Describing the level of unification is more informative than calling the entire architecture unified.

\paragraph{Multimodal Evidence and Representation Sharing}
Vision, touch, proprioception, sound, temperature, and process readings can update a common state while retaining modality-specific noise and timing. Fusion can occur at the input, in intermediate state representations, or at query time. Early fusion preserves rich interactions but couples token budgets and preprocessing. Separate encoders with state-level exchange preserve sensor-specific processing while introducing alignment interfaces. Query-time fusion is modular but can lose information already discarded by each encoder. In a lifting task, disagreement between vision and torque may indicate an incorrect mass hypothesis, a grasp failure, or a sensor fault. Averaging the signals into an unexplained scalar is not a solution.

Understanding and generation also have different representational needs. Show-o combines autoregressive and diffusion-based modeling within a unified transformer. BAGEL studies shared multimodal understanding and generation with specialized representational pathways~\citep{wmXie2025Showo,wmDeng2025}. These attempts show that shared processing need not require identical tokenizers or identical prediction targets. They motivate joint semantic and generative learning, but successful image understanding and synthesis do not independently verify a model's response to physical interventions.

\paragraph{Coupling Future Prediction and Action Generation}
Let $A_{t}^{\mathrm{seq}}=a_{t:t+H_{\mathrm{a}}-1}$ denote an action chunk of length $H_{\mathrm{a}}$, and let $F_{t}^{\mathrm{future}}=o_{t+1:t+H_{\mathrm{o}}}$ denote future observations over $H_{\mathrm{o}}$ observation steps. A statistical joint model admits the factorizations
\begin{equation}
\begin{aligned}
 p_\Theta(A_{t}^{\mathrm{seq}},F_{t}^{\mathrm{future}}\mid h_t,g,c)
 &=p_\Theta(A_{t}^{\mathrm{seq}}\mid h_t,g,c)\,
   p_\Theta(F_{t}^{\mathrm{future}}\mid A_{t}^{\mathrm{seq}},h_t,g,c)\\
 &=p_\Theta(F_{t}^{\mathrm{future}}\mid h_t,g,c)\,
   p_\Theta(A_{t}^{\mathrm{seq}}\mid F_{t}^{\mathrm{future}},h_t,g,c),
\end{aligned}
\label{eq:wm-joint-factorization}
\end{equation}
where $\Theta$ denotes the model's parameters. The action and observation horizons must refer to compatible physical intervals even when their sample counts differ. The first factorization suggests action proposal followed by consequence prediction. The second suggests a future subgoal followed by inverse-dynamics action generation. An observed conditional distribution is not automatically an interventional model, and an attractive generated goal is not necessarily reachable.

Unified World Models (UWM) couples video and action diffusion with independent modality-specific noise levels, using the conditioning pattern to obtain policy, forward-dynamics, inverse-dynamics, and video-generation interfaces~\citep{zhu2025uwm}. This illustrates how the same parameter set can support different queries when the training distribution and masks explicitly cover those queries. It does not imply that every joint generator reliably supports every conditional query. A cascaded design is easier to inspect but can compound errors and latency. A jointly sampled design enables stronger coupling but must control incompatible objectives and future-information leakage.

DreamZero supplies another representative use of a pretrained video prior for joint future-video and action modeling~\citep{ye2026dreamzero}. Together with UWM, it illustrates a broad design axis in which action supervision can shape a model of anticipated observations instead of being added only to a static visual encoder. The systems need not share the same architecture, sampling procedure, or adaptation protocol. OpenWAM's composable framework makes this distinction explicit by separating prediction, action generation, their interaction structure, and inference choices for controlled comparison~\citep{wang2026openwam}. Its methodological lesson is to compare one interface choice at a time, rather than attributing every difference between complete systems to ``world modeling.''

A third route uses future supervision to shape a representation available from present evidence, without generating future observations during deployment. InternVLA-A1.5 trains foresight tokens through a frozen video model and removes the video branch at inference~\citep{wmMa2026A15}. InternW0-$\Delta$ combines a video expert, an action expert, frozen VLM semantics, and training-only 4D distillation. Causal Imprint connects future-derived supervision to observation-conditioned action features~\citep{wmMiao2026Delta}. Such systems unify predictive learning with action generation without requiring a full future rollout for every policy call.

\paragraph{Expert Specialization and Runtime Organization}
A mixture-of-transformers can retain separate parameters and token streams while allowing masked attention between experts. The mask specifies which observations, predicted variables, and action tokens can influence one another. Shared attention does not imply unrestricted bidirectional access. A cache can be reused across action-denoising steps only if its inputs and all upstream dependencies remain unchanged. Architectural independence, gradient isolation during training, and asynchronous execution are three distinct properties.

InternW0 illustrates a complementary runtime design in which slower video prediction supplies cached context, and a faster action expert adapts that context to recent observations through an observation-conditioned routing interface~\citep{wmCai2026InternW0}. This differs from training away the future generator. In InternW0-$\Delta$, future-video tokens are omitted at inference, but the video expert still processes observed context and Causal Imprint once before action-only denoising~\citep{wmMiao2026Delta}. Neither arrangement should be summarized simply as ``the video module is free.'' Their evidence boundaries are taken up again when decision utility is evaluated in Section~\ref{sec:wm-decision-validation}.

\paragraph{Comparative Strengths and Composable Choices}
Table~\ref{tab:wm-architectural-routes} compares the four routes by what they make convenient, what they cost, and what must be validated. The entries describe tendencies, not universal performance rankings. Actual cost depends on backbone size, representation resolution, number of sampling steps, and deployment implementation.

\begin{table}[tbp]
  \centering
  \caption{Architectural routes, practical trade-offs, and principal validation questions. The routes are composable. Multimodal unification is also a cross-cutting organization choice}
  \label{tab:wm-architectural-routes}
  \small
  \renewcommand{\arraystretch}{1.14}
  \begin{tabularx}{\textwidth}{@{}>{\raggedright\arraybackslash}p{0.16\textwidth} >{\raggedright\arraybackslash}X >{\raggedright\arraybackslash}X >{\raggedright\arraybackslash}X@{}}
    \toprule
    \textbf{Route} & \textbf{Useful interface / strength} & \textbf{Main cost or limitation} & \textbf{Critical validation} \\
    \midrule
    Observation generation & Detailed, inspectable futures and synthetic sensor data & Large output budget, decoding and sampling cost, plausible but wrong detail & Action-conditioned change, identity, and long-horizon consistency \\
    Latent dynamics & Compact rollout, inexpensive task readout, planning & Objective-dependent information loss, latent drift, limited new queries & Predictive sufficiency, open-loop stability, and calibrated readouts \\
    3D / object-centric & Metric geometry, entities, and local interactions & Reconstruction and correspondence, structural and simulator assumptions & Contact, topology, identity, physical parameters, and interventions \\
    Multimodal / unified & Shared evidence and priors, multiple query interfaces & Objective interference, alignment and missing-data issues, runtime coupling & Information boundaries, cross-query agreement, missing modalities, and total cost \\
    \bottomrule
  \end{tabularx}
\end{table}

A design can therefore combine an object-centric state, latent stochastic transitions, an optional video decoder, and a policy trained from future supervision. The right combination is determined by the required queries and evidence. Cross-query consistency is useful. Rendering a state and re-estimating it should preserve relevant quantities. However, two heads can share the same bias, so their agreement is not independent physical validation. Section~\ref{sec:wm-outlook} returns to hierarchical, multi-timescale, and selectively activated architectures as future directions rather than treating a single fully shared backbone as inevitable.

\paragraph{Reading Representative Work Through Common Design Questions}
Table~\ref{tab:wm-representative-lineages} summarizes representative implementations of the preceding principles. It is a map of mechanisms, not an exhaustive catalog or a ranking. A new model can be located by its retained state, supervisory target, intervention interface, and deployed computation without adding a new supposedly disjoint category for every paper.

\begin{table}[tbp]
  \centering
  \caption{Representative lineages organized by enduring modeling questions. Examples instantiate the mechanisms discussed in the text. They do not share one evaluation protocol}
  \label{tab:wm-representative-lineages}
  \small
  \renewcommand{\arraystretch}{1.12}
  \begin{tabularx}{\textwidth}{@{}>{\raggedright\arraybackslash}p{0.21\textwidth} >{\raggedright\arraybackslash}X >{\raggedright\arraybackslash}X@{}}
    \toprule
    \textbf{Underlying question} & \textbf{Representative work} & \textbf{Mechanism to compare} \\
    \midrule
    How is history compressed for control? & World Models, PlaNet, Dreamer & State inference, recurrent dynamics, and imagination \\
    Which predictions are sufficient for decisions? & MuZero, TD-MPC2 & Reward/value and task-relative latent consistency \\
    Can useful futures be predicted without pixels? & I-JEPA, V-JEPA, V-JEPA~2 & Target embeddings, masking, and action grounding \\
    How are interactive observations generated? & DIAMOND, Genie, UniSim, Cosmos & Generative targets, controllable inputs, and rollout memory \\
    How are spatial persistence and reconstruction maintained? & Marble, RTFM, Atlas & Explicit scenes or posed-view memory\newline Measured versus completed content \\
    How is physical structure exploited? & HRN, graph/mesh simulators, PhysGaussian, neural operators & Entities, local interactions, physical priors, and fields \\
    How are prediction and action coupled? & UWM, DreamZero, InternVLA-A1.5, InternW0, InternW0-$\Delta$ & Joint generation, predictive features, and training-runtime separation \\
    \bottomrule
  \end{tabularx}
\end{table}

The table also exposes missing interfaces. An image representation learner still needs a temporal, action-conditioned predictor to support control queries. A persistent scene generator still needs dynamics for manipulation. And a successful policy still needs separate tests before being used as a general predictive simulator. These are opportunities for composition, not reasons to exclude the corresponding lines of research from world modeling.

\subsection{Training World Models}
\label{sec:wm-training}

Training must make the internal state useful for future queries rather than merely reconstruct familiar observations. A complete account specifies the data available to each objective, the variables predicted, the parameters updated, the rollout horizon, and the information excluded from deployment. Representation learning, dynamics learning, policy learning, physical constraints, and uncertainty estimation solve related but different problems. Their combination is a design choice, not one mandatory loss shared by all world models.

\begin{figure}[!htb]
  \centering
  \includegraphics[width=\linewidth]{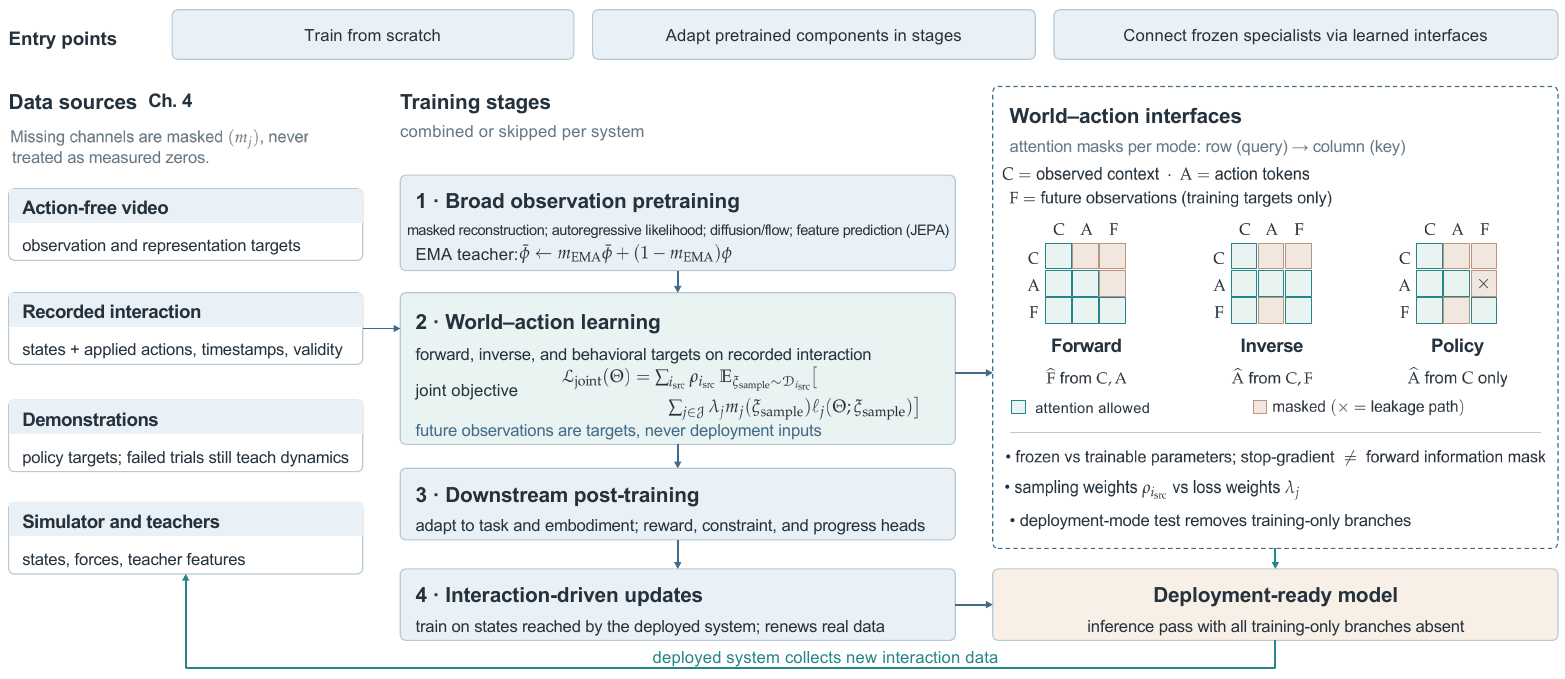}
  \caption{Training world models. Three entry points (top) lead into a staged training pipeline (center), fed by the data sources of Chapter~\ref{ch:physical-data} (left). Interaction-driven updates retrain on the states reached by the deployed system, closing the loop through new data. The right panel defines the world-action interfaces as attention masks over observed-context ($\mathrm{C}$), action ($\mathrm{A}$), and future-observation ($\mathrm{F}$) tokens}
  \label{fig:wm-training}
\end{figure}

A practical training program can start from scratch, adapt pretrained components in stages, or connect frozen specialists through learned interfaces. Broad observation pretraining supplies reusable visual and temporal structure. Action-conditioned training grounds intervention semantics. Downstream post-training adapts the model to a task or embodiment. Interaction-driven updates address the states reached by the deployed system. Some systems combine these stages, and some do not require all of them. PlaNet illustrates joint latent-model learning, V-JEPA~2 separates broad representation learning from action-conditioned post-training, and UWM combines predictive and behavioral supervision within a shared generative framework~\citep{hafner2019planet,assran2025vjepa2,zhu2025uwm}. Figure~\ref{fig:wm-training} summarizes this landscape by showing the entry points along the top, the data sources on the left, the four training stages in the center, and the world-action interfaces on the right.

The data foundations are developed in Chapter~\ref{ch:physical-data}. Their training consequences differ. Action-free videos can supervise observation or representation prediction. Recorded controls can supervise action-conditioned transitions. Demonstrations can supervise a policy. And simulator states, force readings, or teacher features provide additional targets with distinct provenance. A valid failed trajectory can teach dynamics and outcomes without being an appropriate nominal imitation target. Missing channels must be masked rather than treated as measured zeros.

The four stages in the center of Figure~\ref{fig:wm-training} carry different objectives. Broad observation pretraining applies masked reconstruction, autoregressive likelihood, diffusion or flow, and feature-prediction targets, with an exponential-moving-average teacher where the target representation must stay stable. World-action learning adds forward, inverse, and behavioral targets on recorded interaction. Downstream post-training adapts the model to a task and embodiment. And interaction-driven updates retrain on the states the deployed system actually reaches. That last stage closes the loop only through new data, which is what distinguishes continual improvement from repeated fitting of a fixed corpus. In the right panel, the forward mode predicts $\widehat{\mathrm{F}}$ from $\mathrm{C}$ and $\mathrm{A}$, the inverse mode predicts $\widehat{\mathrm{A}}$ from $\mathrm{C}$ and $\mathrm{F}$, and the policy mode predicts $\widehat{\mathrm{A}}$ from $\mathrm{C}$ only, with the future-to-action path masked out as a leakage route (red cross). In all three modes the context rows are barred from reading $\mathrm{F}$, so the context representation remains a pure function of observed evidence even when the action query may consult the future. These interfaces are reading modes over the same token groups, selected by attention masks, rather than three separate models. One network can be trained to serve all of them provided the masks are enforced exactly, as Section~\ref{sec:wm-action-training} discusses. The deployment-ready model is then an inference pass with all training-only branches absent.

\subsubsection{Self-Supervised and Generative Representation Learning}
\label{sec:wm-representation-training}

Self-supervised objectives extract structure from observations without requiring every physical variable to be annotated. Temporal prediction, masked reconstruction, contrastive discrimination, and target-representation prediction impose different information bottlenecks. A reconstruction target retains observable detail. A feature target retains distinctions present in its encoder. A temporal target encourages information predictive over its selected horizon. None automatically identifies mass, contact, or causality when the observations do not distinguish them.

\paragraph{Autoregressive Likelihood and Masked Reconstruction}
Let $x=(x_1,\ldots,x_{N_{\mathrm{tok}}})$ be a target sequence of $N_{\mathrm{tok}}$ tokens, $\kappa$ the available conditioning information, and $\mathcal D$ the training distribution. A maximum-likelihood objective is
\begin{equation}
  \mathcal L_{\mathrm{AR}}
  =-\mathbb E_{(x,\kappa)\sim\mathcal D}
    \left[\sum_{i=1}^{N_{\mathrm{tok}}}\log p_\theta(x_i\mid x_{<i},\kappa)\right].
  \label{eq:wm-ar-training}
\end{equation}
The index $i$ orders tokens, $x_{<i}$ contains preceding target tokens, and $\theta$ denotes trainable parameters. The conditioning can include past observations, a goal, and supplied actions, depending on the advertised query. Discrete tokens yield cross-entropy training. Continuous outputs require an explicit conditional density. During teacher-forced training, preceding targets are real tokens, whereas free generation conditions on the model's own outputs. Thus a low teacher-forced loss does not establish stable recursive prediction.

Masked reconstruction instead withholds selected spatial or temporal observations and predicts them from visible context. A mask that reveals observations on both sides of a hidden frame trains interpolation or smoothing, not causal future prediction. Such supervision can still be valuable for representation pretraining, but a causal predictor needs a corresponding future-blind training or adaptation stage. The mask pattern, temporal window, and conditioning availability should be recorded alongside the loss.

\paragraph{Diffusion and Flow-Matching Objectives}
For a clean continuous target $\mathbf{x}$, a noise-prediction diffusion loss takes the form
\begin{equation}
\begin{aligned}
 \mathbf{x}^{(j)}&=\sqrt{\bar\alpha_j}\,\mathbf{x}+\sqrt{1-\bar\alpha_j}\,\boldsymbol{\epsilon},\qquad
 \boldsymbol{\epsilon}\sim\mathcal N(0,\mathbf{I}),\\
 \mathcal L_{\mathrm{diff}}
 &=\mathbb E_{\mathbf{x},\kappa,j,\boldsymbol{\epsilon}}
   \left[w_j\left\|\boldsymbol{\epsilon}_\theta(\mathbf{x}^{(j)},j,\kappa)-\boldsymbol{\epsilon}\right\|_2^2\right],
\end{aligned}
\label{eq:wm-diffusion-training}
\end{equation}
where $j$ indexes artificial noise levels, $\bar\alpha_j$ is the retained signal fraction, $\mathbf{I}$ the identity covariance, $w_j$ a nonnegative loss weight, and $\boldsymbol{\epsilon}_\theta$ the learned noise predictor. The expectation samples targets, conditions, noise levels, and Gaussian noise. This is a conditional adaptation of denoising diffusion training~\citep{wmHo2020}. The physical timestep remains distinct from $j$. Predicting clean data or a different noise-data combination changes the parameterization and weighting, not the underlying requirement to learn a conditional distribution.

A simple flow-matching construction uses the following linear data-to-noise path
\begin{equation}
\begin{aligned}
 \mathbf{x}^{\sigma_{\mathrm{flow}}}&=(1-\sigma_{\mathrm{flow}})\mathbf{x}+\sigma_{\mathrm{flow}}\boldsymbol{\epsilon},\qquad \sigma_{\mathrm{flow}}\sim\operatorname{Unif}[0,1],\\
 \mathcal L_{\mathrm{FM}}
 &=\mathbb E_{\mathbf{x},\kappa,\sigma_{\mathrm{flow}},\boldsymbol{\epsilon}}
  \left[\left\|\mathbf{v}_{\theta}^{\mathrm{flow}}(\mathbf{x}^{\sigma_{\mathrm{flow}}},\sigma_{\mathrm{flow}},\kappa)-(\boldsymbol{\epsilon}-\mathbf{x})\right\|_2^2\right],
\end{aligned}
\label{eq:wm-flow-training}
\end{equation}
where $\sigma_{\mathrm{flow}}$ is continuous artificial time, $\operatorname{Unif}[0,1]$ the uniform distribution, and $\mathbf{v}_{\theta}^{\mathrm{flow}}$ a learned velocity field in the target space. The target velocity is $\boldsymbol{\epsilon}-\mathbf{x}$ under this convention. Sampling starts at noise and integrates from $\sigma_{\mathrm{flow}}=1$ toward $0$. Reversing the path convention reverses the velocity sign. This is a representative conditional flow-matching objective, not a unique choice of path or solver~\citep{wmLipman2023}.

The target $\mathbf{x}$ can be a future-video latent, an action chunk, or a structured physical quantity. Video and action branches can have separate noise levels and masks, as in UWM~\citep{zhu2025uwm}. This allows examples with missing action labels to train the video objective without fabricating control supervision. Tokenization and normalization determine how the squared loss weights different channels. An angular coordinate, translation, and contact force should not compete merely according to their raw numerical scales. Fewer inference steps lower cost but introduce a solver and approximation trade-off that must be measured at the task interface.

\paragraph{Stochastic State-Space Representation Learning}
A generative latent model must learn both an observation-informed posterior and a prior that can advance without new observations. For clarity, consider a fixed initial latent $z_0$, a Markov prior $p_\theta(z_t\mid z_{t-1},a_{t-1},c)$, an observation likelihood $O_{\omega_{\mathrm{obs}}}(o_t\mid z_t,c)$, and a filtering variational distribution $q_\phi(z_t\mid z_{t-1},h_t,c)$. A representative sequence objective is
\begin{equation}
\begin{aligned}
 \mathcal L_{\mathrm{SSM}}
 =\mathbb E_{\mathcal D}\sum_{t=1}^{N_{\mathrm{step}}}
 \Big[&\mathbb E_{q_\phi(z_{1:t}\mid h_t,c,z_0)}
       [-\log O_{\omega_{\mathrm{obs}}}(o_t\mid z_t,c)]\\
 &+\beta_{\mathrm{KL}}\,\mathbb E_{q_\phi(z_{1:t-1}\mid h_{t-1},c,z_0)}
     D_{\mathrm{KL}}\!\left(
       q_\phi(z_t\mid z_{t-1},h_t,c)
       \,\|\,p_\theta(z_t\mid z_{t-1},a_{t-1},c)
     \right)\Big].
\end{aligned}
\label{eq:wm-ssm-training}
\end{equation}
The variational sequence distribution is the product of its filtering factors. $D_{\mathrm{KL}}$ is Kullback-Leibler divergence, and $\beta_{\mathrm{KL}}$ weights prior-posterior agreement. Elapsed times are suppressed in $c$. With $\beta_{\mathrm{KL}}=1$, this is the negative evidence lower bound for the stated model, conditional on $z_0$. Learning an uncertain initial state adds its corresponding initial term. Other $\beta_{\mathrm{KL}}$ values yield a regularized objective rather than the same exact likelihood bound. The construction combines variational learning with sequential latent dynamics~\citep{wmKingma2014,hafner2019planet}.

The reconstruction term trains the encoder and decoder to retain evidence. The divergence trains the prior to match the observation-informed posterior while limiting information that the prior cannot predict. Reparameterized sampling allows gradients through continuous latent variables. In practice, balancing or stopping selected gradient paths, protecting a minimum latent information budget, and adding reward or event targets can stabilize learning. Their details should be reported, because identical symbolic losses can optimize different parameters under different gradient rules. A deterministic recurrent memory may be included alongside the stochastic state, provided its dependencies are included in the model definition.

\paragraph{Predicting Target Representations}
For a context encoder $E_\phi$, predictor $P_\theta$, and target encoder $E_{\bar{\phi}}$, a generic joint-embedding objective is
\begin{equation}
 \mathcal L_{\mathrm{repr}}
 =\mathbb E_{\mathcal D,\mathcal M_{\mathrm{mask}}}
  \left[d_{\mathrm{feat}}\!\left(P_\theta(E_{\phi}(x_{\mathrm{vis}}),\mathcal M_{\mathrm{mask}},\kappa),
          \operatorname{sg}[E_{\bar{\phi}}(x_{\mathrm{target}})]\right)\right],
 \label{eq:wm-representation-training}
\end{equation}
where $\mathcal M_{\mathrm{mask}}$ specifies target positions or masked regions, $x_{\mathrm{vis}}$ the visible context, $x_{\mathrm{target}}$ the withheld target, $d_{\mathrm{feat}}$ a feature distance, and $\operatorname{sg}$ the stop-gradient operator. The target parameters $\bar{\phi}$ can be frozen or updated by an exponential moving average of $\phi$, rather than by backpropagation through this loss. I-JEPA and V-JEPA~2 instantiate related predictive-embedding mechanisms at different spatial and temporal scales~\citep{assran2023ijepa,assran2025vjepa2}.

The predictor and context encoder learn to make retained target information predictable. Stop-gradient alone is not a universal proof against collapse. The target update, masking, architecture, and representation statistics must work together. A collapsed representation can achieve a trivial small distance, and a semantic target can discard fine motion. Comparing downstream readouts, representation variation, and controlled physical predictions is therefore necessary. The main trade-off concerns which information is retained, how expensive the target is to produce, and which later queries the representation supports.

\paragraph{From Masked JEPA Pretraining to Causal Latent Dynamics}
A typical training iteration first selects a clip and its visible and target regions. The context encoder receives only the visible content. The target encoder processes the target evidence, possibly with the wider context specified by the method, and produces the features to be matched. The predictor is conditioned on the target positions and, when the task requires them, actions and elapsed time. After the student update, an exponential-moving-average teacher can be updated as
\begin{equation}
 \bar{\phi}\leftarrow m_{\mathrm{EMA}}\bar{\phi}+(1-m_{\mathrm{EMA}})\phi,\qquad 0\leq m_{\mathrm{EMA}}<1,
 \label{eq:wm-ema-update}
\end{equation}
where $\phi$ is the updated student-encoder parameter vector, $\bar{\phi}$ the teacher vector, and $m_{\mathrm{EMA}}$ the momentum coefficient. The teacher is not updated by the predictive-loss gradient. A larger $m_{\mathrm{EMA}}$ changes targets more slowly but can leave them behind a rapidly adapting student. This target-update mechanism is used in the JEPA lineage~\citep{assran2023ijepa,bardes2024vjepa}. Freezing an independent pretrained teacher is a different design.

For causal dynamics learning, the visible context must stop at the forecast origin, and the predictor must associate a proposed control sequence with later target features. V-JEPA~2's action-conditioned post-training illustrates this additional step~\citep{assran2025vjepa2}. Bidirectional pretraining remains useful, but its masked-completion score should not be relabeled as open-loop prediction accuracy. A goal-conditioned planner also needs a meaningful discrepancy between predicted and goal features and a valid action-search procedure. Representation quality, transition quality, and search quality are separately testable.

Feature regression has a statistical limit worth making explicit. With a fixed target representation and squared error, the population-optimal deterministic prediction is its conditional mean. With scalar absolute error, it is a conditional median. Neither objective by itself recovers multiple modes or assigns probabilities to them. A scene with two plausible future contact states can therefore require a stochastic latent predictor, multiple hypotheses, or a probabilistic event head. The appropriate objective depends on whether the downstream consumer needs a reusable feature, a most likely outcome, or a calibrated distribution. This applies to JEPA-style targets and to predictive-feature distillation in world-action models alike.

\subsubsection{Action-Conditioned Dynamics and World-Action Learning}
\label{sec:wm-action-training}

Actions connect prediction to intervention, but an action record must have a precise meaning. Distinguish a proposed command, the command actually applied after clipping or safety filtering, and the subsequent measured physical response. A deployable forward model can condition on the applied input or model the actuator transformation from a proposed command. It must not require the future measured response as an input to predict that same response. Controller mode, reference frame, units, duration, delays, and saturation belong in the record. The notation $a_t$ refers to the declared input interface, not an unspecified mixture of command and consequence.

\paragraph{Forward, Inverse, and Behavioral Targets}
A forward model predicts the next observation or state from current evidence and an input. An inverse model predicts an input compatible with two states or observations. A policy predicts an input from current evidence and a goal. These targets are related by the joint distribution in Eq.~\eqref{eq:wm-joint-factorization}, but they are not equivalent. Several actions can reach the same endpoint, and matching endpoints can hide different contact forces or unsafe intermediate states. Inverse-dynamics training therefore needs a multimodal output distribution or an explicit criterion for selecting among compatible actions.

The three targets correspond to the three world-action interfaces in the right panel of Figure~\ref{fig:wm-training}, defined as attention masks over observed-context tokens $\mathrm{C}$, action tokens $\mathrm{A}$, and future-observation tokens $\mathrm{F}$. A \emph{forward dynamics model} (FDM) predicts $\widehat{\mathrm{F}}$ from $\mathrm{C}$ and $\mathrm{A}$. Given what is observed and a candidate input, it forecasts the consequence, which is exactly the transition query of Figure~\ref{fig:wm-concept} and the rollout mechanism used in planning and imagination. An \emph{inverse dynamics model} (IDM) predicts $\widehat{\mathrm{A}}$ from $\mathrm{C}$ and $\mathrm{F}$. It reads how the scene actually changed and infers a compatible input. An IDM that requires the realized future observation cannot use that observation to choose the preceding action online. In the training interface considered here, it labels recorded interaction with action estimates, recovers latent actions from action-free video, and shapes representations that encode how actions change the world. Even here, only the action query may read the future. The context rows of the inverse mask bar $\mathrm{C}$ from attending to $\mathrm{F}$, so context features, which may be cached or shared with the policy mode, remain deployable summaries of observed evidence rather than carriers of future information. A \emph{policy} mode predicts $\widehat{\mathrm{A}}$ from $\mathrm{C}$ alone, possibly with a goal encoded in the context. The future-to-action attention path is therefore masked out entirely (the crossed-out cell in Figure~\ref{fig:wm-training}). A single leaky edge would let action tokens read future observations at training time and silently lose that input at test time.

Sharing one network across the three modes is attractive because the modes provide complementary supervision. The FDM grounds what actions do, the IDM grounds which actions produce observed change, and the policy converts both into executable behavior. The cost is discipline in the masks and in gradient routing. A stop-gradient blocks a parameter update but not a forward information path, so frozen and detached components must be distinguished from masked ones, and a deployment-mode forward pass with all training-only branches absent is the decisive test that the policy mode's inputs are exactly those available at runtime.

Behavior cloning learns demonstrated actions, using a likelihood, regression, diffusion, or flow objective. Diffusion Policy exemplifies a conditional distribution over action sequences rather than a single independent action at each step~\citep{chi2023diffusion}. A failed action can remain useful for a forward model and an outcome predictor while being excluded from a nominal imitation objective. Reward, termination, contact-event, and task-progress heads can make a state more useful for decisions, but they supervise different quantities. A progress label is not a value target unless its connection to future return has been established.

\paragraph{Multi-Source and Multi-Objective Optimization}
A useful way to document joint training is
\begin{equation}
 \mathcal L_{\mathrm{joint}}(\Theta)
 =\sum_{i_{\mathrm{src}}=1}^{N_{\mathrm{src}}}\rho_{i_{\mathrm{src}}}\,
   \mathbb E_{\xi_{\mathrm{sample}}\sim\mathcal D_{i_{\mathrm{src}}}}
   \left[\sum_{j\in\mathcal J}\lambda_j\,m_j(\xi_{\mathrm{sample}})\,
                  \ell_j(\Theta;\xi_{\mathrm{sample}})\right],
 \label{eq:wm-joint-training}
\end{equation}
where $N_{\mathrm{src}}$ is the number of data sources, $\mathcal D_{i_{\mathrm{src}}}$ the distribution of source $i_{\mathrm{src}}$, $\rho_{i_{\mathrm{src}}}\geq0$ its sampling weight with $\sum_{i_{\mathrm{src}}}\rho_{i_{\mathrm{src}}}=1$, and $\xi_{\mathrm{sample}}$ a sampled training record. The set $\mathcal J$ indexes enabled objectives, $\ell_j$ their per-example losses, $\lambda_j\geq0$ their weights, and $m_j(\xi_{\mathrm{sample}})\in\{0,1\}$ indicates whether the target exists and is appropriate. The parameter collection is $\Theta$. Per-dimension masks and normalization by valid target counts are applied inside $\ell_j$ when needed. This equation is a bookkeeping template, not an assertion that the objectives have one common probabilistic interpretation.

The distinction between sampling weights and loss weights matters. Increasing a robot source's sampling rate increases all of its enabled objectives, whereas increasing an action-loss coefficient changes the balance within sampled records. Loss scales also depend on token counts and units. A long video target can dominate a short action chunk unless normalization is deliberate. Joint training can share useful structure but can also let appearance gradients erase precise control information or let narrow action data degrade broad semantics. Freezing pretrained modules, staged unfreezing, separate experts, restricted gradient paths, and alternating updates offer different compromises between preservation and adaptation.

\paragraph{Future Supervision Without Future-Input Leakage}
Future observations are legitimate training targets. The restriction is on the forward information available to the deployed policy. A predictive feature branch can map present history to future-relevant features and align them with a detached future teacher. Alternatively, a frozen generative model can receive learned conditioning and send a loss gradient back through that conditioning while its own weights remain fixed. Freezing weights is not the same as disabling the gradient with respect to the model's input. InternVLA-A1.5 uses this latter mechanism to train foresight conditioning~\citep{wmMa2026A15}.

InternW0-$\Delta$ uses latent-change supervision and detached future-feature alignment for Causal Imprint, alongside training-only 4D teacher targets~\citep{wmMiao2026Delta}. The general lesson is to document every route by which future information could reach action activations, including shared attention, side conditioning, cached features, and preprocessing. A direct mask between future and action tokens is insufficient if an intermediate token first reads the future and then passes it onward. A detached target blocks a gradient path. A causal attention restriction blocks a forward information path. Both can be required, but they solve different problems. Temporal ``causality'' of a mask also does not by itself establish causal identification of the physical system.

\paragraph{A Worked Multi-Supervision Construction}
World-action training becomes clearer when each auxiliary target is written separately. InternW0-$\Delta$ provides a concrete example with video and action flow losses, clean-latent change regression, future-feature alignment, and training-only 4D distillation~\citep{wmMiao2026Delta,wmInternW0DeltaProject}. The following normalized formulation explains their roles using this chapter's notation. It is not a prescription for the paper's exact layer indices, tensor layout, or loss coefficients.

Let $u_{t}^{\mathrm{vid}}$ be a clean video-autoencoder latent for the current observation, and let $U_{t}^{+}=(u_{t+\tau_{\mathrm{stride}}}^{\mathrm{vid}},\ldots,u_{t+N_{\mathrm{slice}}\tau_{\mathrm{stride}}}^{\mathrm{vid}})$ contain $N_{\mathrm{slice}}$ future latent slices with sampling stride $\tau_{\mathrm{stride}}$. The autoencoder is frozen in this construction. The action chunk is $A_{t}^{\mathrm{seq}}$. Let $q_{t}=P^{\mathrm{q}}_\theta(h_t,g,c)$ denote predictive tokens computed only from observed history, instruction, and known conditions. A readout $\widehat{D}_t$ from these tokens is trained against consecutive clean-latent differences given by
\begin{equation}
\begin{aligned}
 D_{t,i}&=u_{t+i\tau_{\mathrm{stride}}}^{\mathrm{vid}}-u_{t+(i-1)\tau_{\mathrm{stride}}}^{\mathrm{vid}},\qquad i=1,\ldots,N_{\mathrm{slice}},\\
 D_{t}&=[D_{t,1};\ldots;D_{t,N_{\mathrm{slice}}}],\qquad
 \mathcal L_{\mathrm{change}}
 =\mathbb E\!\left[\frac{\|\widehat{D}_t-\operatorname{sg}[D_{t}]\|_2^2}{d_{\mathrm{D}}}\right],
\end{aligned}
\label{eq:wm-change-supervision}
\end{equation}
where brackets denote stacking, $d_{\mathrm{D}}$ is the number of valid scalar target entries, and the expectation averages training records. The norm is the Euclidean norm of the concatenated valid scalar latent entries. The readout includes any output reshaping or projection needed to match dimensions. Padded entries are omitted. A latent difference emphasizes change rather than static appearance, but it is not a physical velocity. Its units are feature units, and the sampling interval still matters. Computing differences before noise injection avoids supervising independent diffusion noise as if it were motion. Camera changes and encoder instability can also create latent differences, so this objective needs complementary semantic or geometric checks.

A second target can transfer richer intermediate features. Let $\mathbf{q}_{t,i_{\mathrm{tok}}}^{(\ell_{\mathrm{s}})}$ be a predictive token at student layer $\ell_{\mathrm{s}}$ and spatial position $i_{\mathrm{tok}}$, and let $\mathbf{f}_{t,i_{\mathrm{tok}}}^{+,\ell_{\mathrm{T}}}$ be a spatially corresponding feature of a future slice at target layer $\ell_{\mathrm{T}}$. With compatible feature dimensions, a representative alignment objective is
\begin{equation}
 \mathcal L_{\mathrm{align}}
 =\mathbb E_{t,i_{\mathrm{tok}}}\!\left[
  1-\operatorname{cos}_{\mathrm{sim}}\!\left(\mathbf{q}_{t,i_{\mathrm{tok}}}^{(\ell_{\mathrm{s}})},
                     \operatorname{sg}[\mathbf{f}_{t,i_{\mathrm{tok}}}^{+,\ell_{\mathrm{T}}}]\right)\right].
 \label{eq:wm-future-alignment}
\end{equation}
The function $\operatorname{cos}_{\mathrm{sim}}(\mathbf{u},\mathbf{v})=\mathbf{u}^\top \mathbf{v}/(\|\mathbf{u}\|_2\|\mathbf{v}\|_2)$ measures direction similarity for nonzero feature vectors, with numerical stabilization used in implementation. Layer indices and spatial correspondence are part of the target definition. The future feature can come from another branch of the same trainable model. Detaching this target prevents this loss from updating it, while other objectives may still update its parameters. Such a target is not the same as a permanently frozen teacher. Compared with clean-latent regression, feature alignment transfers the target branch's abstraction but inherits its biases and changes during training.

An external geometry-and-motion teacher supplies a third, complementary signal. Let $\mathbf{r}_t^{\mathrm T}=R_{\mathrm T}(o_{t:t+H_{\mathrm{o}}},c)$ be a descriptor extracted from a recorded clip and $\mathbf{r}_t^{\mathrm S}=R_\theta^{\mathrm S}(h_t,g,c)$ the student descriptor predicted from available observations. The horizon $H_{\mathrm{o}}$ is the observation horizon. $R_{\mathrm T}$ is frozen and may access training-only future frames, while $R_\theta^{\mathrm S}$ may not. A corresponding objective is
\begin{equation}
 \mathcal L_{\mathrm{geom}}
 =\mathbb E_{\mathrm{valid}}\!\left[
   \frac{\|\mathbf{r}_t^{\mathrm S}-\operatorname{sg}[\mathbf{r}_t^{\mathrm T}]\|_2^2}{d_{\mathrm{r}}}\right],
 \label{eq:wm-geometric-distillation}
\end{equation}
where $d_{\mathrm{r}}$ is the descriptor dimension and the expectation includes only samples with valid teacher outputs. Descriptor components should be normalized according to their semantics. Unlike a hard physical constraint, this objective transfers an estimated geometric or motion prior. It does not establish exact metric geometry, correct forces, or conservation. The auxiliary student readout can be removed after training when its role was only to shape a representation already used by the policy.

The combined objective is then
\begin{equation}
\begin{aligned}
 \mathcal L_{\mathrm{WAM}}
 ={}&\lambda_{\mathrm{v}}\mathcal L_{\mathrm{FM}}^{\mathrm{v}}(U_t^+)
    +\lambda_{\mathrm{a}}\mathcal L_{\mathrm{FM}}^{\mathrm{a}}(A_{t}^{\mathrm{seq}})
    +\lambda_{\mathrm{ch}}\mathcal L_{\mathrm{change}}\\
   &+\lambda_{\mathrm{al}}\mathcal L_{\mathrm{align}}
    +\lambda_{\mathrm{ge}}\mathcal L_{\mathrm{geom}}.
\end{aligned}
\label{eq:wm-multisupervision}
\end{equation}
The superscripts $\mathrm{v}$ and $\mathrm{a}$ identify separately conditioned video and action flow objectives of Eq.~\eqref{eq:wm-flow-training}. The nonnegative $\lambda$ coefficients balance their contributions. Source-specific weights and validity masks follow Eq.~\eqref{eq:wm-joint-training}. This combination assigns distinct jobs to reconstructable future detail, executable behavior, change prediction, and teacher knowledge. Ablations should remove or vary these jobs independently, matching data and optimization budgets, rather than comparing only the total loss.

A limitation follows from the conditioning. When $q_t$ does not receive a candidate action sequence, it predicts future-relevant information under the behavior represented in its data, not the separate outcome of every possible intervention. Its value can be improved action features without a standalone counterfactual simulator. Joint video-action training should therefore declare whether its video branch sees supplied actions, generated actions, or neither. The behavioral benefit of future supervision and the correctness of action-conditioned physical prediction are different claims.

\paragraph{Parameters, Information Paths, and a Training Iteration}
Table~\ref{tab:wm-supervision-paths} distinguishes the pathways in this construction. Observation encoders, frozen semantic models, detached targets, and trainable world experts play different roles even when their outputs are all called features. A forward dependency restriction concerns the information available to activations. A stop-gradient concerns which parameters a loss updates. Freezing a module's weights does not necessarily stop derivatives through its computation to an upstream trainable input.

\begin{table}[hbt]
 \centering
 \caption{Information and optimization roles in predictive-feature world-action training. This is a general implementation checklist, not a claim that all world-action models use identical branches}
 \label{tab:wm-supervision-paths}
 \small
 \renewcommand{\arraystretch}{1.10}
 \begin{tabularx}{\textwidth}{@{}>{\raggedright\arraybackslash}p{0.22\textwidth} >{\raggedright\arraybackslash}X >{\raggedright\arraybackslash}X@{}}
  \toprule
  \textbf{Branch or target} & \textbf{Training information and updates} & \textbf{Deployment role} \\
  \midrule
  Observed-context features & Present/past observations\newline Backbone updated if trainable & Available context\newline May be cached \\
  Frozen semantic conditioning & Current observations and instruction\newline Encoder weights fixed & Retained when the action head depends on it \\
  Future video branch & Noised future target and allowed context\newline Video loss updates its expert & Optional\newline Omitted in observation-only predictive-feature execution \\
  Predictive change tokens & Observed context only\newline Future outcomes provide losses, not inputs & Retained as action conditioning \\
  Detached future features & Future branch supplies targets\newline This alignment loss cannot update the target path & Not required as realized future inputs \\
  External geometry teacher and auxiliary readout & Recorded clip targets\newline Student learns and teacher stays fixed & Both removable if used only for distillation \\
  Action expert & Noised actions and deployment-valid context\newline Action loss trains the policy path & Generates the executable action chunk \\
  \bottomrule
 \end{tabularx}
\end{table}

One minibatch update proceeds as follows. Construct a time-consistent observed context, action chunk, future targets, and validity masks. Encode frozen inputs and retrieve any valid cached teacher descriptors. Noise the video and action targets according to their declared schedules, while keeping the observed conditioning and clean change targets separate. Run the experts with the intended directed dependencies, compute only the supported losses, normalize them by valid entries, and update the trainable parameter groups. If a moving-average teacher is used, update it after the student step. Finally, test an inference-mode forward pass with all training-only future and teacher branches absent. The policy must still have every input it requires.

This last test should inspect indirect as well as direct paths. An observed token that reads a future token at one layer can leak that future to the policy at a later layer even when future-to-action attention is masked. Temporal preprocessing can leak in the same way. Encoding an observed frame together with future frames through a noncausal video encoder can contaminate its latent. Cache reuse additionally requires independence from the evolving action and artificial-time inputs, not merely independence from future observations. These checks turn a verbal claim of ``training-only supervision'' into an auditable computation graph.

\paragraph{Learning a World Model Versus Learning Within One}
Model-based reinforcement learning adds a second optimization problem in which a policy is improved using transitions generated by a learned model. In Dreamer, imagined trajectories train behavior while the learned dynamics provide the rollout mechanism~\citep{hafner2019dreamer}. Model parameters can be held fixed during policy optimization even when derivatives through the model are used to update policy parameters. Conversely, updating the model to maximize its own predicted reward would not necessarily improve its physical accuracy.

Dyna-style alternation uses real experience to improve the model and model-generated experience to improve decisions~\citep{wmSutton1991}. MBPO highlights a useful compromise by using short model rollouts branched from real states, limiting reliance on distant model predictions~\citep{wmJanner2019}. Longer imagination expands possible credit assignment but exposes the policy to compounding errors and model exploitation. Offline data create an additional support problem. A proposed policy may select actions absent from the training distribution. Policy improvement should therefore be coupled to rollout-length selection, conservative use of uncertain predictions, renewed real data, and independent physical evaluation rather than judged only by imagined return.

\subsubsection{Physical Constraints, Counterfactuals, and Uncertainty Modeling}
\label{sec:wm-physics-training}

Physical structure can enter through loss penalties, architectural restrictions, or a hybrid of known dynamics and learned corrections. Physics-informed neural networks demonstrate the use of governing-equation residuals as learning signals~\citep{raissi2019physics}. Structured simulators illustrate complementary restrictions arising from a representation and its update rules~\citep{sanchezgonzalez2020learning,wmXie2024}. The appropriate choice depends on which laws, parameters, and boundary conditions are actually known.

\paragraph{Soft Penalties and Structural Constraints}
For a predicted physical trajectory $\widehat S=(\widehat s_t,\ldots,\widehat s_{t+H})$ under actions $A_{t}^{\mathrm{seq}}$, a representative penalty is
\begin{equation}
 \mathcal L_{\mathrm{phys}}
 =\mathbb E\left[
   \sum_{j=1}^{J_{\mathrm{eq}}}\alpha_{j}^{\mathrm{eq}}\|f_{j}^{\mathrm{eq}}(\widehat S,A_{t}^{\mathrm{seq}},c)\|_2^2
  +\sum_{j=1}^{J_{\mathrm{ineq}}}\beta_{j}^{\mathrm{ineq}}\|[g_{j}^{\mathrm{ineq}}(\widehat S,A_{t}^{\mathrm{seq}},c)]_+\|_2^2
   \right].
 \label{eq:wm-physical-constraints}
\end{equation}
The residual $f_{j}^{\mathrm{eq}}=0$ specifies an equality, $g_{j}^{\mathrm{ineq}}\leq0$ an inequality, $[u]_+=\max(u,0)$ applied componentwise, and $J_{\mathrm{eq}},J_{\mathrm{ineq}}$ the numbers of each constraint. The nonnegative coefficients $\alpha_{j}^{\mathrm{eq}},\beta_{j}^{\mathrm{ineq}}$ include suitable scaling by units and tolerances. The expectation averages training records and sampled predictions. These coefficients are local to this equation. Constraints must act on physical readouts or physically defined latent variables, not on arbitrary embedding coordinates without an interpretation.

Examples include joint limits, nonpenetration, contact consistency, and energy or mass balance under known boundary fluxes. A system exchanging heat or liquid with its surroundings should obey a balance equation, not an inappropriate constant-energy or constant-mass penalty. Tolerances should reflect measurement precision and modeling assumptions. A strong incorrect constraint biases the model. A weak penalty may leave violations large. Enforcing symmetries or integrating known mechanics can reduce violation by construction, but only for the constrained property and under the numerical method's assumptions. It cannot identify an unmeasured friction coefficient or certify all downstream behavior.

\paragraph{Counterfactual and Intervention Supervision}
Paired rollouts can compare different actions from matched initial conditions. A resettable simulator can hold state and external inputs fixed while varying an intervention. In a physical experiment, the same comparison requires reproducible initial conditions, controlled confounders, or a justified statistical design. Outcome diversity from an unconditioned video generator is not evidence that it learned how a particular action changes the outcome. Synthetic counterfactuals supervise the simulator's mechanisms. Independent measurements are needed before transferring those conclusions to reality.

A useful training contrast varies the commanded magnitude, direction, or duration while keeping nonintervened conditions comparable. It can supervise the sign and scale of the response as well as which objects should remain unchanged. Care is required near discontinuities. A slightly larger force may change sticking to sliding rather than yield a slightly larger displacement. A smooth penalty that averages these modes can suppress the event the model is supposed to predict.

\paragraph{Predictive Distributions and Uncertainty Sources}
Deterministic regression is cheap but cannot express multiple plausible outcomes. A conditional Gaussian supplies a mean and variance. Mixtures, stochastic latent states, diffusion, or flows can represent more complex distributions. For a scalar measured target $y$ and condition $\kappa$, a Gaussian negative log-likelihood, omitting an additive constant, is
\begin{equation}
 \mathcal L_{\mathrm{Gauss}}
 =\frac12\mathbb E_{\mathcal D}
   \left[\frac{(y-\mu_\theta(\kappa))^2}{\sigma_\theta^2(\kappa)}
          +\log\sigma_\theta^2(\kappa)\right],
 \label{eq:wm-gaussian-uncertainty}
\end{equation}
where $\mu_\theta$ is the predicted conditional mean and $\sigma_\theta^2>0$ the variance. The residual term penalizes inaccurate concentrated predictions, while the logarithmic term discourages arbitrarily inflating variance. Positivity can be enforced through a log-variance parameterization with numerically appropriate bounds. This example illustrates heteroscedastic likelihood learning~\citep{wmKendall2017}. Its Gaussian assumption is inadequate when distinct contact modes produce well-separated outcomes.

An ensemble adds variation across fitted models. For $M$ equally weighted scalar predictors with means $\mu_m$ and variances $\sigma_m^2$ at the same condition, the mixture variance obeys
\begin{equation}
 \bar\mu=\frac1M\sum_{m=1}^{M}\mu_m,\qquad
 \operatorname{Var}(Y\mid\kappa)
 =\frac1M\sum_{m=1}^{M}\sigma_m^2
  +\frac1M\sum_{m=1}^{M}(\mu_m-\bar\mu)^2,
 \label{eq:wm-ensemble-variance}
\end{equation}
where $Y$ is the predicted random quantity and $m$ the ensemble index. The identity separates within-model dispersion from between-model disagreement. Deep ensembles and probabilistic ensemble dynamics use this kind of diversity for predictive uncertainty and planning~\citep{wmLakshminarayanan2017,wmChua2018}. However, disagreement is only a proxy for epistemic uncertainty, and the within-model term can include unresolved state ambiguity as well as observation and process noise. Correlated model errors can leave an ensemble confidently wrong.

Uncertainty must also propagate consistently across time. A hidden but fixed mass should be sampled once for a trajectory hypothesis rather than independently every step. Sensor noise may vary at each observation. More detailed distributions increase sampling and calibration costs. Deterministic models can still be useful when their operating range is narrow and externally monitored, but they should not be presented as calibrated probabilistic simulators. Section~\ref{sec:wm-prediction-validation} explains how uncertainty is evaluated rather than inferred from the mere presence of stochastic layers.

\subsubsection{Multi-Step Prediction, Multiple Time Scales, and Adaptation}
\label{sec:wm-multistep-training}

One-step accuracy can hide recursive drift. A generic horizon-weighted objective is
\begin{equation}
 \mathcal L_{\mathrm{multi}}
 =\mathbb E_{\mathcal D}\left[
   \sum_{k=1}^{H}w_k\,
   \ell\!\left(\widehat y_{t+k\mid t},y_{t+k}\right)\right],
 \qquad w_k\geq0.
 \label{eq:wm-multistep-training}
\end{equation}
The prediction $\widehat y_{t+k\mid t}$ is generated from evidence available at $t$, recursively applying the supplied actions without incorporating intermediate real observations. The target $y_{t+k}$ is a later measurement or explicitly identified teacher target. $w_k$ weights horizon $k$, $H$ is the maximum horizon, and $\ell$ is a suitable regression or probabilistic loss. A probabilistic $\widehat y$ denotes a distribution, not necessarily one sampled value. Supervision should score that distribution appropriately rather than force every sample to match one realized future.

\paragraph{Teacher Forcing, Free Rollout, and Curriculum}
Teacher forcing provides stable local supervision but exposes the model mainly to data states. Free rollout trains on its own evolving predictions and addresses the deployment distribution more directly, at the cost of unstable early trajectories and longer gradient paths. Scheduled sampling mixes real and generated preceding inputs~\citep{wmBengio2015}. It is a training heuristic, not generally the original sequence-likelihood objective. A sensible comparison therefore includes teacher-forced error, open-loop error, and task behavior rather than assuming that mixed inputs always improve all three.

Latent overshooting in PlaNet extends consistency pressure beyond one transition by matching multi-step predictive distributions to observation-informed states~\citep{hafner2019planet}. Related multi-step objectives can supervise physical readouts or future embeddings. Longer windows improve exposure to delayed effects, but require more memory and can overweight a particular target encoder's errors. A curriculum can begin with short horizons and then include longer windows and contact-centered segments, while preserving enough ordinary motion to avoid a distorted training distribution. Truncating backpropagation lowers cost but changes which temporal dependencies receive direct gradient signals.

\paragraph{Multiple Timescales and Temporal Consistency}
Different sensors and physical variables evolve at different rates. Training should include elapsed-time conditioning, modality-validity masks, and action durations rather than assuming one frame equals one control interval. Slow semantic context, faster visual dynamics, and high-rate force feedback can share state through explicit timing interfaces. Interpolation is appropriate only when its assumptions hold. Interpolating across an impact can erase the discontinuity.

For a Markov transition kernel under the same constant control input $a$ and conditions $c$, time consistency takes the form
\begin{equation}
 T_\theta(z''\mid z,a,c,\Delta_1+\Delta_2)
 \approx\int T_\theta(z''\mid u,a,c,\Delta_2)
              T_\theta(u\mid z,a,c,\Delta_1)\,\mathrm d u,
 \label{eq:wm-time-consistency}
\end{equation}
where $z,z''$ are initial and final latent states, $u$ is an intermediate latent state, and $\Delta_1,\Delta_2>0$ are physical durations. This is a distribution-level Chapman-Kolmogorov consistency condition under the stated Markov and fixed-input assumptions. It is not generally valid to replace the intermediate distribution by its mean in nonlinear dynamics. If actions or external conditions change, both sides must use the same time-varying input schedule. If the compressed state is not Markov, residual inconsistency may reveal missing memory rather than numerical error.

\paragraph{Post-Training, Distillation, and Continued Interaction}
Task-specific post-training can refine action interfaces, contact targets, or selected dynamics. InternW0 adds available force and tactile histories and predicts future contact channels during contact-aware post-training~\citep{wmCai2026InternW0}. Such adaptation requires valid contact supervision. Missing physical channels cannot be recovered simply by adding a new output head. Distillation can transfer an expensive teacher's features or predictions to a smaller or less frequently executed branch, but the teacher's uncertainty and errors should remain distinguishable from measured truth.

Local adaptation should separate belief correction, physical-parameter identification, and updates to learned network weights. A failed grasp may revise a contact hypothesis without changing the model. Repeated measured slip may justify identifying friction. Systematic errors across conditions may require a new training round. Replay, regularization, and selective parameter updates address the trade-off between new learning and retention. Elastic weight consolidation is one example of limiting changes to parameters important for previous tasks~\citep{kirkpatrick2017ewc}, not a guarantee against forgetting in an arbitrary world model. New versions should be accepted only after tests of both new and retained capabilities.

\paragraph{Staging Auxiliary Supervision and Adapting the Deployed Interface}
A multi-objective recipe can be organized around a stable behavioral and predictive baseline rather than enabling every target from the first update. InternW0-$\Delta$ first performs joint pretraining and subsequently adds cached 4D supervision during continued pretraining, before task- and embodiment-specific post-training~\citep{wmMiao2026Delta}. The general choice is between early joint shaping and later auxiliary refinement. Early auxiliary losses can guide representation formation, but noisy or mismatched teachers can interfere before the action interface becomes useful. Later introduction reduces that early interference but may require a smaller learning rate or replay to avoid disturbing established behavior. Neither schedule is a universal optimum.

Post-training should match the actual deployment contract. A target robot may change valid action dimensions, normalization, camera configuration, actuation delay, or the relation between predicted and executed chunk lengths. Adapt the relevant input/output mappings and condition the model on these differences. Do not reinterpret an identical numerical vector as an identical physical action. If successive chunks overlap or part of a chunk has already been committed to execution, the training condition should identify that prefix and the loss should target the remaining controllable portion. Where online feedback is used to choose future actions, it must be separated from the fixed action sequence assumed by an open-loop dynamics objective.

Frozen-encoder and teacher caches can reduce repeated training work, but only while their dependencies remain fixed. Changing an encoder checkpoint, a crop, a view layout, an instruction, or the temporal window can invalidate a cached feature. Offline geometric descriptors should retain their source clip, teacher version, confidence, and normalization. Activation checkpointing trades extra computation for memory. Mixed precision reduces storage and arithmetic cost but needs checks for unstable losses and gradients. Report these implementation choices separately from reductions in the amount of training data or the number of optimization steps. A faster experiment pipeline is valuable, but it is not itself evidence of better learned dynamics.

For model selection, monitor the component losses alongside open-loop predictions and closed-loop behavior. A lower teacher-feature error can coexist with worse action accuracy if the representation discards contact detail. Likewise, a lower action loss on demonstration states can conceal failure on states visited by the policy. Compare a frozen baseline, a joint-training variant, and an auxiliary-supervision variant on fixed held-out conditions, with unchanged target preprocessing and comparable budgets. This distinguishes the contribution of predictive supervision from a larger backbone, a different data mixture, or additional optimization.

\paragraph{Training Trade-Offs and Reproducible Reporting}
Table~\ref{tab:wm-training-tradeoffs} summarizes choices that should be reported together rather than hidden inside a total loss. At minimum, an experiment should state its data mixture and splits, temporal sampling, target normalization, trainable and frozen modules, gradient paths, optimizer and learning-rate schedule, effective batch size, sequence length, number of updates, numerical precision, and training and inference budgets. No fixed recipe is optimal across video generation, latent planning, and multimodal control.

The common criterion is whether the resulting model answers the intended queries on evidence it could actually receive. Training losses are useful diagnostics, but the relationships among reconstruction, dynamics, uncertainty, and behavior must be established through the validation protocols below.

\begin{table}[h]
  \centering
  \caption{Training choices and their principal trade-offs. Each comparison requires matched data or compute where feasible, plus evaluation on the actual prediction or control interface}
  \label{tab:wm-training-tradeoffs}
  \small
  \renewcommand{\arraystretch}{1.12}
  \begin{tabularx}{\textwidth}{@{}>{\raggedright\arraybackslash}p{0.22\textwidth} >{\raggedright\arraybackslash}X >{\raggedright\arraybackslash}X@{}}
    \toprule
    \textbf{Choice} & \textbf{Potential benefit} & \textbf{Cost or failure to check} \\
    \midrule
    Observation vs. feature targets & Detail and query breadth vs. compact predictive abstraction & Expensive nuisance modeling vs. omitted physical information \\
    Frozen vs. jointly updated modules & Prior preservation and cheaper optimization vs. task adaptation & Restricted adaptation vs. objective interference and forgetting \\
    Short vs. long rollout training & Stable local learning vs. delayed-effect and drift supervision & Myopic accuracy vs. memory cost and difficult optimization \\
    Deterministic vs. distributional models & Low overhead vs. multiple outcomes and probabilistic queries & Hidden ambiguity vs. calibration and sampling cost \\
    Soft vs. structural physics & Flexible constraints vs. restricted admissible predictions & Residual violations vs. bias from incorrect assumptions \\
    Offline vs. interaction-driven updates & Reproducible reuse vs. coverage of policy-induced failures & Unseen-action errors vs. physical cost and retention risk \\
    \bottomrule
  \end{tabularx}
\end{table}

\subsection{Using and Validating World Models}
\label{sec:wm-validation}

Using a world model means making a query under a declared information boundary and, when appropriate, using its answer to choose an intervention. Validation should separate the quality of a forecast from its physical consistency and from the utility of the decision it supports. A model can produce physically possible but inaccurate futures, accurate short-term forecasts that do not improve a policy, or useful actions without exposing a general-purpose predictive interface. This section develops protocols that make these distinctions measurable.

\subsubsection{Interactive Simulation and Conditional Queries}
\label{sec:wm-queries}

An interactive call starts from an inferred state, belief, or observation history and returns one or more conditional futures. A query may request an image, object trajectory, contact event, temperature, boundary crossing, or task outcome. It should specify the available evidence, intervention interface, physical horizon, environment conditions, and required precision. For a partially known scene, the returned uncertainty should include unresolved state alternatives, not only randomness introduced during decoding.

\paragraph{State Posterior, Future Prediction, and Retrospective Checking}
The state posterior summarizes evidence available up to the prediction time. A future forecast propagates that posterior without using the future measurements to be predicted. For an observation-generating latent model, a representative factorization is
\begin{equation}
\begin{aligned}
 &p_{\theta,\phi,\omega_{\mathrm{obs}}}(o_{t+1:t+H}\mid h_t,A_{t}^{\mathrm{seq}},c)\\
 &\quad=\int q_\phi(z_t\mid h_t,c)
 \prod_{k=0}^{H-1}\Big[
 T_\theta(z_{t+k+1}\mid z_{t+k},a_{t+k},c,\Delta t_k)
 O_{\omega_{\mathrm{obs}}}(o_{t+k+1}\mid z_{t+k+1},c)\Big]
 \,\mathrm d z_{t:t+H},
\end{aligned}
\label{eq:wm-posterior-prediction}
\end{equation}
where $h_t$ is the observed history, $A_{t}^{\mathrm{seq}}=a_{t:t+H-1}$ a supplied action sequence of $H$ intervals, $q_\phi$ the inferred latent-state posterior, $T_\theta$ the learned transition, and $O_{\omega_{\mathrm{obs}}}$ the observation likelihood. The integration marginalizes the current and future latent states $z_{t:t+H}$. $\Delta t_k$ specifies each physical interval. The symbols $\phi,\theta,\omega_{\mathrm{obs}}$ denote fixed fitted parameters in this expression. It is posterior prediction with respect to the current latent state, not necessarily a Bayesian average over uncertain network parameters. The Markov and observation assumptions are those introduced in Section~\ref{sec:wm-belief}. A model with additional dependencies must include them explicitly~\citep{Kaelbling1998,hafner2019planet}.

After real execution, the withheld observations can be used to check the forecast and then update the state. This \emph{retrospective validation of an earlier prediction} is different from reconstructing a future frame after that frame has already entered an inference network. Likewise, smoothing a past state using later evidence can be useful for diagnosis but cannot be reported as an online forecast made with less information. Evaluation logs should retain the prediction, model version, conditioning evidence, and issue time before the target outcome is revealed.

Eq.~\eqref{eq:wm-posterior-prediction} describes a query interface, not a requirement that every architecture evaluate an explicit density. Sample-based models can approximate the predictive distribution. Deterministic models return a restricted point prediction. A latent-only model can answer a physical or task query through a readout rather than an observation decoder. In all cases, a goal specifies what is desired and should not be treated as evidence that the goal will occur. Scenario assumptions, such as a proposed payload or a fixed external temperature, must remain distinguishable from measured conditions.

\paragraph{Interactive Correction and Information Acquisition}
Real feedback should update the inferred state before the next query. A prediction made before a grasp slips may no longer be a valid context for later actions even if its video remains visually coherent. Asynchronous systems should therefore expose the age and observation support of cached predictions, together with the conditions that trigger correction, regeneration, or rejection.

Predictions can also help choose what to observe next. If two action rankings differ mainly because contact is uncertain, an additional tactile measurement may be more useful than another distant camera frame. This is an information-acquisition decision within the partially observable control problem~\citep{Kaelbling1998}. Its value depends on an observation model, the cost of sensing, and whether the proposed measurement can actually resolve the ambiguity. Querying the same model repeatedly does not create independent evidence about the physical state.

\subsubsection{Imagination, Planning, and Decision Calls}
\label{sec:wm-planning}

Planning proposes candidate actions, predicts their consequences, and ranks them using task objectives and constraints. An imagined collection of branches can be written as
\begin{equation}
 \mathcal C_{\mathrm{cand}}=\left\{
 \left[\left(a_{t+k}^{(b)},\widehat z_{t+k+1}^{(b)}\right)\right]_{k=0}^{K_b-1}
 : b=1,\ldots,N_{\mathrm{roll}}\right\},
 \label{eq:wm-imagination-set}
\end{equation}
where $\mathcal C_{\mathrm{cand}}$ is a collection of ordered imagined paths, $N_{\mathrm{roll}}$ is the number of candidate branches, $b$ indexes a branch, and $K_b$ is its number of intervals. The action $a_{t+k}^{(b)}$ produces the imagined next state $\widehat z_{t+k+1}^{(b)}$ from the preceding state in that branch. Branches may also be repeated under different state hypotheses or stochastic outcomes. Increasing $N_{\mathrm{roll}}$ explores alternatives. Increasing $K_b$ reveals later consequences but also exposes more model error.

For candidates with a common horizon $H$, a useful illustrative objective is
\begin{equation}
\begin{aligned}
 U(A_{t}^{\mathrm{seq}}\mid b_t,g,c)
 =\mathbb E_{\mathrm{model}}\!\left[
   \sum_{k=0}^{H-1}\gamma^k\big(r_{t+k}-\lambda C_{t+k}^{\mathrm{run}}\big)
   -\eta_{\mathrm{pen}}\,\mathbf 1\{\mathcal V_{t:t+H}^{\mathrm{viol}}\}
   \,\middle|\,b_t,A_{t}^{\mathrm{seq}},g,c\right].
\end{aligned}
\label{eq:wm-decision-utility}
\end{equation}
The utility $U$ scores the action sequence $A_{t}^{\mathrm{seq}}=a_{t:t+H-1}$ starting from current belief $b_t$, goal $g$, and conditions $c$. The expectation averages the model's predicted state and outcome uncertainty. The transition reward $r_{t+k}$ measures task benefit, $C_{t+k}^{\mathrm{run}}\geq0$ a per-interval resource cost such as time or energy, and $\gamma\in(0,1]$ the discount factor for equally spaced intervals. The coefficient $\lambda\geq0$ converts resource cost into utility units. The indicator $\mathbf1\{\mathcal V_{t:t+H}^{\mathrm{viol}}\}$ equals one when a specified violation occurs anywhere along the trajectory. $\eta_{\mathrm{pen}}\geq0$ penalizes that trajectory-level event once. Both the reward and the per-step cost are inside the sum. For unequal durations, discounts and accumulated costs should instead follow physical time. This is a generic decision objective, not a world-model fitting loss or a safety certificate. Constrained MPC provides the broader control framework~\citep{mayne2000constrained}.

A planner may optimize actions by sampling, tree expansion, or differentiating through a learned transition. PlaNet uses latent-space planning, while probabilistic ensemble methods such as PETS propagate alternative dynamics and state outcomes during action evaluation~\citep{hafner2019planet,wmChua2018}. A separately trained actor may instead amortize this computation into direct action generation. The actor can be trained using imagined trajectories without running a planner at deployment, as in Dreamer~\citep{hafner2019dreamer}. Planning at inference and policy learning through imagination therefore have different runtime costs and failure modes.

Receding-horizon execution applies only an initial portion of the selected sequence and replans from new real evidence. Without this correction, action optimization may deliberately enter regions where prediction errors create unrealistically high reward. Longer imagined rollouts are not automatically better. They can change action ranking for the wrong reason. Model-based policy optimization addresses this issue by controlling the use and length of imagined experience~\citep{wmJanner2019}. Uncertainty-sensitive ranking can compare candidate outcomes across hidden-state hypotheses, but agreement among several copies of a biased model is not independent confirmation.

Hierarchical imagination can compare coarse routes or process phases before refining a selected interval. Candidates should terminate at a common physical horizon or comparable completion event, with appropriate terminal-value treatment. Otherwise a shorter trajectory can appear preferable merely because its later costs are omitted. Efficient execution can also reuse observation features or invoke detailed prediction only near consequential events. These savings must be evaluated together with the error introduced by stale context, reduced branching, or omitted detail, rather than by model throughput alone.

\begin{table}[tb]
  \centering
  \caption{Validation objects, suitable evidence, and common overclaims. The evaluation protocol must preserve the information available at the forecast issue time}
  \label{tab:wm-validation}
  \small
  \renewcommand{\arraystretch}{1.12}
  \begin{tabularx}{\textwidth}{@{}>{\raggedright\arraybackslash}p{0.17\textwidth} >{\raggedright\arraybackslash}X >{\raggedright\arraybackslash}X >{\raggedright\arraybackslash}X@{}}
    \toprule
    \textbf{Object} & \textbf{Evidence or protocol} & \textbf{Example readout} & \textbf{Overclaim to avoid} \\
    \midrule
    Observation forecast & Held-out, time-matched outcomes\newline Declared sample budget & Pixel or perceptual error, conditional and distribution checks & Overall realism establishes action correctness \\
    State persistence & Occlusion, reappearance, and history changes & Identity errors\newline Position, orientation, or attribute drift & Single-frame recognition establishes memory \\
    Action response & Controlled changes and matched initial conditions & Response magnitude, direction, and event timing & Any action sensitivity establishes a correct intervention model \\
    Physical consistency & Independent geometry, force, and process measurements & Penetration, contact consistency, justified balance residuals & Plausibility establishes quantitative accuracy or causality \\
    Predictive probability & Repeated held-out outcomes and separate calibration data & Brier or CRPS, coverage and width, reliability by horizon & Diverse samples establish calibrated uncertainty \\
    Decision utility & Matched-policy and resource comparisons on real outcomes & Task score, completion time, failures, and latency & Prediction gains establish better real decisions \\
    \bottomrule
  \end{tabularx}
\end{table}

\subsubsection{Prediction Quality and Physical-Consistency Tests}
\label{sec:wm-prediction-validation}

Prediction quality asks whether a forecast agrees with the subsequently observed outcome or its conditional distribution. Physical consistency asks whether the predicted trajectory respects relevant identities, mechanisms, and constraints. These are separate axes. Several trajectories may be physically possible, while only one occurs per trial. A visually similar output may still violate contact or respond incorrectly to an action. Table~\ref{tab:wm-validation} organizes the evidence needed for each claim.

\paragraph{Prediction Quality at a Fixed Information Boundary}
Report one-step prediction from newly observed context, open-loop multi-step prediction without future measurements, and feedback-interleaved prediction with explicitly scheduled observations separately. Also specify whether future controls were fixed before the forecast or recorded from a feedback policy. Conditioning retrospectively on a policy's later actions can reveal information about later states. It is not equivalent to forecasting an independently supplied intervention sequence.

Observation forecasts can be compared using pixel errors, perceptual feature distances such as learned perceptual image patch similarity (LPIPS), and distribution-level video measures such as Fréchet video distance (FVD)~\citep{wmZhang2018,wmUnterthiner2018}. These answer different questions. Pixel error is sensitive to small spatial misalignment. Perceptual metrics inherit the features of their evaluator. A good aggregate distribution score does not establish that each sample follows its supplied action. Multimodal futures further complicate paired image error. A legitimate but unobserved outcome can be penalized even when the predicted distribution is appropriate. Report the sample budget and avoid presenting best-of-many selection as the performance of a single unselected prediction.

State and event queries require measurements matched to their meaning. Object positions and velocities can be evaluated in physical units. Orientations need an appropriate rotational distance. Contact or phase-change events require detection and timing errors. For a scalar physical quantity, a horizon-specific summary is
\begin{equation}
 \operatorname{RMSE}(k)=
 \sqrt{\frac{1}{N_{\mathrm{eval}}}\sum_{i=1}^{N_{\mathrm{eval}}}
  \left(\widehat y_{t_i+k\mid t_i}^{(i)}-y_{t_i+k}^{(i)}\right)^2},
 \label{eq:wm-horizon-error}
\end{equation}
where $N_{\mathrm{eval}}$ is the number of independent evaluation trials, $i$ indexes a trial, $t_i$ is its forecast origin, $k$ the forecast offset on a declared sampling grid, $y$ the measured quantity, and $\widehat y_{t_i+k\mid t_i}$ the point forecast issued at $t_i$. Root mean squared error has the same physical unit as $y$. For multiple quantities, report errors separately or declare the normalization instead of adding meters, newtons, and temperatures without scaling. Plotting or tabulating the error against elapsed time exposes drift hidden by an average over horizons. A probabilistic forecast also needs a distributional score, not only error of its mean.

\paragraph{Physical Consistency and Intervention Response}
Evaluate persistence of object identity and hidden attributes through occlusion, compatibility of geometry across views, non-penetration within measurement tolerance, contact creation and release, and agreement between predicted motion and known control changes. When forces, boundary fluxes, and physical parameters are available, check momentum, energy, or material balances under the corresponding assumptions. Conservation of an isolated system must not be imposed on an open or externally actuated process. These tests should use the physical quantities underlying the constraints of Section~\ref{sec:wm-physics-training}, not merely whether an auxiliary loss is small during training.

Contact and other mechanism changes deserve event-centered reporting. Most recorded frames may show easy steady motion, so an aggregate average can conceal incorrect impact timing, slip onset, or release. Partition errors into free motion, contact formation, sustained contact, release, and relevant process transitions. Controlled changes in action direction, magnitude, duration, or timing test whether the model predicts the \emph{right} change, rather than merely any change. Where possible, paired trials should hold the initial conditions and non-intervened factors fixed. Simulation supports precise resets. Physical repeats require reporting reset variability and measurement uncertainty.

Physics judged only from rendered images remains an indirect test. A perception model used to extract trajectories can introduce correlated errors, and a language-model judge can prefer plausible-looking motion. Such evaluators should be validated against independent labels or measurements for the claimed quantity. A state predictor without a visual decoder can instead be assessed through its native physical outputs or a readout fitted only on training data. Requiring every architecture to produce a video would conflate query availability with dynamics quality.

\paragraph{Benchmark Tasks and Information Parity}
Physion illustrates evaluation of a physical event by asking a model to predict, from an initial visible segment, whether specified objects will contact during the unseen continuation, rather than requiring every model to generate a video~\citep{wmBear2021Physion}. The task permits comparisons among learned visual features, explicit states, and dynamical predictions, provided that their input privileges and readout training are reported. A model given simulator particle states has a different information budget from one given RGB observations. Their gap can help diagnose perception versus dynamics, but cannot be interpreted as a matched-input architectural ranking.

The same principle applies across model families. Spatial generators need held-out-view and revisit consistency tests with measured regions distinguished from completed regions. JEPA-style models need fixed, independently trained readouts or action-conditioned rollout tests, not only similarity to their own changing teacher. Joint world-action models need both policy evaluation and any separately claimed forecast evaluation. No single rendering metric, contact classifier, or robot success rate covers all these interfaces. A useful benchmark suite therefore combines common physical questions with native-interface tests and publishes what information each model receives.

\paragraph{Probability, Calibration, and Sharpness}
A forecast distribution should place probability on outcomes in a way that is both accurate and informative. Proper scoring rules reward the predictive distribution rather than a conveniently selected sample~\citep{wmGneiting2007}. For a binary event, a scalar observation, and its predictive distribution, two representative scores are
\begin{equation}
 \operatorname{BS}=\frac{1}{N_{\mathrm{eval}}}\sum_{i=1}^{N_{\mathrm{eval}}}(p_i-e_{i}^{\mathrm{obs}})^2,
 \qquad
 \operatorname{CRPS}(F_{\mathrm{CDF}},y)=
 \mathbb E|Y-y|-\tfrac12\mathbb E|Y-Y'|,
 \label{eq:wm-probability-scores}
\end{equation}
where BS is the Brier score, $p_i\in[0,1]$ the predicted probability of event $e_{i}^{\mathrm{obs}}\in\{0,1\}$, and $N_{\mathrm{eval}}$ the number of trials. CRPS is the continuous ranked probability score for a scalar measurement $y$ and predictive cumulative distribution $F_{\mathrm{CDF}}$. $Y$ and $Y'$ are independent draws from $F_{\mathrm{CDF}}$, assumed to have finite first moments. Lower values are better. CRPS can be estimated from samples without evaluating a density. A log score is appropriate when a normalized predictive density or probability mass is available. A diffusion training loss is not itself the log likelihood of a deployed forecast.

Calibration asks whether probabilities or intervals agree with empirical frequencies. For example, cases assigned a contact probability near a specified value should exhibit contact at approximately that frequency. A nominal prediction interval should be assessed by empirical coverage \emph{and} width, since an excessively broad interval can cover almost every outcome while being unhelpful. Calibration curves and post-hoc calibration techniques are useful diagnostics, but calibration on one distribution need not persist after a shift~\citep{wmGuo2017}. Marginal coverage alone does not establish correctness within each task, contact mode, horizon, or material group. Sample diversity and ensemble disagreement should therefore be tested against realized outcomes rather than interpreted as uncertainty quality by construction.

\paragraph{Splits, Aggregation, and Reproducibility}
Separate model fitting, calibration or threshold selection, and final testing. Group related frames and derived samples by trajectory, scene, object combination, and collection session as appropriate to the generalization claim. Report results by physical horizon, event type, and change of conditions before aggregating them. Confidence intervals should account for trajectory-level dependence. Treating adjacent frames as independent trials substantially overstates the amount of evidence. Include failure cases and invalid predictions in a declared accounting scheme rather than silently removing them from the denominator.

The rows in Table~\ref{tab:wm-validation} are complementary, not interchangeable levels of one score. A model may pass state-persistence tests while failing intervention tests, or yield useful decisions while offering no calibrated probability interface. The evaluation report should identify supported claims and missing evidence separately. Decision-level evaluation is developed next.

\subsubsection{Decision Utility and Applicability Boundaries}
\label{sec:wm-decision-validation}

The operational question is whether model calls improve real outcomes. Compare systems with and without the relevant world-model function while matching the policy proposal mechanism, observation history, action interface, training data, and inference budget as closely as possible. For a specified real-task score $J_{\mathrm{real}}$, define
\begin{equation}
 \Delta J=J_{\mathrm{real}}(\pi_{\mathrm{with\ model}})
          -J_{\mathrm{real}}(\pi_{\mathrm{reference}}),
 \label{eq:wm-utility-gain}
\end{equation}
where $\pi_{\mathrm{with\ model}}$ is the policy or complete decision procedure using the tested model function, $\pi_{\mathrm{reference}}$ the comparison procedure, and $\Delta J$ their difference under the same outcome definition and evaluation distribution. The score may be success rate, measured task quality, or another explicitly specified objective. It is not the model's own predicted reward. Report initial conditions, repetitions, uncertainty intervals, failures, interventions, and resource costs. A positive average gain can conceal systematic failure on the very candidates the model ranks most confidently.

Ablations should replace one function at a time. A persistence predictor tests the value of temporal evolution beyond current state. An action-shuffled predictor tests sensitivity to the supplied control. Removing future supervision tests its contribution to the learned representation. Matched history and parameter budgets help separate predictive learning from a larger encoder or additional context. These controls diagnose mechanisms but do not individually prove a causal explanation for an entire system's improvement.

Efficiency has several boundaries, including observation acquisition and transfer, encoding, predictive prefill, action or video sampling, communication, and physical execution. Report model-side action-update time separately from sensor-to-command latency, full query latency, and successful task throughput. Asynchronous prediction moves work off a critical path but does not remove its computation or the risk of stale predictions. Hardware, precision, batch size, context length, action horizon, denoising steps, and update rates should accompany timings. Neither the inverse of average inference time nor a high low-level servo rate establishes closed-loop responsiveness under all conditions.

Applicability boundaries should cover objects, materials, actions, temporal horizons, modalities, and embodiments. A model may generalize to new appearance while failing under new friction, or predict free motion while failing after a contact switch. Test combinations of mechanism changes as well as isolated shifts. Distinguish recovery through fresh observation, belief correction, physical-parameter identification, and network retraining. These require different resources and support different transfer claims. Uncertainty, modality disagreement, and residuals can trigger shorter horizons, additional sensing, or fallback, but their thresholds need validation against actual errors. Independent execution constraints and safety mechanisms remain system responsibilities.

Deployment validation should establish whether a world model improves outcomes under matched comparisons, within stated boundaries, and at a cost the system can afford. Its query interfaces must specify the limits of those claims.

\subsection{Current Challenges and Future Outlook}
\label{sec:wm-outlook}

The preceding sections suggest that progress should be described through capabilities and their evidence rather than a single architecture label. Broad pretrained priors, explicit physical structure, and efficient action interfaces contribute different strengths. Their integration raises questions about what can be learned from existing observations, what requires new intervention, and how a model can remain useful while its environment and operating conditions change.

\subsubsection{Acquiring Physical Knowledge from Observation to Intervention}
\label{sec:wm-physical-knowledge}

Videos provide rich evidence about appearance and motion but usually omit commanded forces, contact states, and other variables that determine the response to an intervention. Different mechanisms can explain the same observed trajectory. Larger observational datasets can improve coverage without resolving this ambiguity when the distinguishing variable is never measured or excited. The relevant challenge is to connect reusable visual and semantic priors with evidence that identifies executable action consequences.

Action-free pretraining followed by grounded interaction learning offers one division of labor, as illustrated by the separate representation and action-conditioned stages of V-JEPA~2~\citep{assran2025vjepa2}. Other systems can jointly train on several evidence types or transfer a prior through distillation. The choice should follow the gap to be closed. Unfamiliar object semantics, an unobserved material property, a new control interface, and a missing deformation mode are different problems. Transferring a representation does not remove the need to calibrate units, timing, and embodiment-dependent dynamics.

Multimodal measurements and targeted probes can expose distinctions that passive video leaves unresolved. A controlled lift with load measurements constrains mass. A slide under measured force constrains friction within a contact model. A thermal observation distinguishes visually identical process states. Such experiments should be chosen for their ability to separate plausible explanations, not merely to maximize visual novelty. Their cost and risk must be included in learning efficiency. For instruments and infrastructure, the intervention may be a process-setting change rather than a robot motion, but the same requirement holds. Predicted effects must be checked against the measurements relevant to the mechanism.

A further challenge is preserving uncertainty where evidence remains insufficient. A model should represent several compatible explanations instead of inventing a unique physical parameter from appearance. Uncertainty reduction through new sensing, identification through repeated interaction, and correction of structural model error should remain distinct. This distinction gives the closed loop a scientific role. Action tests an explanation, and the resulting discrepancy determines what knowledge should be revised.

\subsubsection{Reliable Consequence Prediction in Complex and Long-Horizon Processes}
\label{sec:wm-long-horizon}

Contact switches, non-rigid deformation, and coupled physical processes can turn small local errors into qualitatively different futures. Long-horizon prediction therefore requires more than a longer context window. It needs persistent state, representations of consequential events, appropriate time resolution, and uncertainty that remains useful as the forecast extends. A model may need to predict a phase boundary accurately without reproducing every intermediate texture.

The architectural routes in Table~\ref{tab:wm-architectural-routes} are likely to be combined in several ways, but no single combination is established as universally preferable. One design direction connects compact predictive states to detailed generative readouts on demand. Another combines object- or field-based physical structure with learned residual dynamics. A third coordinates slower semantic and process models with faster contact or motion models. Each requires an interface that preserves identity, units, physical time, and uncertainty. A hierarchy is not useful when its levels describe incompatible worlds.

Combining routes along any of these directions raises a representation question. A shared substrate can support understanding, prediction, and action, while specialized encoders and heads preserve the information each function needs. Shared parameters are only one implementation. Aligned representations, directed cross-expert communication, and training-only predictive supervision offer alternatives, as discussed in Section~\ref{sec:wm-unified}. The roadmap's distinction between functional roles and architectural choices is useful here. A common physical description should support compatible queries without forcing all modalities into one update frequency or every task into one decoding objective~\citep{wmChen2026Roadmap}.

Future comparisons should therefore test composition rather than only larger monolithic models. Does adding structure improve held-out contact response? Does a generative head retain information needed by a new query? Does selective prediction save computation without making action ranking less reliable? Longer horizons and richer modalities should be assessed at matched resource budgets, with prediction errors decomposed by mechanism. These tests can reveal when an interface is missing information, when a transition is wrong, and when extra detail simply does not help the task.

A persistent challenge is deciding what must remain explicit across abstraction levels. A coarse state may encode that liquid will spill, while a fine field predicts where it flows. The two outputs should refer to the same container, contents, and boundary conditions. Learned refinement must recover detail compatible with the coarse state, not silently revise the initial condition until a preferred prediction becomes possible. The model may retain raw evidence or an external memory so that a new query can request information omitted from its current compact state. This makes memory retrieval and adaptive resolution part of world modeling, rather than equating longer token context with a sufficient physical state.

A second challenge is modular identification. If a prediction fails, the system should distinguish a wrong observation interpretation, an incorrect physical parameter, a missing interaction mechanism, and an inadequate decision objective. Composable models make these distinctions easier to test when their interfaces have clear semantics. A monolithic model can offer the same diagnostic access through controlled readouts and interventions. The long-term comparison is therefore not simply modular versus end-to-end. It is how reliably a system can localize error, acquire the evidence needed to correct it, and preserve previously valid knowledge at an acceptable cost.

\subsubsection{Closing the Real-World Loop: Real-Time Operation, Transfer, and Continual Improvement}
\label{sec:wm-continual-loop}

A deployed world model participates in a continuing loop among observation, understanding, prediction, action, and external feedback. It may help identify an uncertain state, propose informative experiments, screen candidate actions, or supply imagined experience for policy learning. The physical process then produces evidence that the model did not choose. This separation between internal imagination and external verification is essential. Repeated agreement between a generator and an evaluator sharing its assumptions does not establish improved knowledge of the world.

Real-time operation requires allocating computation while the environment continues to change. Cached context, asynchronous experts, selective decoding, and short action chunks can reduce particular costs, but each creates a responsibility for freshness and correction. Transfer adds another problem. A new body, tool, sensor, or process can change what an action means. Reusable physical knowledge should be separated from the local parameters and interfaces that require new evidence. Systems such as InternW0 and InternW0-$\Delta$ show concrete ways to separate training and runtime work, not a guarantee that one update schedule transfers unchanged to every physical system.

Continual improvement should use diagnosis to choose the scale of an update. Fresh evidence may be enough to correct the current belief. Repeated discrepancies may justify identifying a small set of physical parameters. A persistent missing mechanism may require new data and network retraining. These responses should be recorded separately, together with the amount of real interaction, human intervention, and computation required. Updating every model after every observation can be wasteful, while never updating parameters can leave systematic errors unresolved.

A practical improvement cycle retains real failures and corrections, selects data that target identified gaps, updates the relevant model or policy components, and evaluates the candidate version on both new conditions and a retained regression suite. Synthetic rollouts can broaden practice near a known weakness, but acceptance must include evidence not produced by the same model. Replay and parameter regularization can help retain previous skills. They must still be checked for interference and loss of adaptability. Chapter~\ref{ch:physical-data} treats the data-side loop, while the present chapter specifies the predictive interfaces whose validity must survive each update.

The long-term aim extends beyond more elaborate imagined worlds. It is a system in which better representations produce more informative forecasts, forecasts support better real interventions, and the resulting evidence produces measurable correction. Progress should be reported through predictive accuracy, physical consistency, calibrated uncertainty, and improved physical outcomes, together with the costs of training, adaptation, operation, and recovery. This closes the chapter's organizing chain from state and transition to query and evidence.

These challenges converge on a single claim about priorities. Perception models tell an agent what is present. Policy models tell it what to do. A world model has the potential to help an agent understand what will happen, what would have happened instead, and why. In this sense the world model is one of the most important foundation models of Physical AI, providing a substrate on which perception, decision, and control can all stand. Its importance lies in the potential to connect testable predictions of the world with better decisions. More accurate predictions of task-relevant dynamics can support safer exploration, cheaper learning, and more reliable decisions across tasks and embodiments. For a world model to support decisions, its predictions need sufficient accuracy at the relevant resolution. Building machines that act in the world therefore depends, to a large extent, on building models of it and on making the limits of their evidence and predictions explicit.

\clearpage

\section{Foundation Models for Robot Manipulation}
\label{ch:manipulation-models}

Physical AI not only discusses how to perceive the world dynamics, but also instructs robots to actually interact with it. In robotics, manipulation is one of the important forms of physical action, which includes grasping, moving, assembling, and transforming objects through controlled contact, and beyond. This chapter examines foundation models for robot manipulation, which are the core of an embodied system that maps multimodal observations and task instructions to behavioral goals or low-level motor actions. Because open-ended environments present an unbounded variety of objects and tasks, engineering each manipulation skill individually is impractical. Instead, foundation models learn web-scale multimodal priors from physical robot trajectories, enabling a single pretrained model to be reused across tasks, environments and embodiments.

The chapter begins with the evolution of manipulation foundation models, focusing on the vision-language-action (VLA) architectures that dominate current practice. It then examines the pre-training and post-training that shape the model's capabilities, the model's adaptation across embodiments, and its deployment under real-time and resource constraints. It closes by applying these ideas to multi-fingered hands, where high-dimensional actuation, intricate contact transitions, and scarce demonstration data push every stage of the pipeline to its limits.

\begin{figure}[hb]
    \centering
    \includegraphics[width=\linewidth]{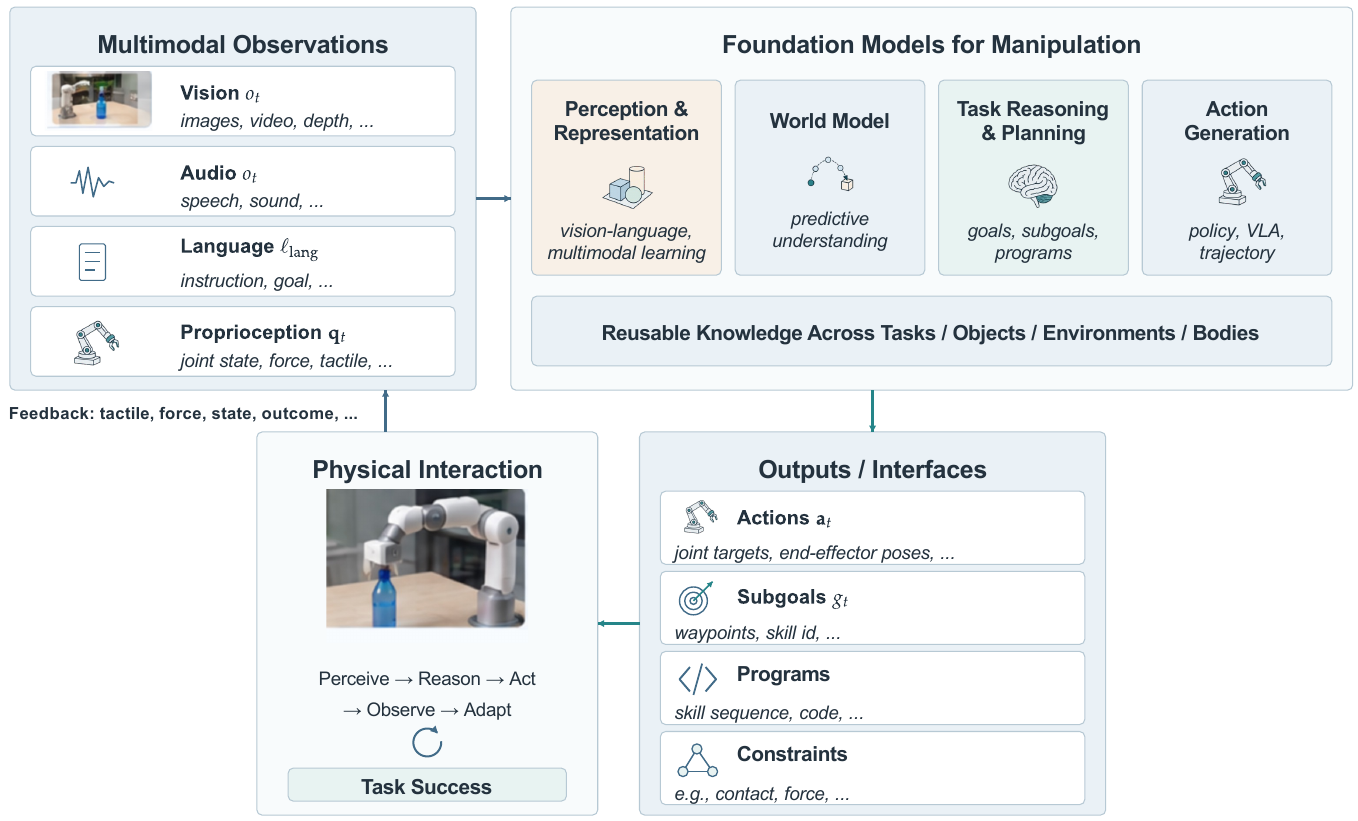}
    \caption{Foundation models for robot manipulation}
    \label{fig:chapter06-61img}
\end{figure}

\subsection{Emergence and Taxonomy}
\label{sec:brain-taxonomy}

An embodied agent should transform an instruction into a sequence of physically meaningful decisions. Recognizing a bottle is only one part of opening it. The agent must identify the requested outcome, choose a grasp and an opening strategy, interpret resistance during contact, and determine whether the cap has actually been released.  Foundation models for robot manipulation aim to make these capabilities reusable across tasks, objects, environments and sometimes even bodies. They bring together multimodal representation learning, generalist robot policies and models of action consequences. Figure~\ref{fig:chapter06-61img} provides an overview of foundation models for robot manipulation. The key question is how broadly acquired knowledge can guide ongoing physical interactions.

\subsubsection{Conceptual Scope and Enabling Conditions}

In this chapter, a \emph{foundation model for robot manipulation} is a pretrained model that can adapt to new tasks. It draws on multimodal evidence and task context to make task-relevant decisions, set behavioral goals, or generate robot actions.  The foundation model property concerns the reuse of learned structure. A single model should provide a useful starting point for many downstream tasks or operating conditions. Neither a large parameter count nor a natural-language interface establishes this property on its own.

Let $o_t$ denote exteroceptive observations, $\mathbf{q}_t$ denote proprioceptive measurements, $\mathbf{a}_t$ denote an action at the model's declared control interface, $\ell_{\mathrm{lang}}$ denote a task instruction, and $e$ describe the embodiment. The interaction history is defined as follows.
\begin{equation}
  h_t=(o_{0:t},\mathbf{q}_{0:t},\mathbf{a}_{0:t-1}),\qquad
  \mathbf{A}_{t}^{\mathrm{seq}}=[\mathbf{a}_t^\top;\ldots;\mathbf{a}_{t+H-1}^\top].
  \label{eq:brain-history}
\end{equation}
An action-producing model implements $\pi_\theta(\mathbf{A}_{t}^{\mathrm{seq}}\mid h_t,\ell_{\mathrm{lang}},e)$. A higher-level model may instead output a subgoal $g_t$, a skill identifier, a program, or a constraint on the desired interaction. These outputs are different interfaces to physical behavior, and should not be treated as interchangeable. An instruction to ``grasp the bottle'' leaves substantially more work to downstream modules than a sequence of end-effector poses or joint targets.

Four developments have made this reuse practical. Vision-language pretraining supplies broad semantic representations. Increasingly diverse robot datasets provide action-grounded supervision. Sequence and generative models capture temporal and multimodal behavior. Training infrastructure and simulation make repeated optimization and controlled data variation feasible. These conditions are complementary. Additional computation cannot recover action semantics that were never recorded, and simulated diversity is useful only to the extent that learned behavior transfers to the target physical conditions.

The distinction from adjacent chapters follows these interfaces. The perception models of Chapter~\ref{ch:perception} extract task-relevant evidence from observations, while Chapter~\ref{ch:physical-data} addresses the data that support learning. The world models of Chapter~\ref{ch:wm} represent conditional evolution and answer questions about possible consequences. Foundation models for robot manipulation use evidence and predictive knowledge to decide what behavior to pursue. The foundation models for whole-body motor control discussed in Chapter~\ref{ch:whole-body-motor-control} regulate the realization of that behavior under body dynamics and contact constraints. One network may implement several of these functions, and a VLA model may extend into motion generation. The distinction concerns the responsibility of an output, not the size, name, or location of a network. Chapter~\ref{ch:physical-agent} then considers how these capabilities are assembled into agents that manage memory, tools, and extended activities.

For example, during cap opening, task reasoning first identifies which bottle to open and what the desired open state is. Prediction then estimates what will happen if the agent grasps and rotates it. Action generation proposes a motion, and local control adjusts that motion based on contact feedback. The same tactile event may cause either a local correction, such as reducing grip force, or a task-level revision, such as regrasping. Effective coordination therefore requires feedback in both directions.

\subsubsection{From Specialized Policies to Reusable Embodied Models}

\begin{figure}[t]
    \centering
    \includegraphics[width=\linewidth]{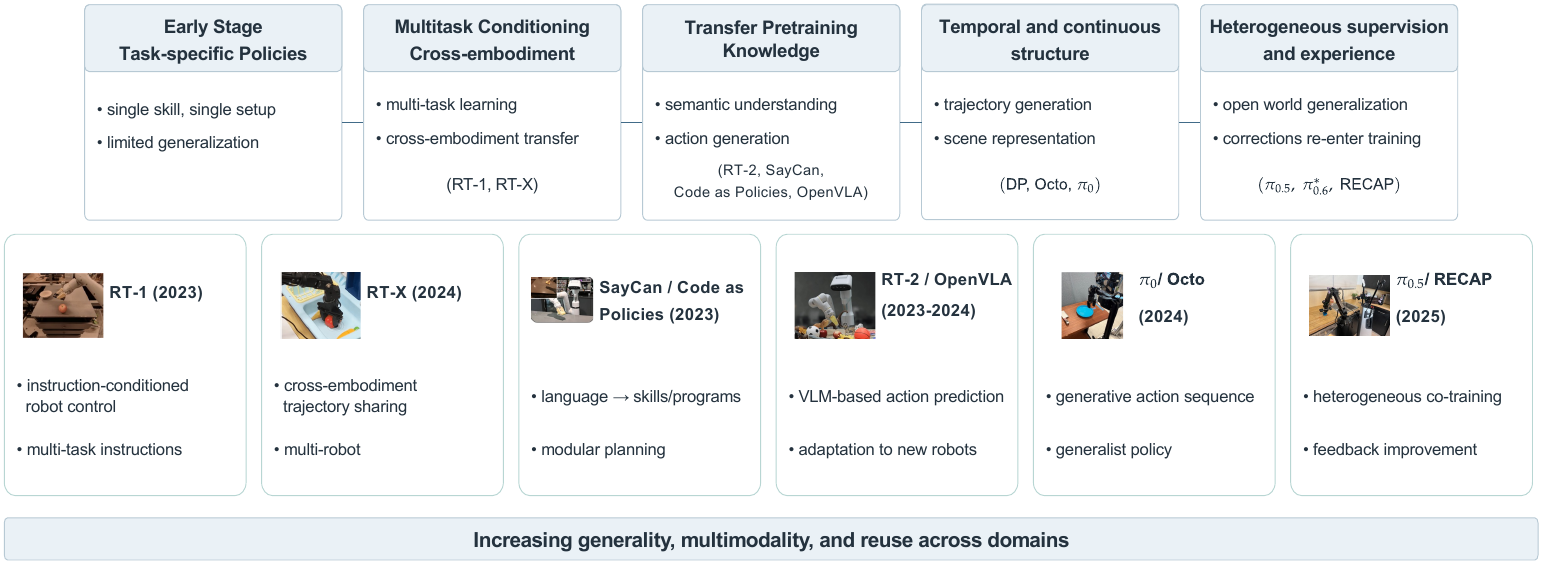}
    \caption{Algorithm development framework}
    \label{fig:chapter06-62img}
\end{figure}
Figure~\ref{fig:chapter06-62img} summarizes the transition from specialized visuomotor policies to reusable embodied models.

These early learned visuomotor policies often acquire one skill on one setup. Their representations are optimized for a narrow demonstration distribution, so even a change in camera, object, instruction, or robot can require substantial retraining. Multitask policies change the unit of learning from an isolated skill to a task-conditioned behavioral repertoire. RT-1 demonstrates instruction-conditioned robot control at scale, while the Open X-Embodiment study investigates the transfer available from pooling trajectories across robot platforms~\citep{brohan2023rt1,openx2024}. These developments make data diversity and interface compatibility central modeling issues.

A second development is the transfer of semantic knowledge from language and vision-language pretraining. In modular systems, a language model can select grounded skills or generate programs that invoke perception and control functions. SayCan and Code as Policies illustrate these two approaches~\citep{ichter2023saycan,liang2023code}. In direct action models, \citet{brohan2023rt2} incorporate robot action prediction into vision-language training with RT-2, and \citet{kim2024openvla} release OpenVLA as a pretrained VLM-based policy available for adaptation. This kind of transfer can broaden object and instruction understanding, but it does not supply the missing dynamics or contact experience of an unfamiliar robot.

A third development focuses on temporal structure and continuous behavior. Diffusion Policy demonstrates multimodal action-sequence generation for visuomotor control, Octo combines a generalist policy with adaptable observation and action interfaces, and $\pi_0$ connects a pretrained VLM to a flow-based action expert~\citep{chi2023diffusion,octo2024,black2024pi0}. These designs make a useful distinction between the representation needed to understand a scene and the computation needed to generate a coherent trajectory. They also show that a foundation model for robotics need not be an autoregressive language model throughout.

Finally, broad pretraining is increasingly connected with heterogeneous supervision and deployment experience. $\pi_{0.5}$ studies open-world generalization through heterogeneous co-training, while the RECAP approach underlying $\pi^{*}_{0.6}$ uses experience and corrections to improve a pretrained policy~\citep{pi2025pi05,physicalintelligence2025b}. The progression is consequently not a simple replacement of control by larger language models. It is the integration of semantic priors, physical action interfaces, predictive representations, and feedback-driven improvement. Each component addresses a different limitation of demonstration-only learning.

\subsubsection{Major Model Lineages and Architectural Evolution}

The main model lineages can be compared along the three axes of primary knowledge source, learning objective, and decision interface. Generalist policies emphasize reusable behavior learned from robot trajectories. VLAs introduce broad vision-language knowledge into action generation. And world-action models couple behavior with predictions of environmental change. Table~\ref{tab:brain-taxonomy} summarizes these emphases. The categories overlap. A generalist policy can be a VLA, and a VLA can contain a world model. Direct, hierarchical, and prediction-assisted execution are architectural choices that cut across all three lineages.

\begin{table}[!htbp]
  \centering
  \caption{Three model lineages compared by knowledge source, learning target, and decision interface. The categories are complementary rather than exclusive}
  \label{tab:brain-taxonomy}
  \small
  \renewcommand{\arraystretch}{1.15}
  \begin{tabularx}{\textwidth}{@{}>{\raggedright\arraybackslash}p{0.17\textwidth} >{\raggedright\arraybackslash}X >{\raggedright\arraybackslash}X >{\raggedright\arraybackslash}X@{}}
    \toprule
    \textbf{Lineage} & \textbf{Knowledge source} & \textbf{Main learning target} & \textbf{Decision interface} \\
    \midrule
    Generalist robot policy & Diverse robot trajectories, task labels & Shared behavioral priors across tasks and setups & Actions, chunks, or skills\newline RT-1, RT-X, Octo \\
    Vision-language-action model & Vision-language pretraining plus robot interaction & Semantic grounding and conditioned action generation & Action tokens or continuous trajectories\newline RT-2, OpenVLA, $\pi_0$ \\
    World-action model & Video and interaction data with available action labels & Coupled action and future-state prediction & Actions with predicted outcomes, or prediction-assisted decisions\newline Unified World Models \\
    \bottomrule
  \end{tabularx}
\end{table}

\paragraph{Generalist Robot Policies}
These policies learn primarily from collections of robot trajectories, usually conditioned on language, goal images, or task identifiers. Shared training allows reaching, grasping, placing, and other recurring structures to contribute to multiple tasks. RT-X explores such sharing across embodiments, and Octo provides a policy initialization that can be adapted to new observation and action spaces~\citep{openx2024,octo2024}. Generality here comes from how broadly the behavior covers and transfers, not from conversational ability. Language may serve as a conditioning representation without making the policy itself a pretrained language model. The main limitation is coverage. Robot trajectories provide physically grounded actions but are costly to collect and can be concentrated in a few environments or control interfaces.

\paragraph{Vision-Language-Action Models}
These models connect visual observations and linguistic goals with a declared action space. A prominent route initializes the policy from a pretrained VLM so that the policy does not have to learn object knowledge and instruction semantics entirely from robot demonstrations. Action supervision then grounds that knowledge in the robot's control interface. The phrase VLA specifies a multimodal policy interface more clearly than it specifies a single architecture. Discrete autoregression, deterministic regression, diffusion, and flow matching are alternative ways of producing actions within this broad class. Their differences concern precision, multimodality, learning signals, and execution cost, as developed in Section~\ref{sec:brain-vla}. Semantic transfer is especially useful for interpreting an unfamiliar instruction, but successful interpretation still requires a motor repertoire capable of realizing it.

\paragraph{Predictive and Joint World-Action Models}
A world model can choose actions in several ways, including evaluating proposed actions, generating subgoal observations, or sharing a representation with a policy. Joint video-action modeling, illustrated by Unified World Models, couples predictive and behavioral learning~\citep{zhu2025uwm}. These couplings should be distinguished. A model that predicts a future image that appears to satisfy the task goal does not thereby show that the robot can reach that image. Likewise, predicting a future and an action together does not automatically provide a calibrated answer to every counterfactual action query. The benefit of prediction must be demonstrated through better decisions or more transferable representations.

One formulation models $p_\theta(\mathbf{A}_{t}^{\mathrm{seq}},O_t^{+}\mid h_t,\ell_{\mathrm{lang}},e)$, where $O_t^{+}$ denotes future observations over a declared horizon. With appropriate conditioning and training, the model can relate a proposed behavior to its expected outcome, or use a desired outcome to help infer behavior. Action-free video can supervise the predictive component while action-labeled robot data ground executable decisions. This is a potential advantage of joint learning, but the missing robot actions in human video cannot be recovered merely by renaming visual motion as an action. Viewpoint, embodiment, and unobserved forces still constrain transfer.

\paragraph{Architectures for Organizing Decisions}
A direct policy maps the current task context straight to actions. A hierarchical system exposes subgoals or skills between task reasoning and motion generation. A prediction-assisted system brings anticipated outcomes into learning or action selection. Language and program-mediated planners can compose existing behaviors, provided that their calls are grounded in perception, skill preconditions, and execution feedback. A higher-level process can also maintain context while a faster process updates motion using recent observations. GR00T N1 provides an example of tightly coupled vision-language and action-generation modules~\citep{bjorck2025gr00t}. Gemini Robotics 1.5 illustrates the coordination of embodied reasoning with action capabilities~\citep{gemini2025robotics15}. Architectural separation does not mean asynchronous execution. Whether modules run independently, share cached features, or wait at synchronization points is a deployment property that must be specified separately.

\paragraph{Connecting Policy, Prediction, and Evaluation}

The relationships among these arrangements can be expressed with the following shared task representation and three possible output interfaces
\begin{equation}
\begin{aligned}
  z_t &= f_\theta(h_t,\ell_{\mathrm{lang}},e), \\
  \mathbf{A}_{t}^{\mathrm{seq}} &\sim \pi_\psi(\cdot\mid z_t,\ell_{\mathrm{lang}},e), \\
  \widehat z_{t+1:t+H} &\sim p_\phi(\cdot\mid z_t,\mathbf{A}_{t}^{\mathrm{seq}},e), \\
  \widehat v_t &= V_{\omega_{\mathrm{critic}}}(z_t,\ell_{\mathrm{lang}},e).
\end{aligned}
\label{eq:brain-policy-world-value}
\end{equation}
Here $z_t$ summarizes available evidence, the policy proposes behavior, the transition model predicts its consequences, and the evaluator estimates a task-related quantity. Not every system contains all three heads, and the equations do not require shared parameters. They distinguish functions that are often conflated in discussion of unified models.

There are several ways to connect them. A policy may use a predicted subgoal as conditioning, a planner may score sampled action chunks through a world model, or predictive training may improve an encoder that is used without online rollout. In a joint generative model, what the model samples (an action, a future observation, or both) depends on how it is conditioned and masked. Each choice changes which questions can be asked at inference time and which computational costs lie on the control path.

Evaluation also requires precise semantics. A progress score estimates how far an observed process has advanced. A success classifier estimates whether a terminal criterion holds. A value function estimates future return under a specified policy and reward. A model trained for one of these tasks is not automatically a substitute for the others. During cap opening, observing rotation may indicate motion without demonstrating release. Outcome feedback must therefore be grounded in the property the task actually requires.

\paragraph{Interpreting Generalization Claims}

Embodied generalization has several independent dimensions. A model may transfer to new object appearances while still relying on the same workspace and robot. It might understand a new composition of familiar instructions while lacking the corresponding motor skill. Or it might adapt to a new embodiment only after target demonstrations. Claims should specify which factors change and which remain fixed. In particular, zero-shot execution, few-shot adaptation, and large-scale fine-tuning are distinct capabilities.

Closed-loop evaluation should test the decisions that make reuse valuable, including selecting the intended object among distractors, completing unfamiliar combinations of known stages, revising behavior after a disturbance, and transferring with a stated amount of new data. A useful experimental design varies task, environment, and embodiment separately before testing their combination. Retaining performance on previously supported conditions matters as much as improvement on the new condition. Offline action prediction error alone cannot establish these properties because it does not measure the states visited by the executing policy.

The remainder of this chapter follows the resulting development path. Section~\ref{sec:brain-vla} examines action representations and multimodal architectures. Section~\ref{sec:manipulation-training} explains how different supervision sources shape capabilities. Section~\ref{sec:brain-embodiment-adaptation} addresses the physical meaning of transfer between bodies. Section~\ref{sec:brain-deployment} then asks how these capabilities can be delivered within the timing and resource constraints of an actual robot.

\subsection{Vision-Language-Action Model}
\label{sec:brain-vla}

Many current VLA models start from a pretrained VLM and extend it to output actions, a modality its pretraining never covered. This section begins with how actions are represented and generated, then turns to how they are combined with visual and linguistic inputs in a single sequence model, and closes with how this model is extended to sensory inputs beyond images and language.

\subsubsection{Action Representation and Generation}
\label{sec:action-representation}

Within a VLA, action is treated as a third modality alongside images and language, although it behaves quite differently from both. Actions are continuous rather than symbolic, their dimensionality and control rate vary with the embodiment, and consecutive actions are sampled at high frequency and are strongly correlated. Since action is also the model's output, how it is represented decides both how demonstrations are encoded during training and how commands are decoded at run time. Writing $\mathbf{a}_t\in\mathbb{R}^{d}$ for the action at step $t$ and $\mathbf{A}_{t}^{\mathrm{seq}}=[\mathbf{a}_t^\top;\ldots;\mathbf{a}_{t+H-1}^\top]$ for an action chunk of horizon $H$, modern VLAs predict a whole chunk at once instead of a single step~\citep{zhao2023,chi2023diffusion}. Doing so limits the accumulation of errors over time, produces smoother motion, and spreads the cost of one forward pass over $H$ control steps. Existing methods fall into two families according to how $\mathbf{A}_{t}^{\mathrm{seq}}$ is fed into the sequence model and how it is generated. The \emph{discrete} route converts $\mathbf{A}_{t}^{\mathrm{seq}}$ into tokens that the language backbone predicts, whereas the \emph{continuous} route lets a dedicated head generate real-valued chunks directly. The two routes trade off training efficiency, inference speed and the preservation of the pre-trained VLM in opposite ways, which is why recent hybrid schemes combine discrete training with continuous execution.

\paragraph{Discrete Action Representations}
Once a chunk is mapped to tokens $z_{1:N_{\mathrm{acttok}}}^{\mathrm{act}}$, the VLA is trained with the ordinary next-token objective
\begin{equation}
  \mathcal{L}_{\mathrm{AR}}
  = -\sum_{k=1}^{N_{\mathrm{acttok}}}\log p_\theta\big(z_{k}^{\mathrm{act}} \,\big|\, z_{<k}^{\mathrm{act}},\, o_t,\, \ell_{\mathrm{lang}}\big).
\end{equation}
This requires no change to the VLM architecture, and since robot data and web data are trained with the same loss, the backbone remains a language model shaped by both. The methods differ mainly in the tokenizer that produces $z_{1:N_{\mathrm{acttok}}}^{\mathrm{act}}$. The simplest, \emph{per-dimension binning}, splits each action dimension into a fixed number of uniform bins (256 in RT-1~\citep{brohan2023rt1}, RT-2~\citep{brohan2023rt2} and OpenVLA~\citep{kim2024openvla}). RT-2 writes the bin indices as text, so the VLM outputs an action as a string of integers, while OpenVLA reassigns the least-used tokens of the Llama vocabulary to action bins. Binning is easy to implement but scales poorly, because a chunk becomes $H\cdot d$ tokens that must be decoded one after another, neighboring tokens at high control rates are nearly identical and therefore carry little learning signal, and precision can never exceed the bin width. \emph{Compression-based tokenization} turns this redundancy into an advantage. FAST~\citep{pertsch2025fast_ch6} applies a discrete cosine transform to each action dimension, quantizes the coefficients so that most high-frequency ones become zero, and compresses the result with byte-pair encoding. The resulting sequences are much shorter than those produced by binning, the same tokenizer transfers across embodiments, and $\pi_0$-FAST reaches the performance of the flow-matching $\pi_0$ with less training compute. \emph{Learned quantizers} take a different approach and train a vector-quantized autoencoder~\citep{van2017neural} whose codebook entries act as short motion primitives. Residual vector quantization (RVQ) increases resolution by quantizing the remaining error over several stages, and VQ-VLA~\citep{wang2025vq} shows that training the tokenizer on more real and simulated trajectories improves both long-horizon success and smoothness. The price is an embodiment-specific decoder that has to be deployed together with the policy. Sequential autoregressive decoding of long action-token chunks through a billion-parameter backbone can incur substantial inference latency. Parallel decoding in OpenVLA-OFT~\citep{kim2025oft_ch6} predicts all action tokens in a single pass and thus raises throughput, but it no longer models the joint distribution over tokens explicitly.

\paragraph{Continuous Action Representations}
On the continuous route, an output module produces $\mathbf{A}_{t}^{\mathrm{seq}}\in\mathbb{R}^{H\times d}$ directly. A simple option is an $L_1$ action-regression head, as used in OpenVLA-OFT~\citep{kim2025oft_ch6}. For direct point regression, expected elementwise $L_1$ loss is minimized by conditional coordinatewise medians. Action Chunking with Transformers (ACT) uses an $L_1$ reconstruction term together with a weighted KL-divergence term within a conditional variational autoencoder~\citep{zhao2023}. Such methods can perform well on benchmarks, but a single point prediction need not capture distinct valid action modes. Other generative heads model the full, possibly multimodal, action distribution. Diffusion Policy~\citep{chi2023diffusion} denoises action chunks conditioned on observations, and Octo~\citep{octo2024}, RDT-1B~\citep{liu2025rdt} and GR00T~N1~\citep{bjorck2025gr00t} scale the idea up with diffusion transformers. Within VLAs, the reference design is $\pi_0$~\citep{black2024pi0}, in which a PaliGemma backbone encodes images and language and a separate, smaller \emph{action expert} attends to the backbone tokens and is trained with conditional flow matching. With an interpolant $\mathbf{A}_{t}^{\mathrm{seq},\tau_{\mathrm{flow}}}=\tau_{\mathrm{flow}} \mathbf{A}_{t}^{\mathrm{seq}}+(1-\tau_{\mathrm{flow}})\boldsymbol{\varepsilon}$, $\boldsymbol{\varepsilon}\in\mathbb{R}^{H\times d}$ with independent entries $\varepsilon_{ij}\sim\mathcal{N}(0,1)$, the expert learns the velocity field
\begin{equation}
  \mathcal{L}_{\mathrm{FM}}
  = \mathbb{E}_{\tau_{\mathrm{flow}},\boldsymbol{\varepsilon}}\big\|\, \mathbf{v}_{\theta}^{\mathrm{flow}}(\mathbf{A}_{t}^{\mathrm{seq},\tau_{\mathrm{flow}}},\tau_{\mathrm{flow}},o_t,\ell_{\mathrm{lang}})
    - (\mathbf{A}_{t}^{\mathrm{seq}}-\boldsymbol{\varepsilon})\big\|_{\mathrm F}^{2},
\end{equation}
where $\|\cdot\|_{\mathrm F}^{2}$ sums squared entries over the $H$ action steps and $d$ action coordinates. At inference the velocity field is integrated with a few Euler steps, and because the backbone prefix is cached, each step requires a forward pass through the expert only. Diffusion and flow-matching action heads avoid action-bin quantization and can represent multimodal action distributions. Compact action expert modules and a small number of sampling steps can make inference efficient. However, in the cited VLA comparisons, training with a continuous action expert converges more slowly than autoregressive action-token training, and gradients from a randomly initialized expert can interfere with the pretrained backbone. Gradient control and co-training with VLM data mitigate this interference in the reported settings~\citep{driess2025insulation_ch6}. Co-training with web data, or interleaving actions with language reasoning, also requires additional machinery. It is therefore natural to use tokens for learning and continuous generation for execution, and several recent systems combine the two in this way. $\pi_{0.5}$~\citep{pi2025pi05} pre-trains the whole model autoregressively on FAST tokens and adds a flow-matching expert only during post-training, where it optimizes $\mathcal{L}=\mathcal{L}_{\mathrm{AR}}+\alpha_{\mathrm{mix}}\mathcal{L}_{\mathrm{FM}}$ with the two action representations masked from each other. At run time the model first decodes text and then runs a few denoising steps of the expert. Knowledge insulation~\citep{driess2025insulation_ch6} merges the two stages into one.  The backbone is trained on FAST tokens and vision-language data while a stop-gradient prevents the flow-matching loss from reaching it. The reported model trains faster, runs faster, and generalizes better than either pure route. These systems point toward a division of labor in which discrete tokens shape the backbone's representation and a lightweight continuous generator handles control. Which tokenizer shapes the backbone best, and how the two representations should be scheduled across pre-training, post-training and continual learning, are still open questions.

\begin{figure}[t]
    \centering
    \includegraphics[width=\linewidth]{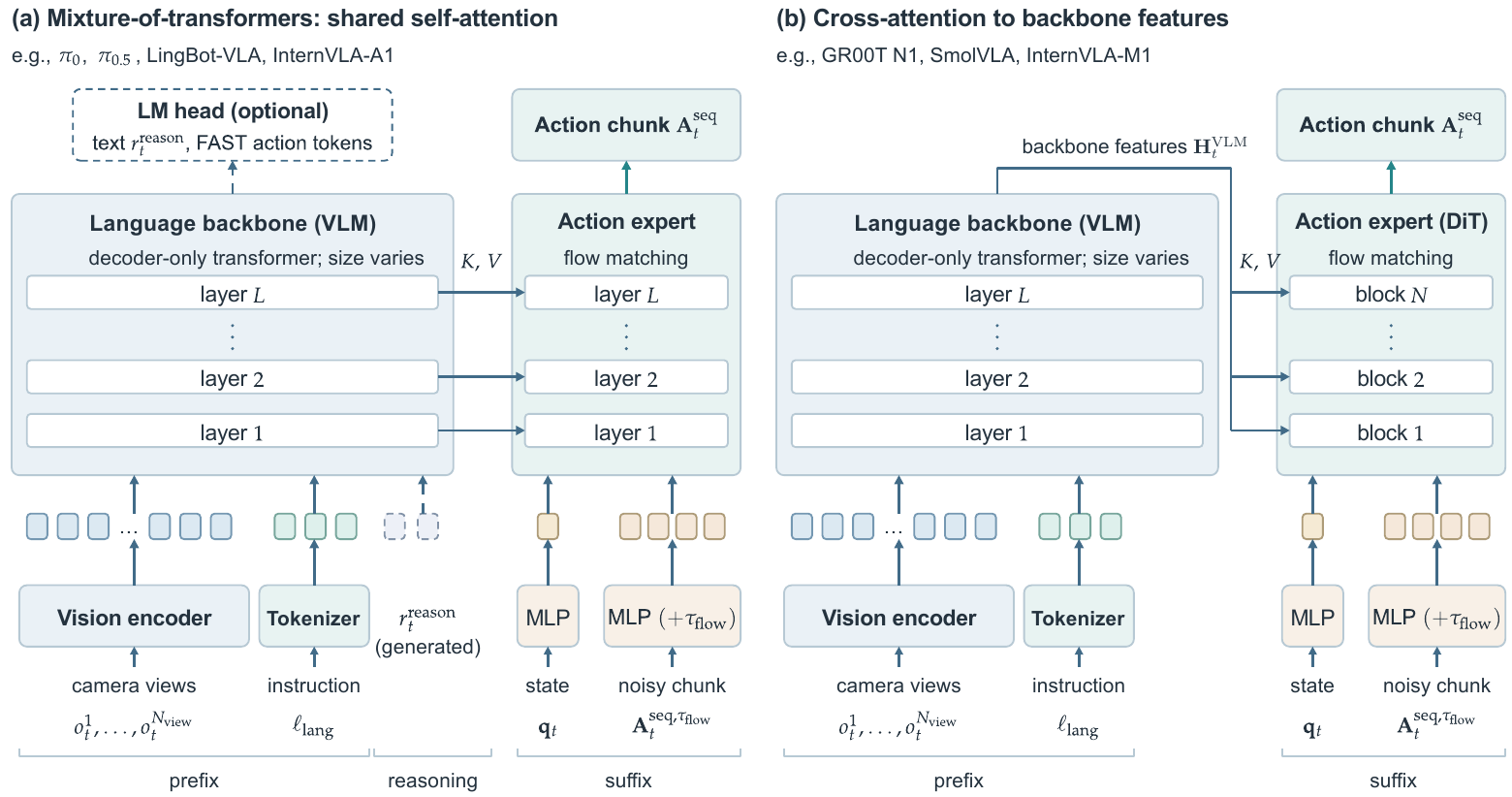}
    \caption{Two mainstream VLA structures. (a)~In a mixture-of-transformers structure, the expert has its own weights but the same depth as the backbone, and in every layer its tokens attend to the backbone's keys and values through a shared self-attention. The backbone can additionally generate reasoning text $r_{t}^{\mathrm{reason}}$ or FAST action tokens through an optional language-model head. (b)~In a cross-attention structure, a separate diffusion transformer reads backbone token-feature matrix $\mathbf H_t^{\mathrm{VLM}}$ through cross-attention, so its depth, size and rate are independent of the backbone}
    \label{fig:chap6-vla-arch}
\end{figure}

\subsubsection{Paradigm of Vision-Language-Action Models}
\label{sec:vla-paradigm}

The VLA paradigm starts from a simple observation. A policy $\pi_\theta(\mathbf{A}_{t}^{\mathrm{seq}}\mid o_t,\ell_{\mathrm{lang}})$ that maps images $o_t$ and an instruction $\ell_{\mathrm{lang}}$ to an action chunk $\mathbf{A}_{t}^{\mathrm{seq}}$ takes the same inputs as a vision-language model, so a policy initialized from a VLM and fine-tuned on robot demonstrations can inherit knowledge that the VLM has learned from billions of web image-text pairs. Almost every current VLA pairs the vision encoder and language backbone of a VLM with an action output module and couples the two in one of two mainstream ways (Figure~\ref{fig:chap6-vla-arch}). This section first describes this common template before tracing its development.

\paragraph{Components}
Both structures in Figure~\ref{fig:chap6-vla-arch} share three components. The first is a \emph{vision encoder}, usually the SigLIP or DINOv2 vision transformer (ViT) inherited from the VLM, which maps each camera image to a few hundred patch tokens. When there are several views, each is encoded separately and the tokens are concatenated. The second is a \emph{vision-language backbone}, typically a decoder-only transformer with 2 to 7 billion parameters in open models (Llama~2, Qwen-VL, PaliGemma) and a frontier-scale model in closed systems, which processes the image tokens together with the tokenized instruction and produces the representation from which actions are read out. The third is an \emph{action output module}, one of the heads described in Section~\ref{sec:action-representation}, which turns this representation into an action chunk. Discrete action tokens need nothing beyond the backbone's own language-model head, as in RT-2 and OpenVLA, whereas current generalist VLAs mostly follow the continuous route and add a smaller action expert trained with flow matching or diffusion. Proprioceptive state $\mathbf{q}_t$ enters alongside them, either as discretized tokens in the prefix or through a small embodiment-specific multilayer perceptron (MLP), and other sensing modalities are deferred to Section~\ref{sec:multimodal-inputs}.

\paragraph{Two Mainstream Structures}
With a continuous expert in place, the key architectural question becomes how the expert reads from the backbone, and current VLAs answer it in one of two ways (Figure~\ref{fig:chap6-vla-arch}). In a \emph{mixture-of-transformers} (MoT) design ($\pi_0$~\citep{black2024pi0}, $\pi_{0.5}$~\citep{pi2025pi05}, LingBot-VLA~\citep{wu2026pragmatic}, InternVLA-A1~\citep{cai2026internvla}), the expert is a narrower transformer with the same depth as the backbone. The two keep separate weights but share a single self-attention operation in every layer, so the action tokens simply form another block of the same sequence and read the backbone's keys and values layer by layer (Figure~\ref{fig:chap6-vla-arch}a). This gives the tightest possible coupling, since the expert can draw on every level of the backbone's representation and the backbone's language-model head remains available for reasoning text or FAST tokens in the same sequence. The downside is that the expert is bound to the backbone's depth and clock, and its gradients reach every backbone layer, which is what knowledge insulation~\citep{driess2025insulation_ch6} is designed to stop. In a \emph{cross-attention} design (GR00T~N1~\citep{bjorck2025gr00t}, SmolVLA~\citep{shukor2025smolvla}, InternVLA-M1~\citep{chen2025internvla}), the expert is a separate diffusion transformer (DiT) over state and action tokens that reads backbone token-feature matrix $\mathbf H_t^{\mathrm{VLM}}$ through cross-attention layers interleaved with self-attention (Figure~\ref{fig:chap6-vla-arch}b). The position and width of this interface are then free design choices, and published systems make them differently. GR00T~N1 conditions its DiT on middle-layer rather than final-layer features of the VLM, and SmolVLA keeps only the lower half of a small VLM's layers so that the whole model has just 0.45B parameters. InternVLA-M1 inserts a querying transformer whose learnable queries cross-attend to intermediate layers of a 3B VLM, passes the resulting fixed set of tokens to an 86M diffusion action expert, and attenuates the expert's gradients before they reach the backbone. Decoupling allows the expert to be sized and clocked independently of the VLM, which dual-system layouts exploit by running a slow VLM above a fast expert, as GR00T~N1 does. The price is a narrower channel, because the expert sees the backbone only through the features or queries it is given and does not take part in the backbone's own attention. In practice, MoT is preferred when the expert and the backbone are trained jointly on one sequence, and cross-attention when the expert has to be small, fast or run at its own rate.

\paragraph{Sequence Organization}
Whatever the structure, the inputs of one control step are organized as one ordered sequence, canonically
\begin{equation}
  \underbrace{[\,\mathrm{img}(o_t^{1}),\ldots,\mathrm{img}(o_{t}^{N_{\mathrm{view}}}),\ \mathrm{txt}(\ell_{\mathrm{lang}})\,]}_{\text{prefix}}\;
  \underbrace{[\,\mathrm{txt}(r_{t}^{\mathrm{reason}})\,]}_{\text{reasoning}}\;
  \underbrace{[\,\mathrm{emb}(\mathbf{q}_t),\ \mathrm{emb}(\mathbf{A}_{t}^{\mathrm{seq},\tau_{\mathrm{flow}}})\,]}_{\text{suffix}},
\end{equation}
where $N_{\mathrm{view}}$ is the number of views, $r_{t}^{\mathrm{reason}}$ is an optional intermediate text output (a subtask, plan or chain of thought), and $\mathbf{A}_{t}^{\mathrm{seq},\tau_{\mathrm{flow}}}$ is the noised continuous action chunk at flow time $\tau_{\mathrm{flow}}$. Discrete action tokens instead use $z_{1:N_{\mathrm{acttok}}}^{\mathrm{act}}$ as defined above. In an MoT the whole sequence passes through shared self-attention, which follows a block mask rather than a plain causal one. Prefix tokens attend to each other bidirectionally, as in prefix-LM training of PaliGemma-style VLMs, while the reasoning block is causal because it is generated token by token. The suffix attends to everything before it and bidirectionally within itself, since all $H$ actions are produced together, but no earlier block attends to the suffix. This keeps the prefix key-value cache independent of the actions, so that it can be reused across denoising steps. A cross-attention design arrives at the same structure by construction, because the backbone processes only the prefix and the reasoning block while the expert processes the suffix and reads the backbone through cross-attention. Here too the backbone features are computed once per observation and reused across denoising steps. When discrete and continuous representations coexist in training, as in $\pi_{0.5}$~\citep{pi2025pi05} and knowledge insulation~\citep{driess2025insulation_ch6}, the two suffixes are additionally masked from each other so that the flow expert cannot copy the FAST tokens.

\paragraph{Development}
The template described above takes shape over several years. RT-1~\citep{brohan2023rt1} predates the use of VLMs but sets conventions that later work retains. A transformer with 35 million parameters, trained on about 130,000 teleoperated episodes from 13 robots, learns to follow several hundred instructions. Its actions are discretized into 256 bins per dimension and emitted as tokens. Its generalization grows with the breadth of tasks and robots rather than with model size. RT-2~\citep{brohan2023rt2} then co-fine-tunes internet-scale VLMs (PaLI-X 55B, PaLM-E 12B~\citep{driess2023palme}) on web data together with robot episodes whose actions are written as integer strings, and the resulting policy generalizes semantically in ways that no robot dataset could have taught it, for instance picking up the object ``that could be used as a hammer''. RT-2, however, is neither open nor backed by shared data. Open X-Embodiment~\citep{openx2024} addresses the data gap by pooling 60 datasets from 22 embodiments into more than a million episodes and showing positive transfer across robots, and OpenVLA~\citep{kim2024openvla} addresses openness by releasing a 7B open-weight policy trained on this corpus.

The current generation keeps this foundation but adopts continuous action experts and trains on data far beyond public collections, and its progress has come on three fronts. The first front is scale, starting with $\pi_0$~\citep{black2024pi0}, which pairs a 3B PaliGemma backbone with a 300M flow-matching expert and trains it on roughly 10k hours of data from 7 robot configurations. $\pi_{0.5}$~\citep{pi2025pi05} adds heterogeneous co-training and subtask prediction, which allows the same policy to work in homes that had not appeared in its training data, GR-3~\citep{cheang2025gr} reproduces the template on a Qwen2.5-VL backbone, and LingBot-VLA~\citep{wu2026pragmatic} reports steady gains as its bimanual data grow from 3k to 20k hours, with no sign of saturation.

The second front is hierarchical control, motivated by whole-body and long-horizon tasks for which a single synchronous model is too slow. GR00T~N1~\citep{bjorck2025gr00t} establishes the open dual-system template for humanoids, with a 2B VLM feeding a diffusion transformer that runs at 120~Hz, and the Gemini Robotics series~\citep{gemini2025robotics15} pairs a frontier VLA with a separate embodied-reasoning model that plans in natural language and orchestrates the VLA.

On the third front, supervision has moved beyond imitation. $\pi^{*}_{0.6}$~\citep{physicalintelligence2025b} learns from its own deployment experience through reinforcement learning, while $\pi_{0.7}$~\citep{intelligence2026pi} conditions on subgoal images generated by a world model and shows early signs of compositional generalization to tasks that have no demonstrations in its training data. InternVLA-A1~\citep{cai2026internvla} unifies understanding, future-frame generation and action in one mixture-of-transformers, and its successor InternVLA-A1.5~\citep{wmMa2026A15} replaces explicit future-frame generation with a small set of learnable foresight tokens supervised by a frozen pre-trained video generator, so that the policy inherits the generator's dynamics priors while the video branch is discarded at inference. X-VLA~\citep{zheng2026x} shows that a 0.9B transformer with embodiment-specific soft prompts can match far larger models. Overall, the field has settled on the template and the two structures described above, and the main axis of scaling has moved from parameters, with open-weight VLAs still at 1 to 7B, to data, which now amounts to tens of thousands of hours supplemented by human video, synthetic trajectories and autonomous experience.

\subsubsection{Extending Multimodal Inputs}
\label{sec:multimodal-inputs}

Images and language are the default inputs of a VLA because they are the modalities used for VLM pretraining, but they leave out much of what manipulation depends on, such as events preceding the current frame, how far away objects are, and what happens at the moment of contact. An additional input can enter the sequence of Section~\ref{sec:vla-paradigm} in one of two places. It can be added to the prefix as \emph{extra tokens} that the backbone attends to together with images and text, or it can be given to the suffix and the action module as a \emph{conditioning input} that bypasses the backbone. Writing $m_{t}^{\mathrm{multi}}$ and $m_{t}^{\mathrm{multi}\prime}$ for inputs on the two routes, the canonical sequence becomes
\begin{equation}
\begin{aligned}
  &\underbrace{[\,\mathrm{img}(o_t^{1}),\ldots,\mathrm{img}(o_{t}^{N_{\mathrm{view}}}),\ \mathrm{enc}(m_{t}^{\mathrm{multi}}),\ \mathrm{txt}(\ell_{\mathrm{lang}})\,]}_{\text{prefix}}\\
  &\quad \underbrace{[\,\mathrm{txt}(r_{t}^{\mathrm{reason}})\,]}_{\text{reasoning}}\;
  \underbrace{[\,\mathrm{emb}(\mathbf{q}_t),\ \mathrm{emb}(m_{t}^{\mathrm{multi}\prime}),\ \mathrm{emb}(\mathbf{A}_{t}^{\mathrm{seq},\tau_{\mathrm{flow}}})\,]}_{\text{suffix}}.
\end{aligned}
\end{equation}
Multiple camera views, including the wrist camera, are already handled by the vision encoder as described in Section~\ref{sec:vla-paradigm}, and proprioceptive state $\mathbf{q}_t$ usually takes the second route, so the rest of this section reviews three further extensions, namely temporal history, 3D geometry and contact.

\paragraph{Temporal History and Memory}
Most VLAs condition only on the current frame, even though many tasks depend on the past, for example when the arm hides the target, when an object has been placed under a container, or when the robot must keep track of which stages of a task are already done. Concatenating past frames into the prefix, as Octo~\citep{octo2024} does, brings sharply diminishing returns beyond one extra frame while adding hundreds of tokens per frame and view, and it also invites causal confusion~\citep{de2019causal}, in which the policy copies the behavior implied by past observations instead of reacting to the current one. Longer memory therefore comes down to compression, and two recent systems approach it in different ways. MEM~\citep{torne2026mem}, which underlies $\pi_{0.6}$~\citep{physicalintelligence2025b} and $\pi_{0.7}$~\citep{intelligence2026pi}, keeps memory inside the sequence at two time scales. Over the short term, a video encoder compresses recent frames, packing six of them into the token budget of a single frame in $\pi_{0.7}$. Over the long term, a high-level policy rewrites a natural-language summary at every step and places it in the reasoning slot $r_{t}^{\mathrm{reason}}$, which supports tasks that require up to 15 minutes of memory. RoboTTT~\citep{jiang2026robottt} instead moves memory out of the sequence and into the weights. Its test-time training layers update a small fast-weight network with one gradient step per timestep, so history is stored in these weights and inference cost stays constant however long the context becomes. With this design the context reaches 8K timesteps, or about 5 minutes, and pre-training with an 8K-timestep context outperforms pre-training with 1K timesteps, which suggests that context length may be a scaling axis for robot policies just as it is for language models.

\paragraph{3D Information}
An RGB image projects the scene onto a plane, so grasp height, clearance and relative depth must be inferred, whereas depth sensors and point clouds supply this geometry directly. The most direct way to use this geometry is to feed it into the model. SpatialVLA~\citep{qu2025spatialvla} turns depth from a monocular estimator into 3D positional encodings for the image tokens, while PointVLA~\citep{li2026pointvla} converts point clouds into native 3D tokens that only the action module consumes. Explicit geometry improves precision on spatially demanding tasks, but the 3D encoder lacks web-scale pre-training, and commodity depth sensors fail on transparent and reflective objects. Spatial Forcing~\citep{li2026spatial} takes a second route and uses geometry only as a training signal. Having found that the visual tokens of an RGB-trained VLA carry little recoverable depth structure, it aligns an intermediate layer of these tokens with the features that the geometry foundation model VGGT~\citep{wang2025vggt} extracts from the same images. Since VGGT is needed only during training, inference is unchanged, and training nevertheless becomes faster and more data-efficient, which shows that an input can shape the backbone's representation without ever entering its sequence.

\paragraph{Force and Tactile Sensing}
Contact-rich tasks such as insertion, screwing and handling fragile objects depend on forces that vision cannot observe, and the wrist camera is usually occluded at exactly the moment of contact. Contact information can enter a VLA in three ways. Images from vision-based tactile sensors such as GelSight~\citep{yuan2017gelsight} can be fed to the vision encoder as an additional view, albeit across a domain gap. Force-torque readings~\citep{cao2021six} can be treated like proprioception and placed in the suffix. And dedicated tactile encoders that cover several sensor types, such as Sparsh~\citep{higuera2024sparsh}, can produce tokens for the backbone. Tactile-VLA~\citep{huang2025tactile} follows the token route, fusing tactile tokens with vision and language inside the VLM and outputting target positions and forces for a hybrid position-force controller, whereas ForceVLA~\citep{yu2026forcevla} follows the conditioning route, fusing force-torque tokens with VLM features in a mixture-of-experts module on the action path. TA-VLA~\citep{zhang2025ta}, a systematic study of torque inputs, comes out in favor of the conditioning route for such signals, finding that torque helps more in the action decoder than in the encoder and that a torque history summarized into a single token works best.

\subsection{Pre-Training, Post-Training, and Capability Shaping}
\label{sec:manipulation-training}
\begin{figure}[t]
    \centering
    \includegraphics[width=\linewidth]{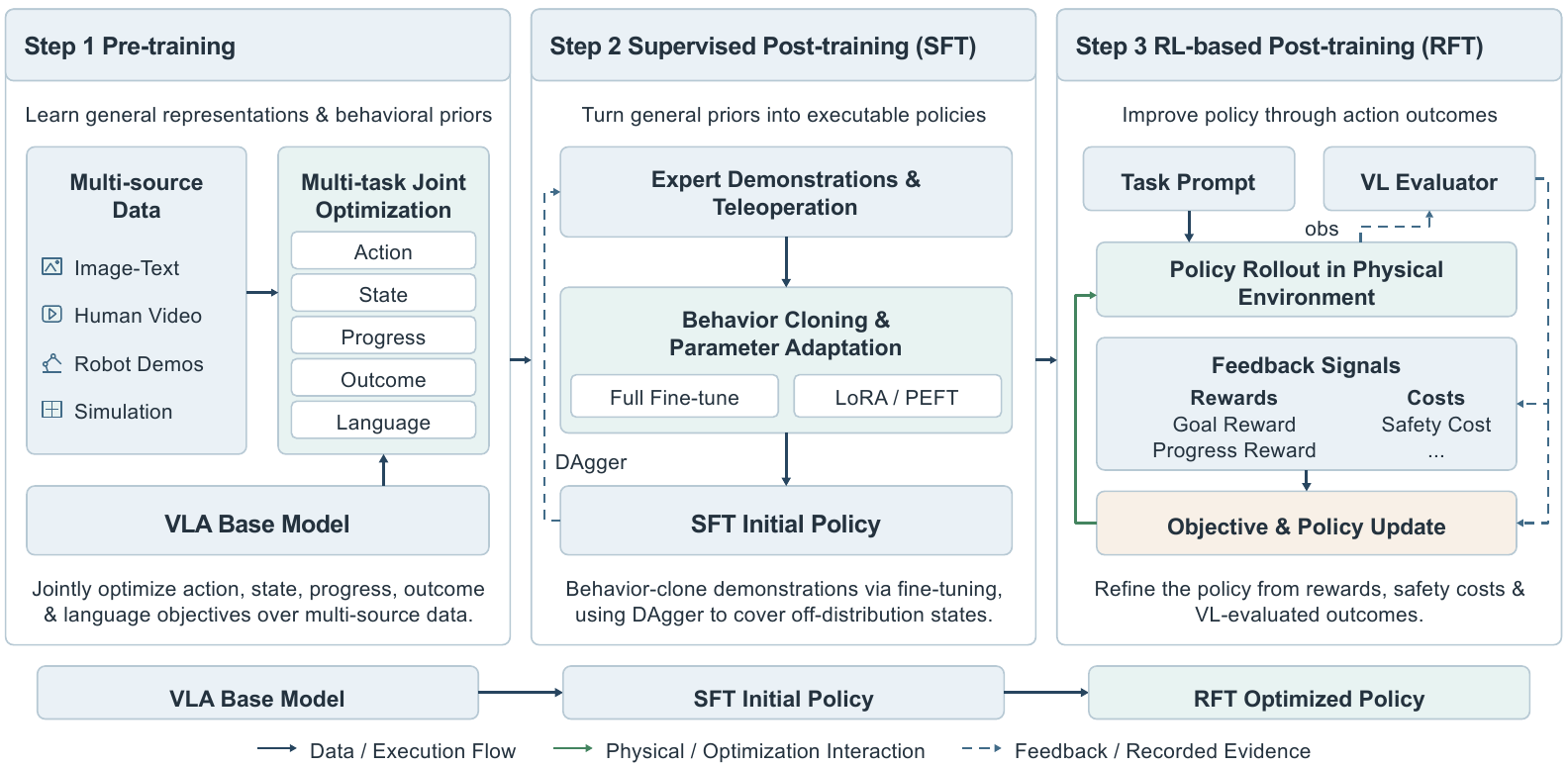}
    \caption{Overview of the embodied VLA training pipeline. Pre-training combines multi-source data under multi-task objectives to produce a VLA base model. Supervised fine-tuning (SFT) during supervised post-training behavior-clones demonstrations into an executable initial policy. And reinforcement fine-tuning (RFT) during reinforcement learning-based post-training refines the policy from rewards, safety costs, and vision-language evaluator feedback. The stages are discussed in Sections~\ref{sec:brain-pretraining} to \ref{sec:brain-rl-posttraining}}
    \label{fig:c6:training-pipeline}
\end{figure}

Figure~\ref{fig:c6:training-pipeline} summarizes the overall training pipeline for embodied VLA models, which proceeds through three stages. Pre-training uses multi-source data to learn general representations and behavioral priors. Supervised post-training, including SFT, turns these priors into executable policies. Reinforcement learning-based post-training, including RFT, improves the policy through action outcomes. The remainder of this section discusses each stage in turn and then addresses continual learning, which converts deployment experience into lasting capabilities (Figure~\ref{fig:c6:continual-learning}).

\subsubsection{Pre-Training: Learning General Embodied Representations and Behavioral Priors}
\label{sec:brain-pretraining}

Vision-language models can recognize objects, interpret instructions, and reason using commonsense knowledge, but these capabilities do not automatically translate into reliable robotic manipulation. Embodied pretraining must integrate language, visual inputs, robot proprioceptive states, actions, and task outcomes within a common temporal framework, enabling the model to learn relationships among instructions, states, actions, and outcomes. Conditioned on observation histories, previous actions, task instructions, and embodiment characteristics, the model learns how to select actions in the current state. Observations may include not only RGB images but also depth, tactile measurements, force and torque readings, joint states, and equipment telemetry.

Embodiment characteristics include degrees of freedom, end-effector type, control frequency, workspace, and action limits. Explicitly incorporating these conditions helps distinguish shared task structure from embodiment-specific execution mechanisms, providing a basis for transfer across tasks and robot platforms. 

Different data sources offer complementary supervision. Internet image-text data provide semantic knowledge about objects, scenes, and tasks. Videos of human activity reveal action sequences, tool use, and state transitions. Robot demonstrations supply action labels associated with specific control interfaces. And simulated interactions provide relatively inexpensive data on contact, forces, collisions, and failures that are difficult to collect at scale. Mixing these sources is intended to combine semantic, temporal, and physical information, rather than merely increase the number of training examples.

This complementarity also offers a way to address the scarcity of labeled robot actions. Video prediction, inverse dynamics, and latent action models can extract task-relevant behavioral representations from human videos, which an embodiment adaptation module can subsequently decode into executable actions. Studies such as UniVLA~\citep{bu2025univla} show that task-centered latent action representations can exploit video data spanning different embodiments and viewpoints. Nevertheless, their physical executability still requires calibration through robot interaction. Acquiring behavioral knowledge from human video therefore also requires aligning that knowledge with robot control interfaces, sensor feedback, and physical outcomes. 

To support this alignment, embodied pretraining commonly combines action prediction, language understanding, state recognition, task progress estimation, and outcome prediction. These objectives correspond to different forms of supervision available in embodied data. Consider a bottle-opening task as an example. Action supervision specifies the robot actions needed to approach, grasp, and rotate the cap, and is typically obtained from teleoperated or scripted demonstrations recorded in the robot's own action space. State supervision describes the relevant object and contact conditions, such as the poses of the bottle and cap and whether the end effector is in contact with the cap. In simulation these labels come from ground-truth physics, while on real robots they are estimated by the perception stack or annotated offline. Progress supervision identifies the current stage of the task, such as approach, grasp, rotation, release, or completion. This supervision can be supplied by scripted stage detectors, human annotators, or a vision-language evaluator of the kind used in reinforcement learning-based post-training. Outcome supervision indicates whether the bottle has ultimately been opened, which can be determined from a terminal physical observation or task-specific success criterion. Language supervision associates the visual and behavioral sequence with the task instruction, such as ``open the bottle'', and is available whenever trajectories are paired with instructions or narrations. These five objectives train distinct, complementary capabilities.

\begin{table}[!b]
\centering
\caption{Correspondence between heterogeneous data sources, the supervision they provide, the objectives of Eq.~\eqref{eq:pretraining_multitask_loss} they support, and the model components primarily updated. Objectives whose labels a source does not provide are masked rather than assigned artificial targets. A dash means that missing labels are not specified}
\label{tab:c6:data-supervision}
\footnotesize
\begin{tabular}{@{}>{\raggedright\arraybackslash}p{\dimexpr.17\linewidth-1.6\tabcolsep\relax}>{\raggedright\arraybackslash}p{\dimexpr.24\linewidth-1.6\tabcolsep\relax}>{\raggedright\arraybackslash}p{\dimexpr.20\linewidth-1.6\tabcolsep\relax}>{\raggedright\arraybackslash}p{\dimexpr.21\linewidth-1.6\tabcolsep\relax}>{\raggedright\arraybackslash}p{\dimexpr.18\linewidth-1.6\tabcolsep\relax}@{}}
\toprule
Data source & Available supervision & Supported objectives & Primary module updated & Missing labels \\
\midrule
Internet image-text & Object/scene semantics, instruction pairing & $\mathcal{L}_{\mathrm{language}}$ ($\mathcal{L}_{\mathrm{state}}$ from captions) & Vision-language backbone & Action, progress masked \\
Human video & Temporal order, task stages, narration & $\mathcal{L}_{\mathrm{progress}}$, $\mathcal{L}_{\mathrm{language}}$, latent-action $\mathcal{L}_{\mathrm{action}}$ & Backbone, latent-action or embodiment adapter & Executable action masked \\
Robot demonstrations, teleoperation & Executable actions, proprioception\newline Stages and outcomes when annotated & $\mathcal{L}_{\mathrm{action}}$, $\mathcal{L}_{\mathrm{state}}$ ($\mathcal{L}_{\mathrm{progress}}$, $\mathcal{L}_{\mathrm{outcome}}$) & Action expert (and adapters) & Unannotated items masked \\
Simulation & Dense state, contact, action, outcome & All five & Backbone and action expert & --- \\
Deployment logs (interventions, failures) & Outcomes, corrections, takeovers & $\mathcal{L}_{\mathrm{outcome}}$\newline Value/critic targets at later stages & Value head, policy via post-training & Per record type \\
\bottomrule
\end{tabular}
\end{table}

These objectives can be summarized by a weighted multi-task loss of the form
\begin{equation}
\mathcal{L}_{\mathrm{pre}} =
\lambda_{\mathrm{a}}\mathcal{L}_{\mathrm{action}} +
\lambda_{\mathrm{s}}\mathcal{L}_{\mathrm{state}} +
\lambda_{\mathrm{p}}\mathcal{L}_{\mathrm{progress}} +
\lambda_{\mathrm{o}}\mathcal{L}_{\mathrm{outcome}} +
\lambda_{\ell_{\mathrm{lang}}}\mathcal{L}_{\mathrm{language}},
\label{eq:pretraining_multitask_loss}
\end{equation}
where the coefficients balance action learning, state understanding, progress estimation, outcome prediction, and language supervision. This schematic loss combines objectives whose labels may be available on different samples. Each loss is evaluated only where its supervision is available, with missing labels handled by masking the corresponding objective.

Table~\ref{tab:c6:data-supervision} summarizes how this masking scheme applies to each data source, which objectives in Eq.~\eqref{eq:pretraining_multitask_loss} it can support, and which model components the resulting gradients primarily update. Because available supervision varies by data source, the same foundation model is trained by applying different subsets of the objectives to different data sources. The resulting training process aligns semantic, temporal, state, and action representations without requiring a homogeneous label space across the entire dataset.

How such multi-task, multi-embodiment data are curated, balanced, and mixture-weighted is a data engineering question treated in Chapter~\ref{ch:physical-data}. This section examines how the composition of the mixture enters optimization. On the training side, a curriculum may progress from short-horizon skills with limited contact to multistage tasks, perturbed states, and failure recovery, and it should specify success and failure boundaries so that the model learns when to continue execution, acquire new observations, or terminate safely. As action learning progresses, retaining existing multimodal knowledge becomes an additional concern. When an action expert is connected to a pretrained vision-language backbone, gradient updates from action learning may interfere with established semantic representations. Staged unfreezing, parameter-efficient fine-tuning, gradient isolation, and knowledge distillation can help preserve language understanding and visual knowledge while new action capabilities are acquired. The heterogeneous joint training used in $\pi_{0.5}$ and the Knowledge Insulation approach illustrate this principle. Training across diverse data sources expands task and environmental coverage while limiting interference with existing multimodal knowledge.

Pretraining should consequently be evaluated using more than training loss or offline action prediction error. Evaluation should also examine closed-loop performance on unseen objects, novel combinations of instructions, and mildly perturbed conditions. These tests assess whether shared representations support perception, decision-making, and action under the new conditions. Pretraining supplies an initialization for the task- and embodiment-specific adaptation discussed next. {Where the prior does not yet support the target task or embodiment, supervised post-training is one route to consistent execution. Outcome-based adaptation provides another, as developed below. Some pipelines also introduce an explicit alignment stage between broad pretraining and task adaptation. Section~\ref{sec:c6:dex-human} examines these choices for dexterous hands through EgoScale on Sharpa Wave.}

\subsubsection{Supervised Post-Training: From General Priors to Executable Policies}

Supervised post-training uses expert demonstrations, teleoperation trajectories, task procedures, and high-quality simulated data to adapt general behavioral priors to specific tasks, embodiments, and environments. Behavioral cloning increases the likelihood of expert actions conditioned on observations, task instructions, and embodiment characteristics, bringing the policy closer to the behavioral distribution represented in the demonstrations. Its effectiveness depends first on whether demonstrations cover the complete task. Training examples should include task initiation, critical contacts, transitions between stages, completion verification, and safe termination, rather than being restricted to locally standardized actions extracted from successful trajectories. For multistep tasks, stage labels, subgoals, and outcome criteria should also be recorded to support learning of temporal relationships among actions.

Adaptation to a new task or embodiment may then use full-parameter fine-tuning, updates to selected modules, or parameter-efficient methods such as low-rank adaptation (LoRA). Full-parameter fine-tuning offers greater scope for parameter adjustment but requires more data and computation. Parameter-efficient methods facilitate rapid iteration, although their effectiveness may be limited by discrepancies between the new task and existing representations. For a VLA initialized from a pretrained vision-language model, parameter-efficient adaptation has a further benefit. The backbone can be kept frozen so that the semantic priors built during pre-training are preserved, while a small number of action-side parameters are learned. Gradient isolation during pre-training addresses the same interference concern. Results reported for OpenVLA and Octo~\citep{kim2024openvla,octo2024} indicate that large-scale multitask pretraining can substantially reduce the data required for downstream adaptation. Performance after adaptation nevertheless depends on appropriate treatment of sensors, action spaces, and embodiment characteristics.

Even with high-quality demonstrations, behavioral cloning faces a discrepancy between training and deployment state distributions. Training typically uses states visited by an expert, whereas deployment states are determined by the model's own actions. A small localization error can cause subsequent observations to depart from the demonstration distribution, allowing errors to compound over time. Interactive dataset aggregation addresses this problem by executing the current policy, requesting expert actions or corrections at the off-distribution states it encounters, and adding these examples to the training set. DAgger exemplifies this approach to distribution shift in sequential decision-making. For a fine-tuned VLA, the expert is typically a teleoperator who shares the policy's action interface, so corrective labels can be appended to the training set without retargeting, and interactive aggregation becomes a natural extension of demonstration collection rather than a separate pipeline. Demonstration collection thus extends beyond recording ideal execution to include difficult states and recovery behaviors encountered by the model itself.

Human intervention data should be differentiated according to their training function. Valid actions preceding a takeover can serve as positive demonstrations. Human corrections can provide supervised labels for off-distribution states. And failure outcomes can support the training of success classifiers or value models. Erroneous actions, however, should not automatically become imitation targets. This treatment preserves both the conditions under which errors arise and the corresponding recovery strategies, enabling supervised post-training to learn normal execution as well as recovery to feasible states. Nevertheless, supervised post-training primarily learns conditional relationships between states and actions in demonstration data. It does not directly optimize completion time, energy consumption, precision, or long-term success rate. It is well suited to establishing a stable initial policy, particularly when operating procedures are clearly defined and expert demonstrations are of high quality, but reductions in training loss cannot substitute for closed-loop validation on the physical system.

Validation should therefore cover instruction following, action executability, critical state transitions, outcome assessment, and termination under abnormal conditions. For tasks such as liquid transfer and precision assembly, success criteria should also incorporate mass, flow rate, applied forces, leakage, and instrument readings, explicitly distinguishing the execution of action commands from the achievement of task objectives. When a model can reliably follow procedures but remains limited in efficiency, contact precision, or recovery from complex disturbances, feedback from action outcomes is needed to support policy improvement beyond imitation.

\subsubsection{Reinforcement Learning-Based Post-Training: Improving Policies Through Action Outcomes}
\label{sec:brain-rl-posttraining}

Behavioral cloning can reproduce demonstrator behavior, but its objective does not directly address efficiency gains subject to success-rate requirements, action adjustment under changing contact conditions, recovery from states outside the demonstration distribution, or improvements in action sequences discovered through autonomous interaction. Reinforcement learning evaluates the long-term consequences of actions through environmental feedback, enabling continued policy improvement toward task objectives. For Physical AI, it is particularly relevant to precise contact interactions, long-horizon tasks, disturbance recovery, and efficiency optimization. Reinforcement learning is not, however, necessary for every task. When procedures are stable, demonstrations provide sufficient coverage, and performance already meets requirements, supervised training and deterministic control may be more appropriate. The decision to introduce reinforcement learning-based post-training should depend on the limitations of the existing policy and whether environmental feedback provides an effective signal for addressing them.

Compared with virtual environments, reinforcement learning on physical robots involves slow interactions, costly resets, equipment wear, delayed rewards, and safety risks. States are often partially observable, contact and friction exhibit discontinuities, and outcomes may also depend on material batches, temperature, and tool condition. Large-scale random exploration in the physical world is therefore generally impractical. Under these constraints, policy optimization typically seeks to maximize expected cumulative return while accounting for task rewards, safety costs, and energy consumption, with discounting controlling the relative weight of returns over different time horizons. For physical robot tasks, this objective can be written as
\begin{equation}
J(\pi_{\theta}) = \mathbb{E}_{\tau_{\mathrm{ep}}\sim\pi_{\theta}}\left[\sum_{t=0}^{T}\gamma^t\left(\lambda_{\mathrm{g}} r_t^{\mathrm{goal}} + \lambda_{\mathrm{p}} r_t^{\mathrm{progress}} - \lambda_{\mathrm{safe}} c_t^{\mathrm{safety}} - \lambda_{\mathrm{e}} c_t^{\mathrm{energy}}\right)\right],
\label{eq:embodied_rl_objective}
\end{equation}
where the terms represent task achievement, progress, safety cost, and energy consumption, respectively. Task rewards, progress, safety constraints, and resource consumption should be recorded separately to determine whether apparent gains reflect improved task performance or exploitation of a particular metric. Safety penalties can help a policy learn risk boundaries, but they cannot replace independent limit enforcement, collision detection, emergency stops, or human takeover mechanisms. Reward design must likewise guide learning while remaining verifiable against physical outcomes. Terminal success signals are simple and reliable but too sparse for many long-horizon tasks. Intermediate rewards can capture distance to the goal, contact quality, trajectory error, completion time, and stage progress, yet may be exploited in ways that diverge from the actual objective. Reward design should therefore prioritize observable, independently verifiable physical quantities, such as changes in mass, position error, peak force, leakage status, and instrument readings, and validate them against independently assessed final outcomes.

Alongside rewards, value functions estimate how current states and actions contribute to future task outcomes, helping the policy identify behaviors with long-term benefits. Vision-language evaluators can infer progress, completion, and stagnation from successive observations, providing denser feedback for physical robots. Such evaluators may nevertheless misclassify outcomes, and policies may exploit evaluation biases through reward hacking. Calibration against physical outcomes is therefore necessary. VLAC~\citep{zhai2025vlac} illustrates an approach that uses vision-language models for progress estimation and asynchronous reinforcement learning on real robots. These methods broaden the sources of feedback while reinforcing the need to cross-check intermediate evaluations against final outcomes.

For a fine-tuned VLA, a practical pipeline begins offline. The demonstrations, successful and failed trajectories, and human corrections already accumulated during supervised post-training form the dataset from which values are estimated and a policy is derived, reducing the risks of physical exploration. A subsequent online stage can collect a limited amount of autonomous execution data, focusing on states associated with model uncertainty, repeated failures, or high informational value. Training should constrain the policy from deviating excessively from behaviors supported by the data, thereby limiting overly optimistic value estimates for actions absent from the dataset. Human intervention can also be adjusted as the policy improves. Takeovers can prevent hazardous exploration early in training, corrective actions can address difficult states at intermediate stages, and later interventions can be confined to infrequent anomalies. HIL-SERL~\citep{luo2025hilserl} combines demonstrations, reward classifiers, and progressively reduced human intervention to improve the sample efficiency of reinforcement learning on physical robots, illustrating the complementary roles of demonstrations, autonomous interaction, and human correction.

For large VLA models, conventional online policy-gradient methods such as PPO~\citep{schulman2017ppo} often entail substantial sample requirements and challenges in training stability. An alternative is to collect experience in batches and update the policy between collection rounds. The RECAP method used in $\pi^{*}_{0.6}$~\citep{physicalintelligence2025b} integrates autonomous execution outcomes, human corrections, and existing demonstrations within iterative offline reinforcement learning. It estimates advantages using a value function and then derives an improved policy through advantage conditioning. This approach illustrates how reinforcement learning on physical robots can alternate between batch data collection, offline policy updates, and validation on the robot, without modifying model parameters after every action. The same workflow supports independent validation between updates to determine whether policy improvements persist during physical execution.

To further reduce physical interaction costs, world models can compare the short-term consequences of candidate actions before execution, helping to screen out behaviors likely to cause collisions, slippage, leakage, or task failure. They can also generate supplementary experience in simulation or latent space. However, world models are subject to modeling errors, which a policy may exploit to obtain inflated estimates of return. They are therefore useful for candidate evaluation, risk prediction, and data augmentation, but cannot replace physical outcome measurements, low-level controllers, or independent safety mechanisms. Regardless of the training approach, evaluation should report success rate, completion time, precision, failure recovery rate, the number of physical interactions, human intervention rate, and the number of safety incidents. Improvements beyond demonstration performance can be established only when gains persist on independent test tasks and under physical disturbances. Retaining and extending those gains during long-term deployment additionally requires continual learning.

\subsubsection{Continual Learning: Turning Deployment Experience into Lasting Capabilities}
\label{sec:brain-continual-learning}

\begin{figure}[t]
    \centering
    \includegraphics[width=\linewidth]{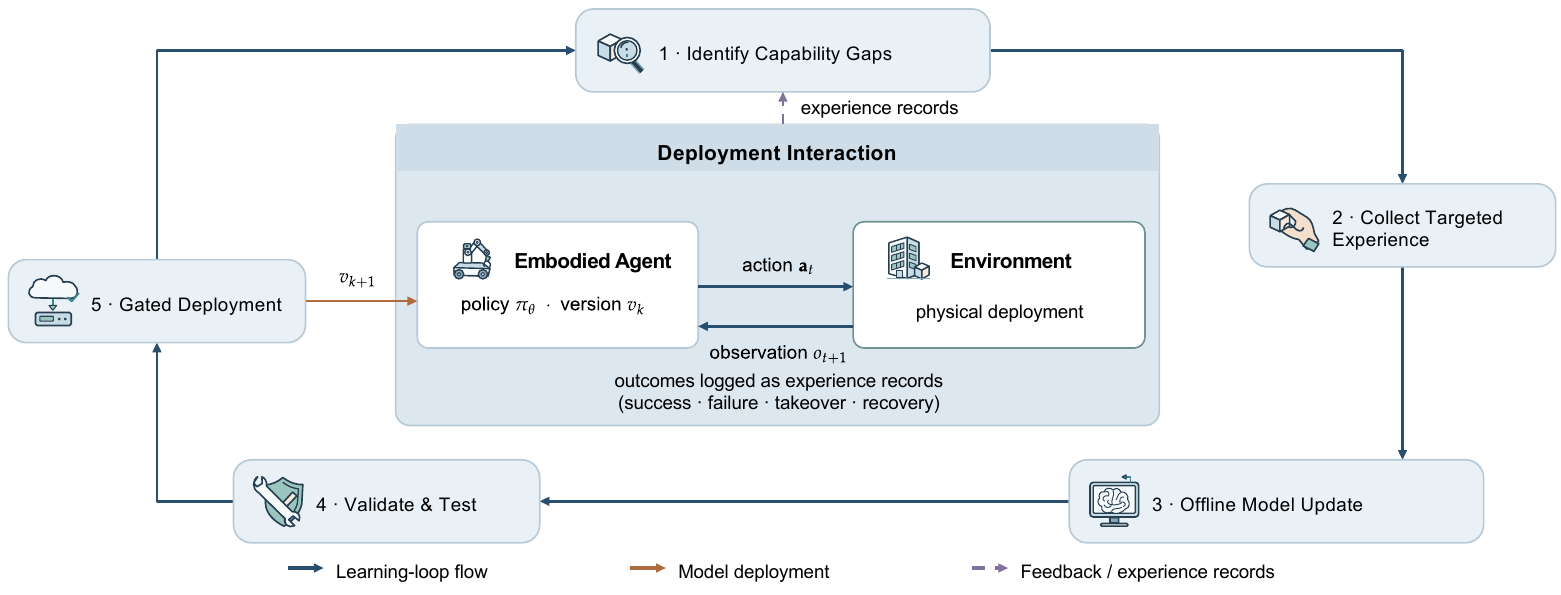}
    \caption{The continual learning loop for embodied agents in long-term deployment. Deployment interactions are logged as experience records (success, failure, takeover, recovery), and each iteration updates the policy version $v_k$ to $v_{k+1}$ through gap identification, targeted collection, offline update, validation, and gated deployment. The process is detailed in Section~\ref{sec:brain-continual-learning}}
    \label{fig:c6:continual-learning}
\end{figure}

Long-term deployment exposes embodied agents to situations that cannot be fully anticipated during pretraining or post-training. The resulting need for continued adaptation can be understood in light of Sutton's Big World Hypothesis. As \citet{JavedSutton2024} articulate, many decision-making environments are far larger and more complex than an agent's perceptual, representational, and computational resources can accommodate. An agent must therefore act on an incomplete understanding of its environment, learning and revising the knowledge most relevant to its current decisions rather than relying on an exhaustive representation of all possible states and optimal actions. For embodied systems, this perspective provides a rationale for continual learning. Capabilities acquired before deployment provide an initial basis for action, but must be evaluated and refined as interaction reveals unfamiliar states, changing task demands, and limitations of the current policy. Such refinement must also preserve previously acquired skills that remain relevant. Supporting this process requires experience records that retain the temporal, spatial, and embodiment context of each interaction, including the model version, task instruction, sensor state, commanded action, observed outcome, and any anomaly or human intervention. Distinguishing successful execution, failure, warnings, takeovers, and recovery within these records allows deployment experience to inform subsequent supervised training, value estimation, and policy improvement.

As experience accumulates, the central challenge in continual learning shifts from increasing data volume to identifying the learning value of individual experiences. Samples can be selected according to failure frequency, model confidence, state novelty, intervention points, and environmental changes, while balancing task value, information gain, collection cost, and risk. Recurrent grasp slippage, for example, calls for additional data on contact states and recovery actions. Occasional instrument anomalies require complete telemetry and procedural context, rather than a final failure label alone. Active sampling should also help identify capability boundaries, allowing the system to distinguish tasks already mastered under current conditions from changes addressable through context updates and changes requiring parameter updates or retraining. This distinction links newly collected data to specific capability gaps and prevents local adaptation from being conflated with improvements in general competence.

Clarifying the form of adaptation is also essential to managing the relationship between new and existing capabilities. Immediate adaptation through context or memory, and runtime calibration through updates to state estimates or embodiment parameters, do not necessarily modify the foundation model. Long-term learning that changes model parameters, by contrast, requires model version control and regression testing. As new tasks arrive and parameters are updated, catastrophic forgetting, interference across tasks, and policy drift may occur. Continual learning therefore needs to optimize new-task adaptation while retaining previously acquired capabilities~\citep{kirkpatrick2017ewc,dohare2024plasticity}, using an objective of the form
\begin{equation}
\mathcal{L}_{\mathrm{CL}} = \mathcal{L}_{\mathrm{new}} + \lambda_{\mathrm{retain}}\mathcal{L}_{\mathrm{retain}} + \lambda_{\mathrm{reg}}\mathcal{L}_{\mathrm{reg}},
\label{eq:continual_learning_objective}
\end{equation}
where $\mathcal{L}_{\mathrm{new}}$ measures adaptation to new tasks, $\mathcal{L}_{\mathrm{retain}}$ preserves previous capabilities, and $\mathcal{L}_{\mathrm{reg}}$ limits undesirable parameter changes. Replay of representative experiences can preserve coverage of previous task distributions. Knowledge distillation can constrain the new model's behavior on earlier tasks. Parameter regularization can limit changes to parameters important for retained knowledge. And modular architectures can isolate new skills within specific components. Methods such as elastic weight consolidation~\citep{kirkpatrick2017ewc} mitigate forgetting by restricting changes to important parameters, although their effectiveness in embodied tasks must still be validated through physical interaction. Continual learning should therefore be evaluated in terms of both gains on new tasks and retention of existing skills.

As illustrated in Figure~\ref{fig:c6:continual-learning}, continual learning should follow a controlled iterative process to make capability gains verifiable. The process should identify capability gaps, collect targeted experience, update the model offline, jointly validate performance on new and existing tasks together with compliance with safety constraints, and deploy only after predefined performance and safety requirements are met. Previous model versions should remain available for rollback. For high-risk tasks, validation should begin in simulation, hardware-in-the-loop settings, or restricted workspaces before the scope of physical operation is gradually expanded. Each update should establish whether performance on new tasks has improved, whether previous capabilities have been retained, how much data and human effort adaptation required, and whether failure recovery has become more reliable. Deployment experience can be regarded as a source of established system capabilities only when these results are reproduced across independent tasks, different batches of objects, and sustained operation. It can then provide a reliable basis for subsequent task adaptation and policy optimization.

As summarized in Figure~\ref{fig:c6:training-pipeline}, capability development in foundation models for robot manipulation thus proceeds through a feedback loop of prior acquisition, task adaptation, outcome-based optimization, and experience accumulation, rather than a direct transition from large-scale pretraining to autonomous operation. Pretraining supports shared semantic and behavioral structure. Supervised post-training enables execution of procedures on specific embodiments. Reinforcement learning-based post-training uses physical outcomes to improve precision, efficiency, and recovery. And continual learning supports the long-term retention of these gains and their transfer to new tasks and environments. Each stage has a distinct training emphasis, while demonstrations, interaction feedback, and deployment experience connect them. Together, they support the progression from demonstration-driven behavioral learning toward continual policy optimization grounded in physical feedback.

\subsection{Multi-Embodiment Modeling and Model Adaptation}
\label{sec:brain-embodiment-adaptation}

The action interfaces in Section~\ref{sec:brain-vla} specify how a model generates behavior, while the pre-training and post-training procedures in Section~\ref{sec:manipulation-training} establish reusable behavioral priors. Realizing these priors on a different physical system requires compatible observation and action interfaces, together with adaptation to its morphology and interaction dynamics. Here, an \emph{embodiment} denotes the robot platform together with its end-effectors, sensing and actuation systems, and control interfaces. Replacing a parallel gripper with a dexterous hand therefore changes the embodiment even if the manipulator remains unchanged. The substitution alters the available contact modes, action dimensionality, sensing geometry, and feasible manipulation strategies. Multi-embodiment modeling seeks to transfer task-relevant knowledge across these configurations while preserving the physical distinctions that determine executable behavior. Cross-robot training in Open X-Embodiment provides empirical evidence that heterogeneous interaction data can support such transfer, although its reported evaluations do not establish generalization to arbitrary unseen robots~\citep{openx2024}.

Following the partially observable formulation in Chapter~\ref{sec:technical-foundations} and the distinction between physical state and internal representation in Chapter~\ref{ch:wm}, let $e\in\mathcal{E}_{\mathrm{emb}}$ index an embodiment and write
\begin{equation}
  \mathcal{M}_e=
  (\mathcal{S}_e,\mathcal{A}_e,P_{e}^{\mathrm{tr}},R_{\ell_{\mathrm{lang}},e},
   \mathcal{O}_e,O_e,\gamma),
  \label{eq:brain-embodiment-process}
\end{equation}
where $\mathcal{S}_e$ includes the physical state of the robot and its environment. $\mathcal{A}_e$ is the action space at the declared control interface. $P_{e}^{\mathrm{tr}}$ is the transition kernel. $R_{\ell_{\mathrm{lang}},e}$ specifies the reward for instruction $\ell_{\mathrm{lang}}$. $\mathcal{O}_e$ is the measurement space. $O_e$ is the observation likelihood, and $\gamma$ is the discount factor. The measurements comprise exteroceptive observations $o_t$ and proprioceptive measurements $\mathbf{q}_t$, retaining their separate notation from Section~\ref{sec:brain-taxonomy}. The kernel $P_{e}^{\mathrm{tr}}$ corresponds to the embodiment-conditioned transition kernel $T$ in Chapter~\ref{ch:wm}, with elapsed time included when required. The measurement pair $(o_t,\mathbf{q}_t)$ belongs to $\mathcal{O}_e$, whereas $O_e(o_{t+1},\mathbf{q}_{t+1}\mid s_{t+1},\mathbf{a}_t)$ denotes its conditional observation likelihood. Actuator memory and other dynamic variables must be included in the physical state when needed for the Markov assumption. The subscript on $\mathcal{S}_e$ allows robot configurations with different state dimensions. A shared task representation does not require identical physical state spaces.

Using the history $h_t=(o_{0:t},\mathbf{q}_{0:t},\mathbf{a}_{0:t-1})$ and action chunk $\mathbf{A}_{t}^{\mathrm{seq}}=[\mathbf{a}_t^\top;\ldots;\mathbf{a}_{t+H-1}^\top]$ defined in Eq.~\eqref{eq:brain-history}, the policy interface remains
\begin{equation}
  \mathbf{A}_{t}^{\mathrm{seq}}\sim\pi_{\Theta}(\cdot\mid h_t,\ell_{\mathrm{lang}},e),
  \label{eq:brain-embodiment-policy}
\end{equation}
where $\Theta$ denotes the complete set of policy parameters, while $\theta$ and $\psi$ below denote representation and action-generation parameters, respectively. The horizon $H$ counts action steps. Its physical duration also depends on the command intervals. A current-observation policy is a restricted instance of this history-conditioned formulation. The modeling objective is to share information about task goals, object relations, and behavioral organization while retaining the embodiment information required for action feasibility and contact regulation.

\subsubsection{Embodiment-Aware Data Representation and Alignment}
\label{sec:brain-embodiment-data}

Pooling trajectories across robots requires compatibility of physical meaning, not merely compatibility of tensor shapes. Let $\mathcal{D}_e$ denote the collection of trajectories associated with embodiment $e$. Each trajectory retains its instruction, ordered observations and actions, outcome annotations, and acquisition metadata. Let $m_{t}^{\mathrm{meta}}$ denote acquisition metadata. Metadata should identify the robot and end-effector configuration, sensor calibration, coordinate frames, units, control mode, timestamps, command duration, latency, and the data-collection policy or operator. These fields determine how a recorded command relates to an observed physical consequence.

\paragraph{Semantic, Geometric, and Dynamic Alignment}
Semantic alignment concerns task goals, object roles, and stages of behavior. Two robots can share the goal of opening a container while requiring different grasps and motion sequences. Geometric alignment concerns reference frames, pose conventions, kinematics, and correspondences between observation and action representations. A command expressed in a camera frame, tool frame, or joint space requires an explicit interpretation before it can be transferred. Dynamic alignment concerns the embodiment-dependent transition kernel $P_{e}^{\mathrm{tr}}$, including actuator response, payload, friction, compliance, and execution delay. Geometric correspondence alone does not establish equivalent contact behavior.

These distinctions also delimit what data standardization achieves. The RT-X experiments use a canonical camera view and a coarsely aligned end-effector action interface, with dataset-specific action normalization. The original control conventions are retained, and action coordinate frames are not fully aligned across datasets~\citep{openx2024}. Thus, numerical normalization facilitates joint training without making identical action vectors physically interchangeable. UMI illustrates a complementary interface-level approach in which relative trajectories and inference-time latency matching support transfer of demonstrations collected with handheld grippers to compatible robot systems~\citep{chi2024umi_ch6}. Neither approach removes the need to verify reachability, tracking, and contact behavior on the receiving embodiment.

\paragraph{Heterogeneous Sensing and Action Interfaces}
Observation interfaces may differ in camera count, viewpoint, and calibration. They may also differ in whether proprioceptive, force, or tactile measurements are available. Modality-specific tokenizers, sensor identity encodings, and validity masks can expose these differences to a shared model. Octo supports adding or removing observation tokenizers and adapting to new sensory inputs. HPT uses embodiment-specific input stems to map heterogeneous visual and proprioceptive inputs into a common token interface~\citep{octo2024,wang2024hpt}. Missing measurements should remain distinguishable from valid zero-valued measurements. For high-rate contact signals, timestamps and observation windows must also preserve the temporal relationship between commands and measured interaction.

Action interfaces require comparable care. A Cartesian displacement, joint target, velocity command, and force target define different interventions even when their numerical dimensions coincide. Padding and normalization can accommodate dimensional heterogeneity, but do not resolve differences in units, control mode, or effective duration. The original $\pi_0$ implementation pads state and action vectors to a common dimensionality and masks unused image slots~\citep{black2024pi0}. Such implementation choices should be specified for the particular model rather than treated as universal conventions of cross-embodiment learning. Likewise, resampling a trajectory does not establish dynamic equivalence between robots operating at different control rates.

\paragraph{Complementary Supervision Sources}
Human video provides evidence about task structure, object transformations, and behavioral stages, but usually lacks robot-executable action labels. Robot demonstrations and teleoperation provide action-grounded supervision under a specified interface. Simulation supports controlled variation of morphology and dynamics, whereas autonomous rollouts expose the state distribution induced by the executing policy. These sources therefore supervise different aspects of transfer, as discussed in Section~\ref{sec:brain-pretraining} and Chapter~\ref{ch:physical-data}. HPT combines data from robot teleoperation, deployed robots, simulation, and human video through heterogeneous input and output interfaces~\citep{wang2024hpt}. For human video without robot action labels, it uses estimated poses and two-dimensional positions as surrogate supervision rather than treating visual motion as a robot command. Failures, interventions, corrections, and recovery segments should remain identifiable because they represent different supervision signals and operating conditions.

Embodiment-aware representation consequently establishes the conditions under which data can be combined or transformed. The desired internal representation $z_t$ should suppress irrelevant variation while preserving information about physical differences that affect the next decision. It is a learned summary of available evidence, following Chapter~\ref{ch:wm}, rather than a physical state or a calibrated belief distribution by definition.

\begin{figure}[!t]
  \centering
  \includegraphics[width=\linewidth]{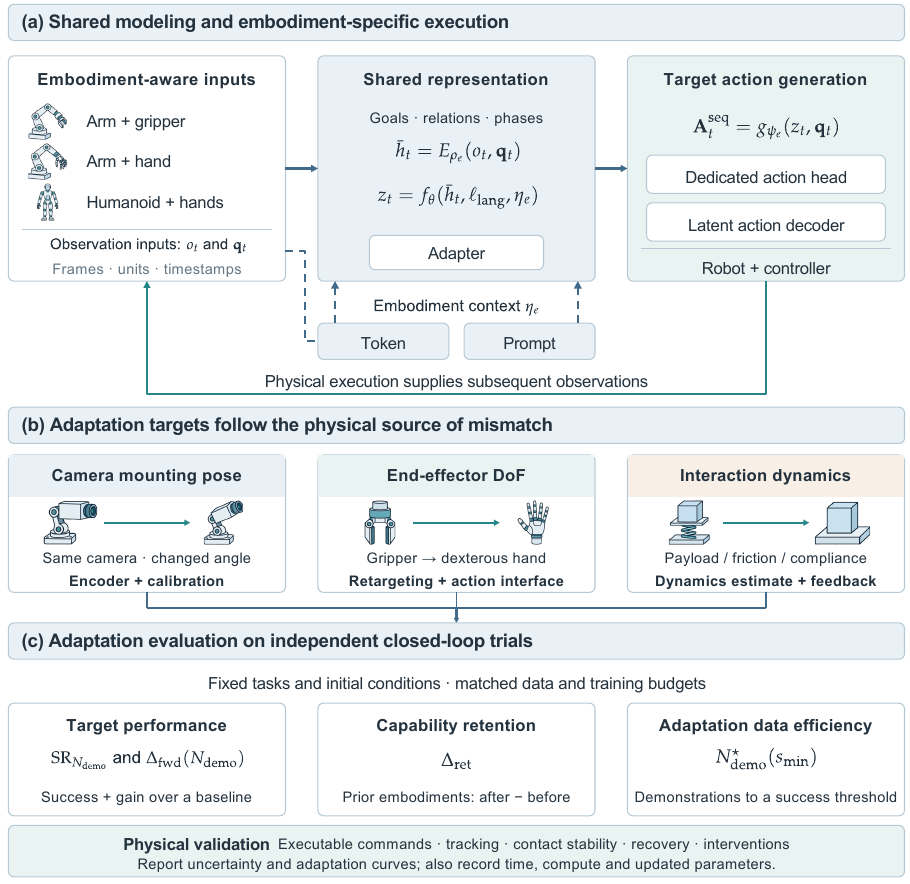}
  \caption{Shared modeling, embodiment-specific adaptation, and adaptation evaluation for manipulation. (a) Observation interfaces, shared task representations, and executable outputs illustrate complementary token, prompt, adapter, action-head, and latent-decoder mechanisms. (b) Changes in camera mounting pose (with camera count unchanged), end-effector degrees of freedom (DoF), and interaction dynamics motivate corresponding adaptation targets. (c) Independent closed-loop trials assess target success and budget-matched transfer gain, retention on previously supported embodiments, adaptation data efficiency, and physical execution. For compactness, panel (a) depicts current-observation inputs and deterministic execution mappings. Eq.~\eqref{eq:brain-embodiment-factorization} gives the general history-conditioned policy formulation}
  \label{fig:chap6-4-Manipulation_Modeling_Adaptation-arch}
\end{figure}

\subsubsection{Shared Representations, Execution Interfaces, and Adaptation}
\label{sec:brain-embodiment-modeling}

The decomposition introduced in Eq.~\eqref{eq:brain-policy-world-value} can be specialized to heterogeneous observation and action spaces. Let $\eta_e$ describe the relevant morphology, sensing, actuation, and control properties of embodiment $e$. A modular policy may be written as
\begin{equation}
  \begin{aligned}
    \bar h_t &= E_{\rho_e}(h_t),\\
    z_t &= f_{\theta}(\bar h_t,\ell_{\mathrm{lang}},\eta_e),\\
    \mathbf{A}_{t}^{\mathrm{seq}} &\sim\pi_{\psi_e}(\cdot\mid z_t,\ell_{\mathrm{lang}},\eta_e),
  \end{aligned}
  \label{eq:brain-embodiment-factorization}
\end{equation}
where $E_{\rho_e}$ is an observation-interface encoder, $f_{\theta}$ is a shared representation model, and $\pi_{\psi_e}$ generates action chunks in $\mathcal{A}_e^H$. This notation retains $\theta$ for representation parameters and $\psi$ for policy parameters, as in Section~\ref{sec:brain-taxonomy}. The subscripts identify the functional interfaces that may depend on embodiment. They do not require a separate network for every robot. Architectures may share some or all of $\rho_e$ and $\psi_e$ and condition their computation on $\eta_e$. A deterministic decoder $\mathbf{A}_{t}^{\mathrm{seq}}=g_{\psi_e}(z_t,\mathbf{q}_t)$, as depicted in Figure~\ref{fig:chap6-4-Manipulation_Modeling_Adaptation-arch}, is a special case. $\mathbf{q}_t$ is already contained in $h_t$ but may also enter the decoder directly.

The descriptor $\eta_e$ may comprise measured calibration quantities, a learned embedding, a structured prompt, or an estimate inferred from interaction history. Explicit identity conditioning and inferred physical parameters serve different purposes. An identity token can index a known configuration, whereas adaptation to unobserved payload or compliance requires evidence about the changed dynamics. Shared features need not be strictly embodiment-invariant. For example, a representation of a feasible grasp should depend on hand geometry even when the object and task are unchanged.

\paragraph{Embodiment Representations and Adaptation Mechanisms}
The descriptor and interface modules in Eq.~\eqref{eq:brain-embodiment-factorization} can be implemented through tokens, prompts, adapters, dedicated action heads, or latent action decoders. These mechanisms differ in where embodiment information enters the model and which components are adapted. They are complementary rather than mutually exclusive. A soft prompt can itself consist of tokens, and a dedicated action head can incorporate a latent decoder.

Embodiment information can be introduced as learned identity embeddings or structured tokens encoding morphology and control properties. Structured tokens expose joint attributes and kinematic relationships to the shared transformer, providing a representation of physical configuration beyond a categorical robot label. Suzuki et al. combine joint-wise kinematic action tokens with topology-aware attention and joint-attribute conditioning~\citep{suzuki2026morphology_ch6}. Their kinematic tokens organize action trajectories, while topology and joint descriptors introduce morphological information. The latter instantiate embodiment context associated with $\eta_e$. The action tokens themselves should be distinguished from an identity descriptor. An identity embedding indexes a known configuration, whereas generalization to an unseen configuration requires a representation of its relevant physical properties.

An embodiment-specific prompt supplies contextual information that modulates shared computation. A structured textual prompt may describe available sensors or the action interface, while a soft prompt represents such context through trainable continuous embeddings. X-VLA introduces separate learnable soft prompts for heterogeneous data sources, together with domain-specific projections for action-related inputs and outputs~\citep{zheng2026x}. This illustrates how embodiment-dependent context can be incorporated without duplicating the shared backbone. Prompt conditioning should be distinguished from the task instruction $\ell_{\mathrm{lang}}$. The former describes the execution configuration, while the latter specifies the desired behavior.

Parameter-efficient adapters introduce a restricted set of trainable parameters within a pretrained model, allowing the shared computation to be adjusted for a target configuration. LoRA parameterizes weight updates through low-rank factors, and OpenVLA evaluates this approach for data-efficient robot-policy fine-tuning~\citep{hu2022lora_ch6,kim2024openvla}. In Eq.~\eqref{eq:embodiment_adaptation_objective}, the adapter parameters can constitute $\Theta_{\mathrm{upd}}$ while the remaining pretrained weights are fixed. The resulting update changes the model's internal computation. It does not by itself define a compatible observation or action interface.

An embodiment-specific head maps shared features to the dimensionality, units, and control semantics required by the target robot, implementing $\pi_{\psi_e}$ or its deterministic counterpart $g_{\psi_e}$. HPT explicitly combines a shared transformer trunk with embodiment-specific input stems and policy heads, reinitializing the head for transfer to a new embodiment~\citep{wang2024hpt}. Octo similarly permits new observation tokenizers and action readouts when sensory inputs or action spaces change~\citep{octo2024}. Interface modularity does not prescribe a head-only update. Input modules, readouts, and shared features can be adapted jointly rather than restricting learning to the action head.

A shared latent action space permits policy learning over a common behavioral representation while an embodiment-specific decoder reconstructs executable actions. Unlike the context representation $z_t$, which summarizes evidence for decision-making, the latent action represents the behavior proposed by the policy. Latent Action Diffusion learns aligned end-effector action representations using contrastive encoders, trains a diffusion policy in that latent space, and selects the corresponding embodiment-specific decoder during execution~\citep{bauer2025latent_ch6}. The action-generation interface is thereby factorized into shared latent prediction and embodiment-dependent reconstruction. Its transfer capacity depends on the learned alignment and the target decoder's ability to realize feasible actions.

These mechanisms instantiate different parts of the shared-execution decomposition. Tokens and prompts principally supply embodiment context, adapters modify selected internal computations, and action heads or latent decoders determine how shared information becomes an executable output. The appropriate combination depends on whether the mismatch concerns sensing, morphology, action semantics, or interaction dynamics.

\paragraph{Adaptation Targets and Learning Regimes}
For a target configuration $e^\star$, adaptation should address the source of mismatch. A change in camera mounting position or orientation, even with the same number of cameras, primarily affects observation encoding and calibration. A change in end-effector degrees of freedom, such as replacing a parallel gripper with a dexterous hand, affects action dimensionality, kinematic retargeting, feasible contact configurations, and action decoding. Changed payload, friction, or actuator response affects dynamics estimation and feedback-dependent execution. Several components may require joint adaptation, but the update scope should follow the identified mismatch rather than being determined solely by a generic fine-tuning recipe.

Zero-shot transfer uses the pretrained policy and a declared target interface without target-specific learning data. Any calibration or robot description available at inference should be reported separately. Few-shot adaptation uses a stated, limited target-data budget. Offline and online adaptation describe when the adaptation evidence is collected and used, and are independent of the number of demonstrations. Moreover, adaptation through context, state estimation, or calibration need not change model weights, whereas fine-tuning does. RoboCat explicitly studies adaptation to new tasks and embodiments using target demonstrations~\citep{bousmalis2023robocat}.

\paragraph{Target Performance and Capability Retention}
The complete parameter set $\Theta$ includes the representation parameters, relevant input and action-interface parameters, and any trainable embodiment descriptors. Let $\Theta_0$ denote their values before target adaptation, including the initialization of newly introduced interface modules, and let $\Theta_{\mathrm{upd}}$ denote the selected trainable subset. Following Eq.~\eqref{eq:continual_learning_objective}, parameter adaptation can be formulated as
\begin{equation}
  \begin{aligned}
    \Theta_{\mathrm{upd}}^\star
      &\in\operatorname*{arg\,min}_{\Theta_{\mathrm{upd}}}
      \mathcal{L}_{\mathrm{adapt}}(\Theta),\\
    \mathcal{L}_{\mathrm{adapt}}(\Theta)
      &=\mathcal{L}_{\mathrm{new}}(\Theta;\mathcal{D}_{e^\star})
       +\lambda_{\mathrm{retain}}
        \mathcal{L}_{\mathrm{retain}}(\Theta;\mathcal{D}_{\mathrm{ret}})
       +\lambda_{\mathrm{reg}}
        \mathcal{L}_{\mathrm{reg}}(\Theta,\Theta_0).
  \end{aligned}
  \label{eq:embodiment_adaptation_objective}
\end{equation}
All parameters outside $\Theta_{\mathrm{upd}}$ remain fixed. The target loss $\mathcal{L}_{\mathrm{new}}$ uses the action-generation objective appropriate to Section~\ref{sec:brain-vla}, such as token likelihood, regression, diffusion, or flow matching. The retention dataset $\mathcal{D}_{\mathrm{ret}}$ represents previously supported embodiments, and its loss can use replay supervision or behavioral distillation. The nonnegative coefficients control the relative weighting of these objectives. Consistent with Section~\ref{sec:manipulation-training}, $\mathcal{L}_{\mathrm{reg}}$ constrains undesirable parameter changes. Data and computational budgets are reported separately rather than identified with this regularizer. Elastic weight consolidation exemplifies parameter-based protection of previously acquired knowledge~\citep{kirkpatrick2017ewc}, but such a regularizer does not guarantee retained closed-loop manipulation performance.

Negative transfer can result from incompatible action semantics, an imbalanced training mixture, or shared features that omit information essential to one embodiment. Embodiment-aware sampling, modular interfaces, replay, and targeted regression evaluation can reduce these sources of interference. Physical feasibility and interaction limits must additionally be enforced and validated at the relevant execution interface. Minimizing Eq.~\eqref{eq:embodiment_adaptation_objective} alone does not establish them.

\subsubsection{Adaptation Validation and Deployment Handoff}
\label{sec:brain-embodiment-validation}

Claims of successful embodiment adaptation require evaluation on independently collected target-embodiment trials. Offline action agreement does not measure the state distribution visited by the executing policy or its response to contact disturbances. Evaluation should distinguish supported embodiments from held-out embodiments and separately control changes in task, object, scene, operator, and acquisition period. Splits should be made at the trajectory or collection-session level, with paired or retargeted versions of one demonstration grouped together, to prevent correlated samples from inflating generalization estimates.

\paragraph{Target Performance and Transfer Gain}
Let $\mathcal{E}_{\mathrm{emb}}^{\mathrm{tgt}}\subseteq\mathcal{E}_{\mathrm{emb}}$ be a finite, nonempty set of target embodiments, and let $\mathcal{K}_{e}^{\mathrm{eval}}$ be the finite, nonempty evaluation task set for embodiment $e$. Write $\mathrm{SR}_{N_{\mathrm{demo}}}(e,k)\in[0,1]$ for the success rate on task $k$ after adaptation with $N_{\mathrm{demo}}$ target demonstrations. Here, $N_{\mathrm{demo}}$ is the adaptation-data budget, distinct from the number of evaluation trials. Task completion criteria, initial-state distributions, and aggregation weights must be fixed before comparison. Report absolute target success separately from improvement over a baseline. A budget-matched transfer-gain measure is
\begin{equation}
  \begin{aligned}
    \Delta_{\mathrm{fwd}}(N_{\mathrm{demo}})
      &=\sum_{e\in\mathcal{E}_{\mathrm{emb}}^{\mathrm{tgt}}}
        \sum_{k\in\mathcal{K}_{e}^{\mathrm{eval}}}w_{e,k}
        \bigl[\mathrm{SR}^{\mathrm{adapt}}_{N_{\mathrm{demo}}}(e,k)
             -\mathrm{SR}^{\mathrm{base}}_{N_{\mathrm{demo}}}(e,k)\bigr],\\
    &\hspace{1em} w_{e,k}\geq0,\qquad
      \sum_{e\in\mathcal{E}_{\mathrm{emb}}^{\mathrm{tgt}}}
      \sum_{k\in\mathcal{K}_{e}^{\mathrm{eval}}}w_{e,k}=1.
  \end{aligned}
  \label{eq:forward_transfer}
\end{equation}
For example, the baseline may use the same architecture trained without cross-embodiment pre-training, under the same target-data and target-training budget. This comparison matches the adaptation budget. Pre-training data and computation must be reported separately. A positive difference then supports the benefit of the initialization under that protocol, rather than being inferred from target success alone. Zero-shot comparisons require a separately specified reference policy. This measure is inspired by transfer evaluation in continual learning~\citep{lopezpaz2017gem_ch6}, but is defined here for embodiment adaptation and is not identical to the original sequential-task forward-transfer metric. Cross-robot specialist comparisons in Open X-Embodiment likewise motivate stating the reference system explicitly~\citep{openx2024}.

\paragraph{Retention on Previously Supported Embodiments}
Let $\mathcal{E}_{\mathrm{ret}}$ be a finite, nonempty set of previously supported embodiments on which retention is assessed, excluding the adaptation target when appropriate. Use the same tasks, initial-state distributions, evaluation criteria, and task-aggregation weights before and after adaptation, and define
\begin{equation}
  \Delta_{\mathrm{ret}}
    =\frac{1}{|\mathcal{E}_{\mathrm{ret}}|}
      \sum_{e\in\mathcal{E}_{\mathrm{ret}}}
      \bigl[\mathrm{SR}_{\mathrm{after}}(e)
           -\mathrm{SR}_{\mathrm{before}}(e)\bigr].
  \label{eq:backward_transfer}
\end{equation}
Negative values indicate average degradation under the stated protocol. Retention testing should also report individual embodiment and task results, because an average gain can conceal a substantial loss on one configuration. Any acceptance tolerance should be chosen before adaptation and interpreted with trial counts and uncertainty. This operational measure is related to backward transfer~\citep{lopezpaz2017gem_ch6}. It evaluates retained behavior rather than directly measuring the training loss $\mathcal{L}_{\mathrm{retain}}$.

\paragraph{Adaptation Data Efficiency}
For a target embodiment $e^\star$, a prespecified success threshold $s_{\min}\in[0,1]$, and a set of tested demonstration budgets $\mathcal{N}_{\mathrm{demo}}\subseteq\mathbb{N}_0$, define
\begin{equation}
  N_{\mathrm{demo}}^\star(s_{\min})
    =\min\bigl\{N_{\mathrm{demo}}\in\mathcal{N}_{\mathrm{demo}}:
      \mathrm{SR}_{N_{\mathrm{demo}}}(e^\star)\geq s_{\min}\bigr\},
  \qquad\min\varnothing:=+\infty,
  \label{eq:adaptation_efficiency}
\end{equation}
where $\mathrm{SR}_{N_{\mathrm{demo}}}(e^\star)$ denotes a prespecified weighted average of $\mathrm{SR}_{N_{\mathrm{demo}}}(e^\star,k)$ over the fixed target task set, with nonnegative weights summing to one. The condition $N_{\mathrm{demo}}=0$ represents zero-shot evaluation when included. This is an empirical summary over tested budgets, not a guarantee of monotonic improvement with additional demonstrations. Report the full adaptation curve, uncertainty across trials and training runs, demonstration duration, and the amount of any additional calibration or interaction data. RoboCat's evaluation across demonstration budgets illustrates why data requirements should accompany final success rates~\citep{bousmalis2023robocat}. Adaptation time, computational resources, and the number of updated parameters provide complementary cost measures. These evaluation quantities assess distinct aspects of adaptation. They are not in one-to-one correspondence with the three optimization terms in Eq.~\eqref{eq:embodiment_adaptation_objective}.

\paragraph{Physical Validation and Deployment Interface}
Target trials should include the failure modes that motivated adaptation, such as grasp loss after an end-effector change, tracking error after a control-interface change, or unstable contact under altered compliance. Alongside task success, relevant measurements include executable-command rate, trajectory deviation, contact stability, recovery success, and intervention frequency. Proposed model outputs and applied controller commands should be recorded separately, so that success obtained through controller corrections can be distinguished from intrinsic policy performance.

The deployment handoff comprises more than a neural checkpoint. It should specify the embodiment and controller configuration, observation and action contracts, normalization statistics, calibration and timing information, descriptor and adapter versions, validated operating conditions, and behavior when required inputs or model outputs are unavailable. Unsupported conditions should remain identifiable, and updates must preserve compatibility among these components. Multi-embodiment modeling thus connects reusable task representations with verified, embodiment-specific physical behavior.

\subsection{Model Deployment and Efficiency Optimization}
\label{sec:brain-deployment}

A trained policy becomes a useful embodied capability only when its decisions arrive in time to influence the physical process. Deployment therefore concerns more than fitting a checkpoint into accelerator memory. It connects sensor acquisition, preprocessing, model execution, action interpretation, and feedback control into an operating system with finite computation and variable delay. A policy with higher offline accuracy can perform worse on a robot if it acts on older observations, pauses between action chunks, or loses contact information during preprocessing. The objective is to improve task performance under a specified resource and timing budget.

\subsubsection{Deployment Architectures and the Execution Pipeline}

\paragraph{Where the Model Runs}
Onboard deployment keeps sensing and inference close to the robot and avoids dependence on a network for every policy update. Its limits are accelerator capacity, memory bandwidth, power, and heat dissipation. A nearby workstation or local server can support a larger model and several robots, but introduces transmission and scheduling delays. Cloud deployment provides elastic computation and centralized model management, while adding a wider-area communication path whose delay and availability may be difficult to bound. These are engineering trade-offs rather than a hierarchy in which a more remote or larger computer is necessarily better.

The computation can also be split. A remote process may interpret instructions or generate task context, while an onboard policy produces actions and a local controller regulates motion. The partition should follow decision time scales and communication dependencies. A task goal can remain useful for seconds, while contact feedback may lose value within a small fraction of that interval. Sending every force sample through a remote language model can consequently increase bandwidth and delay without improving the decision that requires the sample. Conversely, splitting a transformer across machines can create repeated inter-device dependencies and be slower than moving the whole policy to one machine.

Table~\ref{tab:brain-deploy-placements} compares the deployment choices. The appropriate configuration depends on the task's tolerance for delayed decisions, the expected concurrency, and the behavior available when an inference request is late. For a manipulation system, task reasoning and learning can often be remote even when action validation, control, and timeout handling must remain local.

\begin{table}[!htbp]
  \centering
  \caption{Deployment placement and its principal performance constraints}
  \label{tab:brain-deploy-placements}
  \small
  \renewcommand{\arraystretch}{1.12}
  \begin{tabularx}{\textwidth}{@{}>{\raggedright\arraybackslash}p{0.15\textwidth} >{\raggedright\arraybackslash}X >{\raggedright\arraybackslash}X >{\raggedright\arraybackslash}X@{}}
    \toprule
    \textbf{Placement} & \textbf{Useful property} & \textbf{Main constraint} & \textbf{Typical allocation} \\
    \midrule
    Robot onboard & Short communication path, local autonomy & Memory, power, cooling & Compact policy and local feedback \\
    Local server & Larger shared accelerator capacity & Network and queueing variability & VLA inference with local execution \\
    Cloud & Centralized resources and fleet-level services & Wide-area delay and outages & Task reasoning, training, and non-urgent analysis \\
    Split deployment & Different compute budgets for different time scales & Context consistency and interface design & Slow reasoning with fast local action updates \\
    \bottomrule
  \end{tabularx}
\end{table}

\paragraph{The Online Data Contract}
Every inference request needs a reference time, an instruction, an embodiment identifier, the observations it requires, and a clear record of what is missing or delayed. The preprocessing pipeline must reproduce the training convention for image crops, color channels, camera ordering, state normalization, action scaling, and coordinate frames. A model trained on relative tool-frame displacement cannot be deployed by interpreting its output as an absolute base-frame pose. Likewise, reversing the sign of a gripper command or applying normalization statistics from another embodiment can invalidate an otherwise correct prediction.

To align modalities sampled at different times, retain each modality's acquisition time and validity mask and construct the input history according to the model's convention. Interpolation can work well for smoothly varying joint positions, but averaging across a contact onset may erase the very event the policy needs. Training augmentations such as random cropping should also be distinguished from the deterministic preprocessing used for evaluation. These details link deployment directly to the data representation of Section~\ref{sec:brain-embodiment-adaptation}.

For a sequential execution path, a useful latency decomposition is
\begin{equation}
  L_{\mathrm{e2e}} = L_{\mathrm{acq}}+L_{\mathrm{pre}}+
  L_{\mathrm{net}}+L_{\mathrm{queue}}+L_{\mathrm{model}}+
  L_{\mathrm{decode}}+L_{\mathrm{dispatch}},
  \label{eq:brain-latency}
\end{equation}
where the terms describe acquisition, preprocessing, communication, queueing, model execution, action conversion, and dispatch to the control interface. Overlapped implementations should instead measure the critical path without double counting concurrent work. The interval ends once a command is available to the controller. When measuring response to a disturbance, the physical actuator response adds further delay. Even with fast accelerator kernels, waiting for camera frames, host-device copies, and server queues can account for substantial delay.

The deployed controller should retain both the requested and the applied command. Saturation, rate limiting, inverse kinematics, or a lower-level safety filter can alter an action after model inference. Subsequent history and learned dynamics should reflect what actually took effect. Recording only the unmodified output makes later failures appear to be modeling errors even when their cause lies in the execution interface.

\subsubsection{Model Compression and Inference Acceleration}

Optimization begins with profiling the actual execution path. A large VLM backbone may be bottlenecked by weight movement and visual-token processing. An autoregressive policy may be slowed by repeated sequential decoding. A generative action expert may spend most of its time on repeated denoising evaluations. For a small policy, preprocessing, device transfers, and Python-level dispatch can become the dominant cost. The benefit of the same optimization therefore varies across architectures and hardware.

\paragraph{Compression of Parameters and Activations}
Quantization stores selected weights or activations at reduced numerical precision. It can lower memory use and data movement, but speed gains require supported kernels and favorable tensor shapes. OpenVLA studies quantized deployment as part of its efficiency analysis~\citep{kim2024openvla}. Calibration for a robot policy should include the states where precision matters, such as small displacements near contact, rare sensor values, and transitions between action modes. Keeping sensitive output projections or normalization operations at higher precision may be preferable to applying the same precision everywhere.

Structured pruning removes channels, heads, layers, or entire processing branches. Unstructured sparsity does not necessarily accelerate dense hardware kernels, and removing a rarely used feature can damage a rare but important recovery behavior. Knowledge distillation trains a smaller student to reproduce useful teacher behavior or representations~\citep{hinton2015distilling_ch6}. The student can be trained on actions, subgoals, intermediate features, or generated trajectories. For closed-loop control, training should cover perturbed states and the teacher's corrections, and recovery should be tested after the student's own errors.

A lightweight action expert provides another form of computation allocation. The expensive backbone constructs semantic context, while a smaller module performs repeated action-generation steps. This division is exemplified by $\pi_0$ and studied through knowledge insulation~\citep{black2024pi0,driess2025insulation_ch6}. Parameter-efficient methods like LoRA mainly cut the cost of adapting a model. They do not, however, make the original backbone any cheaper to run. An adapter merged into supported weights can avoid extra adapter operations, but the main model still runs.

\paragraph{Reducing Work Per Decision}
For autoregressive policies, action tokenization determines the number of sequential backbone evaluations. FAST compresses temporal action structure and thereby changes this trade-off~\citep{pertsch2025fast_ch6}. Parallel prediction can remove token-by-token dependencies, but must be supported by the training objective and attention structure. OpenVLA-OFT combines parallel decoding, action chunking, and continuous regression~\citep{kim2025oft_ch6}. Its result should not be described as merely running the original autoregressive policy faster.

For diffusion and flow-based policies, fewer sampling steps reduce the number of action-expert evaluations. An excessively coarse solver can change action quality, especially at sharp transitions or for narrow contact tolerances. Few-step distillation, combined with suitable training objectives, can improve this trade-off. Reducing the number of inference steps requires a new closed-loop evaluation. The number of steps generated, how many are actually executed, and how often new observations arrive should all be tuned together.

Visual token reduction can lower the cost of processing several camera views. Cropping, token pooling, selective view use, and compact history encoders are possible choices, but the retained information must match the task. A small contact region can occupy few pixels while determining success. Compression that works for reaching may therefore fail during insertion or when distinguishing two similar tools.

\paragraph{Execution-Level Optimization and Cache Validity}
Operator fusion, compilation, fixed-shape execution, preallocated buffers, and reduced host-device synchronization can remove overhead without intentionally changing the policy function. IO-aware attention implementations such as FlashAttention reduce attention memory traffic~\citep{dao2022flash_ch6}. Their end-to-end benefit still depends on whether attention dominates the deployment workload. Warm-up and compilation cost should be separated from steady-state inference, while occasional recompilation and shape changes should remain visible in tail-latency measurements.

Caching is useful only when the dependency being cached is unchanged. Within a flow or diffusion solve, the observation-language prefix can be reused if it cannot attend to changing action tokens. Across different robot observations, cached visual features generally become stale. Even an unchanged instruction does not imply that every text hidden state can be reused if the backbone mixes text and images bidirectionally. A deployment should distinguish reusable tokenization or independent encodings from context-dependent transformer states. This distinction prevents apparent speed from silently reducing responsiveness.

Efficiency should be reported at a consistent boundary. If the policy generates $H$ actions in $L_{\mathrm{model}}$ seconds, its nominal action-generation throughput is
\begin{equation}
  R_{\mathrm{gen}}=\frac{H}{L_{\mathrm{model}}}.
  \label{eq:brain-throughput}
\end{equation}
This quantity describes generated commands. It does not count independent decisions made from fresh measurements. Comparisons must hold the action duration, prediction horizon, execution horizon, input views, sampling settings, and hardware conditions sufficiently explicit to interpret the number.

\subsubsection{Asynchronous Inference and Real-Time Closed-Loop Control}
\label{subsec:asynchronous-motor-execution}

\paragraph{Four Distinct Frequencies}
The sensor frequency determines how often measurements become available. The policy frequency determines how often a new observation-conditioned prediction is produced. The command frequency determines how often targets are delivered to the lower-level controller. The servo frequency determines how often that controller updates actuator inputs. These frequencies can differ by orders of magnitude. Interpolating a predicted trajectory or streaming a long action chunk raises the command frequency without raising the policy's feedback frequency.

Let $\Delta t$ be the interval between action targets, $H$ the predicted chunk length, and $N_{\mathrm{exec}}\leq H$ the number of actions selected for execution before replacing the remaining plan. The prediction horizon is $H\Delta t$, whereas the nominal replanning interval is $N_{\mathrm{exec}}\Delta t$. Action chunking captures temporal correlations and amortizes inference. Receding-horizon execution retains only a prefix and uses new evidence to reconsider the rest. Longer $N_{\mathrm{exec}}$ reduces inference demand but delays responses to unexpected changes. The appropriate value depends on the physical event, not only on available compute.

Observation age gives a more direct measure of responsiveness. If a command applied at physical time $t_{\mathrm{apply}}$ is based on a reference observation acquired at $t_{\mathrm{obs}}$, define
\begin{equation}
  \Delta_{\mathrm{age}}=t_{\mathrm{apply}}-t_{\mathrm{obs}}.
  \label{eq:brain-observation-age}
\end{equation}
For multimodal policies, ages may also be reported separately by modality. A small model runtime does not guarantee small observation age when old frames are queued or a long action prefix is committed. Dropping obsolete requests can improve feedback freshness even when it reduces the total number of processed frames.

\paragraph{Overlapping Prediction and Execution}
In synchronous execution, a robot may finish a chunk and wait for the next inference. Naively removing this pause creates a different problem. The next chunk was generated from a state that no longer coincides with the state at which it will begin. The remaining trajectory must account for commands executed during inference. Real-time chunking addresses this issue for flow and diffusion policies by conditioning the new chunk on a committed prefix while regenerating the uncommitted future~\citep{black2025rtc_ch6}.

If a request requires $L_{\mathrm{req}}$ seconds from input availability to usable output, the approximate number of command intervals that elapse during the request is
\begin{equation}
  d_{\mathrm{delay}}=\left\lceil\frac{L_{\mathrm{req}}}{\Delta t}\right\rceil.
  \label{eq:brain-delay-steps}
\end{equation}
The overlap procedure should preserve the commands that are already committed during those $d_{\mathrm{delay}}$ intervals and use current timing information to select the replacement suffix. For example, at a command rate of $50\,\mathrm{Hz}$, a $100\,\mathrm{ms}$ request spans five action intervals. A 50-action chunk has a one-second prediction horizon, but neither that horizon nor the command rate implies 50 fresh policy decisions per second. This is an illustrative timing calculation, not a benchmark result.

A simple buffer feasibility condition is
\begin{equation}
  N_{\mathrm{buf}}\Delta t \geq L_{\mathrm{req}}^{(p_{\mathrm{quant}})}+L_{\mathrm{margin}},
  \label{eq:brain-buffer}
\end{equation}
where $N_{\mathrm{buf}}$ is the number of validated buffered commands, $L_{\mathrm{req}}^{(p_{\mathrm{quant}})}$ is a selected latency percentile, and $L_{\mathrm{margin}}$ accounts for dispatch uncertainty. This condition cuts down on buffer underruns under the measured latency distribution. It does not guarantee anything in the worst case. A deeper buffer also commits the robot further into an old plan. Responsiveness requires replacing the uncommitted portion when new evidence arrives and defining task-appropriate behavior when the buffer is exhausted.

Chunk transitions require semantic as well as numerical consistency. Averaging two trajectories may smooth a position jump while mixing incompatible modes, such as opening and closing a gripper or maintaining and releasing contact. Continuity should be checked for the commanded quantity and its physical role, including pose, velocity, force targets, and discrete events. A late result associated with an earlier instruction or aborted task should be rejected rather than appended to the current action stream.

\paragraph{Slow Context and Fast Correction}
A multi-rate system can update task context or predictive features less frequently than local action decisions through the following decomposition
\begin{equation}
  c_{k}^{\mathrm{ctx}}=F_{\theta}^{\mathrm{ctx}}(h_{t_k},\ell_{\mathrm{lang}},e),\qquad
  \mathbf{A}_{t}^{\mathrm{seq}}=G_\psi(c_{k(t)}^{\mathrm{ctx}},o_t^{\mathrm{recent}},\mathbf{q}_t,e),
  \label{eq:brain-multirate}
\end{equation}
where $k(t)$ selects the latest valid context. The fast module uses recent evidence, including contact history when available, while the slow module revises task interpretation or predicted futures. For this separation to work, the action module must be trained to operate in context of varying age and must receive enough current information to correct behavior. Merely caching an old full-policy output does not create such a capability.

Replanning can combine periodic updates with event triggers. Target motion, grasp loss, unexpected force, a completed subtask, or a change in instruction may invalidate the current context. Thresholds and hysteresis can prevent excessive switching around noisy events. The event path itself needs a delay bound. Also, a task-level replan should not interrupt a local stabilization response that is physically necessary. The control principles for this fast feedback layer are developed further in Chapter~\ref{ch:whole-body-motor-control}.

\subsubsection{Deployment Evaluation and Operational Maintenance}

Deployment evaluation should compare complete operating configurations. Alongside model weights, record numerical precision, input resolution and views, action horizon, execution horizon, sampling steps, device, concurrent workload, and communication conditions. Otherwise, a higher throughput may reflect a larger chunk or a less responsive execution schedule rather than an improvement in the same task interface.

The evaluation should connect computation to behavior. Table~\ref{tab:brain-deploy-metrics} links deployment measurements to their interpretations and test conditions. Relevant computational quantities include end-to-end latency, latency percentiles and variability, observation age, missed deadlines, peak memory, and power. Relevant physical quantities include task success, completion time, trajectory and contact accuracy, recovery after disturbances, and human intervention. Energy per completed task is often more informative than accelerator power alone. A faster inference configuration may consume more instantaneous power but finish the task sooner, while a slower configuration may increase robot holding time.

\begin{table}[!htbp]
  \centering
  \caption{Deployment measurements and the questions they answer}
  \label{tab:brain-deploy-metrics}
  \small
  \renewcommand{\arraystretch}{1.12}
  \begin{tabularx}{\textwidth}{@{}>{\raggedright\arraybackslash}p{0.21\textwidth} >{\raggedright\arraybackslash}X >{\raggedright\arraybackslash}X@{}}
    \toprule
    \textbf{Measurement} & \textbf{Interpretation} & \textbf{Useful test condition} \\
    \midrule
    Median and tail latency & Typical response and occasional late decisions & Concurrent requests, warm-up, sustained load \\
    Observation age, deadline misses & Freshness of feedback used in execution & Moving targets, delayed frames, network jitter \\
    Memory, power, task energy & Resource feasibility and operating cost & Long runs at the intended thermal limit \\
    Success and completion time & Whether speed changes useful performance & Matched tasks and initial-state distributions \\
    Contact error and recovery & Precision and response beyond nominal motion & Insertion, slippage, unexpected contact \\
    Interventions and fault recovery & Dependence on operator support & Timeouts, sensor loss, server interruption \\
    \bottomrule
  \end{tabularx}
\end{table}

A staged evaluation can first compare numerical outputs on recorded observations. Next, test timing under replay or hardware-in-the-loop execution, and finally measure closed-loop behavior. Recorded-data agreement is a diagnostic rather than a substitute for robot trials. Dynamic-target tasks expose delayed feedback, contact tasks expose precision loss, and long-horizon tasks expose accumulation of small errors. Comparisons should use matched task conditions and report trial counts and uncertainty. A speedup measured on one accelerator does not establish a hardware-independent property.

Operational logs should connect each input snapshot to its model version, preprocessing version, inference timestamps, proposed action, applied command, and observed outcome. This makes it possible to distinguish recognition errors from stale observations, action conversion errors, or controller overrides. Model weights, tokenizer or action decoder, normalization statistics, embodiment adapters, and runtime configuration form one deployment version. Rolling back weights while retaining incompatible normalization or action decoding is not a complete rollback.

Timeout behavior must be appropriate to the task. Holding position may be reasonable in free space but undesirable during a forceful contact. Stopping a pouring motion can require a controlled change in orientation. Such responses should be implemented and validated at the execution layer, with a clear path to human takeover where needed. After compression, compiler changes, or hardware migration, a fixed regression suite should test both nominal skills and the disturbances that previously caused failures.

Deployment is where this chapter's modeling loop closes. Training creates behavioral priors. Adaptation makes them an embodiment-specific interpretation. The execution system then decides when those priors can actually affect the world. Efficiency is established when reduced computation or communication produces timely, accurate behavior under the intended operating conditions. The resulting execution traces then supply the evidence needed for the next round of capability improvement.

\subsection{Dexterous Manipulation Foundation Models}
\label{sec:c6:dexterous}

Most of the models discussed in this chapter act through a parallel gripper, whose action is a wrist pose and a single opening width. A multi-fingered hand breaks three assumptions behind that design. The action space grows to 20 or more finger joints, and whether an object stays under control depends on how those joints move together. Many decisive events, such as a finger beginning to slide or load shifting onto the palm, happen at contacts the cameras cannot see. Robot data are scarcer still. Teleoperating 20 joints is much harder than teleoperating one, and every hand defines its own action space. Human hands, meanwhile, manipulate objects constantly and on camera, which makes human data the obvious place to look for scale.

The subsections examine how to represent finger actions (Section~\ref{sec:c6:dex-actions}) and how to learn from human hands, with EgoScale as the worked example of the stages in Section~\ref{sec:manipulation-training} (Section~\ref{sec:c6:dex-human}). They then examine how much of a policy can be shared across hands, extending Section~\ref{sec:brain-embodiment-adaptation} (Section~\ref{sec:c6:dex-cross}), where touch helps (Section~\ref{sec:c6:dex-touch}), and what current results establish (Section~\ref{sec:c6:dex-eval}). The hands themselves are described in Section~\ref{sec:c9:dexterous-cases}, and the low-level hand controllers that sit beneath these models in Sections~\ref{subsec:ch07:dexterity} and~\ref{subsec:ch07:contact-balance}.

\subsubsection{Action Representations for Multi-Fingered Hands}
\label{sec:c6:dex-actions}

A dexterous policy has to fix the meaning of its action vector before data collection begins. Current systems use one of three options. Joint targets for a specific hand are directly executable but bind the model to that mechanism. Fingertip or keypoint targets are closer to what human data measure, but need a solver to turn them into joint commands. Functional coordinates give each entry a meaning shared across hands, such as index flexion or thumb opposition, and keep a separate map from those entries to each hand's joints. Most systems also command the wrist separately, as a relative change in hand pose. The choice decides which data can supervise the model and how much of it can be shared between hands.

Human data enter through retargeting, the geometric alignment of Section~\ref{sec:brain-embodiment-data}, which for hands can be written as
\begin{equation}
  \min_{\mathbf{q},\mathbf{T}}\ \sum_i w_i\|\mathbf{p}_i^{\mathrm{robot}}(\mathbf{q},\mathbf{T})
                    -\widehat{\mathbf{p}}_i^{\mathrm{human}}\|^2
             +\lambda\|\mathbf{q}-\mathbf{q}_{\mathrm{prev}}\|^2,
  \quad \mathbf{q}_{\min}\leq \mathbf{q}\leq \mathbf{q}_{\max},
  \label{eq:c6dex:retarget}
\end{equation}
where $\mathbf{q}$ is the robot hand configuration, $\mathbf{T}$ aligns the wrist frames, $\mathbf{p}_i$ is a fingertip position, and $w_i$ weights its importance. The second term penalizes abrupt changes. The objective is illustrative, not the loss of a particular study, and practical versions add collision and contact constraints. As Figure~\ref{fig:c9dex:mapping} shows, a MANO pose vector and a robot joint vector do not correspond one to one. Retargeted human motion is useful supervision only insofar as the mapped action keeps the relevant interaction. More data widen coverage but do not remove a systematic geometric mismatch.

\subsubsection{Learning from Human Hands: Pretraining, Alignment, and Post-Training}
\label{sec:c6:dex-human}

\begin{figure}[!b]
  \centering
  \includegraphics[width=\linewidth]{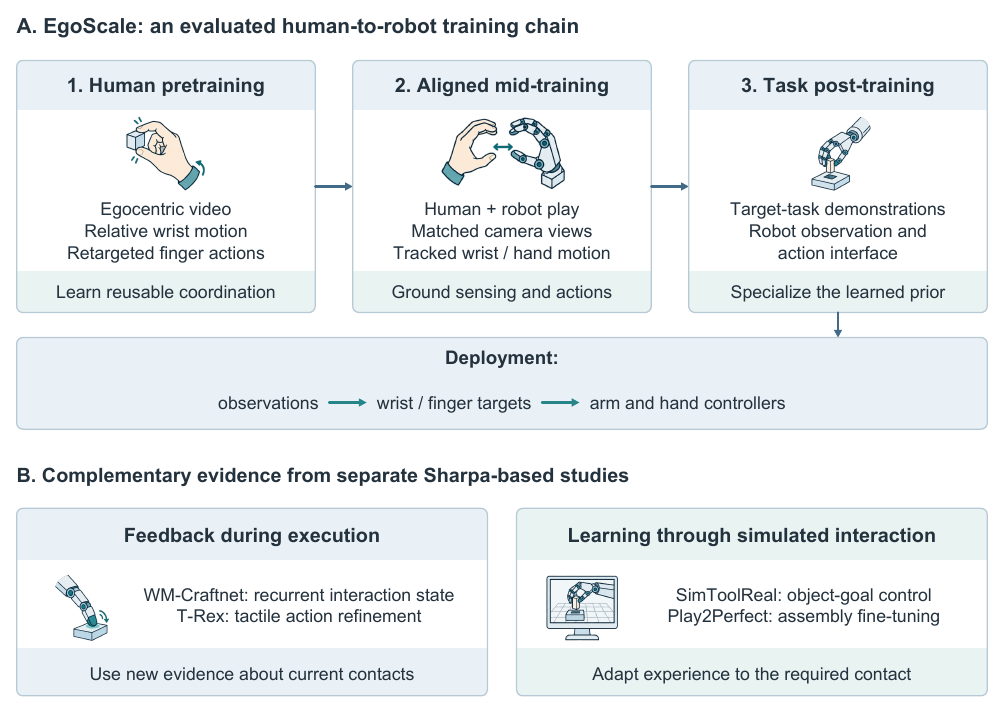}
  \caption{The learning chain on Sharpa Wave. The upper path follows
  EgoScale's training stages, with human and robot observations connected
  through wrist motion and hand-action representations
 ~\citep{zheng2026egoscale,nvidia2026egoscaleproject}. The lower panels
  place separate studies of execution feedback and simulation-based
  adaptation, discussed in Sections~\ref{subsec:ch07:dexterity}
  and~\ref{subsec:ch07:contact-balance}. The grouping is conceptual. The
  studies share no weights or training with EgoScale}
  \label{fig:c6dex:sharpa-learning}
\end{figure}

EgoScale is the clearest evaluated example of the training stages of Section~\ref{sec:manipulation-training} applied to a dexterous hand. It is evaluated on a Galaxea R1Pro with two 22-DoF Sharpa Wave hands and three RGB cameras~\citep{zheng2026egoscale}. The hands are described in Section~\ref{sec:c9:dex-hardware}. Figure~\ref{fig:c6dex:sharpa-learning} lays out the stages.

\paragraph{Pretraining: Learning Reusable Wrist-Finger Coordination}
EgoScale represents human-video motion as relative wrist displacement plus retargeted hand actions~\citep{nvidia2026egoscaleproject}. Predicting articulated actions pushes the model to encode how the wrist and fingers move together. Carrying a bottle without losing the grasp is a different coordination pattern from twisting its cap. Patterns like these recur across tasks and can be learned before any target-robot data are collected. The representation also commits early to an execution interface. Relative wrist motion transfers easily, but the finger component has to be realizable in the target hand's joint space, so the retargeting of Eq.~\eqref{eq:c6dex:retarget} is already present during pretraining.

\paragraph{Alignment: Connecting Human Observations to Robot Actions}
EgoScale's aligned mid-training uses about 50 hours of human and 4 hours of robot play data. The human capture matches the robot's camera viewpoints and intrinsics, with Vive wrist tracking and MANUS hand capture synchronized to video~\citep{zheng2026egoscale}. The question at this stage is which differences in appearance and motion come from the embodiment and which signal a different task state. A human finger and a robot link may occupy different parts of the image while doing the same job of supporting a cup, and similar silhouettes may hide different reachable contact surfaces. In the terms of Section~\ref{sec:brain-embodiment-data}, matched viewpoints and retargeting reduce observation and geometric mismatch, while robot play exposes dynamic mismatch under load. The capture hardware and its calibration are described in Section~\ref{sec:c9:dex-data}.

\paragraph{Post-Training: Specializing the Prior to a Physical Task}
Task-specific robot data then tie the prior to the actual arm, hand, objects, and cameras. For a handle grasp, the prior may already organize approach, closure, and transport, but only the target setup determines how far the fingers must close to engage the handle and which wrist path leaves room for the next step. In EgoScale's scaling experiment, direct post-training without aligned mid-training raises the average task-completion score from 0.30 to 0.71 as human pretraining grows from 1,000 to 20,000 hours~\citep{zheng2026egoscale}. This experiment varies the amount of human data, unlike the comparison with and without pretraining summarized in Section~\ref{sec:embodied-cognition-development}. A separate one-shot protocol uses 1 robot demonstration plus 100 aligned human demonstrations per object, after the earlier stages~\citep{nvidia2026egoscaleproject}. The two experiments answer different questions concerning whether more prior experience helps, and how little robot data is then needed to select a behavior the model already supports. Section~\ref{sec:physical-data-scaling-flywheel} discusses what such scaling variables can and cannot establish.

Two other studies on Sharpa hands use human data in narrower roles. CAIP aligns images with hand-action proxies to pretrain a visual encoder and evaluates policies without tactile input~\citep{sharma2026caip}. Do as I Do reconstructs and retargets human hand-object motion into robot trajectories on a dual-UR3e Sharpa system~\citep{paliwal2026doasido}. Each improves one ingredient, the representation or the execution data. EgoScale shows how the ingredients fit into one training sequence.

\subsubsection{Sharing Actions Across Hands}
\label{sec:c6:dex-cross}

UniDex asks which finger-level action roles can be shared when joint structures differ. Its dataset spans eight hand types and is built by retargeting human demonstrations, with interactive correction and visual alignment. Its Function-Actuator-Aligned Space (FAAS) assigns joints with the same function to common coordinates, the functional-coordinate option of Section~\ref{sec:c6:dex-actions}, and keeps a map to each hand. A policy conditioned on point clouds, language, and proprioception predicts action chunks~\mbox{\citep{unidex2026project}}.

\begin{table}[!t]
  \centering
  \caption{UniDex real-robot results across 5 tasks using Wuji and
  Inspire hands, with 20 trials per task and method. Task progress averages
  stage outcomes. Final success measures task completion. Aggregate means
  reproduced from Fig.~11 of~\citet{zhang2026unidex}. Values are
  percentages}
  \label{tab:c6dex:unidex}
  \small
  \begin{tabularx}{\linewidth}{@{}Xrr@{}}
    \toprule
    Method & Task progress & Final success \\
    \midrule
    $\pi_0$ & 38.0 & 35.0 \\
    UniDex-VLA without pretraining & 32.5 & 23.0 \\
    UniDex-VLA & 81.0 & 76.0 \\
    \bottomrule
  \end{tabularx}
\end{table}

A spray bottle shows what such a mapping has to preserve. The thumb must reach the trigger while the other fingers resist its reaction force. Moving the wrist for better trigger access can weaken the support or misaim the nozzle. And once the trigger moves, the fingers must let it return without dropping the bottle. Minimizing each joint error separately preserves none of these relations.

The Wuji experiments in UniDex use the 20-DoF hand on a Franka arm with RGB-D sensing, on tasks that include a spray bottle, scissors, and a computer mouse, with 50 robot demonstrations per task for adaptation~\citep{zhang2026unidex}. The robot demonstrations correct systematic differences in grasp placement, response, and observation that human data cannot reveal. Table~\ref{tab:c6dex:unidex} reports both outcome measures. Pretraining improves intermediate progress and final success, but because the aggregate pools Wuji and Inspire tasks, it does not isolate the contribution of the Wuji hardware.

Without further training, an Inspire-trained policy transferred to Wuji reaches 40\% average task progress over 10 trials~\citep{zhang2026unidex}. A shared flexion coordinate lets the parameters be shared, but the resulting fingertip motion still depends on link lengths, couplings, and joint ranges, so task structure transfers while contact execution is only partly preserved. SimToolReal (Section~\ref{subsec:ch07:dexterity}) takes the opposite approach. It places the shared interface at the object and leaves finger coordination to the controller.

\subsubsection{Where Tactile Input Helps a Dexterous Policy}
\label{sec:c6:dex-touch}

Section~\ref{sec:multimodal-inputs} describes three ways contact information can enter a VLA. For dexterous hands the narrower question is at which decision a tactile signal changes the outcome. Two XHAND studies give opposite answers.

DexUMI collects demonstrations with an exoskeleton fitted to the target hand and replaces the human hand in the images with the robot hand (Section~\ref{sec:c9:dex-data}). Its policy predicts wrist and finger actions from visual and tactile input~\citep{xu2025dexumi}. In an XHAND kitchen task of turning a knob, moving a pan, and adding salt, touch matters in the salt step because the fingers are hidden from view, and closing the hand to a plausible posture does not show that they reached the salt.\footnote{DexUMI Team, \href{https://dex-umi.github.io/}{\emph{DexUMI: Using Human Hand as the Universal Manipulation Interface for Dexterous Manipulation}}, project page and demonstrations, 2025.} With relative actions and visual adaptation fixed, accumulated success through the salt stage is 75\% with tactile input and 15\% without, over 20 episodes per task~\citep{xu2025dexumi}.

In the shared-autonomy system of~\citet{cui2025sharedautonomy}, DexGrasp-VLA controls the 12-DoF hand from vision and touch during data collection, and the resulting trajectories, supplemented by corrective demonstrations, train an arm-hand VLA with separate arm and hand feature branches. The full system reaches 88.7\% pick-and-place success across 50 objects with 3 trials each. In an ablation, adding tactile features directly to the arm-hand policy lowers success from 95\% to 82\%, although the same features help the hand-only controller~\citep{cui2025sharedautonomy}.

The difference lies in the decision the signal feeds. Touch helps where the next action depends on a contact that vision cannot resolve, and hurts where it becomes one more input to a policy that primarily selects where to reach and place. A third option keeps touch out of the foundation model and uses it in a faster refinement loop beneath it, as T-Rex does (Section~\ref{subsec:ch07:contact-balance}).

\subsubsection{Evaluating Dexterous Foundation Models}
\label{sec:c6:dex-eval}

The studies above report different endpoints (task-completion score, stage-wise progress, final success, and accumulated success through a sequence) on different hands, arms, cameras, objects, and trial counts, and some pool results over several hands. They cannot be ranked against each other. Each supports a narrower claim. EgoScale shows that a larger human prior improves post-trained performance under a fixed protocol. UniDex, that a functional action space allows pretraining across hands and some zero-shot transfer. DexUMI and DexGrasp-VLA, that tactile input pays off only at decisions it can change.

Three questions remain open. Does human pretraining keep scaling once tasks require sustained force, not just reaching the right configuration? Can a shared action space carry contact strategies across hands, or only task structure? And how should a model with slow visual updates hand authority to faster contact-driven control, the interface studied in Chapter~\ref{ch:whole-body-motor-control}? Answering them requires the closed-loop protocols of Chapter~\ref{ch:evaluation}, applied to whole tasks from grasping through recovery on declared hardware.

\clearpage 

\section{Foundation Models for Whole-Body Motor Control}
\label{ch:whole-body-motor-control}

Humanoid and other whole-body robots are an important class of physical embodiments. With improvements in hardware, more recent work focuses on controlling the entire body as a coordinated system. This chapter discusses how to extend current foundation-model algorithms to whole-body motor control. The whole-body control connects task-level decisions to coordinated physical action: it uses control interfaces to regulate motion and contact within the robot's dynamic limits, while adjusting in real time to external forces, terrain changes, and contact transitions. Each of them is discussed in detail as follows.

\subsection{Scope, Taxonomy, and Control Interfaces}
\label{role-and-system-architecture-of-embodied-cerebellum-models}
\label{sec:cerebellum-overview}

\begin{figure}[h]
    \centering
    \includegraphics[width=\linewidth]{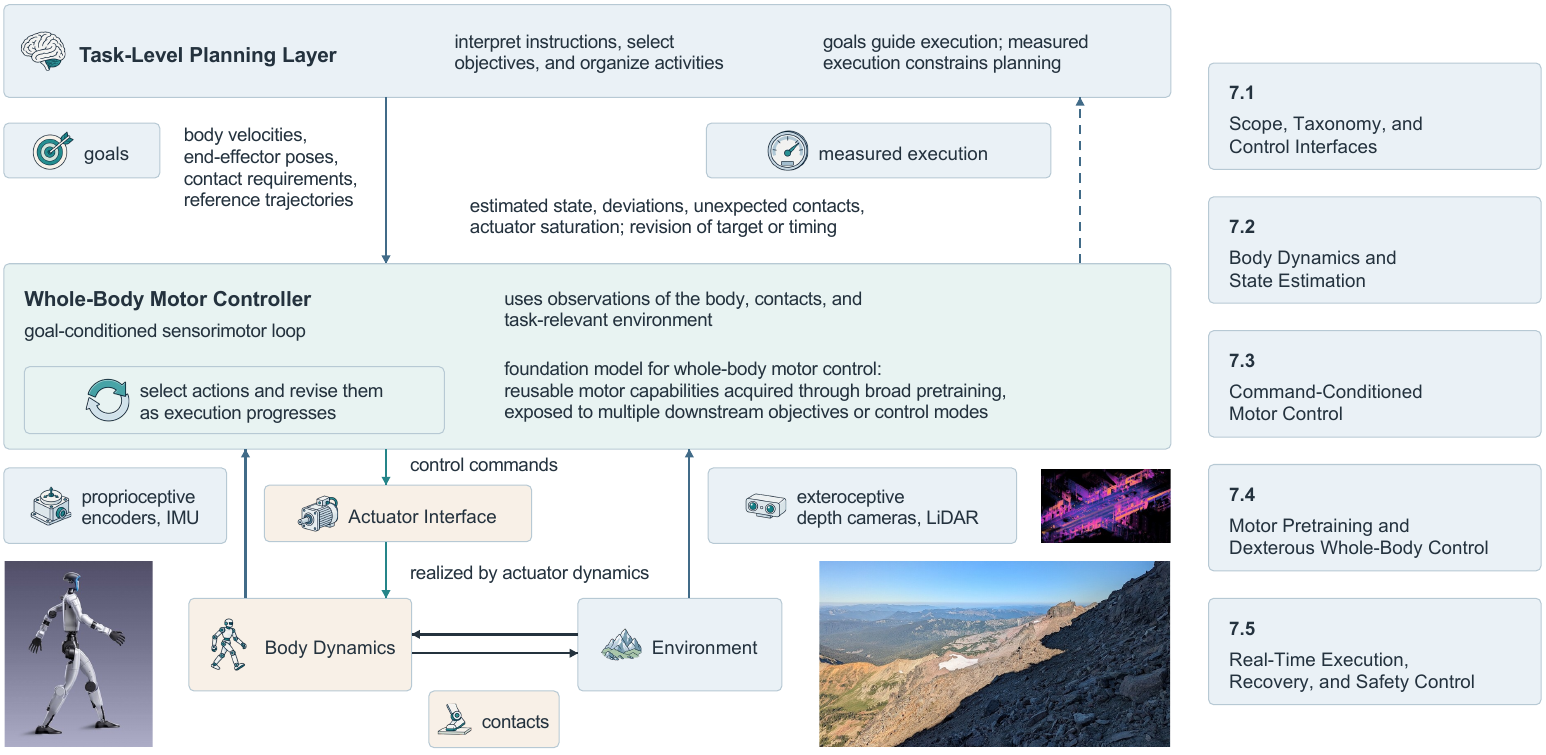}
    \caption{The whole-body motor controller in the robot stack. A task-level planning layer issues goals to a goal-conditioned sensorimotor closed loop, which reads proprioceptive and exteroceptive measurements and issues control commands through the actuator interface. Contacts couple the body to the environment, and measured execution returns to constrain planning. Section markers indicate where Sections~\ref{sec:cerebellum-overview} to \ref{sec:real-time-execution} develop this loop}
    \label{fig:ch07:overview}
\end{figure}

Task-level objectives specify what a robot should accomplish, but generally leave open how its joints and contacts should be coordinated under the body's dynamics and actuation limits. Motor control addresses these execution requirements through feedback from the body and its environment, as Figure~\ref{fig:ch07:overview} illustrates. This section defines the role of whole-body motor control in this process, traces the control principles underlying it, and identifies the interfaces needed to reuse motor capabilities across tasks. In humanoid tasks that combine locomotion and manipulation, body support must be coordinated with arm motion and hand-object contacts.

\subsubsection{Functional Scope and Foundation Models}
\label{functional-definition}

In this chapter, a \emph{whole-body motor controller} is a goal-conditioned sensorimotor model that coordinates robot motion and physical interaction through closed-loop control. It uses available observations of the body, contacts, and task-relevant environment to select and revise actions at high frequency as forces and contact conditions change. Building on this feedback capability, a \emph{foundation model for whole-body motor control} acquires reusable motor capabilities through broad pretraining and exposes them to multiple downstream objectives or control modes~\citep{zeng2025behavior,zeng2026scaling}.

This role complements task-level planning. Task-level models interpret instructions, select objectives, and organize activities. Motion or behavior generators express those objectives as body velocities, end-effector poses, contact requirements, or reference trajectories. The motor controller executes these targets using the available state information. A specified hand position, for instance, may admit different arm configurations, trunk postures, and support arrangements. Selecting a realization requires accounting for its effects on the rest of the body~\citep{khatib1987}.

A motor objective remains subject to execution feedback after it is issued. Small disturbances may be corrected locally. Persistent tracking errors, unexpected contacts, or sustained actuator saturation from torque demands beyond hardware limits may require the upstream planner to revise the target or its timing. Information must therefore flow in both directions. Goals guide execution, while measured execution constrains subsequent planning. A planner must account for estimates of the state reached during execution, including deviations from the preceding reference.

Realizing this feedback loop also requires observation processing, state estimation, target interpretation, actuator interfaces, and mechanisms for enforcing execution limits. Some components may be learned jointly. Others remain explicit. A pretrained tracking policy can provide the core motor capability while relying on external estimation and low-level servos. Such architectures combine explicit control components with learned motor capabilities, drawing on complementary developments in model-based control and reinforcement learning.

\subsubsection{From Control Principles to Reusable Motor Capabilities}
\label{evolution-of-motor-control}

For articulated robots, the main difficulty is that control objectives are coupled through mechanics. Operational-space control expresses motion and force objectives in task coordinates and relates them to joint actuation, making end-effector behavior an explicit object of control~\citep{khatib1987}. Impedance control further regulates the dynamic relationship between motion and interaction force, rather than requiring exact position tracking in all circumstances~\citep{Hogan1985}. Classical work on robot hands similarly analyzes grasping through fingertip geometry, force transmission, and frictional interaction~\citep{mason1985hands}. Together, these methods establish that an appropriate motor interface can specify a physical interaction rather than a complete sequence of joint configurations. In legged robots, the same mechanical coupling must be coordinated across repeated contact transitions.~\citet{raibert1984hopping} decompose hopping into regulation of forward velocity, body attitude, and hopping height.

To account for future motion and support requirements, MPC makes the consequences of candidate actions explicit over a finite horizon. At each update, it optimizes a control sequence from the current state estimate, applies the initial input, and repeats the optimization after receiving new feedback. State and input constraints can be included directly, although feasibility and stability depend on the model, formulation, and assumptions~\citep{mayne2000constrained}. In 2003, zero-moment-point preview control provides a related predictive formulation for biped walking, using future support references to shape the current motion through optimal linear feedback~\citep{kajita2003preview}. By 2018, the MIT Cheetah 3 controller uses convex MPC to optimize ground-reaction forces over a finite horizon, while a gait state machine and swing controller supply contact timing and swing-leg motion~\citep{dicarlo2018}.

In the late 2010s, deep reinforcement learning (RL) expands the use of simulated interaction to acquire feedback policies. DeepMimic combines reference-motion imitation with task rewards to learn coordinated skills for physics-based characters~\citep{peng2018deepmimic}.~\citet{hwangbo2019} train policies for ANYmal using a simulator that combines rigid-body dynamics with an actuator model learned from robot data, then deploy those policies on the physical robot.~\citet{lee2020} extend locomotion to challenging terrain by training a privileged teacher and distilling its behavior into a student that uses proprioceptive history. History-conditioned adaptation further uses recent observations and actions to infer information about unobserved dynamics, including changes in load or ground interaction. Rapid Motor Adaptation (RMA) makes this inference a dedicated learned module that conditions a locomotion policy~\citep{kumar2021}. The estimation methods supporting such adjustment are developed in Section~\ref{sec:dynamics-estimation}.

A parallel line develops motor knowledge that downstream tasks can reuse. Neural probabilistic motor primitives (NPMP) compress expert skills into a shared latent representation~\citep{merel2018npmp}. An adversarial motion prior (AMP) learns a distributional motion prior that shapes policy optimization~\citep{peng2021amp}. Adversarial Skill Embeddings (ASE) learns a skill-conditioned policy whose latent space can serve as the action space of a downstream controller~\citep{peng2022ase}. A distributional prior shapes policy learning, while a skill-conditioned policy generates executable actions. Their shared contribution is to separate the acquisition of coordinated movement from the specification of a particular downstream task.

Modern systems develop both hybrid execution architectures and more reusable command interfaces. Deep Tracking Control combines an online trajectory-optimizing planner with a learned tracking policy that executes its references under model mismatch and disturbances~\citep{jenelten2024}. This combination retains explicit predictive structure while learning the feedback needed during execution. More recent humanoid controllers extend reuse to the control interface. HOVER distills motion-tracking capability into a policy supporting multiple command modes~\citep{he2025hover}. A behavior foundation model (BFM) uses motion imitation and masked online distillation to pretrain a shared behavioral repertoire, while ScaleBFM studies direct goal-conditioned RL pretraining with varied motion data and model capacity~\citep{zeng2025behavior,zeng2026scaling}. These humanoid approaches allow downstream applications to invoke learned movement through different target specifications, as developed in Section~\ref{sec:whole-body-control}.

These approaches reflect two complementary design choices concerning the \emph{execution mechanism} and \emph{how motor knowledge is acquired and reused}. Table~\ref{tab:ch07:motor-taxonomy} shows representative combinations.

\begin{table}[htbp]
\centering
\caption{Selected control approaches compared by execution mechanism and the acquisition and reuse of motor knowledge}
\label{tab:ch07:motor-taxonomy}
\small
\setlength{\tabcolsep}{4pt}
\begin{tabularx}{\linewidth}{@{}
>{\raggedright\arraybackslash}p{0.18\linewidth}
>{\raggedright\arraybackslash}X
>{\raggedright\arraybackslash}X
>{\raggedright\arraybackslash}X@{}}
\toprule
\textbf{Approach} & \textbf{Execution mechanism} & \textbf{Acquisition and reuse} & \textbf{Representative examples} \\
\midrule
Analytical feedback & Task-space motion and force feedback & Explicit models and designed interaction objectives & Operational-space and impedance control~\citep{khatib1987,Hogan1985} \\
\addlinespace
Predictive optimization & Online finite-horizon optimization with feedback & Dynamics and constraints used to recompute motion or force plans & Cheetah 3 convex MPC~\citep{dicarlo2018} \\
\addlinespace
RL feedback policies & Observation- and command- or reference-conditioned actions & Task rewards or motion imitation\newline History-based inference can support adaptation & DeepMimic, ANYmal, RMA~\citep{peng2018deepmimic,hwangbo2019,kumar2021} \\
\addlinespace
Motor priors and skills & Feedback policy, optionally conditioned on a skill latent & Demonstration distributions or distilled experts reused in downstream learning & NPMP, AMP, ASE~\citep{merel2018npmp,peng2021amp,peng2022ase} \\
\addlinespace
Multi-mode motor pretraining & Shared goal- or command-conditioned policy & Distillation or broad RL pretraining exposes a reusable repertoire & HOVER, BFM, ScaleBFM~\citep{he2025hover,zeng2025behavior,zeng2026scaling} \\
\addlinespace
Hybrid planning and tracking & Online optimization with a learned feedback tracker & Planned references executed by a policy trained for tracking & Deep Tracking Control~\citep{jenelten2024} \\
\bottomrule
\end{tabularx}
\end{table}

\subsubsection{Architecture and Interfaces}
\label{architecture-and-interfaces}

\begin{figure}[t]
    \centering
    \includegraphics[width=\linewidth]{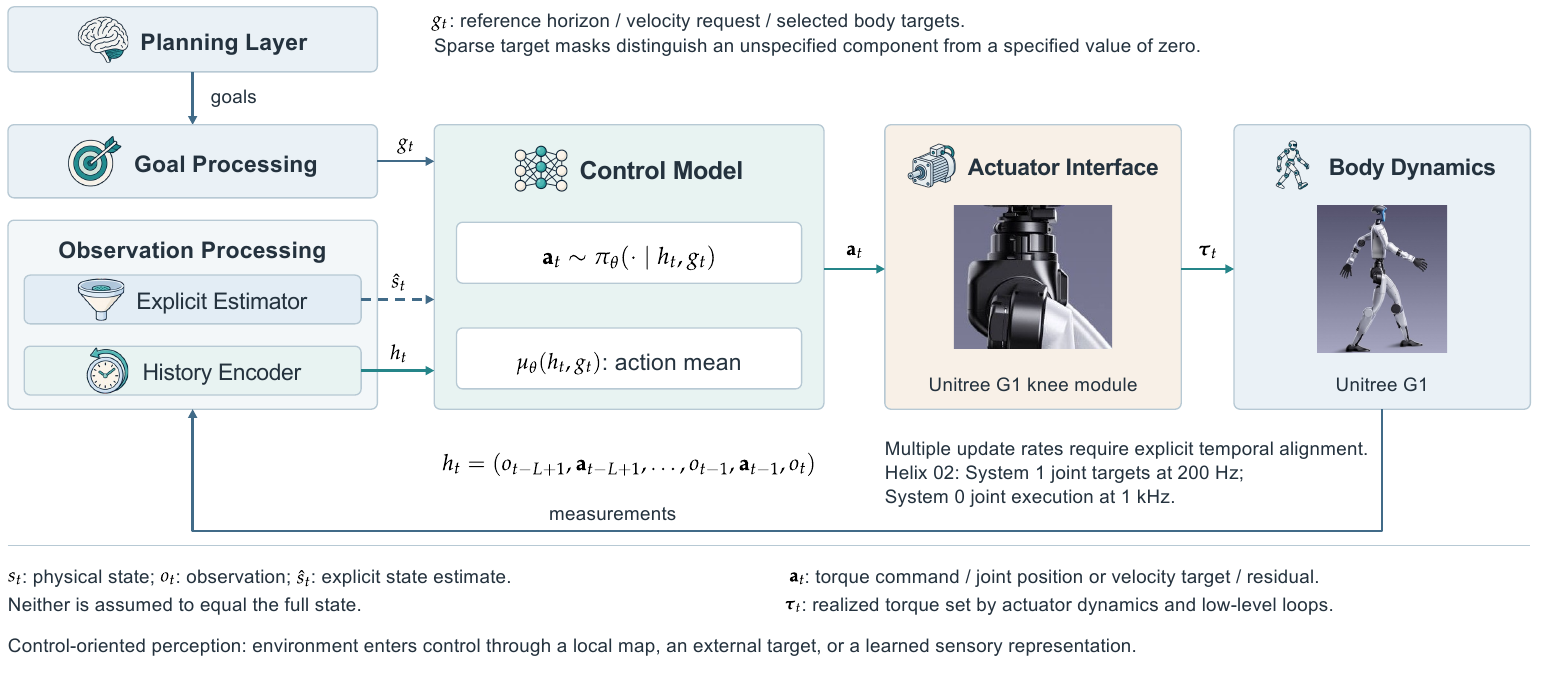}
    \caption{Closed-loop architecture and interfaces of a whole-body motor controller. Goal processing supplies $g_t$, observation processing supplies $h_t$ and $\hat{s}_t$, and the control model outputs $\mathbf{a}_t$, which the actuator interface realizes as joint torques $\boldsymbol{\tau}_t$ through actuator dynamics and low-level loops. Measurements close the loop. Operation at multiple update rates requires explicit temporal alignment}
    \label{fig:ch07:architecture}
\end{figure}

A motor-control architecture connects state representation, goal interpretation, and action generation, as Figure~\ref{fig:ch07:architecture} illustrates. In a humanoid, moving the arms changes whole-body dynamics, while stepping changes the support configuration and the position from which the hands act. A requested behavior therefore has to be realized through coordinated body motion and corrected as the resulting motion and contacts evolve.

Figure's Helix 02 provides a concrete example through its publicly described System 2-System 1-System 0 hierarchy.\footnote{Figure, \href{https://www.figure.ai/news/helix-02}{\emph{Introducing Helix 02: Full-Body Autonomy}}, January 27, 2026.} System 2 interprets scenes and language, sequences behaviors, and supplies latent goals to System 1. Conditioned on these latents, System 1 combines head-camera and palm-camera images, fingertip tactile sensing, and full-body proprioception to generate full-body joint targets at 200~Hz. System 0 is a learned whole-body controller that tracks these targets, using full-body joint state and base motion to produce joint-level actuator commands at 1~kHz. This division connects semantic goals to visuomotor targets and then to feedback execution that handles balance, contact, and whole-body coordination. The additional sensing also serves specific feedback needs. Palm cameras provide in-hand visual feedback when an object is occluded from the head camera, and fingertip tactile sensors detect contact forces.

Throughout this chapter, $s_t$ denotes the physical state at control step $t$, including the robot and any environment variables needed to describe its evolution. The observation $o_t$ contains information available during execution. An explicit state estimate is denoted by $\hat{s}_t$. Neither is assumed to equal the full state. A finite observation-action history is
\begin{equation}
 h_t=(o_{t-L+1},\mathbf{a}_{t-L+1},\ldots,o_{t-1},\mathbf{a}_{t-1},o_t),
 \qquad \mathbf{a}_t\sim\pi_\theta(\cdot\mid h_t,g_t),
 \label{eq:cerebellum-interface}
\end{equation}
where $L\geq1$ is the history length, $\theta$ denotes policy parameters, and $g_t$ is the command representation. The policy $\pi_\theta(\cdot\mid h_t,g_t)$ is a conditional action distribution, and $\sim$ denotes sampling from it. For $L=1$, $h_t=o_t$. A recurrent state or learned history encoding can summarize longer histories.

Privileged observations provide training-time access to variables that can be read directly in simulation but are difficult to obtain reliably from the robot's sensors. HUGWBC uses base linear velocity, ground friction, and foot contact forces as such variables~\citep{xue2025hugwbc}. In its asymmetric actor-critic training, the critic combines these quantities with proprioception and commands to predict returns. Its value estimates then enter the advantage estimates used to update the actor with PPO~\citep{schulman2017ppo,hugwbcCode2026}. This allows the simulator's richer information to contribute to policy optimization through the critic. The actor's execution inputs include recent proprioceptive history, commands, and estimates of selected hidden quantities obtained from that history~\citep{xue2025hugwbc,hugwbcCode2026}.

Command granularity determines how much of the movement the controller must choose. A desired body velocity leaves freedom in foothold placement and posture, as in the locomotion commands of Section~\ref{sec:motion-control}. A full reference motion constrains the coordinated body trajectory more closely, while targets for only selected body links leave the remaining movement free to accommodate support and interaction constraints. Desired reference quantities are denoted by $s_t^{\rm ref}$, with the hat reserved for estimates. The policy receives their command representation $g_t$, which can encode a velocity request, a reference horizon, or selected body targets. For partial targets, a mask identifies the specified components and distinguishes an unspecified target from a specified value of zero, as in the masked-target representation of Section~\ref{sec:whole-body-control}.

Executing these goals requires translating the policy's action $\mathbf{a}_t$ into physical joint actuation. When $\mathbf{a}_t$ contains joint-position targets, a low-level servo converts position error and joint feedback into torque commands. Actuator dynamics then determine the actual joint torque $\boldsymbol{\tau}_t$, including the effects of saturation and delay~\citep{hwangbo2019}. Policies may instead output joint-velocity targets, direct torque commands, or a residual added to an existing controller, changing what the downstream control interface must realize. For a stochastic training policy, $\boldsymbol\mu_\theta(h_t,g_t)$ denotes its action mean, which can also be used for deterministic execution.

Consider a hand target fixed in the world while the humanoid turns or takes a step. The target stays fixed, but its pose relative to the body changes, so a body-frame encoding must be updated using the current body pose. An object-relative target instead follows the object's motion. Timing matters for the same physical reason. A reference computed for one support configuration may need revision after touchdown or lift-off. When perception, goal generation, and actuation update at different rates, a controller can receive observations or references describing an earlier body or contact state. Feedback correction between target updates must therefore relate these inputs to the current motion. A measurement timestamp indicates the time of the observed state, and a reference's validity interval indicates when it is intended to apply. Delayed information can increase tracking error, motivating the latency analysis in Section~\ref{sec:real-time-execution}.

For a humanoid traversing uneven ground through a low passage, successful motion requires both placing the feet and keeping the torso clear of overhead obstacles. Proprioceptive history reflects body motion and the response to contact, supporting correction of ongoing execution. Exteroceptive observations supply the geometry ahead of the robot so that motion can be adapted before contact. Visual servoing uses the discrepancy between current and desired visual features to correct robot motion~\citep{chaumette2006visual}. For perceptive locomotion, Miki et al. combine proprioception with terrain-height observations from a robot-centric elevation map, allowing the controller to adapt its motion before contact while retaining feedback from body motion~\citep{miki2022}. A ground-height map, however, does not preserve all geometry relevant to a humanoid's whole body. Gallant instead uses LiDAR-derived voxel grids together with proprioception to represent lateral obstacles and overhead constraints as well as ground-level terrain~\citep{ben2025gallant}. These examples illustrate \emph{control-oriented perception}, in which the system retains the information needed to select a physical action and correct it as the body interacts with its surroundings.

\subsubsection{Generality, Dexterity, and Evaluation}
\label{general-purpose-motor-models}

Generality concerns reuse under changes in objectives and physical conditions. Reproducing many training motions establishes behavioral coverage. A more demanding test is whether the controller can combine familiar capabilities to meet new task requirements. Goal flexibility allows an existing capability to be invoked through different target representations, while task transfer asks whether it remains useful in a new activity. A controller may track new references yet need substantial retraining for new interaction dynamics. Reuse therefore involves both executing new goal combinations and adapting coordinated behavior to changes in load, terrain, or contact.

These demands become tightly coupled when locomotion and manipulation share the same body. Maintaining an object pose during locomotion imposes constraints that are absent from either unconstrained walking or stationary manipulation~\citep{fu2022}. For a humanoid carrying an object with both hands, body acceleration changes the forces required to support the object, while arm configuration affects both reach and whole-body balance. Dexterous execution additionally requires finger configurations that maintain the grasp. An end-effector pose specifies a hand's position and orientation but leaves its internal contact configuration unresolved. In bimanual manipulation, the two hands interact through the shared object. Their contact forces jointly determine object motion, but different force distributions can produce the same resultant wrench while imposing different grasping loads. Coordinated execution must therefore account for support, object motion, and contact maintenance together, especially during load transfer or changes in the grasp~\citep{bicchi1994forces}. The learned hand-control and force-adaptive balance examples in Sections~\ref{subsec:ch07:dexterity} and~\ref{subsec:ch07:contact-balance} provide concrete examples of contact execution and load-dependent body support.

Extending this reuse to another embodiment introduces changes in the body itself. Robots differ in topology, geometry, actuation, and sensing, so a shared controller must represent these differences or accommodate them through embodiment-specific components. Morphology-conditioned transformers provide one approach to sharing control representations across bodies~\citep{gupta2022metamorph}. Task-space and contact-based goals can reduce dependence on a particular joint layout, but the mapping from those goals to actions remains constrained by each robot's capabilities. Transfer to a physical platform additionally depends on its actuator and sensor characteristics, beyond the morphological variation represented in simulation.

The carrying example makes these forms of generality directly testable. Changing walking speed and object-pose targets tests the composition of movement requirements. Varying the carried load tests adaptation. Using a robot with a different morphology tests embodiment transfer. Tracking error, object slip, and loss of balance assess target execution, grasp maintenance, and body support, respectively. For goals or conditions that require further learning, the benefit of pretraining can be measured by the data and optimization needed to reach a given task performance, compared with learning from scratch~\citep{merel2018npmp,zeng2025behavior}.

\subsection{Body Dynamics and State Estimation}
\label{sec:dynamics-estimation}

\begin{figure}[!t]
    \centering
    \includegraphics[width=\linewidth]{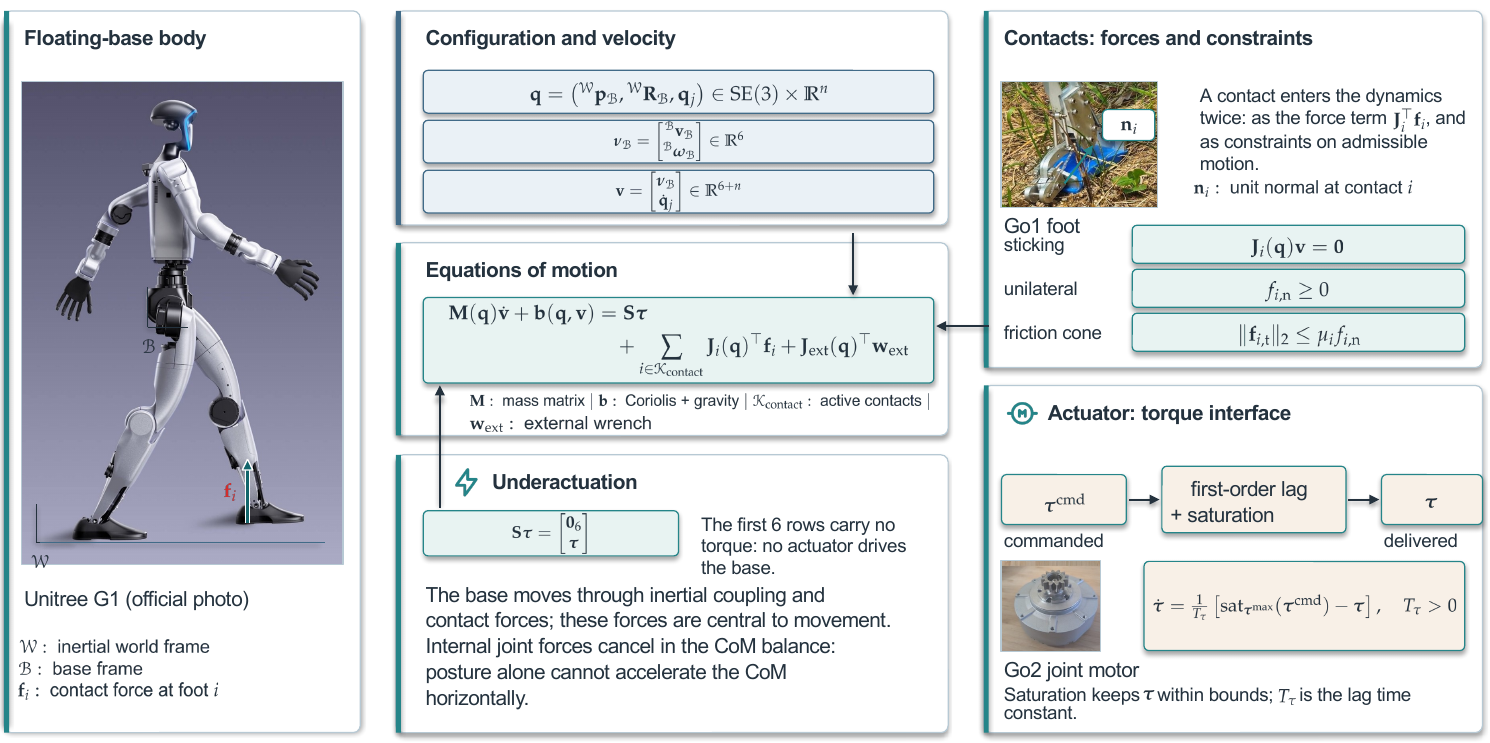}
    \caption{Floating-base dynamics, contacts, and actuation of a legged robot. Joint torques act through the selection matrix on the joint rows only, so contact forces are central to movement. Contacts contribute forces subject to sticking, unilateral, and friction-cone constraints, and the delivered torque follows the commanded torque through a first-order lag with saturation}
    \label{fig:ch07:body-dynamics}
\end{figure}

The motor-execution layer is a high-frequency control loop around a physical system. Figure~\ref{fig:ch07:body-dynamics} depicts floating-base dynamics, contact constraints, and the actuation interface. It receives task commands from the planning layer, reads proprioceptive measurements (joint encoders and an inertial measurement unit, IMU) together with exteroceptive ones such as depth cameras and LiDAR, and issues control commands realized through the actuator interface. Three questions organize this loop, asking how the body moves, where it is now and what it touches, and what the model is getting wrong. This section develops the three answers in turn. Throughout, $s_t$ denotes the actual state, $\hat{s}_t$ its estimate, and $s_t^{\rm ref}$ a commanded reference.

\subsubsection{Motion-Control Foundations: Floating-Base Dynamics, Contacts, and Actuators}
\label{subsec:motion-foundations}

A legged robot is a \emph{floating-base} system. Six base degrees of freedom carry no actuators, and only the joints are driven. Throughout this section, $\mathcal{W}$ denotes the inertial world frame and $\mathcal{B}$ the base frame, and a left superscript marks the frame in which a quantity is expressed. The world vertical is opposite the gravity direction, and ${}^{\mathcal{W}}\mathbf{R}_{\mathcal{B}}$ maps base-frame vectors to world coordinates. With $n$ actuated joints, the configuration combines the base pose with the joint angles, and the generalized velocity stacks the base spatial velocity (linear and angular), expressed in $\mathcal{B}$, with the joint rates, giving
\begin{equation}
\begin{gathered}
\mathbf{q}=\left({}^{\mathcal{W}}\mathbf{p}_{\mathcal{B}},
{}^{\mathcal{W}}\mathbf{R}_{\mathcal{B}},\mathbf{q}_j\right)
\in\mathrm{SE}(3)\times\mathbb{R}^{n},\\
\boldsymbol{\nu}_{\mathcal{B}}=
\begin{bmatrix}{}^{\mathcal{B}}\mathbf{v}_{\mathcal{B}}\\
{}^{\mathcal{B}}\boldsymbol{\omega}_{\mathcal{B}}\end{bmatrix}\in\mathbb{R}^{6},
\qquad
\mathbf{v}=\begin{bmatrix}\boldsymbol{\nu}_{\mathcal{B}}\\
\dot{\mathbf{q}}_j\end{bmatrix}\in\mathbb{R}^{6+n}.
\end{gathered}
\label{eq:body-kinematics}
\end{equation}
The base twist $\boldsymbol{\nu}_{\mathcal{B}}$ contains the linear velocity of the base origin and its angular velocity, both expressed in $\mathcal{B}$. Using this velocity convention consistently is essential. Configuration derivatives are not simply $\dot{\mathbf{q}}=\mathbf{v}$. In particular, ${}^{\mathcal{W}}\dot{\mathbf{p}}_{\mathcal{B}} ={}^{\mathcal{W}}\mathbf{R}_{\mathcal{B}}{}^{\mathcal{B}}\mathbf{v}_{\mathcal{B}}$, and rotation evolves through the base angular velocity. The equations of motion are
\begin{equation}
\mathbf{M}(\mathbf{q})\dot{\mathbf{v}}+\mathbf{b}(\mathbf{q},\mathbf{v})
=\mathbf{S}\boldsymbol{\tau}
+\sum_{i\in\mathcal{K}_{\mathrm{contact}}}\mathbf{J}_i(\mathbf{q})^{\top}\mathbf{f}_i
+\mathbf{J}_{\rm ext}(\mathbf{q})^{\top}\mathbf{w}_{\rm ext},
\label{eq:eom}
\end{equation}
where $\mathbf{M}(\mathbf{q})\in\mathbb{R}^{(6+n)\times(6+n)}$ is the symmetric positive definite mass matrix. The bias term $\mathbf{b}=\mathbf{C}(\mathbf{q},\mathbf{v})\mathbf{v}+\mathbf{g}(\mathbf{q})$ collects Coriolis, centripetal, and gravity forces. The vector $\boldsymbol{\tau}\in\mathbb{R}^n$ contains the joint torques. The selection matrix
\begin{equation}
\mathbf{S}=\begin{bmatrix}\mathbf{0}_{6\times n}\\\mathbf{I}_n\end{bmatrix}
\in\mathbb{R}^{(6+n)\times n}
\label{eq:selection}
\end{equation}
places torques on the joint rows only. $\mathcal{K}_{\mathrm{contact}}$ is the set of active point contacts, with world-frame forces $\mathbf{f}_i\in\mathbb{R}^3$. $\mathbf{J}_i\in\mathbb{R}^{3\times(6+n)}$ maps the generalized velocity to the world-frame velocity of contact point $i$. A contact patch can be represented by multiple such points. The six-dimensional external wrench $\mathbf{w}_{\rm ext}$ and its spatial Jacobian use matching world-aligned coordinates and the same moment reference point. Additional external interactions contribute analogous terms. Written blockwise, the dynamics are
\begin{equation}
\begin{bmatrix}
\mathbf{M}_{\mathcal{B}\mathcal{B}}&\mathbf{M}_{\mathcal{B}j}\\
\mathbf{M}_{j\mathcal{B}}&\mathbf{M}_{jj}
\end{bmatrix}
\begin{bmatrix}\dot{\boldsymbol{\nu}}_{\mathcal{B}}\\\ddot{\mathbf{q}}_j\end{bmatrix}
+\begin{bmatrix}\mathbf{b}_{\mathcal{B}}\\\mathbf{b}_j\end{bmatrix}
=\begin{bmatrix}\mathbf{0}_{6}\\\boldsymbol{\tau}\end{bmatrix}
+\mathbf{J}_{\mathcal{K}_{\mathrm{contact}}}^{\top}\mathbf{f}_{\mathcal{K}_{\mathrm{contact}}}
+\mathbf{J}_{\rm ext}^{\top}\mathbf{w}_{\rm ext},
\label{eq:partitioned}
\end{equation}
where $\mathbf{J}_{\mathcal{K}_{\mathrm{contact}}}$ stacks the Jacobians of the active contacts and $\mathbf{f}_{\mathcal{K}_{\mathrm{contact}}}$ their forces. The equations expose the central structural fact of legged robotics. The first six rows contain no joint torque, so no actuator acts directly on the base. Joint torque moves the base through inertial coupling (an airborne robot can pitch its trunk by swinging its legs) and through contact with the environment. Internal joint forces cancel in the center-of-mass (CoM) balance, so changing limb or trunk posture cannot create horizontal CoM acceleration when the net external horizontal force is zero. Foot friction, hand contacts, and other external interactions supply such forces. This underactuation makes contact forces central to movement~\citep{wensing2024}.

Contacts enter the equations both as forces and as constraints on admissible motion. A foot in sticking contact with stationary ground satisfies $\mathbf{J}_i(\mathbf{q})\mathbf{v}=\mathbf{0}$. Each contact is unilateral. The ground pushes but does not pull, and friction bounds the tangential force. In world coordinates, let $\mathbf{n}_i$ be the unit surface normal and $\mu_i$ the friction coefficient, and split $\mathbf{f}_i$ into a normal part $f_{i,\mathrm{n}}=\mathbf{f}_i\cdot\mathbf{n}_i$ and a tangential part $\mathbf{f}_{i,\mathrm{t}}=\mathbf{f}_i-f_{i,\mathrm{n}}\mathbf{n}_i$. Unilateral contact and Coulomb friction require
\begin{equation}
f_{i,\mathrm{n}}\geq0,\qquad\|\mathbf{f}_{i,\mathrm{t}}\|_2\leq\mu_i f_{i,\mathrm{n}}.
\label{eq:friction-cone}
\end{equation}
The pinned foot turns the stance leg into a reconfigurable strut that couples the base to the ground, and the same coupling is what lets the base state be inferred from joint encoders and an IMU, as the next subsection shows. Contact forces must also be compatible with the available joint torques. Contacts are ultimately hybrid. A contact is open, sticking, or sliding, and impact is discontinuous. Time-stepping complementarity formulations capture this without pre-specifying a mode sequence~\citep{stewart1996,posa2014}, while deployed controllers often fix a contact sequence from a gait or footstep plan and optimize the force distribution within each mode~\citep{dicarlo2018}.

Two further properties of the dynamics recur throughout control and identification. The Coriolis matrix can be chosen with an energy property that underlies passivity-based control arguments, and the dynamics are linear in the inertial parameters, with ten parameters per rigid link. This linearity forms the basis of classical identification~\citep{khalil2002,featherstone2008}. Standard recursive algorithms evaluate inverse dynamics and unconstrained forward dynamics in time linear in the number of joints. Forming a dense mass matrix, resolving contacts, and solving control optimizations have additional costs~\citep{featherstone2008}. Any gap between the simulated body and the real one is a model error the deployed controller must absorb.

So far the joint torque is a free input. On hardware it is not. The actuator sits between the commanded torque and the torque that reaches the joint. Three common designs illustrate different trade-offs. Hydraulic actuation provides high force density, as on HyQ~\citep{semini2011}. Series elastic actuation introduces compliance that protects the drivetrain and makes torque measurable from spring deflection at the cost of bandwidth~\citep{pratt1995}. And quasi-direct drive uses high-torque motors behind a low-ratio gearbox, as on the MIT Cheetah and Mini Cheetah~\citep{seok2015,katz2019}. A simplified torque-interface model uses a first-order lag with saturation,
\begin{equation}
\dot{\boldsymbol{\tau}}
=\frac{1}{T_{\tau}}
\left[\operatorname{sat}_{\boldsymbol{\tau}^{\max}}
\left(\boldsymbol{\tau}^{\rm cmd}\right)-\boldsymbol{\tau}\right],
\qquad T_{\tau}>0,
\label{eq:actuator-model}
\end{equation}
where saturation clips each commanded torque to its admissible interval, and the time constant $T_\tau$ is measured in seconds. Initialized within those intervals, this model preserves the torque bounds. Position-target actions additionally require a model of the intervening servo. Some actuator physics is difficult to capture with a hand-written model. Hwangbo et al.\ train an actuator network on physical-robot data, mapping a history of joint position errors and velocities to delivered torque, and policies trained with this model transfer to hardware~\citep{hwangbo2019}.

\subsubsection{Explicit State Estimation: Attitude, Velocity, Contact, and Local Terrain}
\label{subsec:explicit-estimation}

\begin{figure}[!t]
    \centering
    \includegraphics[width=\linewidth]{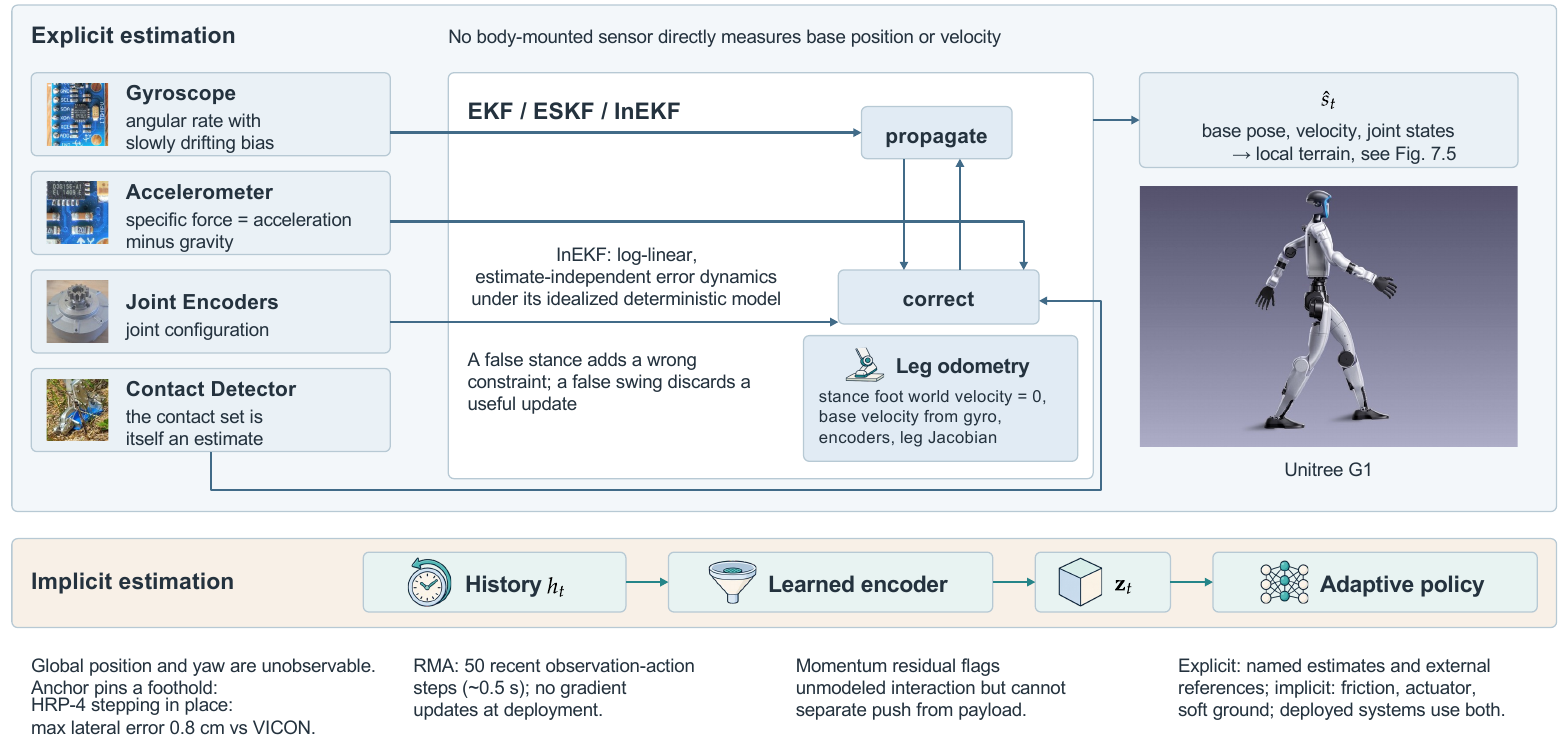}
    \caption{Explicit and implicit state estimation on a legged robot. Explicit estimation fuses inertial and kinematic measurements, joint encoders, and an estimated contact set into a body state estimate through a propagate-correct loop. Implicit estimation maps recent observation-action history to a task-relevant latent. Deployed systems combine the two routes}
    \label{fig:ch07:state-estimation}
\end{figure}

In an explicit state-estimation architecture, the controller acts on physical quantities reconstructed from sensor measurements. Figure~\ref{fig:ch07:state-estimation} contrasts explicit estimation with implicit estimation from observation history. On a legged robot the fused body state includes the base pose and velocity and the joint states, while the estimation pipeline additionally determines the contact set and, on perceptive platforms, maps the local terrain. Joint encoders observe joint configuration, with velocity obtained through sensor processing. An IMU provides angular velocity and specific force. Neither directly measures base position or translational velocity. The estimator must account for sensor orientation, lever arms, timestamps, and coordinate transforms.

A torso-mounted IMU, taken to sit at the base origin, provides two high-rate signals. The gyroscope measures angular velocity up to a slowly drifting bias. Integrating it gives short-term attitude that accumulates drift. The accelerometer measures \emph{specific force}, physical acceleration minus gravity expressed in the sensor frame. At rest it reads not zero but a vector opposite gravity. Aligning the predicted gravity direction with the measured one corrects roll and pitch when other accelerations are modeled or sufficiently small, while yaw requires an external reference such as vision or a magnetometer. Complementary fusion trusts gyro integration at high frequency and the gravity reference at low frequency, implemented on the HRP-4 humanoid as an extended Kalman filter (EKF)~\citep{caron-estimation}. The EKF loop propagates the state and its covariance with the model and then corrects both with each measurement. On the rotation manifold, the error-state Kalman filter (ESKF) propagates a nominal orientation and estimates a small rotation-vector error~\citep{sola2017}. The contact-aided invariant extended Kalman filter (InEKF) has log-linear, estimate-independent error dynamics under its idealized deterministic model and improves convergence and consistency in the reported Cassie comparisons. IMU-bias augmentation weakens some theoretical guarantees, and covariance accuracy still depends on the model and noise assumptions~\citep{hartley2020}.

The contact set is itself an estimate. Simple detectors threshold contact-force estimates obtained from joint-torque measurements using the leg Jacobian and a dynamics model. Classification is particularly sensitive during touchdown and lift-off transients. Probabilistic estimators maintain a per-foot contact probability from proprioceptive signals, with explicit impact detection, on HyQ~\citep{camurri2017}, and filter-based estimators fold contact uncertainty into the measurement update~\citep{bloesch2012}. A false stance introduces an incorrect constraint, while a false swing discards a useful update. With the contact set decided, a foot in firm stance on stationary ground provides a velocity constraint. Its world-frame velocity is zero, so the base velocity follows from the gyroscope, the encoders, and the leg Jacobian. This inference is the legged analog of wheel odometry. Bloesch et al.\ carry the foothold positions and IMU biases in the filter state and fuse these measurements with IMU propagation in one EKF~\citep{bloesch2012}. Leg odometry is only as good as its premise. Slipping feet and misclassified swings bias the estimate. During flight phases, the estimator propagates inertially without contact corrections, and uncertainty grows according to sensor quality and the assumed dynamics.

\begin{figure}[tb]
    \centering
    \includegraphics[width=\linewidth]{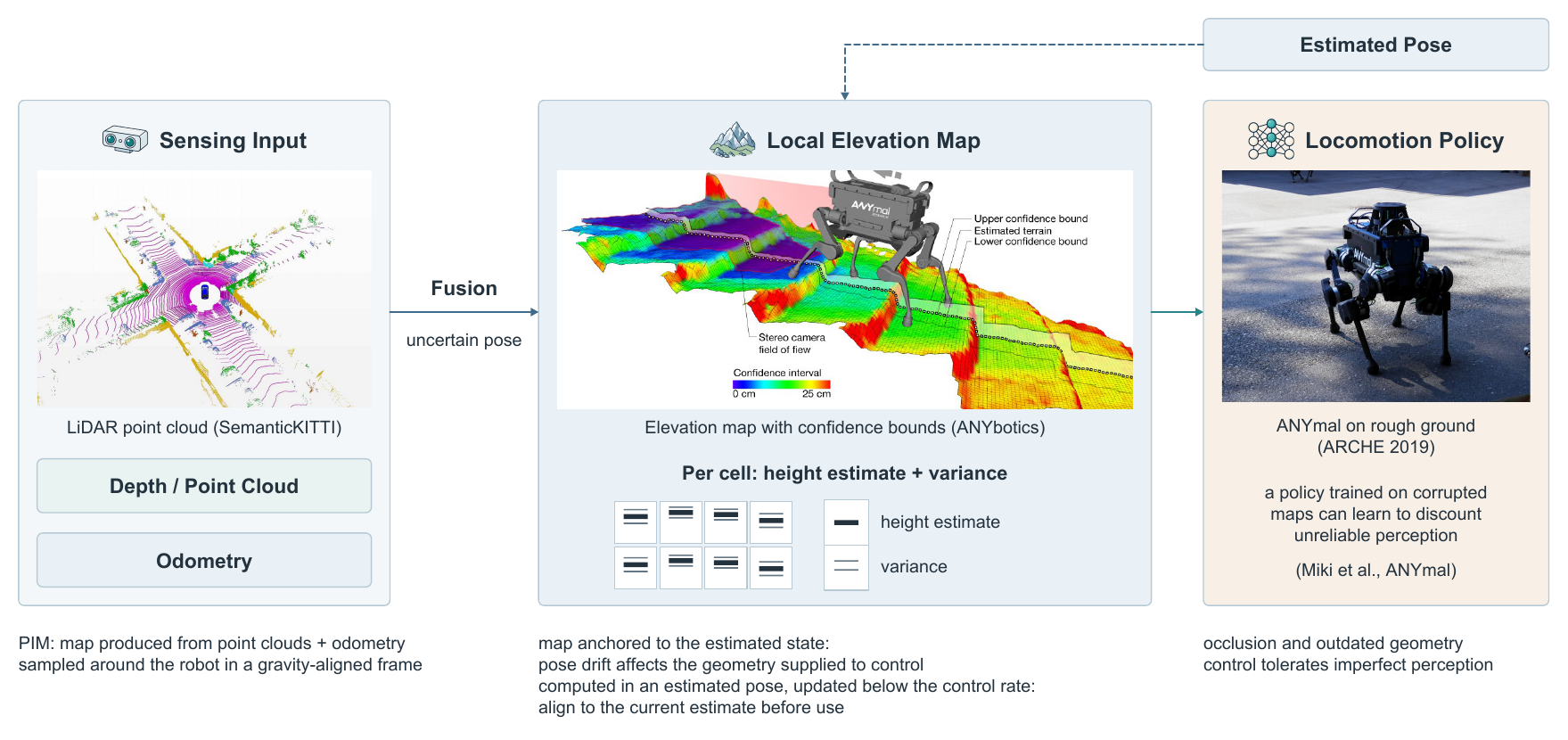}
    \caption{Local terrain estimation for perceptive locomotion. Depth measurements are fused, under an uncertain estimated pose, into a local elevation map in which each cell retains a height estimate and a variance. The map is anchored to the estimated state and updated below the control rate, so it must be aligned to the current estimate before use}
    \label{fig:ch07:terrain-estimation}
\end{figure}

Even perfect fusion leaves global translation and rotation about gravity unobservable when only inertial and relative kinematic measurements are available~\citep{bloesch2012,hartley2020}. Anchor schemes reduce accumulated drift by assuming that a selected stance-foot point occupies its planned world location and combining this with orientation and encoder-based forward kinematics. In Caron's reported HRP-4 \emph{stepping-in-place example}, the maximum lateral base-position error relative to VICON is $0.8\,\mathrm{cm}$. The anchor assumption can fail with sliding, misplaced footholds, or foot flexibility~\citep{caron-estimation}.

In addition to the body state, locomotion over unstructured ground needs a local terrain model, commonly an elevation map that fuses depth measurements under an uncertain robot pose and retains a height estimate and variance for each cell, as illustrated in Figure~\ref{fig:ch07:terrain-estimation}~\citep{fankhauser2018}. Because the map is anchored to the estimated state, pose drift affects the geometry supplied to control. The perceptive internal model (PIM), for example, uses an elevation map produced from point clouds and odometry and sampled around the robot in a gravity-aligned frame~\citep{long2025perceptive}. Miki et al.\ show on ANYmal that a policy trained over corrupted maps can learn to discount unreliable perception~\citep{miki2022}. Maps are computed in an estimated pose and updated at lower rates than the control loop, so they must be aligned to the current estimate before use. Occlusion and outdated geometry further motivate control that tolerates imperfect perception.

\subsubsection{Implicit Estimation from Observation History}
\label{subsec:implicit-estimation}

Explicit estimators rely on nominal dynamics, measurement models, and uncertainty assumptions. Real deployments introduce effects outside these models. Scree is slipperier than the simulated friction coefficient assumes, a carried payload changes the inertial parameters, actuators wear, and encoders are miscalibrated. Classical identification estimates such parameters offline from designed excitation, while a deployed robot can also infer its current response from recent observation-action history $h_t$. In \emph{implicit estimation}, a learned encoder recovers a compact, task-relevant representation $\mathbf{z}_t$ without naming every hidden physical parameter. The history must be temporally aligned and sufficiently informative. Actuator saturation, for example, can make the requested action differ from the input actually experienced by the body.

Training-time randomization varies the physics (masses, friction, and actuation) so that deployment conditions appear as further variations~\citep{tobin2017domain,peng2018}. Randomization alone does not require a dedicated estimate of the current dynamics, although a history-conditioned policy may learn adaptation. A complementary line trains with a targeted adversary in which a learnable disturber selects destabilizing pushes, with an H-infinity-style constraint relating control cost to disturbance intensity~\citep{long2025hinf}. Privileged teacher-student distillation explicitly trains a student to reproduce a privileged teacher's behavior from proprioceptive history~\citep{lee2020}. The history thereby supplies information about hidden conditions that the teacher observes directly.

RMA makes the estimator a dedicated module~\citep{kumar2021}. A base policy acts on the current observation and a latent vector, and an adaptation module produces the latent from 50 recent observation-action steps, corresponding to $0.5$\,s in the reported implementation. During training the latent is supervised to match a privileged encoder. At deployment the module updates it with no gradient computation. The latent encodes how behavior should change without requiring accurate recovery of friction or payload parameters. The hybrid internal model (HIM) combines an explicitly estimated velocity with an implicit response representation trained against successor observations~\citep{long2024hybrid}. Its humanoid extension, the PIM, adds terrain sampled from the robot-centered elevation map described above~\citep{long2025perceptive}. Learned estimation can therefore expose physical state, latent response information, or both. The pattern spans platforms. A history-conditioned humanoid transformer combines sequence-model pretraining with reinforcement learning~\citep{radosavovic2024}. Deep whole-body control unifies manipulation and locomotion on a quadruped with an arm through regularized online adaptation~\citep{fu2022}. And Deep Tracking Control (DTC) pairs a trajectory-optimizing planner with a learned tracking controller~\citep{jenelten2024}.

Generalized momentum, the product of the mass matrix and velocity, provides a dynamics residual through a momentum observer~\citep{deluca2003}. This residual can indicate unmodeled interaction, but also contains torque or model mismatch, so it cannot uniquely separate a push from a payload change. It complements learned response representations with a physics-derived diagnostic.

Several limits shape the division of labor. Distinct physical conditions can produce indistinguishable histories, and conditions outside the training distribution can invalidate a learned latent. A conventional latent vector carries no calibrated confidence measure. An EKF covariance also remains conditional on its model and noise assumptions. History can recover local motion and dynamic response, but inertial and relative kinematic inputs cannot supply a missing absolute position or yaw reference. Implicit representations cover model-resistant effects such as friction variation, actuator response, and soft-ground compliance, while explicit tools provide named physical estimates and external references. Deployed systems combine the two~\citep{miki2022,jenelten2024}. A learned policy can consume the latent directly, while a model-based controller uses physical estimates, parameter corrections, or a momentum residual.

Table~\ref{tab:ch07:state-estimation-interfaces} summarizes sensor inputs, estimated quantities, controller uses, and characteristic failure modes. The control interface specifies the components supplying $\hat{s}_t$, $h_t$, and $\mathbf{z}_t$, together with their frames and timestamps. Persistent estimation or tracking errors may require a slower command, a different contact plan, or upstream replanning, connecting local estimation to the task-level feedback in Section~\ref{sec:cerebellum-overview}.

\begin{table}[tbp]
\centering
\small
\setlength{\tabcolsep}{3pt}
\renewcommand{\arraystretch}{1.12}
\caption{Representative sensor-estimator-controller interfaces and their characteristic failure modes}
\label{tab:ch07:state-estimation-interfaces}
\begin{tabularx}{\linewidth}{@{}
>{\raggedright\arraybackslash}X
>{\raggedright\arraybackslash}X
>{\raggedright\arraybackslash}X
>{\raggedright\arraybackslash}X
>{\raggedright\arraybackslash}X@{}}
\toprule
\textbf{Sensor / input} & \textbf{Estimator} & \textbf{Output} &
\textbf{Controller / use} & \textbf{Failure / limit} \\
\midrule
IMU, encoders, contact cues &
Contact-aided EKF / InEKF &
Base pose and velocity, footholds, biases, covariance &
Legged feedback or trajectory tracking &
Slip and contact errors, global position and yaw drift \\
\addlinespace
Orientation, encoders, planned foothold &
Anchor point and forward kinematics &
Base translation and CoM &
HRP-4 walking stabilization &
Incorrect foothold assumption or foot flexibility \\
\addlinespace
Point clouds, odometry, gravity alignment &
Local elevation mapping &
Terrain heights and available uncertainty &
Footstep planning or perceptive policy &
Pose drift, occlusion, outdated geometry \\
\addlinespace
Proprioceptive and action history &
RMA adaptation module &
Task-relevant latent $\mathbf{z}_t$ &
Adaptive joint-target policy &
Ambiguous history, conditions outside training \\
\addlinespace
History, terrain observations in PIM &
HIM / PIM hybrid estimator &
Velocity estimate and response latent &
Command-conditioned locomotion &
Unseen dynamics, erroneous terrain input \\
\bottomrule
\end{tabularx}
\end{table}

\subsection{Command-Conditioned Motor Control}
\label{sec:motion-control}

A motion controller converts an abstract request such as ``walk forward at a given speed'' into a time sequence of coordinated joint actions. For a humanoid, this conversion is difficult because the request is multi-dimensional, the body is underactuated during foot contact, and the same joint motion simultaneously affects translation, turning, balance, foot clearance, and upper-body posture. A useful controller must therefore expose commands at a semantic level while learning (or solving) the coordination needed at the joint level. A velocity command leaves joint configurations and foot placement largely unspecified, while posture and phase commands specify body configuration and support timing. Reference-conditioned tracking, developed in Section~\ref{sec:whole-body-control}, supplies more detailed time-varying motion targets.

\subsubsection{Command Interfaces and Controller Families}

\begin{figure}[tbp]
  \centering
  \includegraphics[width=\linewidth]{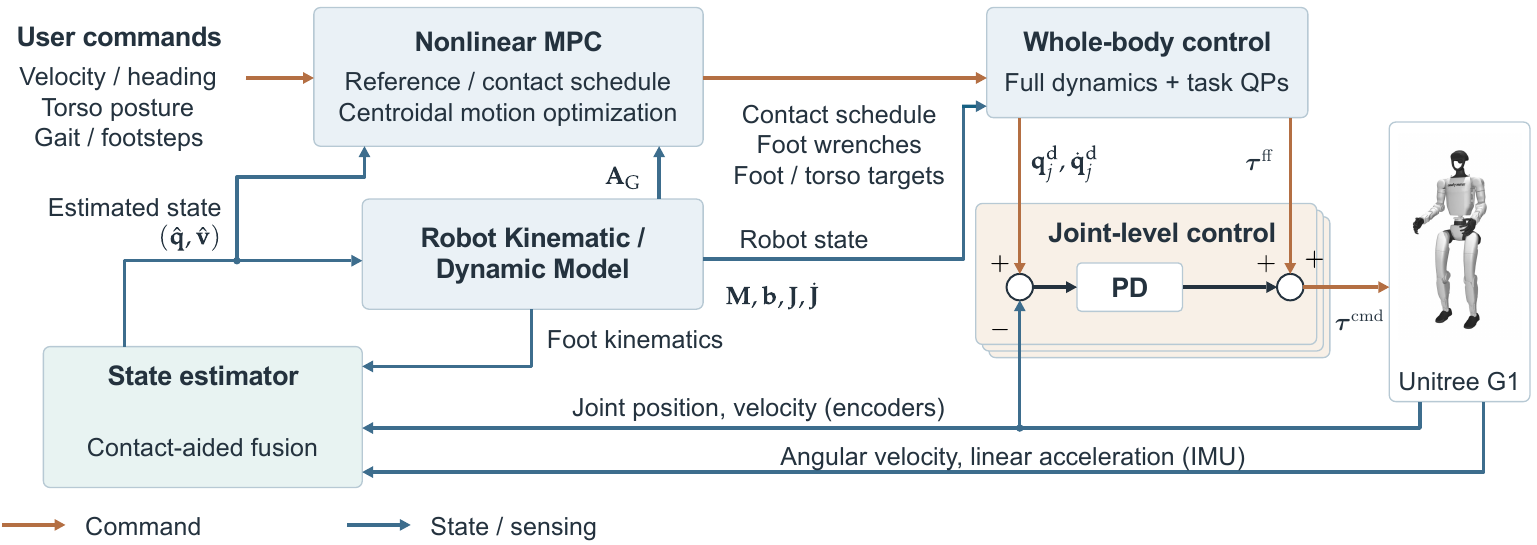}
  \caption{Model-based humanoid motion control, illustrated with Unitree G1. Command paths connect reference generation, NMPC, WBC, and joint control. Feedback paths supply estimated state, robot-model quantities, and local joint measurements}
  \label{fig:humanoid-nmpc-wbc}
\end{figure}

\paragraph{Framework and Modules}
A model-based controller separates prediction over a future horizon from instantaneous enforcement of whole-body dynamics. A reference generator converts velocity, posture, and gait commands into trajectories and a contact schedule. Nonlinear model predictive control (NMPC) plans the motion and support wrenches. And a faster whole-body controller (WBC) realizes them through joint commands. State estimation combines IMU and encoder measurements with contact kinematics, while a robot-model module supplies the dynamics and Jacobians. Figure~\ref{fig:humanoid-nmpc-wbc} adapts this organization to a humanoid robot.

\paragraph{NMPC: Anticipating Bipedal Support}
Use the configuration $\mathbf q$ and generalized velocity $\mathbf v$ of Section~\ref{subsec:motion-foundations}, with $n$ actuated joints. A centroidal formulation~\citep{LocManMPC2021RAL} uses the state $\mathbf x=(\mathbf h_{\mathrm{G}},\mathbf q)$ and input $\mathbf u=(\mathbf w_{\mathrm{L}},\mathbf w_{\mathrm{R}},\dot{\mathbf q}_j)$, where $\mathbf h_{\mathrm{G}}=(\mathbf p_{\mathrm{G}},\mathbf k_{\mathrm{G}})$ contains linear and angular momentum about the center of mass, and $\mathbf w_i=(\mathbf f_i,\boldsymbol\mu_i)$ is a six-dimensional foot wrench. Here $\mathbf p_{\mathrm{G}}$ is linear momentum in $\mathrm{kg}\,\mathrm{m}/\mathrm{s}$, $\mathbf k_{\mathrm{G}}$ is angular momentum in $\mathrm{kg}\,\mathrm{m}^{2}/\mathrm{s}$, and $m_{\mathrm{robot}}$ in the centroidal equations is robot mass in $\mathrm{kg}$. Here momentum is unnormalized and each foot-wrench moment is taken about its foot-frame origin. With all terms expressed in world coordinates,
\begin{equation}
  \dot{\mathbf p}_{\mathrm{G}}=m_{\mathrm{robot}}\mathbf g_{\mathrm{acc}}+\sum_{i\in\{\mathrm{L},\mathrm{R}\}}\mathbf f_i,
  \qquad
  \dot{\mathbf k}_{\mathrm{G}}=\sum_{i\in\{\mathrm{L},\mathrm{R}\}}
  \bigl[(\mathbf r_i-\mathbf c_{\mathrm{G}})\times\mathbf f_i+\boldsymbol\mu_i\bigr],
  \label{eq:humanoid-centroidal-mpc}
\end{equation}
where $\mathbf c_{\mathrm{G}}$ is the CoM position and $\mathbf r_i$ the foot origin. External loads are omitted here. The centroidal momentum matrix $\mathbf A_{\mathrm{G}}(\mathbf q)$ relates momentum to velocity by $\mathbf h_{\mathrm{G}}=\mathbf A_{\mathrm{G}}\mathbf v$. Together with joint rates and configuration kinematics, this determines the state flow $\dot{\mathbf x}=F_{\sigma_{\mathrm{contact}}}(\mathbf x,\mathbf u)$ for a prescribed contact schedule $\sigma_{\mathrm{contact}}$.

At each update, NMPC minimizes
\begin{equation}
\begin{aligned}
 \min_{\mathbf u(\cdot)}\quad &
 \Phi_{\mathrm{term}}(\mathbf x(t+T))+\int_t^{t+T}\ell(\mathbf x(\tau),\mathbf u(\tau),\tau)\,\mathrm{d}\tau,\\
 \ell={}&\|\mathbf v_{\mathrm{G},\mathrm{xy}}-\mathbf v^{\rm cmd}_{\mathrm{xy}}\|_{\mathbf{Q}_v}^{2}
 +\|\omega_{\mathrm{B},\mathrm{z}}-\omega_{\mathrm{z}}^{\rm cmd}\|_{Q_\omega}^{2}
 +\|\mathbf e_{\rm torso}\|_{\mathbf{Q}_{\mathrm{R}}}^{2}\\
 &+\|c_{\mathrm{G},\mathrm{z}}-c_{\mathrm{G},\mathrm{z}}^{\rm ref}\|_{Q_c}^{2}
 +\|\mathbf q_{\rm upper}-\mathbf q_{\rm upper}^{\rm ref}\|_{\mathbf{Q}_a}^{2}
 +\|\mathbf u-\mathbf u^{\rm ref}\|_{\mathbf{R}_{\mathrm{u}}}^{2},
\end{aligned}
\label{eq:humanoid-nmpc-objective}
\end{equation}
subject to the estimated initial state, centroidal dynamics, joint limits, and mode-dependent contact constraints. Here $\|\mathbf e\|_{\mathbf{Q}}^2=\mathbf e^\top \mathbf{Q}\mathbf e$, $\Phi_{\mathrm{term}}$ is a terminal tracking cost, $\mathbf v_{\mathrm{G}}=\mathbf p_{\mathrm{G}}/m_{\mathrm{robot}}$, and $\mathbf e_{\rm torso}$ is a torso-orientation error. The nonnegative scalars $Q_\omega$ and $Q_c$ weight the angular-velocity and CoM-height errors, respectively. Their weighted squares are $\|e\|_{Q_\omega}^2=Q_\omega e^2$ and $\|e\|_{Q_c}^2=Q_c e^2$ for a scalar error $e$. Velocity commands are transformed into the same frames as the predicted velocities before comparison. The upper-body target specifies waist and arm posture. Desired support wrenches and joint rates define $\mathbf u^{\rm ref}$. The humanoid objective therefore couples commanded travel to upright posture, CoM height, and upper-body motion.

A stance sole remains stationary. A swing foot carries zero wrench and follows a clearance trajectory. Each supporting foot must satisfy unilateral contact, friction, center-of-pressure bounds inside its sole, and admissible yaw moment. These surface-wrench constraints are essential when transferring load between one and two supporting feet. A prescribed schedule fixes contact order and timing. Optimizing those discrete choices requires an additional planner. Multiple shooting and sequential quadratic programming provide one implementation route. Only the current portion of the optimized trajectory is passed to WBC before replanning.

\paragraph{WBC: Realizing the Plan on the Full Body}
Between contact switches, WBC solves at the current estimated configuration for $\mathbf z_{\mathrm{WBC}}=(\dot{\mathbf v},\mathbf w_{\mathrm{L}},\mathbf w_{\mathrm{R}},\boldsymbol\tau)$. A representative task objective is
\begin{equation}
 \min_{\mathbf z_{\mathrm{WBC}}}\quad
 \sum_{k\in\mathcal T_{\mathrm{WBC}}}\|\mathbf J_k\dot{\mathbf v}
       +\dot{\mathbf J}_k\mathbf v-\mathbf a_k^{\rm des}\|_{\mathbf W_k}^{2}
 +\|\mathbf w-\mathbf w^{\star}_{\rm NMPC}\|_{\mathbf W_{\mathrm f}}^{2}
 +\|\boldsymbol\tau\|_{\mathbf W_\tau}^{2},
 \label{eq:humanoid-wbc-objective}
\end{equation}
where $\mathbf W_k\in\mathbb R^{d_{\mathrm{task},k}\times d_{\mathrm{task},k}}$, $\mathbf W_{\mathrm f}\in\mathbb R^{12\times12}$, and $\mathbf W_\tau\in\mathbb R^{n\times n}$ are symmetric positive-semidefinite weight matrices, with $\|\mathbf e\|_{\mathbf W}^2=\mathbf e^\top\mathbf W\mathbf e$. The positive integer $d_{\mathrm{task},k}$ is the dimension of task $k$'s acceleration residual. The vector $\mathbf w=(\mathbf w_{\mathrm L},\mathbf w_{\mathrm R})\in\mathbb R^{12}$ stacks the two six-dimensional foot wrenches, and $\boldsymbol\tau\in\mathbb R^n$ contains the torques of the $n$ actuated joints. The optimization is subject to the hard constraints
\begin{equation}
\begin{aligned}
 \mathbf M\dot{\mathbf v}+\mathbf b
   &=\mathbf S\boldsymbol\tau+\sum_i\mathbf J_{w,i}^{\top}\mathbf w_i,\\
 \mathbf J_{w,i}\dot{\mathbf v}+\dot{\mathbf J}_{w,i}\mathbf v&=\mathbf0
       &&(i\text{ in stance}),\\
 \mathbf w_i&\in\mathcal W_{i}^{\mathrm{wrench}}(\sigma_{\mathrm{contact}}),\qquad
 \boldsymbol\tau_{\min}\leq\boldsymbol\tau\leq\boldsymbol\tau_{\max}.
\end{aligned}
\label{eq:humanoid-wbc-constraints}
\end{equation}
The six-row Jacobian $\mathbf J_{w,i}$ maps generalized velocity to foot twist, extending the point-contact model of Section~\ref{subsec:motion-foundations}. $\mathcal W_{i}^{\mathrm{wrench}}$ is the admissible wrench set, reduced to $\{\mathbf0\}$ in swing. The torque bounds in Eq.~\eqref{eq:humanoid-wbc-constraints} apply componentwise. Tasks $\mathcal T_{\mathrm{WBC}}$ track CoM motion, torso orientation, swing-foot pose, and waist/arm posture, with feedback embedded in $\mathbf a_k^{\rm des}$. Linearized friction bounds give a quadratic program. Weighted tasks express trade-offs. Strict priorities instead require hierarchical quadratic programs (QPs) that preserve higher-priority optima~\citep{2016RasHermiteSplines}. Joint position and velocity limits can be enforced through bounds on predicted motion over the control step.

The resulting torque is combined with local joint feedback,
\begin{equation}
 \boldsymbol\tau^{\mathrm{cmd}}=\boldsymbol\tau^{\mathrm{ff}}
 +\mathbf K_{\mathrm{p}}(\mathbf q_j^{\mathrm{d}}-\mathbf q_j)
 +\mathbf K_{\mathrm{d}}(\dot{\mathbf q}_j^{\mathrm{d}}-\dot{\mathbf q}_j).
\end{equation}
Desired joint motion is obtained from the whole-body tracking solution, with the feedforward and feedback terms sharing the actuator torque budget. Predictive planning anticipates support changes. WBC enforces instantaneous feasibility. Their performance remains dependent on state estimates, contact assumptions, and timely optimization. A learned controller moves much of this coordination into training, while retaining a comparable command interface.

\subsubsection{A Unified Command Space}

The HUGWBC controller provides a compact example~\citep{xue2025hugwbc}. Its policy receives task commands, behavior commands, proprioceptive history, and an intervention indicator, and emits the whole-body action at 50~Hz. Its design brings command following, disturbance rejection, and motion quality into a common training framework. In the released implementation, the task command contains desired base-frame longitudinal and lateral linear velocity and the base-frame angular-velocity $z$ component, which need not equal the Euler yaw-angle derivative,
\begin{equation}
 \mathbf v_t^{\rm cmd}
 = [v_{\mathrm{x},t}^{\rm cmd},v_{\mathrm{y},t}^{\rm cmd},\omega_{\mathrm{z},t}^{\rm cmd}]^\top.
\end{equation}
This representation allows the policy to select footholds and body motion while being judged by the achieved velocity. The command is augmented with behavior variables including gait frequency $f_{t}^{\mathrm{gait}}$, maximum foot clearance $\ell_{t}^{\mathrm{clear}}$, body-height command $z_t^{\rm cmd}$, body pitch $\beta_{t}^{\mathrm{pitch}}$, waist yaw $\chi_t$, phase offset $\psi_{t}^{\mathrm{phase}}$, leg phases $\phi_{i,t}$, and stance duty factor $d_{t}^{\mathrm{stance}}$, forming
\begin{equation}
 \mathbf c_t^{\rm beh}=[f_{t}^{\mathrm{gait}},\ell_{t}^{\mathrm{clear}},z_t^{\rm cmd},\beta_{t}^{\mathrm{pitch}},\chi_t,
              \psi_{t}^{\mathrm{phase}},\phi_{1,t},\phi_{2,t},d_{t}^{\mathrm{stance}}]^\top.
\end{equation}
The clearance command is measured relative to a contact-height reference, and the duty factor specifies the fraction of a periodic cycle allocated to stance. The phase variables provide a low-dimensional clock shared by the legs and the reward terms. They are internal clocks advanced from frequency and phase offset, rather than independent operator inputs. The periodic mapping below uses $0<d_{t}^{\mathrm{stance}}<1$. Standing instead fixes the phases. For a control interval $\Delta t$,
\begin{equation}
 \phi_{1,t+1}=(\phi_{1,t}+f_{t}^{\mathrm{gait}}\Delta t)\bmod1,
 \qquad
 \phi_{2,t+1}=(\phi_{1,t+1}+\psi_{t}^{\mathrm{phase}})\bmod1.
\end{equation}
A walking gait uses half-cycle phase separation, whereas a jumping gait synchronizes both legs. In the paper's nominal definitions, standing fixes both phases at $0.25$, and hopping fixes the airborne leg at $0.75$ while the other leg cycles. HUGWBC uses one policy for standing, walking, and jumping, and a separate policy for hopping. Both the reported settings and the public default configuration fix $d_{t}^{\mathrm{stance}}=0.5$.

The raw phase is normalized so that stance and swing occupy equal phase intervals according to
\begin{equation}
 \bar\phi_{i,t}=\begin{cases}
 \displaystyle\frac{\phi_{i,t}}{2d_{t}^{\mathrm{stance}}},&0\leq\phi_{i,t}<d_{t}^{\mathrm{stance}},\\[3pt]
 \displaystyle\frac12+\frac{\phi_{i,t}-d_{t}^{\mathrm{stance}}}{2(1-d_{t}^{\mathrm{stance}})},
     &d_{t}^{\mathrm{stance}}\leq\phi_{i,t}<1.
 \end{cases}.
\end{equation}
Thus stance maps to $[0,\frac12)$ and swing to $[\frac12,1)$. The actor receives an encoded command $\mathbf c_t=[\mathbf v_t^{\rm cmd};\tilde{\mathbf c}_t^{\rm beh}]$, where $\tilde{\mathbf c}_t^{\rm beh}$ replaces the two phase entries by $\sin(2\pi\bar\phi_{i,t})$. The complete command separates \emph{what} the robot should accomplish from \emph{how} the motion should look, while leaving the policy free to discover the joint-level realization.

\subsubsection{From Commands to Coordinated Actions}

The policy observes a history of joint positions and velocities, base angular velocity and gravity direction, and its previous action, using $h_t$ defined in Eq.~\eqref{eq:cerebellum-interface}. Write $g_t=(\mathbf c_t,I_{t}^{\mathrm{ext}})$ for the command input, where $I_{t}^{\mathrm{ext}}$ indicates whether an external upper-body controller is active. The deployed action mean is
\begin{equation}
 \mathbf{a}_t=\boldsymbol\mu_\theta(h_t,g_t).
\end{equation}
During training, exploration uses a stochastic policy around this mean. A history encoder and state estimator supply estimated translational velocity, body height, and foot clearance. The privileged critic additionally sees linear velocity, body-height error, foot clearance, contact forces, friction, terrain samples, and collision indicators during training. The history allows the policy to infer quantities that are not directly measured or are delayed, a practical form of implicit state estimation.

The action is interpreted as a joint target (or a residual around a nominal posture) and passed through the robot's low-level actuator loop. Consequently, a velocity command is not converted to a fixed stride table. The policy coordinates stance-leg actuation, swing-leg trajectories, trunk inclination, and arm motion to reduce the velocity error. Contact forces are controlled indirectly through body motion and joint actuation.

To describe a smooth contact target, let $\Phi$ be the standard normal cumulative distribution function and $\sigma_\phi>0$ a dimensionless smoothing width. For normalized phase $z\in[0,1)$, define
\begin{equation}
 \begin{aligned}
 C_{\mathrm{contact}}(z)={}&\Phi(z/\sigma_\phi)
       [1-\Phi((z-\tfrac12)/\sigma_\phi)]\\
       &+\Phi((z-1)/\sigma_\phi)
       [1-\Phi((z-\tfrac32)/\sigma_\phi)],
 \qquad C_{i,t}^{\mathrm{contact}}=C_{\mathrm{contact}}(\bar\phi_{i,t}).
 \end{aligned}
\end{equation}
$C_{i,t}^{\mathrm{contact}}$ is high in the intended stance interval and low in swing, with smooth transitions near phase boundaries. This command-derived expected contact weight is distinct from the sensor-based contact estimates of Section~\ref{sec:dynamics-estimation}. The smooth weighting gives the contact objective a finite transition interval between stance and swing.

\subsubsection{Contact Penalties and Swing Targets}

Command following alone can admit motions that track velocity on average while making unnecessary contacts or moving the support foot. HUGWBC adds a periodic contact-swing objective. The public implementation uses nonnegative saturating functions~\citep{hugwbcCode2026} given by
\begin{equation}
 g_f(x)=1-\exp(-x^2/s_{\mathrm{F}}),\qquad
 g_v(x)=1-\exp(-x^2/s_{\mathrm{V}}),\qquad s_{\mathrm{F}},s_{\mathrm{V}}>0.
\end{equation}
The denominators have units compatible with squared force and squared speed. With positive effective weights $w_{\mathrm{F}},w_{\mathrm{V}}$ that absorb control-interval scaling by $\Delta t$, the combined penalty can be written as
\begin{equation}
 r_t^{\rm contact}=-\frac12\sum_{i=1}^2
 \left[
 w_{\mathrm{F}}(1-C_{i,t}^{\mathrm{contact}})g_f(\|\mathbf f_{i,t}^{\rm foot}\|_2)
 +w_{\mathrm{V}}C_{i,t}^{\mathrm{contact}}g_v(\|\mathbf v_{i,t}^{\rm foot}\|_2)
 \right],
\end{equation}
where $\mathbf f_{i,t}^{\rm foot}$ is the three-dimensional contact force and $\mathbf v_{i,t}^{\rm foot}$ the three-dimensional world-frame foot velocity. This follows the released implementation. The paper expresses its velocity term using horizontal components. The code defaults are $s_{\mathrm{F}}=50$, $s_{\mathrm{V}}=5$, and reward weights $2$ and $4$, respectively, before control-interval scaling. During stance the penalty encourages a planted support foot. During swing it discourages premature contact. The phase-dependent weight smooths the transition between these soft objectives.

In the paper, the vertical swing target is generated with a piecewise quintic polynomial whose coefficients satisfy position, velocity, and acceleration boundary conditions. A convenient mathematical expression uses
\begin{equation}
 q_{\mathrm{interp}}(u_{\mathrm{interp}})=10u_{\mathrm{interp}}^{3}-15u_{\mathrm{interp}}^{4}+6u_{\mathrm{interp}}^{5},\qquad 0\leq u_{\mathrm{interp}}\leq1,
\end{equation}
which satisfies $q_{\mathrm{interp}}(0)=0$, $q_{\mathrm{interp}}(1)=1$, and zero first and second derivatives at both endpoints. For equal starting and ending contact heights, the relative clearance is
\begin{equation}
 \ell_{i,t}^{\rm target}=\ell_{t}^{\mathrm{clear}}\begin{cases}
 0,&0\leq\bar\phi_{i,t}\leq\frac12,\\
 q_{\mathrm{interp}}(4\bar\phi_{i,t}-2),&\frac12<\bar\phi_{i,t}\leq\frac34,\\
 q_{\mathrm{interp}}(4-4\bar\phi_{i,t}),&\frac34<\bar\phi_{i,t}<1.
 \end{cases}.
\end{equation}
The target is zero in stance, reaches the commanded height $\ell_{t}^{\mathrm{clear}}$ in mid-swing, and returns to the contact reference before touchdown. With fixed clearance and duty commands and constant phase rate during a cycle, height, velocity, and acceleration match at lift-off, the apex, and touchdown. For different endpoint heights, each half is written as $p_0+(p_1-p_0)q_{\mathrm{interp}}(u_{\mathrm{interp}})$ using its time-ordered endpoints and a local coordinate $u_{\mathrm{interp}}$ increasing from zero to one.

The paper's folded variable $u_{\rm fold}=\frac14-|\bar\phi_{i,t}-\frac34|$ gives the equivalent swing expression $\ell_{t}^{\mathrm{clear}}\sum_{k=0}^5a_k^{\mathrm{swing}}u_{\rm fold}^k=\ell_{t}^{\mathrm{clear}} q_{\mathrm{interp}}(4u_{\rm fold})$, with $(a_0^{\mathrm{swing}},\ldots,a_5^{\mathrm{swing}})=(0,0,0,640,-3840,6144)$. This form makes the boundary conditions explicit. A swing penalty is
\begin{equation}
 r_t^{\rm swing}=-w_\ell\sum_{i=1}^2(1-C_{i,t}^{\mathrm{contact}})
      (\ell_{i,t}^{\rm target}-\ell_{i,t}^{\rm foot})^2,
 \qquad w_\ell>0,
\end{equation}
where $\ell_{i,t}^{\rm foot}$ is actual clearance in the same convention. The polynomial supplies a smooth vertical reference, while the policy learns the coordinated foothold and body motion. The released implementation includes both linear and polynomial clearance rewards, with the default configuration enabling the linear term and assigning zero weight to the polynomial term~\citep{hugwbcCode2026}.

\subsubsection{Environment Rewards and Mirror Regularization}

The environment reward contains task tracking, behavior tracking, and regularization terms, combined as
\begin{equation}
 r_t=r_t^{\rm task}+r_t^{\rm behavior}+r_t^{\rm reg}.
\end{equation}
Task terms track the three velocity components with dense exponential errors. Behavior terms track height, pitch, and waist yaw, and include the contact and swing terms above. Regularization penalizes torque, joint-speed limit violations, joint acceleration, and action increments. Collision-based termination discourages failed execution. In particular, an action-rate term $-\|\mathbf{a}_t-\mathbf{a}_{t-1}\|^2$ discourages frame-to-frame jitter. These rewards define the rollout return and advantages used by PPO.

Mirror consistency is a separate training loss. Let $\mathcal F_h$ mirror body observations and past actions at each history step, $\mathcal F_g$ mirror the associated commands, and $\mathcal F_a$ mirror actions across the robot's sagittal plane. The intervention indicator is unchanged. For rollout minibatches $\mathcal D_{\rm roll}$, the released implementation uses
\begin{equation}
 \mathcal L_{\rm sym}=\lambda_{\rm sym}
 \mathbb E_{(h_t,g_t)\sim\mathcal D_{\rm roll}}
 \left[\frac1{n_a}\left\|
 \operatorname{sg}[\boldsymbol\mu_\theta(h_t,g_t)]
 -\mathcal F_a\!\left(
  \boldsymbol\mu_\theta(\mathcal F_h(h_t),\mathcal F_g(g_t))\right)
 \right\|_2^2\right],
\end{equation}
where $n_a$ is the action dimension and $\operatorname{sg}$ denotes stop-gradient. The default $\lambda_{\rm sym}=0.5$. Symmetry is applied to observations, past actions, and commands as well as the action output, so it remains useful when the same policy is asked to produce left-right variants of an otherwise identical behavior~\citep{symmetry2018tog}.

The optimizer minimizes the PPO actor and value losses minus the entropy bonus, plus $\mathcal L_{\rm sym}$ and the state-estimation loss when jointly trained. The mirror loss does not enter each-step reward, discounted return, or advantage. An independent foot-position symmetry reward used for non-walking gaits can enter the environment return. It is distinct from policy mirror consistency~\citep{hugwbcCode2026}.

\subsubsection{Couplings Between Commands}

The learned controller exposes several useful couplings. At approximately steady forward speed, the pair $(v_{\mathrm{x}},f^{\mathrm{gait}})$ controls displacement per cycle. At fixed speed, raising frequency shortens each cycle's displacement. At fixed frequency, raising speed requires greater displacement. The pair $(f^{\mathrm{gait}},\ell^{\mathrm{clear}})$ trades clearance against swing speed and work. Lateral velocity and turning interact with foot placement, phase offset, and waist orientation. Body height $z^{\rm cmd}$ changes leg configuration and effective workspace, while pitch $\beta^{\mathrm{pitch}}$ changes body geometry and the relation between base-frame and world-frame velocity. Finally, phase offset $\psi^{\mathrm{phase}}$ and duty factor $d^{\mathrm{stance}}$ specify the intended contact rhythm. A unified policy can learn these trade-offs from the return, while coordinated actuation and contact still limit the command combinations it can realize.

The reward weights also interact with these commands. For example, a high-frequency, high-clearance target increases the acceleration required by the swing trajectory, so an action-rate penalty suitable for walking may over-damp hopping. Curriculum sampling gradually expands command ranges and intervention magnitude. Massive parallel simulation makes such coverage practical~\citep{rudin2022learning}. Periodic reward composition and multiplicity of behavior provide related examples of diverse gait learning through explicit command spaces~\citep{2021BipedalPeriodicReward,margolis2022walktheseways}.

A useful diagnostic is therefore a command-interaction map rather than a list of single-command errors. Evaluate tracking error and smoothness on a grid of $(v_{\mathrm{x}},v_{\mathrm{y}},\omega_{\mathrm{z}},f^{\mathrm{gait}},\ell^{\mathrm{clear}},z^{\rm cmd},\beta^{\mathrm{pitch}},\chi)$, then inspect where one command degrades another. Such tests assess behavior between command combinations as well as at isolated settings. They also expose saturation. If increasing $f^{\mathrm{gait}}$ no longer changes measured cadence, actuator or contact limits may be responsible, but further tests are needed to distinguish these limits from inadequate command conditioning.

\subsubsection{Disturbance Rejection and External Control}

Robustness is trained as part of motion generation. Upper-body intervention replaces the policy's arm actions with sampled joint targets or bounded noise, while the lower-body command remains active. The intervention indicator tells the policy which control mode is present, and the intervention is interpolated over a finite interval rather than switched instantaneously. Regularization rewards for the intervened joints are masked, while velocity tracking, contact objectives, collision-based termination, and lower-body regularization remain active. A curriculum gradually increases the contribution of external actions once baseline velocity tracking is reliable. This trains trunk and leg compensation for arm motion while preserving the commanded gait, supporting loco-manipulation.

The resulting architecture closes the feedback loop at each control step. A semantic command is encoded as velocity, posture, and phase variables. Temporal proprioception supports state estimation. A command-conditioned policy maps the combined input to joint targets. And contact and motion-quality objectives shape the solution. Low-level torque limits and deployment checks complement this learned control. For deployment, the most important tests jointly assess command tracking under phase changes, recovery after external impulses, action-rate and contact behavior, and retention of lower-body stability during upper-body intervention. Maintaining this coordination on hardware also depends on the delays and actuator limitations examined in Section~\ref{sec:real-time-execution}.

\subsection{Motor Pretraining and Dexterous Whole-Body Control}
\label{sec:whole-body-control}

Command-conditioned control exposes a specified behavior domain. Motor pretraining seeks capabilities that can be reused across downstream objectives. Three approaches serve different roles. Reference tracking learns to execute supplied motions. Learned priors and skill representations support downstream policy learning. And multi-mode pretraining or distillation exposes a shared repertoire through varied commands~\citep{peng2018deepmimic,peng2021amp,peng2022ase,he2025hover,zeng2025behavior}. These approaches can be combined. This section first develops tracking as a common learning objective, then examines behavior foundation models, and finally shows that the same dynamics, estimation, tracking objectives, and command interfaces carry over to the multi-fingered hand, where the held object rather than the robot's base is the unactuated body.

\subsubsection{Motion Tracking}

Motion tracking formulates humanoid whole-body control as the reproduction of a reference motion under the robot's dynamics. Reference trajectories provide a time-varying description of the desired behavior, capturing the coordinated movement of the body across different skills. DeepMimic~\citep{peng2018deepmimic} establishes a reinforcement-learning formulation of this approach in physics-based character animation. Recent humanoid systems, including BeyondMimic~\citep{liao2026beyondmimic} and SONIC~\citep{luo2026sonic}, build on this principle to acquire physically executable skills and broad behavior repertoires. The reference specifies how the robot should move, while reinforcement learning discovers the feedback control needed to reproduce that movement under physical dynamics.

Formally, consider a reference motion sequence
\begin{equation}
    \mathcal{M}_{\mathrm{motion}}=(s_0^{\rm ref},s_1^{\rm ref},\ldots,s_T^{\rm ref}),
\end{equation}
where $s_t^{\rm ref}$ specifies the target kinematic quantities at timestep $t$, such as joint configurations, body poses, and velocities. At each timestep, the policy selects an action conditioned on the available history and reference command, and the system evolves according to
\begin{equation}
    \mathbf{a}_t\sim\pi_\theta(\cdot\mid h_t,g_t),\qquad
    s_{t+1}\sim P(\cdot\mid s_t,\mathbf{a}_t),
    \label{eq:tracking_dynamics}
\end{equation}
where $\pi_\theta$ is the policy parameterized by $\theta$, $\mathbf{a}_t$ is the action, $h_t$ is the observation-action history defined in Eq.~\eqref{eq:cerebellum-interface}, $g_t$ is the reference-derived command, and $P$ denotes the transition dynamics. The full physical state $s_t$ may include privileged information accessible in simulation but unavailable to the deployed policy.

The discrepancy between the executed and reference motions provides a dense learning signal. A generic tracking reward takes the form
\begin{equation}
    r_t^{\mathrm{track}}
    =\sum_{k\in\mathcal{I}_{\mathrm{track}}}w_k
      \exp\!\left[-\frac{d_k(s_{t+1},s_{t+1}^{\rm ref})}{{\sigma_{k}^{\mathrm{track}}}^{2}}\right],
    \qquad w_k\geq0,
    \label{eq:tracking_reward}
\end{equation}
where $\mathcal{I}_{\mathrm{track}}$ indexes the tracked quantities, $d_k$ measures the corresponding squared discrepancy, $w_k$ determines its relative contribution, and $\sigma_{k}^{\mathrm{track}}>0$ controls the sensitivity to tracking errors. Additional costs can regularize physical execution through a reward of the form
\begin{equation}
    r_t=r_t^{\mathrm{track}}
       -\sum_j\lambda_j c_j(s_t,\mathbf{a}_{t-1},\mathbf{a}_t,s_{t+1}),
    \qquad \lambda_j\geq0,
\end{equation}
where $c_j$ denotes a nonnegative regularization cost and $\lambda_j$ its weight. The preceding action is included to allow action-rate penalties. Such costs can penalize undesirable behavior, including abrupt action changes or joint-limit violations. Motion tracking therefore provides a shared behavioral objective while retaining regularization appropriate for physical execution.

The policy is trained to maximize the expected discounted return according to the objective
\begin{equation}
    \max_\theta\;
    \mathbb{E}_{\mathcal{M}_{\mathrm{motion}}\sim\mathcal{D},\,
    \zeta_{\mathrm{traj}}\sim p_\theta(\zeta_{\mathrm{traj}}\mid\mathcal{M}_{\mathrm{motion}})}
    \left[\sum_{t=0}^{T-1}\gamma^t r_t\right],
    \label{eq:tracking_rl}
\end{equation}
where $\gamma\in[0,1)$ is the discount factor, $\mathcal{D}$ is the reference-motion distribution, and $\zeta_{\mathrm{traj}}$ is an executed state-action trajectory. The trajectory distribution $p_\theta$ is induced by the policy, transition dynamics, and initialization distribution. For a single-clip policy, $\mathcal{D}$ concentrates on one reference motion. For a controller trained on a motion corpus, it samples across multiple clips. In both settings, reference motions specify the desired outcomes, while interaction with the environment supplies the experience used for policy optimization. PPO~\citep{schulman2017ppo} is commonly used to optimize this objective, as exemplified by SONIC~\citep{luo2026sonic}.

Several practical strategies support effective learning. \emph{Reference state initialization} starts episodes from sampled reference configurations and velocities, exposing the policy to later phases of a motion before it can reliably reach them from the beginning. \emph{Early termination} ends episodes when predefined failure conditions are met, such as excessive tracking errors, avoiding prolonged rollouts in unsuccessful states. \emph{Adaptive sampling} further directs training effort toward challenging motions or segments. For example, BeyondMimic~\citep{liao2026beyondmimic} prioritizes temporal regions with high estimated failure rates while retaining a uniform sampling component to preserve coverage. Together, these strategies determine where rollouts begin, when they end, and how experience is distributed across the reference corpus.

\subsubsection{Behavior Foundation Model} 

Motion tracking provides a shared learning objective, but a general-purpose controller must also accommodate the diverse commands used in downstream applications. A complete reference sequence may be suitable for motion replay, whereas interactive control may specify targets for only a few body links. This motivates behavior foundation models, humanoid controllers pretrained on large-scale motion data to acquire a broad repertoire of reusable whole-body behaviors and execute them in response to diverse behavioral specifications~\citep{zeng2025behavior,zeng2026scaling}. By combining broad behavioral coverage with a versatile control interface, BFMs enable learned motor capabilities to be shared across multiple control modes and downstream tasks.

In this formulation, a behavior is defined as a trajectory of body states and actions, where $s_t^{\mathrm{p}}$ denotes the proprioceptive body variables selected from $s_t$. The trajectory is written as
\begin{equation}
    \mathbf{B}^{\mathrm{seq}}\triangleq
    (s_0^{\mathrm{p}},\mathbf{a}_0,s_1^{\mathrm{p}},\mathbf{a}_1,\ldots,s_{T-1}^{\mathrm{p}},\mathbf{a}_{T-1},s_T^{\mathrm{p}}).
\end{equation}
Goal specifications are treated as external conditions that guide or constrain the execution of a behavior. A single behavior may satisfy different specifications, while a given specification may admit multiple physically valid behaviors. For example, specifying a hand target leaves considerable freedom in the robot's stance and the movement of its other arm. Separating goals from the definition of $\mathbf{B}^{\mathrm{seq}}$ therefore distinguishes the executed behavior from the conditions used to specify it.

From this perspective, BFM pretraining requires both a shared objective for acquiring diverse behaviors and a versatile interface for specifying their execution. Motion tracking provides the former through coordinated demonstrations and dense tracking rewards. The control interface provides the latter by allowing downstream applications to specify only the aspects of movement relevant to their objectives. Motion tracking thus serves as a proxy task for learning reusable motor capabilities, which can subsequently be invoked through commands that need not describe a complete whole-body trajectory.

One practical realization of this interface uses masked Cartesian body targets. Let $\mathbf{x}_{t}^{\mathrm{target}}$ contain target poses and tracking errors expressed in the robot's current root frame, and let $\mathbf{m}_t$ be a binary mask selecting the commanded body links. The command and policy are then
\begin{equation}
    g_t=(\mathbf{m}_t\odot \mathbf{x}_{t}^{\mathrm{target}},\mathbf{m}_t),\qquad
    \mathbf{a}_t\sim\pi_\theta(\cdot\mid h_t,g_t),
    \label{eq:masked_command}
\end{equation}
where $h_t$ retains the history definition in Section~\ref{architecture-and-interfaces}, and the mask is applied to the features associated with each link. An entry $m_t^\ell=1$ indicates that link $\ell$ is specified. An all-one mask requests full-body tracking, while a sparse mask specifies selected links and allows the controller to coordinate the remaining body movements. Including the mask explicitly distinguishes an unspecified target from a valid target whose encoded value is zero. Sampling different masks during pretraining enables a single controller to learn from multiple forms of behavioral specification. For globally specified targets, root-relative encoding uses the robot pose relative to the target.

\begin{figure}[tb]
    \centering
    \includegraphics[width=\linewidth]{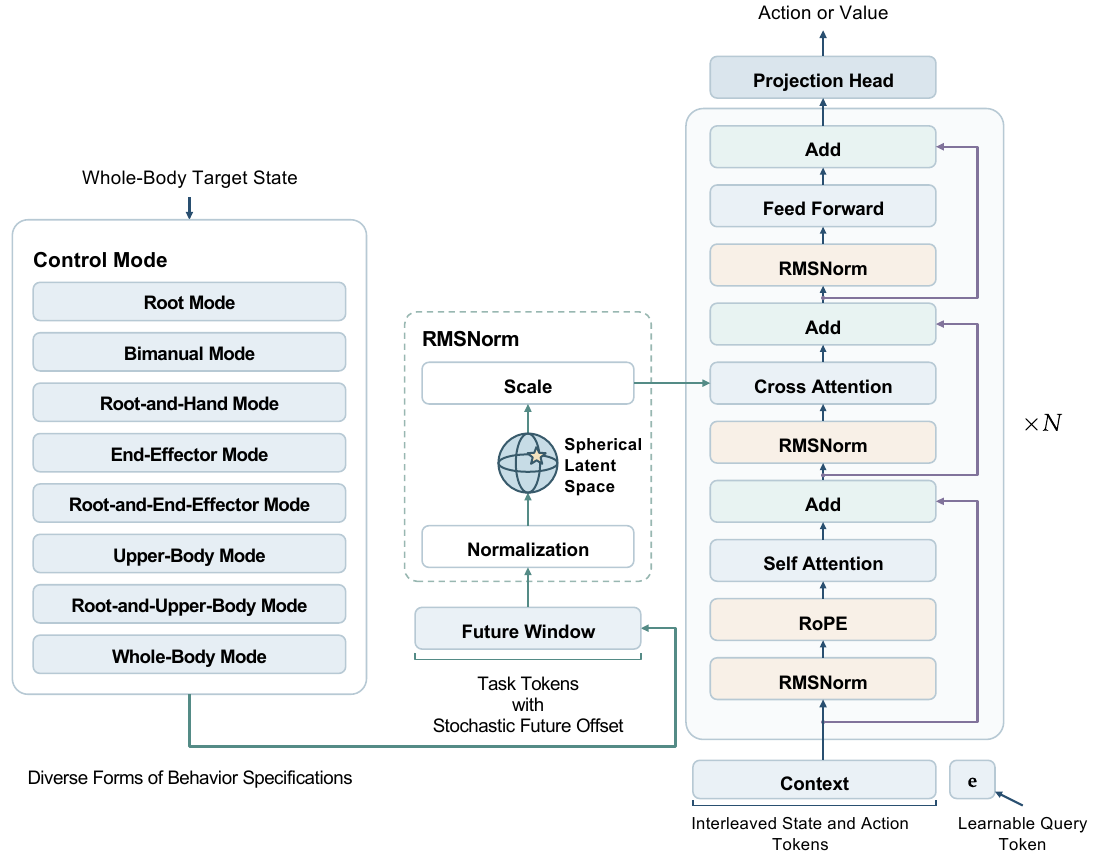}
    \caption{Visualization of the Humanoid Transformer architecture applied in ScaleBFM~\citep{zeng2026scaling}}
    \label{fig:c7bfm:humanoid_transformer}
\end{figure}

The original BFM implementation trains a motion-imitation policy and transfers its behavior through masked online distillation. It also evaluates learning new behaviors through a residual decoder with the BFM frozen~\citep{zeng2025behavior}. ScaleBFM studies direct goal-conditioned reinforcement-learning pretraining, examining the joint effects of training data and model capacity~\citep{zeng2026scaling}. Its findings highlight the complementary roles of reference-motion diversity and on-policy experience. Diverse references expand the range of behaviors available for learning, while rollouts supply the interaction experience needed to acquire them. Effective scaling therefore calls for coordinating these two resources with an architecture capable of translating diverse behavioral specifications into coordinated actions.

As its model backbone, ScaleBFM introduces the \emph{Humanoid Transformer}, which combines temporal context with goal-conditioned control, as illustrated in Figure~\ref{fig:c7bfm:humanoid_transformer}. Proprioceptive states and past actions are encoded as interleaved tokens and processed through self-attention to capture the recent execution history. A learnable query token aggregates this information for action or value prediction. Future goal states, together with their temporal offsets, are encoded into goal tokens and incorporated through cross-attention. This design allows the controller to interpret the requested movement in the context of its ongoing behavior, directly accommodating the history and commands in Eq.~\eqref{eq:masked_command}. 

ScaleBFM also studies the representations learned through goal conditioning. It applies root mean square normalization (RMSNorm)~\citep{zhang2019root} to goal embeddings and analyzes their normalized directions. For the selected motion sequences, neighboring targets yield nearby representations, while behaviors such as forward and backward walking occupy distinguishable regions. Perturbation experiments test how changes in embedding direction affect control performance. These findings provide evidence of local organization and robustness in the learned goal representation.

Together, these components illustrate how the general BFM framework can be instantiated in practice. Motion tracking supplies a shared learning objective, masked targets provide a versatile control interface, and the Humanoid Transformer relates behavioral specifications to the robot's execution history. ScaleBFM thus offers a concrete example of learning coordinated motor skills and structured goal representations within a single pretrained controller, supporting the reuse of its behavioral repertoire across multiple control modes.

\subsubsection{Evaluation Metrics}

Evaluating a BFM requires measuring how reliably and accurately it realizes the behavioral specifications supplied through its control interface. Motion tracking provides a direct setting for this assessment because the executed behavior can be compared with a time-aligned reference. Using the tracking metrics reported by ScaleBFM~\citep{zeng2026scaling}, this chapter considers sequence-level success rate together with mean per-keypoint position and rotation errors in global and local coordinates. These metrics describe complementary aspects of control, including persistent compliance with the command, accuracy in the world frame, and fidelity of the body configuration after horizontal root displacement and heading differences are removed.

Consider a reference motion $\mathcal{M}_{\mathrm{motion}}$ and the corresponding rollout of $\pi_\theta(\cdot\mid h_t,g_t)$. With the initial state indexed by $0$ as above, let $t=1,\ldots,T$ index the frames actually executed and evaluated. For a fixed control mode, the command mask satisfies $\mathbf{m}_t=\mathbf{m}$, and the set of evaluated links is
\begin{equation}
    \mathcal{L}_{\mathbf{m}}^{\mathrm{links}}=\{\ell\mid m^\ell=1\}.
    \label{eq:bfm_eval_links}
\end{equation}
Let $\mathbf{p}_{t,\ell}\in\mathbb{R}^3$ and $\mathbf{R}_{t,\ell}\in\mathrm{SO}(3)$ denote the world-frame position and orientation of link $\ell$, extracted from $s_t$, and let $\mathbf{p}_{t,\ell}^{\rm ref}$ and $\mathbf{R}_{t,\ell}^{\rm ref}$ denote their counterparts in $s_t^{\rm ref}$. In SI units, these position quantities are measured in $\mathrm{m}$. Each rotation matrix maps vectors from the corresponding link frame to the world frame. The reference and executed states refer to the same robot embodiment and the same timestep. Only links in $\mathcal{L}_{\mathbf{m}}^{\mathrm{links}}$ contribute to the tracking metrics. Sequences with an all-zero command mask are excluded from the tracking-error means below, and the number excluded is reported. An unspecified link may move differently from the reference while helping the robot satisfy the commanded targets.

\paragraph{Success Rate} For an evaluation set containing $N_{\mathrm{eval}}>0$ reference sequences, let $T_i$ denote the full reference length of sequence $i$ in frames, with $T_i>0$ and a nonempty selected-link set. Given a positional tolerance $\delta>0$, the success rate is
\begin{equation}
    \operatorname{Succ}(\delta)
    =\frac{1}{N_{\mathrm{eval}}}\sum_{i=1}^{N_{\mathrm{eval}}}
    \mathbf{1}\!\left[
        \max_{\substack{1\leq t\leq T_i\\ \ell\in\mathcal{L}_{\mathbf{m}}^{\mathrm{links}}}}
        \left\|\mathbf{p}_{t,\ell}^{(i)}
        -\mathbf{p}_{t,\ell}^{\mathrm{ref},(i)}\right\|_2
        \leq\delta
    \right],
    \label{eq:bfm_eval_success}
\end{equation}
where $\mathbf{1}[\cdot]$ is the indicator function. A sequence succeeds only if every selected link remains within the tolerance throughout its execution. A rollout terminated by failure before completing the reference is counted as unsuccessful. ScaleBFM uses $\delta=0.5\,\mathrm{m}$. The maximum over links and time makes this criterion sensitive to even a single large deviation. Success rate therefore complements average tracking errors, which can conceal brief failures. Since the criterion constrains position alone, orientation accuracy must be assessed separately. The score lies in $[0,1]$ and may be multiplied by $100$ when reported as a percentage.

\paragraph{Global Tracking Errors} For $T>0$ and a nonempty selected-link set, the global mean per-keypoint position error averages the Euclidean distance between corresponding link positions. The error is computed as
\begin{equation}
    \mathrm{G\mbox{-}MPKPE}
    =\frac{1}{T|\mathcal{L}_{\mathbf{m}}^{\mathrm{links}}|}
      \sum_{t=1}^{T}\sum_{\ell\in\mathcal{L}_{\mathbf{m}}^{\mathrm{links}}}
      \left\|\mathbf{p}_{t,\ell}-\mathbf{p}_{t,\ell}^{\rm ref}\right\|_2.
    \label{eq:bfm_eval_global_position}
\end{equation}
This error measures positional agreement in the world frame and is affected by both root motion and body articulation. For example, a controller may reproduce the reference limb configuration while gradually drifting away from the intended path, resulting in a large global position error.

For orientations, define the angular distance between two rotation matrices as
\begin{equation}
    d_{\mathrm{R}}(\mathbf{R}_1,\mathbf{R}_2)
    =\arccos\!\left(
        \operatorname{clip}\!\left(
            \frac{\operatorname{tr}(\mathbf{R}_1^\top \mathbf{R}_2)-1}{2},
            -1,1
        \right)
    \right),
    \label{eq:bfm_eval_rotation_distance}
\end{equation}
where $\operatorname{tr}$ denotes the matrix trace and clipping accommodates numerical roundoff. This distance is the magnitude of the shortest relative rotation, taking values in $[0,\pi]$. It gives a geometric orientation error without depending on the choice or wrapping of Euler angles. The global mean per-keypoint rotation error is then
\begin{equation}
    \mathrm{G\mbox{-}MPKRE}
    =\frac{1}{T|\mathcal{L}_{\mathbf{m}}^{\mathrm{links}}|}
      \sum_{t=1}^{T}\sum_{\ell\in\mathcal{L}_{\mathbf{m}}^{\mathrm{links}}}
      d_{\mathrm{R}}\!\left(\mathbf{R}_{t,\ell},\mathbf{R}_{t,\ell}^{\rm ref}\right).
    \label{eq:bfm_eval_global_rotation}
\end{equation}
Position errors are measured in meters and rotation errors in radians. Both are means of per-link distances, without squaring the errors.

\paragraph{Local Tracking Errors} Local evaluation removes the horizontal root position and heading of each trajectory at each timestep. To make this convention explicit, let
\begin{equation}
    \mathbf{c}_{t}^{\mathrm{root}}=
    \begin{bmatrix}
        p_{t,\mathrm{x}}^{\rm root} &
        p_{t,\mathrm{y}}^{\rm root} &
        0
    \end{bmatrix}^{\!\top},
    \qquad
    \mathbf{H}_t=\mathbf{R}_{\mathrm{z}}(\psi_{t}^{\mathrm{head}}),
    \label{eq:bfm_eval_heading_frame}
\end{equation}
where $\mathbf{p}_t^{\rm root}$ is the robot root position, $\psi_{t}^{\mathrm{head}}$ is its heading angle, and $\mathbf{R}_{\mathrm{z}}(\psi_{t}^{\mathrm{head}})$ is a rotation about the vertical axis. Define $\mathbf{c}_{t}^{\mathrm{root},\mathrm{ref}}$ and $\mathbf{H}_t^{\rm ref}$ analogously from the reference root state. The aligned link poses are
\begin{equation}
    \begin{aligned}
        \bar{\mathbf{p}}_{t,\ell}
        &=\mathbf{H}_t^\top(\mathbf{p}_{t,\ell}-\mathbf{c}_{t}^{\mathrm{root}}),
        &\qquad
        \bar{\mathbf{R}}_{t,\ell}
        &=\mathbf{H}_t^\top \mathbf{R}_{t,\ell},\\
        \bar{\mathbf{p}}_{t,\ell}^{\rm ref}
        &=(\mathbf{H}_t^{\rm ref})^\top
          (\mathbf{p}_{t,\ell}^{\rm ref}-\mathbf{c}_{t}^{\mathrm{root},\mathrm{ref}}),
        &
        \bar{\mathbf{R}}_{t,\ell}^{\rm ref}
        &=(\mathbf{H}_t^{\rm ref})^\top \mathbf{R}_{t,\ell}^{\rm ref}.
    \end{aligned}
    \label{eq:bfm_eval_local_poses}
\end{equation}
The executed and reference poses are thus expressed in their respective heading frames, with each frame origin placed at the root's horizontal projection onto the ground plane. Because this transformation removes only horizontal translation and yaw, discrepancies in height, roll, and pitch remain measurable. Applying the same position and rotation distances to these aligned quantities yields
\begin{equation}
    \begin{aligned}
        \mathrm{L\mbox{-}MPKPE}
        &=\frac{1}{T|\mathcal{L}_{\mathbf{m}}^{\mathrm{links}}|}
          \sum_{t=1}^{T}\sum_{\ell\in\mathcal{L}_{\mathbf{m}}^{\mathrm{links}}}
          \left\|\bar{\mathbf{p}}_{t,\ell}
          -\bar{\mathbf{p}}_{t,\ell}^{\rm ref}\right\|_2,\\
        \mathrm{L\mbox{-}MPKRE}
        &=\frac{1}{T|\mathcal{L}_{\mathbf{m}}^{\mathrm{links}}|}
          \sum_{t=1}^{T}\sum_{\ell\in\mathcal{L}_{\mathbf{m}}^{\mathrm{links}}}
          d_{\mathrm{R}}\!\left(\bar{\mathbf{R}}_{t,\ell},
          \bar{\mathbf{R}}_{t,\ell}^{\rm ref}\right).
    \end{aligned}
    \label{eq:bfm_eval_local_errors}
\end{equation}
A low local error indicates accurate reconstruction of the commanded body configuration after this alignment. A substantial global position error together with a small local position error can therefore indicate difficulty following the desired root trajectory despite accurate local pose reproduction. Global and local scores should consequently be interpreted together. Their coordinate conventions describe how a rollout is scored and are separate from the coordinate encoding used to provide $g_t$ to the policy.

For meaningful comparisons, results should be reported separately for each control mode using matched reference motions, selected links, initialization, and sampling frequency. The formulas above define errors for one sequence. The reported mean is the arithmetic mean of the per-sequence errors, with each included sequence receiving equal weight. The error statistics retain the evaluated frames from failed and truncated rollouts. Because early termination can omit harder parts of a reference, mean errors should be interpreted together with success rate and rollout duration. Applying the same protocol to held-out motions from familiar sources and motions from unseen sources further distinguishes tracking accuracy from cross-source generalization. These measures characterize the realization of behavioral specifications. Downstream scene-interaction tasks additionally require evaluation of their own task outcomes.

\subsubsection{Dexterous Hands: Motor Control at the Finger Level}
\label{subsec:ch07:dexterity}

A Cartesian hand target in a body controller specifies where the hand should move, but leaves its finger configuration and object contacts unresolved. Extending motor pretraining to dexterity therefore requires examples of contact transitions, together with an interface that can request useful movement without prescribing every contact.

\paragraph{The Held Object as a Floating Base}
A multi-fingered hand needs the same layers of low-level control as the humanoid body, because its control problem has the same structure. Recall from Eq.~\eqref{eq:partitioned} that the first six rows of a legged robot's dynamics contain no joint torque. The base moves only through inertial coupling and contact. In in-hand manipulation the unactuated body is the held object. With the wrist pose given by the arm, the object twist $\boldsymbol{\nu}_{\mathrm{o}}\in\mathbb{R}^6$ and the finger joints $\mathbf{q}_{\mathrm{h}}$ evolve as
\begin{equation}
\mathbf{M}_{\mathrm{o}}\dot{\boldsymbol{\nu}}_{\mathrm{o}}+\mathbf{b}_{\mathrm{o}}
=\mathbf{G}(\mathbf{q}_{\mathrm{h}},\mathbf{x}_{\mathrm{o}})\,\mathbf{f}_{\mathcal{K}_{\mathrm{hand}}}
+\mathbf{w}_{\rm ext},
\qquad
\mathbf{M}_{\mathrm{h}}(\mathbf{q}_{\mathrm{h}})\ddot{\mathbf{q}}_{\mathrm{h}}+\mathbf{b}_{\mathrm{h}}
=\boldsymbol{\tau}-\mathbf{J}_{\mathrm{h},\mathcal{K}_{\mathrm{hand}}}^{\top}\mathbf{f}_{\mathcal{K}_{\mathrm{hand}}},
\label{eq:ch07:hand-object}
\end{equation}
where $\mathcal{K}_{\mathrm{hand}}$ is the set of active finger-object contacts, $\mathbf{x}_{\mathrm{o}}$ is the object pose, $\mathbf{M}_{\mathrm{o}}$ and $\mathbf{b}_{\mathrm{o}}$ its spatial inertia and bias wrench, $\mathbf{f}_{\mathcal{K}_{\mathrm{hand}}}$ stacks the forces that the fingers apply at the active contacts, and the grasp matrix $\mathbf{G}$ maps them to the resultant wrench on the object~\citep{lynch2017modern}. Each contact force must lie in the friction cone of Eq.~\eqref{eq:friction-cone}. As for the floating base, no actuator acts on the object directly. Joint torque $\boldsymbol{\tau}$ reaches it only through $\mathbf{J}_{\mathrm{h},\mathcal{K}_{\mathrm{hand}}}$ and $\mathbf{G}$, and only while the corresponding contacts are closed. Contact forces in the null space of $\mathbf{G}$ squeeze the object without moving it. They set the friction margins that keep the grasp stable~\citep{bicchi1994forces}.

\begin{table}[!b]
  \centering
  \caption{Correspondence between humanoid whole-body control and dexterous in-hand control. References in the humanoid-body column point to the relevant formulations and sections of this chapter. The dexterous-hand column shows analogous quantities for a multi-fingered hand holding an object}
  \label{tab:ch07:body-hand}
  \small
  \begin{tabularx}{\linewidth}{@{}>{\raggedright\arraybackslash}p{0.22\linewidth}
      >{\raggedright\arraybackslash}X
      >{\raggedright\arraybackslash}X@{}}
    \toprule
    Concept & Humanoid body & Dexterous hand \\
    \midrule
    Unactuated body & Floating base, Eq.~\eqref{eq:partitioned}
      & Held object, Eq.~\eqref{eq:ch07:hand-object} \\
    Contacts & Feet or hands on the environment\newline Friction cone
      & Fingertips, phalanges, and palm on the object\newline Same cone \\
    Contact schedule & Gait phase, stance and swing (Section~\ref{sec:motion-control})
      & Finger gaiting (release, reposition, re-contact) \\
    Stability objective & Balance of the center of mass over the support
      & Grasp stability (resultant wrench and friction margins) \\
    Hidden state & Base velocity, foot contact, terrain
      & Object pose, contact locations, incipient slip \\
    Command interface & Velocity commands, masked link targets, Eq.~\eqref{eq:masked_command}
      & Fingertip targets, object-pose goals, functional finger coordinates \\
    Reference data & Retargeted human motion capture
      & Retargeted human hand motion from gloves or video \\
    Failure and recovery & Fall and get-up (Section~\ref{sec:real-time-execution})
      & Slip, drop, and regrasp \\
    \bottomrule
  \end{tabularx}
\end{table}

Table~\ref{tab:ch07:body-hand} extends the correspondence to the rest of the chapter. Two points go beyond the table. First, rolling or sliding fingertips change $\mathbf{G}$ and $\mathbf{J}_{\mathrm{h},\mathcal{K}_{\mathrm{hand}}}$ during a manipulation, much as foot placement changes the stance Jacobian, so the contact mode is part of the state to be estimated. Second, the hand differs in scale and instrumentation. Finger actuators are small, often heavily geared, and work close to their force limits, so the gap between commanded and realized torque in Eq.~\eqref{eq:actuator-model} matters more than it does for a leg. Available sensing ranges from motor current to fingertip tactile images (Section~\ref{sec:c9:dex-hardware}). And the hand's ``floating base'' has a different mass and shape in every task.

The three routes to reusable motor skills named at the start of this section, namely reference tracking, learned priors, and goal-conditioned interfaces, carry over to the hand with the object in place of the base.

\paragraph{Tracking Human Hand References}
TeleDexter makes the desired object motion explicit~\citep{li2026teledexter}. The operator specifies fingertip positions and object pose in the wrist frame. A recurrent hand policy maps these goals and the observed hand-object state to residual joint-position commands, while independent inverse kinematics tracks the operator's wrist with a fixed-base Franka FR3 arm. Training uses consecutive hand-object subgoals, with sparse goal-reaching rewards supplemented by dense tracking rewards. The hand can discover contact strategies between subgoals rather than reproduce every reference frame. Deployment runs at 30\,Hz. Reaching and initial grasping use kinematic retargeting before the operator switches to the learned in-hand controller. Its training signal follows Eq.~\eqref{eq:tracking_reward}, but the reference covers the object and fingertips instead of every joint, much like the sparse masked targets of Eq.~\eqref{eq:masked_command}.

On a SharpaWave hand, TeleDexter achieves 75.2\% mean success over seven reorientation and tool-use tasks, with 15 trials per task. Reusing the human reference data after retargeting and training a separate LeapHand controller yields 60\% and 73.3\% success on two reorientation tasks. This is reuse of data and a training procedure across hand morphologies. The current controller remains object-specific. A new object requires reference collection and policy training.

\paragraph{A Learned Motion Prior for Fingers}
DexterityGen gives a concrete pretraining implementation~\citep{yin2025dexteritygen}. Simulation reinforcement learning generates transitions between nearby sampled grasps, supplemented by free-finger and precision-manipulation behaviors. Varying object geometry, wrist motion, and physical parameters broadens the resulting dataset. A diffusion model then learns a distribution of future finger-keypoint motion conditioned on proprioceptive history and a mode indicator. An inverse dynamics model maps sampled motion to joint-position targets. At deployment, an external motion command guides diffusion sampling toward the requested finger movement while the learned prior favors motions represented in the training data. Thus, the command enters through sampling guidance. It is not simply copied to the joints. As in a behavior foundation model, a prior learned from broad simulated experience supplies coordinated motion, and a sparse command selects among behaviors it already supports.

The experiments demonstrate reuse on an Allegro hand attached to a separately controlled Franka Panda arm, with human teleoperation supplying motion commands. Test objects are absent from the pretraining dataset. Assisted teleoperation improves object retention and completes reorientation and regrasping tasks, but complete screwdriver and syringe tasks succeed in only 3/20 and 4/20 trials, respectively. These results support a reusable hand-motion prior while identifying the difficulty of composing skills into long tasks. The low-level controller uses proprioception rather than visual or tactile input. Control errors provide only an indirect contact cue.

\paragraph{Object-Goal Control Learned in Simulation}
SimToolReal asks whether simulated experience can produce a controller that follows object goals taken from a human demonstration. It uses a 22-DoF Sharpa hand on a KUKA iiwa arm and trains a recurrent goal-reaching policy on procedurally generated tool-like objects. At deployment, a human RGB-D demonstration supplies a sequence of object poses. The policy observes proprioception, the current object pose, a grasp-region bounding box, and the current goal, and commands arm and hand joint targets~\citep{kedia2026simtoolreal}. A demonstration can thus specify how a brush should be oriented without saying which finger releases first, and the controller is free to rotate the brush in the hand when moving the whole arm would hit the table. This is the object-centered goal representation of Section~\ref{subsec:goal-grounding}. The study evaluates 120 real rollouts over 24 trajectories, 12 tool instances, and six categories. Its Task Progress metric counts the fraction of demonstrated poses reached within a distance threshold and does not measure each tool's effect~\citep{kedia2026simtoolreal}. Reaching the intended hammer trajectory is not the same as driving a nail to depth. In the terms of Table~\ref{tab:ch07:body-hand}, SimToolReal tracks a reference for the unactuated body itself and leaves the contact schedule to the policy, as a velocity command leaves footstep placement to a locomotion controller.

\paragraph{Adapting a Prior to Tight Contact}
Play2Perfect, also on a Sharpa hand and KUKA arm, pretrains object manipulation in free space and then fine-tunes in CAD-derived assembly environments with sparse goal rewards. In 10 real trials at 0.5\,mm insertion clearance, the fine-tuned policy succeeds 6 times and the play-only policy never~\citep{lum2026play2perfect}. The deployment is zero-shot sim-to-real, but the policy for the assembly task itself was trained in simulation. Free-space reorientation can reach the right pose through many contact arrangements. Insertion cannot, since a small angular error produces an edge contact that jams the part, and the grasp must still allow corrective motion.

This is the hand counterpart of adapting a pretrained body controller to a new behavior, as the residual decoder of the original BFM does. The prior makes the approach reachable, and task-specific training concentrates on the final contact.

The interfaces exposed by the four controllers to the layer above are fingertip and object goals (TeleDexter), guided finger-keypoint motion (DexterityGen), object-pose sequences (SimToolReal), and a task reward in simulation (Play2Perfect). Any of them could be where a manipulation foundation model (Section~\ref{sec:c6:dexterous}) hands a goal to a hand controller, as a semantic planner hands targets to a humanoid controller (Section~\ref{architecture-and-interfaces}). None yet combines broad object coverage with contact sensing, the subject of the next subsection.

\subsubsection{Contact Feedback and Force-Adaptive Body-Hand Coordination}
\label{subsec:ch07:contact-balance}

Contact execution requires feedback that distinguishes mechanically different states. Similar finger positions can occur during pressing, sliding, or a failed regrasp. TeleDexter excludes contact-force and tactile observations. Its failure analysis links tracking stalls to the inability to distinguish pressing against an object from sliding past it. Tool impacts also perturb grasps beyond its free-space hand-object training distribution. Motion targets specify intended interaction, while contact feedback helps determine whether it is being realized.

One response is to give the controller a state that keeps evidence the current image has lost. WM-Craftnet combines wrist depth, binary tactile contact, proprioception, and previous actions in a recurrent World Synesthesia Model, whose state conditions a PPO policy that outputs relative targets for the 22 joints of a Sharpa Wave hand. Simulation supplies clean-depth reconstruction targets during training. At deployment the state is updated from noisy sensor data~\citep{yin2026wmcraftnet}. The model gives the policy context and does not generate imagined rollouts, so it is a history-based state estimator in the sense of Section~\ref{sec:wm-belief}, serving as the hand counterpart of the RMA-style estimators of Section~\ref{subsec:implicit-estimation}, with the object and its contacts as the hidden state. On an unseen double-notched block, WM-Craftnet meets the rotation criterion (more than half a revolution within 20 seconds) in 8 of 10 real trials, while a policy using WSM-denoised depth succeeds in none~\citep{yin2026wmcraftnet}. This favors recurrent interaction context over image denoising alone, although matched comparisons of history, capacity, and objectives would be needed to pin down the source of the gain.

Touch can also correct execution between visual updates. T-Rex, evaluated on a fixed-base dual-arm system with two Sharpa Wave hands, refines actions asynchronously from recent fingertip deformation and force history, reusing the visual-language context instead of recomputing it for every refinement~\citep{niu2026trex}. A separate 300\,Hz thread tracks the updated targets. Recent contact evidence can therefore change finger motion while the action chunk remains valid at the task level, the multirate interface of Section~\ref{architecture-and-interfaces}. In a six-task ablation, the average score is 65 for the full system, 42 without touch, and 60 without asynchronous refinement, under the paper's task-specific scoring~\citep{niu2026trex}. Both the tactile signal and the way it is integrated contribute. The split between slow refinement and a fast servo is the hand version of the latency budget of Section~\ref{sec:real-time-execution}. The tactile rates that make it necessary are given in Section~\ref{sec:c9:dex-hardware}.

Body-hand coordination introduces a further mechanical constraint. When a held object accelerates with the body, its required net contact force changes. Arm configuration can also change the moment transmitted to the base. With both hands on one object, their forces jointly determine its resultant wrench. Force components that cancel at the object can still change internal grasp loading and friction margins~\citep{bicchi1994forces}. Coordinating a feasible object trajectory therefore also requires maintaining contacts and body support.

FAME provides direct evidence for force-adaptive standing~\citep{pudasaini2026fame}. An encoder jointly maps 15 torso-and-arm joint positions and the two three-dimensional hand forces to an eight-dimensional latent, which conditions a lower-body policy. Training combines an upper-body pose curriculum with sampled hand loads. Deployment estimates each hand force from measured joint torques $\boldsymbol{\tau}_t$, rigid-body inverse dynamics, and the hand Jacobian, without wrist force/torque sensors. The policy outputs 12 leg-joint position offsets at 50\,Hz. Upper-body motion remains externally specified. This force estimate describes the load transmitted through an arm.

In simulation, FAME is evaluated on 100 upper-body configurations sampled from the training pose distribution, with 25 load draws per configuration. Success requires remaining upright, preserving the stance, and limiting force-induced hand displacement. At a 150\,mm tolerance, success is 38.9\%, compared with 16.6\% when the same force information is supplied without jointly encoding force and upper-body configuration. On a Unitree H1-2, FAME maintains stance in all three asymmetric single-arm and all three symmetric bimanual loading trials. The pose-only baseline falls in all corresponding trials. These limited hardware tests establish load-dependent leg compensation. Together, the cases identify complementary interfaces. Object-and-finger goals request contact execution, tactile observations refine local actions, and hand-load estimates inform body support. An integrated system would need to be evaluated on simultaneous grasp maintenance, object motion, and balance.

\subsection{Real-Time Execution, Recovery, and Safety Control}
\label{sec:real-time-execution} 

Deploying learned control policies from simulation to real humanoid robots involves challenges that extend beyond the performance of the policy itself. In practice, the system must operate under limited computational resources, perception and communication delays, actuator dynamics, and unexpected external disturbances while maintaining stable, continuous, and safe control. This is particularly challenging for humanoid robots due to their high-dimensional, strongly coupled, and partially underactuated dynamics, where local errors can propagate into whole-body deviations or even loss of balance. Therefore, real-world deployment requires jointly addressing real-time inference, motion smoothing, sim-to-real adaptation, anomaly monitoring, disturbance rejection, and recovery, ultimately forming a closed-loop safety framework from normal execution to failure handling and task recovery. Standing up from a fallen, seated, or leaning posture brings several of these requirements together. The robot must reorganize its contacts, control the resulting motion, and settle into stable support. HoST, introduced in \emph{Learning Humanoid Standing-up Control across Diverse Postures}, provides a concrete example of learning this recovery capability~\citep{huang2025host}.

\subsubsection{Real-Time Execution and the Sim-to-Real Gap}
Deploying learned control policies on physical humanoid robots requires bridging a substantial gap between simulation and the real world. This gap arises from multiple mismatches between the simulated and physical systems across dynamics, actuation, sensing, computation, and environmental interaction. An implementation with explicit state estimation can be summarized as
\begin{equation}
 o_t\longrightarrow\hat{s}_t
 \longrightarrow\pi_\theta(\cdot\mid\hat{s}_t,g_t)
 \longrightarrow \mathbf{a}_t
 \longrightarrow\text{actuator and physical system}
 \longrightarrow o_{t+1},
\end{equation}
where $o_t$ denotes the observation at time $t$, $\hat{s}_t$ is the estimated robot state, $\pi_\theta$ is the learned policy, $g_t$ is the command, and $\mathbf{a}_t$ is the resulting action. The schematic suppresses the estimator's memory and any additional policy history. A history-conditioned policy can instead use $h_t$ directly or combine it with explicit estimates. In practice, the discrepancy between the simulated and real execution of this loop can progressively degrade motion tracking, stability, and task performance. The major sources of the sim-to-real gap are discussed below.

\paragraph{Dynamics and Physical Parameter Mismatch} The physical dynamics of a humanoid robot are difficult to reproduce exactly in simulation. Differences in mass distribution, center of mass, inertia, joint friction, damping, actuator strength, gear transmission, and joint limits can cause the simulated robot to respond differently from its physical counterpart under the same control command. Contact-related parameters, such as ground friction, restitution, and collision geometry, further introduce discrepancies during locomotion, jumping, and other highly dynamic motions. Although system identification and domain randomization can reduce these differences, the real robot inevitably exhibits unmodeled dynamics and parameter variations that are difficult to capture completely during training.

\paragraph{Actuation and Low-Level Control Mismatch} The action produced by a learned policy is not directly realized as an idealized control input on a physical robot. Instead, it is typically converted through a low-level controller, actuator dynamics, communication interfaces, and hardware-specific constraints. Differences between simulated and real actuator response, including torque tracking errors, motor bandwidth, saturation, dead zones, and control frequency, can therefore introduce discrepancies between the commanded and realized motion. These errors become particularly significant for fast motions, where accurate timing and high-frequency torque responses are essential.

\paragraph{Perception and State-Estimation Errors} In simulation, the policy can often access accurate robot states, object positions, or environment information, whereas real systems must estimate these quantities from noisy and delayed sensor measurements. Camera noise, depth errors, occlusions, motion blur, imperfect calibration, and tracking failures can lead to discrepancies between the true physical state and the state provided to the policy. Such errors can alter both the magnitude and timing of the resulting control actions, especially for tasks involving fast-moving objects or rapidly changing contacts.

\paragraph{Control Latency and Asynchronous Execution} Real-time execution also introduces computational and communication delays across perception, state estimation, policy inference, communication, and actuation. For a sequential processing path, the total latency can be approximated as
\begin{equation}
 d_{\rm total}=d_{\rm perception}+d_{\rm estimation}
 +d_{\rm policy}+d_{\rm communication}+d_{\rm actuation},
\end{equation}
where each $d$ is a time duration. In asynchronous or overlapping pipelines, the observation timestamp and actual execution time determine information age. The sum alone is insufficient. Given a control period $\Delta t$, a large $d_{\rm total}$ means that the action is effectively computed from an outdated state, introducing a phase lag into the control loop. This effect is particularly detrimental to fast dynamic motions and balance control, where even a small temporal mismatch can result in noticeable tracking errors. In addition, asynchronous execution between perception, policy inference, and low-level control can introduce variable delays, making the effective control dynamics different from those assumed during simulation.

HoST illustrates how policy updates and joint feedback operate at different rates. On the 23-DoF Unitree G1, a three-layer MLP uses a short proprioceptive history, including body roll and pitch, angular velocity, and joint states, together with the previous action and action scale. The policy updates joint-position targets at 50\,Hz, while the physical robot's proportional-derivative (PD) servos run at 500\,Hz~\citep{huang2025host}. Joint feedback thus continues between policy updates. The simulation uses a 200\,Hz PD loop, illustrating why transfer depends on the response of the complete policy-servo system as well as the policy update rate.

\paragraph{Contact and Environmental Mismatch} Physical interaction with the environment is another important source of discrepancy. Contact locations, collision geometry, ground compliance, friction conditions, and external disturbances are often simplified in simulation but can vary substantially in the real world. For humanoid robots, small differences in foot-ground contact or object interaction can change the resulting contact forces and consequently alter the entire-body motion. Such discrepancies are particularly challenging for behaviors involving impacts, rapid changes of support, or interaction with non-rigid or unstructured environments.

\paragraph{Reference-Tracking Error} For systems that generate reference motions and subsequently track them with a learned controller, the sim-to-real gap can be further amplified by tracking errors. Even if a kinematic trajectory generated by a planner is physically plausible, the real robot may not reproduce it exactly due to actuator limitations, modeling errors, latency, and state-estimation noise. The resulting deviation between the reference trajectory and the executed motion can accumulate over time, causing the robot to gradually drift away from the intended state. This issue is especially important in hierarchical architectures, where the output of one module becomes the input or conditioning signal of another.

\paragraph{Error Accumulation in Hierarchical and Autoregressive Systems} The effect becomes more pronounced when motion generation and control are coupled through multiple sequential modules. Consider a hierarchical system consisting of a high-level motion planner and a low-level tracking controller. If the planner generates a reference trajectory $\mathbf{q}_{t:t+H}^{\rm ref}$ spanning the current step and $H$ future control steps and the tracker executes it with tracking error, the actual state can deviate from the planned trajectory. When subsequent planning decisions use execution feedback, this deviation can propagate to future references. Errors in previously generated references can also propagate through the history used by an autoregressive planner.

\begin{figure}[tb]
  \centering
  \includegraphics[width=\linewidth]{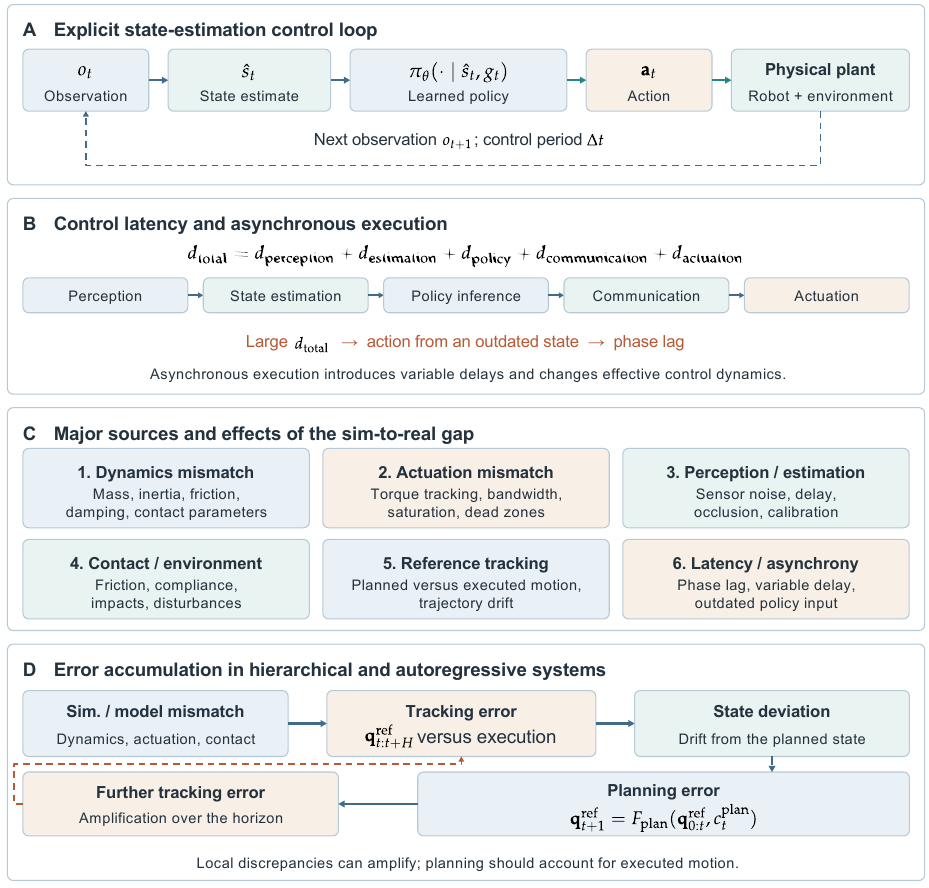}
  \caption{Sources of sim-to-real mismatch and their propagation through a planner-tracker system. The latency sum depicts a sequential processing path. The dashed state-to-planning connection applies when execution feedback conditions the planner}
  \label{fig:ch07:sim2real-gap}
\end{figure}

This problem is particularly evident in autoregressive kinematic planners. Instead of generating an entire trajectory independently, an autoregressive planner predicts future motion conditioned on previously generated reference states and a planning condition $c_t^{\rm plan}$ according to
\begin{equation}
 \mathbf{q}_{t+1}^{\rm ref}
 =F_{\rm plan}(\mathbf{q}_{0:t}^{\rm ref},c_t^{\rm plan}),
\end{equation}
where $F_{\rm plan}$ is the motion generator. Consequently, an error in an earlier prediction can become part of the input to subsequent predictions, resulting in accumulated prediction error over a long horizon. In a hierarchical planner-tracker architecture, tracking errors introduce another source of deviation between the planned and physically executed states. When execution feedback enters $c_t^{\rm plan}$ or another planning input, these state deviations can change future references. This coupling is particularly important for long-horizon and highly dynamic behaviors, where small local discrepancies can progressively amplify over time. Deep Tracking Control combines an optimization-based planner with a learned tracker~\citep{jenelten2024}. AdaPT conditions a tennis planner on robot and ball information and adapts the reference execution speed~\citep{huang2026towards}. These examples illustrate how planner-tracker interfaces connect motion generation with physical execution. Figure~\ref{fig:ch07:sim2real-gap} summarizes sources of sim-to-real mismatch and their propagation through a planner-tracker system.

\subsubsection{Motion Smoothing and Continuous Execution}
In addition to inference speed, the temporal continuity of policy outputs is critical for stable real-world execution. Directly executing the actions produced by a learned policy at each timestep may result in high-frequency oscillations, particularly under observation noise, out-of-distribution states, or discrepancies between simulated and real dynamics. Such oscillations can introduce unnecessary joint acceleration, actuator load, and mechanical wear.

\emph{Locally Lipschitz Continuous Constraint (L2C2)} regularization encourages local smoothness of policy and value functions during training~\citep{kobayashi2022l2c2}. It penalizes changes under nearby input perturbations, helping reduce sensitivity to small observation changes. For highly dynamic behaviors, the resulting action continuity must be considered together with the responsiveness needed to preserve task-relevant motion components.

HoST applies L2C2 to the actor and each critic, constructing nearby inputs from differences between successive observed states. This training objective encourages smoother actor and value outputs. In hardware experiments with 5 trials in each of 4 settings (ground, platform, wall, and slope), HoST succeeds in 20/20 trials, compared with 11/20 after removing L2C2~\citep{huang2025host}. These results connect local function smoothness to the reliability of physical execution under the tested conditions.

Motion amplitude introduces a complementary trade-off. Large actions aid exploration but can produce violent transitions. HoST bounds the policy output and constructs each joint target relative to the current joint position according to
\begin{equation}
 \mathbf q_{j,t}^{\mathrm{d}}=\mathbf q_{j,t}+\alpha_{\rm act}\mathbf{a}_t,
 \qquad \mathbf{a}_t\in[-1,1]^n,
 \label{eq:host-action-target}
\end{equation}
where $\mathbf q_{j,t}$ and $\mathbf q_{j,t}^{\mathrm{d}}$ are the measured and desired positions of the $n$ actuated joints, and $\alpha_{\rm act}$ scales the target increment in radians. Meeting an episode-end head-height criterion advances an action-scale curriculum that reduces $\alpha_{\rm act}$ from 1 toward 0.25~\citep{huang2025host}. This bounds the requested displacement at each update and indirectly moderates motion through the PD response. Gradual tightening retains early exploration while encouraging less abrupt execution as the skill develops.

For tracking a reference motion with specified timing, smoothing must also preserve the required movement speed. The objective is to suppress high-frequency noise while preserving the temporal structure and key dynamic characteristics of the original motion. Excessive smoothing may, for example, reduce the speed of a tennis swing or alter the timing of a dynamic motion. Therefore, motion smoothing should be designed jointly with motion phase, robot dynamic capabilities, and task constraints. The planner-tracker decomposition in AdaPT converts generated motions into executable commands through a dedicated tracking controller.

\subsubsection{Sim-to-Real Adaptation} The deployment mismatches described above motivate training and adaptation across variations in dynamics, sensing, and latency.

A widely used approach is domain randomization, where physical and environmental parameters are randomized during training by sampling from a distribution
\begin{equation}
 \boldsymbol{\xi}_{\rm sim}\sim p(\boldsymbol{\xi}_{\rm sim}),
\end{equation}
where $\boldsymbol{\xi}_{\rm sim}$ may include robot mass, friction coefficients, joint damping, actuator strength, observation noise, control latency, and external disturbances. By exposing the policy to a distribution of possible environments, the learned controller can become less sensitive to modeling errors.

Another approach is to calibrate the simulation using real-robot data through system identification, thereby reducing the discrepancy between simulated and physical dynamics. In practice, these approaches can be combined. Real-robot data are first used to estimate the dominant physical parameters, while domain randomization is then applied to cover the remaining uncertainty.

HoST combines coverage of initial support conditions with randomization of physical parameters. Ground, platform, wall, and slope settings expose training to different starting postures and contacts. Within these settings, randomization includes base center-of-mass offsets of up to $\pm3$\,cm and control delays from 0 to 100\,ms, alongside friction, actuator strength, and PD gains~\citep{huang2025host}. Hardware deployment also adjusts hip and knee PD stiffness to compensate for differences in torque response. The example shows how broad simulation coverage and servo-gain adjustment contribute jointly to transfer.

For humanoid motion control, an important observation is that a motion that is kinematically feasible in simulation is not necessarily dynamically executable on the real robot. Consequently, the training pipeline should account not only for task-level objectives but also for the actual controller, state estimation process, and execution constraints used during deployment. For systems that generate high-level motion references, the generated motion must ultimately be converted into executable behavior by a tracking controller that accounts for these constraints.

\subsubsection{Balance Protection and Fall Recovery}
For humanoid robots, a particularly important aspect of safety control is the possibility of losing balance and falling. A complete safety framework should therefore not assume that the robot never falls, but should also provide mechanisms for controlled falling, post-fall recovery, and subsequent task resumption.

Balance protection can be organized into several stages according to the severity of the instability. The first stage is rapid disturbance rejection, such as adjusting foot placement, torso orientation, and motion speed. If the robot can no longer recover its standing state and a validated protective controller is available, the system can enter a controlled fall mode and modify the body configuration to reduce impact. After falling, a dedicated, trained and validated get-up controller can bring the robot back to a stable standing configuration.

Learning such a controller requires objectives that guide the robot through changing support configurations. HoST learns standing-up motion without reference trajectories, using height-dependent reward activation to organize righting, rising, and stable standing~\citep{huang2025host}. Its four reward groups (task, style, regularization, and post-task) describe progress, motion quality, and the stability of the achieved posture. Each group has a separate critic. Its generalized advantage estimates are normalized within the training batch, then combined in a weighted sum to update a shared actor with PPO. This separates return estimation for signals with different scales and temporal structure while learning a policy that coordinates their requirements.

Exploration is especially difficult before the robot discovers how to raise its body into a supported posture. HoST therefore applies an upward force to the base during training when the trunk is nearly vertical. As the episode-end head-height criterion is met, the assistance decreases from 200\,N in 20\,N steps until it reaches zero. Together with the action-scale curriculum, this lets training first discover the recovery motion and then refine its execution with reduced assistance and smaller target increments. Post-task rewards favor low base linear and angular velocities and the maintenance of body height and posture. Consistently, simulation success requires the base to reach the target height and remain above it for the rest of the episode~\citep{huang2025host}. Recovery thus includes stabilizing the terminal posture.

Executing a learned recovery skill also depends on the available space around the body. HoST reports collisions with the surroundings in some seated-start trials using proprioception alone~\citep{huang2025host}. Such cases motivate environmental checks when selecting and initiating a recovery motion. Incorporating the skill into an autonomous system further requires fall detection, controller handover, and task resumption. The HoST study leaves this integration for future work.

Task recovery should go beyond simply returning the robot to a standing posture. The system should also restore the task context. For example, if the robot loses balance during a long-horizon motion imitation task, the recovery module should determine the current task phase and decide whether to restart the current motion, resume from an intermediate state, or replan a new executable trajectory. Integrating these decisions with balance protection makes recovery part of the task-level control process.

Future humanoid control systems should account for dynamic feasibility while integrating motion generation, real-time control, failure detection, risk mitigation, recovery, and task resumption. These functions need validation within the intended operating conditions.

\clearpage

\section{Physical Agents}
\label{ch:physical-agent}
For a long time, research on the interaction between AI and the physical world has focused primarily on short-term actions and isolated skills, such as executing a single grasp or traversing a short stretch of terrain. Yet a true embodied agent must be able to operate continuously in real environments over extended periods of time. Achieving this requires more than low-level perception, motor skills, and future prediction. Instead, it demands a capable cognitive core that can perform long-horizon planning and adapt as conditions change.

This chapter discusses how to build such an integrated system, which is a physical agent, that can interact with the physical world continuously in a closed loop. To this end, the chapter first defines the scope of physical agents and clarifies how they differ from purely digital agents and conventional robots. It then discusses how physical agents scale to adapt to entity-level, composite, and system-level constraints. Building on this foundation, it explains how physical goals become assessable task specifications, how available skills are organized into feasible behavior, and how execution feedback supports verification and recovery. Finally, it examines collaboration among heterogeneous agents and the emergence of distributed capabilities.

\subsection{Definition of Physical Agent}
\label{sec:physical-agent-definition}

As artificial intelligence extends from digital information spaces into the physical world, the notion of an intelligent agent acquires additional requirements. A digital agent may pursue goals, maintain memory, reason over observations, and invoke software tools, while its actions primarily alter informational states. A physical agent must additionally operate through an embodied system whose actions are constrained by geometry, dynamics, contact, sensing limitations, actuation uncertainty, time, energy, and safety ~\citep{RussellNorvig2020,PfeiferBongard2006,SicilianoKhatib2016}. Its decisions can change not only the state of the task, but also its own physical configuration, the surrounding environment, and the observations and actions that become available afterward.

A physical agent should therefore not be understood simply as a robot equipped with an artificial intelligence model. The relevant distinction lies in how the system organizes perception and action around an objective under physical constraints. In this chapter, the term refers to systems that are physically situated, possess some degree of autonomy in selecting their behavior, and continuously use the consequences of physical interaction to guide subsequent decisions. This perspective provides a common basis for discussing agents implemented as mobile robots, manipulators, legged systems, aerial vehicles, autonomous machines, or more distributed physical systems ~\citep{brooks1991intelligence,Liu2025EmbodiedAI}.

\subsubsection{Concept and Boundary of Physical Agent}
\label{subsec:physical-agent-concept}

\paragraph{A Working Definition}
In this chapter, \emph{a physical agent is defined as a goal-directed, relatively autonomous unit that perceives physical conditions, selects and executes feasible actions through its embodiment, and uses feedback from their physical effects to adjust subsequent behavior.} Its embodiment determines what it can sense and how it can act, while environmental conditions, resource limits, and assigned authority constrain admissible behavior. Feedback connects intended outcomes to observed changes in the agent and its surroundings, enabling it to continue execution, correct its actions, or revise its plan.

This definition treats physical agency as a relationship among objective, embodiment, environment, action, and feedback. The agent is \emph{situated} because its behavior takes place in an environment governed by spatial and physical relationships rather than solely by symbolic or computational rules. It is \emph{goal-directed} because its behavior is organized around a desired outcome rather than a fully prescribed sequence of actions. It is autonomous or semi-autonomous because at least part of the mapping from a task objective to concrete actions is determined during operation rather than completely specified in advance.

\begin{figure}[!t]
    \centering
    \includegraphics[width=\linewidth]{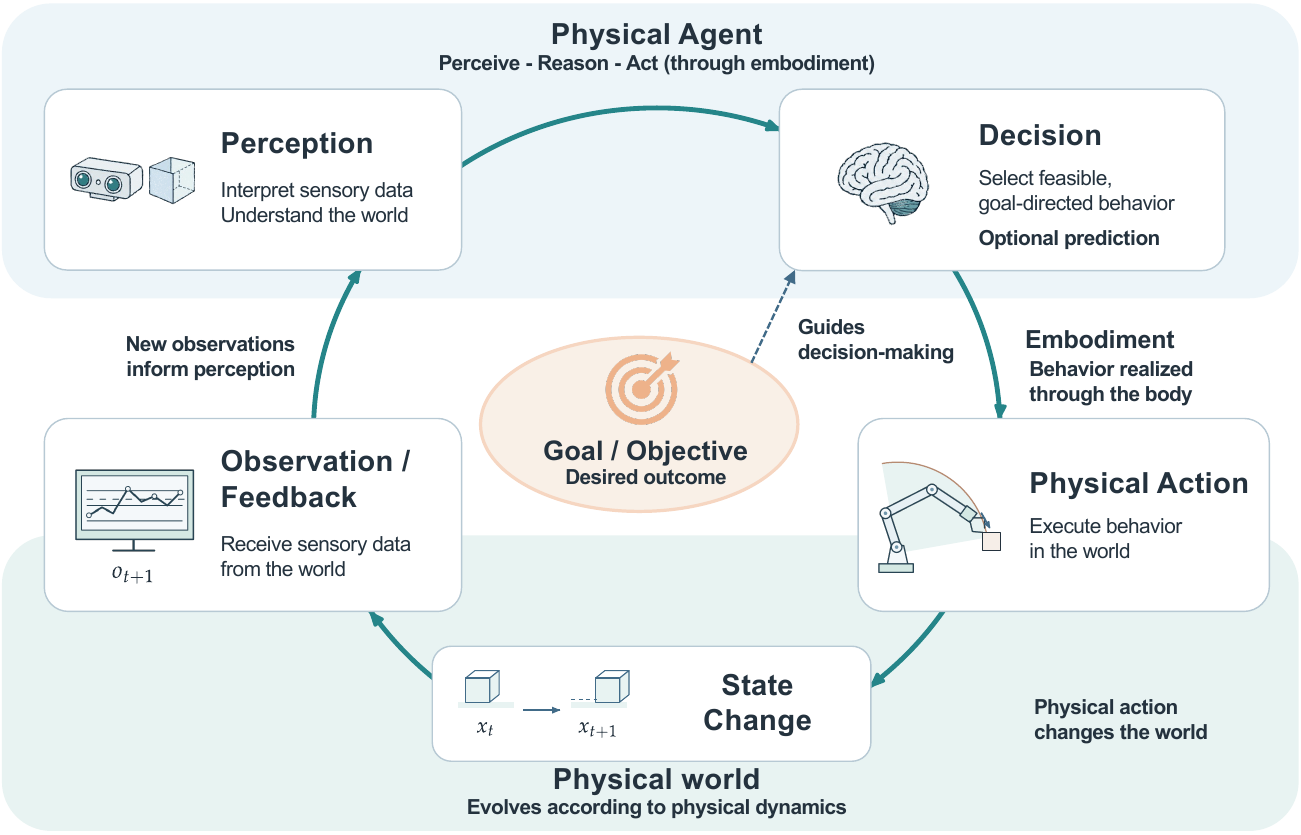}
    \caption{Closed-loop interaction between a physical agent and the physical world. Perception and task-relevant representations inform decision-making, which may include optional prediction and scenario simulation. Decisions are realized through embodied physical actions, and these actions change the physical state of the agent and/or environment. The resulting state gives rise to new observations that inform subsequent decisions. Here $x_t$ is the joint physical state at step $t$, $x_{t+1}$ its next state, and $o_{t+1}$ the resulting observation}
    \label{fig:physical-agent-loop}
\end{figure}

Physical embodiment is essential to this definition. The embodiment includes the system's mechanical structure together with its sensors, actuators, onboard computation, and the interfaces through which it exchanges information and physical influence with the environment. It determines what the agent can observe, which actions are feasible, and what physical consequences those actions may produce. The embodiment is therefore not merely a carrier on which an intelligent algorithm is deployed. It constitutes part of the agent's sensing and action capabilities and thereby defines part of its operational boundary ~\citep{PfeiferBongard2006,SicilianoKhatib2016}.

The definition also requires closed-loop interaction ~\citep{brooks1991intelligence,AstromMurray2008}. An action produces a physical consequence, that consequence gives rise to new observations, and those observations may alter subsequent behavior. This interaction is illustrated in Figure~\ref{fig:physical-agent-loop}.

Perception and action are coupled in this loop. Actions change the world from which future observations are obtained, and new observations change which actions are subsequently selected.

\paragraph{Physical Agents and Digital Agents}
A physical agent and a digital agent share several fundamental properties. Both may be goal-directed, maintain internal state, use memory, reason over uncertain information, invoke tools, and revise their behavior according to feedback. The distinction therefore does not lie in whether the system performs reasoning or planning.

The more consequential difference concerns the domain in which actions take effect. A digital agent primarily changes informational states, for example by generating text, editing files, querying a database, executing code, or invoking a software service. These actions may indirectly lead to physical consequences when mediated by an external system, but the agent's immediate action space is digital.

A physical agent acts through a physical embodiment. Its actions may change its position, orientation, velocity, configuration, contact state, or energy state, as well as the location, configuration, or physical condition of external objects. Such actions are subject to feasibility constraints and may produce delayed, uncertain, or irreversible consequences. A manipulator cannot instantaneously move its end effector to an arbitrary pose. An aerial vehicle cannot change velocity without satisfying its dynamics. A mobile robot cannot traverse an occupied region merely because a planner identifies it as geometrically desirable.

The distinction can therefore be expressed in terms of the state directly affected by action. Digital agents act primarily on informational states, whereas a physical agent directly participates in the evolution of physical states. This difference introduces requirements that are less prominent in purely digital settings, including dynamic feasibility, physical safety, execution uncertainty, limited sensing, resource constraints, and the need to verify the actual outcome of an action rather than only whether a command was issued.

\paragraph{Physical Agents and Robots}
The terms \emph{robot} and \emph{physical agent} also overlap without being synonymous. A robot primarily identifies a class of embodied machines capable of sensing, computation, and actuation. The concept of a physical agent places greater emphasis on how such capabilities are organized around an objective and how much of the resulting behavior is selected during interaction with the environment.

Consider an industrial manipulator that repeatedly follows a predetermined welding trajectory. The system is clearly a robot, and it may employ sophisticated feedback control to achieve high precision. Nevertheless, if the task, sequence, and responses to relevant situations are almost entirely specified in advance, its behavior exhibits relatively little agent-level autonomy. By contrast, a manipulation system that is given a desired outcome, selects an appropriate grasp and motion strategy from the observed scene, detects unsuccessful execution, and chooses an alternative course of action has stronger agentic properties even if the underlying physical platform is similar.

The distinction is therefore one of emphasis rather than a strict separation of system categories. A robot characterizes the physical means through which sensing and action are realized, whereas a physical agent characterizes a mode of operation in which perception, decision-making, physical action, and feedback are organized in pursuit of a goal. Many modern robotic systems satisfy both descriptions, while some robots remain primarily automated machines with narrowly prescribed behavior.

\subsubsection{Functional Organization and Interaction Loop}
\label{subsec:physical-agent-loop}

Building on the four elements in Chapter~\ref{ch:introduction} (Section~\ref{sec:core-elements}), this section distinguishes five functional roles by discussing world representation and physical prediction separately. These roles comprise physical state perception and estimation; physical representation and world models; physical prediction and simulation; physical decision-making, planning, control, and execution; and closed-loop feedback and multiscale adaptation. A physical agent realizes this same loop within an identifiable operational boundary. The following account applies those elements to agent-level behavior, without prescribing separate software modules. Their functions may be explicitly implemented, learned jointly, or distributed across computational and physical subsystems~\citep{SicilianoKhatib2016,Liu2025EmbodiedAI}.

Embodiment, objectives, and operating constraints provide the conditions under which all five functional roles function. The embodiment includes mechanical structure, sensors, actuators, computation, and environment interfaces. It determines which observations and physical actions are available. A wheeled platform, manipulator, and aerial robot may therefore use different strategies for the same objective. Goals specify desired outcomes, while kinematic and dynamic limits, collision avoidance, energy, time, permissions, and safety requirements delimit acceptable behavior. These conditions guide the entire loop. Memory supports continuity across its elements, and safety mechanisms constrain and monitor their operation.

\paragraph{Physical State Perception and Estimation}
A physical agent generally lacks direct access to the complete state of the world. Cameras, depth sensors, LiDAR, tactile and force sensors, inertial units, encoders, and other instruments provide incomplete measurements affected by noise, occlusion, calibration errors, and latency~\citep{SicilianoKhatib2016}. Perception and estimation construct task-relevant descriptions of the body's state and surroundings, including pose, motion, object configurations, contact, free space, or process quantities such as temperature. The required information and its accuracy depend on the decisions and control actions that follow.

\paragraph{Physical Representation and World Models}
Representations organize estimated information into spatial, semantic, relational, or physical descriptions that support action. Memory retains earlier observations, explored regions, attempted actions, and outcomes, enabling the physical agent to reason beyond its current view. A world model may encode object relationships, affordances, and transition regularities. Such information can be explicit, implicit in a learned policy, or distributed across components. A comprehensive internal simulator is therefore unnecessary. The relevant requirement is a representation adequate for the task and operating conditions.

\paragraph{Physical Prediction and Simulation}
Prediction uses available representations to anticipate how candidate actions, disturbances, or other participants may change physical conditions. Scenario simulation can compare alternative sequences before committing to execution, using analytical dynamics, learned models, or other approximations. Its usefulness depends on the prediction horizon, uncertainty, and computational cost. This element need not be a separate predictive module. A reactive controller or learned policy may incorporate local expectations implicitly, while explicit prediction supports longer horizons and more complex dependencies. Forecasts remain conditional on their assumptions and must be checked against interaction outcomes.

\paragraph{Physical Decision-Making, Planning, Control, and Execution}
This element connects estimated conditions and the objective to feasible physical intervention~\citep{RussellNorvig2020,Kober2013}. A complex objective can be decomposed into intermediate goals and realized through reusable skills, such as approaching, grasping, transporting, or inspecting. Planning coordinates these skills and their dependencies. Control converts the selected behavior into trajectories, velocities, forces, torques, or actuator commands compatible with the embodiment. An instruction to grasp a container may thus require approach motion, alignment, contact establishment, closure, and force regulation. Modular architectures may expose these stages individually, while learned policies may combine them. In either case, action selection remains conditioned on the task, current information, and admissible operating conditions.

\paragraph{Closed-Loop Feedback and Multiscale Adaptation}
Execution changes the physical state and produces new evidence. Feedback compares intended and observed outcomes and informs corrections at different scales~\citep{AstromMurray2008}. A controller can regulate contact force, a skill can adjust its motion, and a task planner can replace an unsuccessful operation. Over repeated operation, calibration and retained experience can support adaptation to equipment drift or changing conditions. Adaptation may update estimates, representations, predictions, or behavior without necessarily changing learned parameters online. Safety mechanisms operate throughout the loop by restricting candidate actions, monitoring execution, and triggering recovery, fallback, or intervention when limits are approached. Feedback therefore connects local regulation, task-level revision, and longer-term adaptation, sustaining the closed-loop organization described in Chapter~\ref{ch:introduction} and its application loop (Section~\ref{sec:app-loop}).

\paragraph{A Formal View of Physical Interaction}
The interaction of a physical agent with its environment can be represented as a partially observable closed-loop decision process~\citep{Kaelbling1998}. Let $t$ be a nonnegative integer indexing interaction steps, $B_{\mathrm{emb}}$ describe the embodiment and its sensing and actuation capabilities, and $\mathcal{X}$ be the space of joint physical states of the body and environment. Then
\begin{equation}
x_t \in \mathcal{X}
\end{equation}
denotes the physical state at step $t$. The physical agent receives an observation $o_t$ in an observation space $\mathcal{O}$ according to
\begin{equation}
o_t \sim p(o_t \mid x_t,B_{\mathrm{emb}}),
\label{eq:physical-agent-observation}
\end{equation}
where $p(o_t\mid x_t,B_{\mathrm{emb}})$ is the conditional observation distribution and $\sim$ denotes sampling from that distribution. It accounts for the sensing capabilities of the embodiment and measurement uncertainty.

Let $m_{t}^{\mathrm{mem}}$ denote retained information from earlier interactions, $g$ the task objective, $c_{t}^{\mathrm{constr}}$ the currently relevant constraints, and $o_{\leq t}=(o_0,\ldots,o_t)$ the observation history through step $t$. The policy $\pi$ is a conditional action distribution that selects an action $a_t$ from the embodiment's action space $\mathcal{A}_{B_{\mathrm{emb}}}$, subject to $c_{t}^{\mathrm{constr}}$:
\begin{equation}
a_t \sim
\pi
\left(
a_t
\mid
o_{\leq t},
m_{t}^{\mathrm{mem}},
g,
c_{t}^{\mathrm{constr}},
B_{\mathrm{emb}}
\right).
\label{eq:physical-agent-policy}
\end{equation}
The physical state then evolves according to
\begin{equation}
x_{t+1}
\sim
p(x_{t+1}\mid x_t,a_t,B_{\mathrm{emb}}),
\label{eq:physical-agent-dynamics}
\end{equation}
where $p(x_{t+1}\mid x_t,a_t,B_{\mathrm{emb}})$ is the conditional state-transition distribution, including environmental uncertainty and the effects of embodied actuation. The notation $p$ denotes the appropriate conditional distribution in each equation. Deterministic sensing, policies, or dynamics are special cases of these distributions.

Eqs.~\eqref{eq:physical-agent-observation} to \eqref{eq:physical-agent-dynamics} specify the observation, action-selection, and physical-transition relationships, rather than a particular control or learning architecture. Estimation, representation, and any explicit prediction support the policy. Feedback supplies subsequent observations and updates. The resulting interaction is
\begin{equation}
x_t
\rightarrow
o_t
\rightarrow
a_t
\rightarrow
x_{t+1}
\rightarrow
o_{t+1},
\label{eq:physical-agent-closed-loop}
\end{equation}
where arrows indicate the causal and informational progression of one cycle. Actions affect both the physical state and the evidence available for later decisions, coupling sensing and acting throughout the loop.

\subsubsection{Autonomy and the Transition from Automation to Agency}
\label{subsec:physical-agent-autonomy}

\paragraph{Autonomy as a Continuum}
The distinction between a physical agent and a conventional automated system is not binary. Physical systems exhibit different degrees of autonomy, and many useful machines occupy intermediate positions between fixed automation and general goal-directed agency ~\citep{Parasuraman2000}.

In fixed automation, task structure and action sequences are largely specified before deployment. A production system may repeatedly execute the same sequence with high precision, while conventional feedback control compensates for deviations in variables such as position, velocity, force, or temperature. Such feedback is essential to reliable physical operation, but local error correction alone does not imply that the system selects its behavior at the task level.

As sensing, computation, and adaptive control become more capable, physical systems can respond to a wider range of environmental variations. A mobile robot may choose a different route when an aisle is blocked, and a manipulator may adjust its motion when an object is displaced. In such cases, some aspects of behavior are determined online, although the range of possible tasks and responses may remain strongly bounded by predefined representations and skills.

A more agentic system shifts the specification from actions toward outcomes. Rather than being provided with a complete trajectory or procedure, it may receive a target state or higher-level objective and determine an appropriate sequence of actions from the current situation. The progression can be represented schematically as
\begin{equation}
\text{Fixed Automation}
\rightarrow
\text{Feedback Control}
\rightarrow
\text{Adaptive Robotics}
\rightarrow
\text{Goal-Directed Agency}.
\label{eq:automation-to-agency}
\end{equation}
This sequence should be interpreted as an organizing continuum rather than a strict taxonomy. Real systems may combine characteristics from several stages and may exhibit different levels of autonomy in different parts of the same task.

\paragraph{Goal-Conditioned Action Selection}
A central feature of agency is that the same objective can lead to different actions depending on the current physical situation. If a direct route becomes obstructed, a mobile agent may select an alternative path. If an intended grasp is infeasible, a manipulation system may choose another grasp configuration or reposition itself before trying again. The objective remains unchanged while the means of achieving it are revised.

This distinguishes goal-conditioned action selection from a fixed input-output response. The agent's action depends jointly on the current state, the desired outcome, the available capabilities, and the constraints under which it operates.

\paragraph{Adaptation Under Uncertainty}
Physical environments are only partially observable and are rarely perfectly predictable. Objects may move, materials may behave differently from expected models, sensing may degrade, and the effect of an action may differ from its nominal prediction. A physical agent must therefore operate despite uncertainty about both the current state and the consequences of its own actions ~\citep{Kober2013,Brunke2022}.

Adaptation in this context does not necessarily imply online learning of model parameters. It more generally refers to the ability to revise behavior as new evidence becomes available. The agent may update a state estimate, change a planned path, select another skill, reduce speed, request additional observations, or transfer control to a human operator. These mechanisms differ in implementation but serve the same role by preventing execution from being tied irreversibly to assumptions made before interaction.

\paragraph{Outcome-Driven Correction and Recovery}
An important consequence of closed-loop autonomy is that issuing an action is not equivalent to completing it. Physical execution must be assessed in terms of outcomes. A command to close a door is not successful merely because the corresponding motor command was sent. The agent must establish, to the extent required by the task, that the resulting physical state satisfies the intended condition.

When execution does not produce the intended outcome, an agent may need to distinguish among transient error, action failure, environmental change, or an infeasible objective. Recovery can then involve repeating the action, selecting an alternative action, revising the plan, modifying the objective within permitted limits, or requesting external intervention. The ability to make such distinctions becomes increasingly important as tasks extend over longer horizons and involve larger numbers of dependent actions.

\paragraph{Autonomy Does Not Imply Unrestricted Operation}
Autonomy should not be equated with the absence of human control. In practical physical systems, autonomous behavior is typically bounded by authority, safety, and operational constraints. A robot may plan its own route while requiring authorization for a safety-critical manipulation, or it may execute a task independently while requesting human assistance when uncertainty exceeds an acceptable threshold.

For this reason, a physical agent is more appropriately described as an autonomous or semi-autonomous system than as a necessarily fully autonomous machine. Its autonomy concerns the degree to which action selection and adaptation are delegated to the system within a specified operating envelope.

Autonomy should also not be treated as synonymous with intelligence. A narrowly specialized machine may operate with little human intervention while possessing limited ability to generalize beyond a predefined task. Conversely, a system with sophisticated perception, reasoning, or learning capabilities may intentionally operate under substantial human supervision. The development of Physical AI should therefore be considered along multiple dimensions, including autonomy, task diversity, environmental openness, uncertainty, temporal horizon, and the complexity of adaptive behavior.

From this perspective, the transition from automation to physical agency is characterized less by the addition of any single algorithm than by a change in how behavior is organized. The system moves from executing actions specified in advance toward selecting and revising actions in response to goals, physical constraints, and interaction outcomes. This concept provides the foundation for the subsequent discussion of goal representation, perception, world modeling, planning, learning, control, and other capabilities required for a physical agent.

\subsection{Physical Forms of Physical Agent}
\label{pai82:sec:physical-forms}

A physical agent can take different forms, distinguished by how its body provides access to the world. A manipulator changes an object's pose through contact, a mobile robot changes position and viewpoint, and a laboratory installation transforms and measures samples. These differences arise from both bodily morphology and the arrangement of sensing, actuation, computation, and resources. Accordingly, \emph{physical form} denotes the organization through which an agent acts, including systems whose action interfaces are distributed across several devices. Situated robotics and embodied intelligence provide the conceptual foundations for relating behavior to the coupling of controller, body, and environment~\citep{brooks1991intelligence,pfeifer2007self}.

Under the working definition adopted in this chapter, a physical agent is a relatively autonomous, goal-directed unit that perceives physical conditions, selects feasible actions, changes those conditions, and adapts through feedback. A robot is one possible realization of its body. Agency is a property of the complete operational loop, including the executive and its authority over physical interfaces. Software that recommends an action remains a digital agent at that boundary. Attributing physical agency to a software-equipment installation requires including the equipment that executes the action and the measurements used to evaluate it.

\begin{figure}[!b]
\centering
\includegraphics[width=\linewidth]{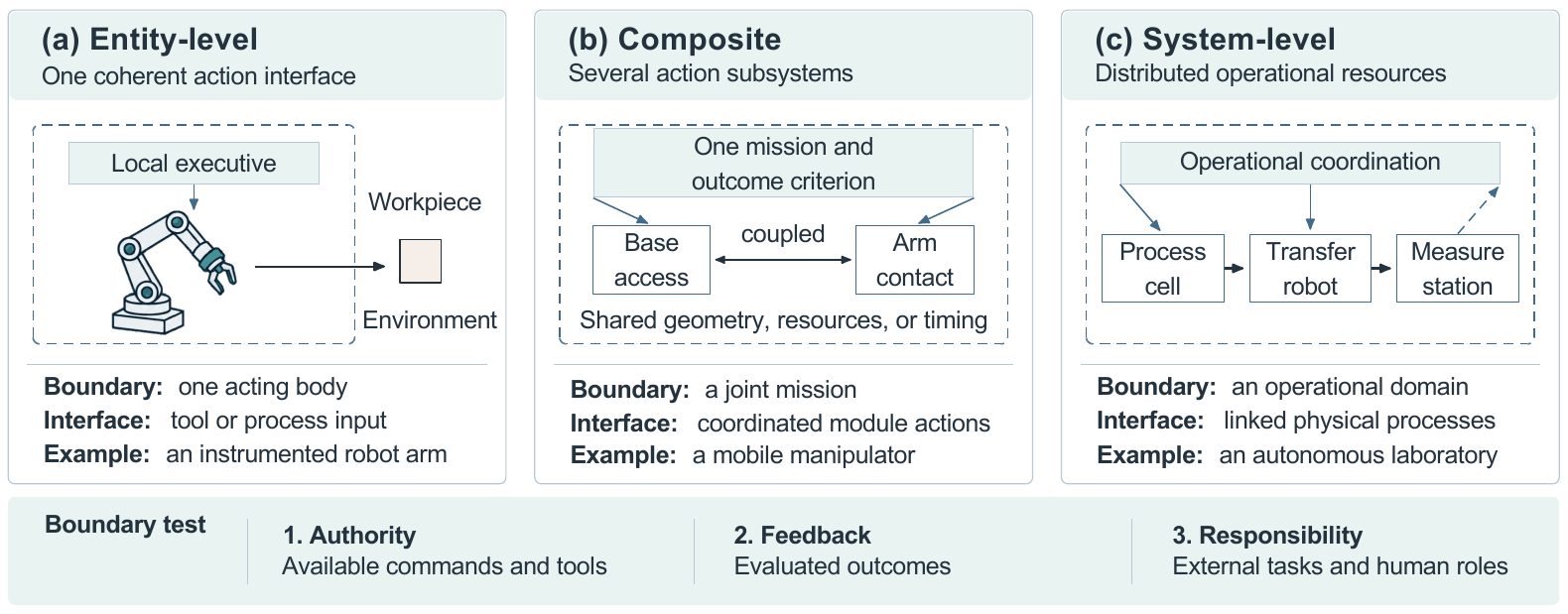}
\caption{Physical forms distinguished by the boundary of attribution, the organization of action, and the scope of the task. Dashed outlines delimit the selected agent. Solid dark arrows denote physical action or flow, green arrows commands, and dashed blue arrows information. The three views can be nested. Original analytical synthesis informed by~\citet{pfeifer2007self,yim2007,monostori2016}}
\label{pai82:fig:taxonomy}
\end{figure}

Figure~\ref{pai82:fig:taxonomy} introduces three complementary views. An \emph{entity-level} view treats a coherent acting body as one unit. A \emph{composite} view resolves functionally separable action subsystems that fulfill a joint mission. A \emph{system-level} view encompasses distributed resources coordinated across an operational domain. This is the chapter's analytical framework, rather than a standardized taxonomy or an autonomy ranking. For example, a mobile manipulator can be one entity in a transport study, a base-arm composite in a manipulation study, and a resource within an autonomous laboratory. The chosen view should follow the capability being explained.

Three criteria make that choice explicit, namely command authority, observable outcomes, and the allocation of responsibility to people or other systems. They distinguish an operational boundary from a building perimeter or a network membership list. Task objects usually remain part of the environment, even while grasped. Materials and mechanisms can themselves support adaptive behavior, as emphasized in physical intelligence research~\citep{sitti2021physical}. The working definition here additionally requires a goal-directed action loop. With this boundary established, physical forms can be examined through the interfaces and constraints that make particular actions possible.

\subsubsection{Entity-Level and Composite Forms of Physical Agent}
\label{pai82:subsec:entity-composite}
\paragraph*{Entity-Level Forms and Their Action Interfaces}

\begin{figure}[htbp]
\centering
\includegraphics[width=\linewidth]{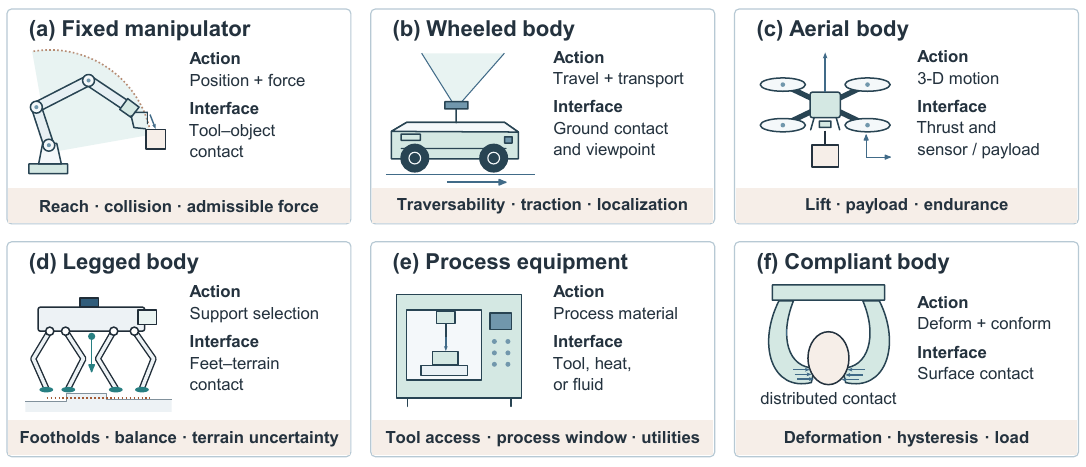}
\caption{Morphology, action interface, and limiting conditions for six representative forms. Shaded reach and sensing regions are qualitative. Arrows indicate motion or interaction directions. The compliant gripper contrasts nominal straight fingers with deformed contact. These original schematics describe physical affordances, not the autonomy of a particular product}
\label{pai82:fig:entities}
\end{figure}

An entity-level physical agent has a coherent body and an identifiable action interface. Mobility, contact geometry, compliance, workspace, payload, and sensing arrangement describe how that body interacts with its surroundings. Figure~\ref{pai82:fig:entities} pairs representative morphologies with the physical variables through which they act.

\paragraph*{Fixed Manipulators}
An arm controls end-effector pose and interaction forces within a bounded workspace. Operational-space control formalizes the relationship between joint actuation and end-effector motion or force~\citep{khatib1987}. Geometric reach is only one requirement. Tool shape, friction, collision clearance, and admissible contact loads also determine whether an operation is feasible. Pairing two arms adds complementary contact roles, such as stabilizing and inserting. ALOHA demonstrates learned fine manipulation using such a bimanual arrangement and human demonstrations~\citep{zhao2023}. At the installation scale the pair can be treated as one manipulation unit. A composite view makes its coordination requirements explicit.

\paragraph*{Ground-Mobile Robots}
Wheeled bodies provide efficient transport where supporting surfaces are traversable. Legged bodies can select footholds across uneven terrain, with balance and contact transitions forming part of the action problem. \citet{miki2022} integrate terrain information with proprioception for quadrupedal locomotion under unreliable perception. Terrain estimates support action through the footholds and dynamics available to the body. Humanoids extend this family with human-like proportions and manipulation interfaces. Resemblance to people is not a criterion for physical agency.

\paragraph*{Aerial Robots}
Flight provides three-dimensional access and changes the sensor's viewpoint without continuous ground support. This capability couples navigation to lift, attitude stabilization, obstacle clearance, and energy availability. \citet{loquercio2021} demonstrate high-speed autonomous flight using onboard sensing and computation. Inspection and delivery build on aerial access but additionally depend on task-specific sensors, payload interfaces, and operating conditions.

\paragraph*{Process Equipment and Compliant Bodies}
Stationary equipment can change geometry, temperature, pressure, or composition through tools and process inputs. Cyber-physical manufacturing architectures connect such equipment to state assessment and adaptive responses~\citep{lee2015}. Soft structures, meanwhile, support conformal interaction by distributing deformation over a body~\citep{rus2015}. Passive dynamic walking offers a complementary demonstration of mechanics contributing to organized motion~\citep{mcgeer1990}.

\paragraph*{Composite Forms: Coordinated Action Across Modules}

A composite physical agent combines functionally separable action subsystems under one mission and outcome criterion. Typical combinations join locomotion with manipulation, ground transport with flight, or sample handling with experimental processing. The analytical emphasis lies on the interfaces between these capabilities. Mechanical attachment is one coupling mechanism. Shared timing, resources, and responsibility can also bind modules that separate in space.

\begin{figure}[!b]
\centering
\includegraphics[width=\linewidth]{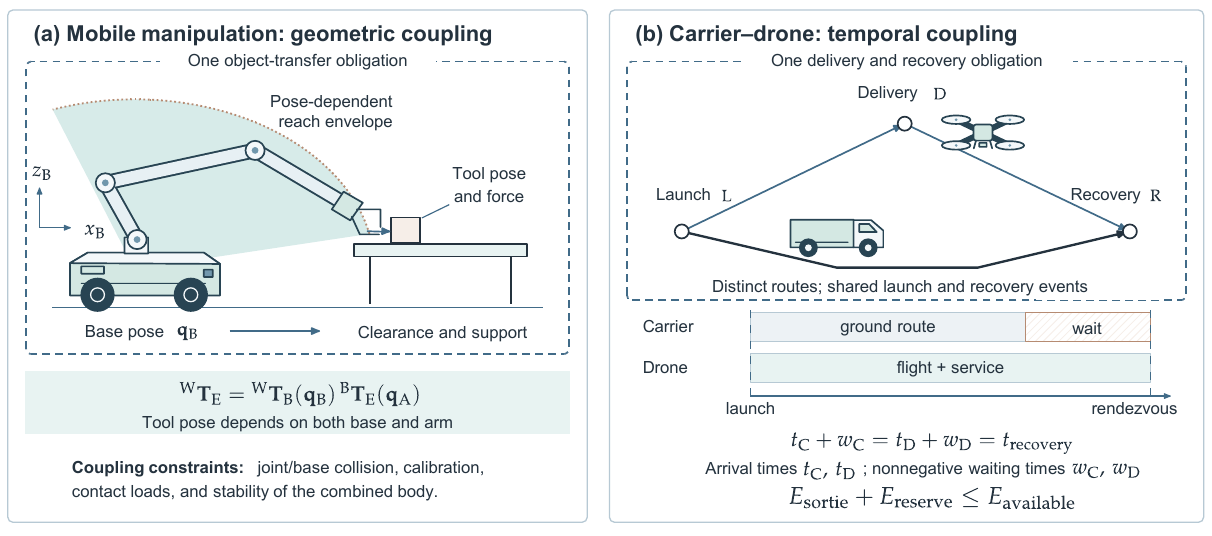}
\caption{Two forms of inter-module dependence. (a) Base configuration $\mathbf{q}_{\mathrm{B}}$ and arm configuration $\mathbf{q}_{\mathrm{A}}$ jointly determine tool pose. $\mathrm{W}$, $\mathrm{B}$, and $\mathrm{E}$ denote world, base, and end-effector frames. (b) Separate routes require compatible recovery times and sufficient flight energy. Symbols within the schematic are defined in the accompanying text. The reach boundary and waiting timeline are qualitative. Original schematics informed by mobile manipulation~\citep{fu2025} and truck-drone routing~\citep{murray2015}}
\label{pai82:fig:composites}
\end{figure}

The schematic in Figure~\ref{pai82:fig:composites} uses $x_{\mathrm{B}}$ and $z_{\mathrm{B}}$ for base-frame axes. For frame labels $\mathrm{F}$ and $\mathrm{H}$, the homogeneous transform ${}^{\mathrm{F}}\mathbf{T}_{\mathrm{H}}$ maps coordinates from $\mathrm{H}$ to $\mathrm{F}$, so ${}^{\mathrm{W}}\mathbf{T}_{\mathrm{E}}={}^{\mathrm{W}}\mathbf{T}_{\mathrm{B}}(\mathbf{q}_{\mathrm{B}}){}^{\mathrm{B}}\mathbf{T}_{\mathrm{E}}(\mathbf{q}_{\mathrm{A}})$ composes the base and arm transformations. In the routing example, $\mathrm{L}$, $\mathrm{D}$, and $\mathrm{R}$ label launch, delivery, and recovery sites. The arrival times $t_{\mathrm{C}},t_{\mathrm{D}}$ and nonnegative waiting times $w_{\mathrm{C}},w_{\mathrm{D}}$ belong to the carrier and drone, respectively. $t_{\mathrm{recovery}}$ is their common recovery time. The terms $E_{\mathrm{sortie}}$, $E_{\mathrm{reserve}}$, and $E_{\mathrm{available}}$ denote required sortie energy, required reserve, and available drone energy.

\paragraph*{Mobile Manipulation}
Base motion extends access beyond a fixed arm's workspace while changing its reference frame, sensing viewpoint, and collision envelope. The pose relation in Figure~\ref{pai82:fig:composites}(a) expresses this geometric dependence. Contact forces and stability introduce additional dynamic constraints. Mobile ALOHA combines a mobile base, two arms, and whole-body demonstration collection to learn coordinated tasks, including cabinet operation and use of a kitchen faucet~\citep{fu2025}. Locomotion establishes access to a work location, while manipulation determines where the base can operate. The cited study evaluates learned execution after demonstration collection, so its teleoperation apparatus and its autonomous policy serve distinct roles.

\paragraph*{Vehicle-Drone Systems}
A carrier can move a drone between launch sites while the drone serves locations through complementary aerial routes. The flying sidekick traveling salesman problem models the resulting route, endurance, and rendezvous dependencies~\citep{murray2015}. Figure~\ref{pai82:fig:composites}(b) makes one consequence visible. Arrival at the recovery site must be synchronized, potentially through waiting. This is an optimization formulation, with human driving of the truck retained in the original problem setting. A composite interpretation becomes useful when one mission executive is responsible for both delivery and recovery. Separate vehicles with independently assigned goals invite a multi-agent interpretation.

\paragraph*{Multi-Device Experimental Platforms}
Dispensing, processing, transfer, and measurement modules can form a composite when they execute a bounded protocol. \citet{burger2020} connect laboratory stations through a dexterous mobile robot in a photocatalyst search. Transporting one vial can be attributed to the robot, whereas conducting the experimental search also requires the stations and experiment-selection software. The example shows how changing the attributed outcome changes the necessary physical boundary.

Composite form can itself be variable. Modular self-reconfigurable robots alter the connectivity and arrangement of their modules~\citep{yim2007}. Self-modeling robots can revise their behavior after bodily damage~\citep{bongard2006selfmodeling}. Capability consequently depends on current connectivity, calibration, module condition, and coordinated skills. This dependence becomes still more prominent when the acting resources form a persistent, distributed installation.

\subsubsection{System-Level Physical Agents}
\label{pai82:subsec:system}
\paragraph*{A Distributed Body with an Operational Boundary}

A system-level physical agent acts through distributed sensors, robots, process equipment, computing nodes, and control networks. Its operational unity comes from an objective and the coordinated use of resources across a domain. Physical distribution and controller organization are separate design choices. A distributed body may use centralized, distributed, or hybrid coordination. Holonic manufacturing provides an early account of autonomous units participating in a larger production organization~\citep{vanbrussel1998}. Cyber-physical production connects computational coordination with physical processes~\citep{monostori2016}.

\begin{figure}[htbp]
\centering
\includegraphics[width=\linewidth]{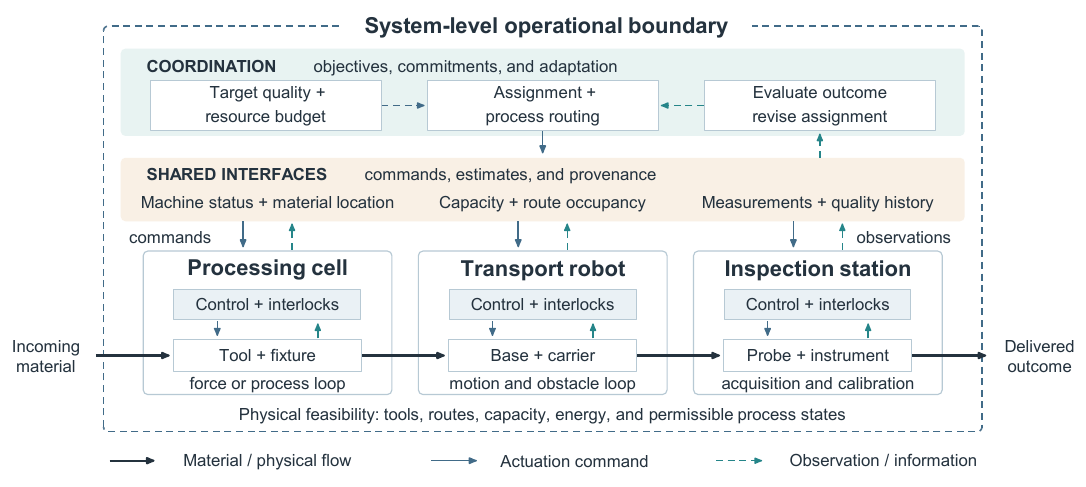}
\caption{Physical and informational organization of a system-level physical agent. Coordination functions operate through shared interfaces, while individual resources retain local feedback and interlocks. Commands, observations, and material flows are distinguished explicitly. The layers describe functional responsibilities and do not prescribe a centralized network architecture. Original synthesis based on holonic and cyber-physical manufacturing~\citep{vanbrussel1998,lee2015,monostori2016}}
\label{pai82:fig:system}
\end{figure}

Two related graphs clarify the organization of such a body:
\begin{equation}
 \mathcal{G}_{\mathrm{phys}}=(V_{\mathrm{phys}},E_{\mathrm{phys}}),\qquad
 \mathcal{G}_{\mathrm{info}}=(V_{\mathrm{phys}}\cup V_{\mathrm{comp}},E_{\mathrm{info}}),
 \label{pai82:eq:graphs}
\end{equation}
where $\mathcal{G}_{\mathrm{phys}}$ and $\mathcal{G}_{\mathrm{info}}$ are the physical-dependency and information-exchange graphs. The resource vertex sets $V_{\mathrm{phys}}$ and $V_{\mathrm{comp}}$ contain sensing/acting and logical computation resources, respectively. Computation may be hosted on the same devices. The edge set $E_{\mathrm{phys}}$ encodes mechanical, material-transfer, or process dependencies. The edge set $E_{\mathrm{info}}$ encodes sensing, coordination, or command exchange. The union $V_{\mathrm{phys}}\cup V_{\mathrm{comp}}$ comprises both resource types. Figure~\ref{pai82:fig:system} illustrates the organization of a system-level physical agent. This descriptive model distinguishes the existence of a communication path from the existence of an executable physical operation. For example, rerouting a part requires a compatible transport path as well as a revised assignment.

Attribution of agency at this scale requires an actionable system objective, authority over the necessary equipment, and measured outcomes that change subsequent interventions. A production objective, for example, becomes operational through coordinated transformation, transport, and quality evaluation. Sensor aggregation alone supplies only part of that loop. Similarly, a fixed command sequence can automate execution while leaving adaptation to an operator. The relevant distinction concerns who selects and revises physical actions when conditions change.

The operational boundary also spans different timescales. Fast contact and motion loops coexist with slower production and experiment cycles. Local stabilization and protective responses must remain effective within the coordination architecture. Human supervisors can set objectives, replenish resources, and authorize exceptional operations, provided these responsibilities are stated. A system-level physical agent can therefore contain local agents and depend on specified human support. What justifies the higher-level description is the additional feedback loop that coordinates resources to achieve an installation-wide outcome. Smart factories, intelligent mines, and autonomous laboratories are representative realizations of a system-level physical agent. 

\paragraph*{Protocols and Standards for a Physical Agent}

For a system-level physical agent, interoperability determines whether independently developed resources can participate in a common course of action. Robots, instruments, and process equipment need shared ways to describe their capabilities, accept delegated work, and report what occurred. Protocols specify these exchanges, while standards establish common information models and requirements against which implementations can be checked. Together, they make the relationships between reasoning, coordination, and physical execution explicit enough to support integration across manufacturers and applications.

A useful organization separates objective formation and revision, task coordination, and local execution. Protocols connect these responsibilities by carrying capability descriptions, task requests, and execution feedback. This separation allows planning methods and equipment implementations to evolve while preserving agreed interfaces. The Model Context Protocol (MCP), for example, specifies how AI applications discover and invoke tools with structured input and output descriptions~\citep{mcp2025}. Applying such interfaces to physical resources requires their descriptions and results to carry the operational meaning needed by the task, including which objects are affected, which conditions permit execution, and what observations establish completion. Anthropic's Model Hardware Standard (MHS), introduced as a research preview in collaboration with HHMI Janelia Research Campus in 2026, addresses the discovery and operation of programmable physical devices.\footnote{Anthropic, \href{https://www.anthropic.com/news/model-hardware-standard-research-preview}{\emph{Previewing the Model Hardware Standard}}, research preview, August 27, 2026.} MHS combines standardized drivers with descriptions of device characteristics and operating limits, and supports access through MCP, command-line interfaces, and code. It thus provides a concrete example of connecting general agent interfaces to laboratory and manufacturing equipment.

The central abstraction is the capability that a resource makes available to the wider system. This may be a measurement, a material-processing operation, a robotic skill, or an established procedure involving several devices. A common description helps an agent determine what can be requested and what information it can expect in return. The Web of Things Thing Description provides one foundation for describing interactions with heterogeneous devices~\citep{w3c2023wot}. At the system level, this suggests a way to organize access around useful functions while retaining the specialized knowledge and control methods within each resource. The same planning system could then work with different combinations of equipment through compatible descriptions.

An autonomous laboratory illustrates how this organization connects reasoning with diverse physical capabilities. In an illustrative materials study, a scientific agent might propose candidate compositions and the properties to be measured. A laboratory coordination system would relate this plan to available preparation, processing, and characterization capabilities. Dedicated instruments would perform established procedures, while robots could contribute manipulation or transport where needed. The protocol would connect these contributions to the same experimental objective and return their results in an interpretable form. The OPC UA Laboratory and Analytical Device Standard supports this kind of integration through common models for device functions, program execution, and results~\citep{opc2023lads}. Here OPC UA denotes OPC Unified Architecture.

The significance of this connection extends to what the agent can learn from an intervention. In the laboratory example, measurements become useful feedback when they remain associated with the candidate material, preparation history, and conditions under which they were obtained. The agent can then compare outcomes, revise an explanation, or select a subsequent experiment. Similar relationships arise when a production system uses inspection results to adjust processing conditions, or an environmental monitoring system directs further measurements toward an uncertain region. Across these settings, a protocol should preserve the context that makes returned observations relevant to the objective. This connects access to physical resources with the continuing development of the agent's understanding.

Shared descriptions also create opportunities for reuse across tasks and facilities. A preparation-and-measurement procedure could be offered as a composite capability for later studies, while its constituent functions remain available for other combinations. Another facility could contribute a complementary measurement or a compatible processing operation. Industrial asset representations such as the Asset Administration Shell provide standardized structures for accessing asset information and services~\citep{iec2023aas}. Research linking these representations with capability models and OPC UA communication explores how such descriptions can support reconfigurable systems~\citep{justmann2024mapping}. Reuse still depends on whether the available methods, materials, and operating conditions fit the new task.

From this perspective, protocols and standards provide shared infrastructure through which physical capabilities can accumulate and be used by different agents. A common core can organize capability discovery, task interaction, and contextualized feedback, while domain-specific descriptions express the knowledge needed in manufacturing, laboratories, or other settings. As models, instruments, and procedures improve, compatible interfaces could allow those advances to benefit multiple workflows. The broader opportunity is to connect independently developed intelligence and physical resources into systems that can pursue objectives, evaluate their interventions, and draw on a growing range of capabilities. Autonomous laboratories offer a concrete setting in which to examine these relationships.

\subsubsection{Correspondence Between Form, Capability, and Application}
\label{pai82:subsec:correspondence}

Reaching an object, exerting a controlled force, and checking the result require different physical provisions. A long reach does not establish that an operation can be performed at every reachable position. A large payload does not establish precise control of contact. Connecting form to application therefore requires examining how the body produces the actions a task needs, and under what conditions it can produce them reliably.

\paragraph*{Geometry and Motion}
Link lengths, joint arrangements, and movement limits determine which positions and orientations a mechanism can reach. Its current posture also matters. Joint motion can produce end-effector motion more readily in some directions than in others, and this relationship changes as the mechanism moves. This posture dependence is captured by manipulability measures~\citep{yoshikawa1985}. For capability assessment, the implication is that access to a position and freedom to act there should be examined separately. The relevant description includes the required working posture and motion direction, alongside collision clearance and load conditions. Manipulability supplies one local kinematic measure. The other requirements need their own checks.

\paragraph*{Contact and Material Response}
Surface geometry, friction, and compliance affect how a body establishes contact and transmits force. Compliance allows a body to deform under contact, sometimes accommodating differences in object shape through its mechanics. A deformable gripping interface, for instance, can conform to an object before stiffening to hold it. Its effectiveness still depends on obtaining suitable contact~\citep{brown2010}. Changing the object or contact conditions can therefore change the result even when the controller remains the same. Thus, describing a body as compliant or adaptable should be accompanied by the object properties and loads for which that behavior is useful.

\paragraph*{Sensing Geometry}
The placement and mobility of sensors determine which information the body can acquire. A camera may have adequate resolution while lacking a useful view of the interaction. Moving the sensor or the body can expose an occluded region. Active perception explicitly connects sensing to such purposeful actions~\citep{bajcsy2018}. Physical form consequently affects both the execution of an action and the possibility of obtaining the feedback needed to guide it. Relevant capabilities include reaching an informative viewpoint, maintaining visibility during movement, and accessing the measurements required to verify the result.

\paragraph*{Relating Task Requirements to Physical Arrangements}

These mechanisms give the form-\allowbreak{}capability-\allowbreak{}task-\allowbreak{}constraint relationship a practical meaning. \emph{Form} specifies the available physical means, \emph{capability} describes the actions and performance those means support, \emph{task} specifies the required effect, and \emph{constraints} establish the conditions that must hold during execution. A useful comparison starts with requirements such as a working region, permissible contact force, positional tolerance, or completion time, and then examines whether a proposed arrangement can satisfy them together.

Table~\ref{pai82:tab:mapping} organizes this comparison into five dimensions. It is an analytical synthesis of the mechanisms and performance-assessment approaches discussed here, rather than a standardized classification. The task determines which dimensions and criteria need evaluation.

\begin{table}[htbp]
\centering
\caption{Connecting task requirements to physical form and evidence of capability. The dimensions provide a framework for comparison, not a single score of intelligence}
\label{pai82:tab:mapping}
\small
\setlength{\tabcolsep}{4pt}
\renewcommand{\arraystretch}{1.14}
\begin{tabularx}{\linewidth}{>{\raggedright\arraybackslash}p{0.20\linewidth}*{2}{>{\raggedright\arraybackslash}X}}
\toprule
\textbf{Task requirement} & \textbf{Physical properties to examine} & \textbf{Evidence of a suitable match}\\
\midrule
Spatial access and motion & Dimensions, degrees of freedom, workspace, support and clearance & Required positions and orientations can be reached and the intended motion can be executed\\
\addlinespace[3pt]
Physical action and load & Tool or process interface, actuation, stiffness, friction and contact geometry & Required loads or process conditions are maintained within the specified tolerances\\
\addlinespace[3pt]
Observation and verification & Sensor placement, field of view, measurement access and calibration & Task-relevant states and outcomes can be measured with sufficient accuracy\\
\addlinespace[3pt]
Sustained operation & Energy supply, thermal limits, replenishment and available capacity & Performance can be maintained for the required duration or rate of work\\
\addlinespace[3pt]
Handover between operations & Physical connections, transfer paths, interface compatibility and shared resources & The output of one operation meets the input conditions of the next\\
\bottomrule
\end{tabularx}
\end{table}

The requirements interact. Extending a mechanism's reach can alter the actuator effort needed for a motion. Adding mobility changes the relationship between the working tool, its support, and the surrounding space. This interaction motivates the joint design of morphology and control. Research on morphology-control co-design makes this relationship explicit by optimizing mechanical dimensions and actuator placement together with motion trajectories and forces, subject to physical and task constraints~\citep{ha2018}. This connects design choices to the behavior needed for a specified task. It also explains why improvements in one hardware specification need to be evaluated in the context of the complete operation.

For a composite or system-level physical agent, the same reasoning extends to the physical connections between resources. Cyber-physical production research situates computational coordination within ongoing physical processes~\citep{monostori2016}. Applying that perspective here, compatible connections between local operations are a prerequisite for system-level capability. A transfer must deliver an object in a state and location that the receiving resource can accept, and shared equipment must be available when required. These conditions must be combined with effective coordination and feedback to achieve the overall task. Communication enables coordination, while physical interfaces and material routes determine which coordinated operations can actually occur. The relevant scale of embodiment follows these dependencies and may take the form of one coherent body, several coupled modules, or a distributed arrangement of resources.

\paragraph*{Comparing Forms Through Demonstrated Performance}

Capability descriptions should distinguish the physical means available, the performance of the implemented agent, and the consistency of that performance. End-effector benchmarking, for example, measures grasp strength, cycle time, and repeatability separately because they answer different questions about application suitability~\citep{falco2020}. At the task level, the relevant evidence may include success rate, completion time, error, and required human intervention. Hardware specifications and task results consequently provide complementary evidence when comparing forms.

The conditions of comparison must also be explicit. The YCB benchmarking framework records the task, environment and object states, hardware, prior information, procedure, and execution constraints~\citep{calli2015}. Following this approach, alternative embodiments should be compared against stated outcome criteria and documented conditions, while allowing each an appropriate means of execution. Repeated trials establish how consistently the implemented system meets those criteria. They also make clear the operating range within which a capability has been demonstrated, rather than assuming that one successful demonstration establishes performance across settings.

The correspondence between form and application is therefore established through a traceable connection. Task requirements identify the necessary physical interactions, bodily organization makes those interactions possible, and evaluation establishes the conditions under which they are achieved. This provides a basis for selecting among entity-level, composite, and system-level forms, and for expressing the physical goals developed in the next section.

\subsection{Physical Goal Representation and Task Grounding}
\label{sec:physical-goals}

A physical agent realizes intended changes in the world through actions constrained by its embodiment and environment. A request such as ``store the sample safely'' must therefore identify an acceptable outcome, admissible interactions, and evidence of completion. This connection applies both to individual robots and to system-level implementations of a physical agent acting through machines, instruments, and control interfaces.

Task realization involves specifying the required outcome, organizing available capabilities to achieve it, and using feedback to assess and revise execution. These functions are examined through the instruction ``Place the sealed sample container in cabinet B, keep it upright, and close the cabinet.'' The example connects goal semantics and task-relevant state representations to feasible behavior and observed results.

\subsubsection{Physical Goal Representation and Semantic Grounding}
\label{subsec:goal-grounding}

\paragraph{From an Instruction to a Physical Specification}
A goal specifies a desired outcome. A plan describes how available actions may achieve it. The container-storage goal should remain meaningful whether executed by a mobile manipulator or by a fixed manipulator supplied through a conveyor. Separating outcomes from procedures supports alternative plans and transfer across embodiments.

For exposition, a physical task can be summarized as
\begin{equation}
    \mathcal{T}_{\mathrm{task}}
    = \left(\mathcal{G}_{\mathrm{goal}},\mathcal{C}_{\mathrm{exec}},\mathcal{P}_{\mathrm{pref}},\mathcal{V}_{\mathrm{verify}}\right),
    \label{eq:task-specification}
\end{equation}
where $\mathcal{T}_{\mathrm{task}}$ is the task specification. Here $\mathcal{G}_{\mathrm{goal}}$ specifies acceptable outcomes, $\mathcal{C}_{\mathrm{exec}}$ constrains execution, $\mathcal{P}_{\mathrm{pref}}$ expresses preferences among acceptable solutions, and $\mathcal{V}_{\mathrm{verify}}$ specifies how outcomes and constraints are assessed. This is an organizing abstraction rather than a prescribed data structure. Reaching the storage region is an outcome. Preserving the seal is a constraint. Minimizing travel time is a preference. Checking placement and closure belongs to verification. Timing and resource requirements may be constraints or preferences, depending on the application.

Let $x_t$ denote the joint physical state of the physical agent and environment at decision index $t$. In this example, $c_{\mathrm{cont}}$ labels the container and $B_{\mathrm{cab}}$ the cabinet. An illustrative Boolean terminal-goal predicate $g$ is
\begin{equation}
\begin{aligned}
    g(x_T) ={}& \operatorname{Inside}(c_{\mathrm{cont}},B_{\mathrm{cab}})
    \land \operatorname{Supported}(c_{\mathrm{cont}})
    \land \operatorname{Released}(c_{\mathrm{cont}}) \\
    &\land \operatorname{Upright}(c_{\mathrm{cont}})
    \land \operatorname{Closed}(B_{\mathrm{cab}}),
\end{aligned}
\label{eq:container-goal}
\end{equation}
where $T$ is the terminal decision index, $x_T$ the corresponding physical state, and $\land$ logical conjunction. Evaluated on $x_T$, the predicates require the container to be inside the intended storage region, physically supported, released by the manipulator, and upright. They also require the cabinet to be closed. Keeping the container upright throughout transport additionally constrains the intervening trajectory. Operational definitions must specify tolerances and evidence appropriate to the task and sensing capabilities. Support and release distinguish completed placement from holding the container at the target.

\paragraph{Complementary Representations and Task Abstractions}
Goals may originate from language, images, demonstrations, geometric designs, or numerical process specifications. They may require a terminal state, sustained conditions, or repeated service. Symbolic descriptions express objects, relations, and logical requirements. Geometric representations specify poses, regions, and clearances. Physical variables describe forces, temperatures, or velocities. Temporal specifications distinguish eventual achievement from requirements that must persist throughout execution~\citep{kressgazit2009temporal}. These descriptions serve different decisions within the same task. A target image communicates arrangement but may omit hidden contact conditions. A learning reward guides optimization but need not fully specify acceptable behavior.

Task-relevant abstractions connect these descriptions to observations. Object properties and relations indicate which conditions currently hold, while action models describe how those conditions may change. Such abstractions can be designed or proposed from language, video, and interaction records, then checked for consistency and against available data. A learned detector may assess whether a cabinet is open, while an action model associates opening with access to its interior. The representation provides an adjustable interface for reasoning, learning, and verification. It need not impose a fixed symbolic vocabulary or a particular architecture. Learning this interface must preserve the task's specified requirements.

\paragraph{Grounding Entities and Conditions in the Current Scene}
Semantic grounding identifies the intended container and cabinet and estimates their task-relevant properties and relations. Recognition alone does not establish that the cabinet is accessible or has sufficient storage space. Even a small observational difference, such as an open versus closed door, changes the conditions from which action must begin. Grounding therefore relates instruction semantics to the current scene and the agent's capabilities.

\begin{figure}[h]
    \centering
    \IfFileExists{figs-new/8-6.pdf}{%
        \includegraphics[width=\linewidth]{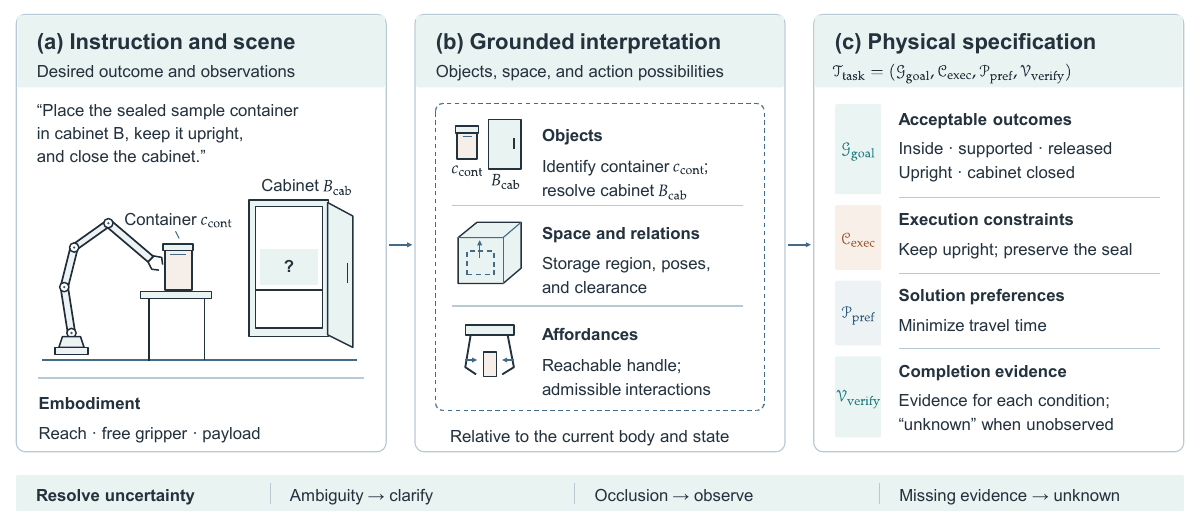}%
    }{%
        \fbox{\parbox[c][3.6cm][c]{0.94\linewidth}{\centering
        Figure placeholder: goal representation and grounding.\\[4pt]
        Instruction and scene observations $\rightarrow$\\
        grounded objects, goal conditions, constraints, and verification.}}
    }
    \caption{Grounding an instruction in task-relevant objects, state conditions, constraints, and completion evidence. Observations and embodiment capabilities inform interpretation. Unresolved uncertainty may require further observation or clarification}
    \label{fig:goal-grounding}
\end{figure}

Language and vision-language models can supply useful priors. SayCan combines language-based skill relevance with skill value estimates grounded in the robot's situation~\citep{ichter2023saycan}. VoxPoser constructs observation-linked spatial value maps to guide motion generation~\citep{huang2023voxposer}. They illustrate different interfaces between semantic interpretation and physical behavior, each relying on appropriate perception and capability estimates. Figure~\ref{fig:goal-grounding} summarizes the grounding process. Action organization is developed in the next subsection.

\paragraph{Uncertainty and Evidence Requirements}
Physical states are partially observed. The belief state $b_t(x)=p(x_t=x\mid o_{0:t},u_{0:t-1})$ represents the conditional probability mass or density of a candidate physical state $x$ at decision index $t$. Here $p$ is the conditional state distribution, $o_{0:t}$ denotes observations from indices $0$ through $t$, and $u_{0:t-1}$ denotes the executed control inputs over preceding transitions. $u_t$ is the control input implementing the selected action $a_t$. This representation distinguishes measured information from inferred conditions and supports information-seeking actions~\citep{Kaelbling1998}. Implementations may instead use confidence-tagged records, alternative hypotheses, or other uncertainty estimates.

If two containers match the instruction, further evidence may be needed to resolve their identity. An occluded cabinet interior may require inspection before placement. Clarification is appropriate when ambiguity materially changes the intended outcome or acceptable risk. Observation should resolve the uncertainty relevant to the next decision.

Verification requirements must reflect what can be established from evidence. A visually closed door does not establish latching without suitable observations. Conditions should therefore admit an ``unknown'' assessment when evidence is insufficient. The specification must identify both the required conditions and the evidence that would justify claiming their satisfaction.

\subsubsection{Task Decomposition and Action Organization}
\label{subsec:task-organization}

\paragraph{A Functional Hierarchy from Goals to Control}
Task realization connects decisions at different abstractions and timescales. A useful functional hierarchy is
\begin{equation}
\begin{gathered}
    \text{goal specification}
    \longrightarrow \text{subgoals and dependencies}
    \longrightarrow \text{skills} \\
    \longrightarrow \text{motion or process references}
    \longrightarrow \text{control inputs}.
\end{gathered}
\label{eq:action-hierarchy}
\end{equation}
Upper levels determine desired outcomes and their dependencies. Intermediate levels select skills and parameters. Lower levels generate movement or process evolution. These are functional responsibilities. A learned policy may combine several levels, while a system-level physical agent may refine tasks into instrument commands rather than joint trajectories.

For the storage example, decomposition includes preparing access, acquiring and transporting the container, placing it, and closing the cabinet. The order depends on capability and resource availability. A single-gripper robot may need to open the cabinet before acquiring the container or use an intermediate support surface. An additional manipulator permits other arrangements. Decomposition therefore accounts for resources occupied by each action and conditions needed by subsequent actions.

\paragraph{Task Planning and Dependency Structure}
Task planning uses action conditions and predicted effects to identify dependencies and construct plans, as exemplified by the Stanford Research Institute Problem Solver (STRIPS)~\citep{fikes1971strips}. Hierarchical decomposition organizes recurring procedures into reusable task structures. The planning model should reflect the agent's available skills and the current environment, so that task steps correspond to capabilities the agent can invoke.

A plan need not impose a total order. Access must precede insertion, and the manipulator must leave the door's swept region before closure, whereas independent operations may be reordered or run concurrently when resources permit. Logically independent steps may still compete for the same manipulator or instrument. Conditional branches accommodate initially unknown states, such as whether the cabinet is open. The resulting structure separates persistent dependencies from ordering choices that vary with initial conditions and resources.

\paragraph{Coupling Task Planning with Physical Feasibility}
Grasp choice, approach configuration, and placement pose affect which task sequences are feasible. A grasp may obstruct passage through the opening, or a placement may leave insufficient room for withdrawal. Task and motion planning (TAMP) addresses these interactions by coupling discrete action and contact choices with continuous configurations, grasps, and trajectories~\citep{garrett2021integrated}.

The coupling is bidirectional. A task planner proposes actions and required conditions, while geometric and physical reasoning supplies feasible parameters or revises those choices. PDDLStream connects symbolic planning to black-box procedures that generate values satisfying continuous constraints~\citep{garrett2020pddlstream}. Failure to find suitable parameters can prompt a different grasp, approach, or task sequence. These parameters must support the entire relevant interaction, including approach, contact, transport, and release, rather than merely reaching an isolated pose.

\begin{figure}[tbp]
    \centering
    \IfFileExists{figs-new/8-7.pdf}{%
        \includegraphics[width=\linewidth]{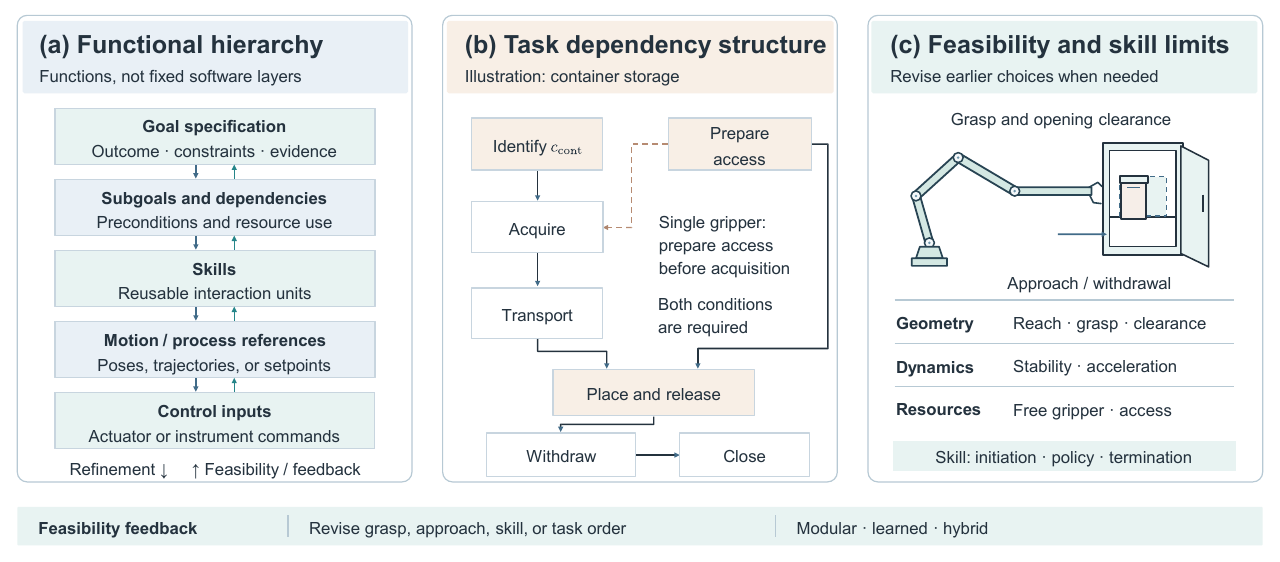}%
    }{%
        \fbox{\parbox[c][3.8cm][c]{0.94\linewidth}{\centering
        Figure placeholder: hierarchical action organization.\\[4pt]
        Goal $\leftrightarrow$ task graph $\leftrightarrow$ skills\\
        $\leftrightarrow$ motion/process planning $\leftrightarrow$ control.\\[4pt]
        Downward refinement, upward feasibility and execution feedback.}}
    }
    \caption{Hierarchical action organization with downward task refinement and upward feasibility feedback. Geometric, dynamic, and resource constraints can revise higher-level choices within modular, learned, or hybrid implementations}
    \label{fig:action-hierarchy}
\end{figure}

Relevant constraints extend beyond collision avoidance. Container transport requires stability and acceleration limits. Door manipulation involves contact and changing geometry. Manufacturing and laboratory workflows similarly couple operation ordering to machine availability, instrument access, and admissible process or sample states. These conditions must be represented where they affect planning. Geometric feasibility alone cannot capture them all. Figure~\ref{fig:action-hierarchy} summarizes this organization.

\paragraph{Skills as Reusable Units of Interaction}
Skills such as grasping, opening, docking, and measuring encapsulate temporally extended interactions. In the options framework, a skill comprises an initiation set, an internal policy, and a termination rule~\citep{sutton1999options}:
\begin{equation}
    \sigma_{k}^{\mathrm{opt}} = (\mathcal{I}_{k}^{\mathrm{init}},\pi_k^{\mathrm{skill}},\beta_{k}^{\mathrm{term}}),
    \label{eq:skill-option}
\end{equation}
where $\sigma_{k}^{\mathrm{opt}}$ denotes skill $k$ in the skill library, $\mathcal{I}_{k}^{\mathrm{init}}$ is its set of admissible starting states, $\pi_k^{\mathrm{skill}}$ its internal action policy, and $\beta_{k}^{\mathrm{term}}$ its state-dependent termination probability in $[0,1]$. Practical interfaces may expose parameters, expected effects, operating limits, and completion or failure statuses. Such interfaces support composition independently of whether a skill is programmed or learned.

Composition requires compatible boundary conditions. Grasping must leave the container suitable for transport. Placement must preserve access for withdrawal and closure. Skill quality includes its contribution to subsequent actions. Behavior trees organize conditions, skills, and alternative branches into a modular execution structure~\citep{ColledanchiseOgren2018}, providing one way to assemble reusable capabilities into longer tasks.

\paragraph{Learning, Adaptation, and Hybrid Organization}
Learning can acquire skills from demonstrations or interaction, estimate skill parameters, and support action selection. It can also be used to adapt a planner to a particular task domain and skill library. Learning a planner concerns the selection and organization of actions, whereas learning a skill concerns how an interaction is executed. The two can share experience while retaining distinct objectives. Such adaptation should preserve the relationship between proposed steps, available capabilities, and task constraints.

Generative policies offer expressive representations for skill execution. Diffusion Policy produces observation-conditioned action sequences with receding-horizon updates~\citep{chi2023diffusion}. OpenVLA connects visual observations and language instructions to learned robotic actions~\citep{kim2024openvla}. Their use in new settings depends on data coverage and compatibility with the agent's embodiment and operating conditions. Adaptation may therefore involve updating selected components rather than redesigning the entire hierarchy.

Hybrid organization assigns complementary functions to planning and learned behavior. A planner can select subgoals and skills, while a learned policy handles contact-rich interaction. Code as Policies illustrates another interface by generating programs that compose perception and control functions~\citep{liang2023code}. The allocation of functions should follow task demands and the evidence available for validating each component.

\subsubsection{Closed-Loop Execution and Failure Recovery}
\label{subsec:closed-loop-realization}

\paragraph{Execution as Repeated State-Dependent Commitment}
Execution tests a plan's predictions against a changing world. At each decision point, the physical agent updates its state estimate and task context, then decides whether to continue, obtain information, adjust behavior, or replan. A compact representation is
\begin{equation}
\begin{aligned}
    b_{t+1} &= \operatorname{Update}(b_t,u_t,o_{t+1}), \\
    d_{t+1}^{\mathrm{skill}} &= \operatorname{Supervise}(b_{t+1},\mathcal{T}_{\mathrm{task}},\zeta_{t}^{\mathrm{exec}}),
\end{aligned}
\label{eq:execution-loop}
\end{equation}
where $t$ indexes decision steps, $b_t$ is the task-relevant belief state, $u_t$ is the control applied during step $t$, and $o_{t+1}$ is the resulting observation. The function $\operatorname{Update}$ revises that belief, while $\operatorname{Supervise}$ selects the next supervisory decision $d_{t+1}^{\mathrm{skill}}$ using the task specification $\mathcal{T}_{\mathrm{task}}$ and execution context $\zeta_{t}^{\mathrm{exec}}$. This context records the active skill, verified subgoals, and relevant constraint history. The decision may authorize continuation, information gathering, recovery, replanning, or termination. History matters because terminal observations cannot reveal every earlier constraint violation. If another actor closes the cabinet during transport, the remaining plan must accommodate the changed access condition.

Feedback operates at several timescales. Controllers regulate physical variables, skills assess progress and termination, and task-level supervision revises dependencies. Model predictive control repeatedly solves a finite-horizon problem and executes an initial portion before updating it~\citep{mayne2000constrained}. Inner Monologue uses success feedback, scene descriptions, and human feedback to inform language-based task planning~\citep{huang2023inner}. These levels provide complementary corrections. Timely control feedback regulates execution, while task supervision assesses its relevance to the goal.

\paragraph{Monitoring Effects and Diagnosing Discrepancies}
Monitoring tests whether prerequisites remain valid, expected changes occur, constraints hold, and goals are achieved. Evidence may include images, joint states, contact measurements, instrument readouts, and elapsed time. Gripper closure does not establish acquisition, a placement pose does not establish support, and a door-close command does not establish closure. The relevant checks concern physical effects.

Diagnosis distinguishes semantic mistakes, state-estimation errors, infeasible plans, skill failures, and external changes. A small alignment error may be corrected locally. Persistent lack of progress may indicate a blocked mechanism or an incorrect condition estimate. Repeated discrepancies also warrant checking the state abstractions and action models used for planning. Execution records can support revisions to initiation conditions or effect predictions, which should be validated before reuse.

\begin{figure}[tbp]
    \centering
        \includegraphics[width=\linewidth]{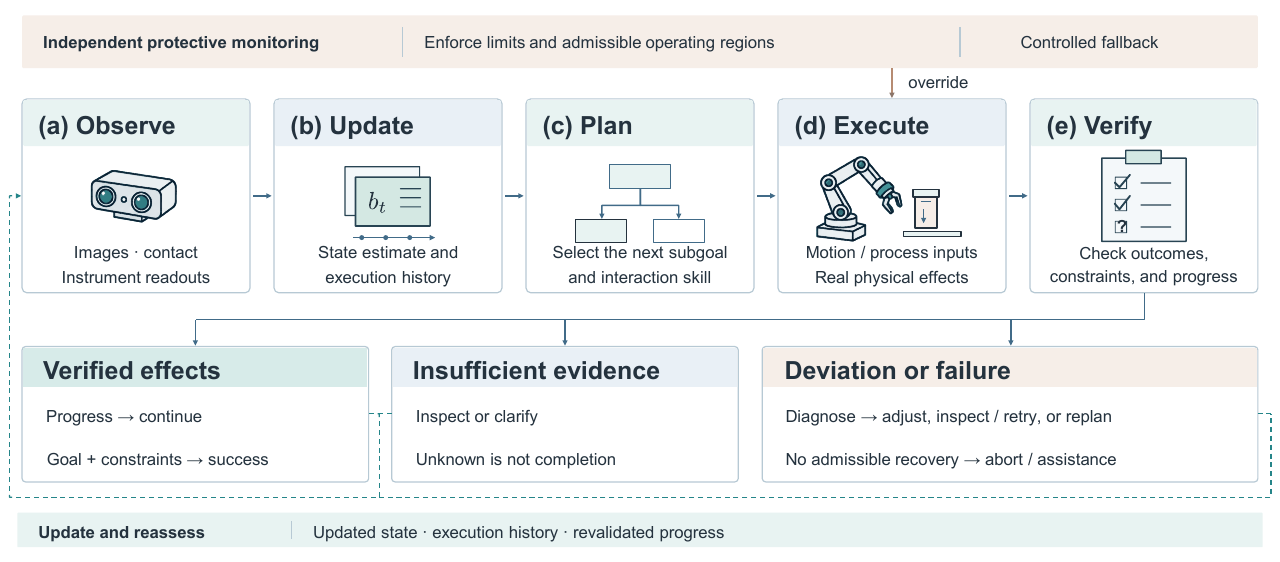}%
    \caption{Observed effects update state estimates and execution context. Verified progress supports continuation. Discrepancies trigger diagnosis and recovery. Insufficient evidence prompts further observation. Protective monitoring can interrupt the loop independently of task planning}
    \label{fig:closed-loop-recovery}
\end{figure}

Monitoring itself is uncertain. Confirmation over time, complementary sensors, and explicit ``unknown'' assessments can reduce false alarms and missed detections. Confirmation must nevertheless remain compatible with timely protective action. Figure~\ref{fig:closed-loop-recovery} relates these checks to subsequent decisions.

\paragraph{Recovery as Constrained Continuation}
Recovery addresses the diagnosed cause while preserving progress that remains valid. Responses range from local correction and renewed observation to another grasp, repositioning, or task-level replanning. Each response starts from the current physical state. A failed placement may leave the container tilted or partially supported, so the conditions for retrying must be reassessed. Previously verified subgoals must also be rechecked when subsequent actions may have disturbed them.

Physical actions can consume resources or cause irreversible changes. Retries should therefore depend on remaining resources, acceptable risk, and evidence that another attempt can address the cause. If no admissible recovery exists, controlled termination or assistance may be required. Changing the destination or relaxing upright transport changes the task specification and requires authorization unless such alternatives were already permitted.

\paragraph{Safety and Bounded Autonomy}
Protective mechanisms constrain commands and monitor operating limits independently of unrestricted task-level decisions. Control barrier functions provide one approach to enforcing safety-related state constraints under modeling and feasibility assumptions~\citep{ames2019control}. Their guarantees depend on those assumptions, including valid state estimates and available corrective control. Safety requirements therefore need explicit physical conditions and appropriate protection.

Fallback behavior must suit the embodiment and situation. A loaded manipulator may need to maintain support. A moving platform may require controlled deceleration. The scope of autonomy should specify when task execution yields to protective control or an operator, as well as which actions may be initiated.

\paragraph{Execution Evidence and Broader Evaluation}
Recorded observations, verified subgoals, constraint violations, and interventions preserve the evidence behind execution decisions. Chapter~\ref{ch:evaluation} develops their use in systematic task evaluation, capability comparison, and benchmarking. Here, verification supplies the evidence needed to continue, recover, or terminate the current task.

The connection between task specifications, capability-dependent plans, and execution evidence also underpins the coordination among multiple locally autonomous participants discussed next.

\subsection{Multi-Agent Collaboration and Distributed Physical AI}

Section~\ref{pai82:subsec:system} describes a \emph{system-level physical agent} through its operational boundary. Distributed sensors, robots, equipment, and computing resources act toward a system objective, with measured outcomes informing subsequent interventions. This section examines the \emph{collaboration among multiple physical agent instances} that retain local sensing, decision-making, and action capabilities. The distinction concerns the level at which agency and coordination are analyzed. System-level agency does not require every component to be autonomous, whereas multi-agent collaboration concerns the relationships among locally autonomous participants. A laboratory or production installation may therefore be a system-level physical agent as a whole and a multi-agent organization internally. Centralized task allocation is compatible with local agency. An instrument accessible through an application programming interface (API) that only executes commands can instead remain a supporting resource.

\emph{Distributed Physical AI} denotes the collective ability of agents to combine observations, complementary capabilities, and coordinated actions across a shared physical environment~\citep{parker2007distributed,rizk2019cooperative,wu2025generativeEMAS}. Coordination must reconcile local autonomy with collective commitments when decisions are coupled through space, time, resources, and the effects of physical action. Communication supports the exchange of observations and coordination information~\citep{zhou2023racer}. Shared infrastructure and observed actions also inform coordination. Humans may contribute objectives, expertise, supervision, and intervention. Building on the individual-agent mechanisms in Sections~\ref{sec:physical-agent-definition} and~\ref{sec:physical-goals}, the following discussion focuses on the distribution of decision authority, inter-agent coordination, and the capabilities and limitations of the resulting teams.

\subsubsection{Organizational Architectures for Distributed Physical AI}
\label{subsec:distributed-architectures}

Consider a partially collapsed tunnel in which a heterogeneous robot team searches for trapped workers and delivers supplies. An aerial vehicle surveys accessible openings. A quadruped traverses rubble and carries relay equipment. A mobile manipulator transports supplies and removes small obstacles where feasible. Their tasks depend on shared access routes, communication coverage, and complementary capabilities. Organizing the team requires specifying where coordination state, computation, and decision authority reside~\citep{yan2013survey}. Four representative architectures are centralized, hierarchical, decentralized, and hybrid.

\paragraph{Centralized Organization}
A central node aggregates reported state and performs team-level allocation, resource coordination, or joint planning. It may combine automated planning with human supervision. In the tunnel example, it assigns search, relay, and delivery responsibilities using the available map and robot status. Agents retain local execution authority within these assignments. The center's decisions depend on the coverage and freshness of the information it receives.

\paragraph{Hierarchical Organization}
Coordination responsibilities are divided across mission, team, and agent levels. A mission coordinator sets rescue priorities. Team coordinators organize operations within assigned regions. Agents execute their responsibilities locally. Decisions that affect several teams are escalated, while regional details remain within the relevant team. Effective hierarchy requires explicit interfaces between levels and sufficient information to recognize cross-team dependencies.

\paragraph{Decentralized Organization}
Agents coordinate without a single team-level decision center, using local observations and information from other agents. After discovering a trapped worker, an explorer can request assistance. Candidates exchange capabilities and costs and agree on responsibility. Coordination may use negotiation, consensus, or distributed optimization. Agents must track the validity of agreements and revise or suspend them when information or connectivity changes.

\begin{figure}[tb]
    \centering
    \IfFileExists{figs-new/8-9.pdf}{%
        \includegraphics[width=\linewidth]{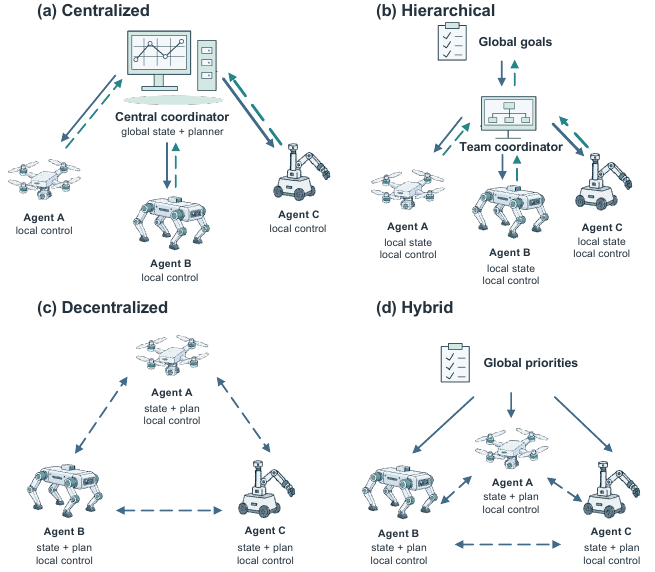}%
    }{%
        \fbox{\parbox[c][3.8cm][c]{0.94\linewidth}{\centering
        Figure placeholder: organizational architectures.\\[4pt]
        Centralized / hierarchical / decentralized / hybrid.\\[4pt]
        Distribution of state, computation, and decision authority.}}
    }
    \caption{Four organizational architectures for distributed Physical AI, comprising centralized, hierarchical, decentralized, and hybrid}
    \label{fig:distributed-architectures}
\end{figure}

\paragraph{Hybrid Organization}
Different functions or operating conditions use different arrangements. A team may build maps cooperatively while a mission center assigns priorities, or negotiate passage access locally while following centralized scheduling. If communication is interrupted, agents may continue within agreed limits. Such changes require rules for transferring authority and reconciling outstanding commitments when coordination is restored. Figure~\ref{fig:distributed-architectures} summarizes the four forms.

Table~\ref{tab:distributed-architectures} compares their typical tradeoffs. Actual performance depends on task coupling, network conditions, and the implementation of coordination protocols.

\begin{table}[!t]
    \centering
    \small
    \setlength{\tabcolsep}{3pt}
    \renewcommand{\arraystretch}{1.08}
    \caption{Typical tradeoffs among organizational architectures for distributed Physical AI}
    \label{tab:distributed-architectures}
    \begin{tabularx}{\linewidth}{@{}>{\raggedright\arraybackslash}p{0.15\linewidth}*{4}{>{\raggedright\arraybackslash}X}@{}}
        \toprule
        Dimension & Centralized & Hierarchical & Decentralized & Hybrid \\
        \midrule
        Global coordination
        & Joint decisions from reported state, with sensitivity to missing or stale information.
        & Cross-team coordination through upper levels, where abstraction can hide dependencies.
        & Global constraints require agreements across local views.
        & Coordination concentrated on selected global dependencies. \\
        \addlinespace
        Communication dependence
        & Requires access to the center for team-level decisions.
        & Depends on links within teams and between levels.
        & Depends on the connectivity required by local protocols.
        & Varies by function and operating mode. \\
        \addlinespace
        Scalability
        & Central computation and reporting can become bottlenecks.
        & Grouping reduces load, while cross-team coupling adds overhead.
        & Sparse interactions scale more readily than strongly coupled tasks.
        & Flexible grouping adds management complexity. \\
        \addlinespace
        Response time
        & Team decisions incur center communication and processing delays.
        & Local decisions are fast, while escalation adds delay.
        & Local response is fast, while negotiation may delay agreement.
        & Depends on authority transfer and coordination mode. \\
        \addlinespace
        Fault tolerance
        & Center failure requires redundancy or local fallback.
        & Fault isolation depends on backup paths and coordinators.
        & Avoids a sole center, though partitions can disrupt commitments.
        & Alternative modes help when transitions are reliable. \\
        \bottomrule
    \end{tabularx}
\end{table}

These architectures describe runtime coordination authority. When coordination policies are learned, centralized training with decentralized execution (CTDE) describes a separate choice about information available during learning and deployment~\citep{amato2024introduction}. Training may use global state or centralized value estimates, while deployed policies act on locally available information, potentially including communicated messages. CTDE alone does not determine the team's runtime allocation or coordination architecture.

\subsubsection{Coordination Mechanisms for Distributed Physical AI}
\label{subsec:distributed-coordination}

Organizational architecture determines where coordination decisions are made. Coordination mechanisms determine how agents align their objectives, information, and actions. The central issue is how decisions made by different agents remain compatible when they depend on shared resources or on one another's behavior.

\begin{figure}[!b]
    \centering
    \IfFileExists{figs-new/8-10.pdf}{%
        \includegraphics[width=0.7\linewidth]{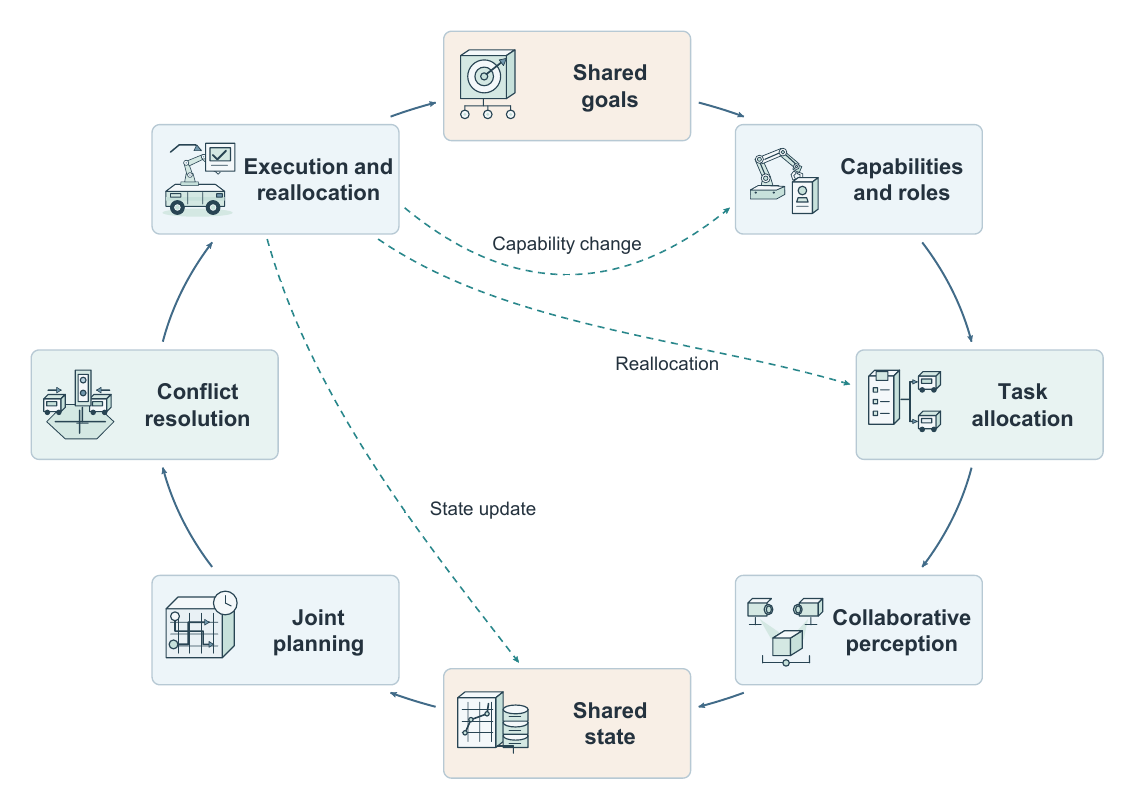}%
    }{%
        \fbox{\parbox[c][4.2cm][c]{0.94\linewidth}{\centering
        Figure placeholder: interdependent coordination mechanisms.\\[4pt]
        Shared goals, capabilities and roles, task allocation\\
        Collaborative perception, shared state, joint planning\\
        Conflict resolution, execution and reallocation\\[4pt]
        Information exchange and changing conditions connect the mechanisms.}}
    }
    \caption{Eight interdependent mechanisms of multi-agent collaboration. The mechanisms may operate concurrently, exchanging information and revising decisions as task conditions and agent capabilities change}
    \label{fig:distributed-coordination}
\end{figure}

Eight interdependent mechanisms organize this discussion, namely establishing shared goals, capability and role modeling, task allocation, collaborative perception, shared state, joint planning, conflict resolution, and execution and reallocation~\citep{rizk2019cooperative,wu2025generativeEMAS}. They operate concurrently and revise one another as conditions change. Their ordering in the figure is schematic. Figure~\ref{fig:distributed-coordination} summarizes their relationships.

\paragraph{Establishing Shared Goals}
Agents need a common understanding of overall objectives, priorities, completion criteria, and collaboration protocols~\citep{kannan2024smart,obata2024lip}. Their assigned subgoals can differ while contributing to the same outcome. In tunnel rescue, searching for trapped people, maintaining communication, and delivering supplies must be coordinated according to agreed priorities. A human decision to prioritize confirmation of vital status over delivery should propagate to affected agents, so that their local objectives remain consistent with the revised mission.

\paragraph{Capability and Role Modeling}
Coordination requires descriptions of what each agent can currently contribute. These combine configuration-dependent capabilities, such as locomotion, sensing range, and payload, with changing conditions, including battery energy, communication connectivity, equipment, and health~\citep{chen2025emos}. Roles express how those capabilities serve the mission. A quadruped carrying relay equipment may maintain communication rather than search, while another takes over exploration. Distinguishing capability from role allows responsibilities to change without assuming that robots of the same category are interchangeable.

\paragraph{Task Allocation}
Allocation assigns tasks to agents or temporary groups, accounting for capability matching, dependencies, energy, risk, and completion time~\citep{gerkey2004allocation}. A delivery task may require both a carrier and a relay, making the assignment a joint choice rather than two independent selections. Centralized optimization, auctions, negotiation, and distributed optimization offer different ways to establish these assignments. Allocation also determines which agents remain available for subsequent demands. Minimizing the immediate travel distance can therefore conflict with preserving scarce capabilities for later tasks.

\paragraph{Collaborative Perception}
Collaborative perception combines complementary observations and coordinates sensing effort~\citep{xu2024d2slam}. An aerial vehicle can survey the tunnel structure, while ground robots inspect occluded regions and assess traversability. Agents can divide observation responsibilities according to uncertainty and information value, reducing unnecessary duplication. Time synchronization, coordinate transformations, and target association support observation fusion across viewpoints~\citep{hu2022where2comm}. Fusion should preserve relevant uncertainty and identify conflicting observations.

\paragraph{Shared State}
Shared state maintains information needed for joint decisions, including maps, targets, agent locations, resource occupancy, and task progress. Unlike collaborative perception, which concerns acquiring and combining observations, this mechanism concerns distributing and updating the resulting operational information. State may reside centrally or in local copies. Limited bandwidth favors task-relevant abstractions over continuous transmission of raw sensor data~\citep{chang2023hydra}. Agents need sufficiently current and consistent information about the resources and dependencies on which their decisions rely. Identical knowledge of the entire environment is unnecessary. The required consistency depends on how tightly the agents' actions are coupled.

\paragraph{Joint Planning}
Joint planning addresses dependencies among agents' actions in space, time, and resource use. Two robots sharing a narrow passage must coordinate traversal order. A delivery robot may depend on a relay reaching its coverage position first. Plans can be generated centrally or reconciled through iterative negotiation~\citep{mandi2024roco}. Coordination can concentrate on strongly coupled near-term actions while leaving independent actions locally planned. This limits the need to fix every agent's full trajectory while retaining explicit commitments where another agent's progress depends on them.

\paragraph{Conflict Resolution}
Conflicts concern spatial occupancy, schedules, power, bandwidth, or incompatible task assignments. They may arise from competing objectives, outdated information, or deviations from agreed plans. Priorities, reservations, negotiation, and replanning can resolve resource contention. Protective intervention may be required for imminent physical conflicts~\citep{zhang2025gcbf+,lee2025merry}. Resolution must propagate changes to affected commitments so that other agents can revise their dependent actions.

\paragraph{Execution and Reallocation}
Execution reports reveal whether assignments and inter-agent commitments remain viable. Agents share progress, exceptions, and remaining resources when these affect collaborators. A blocked route, lost connection, or unavailable executor may require transferring a task or reorganizing a group~\citep{liu2025coherent}. Reallocation should account for work already completed, current resource reservations, and dependencies on other agents. Updating roles and shared state alongside the assignment prevents a replacement agent from duplicating completed work or continuing under obsolete assumptions.

\subsubsection{Open Problems and Trends in Distributed Physical AI}
\label{subsec:distributed-open-problems}

Coordination can extend the coverage, throughput, and range of tasks achievable by a team. These gains depend on how sensing, task dependencies, resources, and failures are managed across its members.

\paragraph{Collective Capabilities}
Complementary sensing and coordinated revisits extend spatial coverage and the frequency of environmental updates~\citep{francos2023role}. Independent subtasks can proceed concurrently, while ordered operations can be pipelined. Under the idealized assumption of equal per-participant execution rates, let $W_{\mathrm{work}}$ denote total workload measured as the time required by one physical agent, $N_{\mathrm{agent}}\geq1$ the number of participating instances, $L_{\mathrm{crit}}$ the duration of the critical path induced by task dependencies, and $T$ the team's completion duration. In this bound, $W_{\mathrm{work}}$, $L_{\mathrm{crit}}$, and $T$ use the same time unit, and
\begin{equation}
    T \geq \max\left(\frac{W_{\mathrm{work}}}{N_{\mathrm{agent}}},L_{\mathrm{crit}}\right).
    \label{eq:distributed-completion-bound}
\end{equation}
This lower bound exposes the limits of parallelism~\citep{brent1974parallel}. Physical contention and coordination overhead can further restrict the benefit of additional agents. Heterogeneous capabilities also expand the feasible task set. Aerial reconnaissance, ground access, manipulation, and infrastructure support can form task chains unavailable to any member alone. Alternative executors and routes can preserve service after failures~\citep{bossens2022resilient}, but resilience requires substitutes for critical tools, capabilities, and communication links. Realizing these gains raises the following open problems.

\paragraph{Communication, Information Freshness, and Consistency}
Communication range changes with robot motion and environmental conditions, while bandwidth, delay, and packet loss constrain information exchange~\citep{gielis2022critical}. Collaboration requires deciding which observations, task commitments, and resource states to share, and when their expected benefit justifies transmission. Agents must also track whether information remains valid when making coupled decisions. A delayed passage-occupancy update can be more consequential than an incomplete distant map. Maintaining sufficient consistency requires task-specific priorities and rules for conflicting or outdated updates. Because motion changes network topology, communication and physical planning may need to be optimized jointly.

\paragraph{Cross-Embodiment Knowledge and Semantic Interfaces}
Different observation spaces, action spaces, and dynamics prevent direct transfer of many policies between embodiments. Cross-embodiment learning provides useful experience~\citep{openx2024}, but collaboration additionally requires alignment of object identities, task states, capability descriptions, and expected skill outcomes. An aerial vehicle and a mobile manipulator may interpret the same destination while requiring different access conditions and control policies. Shared interfaces should expose these differences so that task assignment and composition remain meaningful. Compatible task semantics must therefore support coordination while preserving embodiment-specific actions and requirements.

\paragraph{Collaborative Safety and Runtime Assurance}
Individually acceptable actions can conflict through shared space, resource occupancy, or coupled task dependencies. Updating one learned policy can also change the operating conditions encountered by other agents~\citep{garg2024SafeMultiRobot}. Assurance must therefore address joint behavior, including the validity of reservations and commitments when communication is delayed or an agent fails. Local protective responses need to respect team interactions. Stopping one robot may obstruct another or interrupt a jointly supported load. Coordination protocols should specify how such responses are communicated and how affected commitments are revised, including the respective authority of agents, coordinators, and human supervisors.

\paragraph{System-Level Evaluation}
Embodied multi-agent benchmarks provide a basis for assessing collaborative task performance~\citep{chang2025partnr}. Evaluation should also measure the benefit and cost of coordination, including team completion time, duplicated work, workload distribution, communication demand, inter-agent conflicts, and recovery after member failures. Tests should vary team size, embodiment combinations, connectivity, and capability availability. Comparisons with suitable single-agent or independently operating baselines help identify gains attributable to coordination. Recovery results should distinguish loss of a replaceable member from loss of a unique capability. These protocols reveal whether performance reflects effective distributed collaboration and whether it persists as the team changes.

The following directions seek to address these limitations.

\paragraph{Shared World Models}
Shared world models extend exchanged state records toward predictions of how one agent's actions affect other agents and the environment~\citep{zhang2025combo}. For example, moving a relay robot changes the communication conditions under which other agents can proceed. The challenge is to coordinate local models with different spatial resolutions and prediction horizons without requiring a single central model. Their predictions should expose coupled constraints and support comparison of alternative joint actions. Learning these interactions from team experience could improve planning when capability, resource, and communication dependencies cannot be handled independently.

\paragraph{Multi-Agent Foundation Models}
Learned coordination components, including collaborative perception~\citep{blumenkamp2025covis}, suggest a route toward broader models of team behavior. Models trained across tasks and team configurations could acquire priors about capability differences, task dependencies, and other agents' likely actions. Transfer to new teams would require representations that accommodate variable membership and embodiment-specific capabilities. Such models should provide reusable coordination knowledge for allocation and joint planning, assessed through physical team performance.

\paragraph{Self-Organizing Collaboration in Open Environments}
Long-running teams must accommodate agents joining or leaving, newly arising tasks, changing connectivity, and human intervention. Self-organization can form temporary teams, select coordinators, and revise roles according to available capabilities~\citep{zhu2024self}. Such changes must preserve task ownership and outstanding commitments across organizational transitions. The resulting direction is adaptive coordination that maintains core services while membership and architecture evolve, extending distributed Physical AI beyond fixed teams and predefined operating conditions.

\clearpage

\section{Hardware and Systems}
\label{ch:hardware-systems}
At last, every Physical AI program ultimately depends on an actual machine to execute proposed actions in the real world. Therefore, this chapter discusses the actual hardware and system through five questions: what physical interactions a body's morphology can support (Section~\ref{sec:c9:embodiment}), how sensing and actuation close the feedback loop (Section~\ref{sec:c9:sensing-actuation}), where computation can run without missing the time to act (Section~\ref{sec:c9:device-edge-cloud}), how software organizes these capabilities (Section~\ref{sec:software-infrastructure}), and how they remain reliable through repeated use and updates (Section~\ref{sec:c9:reliability}). For the ease of understanding, there are several case studies: a simple component case in Section~\ref{sec:c9:sensing-actuation}, a dexterous-hand case study in Section~\ref{sec:c9:dexterous-cases} to illustrate how hardware shapes interaction, and a more complex humanoid case that stitches all components together in Section~\ref{sec:c9:humanoid-cases}.

\subsection{Hardware and Embodiment Taxonomy for Physical AI}
\label{sec:c9:embodiment}


A robot opening an unfamiliar drawer must do more than recognize a handle. It must reach it, establish a grip, and discover how the drawer responds to a pull. Resistance may mean that the drawer is heavy, jammed, or still locked. Continuing the same movement could damage the handle. The robot needs evidence that distinguishes these possibilities and a body capable of changing its action. This connection between physical interaction, observation, and correction is central to Physical AI.

\subsubsection{From Hardware Components to Embodiment}

Cameras, motors, encoders, transmissions, and processors are hardware components. Their arrangement gives a robot its \emph{embodiment}, comprising its shape and mass distribution, available movements, sensing locations, and ways of making contact. The platform adds power, computation, communication, and the controllers that regulate those movements. Together, these correspond to the broader notion of the embodiment $B_{\mathrm{emb}}$ in Section~\ref{sec:physical-agent-definition}. A change anywhere in this system can change what an action means. Replacing a rigid fingertip with a soft pad alters both the grip and the measurements produced during contact.

The drawer example makes these dependencies visible. An external camera may locate the handle but lose sight of it behind the gripper. Joint encoders describe arm movement, while force or tactile measurements can reveal resistance or slip. A small initial pull then serves two purposes by moving the drawer and testing the robot's prediction about its motion. The outcome can justify continuing, changing direction, or releasing the handle. Such exploratory actions require suitable force limits and feedback. Trial and error is useful only when the body can tolerate the trial and the system can interpret its outcome.

The same mechanics constrain possible corrections. Reach does not guarantee that an arm can sustain a load, and a grip that supports lifting may fail when twisting a cap. Grasp analysis captures this distinction through the forces and torques a set of contacts can resist~\citep{ferrari1992grasp}. A mobile robot may reach the handle yet lack room to reposition or sufficient support to resist the pull. Planning must therefore consider how to remain controllable after the first movement, as well as whether that movement is possible.

Useful feedback must also arrive before the opportunity to correct is lost. A detailed image delivered after an object has slipped provides evidence for later learning but cannot rescue that grasp. Sensor placement and task-state observability (Section~\ref{sec:c9cao:observability}) must be considered together with control delay and actuator response (Section~\ref{subsec:measurement-motion-loop}). Better inference can extract more from available signals. Sustained performance still depends on obtaining the right evidence and acting within the body's limits.

\begin{table}[!t]
  \centering
  \caption{Common embodiments and the feedback their tasks require.
  Categories overlap. A hand belongs to a carrier, and a laboratory may
  combine several kinds of machine}
  \label{tab:c9cao:spectrum}
  \footnotesize
  \setlength{\tabcolsep}{4pt}
  \begin{tabular}{@{}>{\raggedright\arraybackslash}p{2.7cm}
                  >{\raggedright\arraybackslash}p{5.0cm}
                  >{\raggedright\arraybackslash}p{6.1cm}@{}}
    \toprule
    Embodiment & Physical advantage & What the feedback must resolve \\
    \midrule
    Fixed instruments
      & Fixtures and specialized tools support repeatable operations
      & Whether a process achieved its target volume, temperature, or other measured result \\
    Fixed-base arms
      & Stable support for manipulation within reach
      & Tool position and contact loading, including errors hidden by visual occlusion \\
    Mobile manipulators
      & Access to several workstations
      & Base location and support as well as arm and object state \\
    Legged and humanoid robots
      & Movement across steps and uneven ground
      & Balance and changing support contacts while moving or manipulating \\
    Dexterous hands
      & Multiple contacts permit adjustment within the grasp
      & Finger configuration, contact distribution, and incipient slip \\
    Soft and continuum robots
      & Deformation accommodates confined spaces or delicate objects
      & Changing shape and loading despite material hysteresis and distributed contact \\
    Aerial and underwater robots
      & Access beyond ground routes
      & Body motion under fluid disturbances, with constrained energy and communication \\
    Robot-instrument systems
      & Complementary machines complete experimental workflows
      & Sample identity, transfers, and measured outcomes across machine interfaces \\
    \bottomrule
  \end{tabular}
\end{table}

Table~\ref{tab:c9cao:spectrum} compares embodiments through the interactions they support. A liquid handler is an example of a fixed instrument. An arm on a wheeled base is a mobile manipulator. A biped with arms belongs to the legged and humanoid category. These categories describe different physical advantages, rather than successive levels of intelligence. Suitability depends on the task and on what each body can measure while performing it.

Autonomous laboratories illustrate why capability often comes from a combination of bodies. Instruments provide controlled processes and measurements, while arms or mobile robots transfer samples and operate equipment~\citep{burger2020,szymanski2023alab,dai2024}. The resulting system must track whether each transfer and operation actually succeeded. These interfaces connect embodiment to the system-level agents in Section~\ref{pai82:sec:physical-forms}.

\subsubsection{Embodied Model-Hardware Co-Design and Multi-Embodiment Adaptation}

Model-hardware co-design starts from a coupled choice. The body determines which states can be observed and which actions can be realized, while the model and controller determine how much of that capability is used. A camera added to expose a contact changes the policy's inputs and training data. A softer fingertip changes contact dynamics and may reduce the precision the policy must achieve. A larger model can improve inference yet lose that advantage if its delay exceeds the time available to correct a slip. These choices must therefore be evaluated as a system, against the same physical task and operating limits.

\begin{figure}[tbp]
  \centering
  \includegraphics[width=\linewidth]{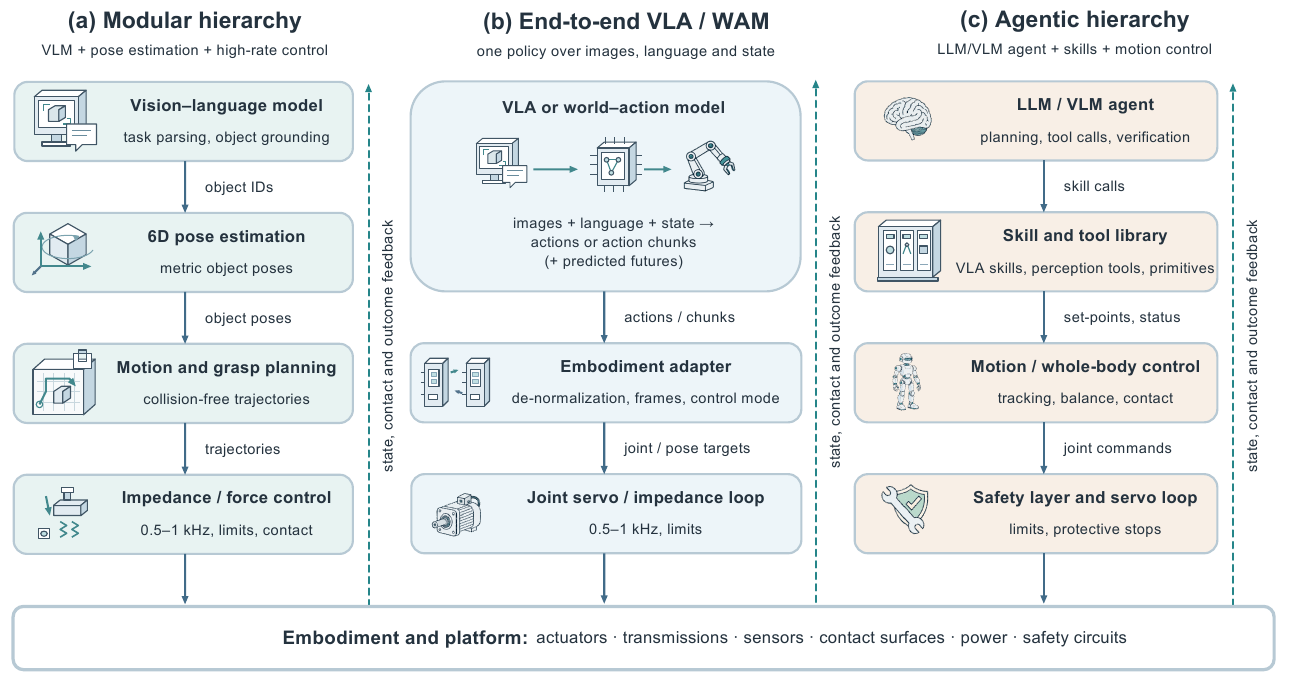}
  \caption{Models reach the world through different control arrangements.
  A modular hierarchy exposes intermediate estimates and plans. An
  end-to-end policy learns a more direct observation-to-action mapping. An agent calls skills and checks their outcomes. Local feedback supports
  timely motion and contact regulation in each arrangement}
  \label{fig:c9cao:arch}
\end{figure}

A model influences the world through a control interface. It may request a tool movement, joint positions, or forces. Local controllers translate those targets into actuator behavior. An \emph{embodiment adapter} handles the robot's command conventions, coordinate frames, and limits. Figure~\ref{fig:c9cao:arch} shows three arrangements of these responsibilities. All require a return path from the physical outcome to the next decision.

A modular pipeline estimates an object's pose, plans a motion, and tracks it with a controller. Its intermediate quantities make some failures easier to locate. The object may have been misidentified, the path may have been infeasible, or contact may have differed from the prediction. An end-to-end vision-language-action policy learns a more direct mapping from images and instructions to actions, as in OpenVLA~\citep{kim2024openvla}. It can exploit visual cues without requiring an explicit object pose at every step, but its output still has the units and physical consequences defined by the controller. Without incorporating new observations, neither arrangement can revise its task-level response to an unexpected event.

An agentic hierarchy adds task-level decisions, such as selecting a tool or deciding to retry. Systems that compose skills from language illustrate this separation~\citep{ichter2023saycan,liang2023code}. During drawer opening, a fast controller may limit contact force while the higher-level system decides whether resistance calls for another strategy. A completion message from a skill is useful only insofar as it is supported by evidence about the drawer. Execution checking in Section~\ref{sec:software-infrastructure} and reliability assessment in Section~\ref{sec:c9:reliability} develop this requirement further.

Transfer to another robot changes the relationship between observation and action. Knowledge that a drawer translates along a guide may remain useful. Camera appearance, reachable poses, finger contacts, motor limits, and the forces generated by a command may all change. Expressing a desired movement relative to the tool can reduce dependence on a laboratory's coordinates, but does not make the new robot mechanically equivalent. The receiving body must still support the intended motion and contact.

Training across robots can reveal useful shared regularities. The Open X-Embodiment study demonstrates benefits from combining data across robots~\citep{openx2024}. Octo supports adaptation to new observation and action spaces through fine-tuning~\citep{octo2024}, while HPT combines a shared model with embodiment-specific input and output modules~\citep{wang2024hpt}. These approaches organize what can be shared and what must be adapted. Their physical value must be established on the receiving robot, including its contact behavior and recovery from errors (Section~\ref{sec:brain-embodiment-adaptation}). Success on familiar images alone does not establish that the transferred commands produce the intended forces.

Hardware design can make this adaptation easier. ALOHA's matched leader and follower arms give demonstrations a direct relationship to robot joint movements~\citep{zhao2023}. UMI coordinates a handheld gripper, camera geometry, and action representation to support transfer from portable demonstrations~\citep{chi2024umi_ch6} (Section~\ref{subsec:physical-experience-acquisition}). A compliant fingertip can also tolerate small placement errors, while tactile sensing reveals how the contact develops. Joint optimization can extend to the mechanism itself.~\citet{caoChen2021HardwarePolicy} optimize mechanical and control parameters together and build an underactuated hand from the resulting design. Such optimization still depends on whether the modeled contacts and actuator response describe the physical machine.

For the drawer task, a useful co-design cycle begins by identifying the failed correction. If the handle disappears behind the fingers, changing the viewpoint or adding touch may resolve missing evidence. If resistance is measured but force cannot be reduced promptly, transmission, compliance, and controller bandwidth become the relevant design variables. If the state is observed and the action is feasible but the policy chooses poorly, additional training may be the appropriate change. Each candidate requires data collected or transformed for its own observation and action interface, followed by evaluation on held-out physical trials. Comparing an unchanged policy across bodies measures transfer. Allowing each body the same adaptation budget tests the quality of the co-designed system. Reporting both prevents a favorable match to one policy from being mistaken for a universal hardware advantage. The final comparison is cost per valid output, including collection, computation, interventions, and maintenance, as developed below and in Section~\ref{sec:c9:reliability}.

\subsubsection{Task-Oriented Embodiment Selection, Validation, and Evaluation}

\begin{figure}[h]
  \centering
  \includegraphics[width=\linewidth]{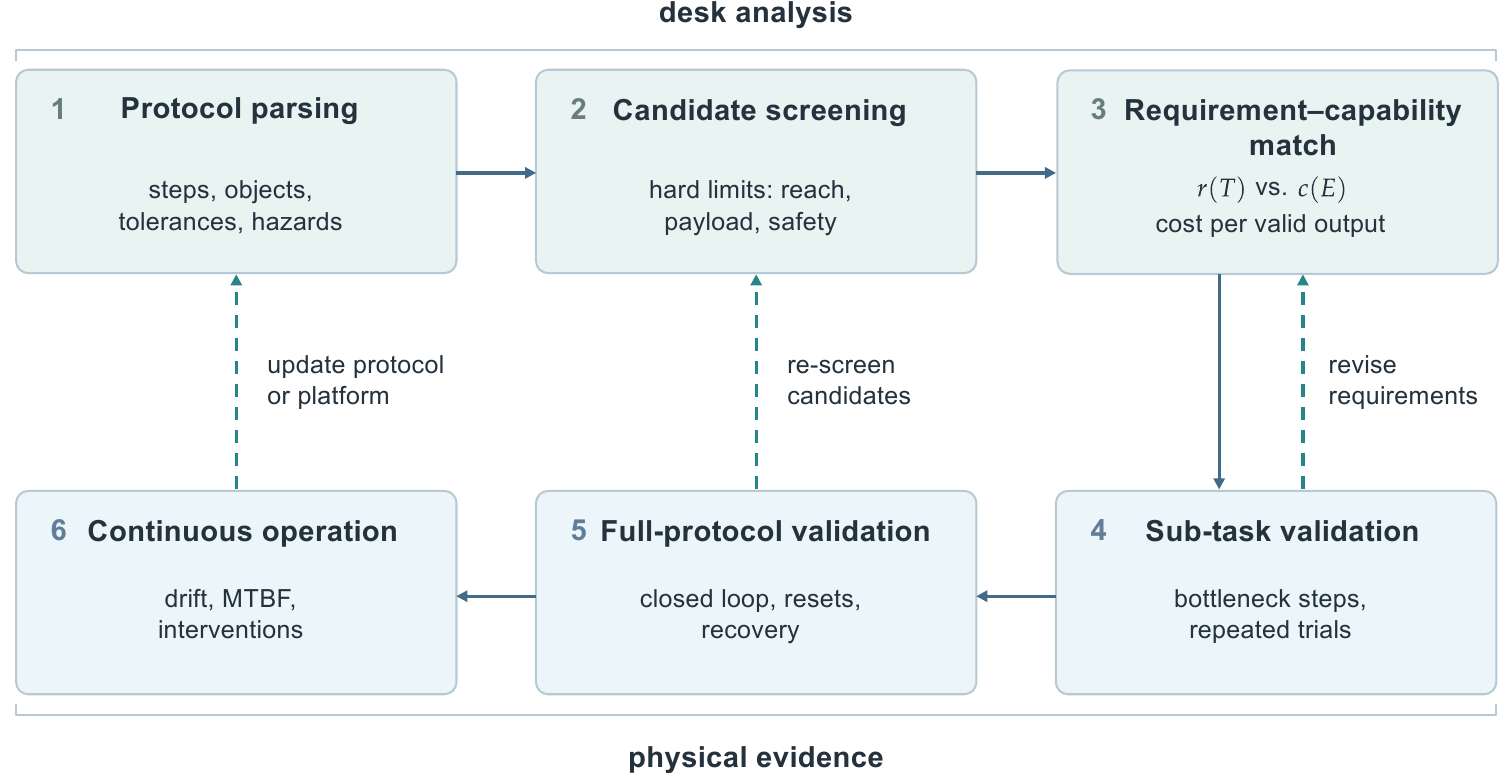}
  \caption{Embodiment selection develops through increasingly realistic
  evidence. Task analysis and specifications identify candidates. Operation-level tests, complete workflows, and continued use reveal
  whether the proposed body and controller meet the requirements}
  \label{fig:c9cao:select}
\end{figure}

Selecting a robot begins with a complete task, including the measurements that establish success. Reach and payload can exclude an unsuitable arm, but do not establish whether it can thread a cap without damaging a vial. Figure~\ref{fig:c9cao:select} moves from requirements to tests of individual operations, complete workflows, and continued use. Failures can motivate a change in the policy, sensing, mechanics, or protocol.

Candidates should meet the same requirements for workspace, load, precision, contact force, time, and recovery. Among feasible options, the cost per valid output introduced in Section~\ref{sec:application-landscape} gives a useful comparison. Divide the fully allocated average cost of an attempt by the measured fraction that produces a valid result. The cost includes failed attempts, human interventions, and recovery. Early estimates remain provisional until tested under the intended workload.

A laboratory preparing metal-organic frameworks (MOFs), porous materials built from metal-containing units and organic linkers~\citep{furukawa2013mof}, provides a concrete example. A synthesis workflow may involve opening vials, transferring solids and liquids, heating, separation, and loading instruments~\citep{tom2024sdl}. Successful motion is insufficient. The sample must retain its identity, composition, and integrity. A balance may verify a transfer that the arm's camera and encoders cannot establish by themselves.

Consider opening a sealed vial, dispensing a specified quantity, and seating the vial in an instrument. Table~\ref{tab:c9cao:mof} compares four possible configurations through the questions these operations raise. Their exact arms, grippers, sensors, and controller settings must be specified before testing. Additional fingers or lower transmission ratios change the possibilities for interaction. Their practical benefit depends on how well that interaction is observed and controlled.

\begin{table}[tbp]
  \centering
  \caption{Questions for evaluating four candidate configurations in a
  MOF workflow. These are mechanisms to test, not measured rankings.
  The first three configurations use parallel grippers}
  \label{tab:c9cao:mof}
  \footnotesize
  \setlength{\tabcolsep}{4pt}
  \begin{tabular}{@{}>{\raggedright\arraybackslash}p{3.0cm}
                  >{\raggedright\arraybackslash}p{5.2cm}
                  >{\raggedright\arraybackslash}p{5.6cm}@{}}
    \toprule
    Configuration & Interaction to evaluate & Evidence needed \\
    \midrule
    Torque-sensing dual arm with impedance control
      & Yielding during cap engagement and vial seating
      & Contact loads, alignment errors, damage, and completion under matched tuning \\
    Position-controlled dual arm without added force sensing
      & Precise free motion followed by constrained contact
      & Contact-force peaks, effects of passive compliance, and recovery from misalignment \\
    Low-reduction dual arm with motor-current feedback
      & Responsive motion and current-based loading estimates
      & Accuracy under load, effects of friction and acceleration on force estimates \\
    Arms carrying dexterous hands
      & Reorientation and cap handling through multiple contacts
      & Slip, successful in-hand adjustments, and whether fewer regrasps offset coordination costs \\
    \bottomrule
  \end{tabular}
\end{table}

Tests should use the same protocol, object batches, starting-condition distribution, and definition of success. Repeated trials distinguish a reliable advantage from a fortunate grasp (Section~\ref{subsec:sustained-physical-output}). Recording observations, commands, and measured outcomes helps explain failures by distinguishing whether the robot lacked information, chose an unsuitable action, or could not physically realize it. The diagnosis then guides a concrete change and another test.

The preferred solution may combine specialized machines, provided their transfers and coordination remain reliable. Continued operation must then test calibration drift, worn contact surfaces, thermal limits, and recovery after interruptions. This closes the design loop. Experience changes the model and, where necessary, the body that supplies its evidence and carries out its decisions.

\subsection{Sensors, Actuators, and Intelligent Materials}
\label{sec:c9:sensing-actuation}


To lift a glass vial, a robot must first find it, then recognize contact, and finally hold it firmly without breaking it. A camera helps with the approach. Joint encoders report how the fingers and arm have moved. Touch and force measurements reveal what happens when the fingers meet the glass. Electric motors supply the motion, but their response depends on the transmission, the controller, and the load. Physical AI emerges through this interaction. The robot acts, measures the consequences, and adjusts its behavior. A model can infer an object's properties from experience only when the hardware exposes useful evidence and can carry out the resulting decisions.

\begin{figure}[tbp]
  \centering
  \includegraphics[width=\linewidth]{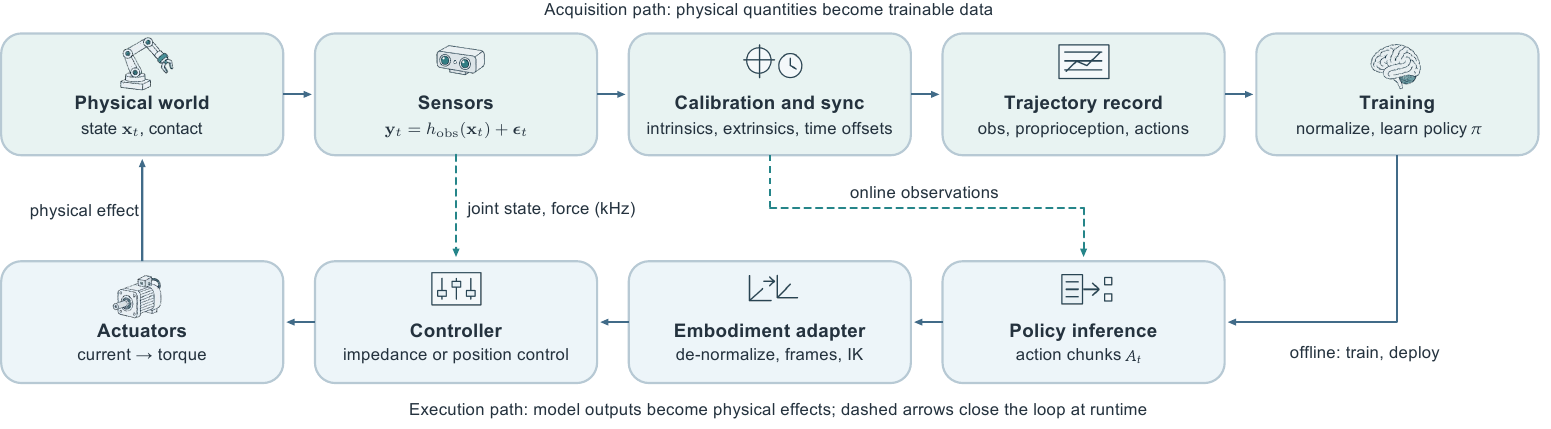}
  \caption{Measurements connect learning to physical action. Calibrated,
  synchronized observations become training trajectories. During execution,
  the policy supplies targets, while local controllers use fresh body and
  contact measurements to drive the actuators. The commanded movement and
  the movement actually achieved need not be identical}
  \label{fig:c9cao:loopfig}
\end{figure}

The relevant hardware therefore extends from visible cameras and grippers to the electronics inside a joint, including encoders, current sensors, power transistors, and local control processors. These components determine both what a learning system can observe and how its actions affect the world. Figure~\ref{fig:c9cao:loopfig} connects the measurements collected during learning to the feedback used during execution. Section~\ref{sec:multimodal-perception} discusses sensor physics and fusion more generally. Here the emphasis is on their consequences for robot learning and physical interaction.

\subsubsection{Acquiring Physical Experience for Learning}
\label{subsec:physical-experience-acquisition}

Before a robot can learn from a recorded movement, its sensors must agree on where and when that movement occurred. Camera calibration relates image coordinates to the robot. Joint positions and a kinematic model locate the tool. For a wrist camera, hand-eye calibration identifies the fixed relationship between camera and gripper~\citep{zhang2000flexible,tsai1989handeye,park1994handeye}. Timing is equally important. A 20\,ms mismatch for a tool moving at 0.5\,m/s corresponds to 1\,cm of displacement. Hardware triggering, synchronized clocks, and measured delays help align images, joint states, and forces. A receipt timestamp is not necessarily the acquisition time. These details can determine whether a demonstration teaches the robot to react to contact or, mistakenly, to a state that occurred afterward.

A useful trajectory records observations, measured body state, and the commands sent to the controller together. It also preserves their units, coordinate frames, calibration, and control mode. A command to advance the tool by a centimeter means something different from a command to exert a force. Even identical position targets can produce different contact behavior under different controller gains. Changes to the camera, gripper, or controller consequently change the learning problem, not just the equipment inventory. If actions are normalized numbers or relative tool motions, deployment must preserve the same scaling and coordinate conventions when converting them into controller targets.

Different collection methods preserve different parts of this physical experience. Robot teleoperation, as in ALOHA and DROID~\citep{zhao2023,khazatsky2024droid}, records demonstrations through a robot's own body and controller. Handheld devices such as UMI make collection more portable by combining a gripper with a camera and motion estimation~\citep{chi2024umi_ch6}. Their trajectories can describe real contact, but the human holding the device does not have the robot's inertia, reach, or torque limits. Human video supplies broader examples of activities and object use, while motion-tracking gloves and exoskeletons add hand or arm kinematics~\citep{grauman2022ego4d,wang2024dexcap,fang2024airexo}.

\begin{figure}[!t]
  \centering
  \includegraphics[width=0.92\linewidth]{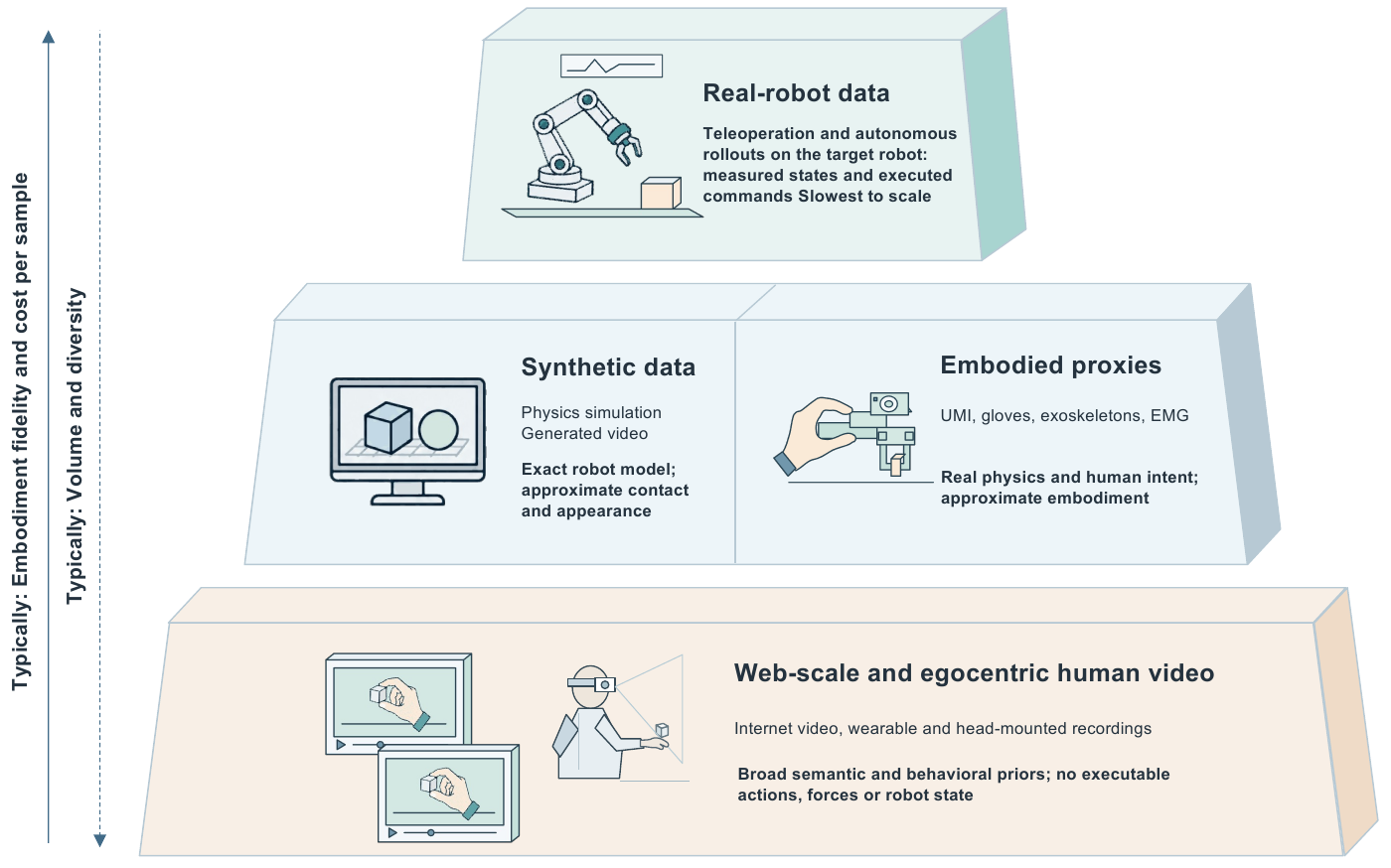}
  \caption{Sources of experience for Physical AI. Human video
  offers broad coverage. Simulation and handheld devices address different
  parts of the gap between observation and execution. Robot recordings
  capture the target body's behavior. The levels describe complementary
  evidence, not a universal ranking of dataset quality. EMG denotes
  electromyography. It records muscle activity rather than robot contact
  forces}
  \label{fig:c9cao:pyramid}
\end{figure}

Surface electromyography (sEMG) adds a different kind of observation. Electrodes on the skin record the electrical activity associated with muscle recruitment~\citep{farina2014emg}. This is the modality denoted EMG (electromyography) in Figure~\ref{fig:c9cao:pyramid}. It can reveal motor intent when movement is slight or visually occluded, making it a useful complement to video during demonstration collection~\citep{ctrllabs2025emg}. Its interpretation nevertheless depends on the person, posture, and electrode placement. Neither visible motion nor muscle electrical activity directly specifies the contact forces a robot must reproduce. Recordings should therefore synchronize EMG with kinematics and video, preserve the electrode configuration and calibration, and include force or tactile measurements when contact-force supervision is required. These sources complement robot recordings rather than serving as mechanically interchangeable substitutes.

The data pyramid in Figure~\ref{fig:c9cao:pyramid}, motivated by GR00T~N1~\citep{bjorck2025gr00t}, places abundant human video below synthetic and robot-free demonstrations, with robot execution at the top. Its distinction is the kind of evidence available, as well as its volume. Simulation can supply complete robot actions while approximating friction and contact. Human recordings contain real interactions without the robot's execution dynamics. A weighted training objective retains these differences through
\begin{equation}
  \min_{\theta}\sum_j w_j\,
  \mathbb{E}_{\zeta_{\mathrm{sample}}\sim\mathcal{D}_j}
  [\mathcal{L}_j(\theta;\zeta_{\mathrm{sample}})],
  \qquad w_j\geq 0,\quad \sum_j w_j=1.
  \label{eq:c9cao:mixture}
\end{equation}

Here each source $\mathcal{D}_j$ supports its own learning objective $\mathcal{L}_j$. Robot actions can supervise action prediction, whereas human video can support visual prediction or activity understanding. Reweighting cannot turn an unmeasured quantity into direct supervision (Section~\ref{sec:physical-data-foundations}).

\subsubsection{Closing the Loop from Measurements to Motion}
\label{subsec:measurement-motion-loop}

When a finger touches a hard surface, force may rise before a large model has processed another image. The robot must therefore respond at several timescales. A learned policy can choose a grasp, a movement, or a desired compliance. A local controller follows that target. The motor drive regulates electrical current. Fast joint feedback remains useful even when the policy is learned end to end. Playing a predicted sequence of actions rapidly is also different from reacting rapidly to new evidence.

The electronics make this distinction concrete. A drive switches power transistors to apply voltages to the motor windings. Current-sensing circuits and analog-to-digital converters report the resulting currents. An encoder or other rotor-position sensor tells the controller how the rotor is oriented. The drive adjusts these voltages to produce the desired torque. Filtering suppresses noise but introduces delay, and electrical switching can contaminate small sensor signals. Consequently, sensing, computation, and power delivery must be designed together~\citep{ti2025humanoidcurrent,kollmorgen2020servo}.

The total delay includes sensing, communication, inference, and actuation. In a linearized feedback loop, a delay $\Delta$ adds a phase lag $\omega\Delta$ at angular frequency $\omega$. The controller increasingly acts on an outdated state. This explains why contact regulation usually stays close to the hardware, while slower reasoning can run elsewhere (Section~\ref{sec:c9:device-edge-cloud}). Each interface must specify reference frames, units, timestamps, control modes, and limits. The return path should report the measured motion and available forces or currents, together with faults and temperature. Without this feedback, a learning system can confuse a saturated actuator or a delayed measurement with an incorrect plan.

\subsubsection{Task-State Observability: Cameras, Encoders, and Contact Sensing}
\label{sec:c9cao:observability}

The camera is often the most recognizable sensor on a physically intelligent robot. An RGB camera measures light and provides the appearance cues used to recognize objects, interpret a scene, and connect language to action. Depth cameras add geometric information. Stereo systems compare two viewpoints. Structured-light systems observe a projected pattern. Time-of-flight systems estimate distance from the travel time or phase of returned light. These approaches have different sensitivities to distance, illumination, and surface appearance. An integrated RGB-D sensor makes color and depth available together, as illustrated by the RealSense architecture~\citep{keselman2017intel}. However, transparent glass, shiny metal, and low-texture surfaces can leave missing or misleading depth. ClearGrasp addresses one such problem by inferring geometry for transparent objects~\citep{sajjan2020cleargrasp}. Inferred geometry still requires validation when an action can damage the object.

Placement determines what the camera can contribute. A head or external camera sees the overall workspace but may lose the contact behind the hand. A wrist camera follows the approach and reveals small features, although its view moves with the arm and may become obstructed at the final instant. Exposure, motion blur, field of view, and timing can matter more than the advertised pixel count. For a vision-language-action model, moving the camera also changes the relationship between an image and a successful action. Calibration aligns geometry. Training must still account for the change in appearance.

While cameras look outward, encoders measure the body's own movement. A rotary encoder converts shaft rotation into an electrical position reading. Optical encoders read a patterned disk. Magnetic encoders sense the orientation of a magnet. Inductive designs use electromagnetic coupling. Incremental encoders count changes and generally need a reference after startup, whereas absolute encoders identify position within their specified measurement range. More digital counts do not by themselves guarantee more accurate position. Mounting error, alignment, mechanical deformation, and calibration also matter~\citep{heidenhain2023dualencoders}. Velocities are often estimated from successive position measurements, which introduces a trade-off between noise and delay.

\begin{figure}[tb]
  \centering
  \includegraphics[width=0.92\linewidth]{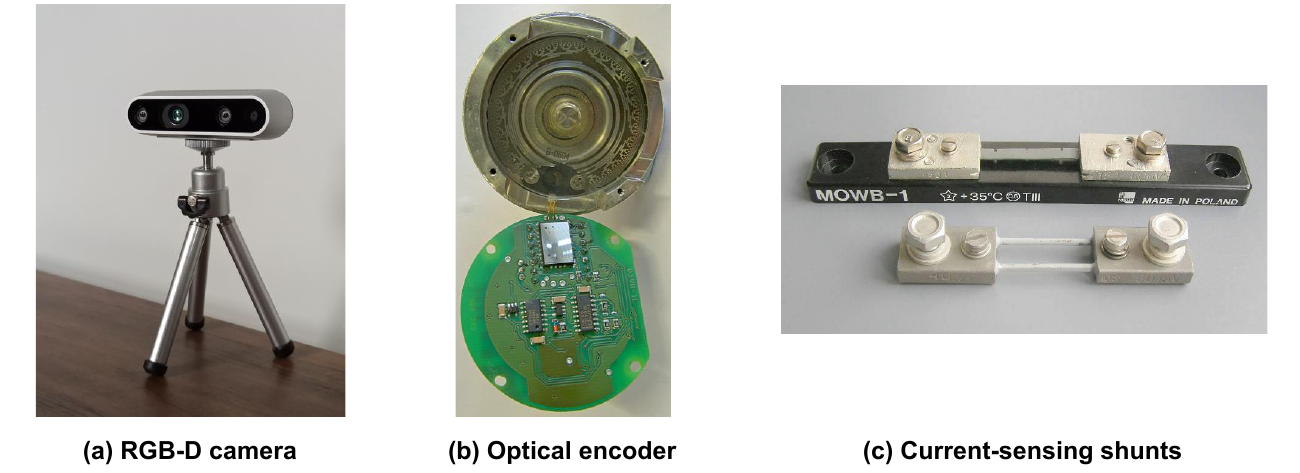}
  \caption{Different measurements support different parts of an action.
  A camera observes the scene, an encoder reads shaft angle, and a shunt
  converts electrical current into a small voltage. The opened encoder
  shows its coded disk. These examples illustrate principles rather than
  a prescribed robot sensor suite. Photos: (a)
  \href{https://commons.wikimedia.org/wiki/File:Intel_Realsense_depth_camera_D435.jpg}{Marc Auledas},
  \href{https://creativecommons.org/licenses/by-sa/4.0/}{CC BY-SA 4.0};
  (b) \href{https://commons.wikimedia.org/wiki/File:Gray_code_rotary_encoder_13-track_opened.jpg}{Mike1024/Dicklyon}, public domain;
  (c) \href{https://commons.wikimedia.org/wiki/File:Bocznik01.jpg}{Marek bydg},
  \href{https://creativecommons.org/licenses/by-sa/3.0/}{CC BY-SA 3.0}}
  \label{fig:c9cao:sensorhardware}
\end{figure}

A motor encoder and a joint-output encoder observe different shafts. With an ideal rigid gearbox, motor angle divided by the reduction ratio gives joint angle. Real transmissions have friction, lost motion, and elastic deformation. An output encoder can reveal movement errors that the motor encoder misses, particularly during reversals and under load. Joint measurements and the robot's geometry then estimate the tool's pose, but do not directly reveal whether an object is slipping between the fingers. On a walking or mobile robot, an IMU adds angular velocity and specific force from gyroscopes and accelerometers. Combined with joint and contact measurements, it helps estimate body motion. Integration alone accumulates error, and gravity must be accounted for when interpreting acceleration. Figure~\ref{fig:c9cao:sensorhardware} presents examples of cameras, encoders, and current-sensing shunts.

\begin{figure}[tb]
  \centering
  \includegraphics[width=0.92\linewidth]{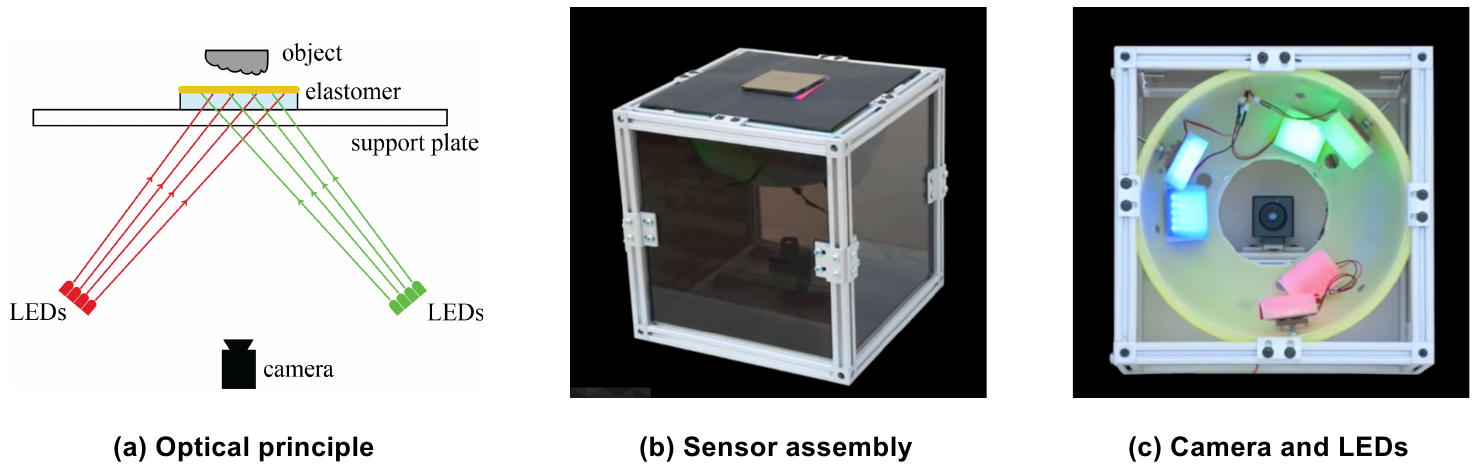}
  \caption{Optical touch converts contact into an image of a deformable
  surface. The reflective coating and illumination allow surface shape to
  be recovered independently of the contacted object's optical appearance.
  Reproduced from Figure~2 of~\citet{yuan2017gelsight}, under
  \href{https://creativecommons.org/licenses/by/4.0/}{CC BY 4.0}.
  The illustration describes the sensing principle, not the packaging of
  every modern fingertip}
  \label{fig:c9cao:gelsight}
\end{figure}

Touch becomes especially valuable once the hand hides the contact. In a vision-based tactile sensor, a small camera looks inward at a soft surface rather than outward at the scene. Pressing an object against that surface changes its shape and, in some designs, moves printed markers. GelSight uses controlled illumination to recover fine contact geometry. DIGIT packages a related optical approach into a compact fingertip~\citep{yuan2017gelsight,lambeta2020digit}. Figure~\ref{fig:c9cao:gelsight} shows why optics and material design are inseparable here. The camera observes the deformation of a purpose-built skin. Images can support estimates of local shape, shear deformation, and incipient slip. They are not direct force maps. Recovering force requires calibration or a learned model of the skin and its loading.

This information changes what a policy can learn. A gripper can tighten when the contact patch starts to slip, or adjust the direction of insertion when an edge presses unevenly. A compliant skin may also wear, creep, or change with temperature, shifting the relationship between image and force. Other tactile technologies measure changes in resistance, capacitance, or magnetic field. ReSkin, for example, uses a deformable magnetic skin~\citep{bhirangi2021reskin}. The choice is therefore between useful measurements, spatial coverage, durability, and processing demands, rather than between a sensor that ``has touch'' and one that does not.

Force sensing complements this local picture. A six-axis wrist force-torque sensor measures three force components and three moments transmitted through the wrist. Strain gauges detect tiny deformations of an elastic structure. Calibrated combinations of their signals recover the load. The reading includes the tool's weight and inertial loading, which must be compensated when estimating contact force~\citep{atiAxiaEtherCAT}. A joint torque sensor measures twisting load about one joint axis, often on the output side of the transmission. Its location matters. A sensor upstream of a lossy gearbox does not measure the same load as one downstream. Multiple joint measurements can support estimates of external contact, but these depend on the robot's dynamics and contact assumptions. Specialized joint force-torque sensing has even supported touch localization over a robot's body~\citep{iskandar2024intrinsictouch}. An ordinary single-axis torque sensor does not automatically provide that capability.

Current sensors offer another source of evidence already needed by the drive. A small shunt resistor produces a voltage proportional to current. Magnetic current sensors infer it from the surrounding magnetic field. For a suitable motor model, torque-producing current estimates motor torque. It does not directly measure force at the fingertips. Friction, acceleration, gravity, and transmission losses contribute to the same signal. Current remains valuable for control, overload detection, and learning about loading, provided these ambiguities are recognized~\citep{ti2025humanoidcurrent,wensing2017proprioceptive}.

The purpose of combining sensors is to resolve uncertainty that matters for an action. In the Jenga experiments of~\citet{fazeli2019seefeelact}, vision locates blocks and force measurements during interaction help infer how they can move. The contact supplied information unavailable from an image alone. A similar principle applies when opening a drawer or seating a connector. The robot must distinguish free motion from resistance, then decide whether to continue, change direction, or retreat. This is task-state observability in practical terms. The most useful sensor is the one that reveals the distinction in time to change the action, at a precision appropriate to the task.

\subsubsection{Electric Motors and the Mechanics of Actuation}
\label{subsec:motors-actuation-mechanics}

An actuator converts supplied energy into mechanical work. Electric motors are a practical default for many robot arms, grippers, mobile bases, and legged platforms because electrical power is readily distributed, motor drives are mature, and feedback can regulate motion and torque. These advantages depend on scale and duty. Hydraulic, pneumatic, and material-based actuators remain useful where their force capacity or mechanical behavior better fits the task.

Inside an electromagnetic motor, current in windings interacts with a magnetic field to turn a rotor. The stationary part is the stator. Sustained rotation requires the winding currents to change as the rotor turns, a process called commutation. In a brushed direct-current (DC) motor, physical brushes and a segmented commutator perform this switching. Such motors are straightforward to drive and remain useful in compact or inexpensive mechanisms, but the brushes wear and create electrical noise. A coreless brushed motor removes the iron core from the rotating winding, reducing rotor inertia and cogging. It still needs careful thermal design. ``Coreless'' describes construction, rather than a separate feedback control method~\citep{faulhaberDCtechnical}.

Brushless permanent-magnet motors place commutation in the electronics. The rotor carries magnets and the stator carries windings. The drive energizes the phases according to rotor position. Eliminating brushes removes a wear mechanism and enables efficient, controllable drives, at the cost of more complex electronics. The labels brushless direct-current (BLDC) motor and permanent-magnet synchronous motor (PMSM) describe closely related motor families. In a common convention, BLDC designs are associated with trapezoidal back-electromotive force and block commutation, whereas PMSMs are associated with sinusoidal waveforms. Products and control methods overlap, so the label alone is insufficient to predict torque quality. Field-oriented control regulates current components relative to the rotor field, allowing smooth torque control over a useful operating range~\citep{ti2020bldcpmsm,ti2025humanoidcurrent}.

\begin{figure}[tb]
  \centering
  \includegraphics[width=0.92\linewidth]{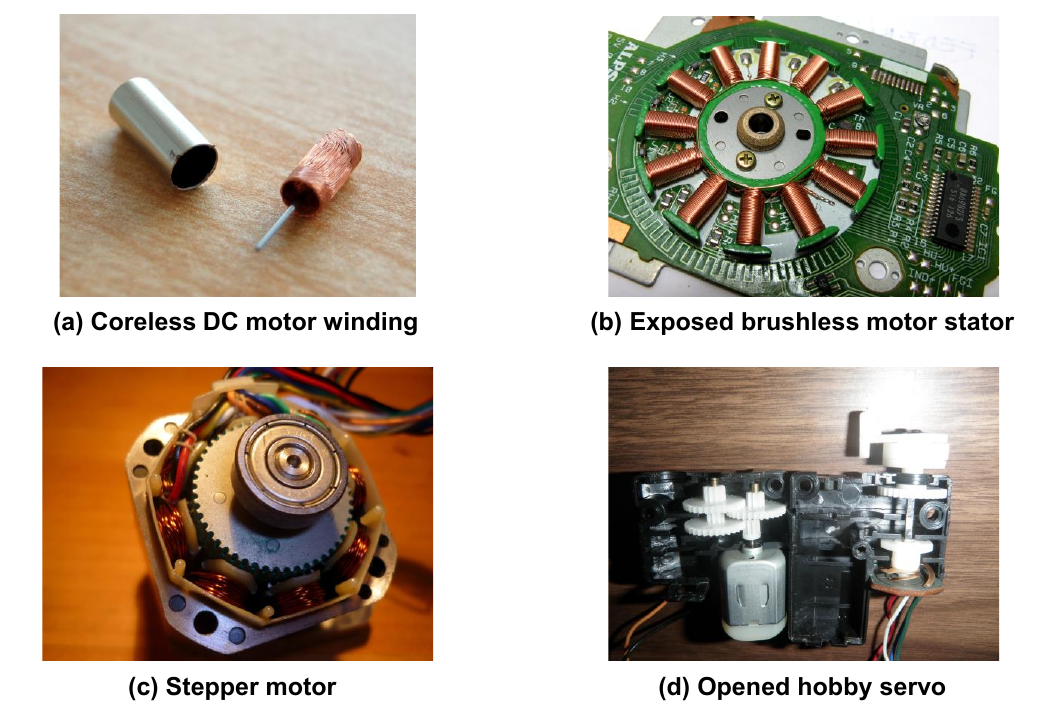}
  \caption{Motor construction and feedback integration. (a) A lightweight
  winding removed from a coreless DC motor. (b) Stationary windings of a
  brushless motor with its rotor removed. (c) A stepper motor, whose motion
  follows commanded magnetic alignments. (d) A hobby servo showing the
  motor, reduction gears, and output-position sensor. The servo is an
  integrated feedback system, not a separate electromagnetic motor family.
  Photos: (a) \href{https://commons.wikimedia.org/wiki/File:Miniature_Coreless_DC_Motor.jpg}{Avsararas},
  (b) \href{https://commons.wikimedia.org/wiki/File:Floppy_brushless_motor.jpg}{DnaX},
  and (d) \href{https://commons.wikimedia.org/wiki/File:Inside_a_servo_motor.JPG}{Jstapko},
  \href{https://creativecommons.org/licenses/by-sa/3.0/}{CC BY-SA 3.0};
  (c) \href{https://commons.wikimedia.org/wiki/File:Stepper_motor.jpg}{Dolly1010},
  \href{https://creativecommons.org/licenses/by/3.0/}{CC BY 3.0}}
  \label{fig:c9cao:motorhardware}
\end{figure}

Packaging creates further options within these families. An inrunner places the rotor inside the stator. An outrunner rotates an outer shell, often providing a larger torque-producing radius. A frameless motor supplies the rotor and stator for integration into the robot's own bearings and housing. A hollow center can accommodate cables, while different winding and rotor designs trade speed, torque, inertia, cooling, and space. These are engineering choices rather than a universal ranking. An easily accelerated rotor can benefit a responsive finger, whereas a large joint must also carry sustained load and reject heat. Figure~\ref{fig:c9cao:motorhardware} illustrates motor construction and feedback integration.

A stepper motor also uses electronically switched windings, but is designed to advance through discrete angular equilibria. Counting commanded steps can provide simple positioning in lightly loaded fixtures, dispensers, and laboratory mechanisms. The count is a command history, however, not proof of actual position. An overloaded open-loop stepper can lose steps. Microstepping makes the commanded motion smoother without guaranteeing correspondingly finer absolute accuracy. Closed-loop steppers add feedback, and the best choice depends on speed, load, efficiency, and required disturbance response~\citep{orientalmotorStepperBasics}. For Physical AI, this distinction affects the data as well as the motion. A dataset should not silently treat requested steps as measured joint states.

The word servo describes feedback-controlled operation, not a unique electromagnetic motor type. A servo system combines a motor, a drive, a position or velocity sensor, and a controller. It may also contain a gearbox~\citep{kollmorgen2020servo}. A small hobby servo and an industrial brushless servo share this principle while differing greatly in precision, torque control, response, and access to measurements. For a learning system, the interface matters. Position-only commands offer different possibilities from current or torque commands, and a nominally identical target can behave differently under another drive's limits and gains.

The transmission then converts motor motion into joint motion. Planetary, strain-wave, and cycloidal gearboxes are transmission mechanisms, not types of motor. They allow a relatively fast motor to turn a slower joint with greater torque. Direct drive couples the motor to the load without reduction. Quasi-direct drive combines a high-torque motor with a modest reduction. Neither approach is free of trade-offs. Removing reduction requires the motor itself to supply more torque, often increasing its size or thermal burden. Under an idealized constant-efficiency approximation,
\begin{equation}
  \tau_{\mathrm{j}} \simeq \eta_{\mathrm{gear}} N_{\mathrm{gear}} k_{\mathrm{t}} i_{\mathrm{q}},\qquad
  \omega_{\mathrm{j}} = \frac{\omega_{\mathrm{m}}}{N_{\mathrm{gear}}},\qquad
  I_{\mathrm{refl}} = N_{\mathrm{gear}}^{2} I_{\mathrm{m}} ,
  \label{eq:c9cao:gear}
\end{equation}
where $N_{\mathrm{gear}}$ is motor speed divided by joint speed, $\eta_{\mathrm{gear}}$ is transmission efficiency, and $k_{\mathrm{t}} i_{\mathrm{q}}$ is electromagnetic torque in an appropriate torque-current convention. For a simple brushed motor, armature current replaces $i_{\mathrm{q}}$. The last relation shows why a large reduction can make the rotor's inertia strongly felt at the joint. Actual output also depends on acceleration, friction, elasticity, and direction of power flow, so this equation is not an external-force measurement.

These effects become apparent when a foot lands or a hand meets an unexpected obstacle. A transmission that can be driven readily from the joint side is called backdrivable. Low reduction and low friction can improve this response and make motor current more informative about joint loading, as explored in the MIT Cheetah's proprioceptive actuators~\citep{wensing2017proprioceptive}. High-reduction joints can instead use output torque sensing and feedback to yield under external load, as in the DLR lightweight robot~\citep{albuschaffer2007dlr}. A series elastic actuator places a spring in the force path. Its calibrated deflection estimates transmitted torque and cushions disturbances, while introducing additional dynamics that the controller must handle~\citep{pratt1995}.

Peak torque is only part of the operating envelope. Winding heat grows approximately with current squared times resistance, so a joint may produce a large torque briefly but be unable to sustain it. At higher speed, the motor's back-electromotive force and the supply voltage limit the current the drive can establish. A policy that repeatedly asks for unavailable torque experiences delayed or incomplete movement. Temperature-dependent limits can make an identical command behave differently later in a task. Training and evaluation should therefore include the relevant torque-speed limits, thermal duty, friction, and control delay, rather than treating an actuator as an unlimited source of prescribed motion~\citep{maxon2024continuous,gealy2019quasidirect}. In the ANYmal experiments of~\citet{hwangbo2019}, a learned actuator model helps bridge this gap by representing the real response between commands and torque. The result shows why actuator and controller dynamics must be accounted for when transferring behavior from simulation to the physical robot.

For contact, the controller must also decide how strongly to resist position error. Impedance control makes a robot respond like a chosen spring and damper~\citep{Hogan1985}. A softer response can absorb a small alignment error. A stiffer response may be needed to support a load. In a simple, slowly varying contact model,
\begin{equation}
  F_{\mathrm{contact}}\approx k_{\mathrm{eff}}e_{\mathrm{contact}},\qquad
  k_{\mathrm{eff}}=\left(\frac{1}{k_{\mathrm{r}}}+\frac{1}{k_{\mathrm{e}}}\right)^{-1},
  \label{eq:c9cao:series}
\end{equation}
where $e_{\mathrm{contact}}$ is the position error along the contact direction, and the robot and environment stiffnesses, $k_{\mathrm{r}}$ and $k_{\mathrm{e}}$, act in series. An effective stiffness of $10^5$\,N/m turns a 1\,mm error into about 100\,N. Lower stiffness reduces this quasi-static force, although collision severity also depends on mass and speed. Learning can select compliance together with motion~\citep{martinmartin2019vices,hou2025acp}. Fast feedback must still realize that choice within the actuator's limits. A position-controlled robot can instead use force feedback to adjust its position target, an approach called admittance control. This is how a learned intention to ``insert gently'' acquires a measurable physical meaning.

\subsubsection{Fluid Power, Compliance, and Functional Materials}

Electric motors do not cover every useful form of actuation. Hydraulic systems use pressurized liquid to move pistons or hydraulic motors. Large force can be generated in a compact actuator, making fluid power attractive for demanding loads, but the complete machine also needs pumps, valves, seals, and fluid routing. Valve dynamics, leakage, friction, and fluid temperature affect the response a controller must learn or model. Electrically powered pumps and local hydraulic circuits can combine electrical distribution with hydraulic force generation. ``Hydraulic'' does not necessarily mean a remote central pump.

Pneumatic systems use compressed gas. Air can inflate a soft finger that wraps around an object with relatively simple mechanics, but its compressibility, flow restrictions, and material hysteresis complicate precise positioning. Pressure is useful feedback, yet does not uniquely determine the fingertip force when shape and contact change. Both fluid technologies therefore need sensing appropriate to their dynamics, rather than relying on pressure commands alone~\citep{rus2015,laschi2016soft}.

\begin{figure}[!t]
  \centering
  \includegraphics[width=\linewidth]{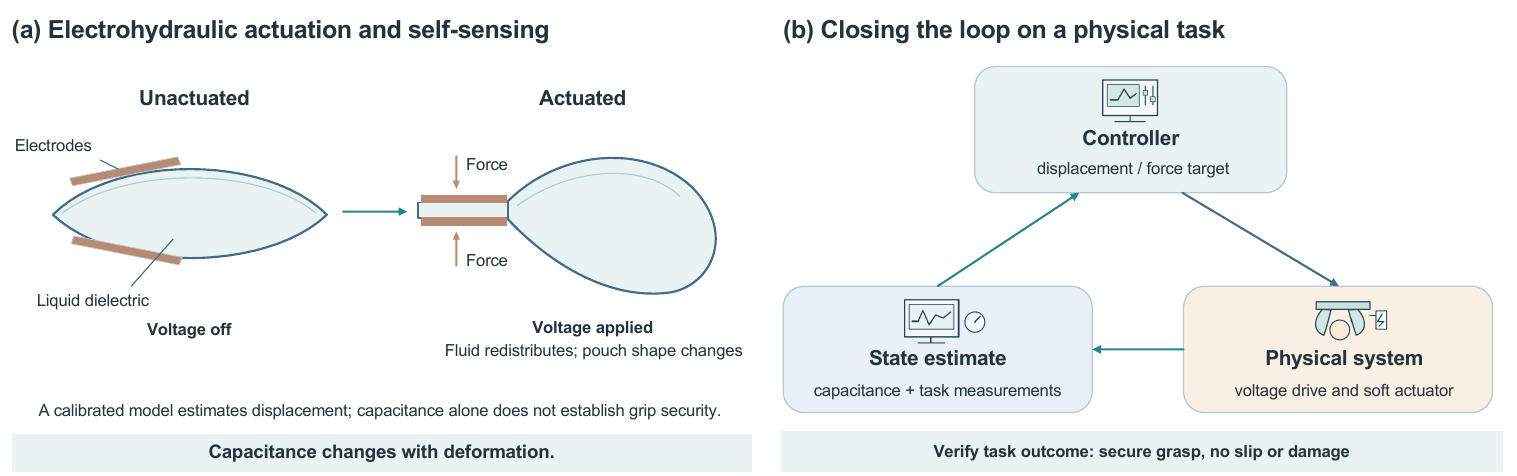}
  \caption{Functional materials become useful to physical intelligence
  through measured interaction. (a) In a conceptual HASEL cross-section, electrostatic attraction redistributes liquid and deforms the pouch.
  (b) Self-sensing contributes to feedback, while task measurements
  establish whether the interaction achieved its purpose
 ~\citep{acome2018hasel,ly2021hasel}. Shapes are illustrative, and the
  diagram does not imply that capacitance alone measures contact force}
  \label{fig:c9cao:materials}
\end{figure}

Functional materials offer an emerging route to combining actuation, sensing, and compliance within the body. They remain under active investigation alongside the motor-driven platforms discussed above. Thermal and electrochemical artificial muscles offer distinctive combinations of strain, force, and response time~\citep{mirvakili2018muscles}, but their value for physical intelligence depends on the interaction they enable. Hydraulically amplified self-healing electrostatic (HASEL) actuators, for example, use electrostatic forces to redistribute a liquid dielectric inside flexible pouches, producing motion. Changes in capacitance also reveal deformation~\citep{acome2018hasel}. By integrating driving electronics, capacitive self-sensing, and computation, \citet{ly2021hasel} demonstrate feedback control of a Peano-HASEL-powered arm. The actuator therefore supplies both mechanical work and a measurement that can guide the next action.

Figure~\ref{fig:c9cao:materials} connects this component behavior to task feedback. A compliant finger may accommodate an object's shape mechanically, while its deformation provides evidence for adjusting the grasp. Yet compliance alone does not reveal whether the object is secure, slipping, or being compressed excessively. That distinction requires calibrated sensing and a controller whose response remains useful under changing loads. High-voltage electronics, nonlinear mechanics, material drift, and fatigue must be considered together with control bandwidth and replacement needs. Progress should therefore be measured through repeatable task performance, including sensing, power delivery, and maintenance, as well as the properties of an isolated material sample.

Even without an unusual actuator, the body's structure can simplify control. A compliant insertion fixture can guide a misaligned peg, and an underactuated hand can conform to an object using fewer motors than moving joints~\citep{whitney1982rcc,dollar2010sdm}. This is a practical form of sharing the work between mechanics and computation. The same softness that accommodates contact can make shape harder to estimate. Electronic skins and learned sensing models help address that problem~\citep{shih2020electronicskins}. For physical intelligence, the useful question is which part of an interaction the body can handle reliably, and which changing circumstances still require observation, inference, and a new action.

\begin{figure}[tb]
  \centering
  \includegraphics[width=\linewidth]{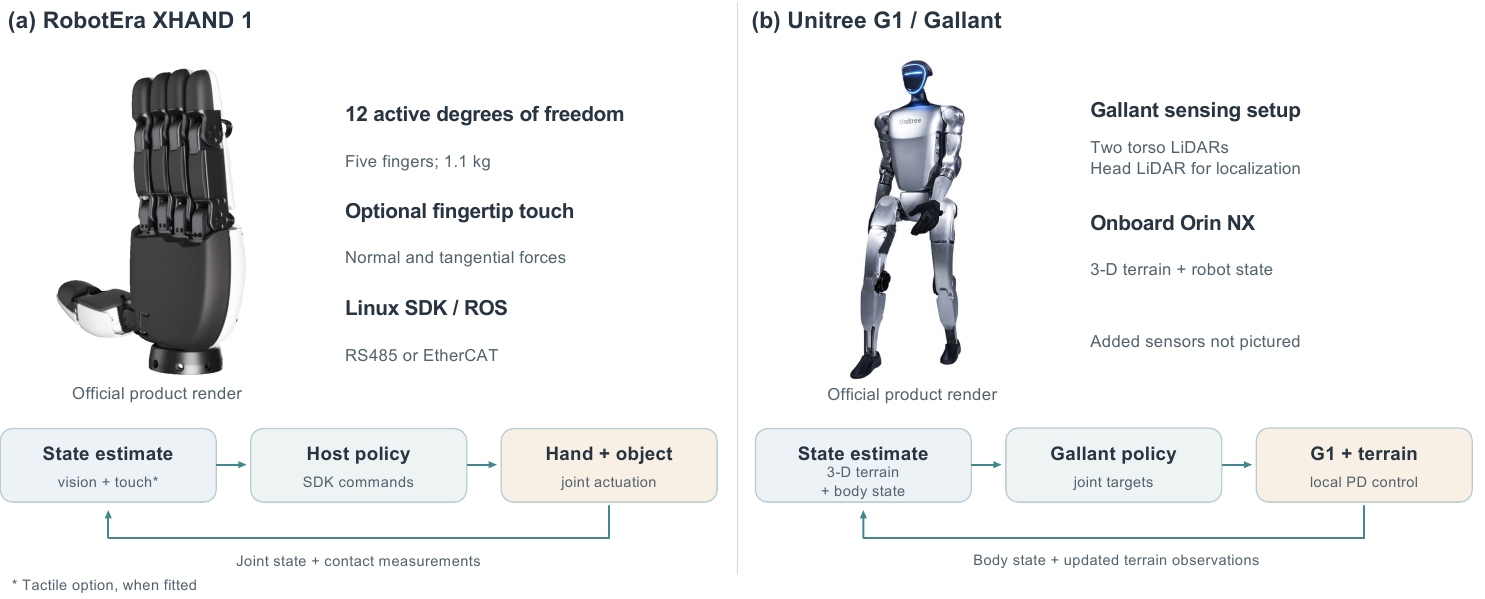}
  \caption{Product hardware and the feedback needed for Physical AI.
  (a) RobotEra XHAND~1 links visual and optional tactile observations to
  commanded finger motion and renewed contact measurements.
  (b) Gallant on Unitree G1 links 3D terrain perception to learned joint
  targets and local control~\citep{ben2025gallant}.
  Official product renders from \href{https://www.robotera.com/product/XHAND}{RobotEra} and \href{https://shop.unitree.com/products/unitree-g1}{Unitree Robotics}. The G1 image shows the base
  platform, without Gallant's additional LiDARs. The control diagrams
  summarize information flow, and the products are not shown to scale}
  \label{fig:c9cao:hardwarecases}
\end{figure}

The following cases examine a hand and a humanoid as complete sensing and actuation systems (Figure~\ref{fig:c9cao:hardwarecases}). Section~\ref{sec:c9:dexterous-cases} then compares three dexterous hands at the level of actuation, tactile sensing, and demonstration hardware.

\subsubsection{Case Study: RobotEra XHAND 1}
\label{sec:ch9-xhand-case}

RobotEra's XHAND~1 illustrates the measurement-motion loop at the level of one component. Its host sends joint targets, controller gains, and torque limits. The hand returns joint positions, motor-current information, temperature, and error status, together with contact measurements when the optional fingertip sensors are fitted~\citep{robotera2026xhandmanual,robotera2024xhandprotocol}. Section~\ref{sec:c9:dex-hardware} compares the mechanism, loading capacity, and communication timing with other hands. The question here is how these signals support correction during a grasp.

Vision can locate an object before contact, while joint feedback records how the fingers move toward it. Once a finger meets the object, target position alone no longer describes the interaction. Controller gains, effort limits, and transmission mechanics determine the load. Tactile measurements, when available, can reveal changes in contact that require a revised finger target. Motor current supplies useful actuator telemetry but does not locate that contact. The host must align these observations with the commands that produce them and respond while the grasp is still recoverable.

The same boundary applies to transferred demonstrations. ManipTrans uses Realman arms and Inspire hands, fitting 12 simulated hand coordinates to six hardware coordinates before smoothing and replay~\citep{li2025maniptrans}. Applying such motion to XHAND requires a new mapping through its joint axes, ranges, and contact surfaces. Matching a pose vector is only one step. The deployed hand must sustain the intended contact under load. Section~\ref{sec:c9:dex-xhand} examines how demonstration hardware and collection interfaces address this correspondence.

\subsubsection{Case Study: Unitree G1 and Perceptive Locomotion}
\label{sec:c9cao:g1case}
Gallant illustrates how a humanoid's useful workspace depends on the sensing and control system built around its body~\citep{ben2025gallant}. The task is to move a 29-degree-of-freedom Unitree G1 through stairs, gaps, narrow passages and overhead obstacles. Two torso-mounted LiDARs provide complementary views of the surrounding geometry. A head-mounted Livox Mid360 supports localization. A voxel representation preserves the vertical structure that a ground-height map discards, allowing the controller to account for both footholds and body clearance.

The software connects point-cloud processing, proprioception and learned control on an onboard NVIDIA Orin NX. A convolutional encoder extracts spatial features, which are combined with the robot state to generate joint commands. Sensor placement and the representation used during training must therefore remain consistent at deployment. Changing the visible space changes the information available for each action. In the initial deployment, LiDAR traffic and robot-state messages share a software pipeline and incur substantial communication delay. Restricting transmitted point clouds and using shared memory reduce this overhead~\citep{ben2025gallant}. These changes concern the age of the feedback reaching the controller, rather than the size of the policy alone.

The real-robot evaluation uses 15 trials per terrain and compares the full system with perception and training ablations~\citep{ben2025gallant}. This provides evidence for the integrated locomotion loop within the tested environments. On approach to a low ceiling, observed clearance and current body configuration must jointly determine posture. A correct map delivered too late cannot support that correction. The useful capability belongs to the complete perception-action loop.

\subsection{Device-Edge-Cloud Collaboration and Real-Time Communication}
\label{sec:c9:device-edge-cloud}
\subsubsection{Distributed Architecture}
The purpose of Physical AI is to enable intelligent agents, under finite computational resources, to interact continuously with the physical environment through perception, decision-making, and action and to accomplish a given physical objective in real time, effectively and safely. Resources in computation, energy, storage, communication, and control cycles must be allocated across perception, state estimation, prediction, planning, policy generation, and execution control at different nodes. These functions have different temporal scales, resource requirements, and spatial distances. This creates a central challenge for Physical AI. High-level computation tends to be low-frequency, computationally intensive, and spatially remote, whereas physical perception and interaction must be high-frequency, computationally limited, physically proximate, and temporally continuous.

\begin{figure}[!b]
  \centering
  \includegraphics[width=\linewidth]{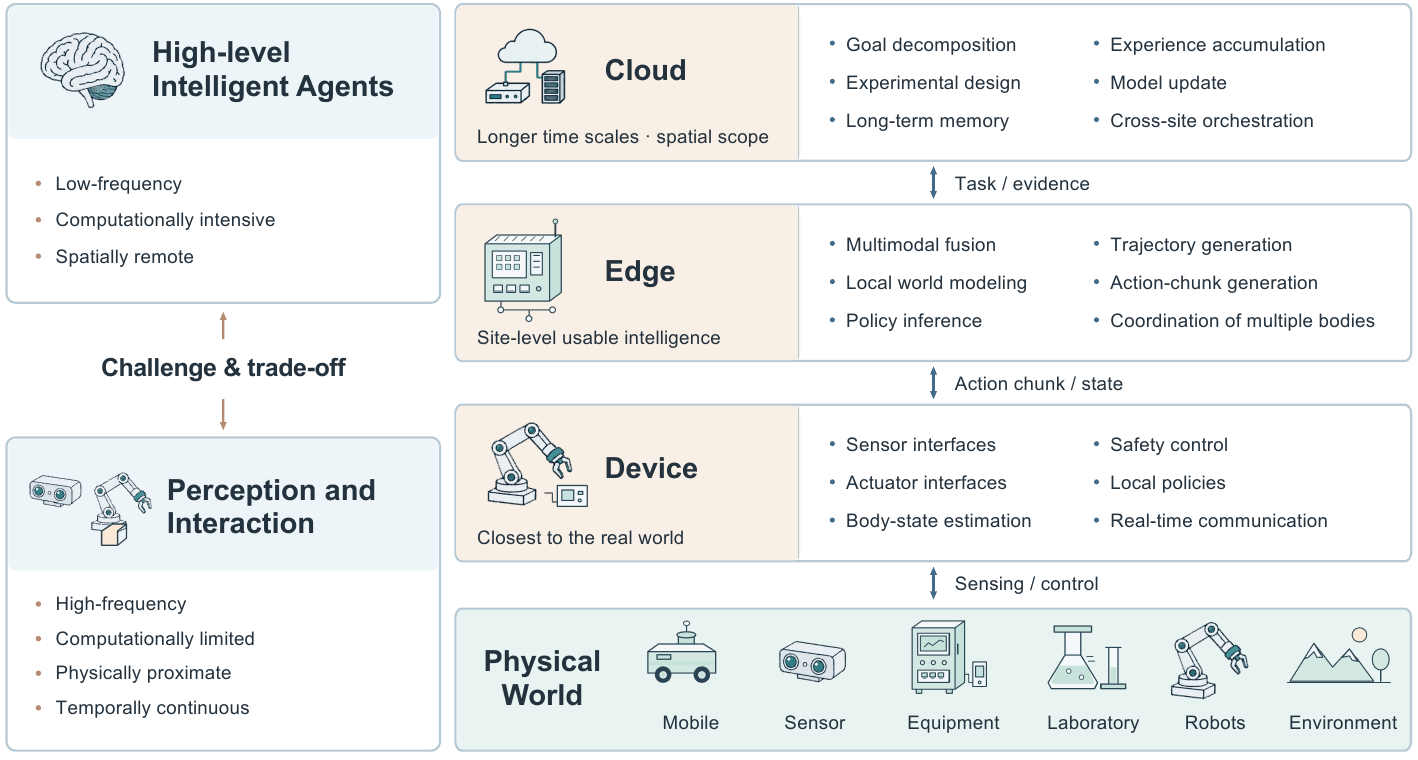}
  \caption{The challenge of physical AI and device-edge-cloud collaborative
  Architecture}
  \label{fig:c9yuan:challenge}
\end{figure}

The device-edge-cloud distributed architecture provides a way to balance these requirements (Figure~\ref{fig:c9yuan:challenge}). It decomposes a unified perception-decision-action loop according to distinct \emph{timescales, spatial scopes, and responsibility boundaries}~\citep{lee2026middleware}, organized around \emph{validity deadlines, distributed networks, and responsibility horizons}. The \emph{device} connects directly to the body, sensors, and actuators. It handles the fastest state perception, servo control, contact responses, and safety constraints so that the minimum safety-critical loop does not depend on the network. The \emph{edge} operates at the physical site with greater computational resources. It performs multimodal fusion, local world modeling, policy inference, trajectory generation, action-chunk generation, and coordination of multiple bodies, transforming site-level state into executable commitments that remain valid. The \emph{cloud} operates over longer timescales across tasks, devices, and sites. It supports goal decomposition, experimental design, long-term memory, experience accumulation, model updates, and cross-site orchestration. Computation can migrate according to task requirements, but physical responsibility cannot migrate arbitrarily. The closer a responsibility is to a physical consequence, the more it must be assumed by a trusted node close to the body.

Accordingly, device-edge-cloud collaboration is better understood in terms of \emph{migratable responsibility boundaries} rather than as a fixed three-layer computational hierarchy. When real-time communication is healthy, prediction, planning, and policy generation can migrate toward the edge or cloud in exchange for stronger model capabilities. When real-time communication becomes abnormal, responsibility must contract toward the physical object and switch to validated degraded modes.

Furthermore, device-edge-cloud collaboration exchanges node state and commitments with physical semantics. At least three categories of state should be distinguished. \emph{Physical state} includes pose, joint state, force, contact, limits, environmental conditions, and execution mode. It must carry sampling time, coordinate frame, units, and effective range. \emph{Action state} includes action chunks, task phase, resource occupancy, execution progress, and version information. It must distinguish among ``generated,'' ``sent,'' ``received,'' ``acknowledged,'' and ``physically acted upon.'' \emph{Evidence state} includes credit, measurement identity, experimental conditions, execution confirmation, anomalies, and recovery records. It determines whether a result is sufficiently grounded to be treated as a fact by subsequent reasoning. Confusing these categories produces characteristic failures in Physical AI closed loops, including treating a plan as a fact, transmission as execution, or an old observation as the current state.

The following sections address four questions within this Physical AI framework. What information must the device maintain, and what must it guarantee? How does the edge transform site-level state into valid action? How does the cloud provide low-frequency, high-level intelligence? How can real-time communication ensure that information transferred across these responsibility boundaries remains physically valid?

\subsubsection{Device-Side Computing}
The device is the computational and control node closest to the point of physical action in the Physical AI loop. It typically includes sensor interfaces, actuator interfaces, body-state estimation, safety control, local policies, and an adaptation layer for real-time communication with higher-level systems~\citep{ortega-kral_rio_2026}. Device-side computing must ensure that the physical system does not enter an unacceptable state simply because it is waiting for remote information, even under the worst real-time communication conditions. Hosting the largest model is secondary to this requirement.

The device must first establish a \emph{time-aligned multimodal state}. Cameras, encoders, force sensors, tactile sensors, and motor feedback do not inherently share the same sampling instant. A visual-frame timestamp may correspond to exposure or acquisition time. Joint state may originate from a controller cycle. Force signals may contain additional delay introduced by filtering, transmission, and mechanical response. Device state should therefore be represented by states carrying timestamps, source identities, safety attributes, spatial coordinate frames, units, and quality indicators, rather than simply by a collection of ``current values.'' Only then can the system determine whether different trusted observations describe the same contact event and, when a sensor is missing, represent the state as ``unknown'' rather than incorrectly reusing the previous frame.

The device must also maintain a \emph{local execution state machine}. The state machine should at least distinguish standby, preparation, execution, constrained execution, pause, fault, and recovery, while specifying which actuator actions are permitted in each state. Its purpose is to make device-native safety enforcement independent of interpretation by a high-level model. A policy model may recommend ``continue grasping,'' but if a joint is approaching its limit, force exceeds a threshold, or the real-time communication lease has expired, the device-side state machine must be able to reject the action. Policy output and execution permission are distinct. The policy proposes an action, whereas the safety layer determines whether it is currently allowed.

For real-time control, the device typically uses a fixed-period or event-triggered local closed loop. The real-time communication layer provides references, targets, parameters, or action chunks. It should not require every servo cycle to wait for a remote system to generate the next joint command. This prevents network jitter from entering the innermost control loop and turns network failure into an explicit operating state rather than an accident handled only after it occurs. Based on the remaining validity of an action chunk, local state, and safety rules, the controller can continue execution, execute in a degraded mode, perform a controlled stop, or stop immediately, thereby implementing device-native safety.

Device computation is also constrained by physical considerations. Larger models imply higher memory consumption, power draw, thermal load, and computational latency, and may also alter the weight distribution, power capacity, and onboard compute budget of a mobile system. Model size therefore cannot be discussed independently of the control cycle. For tasks requiring rapid contact feedback, a more complex but slower model may be unsuitable for direct participation in the execution loop. It can instead support slower action generation, candidate selection, or anomaly interpretation, while the fast loop remains the responsibility of lightweight models, controllers, and safety logic.

\subsubsection{Edge-Side Computing}
Edge-side computing provides site-level computation and coordination. It is closer to the physical process than the cloud while offering substantially more computational and coordination capability than the device. Edge nodes are well suited to policy inference, local world models, trajectory generation, resource coordination, and multimodal fusion \citep{community2026xpolicylab}. These computations, however, have physical control value only when their outputs still correspond to the current state of the site.

In this architecture, an \emph{action chunk} serves as a time-bounded execution commitment with explicit semantics. An action chunk should include, at minimum, the target device or body identity, reference time, validity interval, sequence number, control units, execution mode, and necessary constraints. The device can then determine whether the action still belongs to the current task, whether it has expired, and whether it has been superseded by a newer version. The key metric for an edge closed loop is not average inference time, but the \emph{end-to-end interval between the observation reference time and the instant at which the action begins to take effect}. This interval must be shorter than the validity deadline permitted by the physical action, rather than merely satisfying a requirement that model inference itself be fast enough. Loss of an action chunk must not be interpreted by default as ``continue executing the previous chunk,'' because doing so hides communication failure inside the control state. A more appropriate execution rule uses a \emph{lease}. When the lease expires without a new valid commitment, the device enters a predefined safe state.

The edge should also keep the \emph{world model decoupled from the controller}. A world model may predict ``what might happen if this grasp continues,'' but that prediction does not thereby acquire direct control authority. Its output can condition action generation, candidate-action selection, or risk estimation, while final execution must still pass through the device-control interface and safety state machine. This allows independent checks before execution and lets the world model be upgraded independently.

In site-level distributed collaboration, the edge is better suited to maintaining \emph{resource semantics} than to replicating all internal device data. For example, when multiple robots compete for the same tool, the edge needs information about tool occupancy, reservation expiry, task priority, and current execution phase. It does not need to continuously aggregate high-frequency joint data from every robot. Resource state can be encoded using leases, version numbers, or monotonic sequence numbers, making ``who owns the resource'' a state that can be checked rather than a conclusion inferred from whichever message arrives last.

\subsubsection{Cloud-Side Agents}
Cloud-based agents operate on the longest time scale and are suited to processing information across experiments, devices, and sites. They can maintain long-term memory, compare historical results, generate scientific hypotheses, plan resources, and select subsequent tasks, but should not directly assume millisecond-scale contact control. Communication between the cloud and the physical site therefore primarily concerns \emph{task state and evidence state}, rather than the next joint command.

The cloud must distinguish \emph{plan, request, confirmation, and fact}. A plan describes what is intended to happen. A request specifies what the site is asked to execute. A confirmation indicates that the site has accepted the request. A fact originates from an identified, timestamped physical measurement. These four concepts cannot substitute for one another. For example, when the cloud generates a task to ``close the valve,'' this does not mean that the valve has closed. Even if the actuator reports ``completed,'' sensor evidence may still be required to confirm the final physical state. Only states supported by site-level measurements should enter the evidence set used for subsequent experimental reasoning.

Accordingly, communication between cloud and site is better implemented through \emph{event-driven state synchronization} than through continuous replication of all raw signals. High-frequency data can be aggregated, compressed, or selectively uploaded on site. The cloud receives task-relevant measurements, summaries, anomalies, and evidence. For experiments that must be reproducible, uploaded records must also carry identities for samples, devices, experimental conditions, and software versions. Otherwise, historical results become difficult to interpret reliably.

When the cloud becomes unavailable, an action that has already started should not be extended indefinitely while waiting for a cloud response. The site should complete, pause, or stop according to device- and edge-level state. After recovery, the cloud should first re-read site-level facts and only then select the next action. In particular, ``the last transmission succeeded'' must never be interpreted as ``the experiment has completed.''

\subsubsection{Real-Time Communication}
The central metric for real-time communication in Physical AI is the physical validity of information when it arrives relative to the physical process, rather than throughput alone. Average latency, average bandwidth, and average packet-loss rate are therefore insufficient to characterize a physical closed loop. At minimum, one must also consider latency bounds, jitter, queueing time, observation age, action validity, reordering, duplication, and the duration of disconnection~\citep{sajid2026embodied}.

A simple serial end-to-end real-time constraint~\citep{diao2025task} can be written as
\begin{equation}
  T_{\mathrm{e2e}}
  =
  T_{\mathrm{sensor}}
  +
  T_{\mathrm{encoder}}
  +
  T_{\mathrm{uplink}}
  +
  T_{\mathrm{compute}}
  +
  T_{\mathrm{downlink}}
  +
  T_{\mathrm{decoder}}
  +
  T_{\mathrm{actuator}}.
\label{eq:c9yuan:e2e}
\end{equation}

The terms correspond, in order, to the stages of one remote closed loop, namely on-device sensing, observation encoding, uplink transport, computation at the edge or in the cloud, downlink transport, command decoding, and the onset of actuation. For one observation and its resulting action, the stage intervals cover sensing start to actuation onset without gaps or overlaps, including any queueing time. Their sum is $T_{\mathrm{e2e}}$, with stage boundaries specified and timestamps mapped to a common clock domain. $T_{\mathrm{deadline}}$ is not an arbitrary parameter chosen by the real-time communication system, but a deadline determined by the physical process, and $T_{\mathrm{e2e}}$ must be strictly less than it. Contact, balance, and high-speed motion may require deadlines close to the control cycle, whereas experimental steps and resource scheduling may tolerate delays of seconds, minutes, or even longer. A distributed physical AI can therefore contain multiple real-time classes simultaneously. A single ``real-time/non-real-time'' label cannot adequately describe all of its communication.

In device-edge-cloud collaboration, asynchronous execution and queueing affect which observation an action uses. The \emph{age of information (AoI)}~\citep{yates2021aoi} complements end-to-end latency by measuring information freshness. At an evaluation time, standard AoI is the interval since generation of the freshest state update available at the receiver. The observation age of an action measures the interval from generation of the observation actually used for that action to the action's execution. Both age calculations use timestamps mapped to a common clock domain. For the same update stream and receiver, the two ages coincide at action execution if the action uses the freshest available update. The action's observation age is greater if it uses an older update. These quantities should not be conflated with the simple serial latency estimate above. A backlog of action chunks can make nominal network latency appear normal while delaying execution and increasing the age of the observation used by the current action, potentially beyond its validity period. For real-time action generation, skipping queued observations that are no longer useful to the current control computation can prevent queue buildup. Older observations may still be retained for state estimation, sensor fusion, or historical analysis. For example, a model requiring only 5\,ms for inference can still produce an unusable action if its input waits 100\,ms in a queue. Conversely, a slower model may be entirely acceptable for second-scale task planning.

The action side requires complementary execution rules. An action prefix that has already begun execution generally cannot be arbitrarily rolled back, whereas an unexecuted suffix can be regenerated according to new observations. This distinction avoids excessive action oscillation while allowing new information to modify future behavior. When an action chunk expires, the system should stop extending the commitment rather than repeatedly copying the last action.

Real-time communication must also handle \emph{semantic loss}. Undetected message loss can leave the receiver with an incorrect view of task or safety state. Losing a frame used only for visualization may be harmless, whereas silently losing an event indicating that a device has entered a safe state could cause the cloud to schedule the next experiment incorrectly. Communication protocols must therefore distinguish discardable information from information that requires acknowledgment, and provide the latter with confirmation, sequence numbers, timeouts, and idempotent semantics \citep{zhang2024middleware,kluner2025automotive,chovet2024middleware,liang2023zenoh}.

In multi-body collaboration, real-time communication should likewise not force all bodies to share a single high-frequency state. When two robots compete for the same tool, the edge must make tool-occupancy and release events visible \emph{before a conflict occurs}. If occupancy information arrives late, its effect is equivalent to having no information. Force feedback within a body should remain in a local closed loop, because elevating it to an inter-body network only increases real-time communication latency and the risk of failure.

A real-time communication system with explicit responsibility boundaries should provide at least four capabilities, including \emph{timestamping, state versioning, validity control, and explicit fault representation}. Timestamping identifies when a state was observed, versioning identifies the plan or policy that produced it, and validity control determines whether it remains usable. Explicit fault representation identifies who is responsible for entering the safe state when the information is no longer usable. Together, these capabilities establish state and responsibility semantics for maintaining the Physical AI closed loop in addition to transporting messages.

\subsubsection{Self-Driving Laboratories}
A self-driving laboratory (SDL) provides a validation setting for device-edge-cloud collaboration \citep{gao2025unilabos}. A typical dry/wet experimental SDL simultaneously contains millisecond-scale device control, second-scale experimental operations, and minute-to-hour-scale scientific decision-making. Instruments and robots must execute rapidly. The site must continuously maintain the state of samples, devices, and the environment, while higher-level scientific-discovery agents typically operate in a relatively \emph{predefined experimental world}, running a ``hypothesis, execution, measurement, and next selection'' loop. The objectives, workflow, instruments, and basic interpretive framework for measurements are specified before an experimental campaign begins, and the system primarily optimizes parameters and actions within those predefined boundaries.

An open-governance architecture for SDLs emphasizes a transition from \emph{closed-loop optimization in a predefined world} to \emph{open-world scientific discovery}. This motivates designing the InternLab platform~\citep{li2026internlab} to advance Physical AI from a single-laboratory closed loop to a governable capability network across laboratories and institutions.

InternLab's device-edge-cloud distributed architecture organizes physical capabilities into a governed network without centralizing all laboratories into a single cloud. Scientific agents can use it to discover and compose experimental capabilities dynamically from nodes across the devices, sites, and cloud systems of multiple laboratories and institutions. Its three mutually coupled layers are \emph{Physical Intelligence \& Distributed Network} (the capability layer), \emph{Lab OS Runtime} (the execution and orchestration runtime connecting the capability and governance layers), and \emph{Native Governance} (the governance foundation).

The \emph{Capabilities Network Layer} transforms ``equipment owned by a laboratory'' into ``capabilities callable by a network.'' Robotic manipulation, scientific instruments, analytical models, and experimental skills are no longer merely private components within a site. They can be described as capabilities with explicit applicability conditions. The \emph{Capability Graph} describes which nodes can provide which capabilities under which conditions. \emph{Scheduling \& Routing} selects qualified nodes according to sample state, temporal constraints, instrument availability, and governance conditions. The \emph{Composition Engine} combines multiple capabilities serially or in parallel into an experimental route.

The \emph{Lab OS Runtime Layer} transforms continuously changing scientific intent into executable physical actions and brings the resulting outcomes back into the scientific loop. Its core can be abstracted into four runtime functions, namely \emph{Task Routing, Live State, Authorization Checks, and Trace Return}. Task Routing delivers tasks or action chunks to bodies, instruments, or sites with the required capabilities, rather than requiring every servo cycle to wait for a remote command. Live State maintains site state with timestamps, coordinate frames, and validity intervals, enabling higher-level decisions to determine whether observations remain fresh. Authorization Checks verify the necessary governance prerequisites before an action enters the physical system. Trace Return brings execution results, anomalies, and recovery information back into the next round of scientific reasoning.

The \emph{Native Governance Layer} is the foundation that enables InternLab to move from a conventional automation network to an open experimental network. InternLab explicitly contains at least five inseparable conditions, including \emph{Permission, Safety, Calibration, Provenance, and Credit}. Permission specifies who may invoke a capability, for what purpose, over which sample scope, and within which risk boundary. Safety is not a local property of a single action, but an end-to-end property spanning hardware, environmental monitoring, action validation, and fail-safe recovery. Calibration determines whether an instrument or sensor is currently in a measurement state that can be trusted. Provenance records sample identity, protocol version, data source, and operational chain so that cross-site evidence can be audited and reproduced. Credit preserves attribution for contributions across institutions. Most importantly, the design goal is not merely ``post hoc auditing,'' but governance-native execution. Governance conditions are part of the interface when a capability is defined and invoked, and they propagate together with the execution context. When any prerequisite fails, the capability should be rejected, degraded, re-authorized, or escalated to human handling.

The two InternLab pilot deployments~\citep{li2026internlab} demonstrate that the device-edge-cloud collaboration and responsibility model extends to an \emph{open spatial scale}. The local safety loop at the device, site-level state and resource coordination at the edge, and long-horizon scientific reasoning in the cloud therefore do not disappear when collaboration becomes cross-site. They become embedded within a network of capability discovery, task routing, governance checks, and evidence return. The evolution from SDL to the InternLab design extends Physical AI from a \emph{single, predefined, closed experimental loop} to an \emph{open-world loop that spans bodies, sites, and institutions and supports re-planning, re-composition, and governance-native execution}. This makes device-edge-cloud collaboration and real-time communication operational infrastructure for open-world, distributed Physical AI.

\subsection{Software Stacks and Development Infrastructure}
\label{sec:software-infrastructure}

A robot carrying a vial cannot wait for a language model to finish reasoning before responding to a slipping grasp. Nor does a completed arm trajectory establish that the vial reached the correct rack. These two observations define the software problem. Physical regulation must continue at the rate required by the body, while task execution must remain accountable to measured outcomes. Software connects these timescales through interfaces that specify what a component may command, what evidence it consumes, and what happens when that evidence or the component itself becomes unavailable.

Section~\ref{sec:c9:device-edge-cloud} establishes where computation can run across the device, edge, and cloud. This section examines the software boundaries that make that distribution workable. The architecture comprises nested feedback loops, hardware and data interfaces, a runtime for controllers and learned policies, and a task execution framework. This chapter uses \emph{Embodied Intelligence Harness} as its organizing term for the last of these. It brings together established mechanisms for skill composition, state management, execution monitoring, and recovery. The term denotes their integration for this chapter rather than a new middleware standard.

\subsubsection{Layered Software Architecture and Physical Interaction Loops}

Robotics software separates perception, estimation, planning, and control because these functions have different computational and physical requirements. Robot Operating System 2 (ROS~2) provides a widely used framework for composing such components through explicit interfaces~\citep{macenski2022robot}. Learning changes what can be implemented inside a component, and foundation models expand how tasks can be specified and decomposed. Neither change removes the need to assign ownership of state, control authority, and failure responses. A hierarchy is useful only if its feedback paths and failure boundaries are equally explicit.

\begin{figure}[htb]
  \centering
  \includegraphics[width=\linewidth]{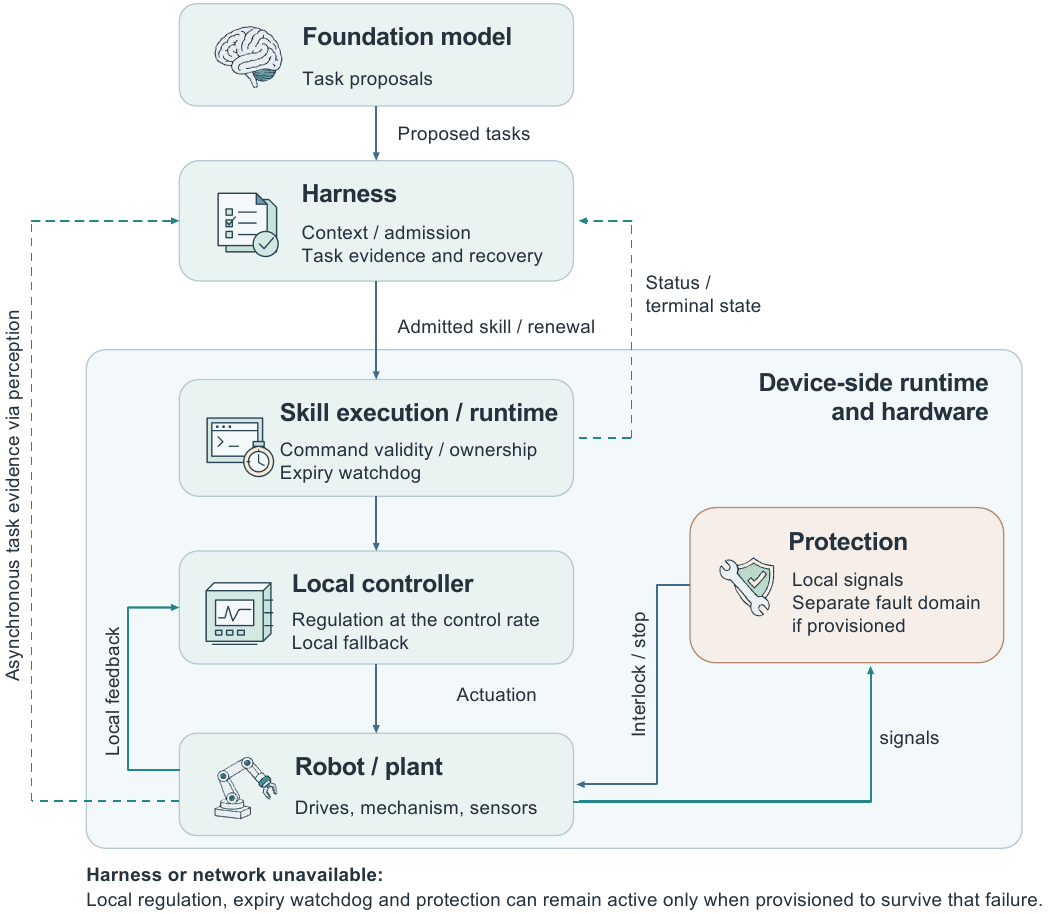}
  \caption{Nested feedback loops with distinct control authority. Task proposals pass through the Harness and a skill interface before reaching local execution. Sensor feedback closes the control loop directly on the device. A separate protection path can inhibit actuation. Dashed paths carry perception evidence and executor status to the Harness. Protection-event reporting to the supervisor is omitted for clarity. Intervention uses the local path}
  \label{fig:software:stack-hierarchy}
\end{figure}

Figure~\ref{fig:software:stack-hierarchy} distinguishes three loops. The \emph{task loop} interprets instructions, selects skills, checks their applicability, and evaluates their effects. It can revise a plan when a rack is occupied or an object's identity is uncertain. The \emph{local control loop} regulates motion or contact using measurements available on the device. Its execution period follows the plant dynamics and the controller design. The \emph{protection path} detects conditions requiring intervention and can inhibit or limit actuation within its specified response time. It may use drive electronics, a separate controller, or other independently provisioned hardware. These functions can exchange evidence without sharing a single execution path.

The task loop admits a request such as ``transfer vial A to slot B'' while the local controller processes encoder samples and motor updates. At admission, the executor assigns a skill instance a lease, which grants authority to issue task commands until a deadline on the executor's monotonic clock. Each command or action chunk also has an expiry determined by its intended execution interval and the validity of its observations. The executor applies task commands only while both the lease and the command remain valid. A new policy output cannot renew the lease. An admitted segment must fit within this joint interval. The executor begins its preconfigured fallback no later than expiry, allowing for detection and switching delay. That fallback has separate local authority and a specified completion or holding limit.

Crash, stall, and network partition exercise this boundary differently. A crashed Harness stops renewing its lease. A stalled task monitor must also stop renewal, even if a separate heartbeat thread is alive. Renewal follows a completed monitoring cycle with valid evidence. A partition can leave both sides running with divergent state. The executor therefore checks skill identity, supervisor epoch, sequence, and freshness before accepting a renewal. One implementation uses an executor-issued challenge with a local response deadline. Delayed responses cannot start a fresh lease merely by arriving. Expired or superseded authority requires re-admission. The lease duration must accommodate the admitted renewal delay while leaving time for the tested fallback. Until expiry, locally available regulation and protection continue within the existing authority.

The correct fallback depends on the physical state. A moving base may brake. A loaded gripper may need to maintain a limited holding force. A gravity-loaded joint may require controlled braking rather than abrupt removal of torque. Holding has thermal and energy limits and needs an escalation or unloading procedure. If sensing or the controller fails, an independent protection channel handles the faults assigned to it. Merely placing two processes on the same computer does not establish independence from kernel failure, power loss, or resource exhaustion. A task-level cancellation request is not an emergency-stop mechanism. Industrial implementations specify and validate the safety-related control and drive functions under applicable standards, including ISO~13849-1 and IEC~61800-5-2~\citep{iso138492023software94,iec61800}. The required functions follow the machine's risk assessment.

Runtime assurance provides a related logical architecture. An advanced controller operates while a monitor retains authority to transfer control to a verified fallback under stated assumptions~\citep{desai2019soter}. Its guarantees depend on the plant model, measurements, recovery region, and timely execution of the monitor and fallback. Implementing that logic in a separate fault domain is an additional engineering decision. The local sensing, compute, power, and actuation paths must survive the faults for which continued operation is claimed. This separation preserves the autonomous device loop emphasized in Section~\ref{sec:c9:device-edge-cloud} while slower layers improve task selection and recovery.

\subsubsection{Hardware Abstraction, Sensor Data, and Robotics Middleware}

A hardware interface should hide transport details while exposing the physical properties on which correct action depends. The \texttt{ros2\_control} framework, for example, separates controllers from hardware components through state and command interfaces, with a controller manager coordinating the read-update-write cycle~\citep{roscontrol2026architecture}. An interface named ``position'' is still incomplete without joint units, coordinate conventions, limits, supported modes, calibration state, and the conditions for switching modes. A simulated actuator and a physical drive can satisfy the same software signature while differing in delay, saturation, compliance, and failure behavior. Closed-loop substitution consequently requires testing the dynamics as well as the API.

Sensor interfaces need similar precision. An observation should carry its acquisition time, clock domain, frame, calibration identity, and validity information alongside the measurement. Covariance or another uncertainty representation is needed when the estimator relies on it. Receipt time alone conceals camera exposure, transport, and queueing delays. Combining a recent force sample with an old image can create a plausible state that never existed. The estimator must either align the measurements to a common time, propagate them using an explicit motion model, or mark the combination as unsuitable for the intended decision. Missing and invalid measurements must remain distinguishable from a measured zero.

Sensor validity can be expressed on the executor's clock. Let $\hat{t}_{\mathrm{s}}$ be the mapped acquisition time, with a clock-mapping error bounded by $\epsilon_{\mathrm{sync}}$. The estimated age of this particular observation at use time $t$, rather than standard AoI for the latest available update, is
\begin{equation}
  \Delta_{\mathrm{age}}(t)=t-\hat{t}_{\mathrm{s}}, \qquad \Delta_{\mathrm{age}}(t)+\epsilon_{\mathrm{sync}}\leq \Delta_{\mathrm{age}}^{\max},
  \label{eq:software:age}
\end{equation}
where $\Delta_{\mathrm{age}}^{\max}$ is a configured age limit for the particular observation and use. It can differ for a cached object label and a force estimate. The test applies at admission and at each intended actuation time, so age continues to grow across an action chunk. For commands that depend on this observation, it limits expiry to $\hat{t}_{\mathrm{s}}+\Delta_{\mathrm{age}}^{\max}-\epsilon_{\mathrm{sync}}$. Additional constraints may shorten the interval. If the clock mapping or its error bound is unavailable, the receiver needs another validated freshness mechanism or must reject the observation for that use. ROS~2 distinguishes ROS time, which can follow simulation, from steady time~\citep{foote2018clock94}. Physical watchdogs and leases use the latter. Simulation tests must state how their virtual timing maps to that mechanism.

Middleware carries these interfaces but does not supply their physical meaning. ROS~2 offers topics for streams, services for request-response interactions, and actions for longer operations with feedback and cancellation~\citep{macenski2022robot,ros22026actions}. The Data Distribution Service (DDS) specifies data-centric publish-subscribe communication and quality-of-service (QoS) policies, including reliability, history, deadline, and liveliness~\citep{omg2015dds94}. Zenoh combines publish-subscribe with query-based data access, and \texttt{rmw\_zenoh} integrates it beneath ROS~2's middleware interface~\citep{zenoh2026overview,rmwzenoh2026design}. These technologies occupy different layers. ROS~2 defines the action protocol above the underlying DDS or Zenoh transport.

QoS choices should follow the consequences of delay and loss. A high-rate state stream may benefit from a shallow queue that favors recent samples. A configuration transaction may require acknowledged delivery. Retransmission can improve delivery while increasing age, and a growing queue can turn a responsive estimator into a consumer of historical data. DDS deadline and liveliness events make some communication failures observable. They do not stop a motor or prove that a task is progressing. Backend support also differs. The Jazzy implementation of \texttt{rmw\_zenoh} does not implement deadline or lifespan policies. A design that depends on those policies must verify the selected release and enforce the necessary command-expiry behavior locally~\citep{rmwzenoh2026design}.

The resulting boundary is practical. Middleware transports typed information. The application establishes freshness, ownership, and permissible use. A command should identify its producing skill and sequence, its coordinate and control modes, and its validity interval. The receiver rejects incompatible or expired commands before they enter the actuator path. Configuration changes are versioned and checked before activation. These conventions allow a developer to distinguish a transport failure from a stale observation, a conflicting controller, or an invalid request, each of which calls for a different response.

\subsubsection{Real-Time Runtimes and Learned Policy Integration}

A controller needs its next input before a deadline, whereas a learned policy may have variable inference time. Running both in one callback chain makes physical timing depend on model execution, memory allocation, and accelerator contention. ROS~2's real-time guidance identifies page faults, dynamic allocation, and indefinitely blocking synchronization as hazards in a time-critical path~\citep{ros22026realtime}. Deployment therefore separates the bounded control cycle from inference, logging, visualization, and network-facing task services. Separation may require dedicated threads, priorities, memory discipline, processor allocation, or a separate controller. The necessary arrangement depends on the faults and interference the system must tolerate.

For a periodic local update, a useful timing budget is
\begin{equation}
  C_{\max}+B_{\max}+J_{\max} < T_{\mathrm{deadline}} \leq T_c,
  \label{eq:software:deadline}
\end{equation}
where $C_{\max}$ bounds computation and input/output (I/O) on the selected path, $B_{\max}$ bounds blocking and scheduling interference, $J_{\max}$ bounds release jitter, $T_{\mathrm{deadline}}$ is the deadline, and $T_c$ is the control period. This is an accounting condition for that path, not a complete schedulability or stability proof. The bounds require evidence for the actual workload and hardware. A finite benchmark's largest observed latency is not automatically a worst-case bound. A complete assessment also includes sensing delay, communication, actuator response, and the controller's tolerance to missed updates.

Learned policies fit this structure through explicit adapters. OpenVLA illustrates a vision-language-action model that maps observations and instructions to robot actions~\citep{kim2024openvla}. ACT predicts sequences of actions for fine-grained manipulation~\citep{zhao2023}. Their outputs may be joint targets, end-effector increments, or other embodiment-specific commands. The adapter must preserve the model's normalization, reference frames, action period, and control mode, and establish compatibility with the receiving controller. Position, velocity, and force checks constrain admission. Collision avoidance and contact stability additionally depend on the controller and its operating envelope.

\begin{figure}[htb]
  \centering
  \includegraphics[width=\linewidth]{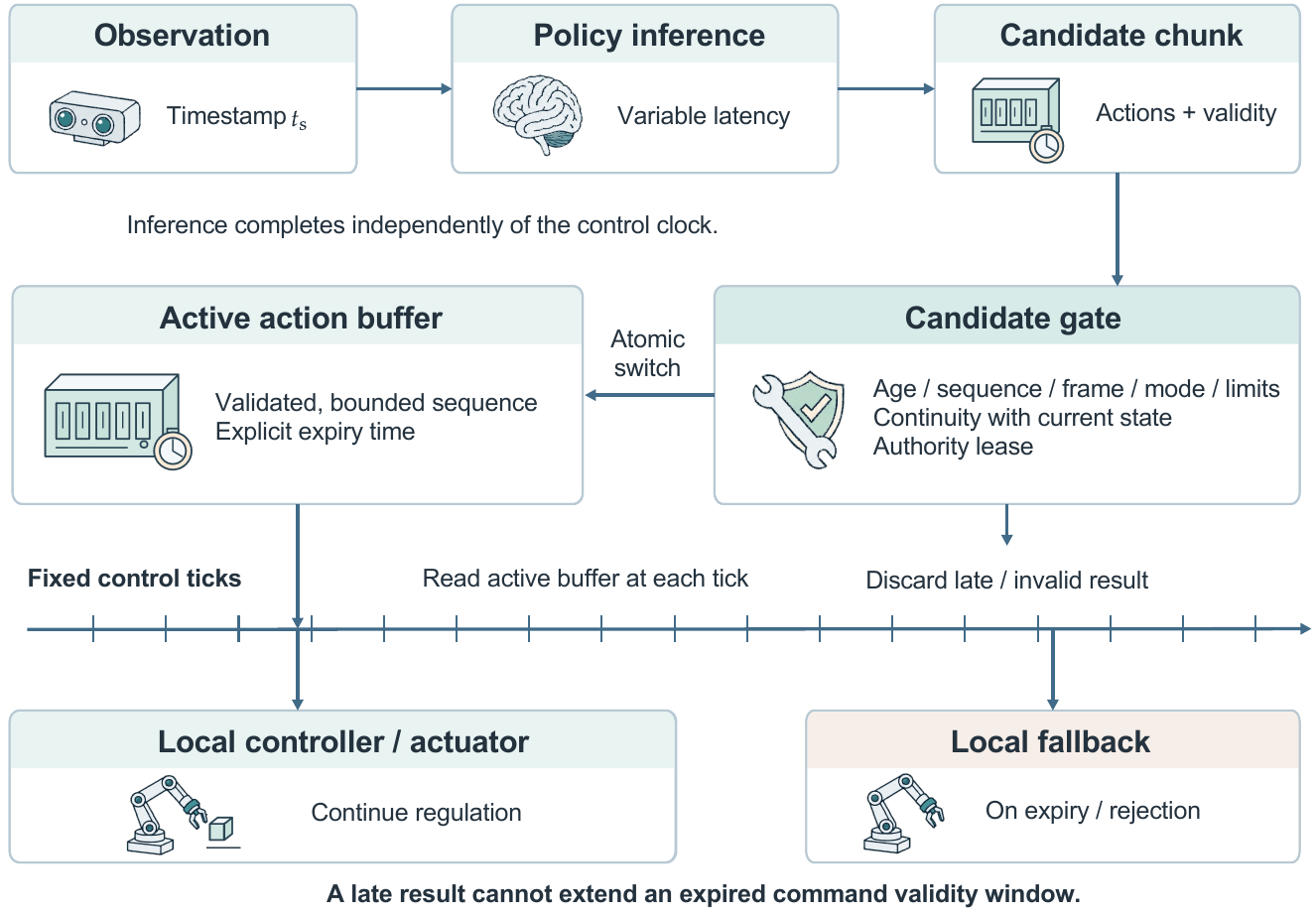}
  \caption{Asynchronous inference coupled to local execution. Variable-duration policy computation produces candidate action chunks, while the local loop consumes a validated buffer at its own period. Admission and a controlled buffer switch connect the two. Expired results are discarded. Loss of valid command authority activates a local fallback. The control ticks are schematic. The gate checks continuity as well as validity before the atomic buffer switch}

  \label{fig:software:policy-runtime}
\end{figure}

Figure~\ref{fig:software:policy-runtime} shows an asynchronous arrangement. An inference worker consumes a timestamped observation and produces a candidate action sequence with a sequence identity and validity horizon. A local executor continues its fixed-rate cycle from an already admitted buffer while the candidate is computed. Before a buffer switch, the receiver checks observation age at the scheduled use times, command order, frame, operating mode, lease, and relevant limits. It rejects late results whose assumptions are no longer valid. The gate also applies a controller-specific continuity rule before the buffer is switched atomically. Otherwise the handover can introduce a discontinuity in velocity, force, or target pose. To avoid fallback, a replacement must finish inference, transport, and admission before the current buffer expires. When inference runs at the edge, the freshness budget includes both outward observation transport and returning commands. A small learned controller with a demonstrated execution bound can instead run inside the periodic path and is subject to Eq.~\eqref{eq:software:deadline}.

Action chunks reduce the frequency at which a model must be invoked, but they also commit the robot to predictions made from an earlier observation. A longer chunk can absorb inference delay while delaying the response to a changed scene. Real-time chunking addresses this interface by generating a new chunk asynchronously while accounting for actions committed during inference~\citep{black2025rtc_ch6}. It coordinates policy execution under delay through a continuation consistent with the committed prefix. Deployment still requires measurement and validation of the complete runtime. Contact-sensitive control and protection still require timely local measurements, even when the high-level action sequence remains valid.

The runtime should make buffer expiry, inference overrun, and observation dropout first-class events. Repeating the last action indefinitely can be especially dangerous when the action is a velocity or torque. At the earlier of lease and command expiry, the executor begins the plant-specific fallback described above, records the transition, and prevents a delayed response from restoring authority. Resumption requires fresh state and an explicit re-admission. Thus learning can change the motion strategy without making the availability of a large model a prerequisite for every control update.

\subsubsection{The Embodied Intelligence Harness and Skill Contracts}

\begin{table}[!t]
\centering
\small
\caption{An illustrative \texttt{transfer\_vial} skill contract. Symbols denote configured bounds, not universal thresholds. The local executor enforces command validity and fallback independently of the task supervisor}
\label{tab:software:skill-contract}
\renewcommand{\arraystretch}{1.18}
\begin{tabularx}{\linewidth}{@{}>{\raggedright\arraybackslash}p{0.19\linewidth}>{\raggedright\arraybackslash}X@{}}
\toprule
Contract field & Concrete interface and obligation \\
\midrule
Input & \texttt{request\_id}, vial identity, source and destination fixture IDs, pose and frame, calibration version, payload class, and maximum task duration $T_{\mathrm{task}}$. Duplicate IDs query the recorded execution state rather than starting another transfer. \\
Preconditions & Identity and source occupancy confirmed, destination empty, pose age satisfying Eq.~\eqref{eq:software:age}, compatible gripper and payload, local controller ready, required interlocks valid. Unknown conditions trigger inspection or rejection. \\
Resources & Exclusive arm and gripper authority, reservation of the destination slot and shared motion region. Acquisition is bounded. Partial acquisition is released on rejection. \\
Timing & Task duration $T_{\mathrm{task}}$ starts at local admission. Each phase has its own timeout. Task commands expire at the earlier of the local lease deadline and their observation-limited expiry. Fallback begins no later than that point, independently of Harness cancellation. \\
Progress & Timestamped phase, controller status, grasp/contact evidence, observation age, and remaining authority. ``Goal accepted'' and ``target pose reached'' are intermediate states. \\
Completion evidence & After release and retreat, destination occupancy and vial identity are observed, the source is empty, and the gripper no longer holds the vial. Return evidence timestamps, confidence or validity flags, and the relevant log references. \\
Failure and recovery & The executor reports \texttt{aborted} or \texttt{canceled} with the last verified state. A missing result leaves the Harness with an application-level \texttt{outcome\_unknown}. Apply the local fallback and re-observe before retrying or releasing resources. \\
\bottomrule
\end{tabularx}
\end{table}

The Harness governs the task loop across these interfaces. It maintains the current goal and evidence, resolves requests against available skills, arbitrates resources, and tracks execution to an observable result. Hierarchical reinforcement learning formalizes temporally extended actions through initiation, policy, and termination components~\citep{sutton1999options}. Behavior trees provide established mechanisms for composing and monitoring robot behaviors~\citep{IovinoEtAl2022}. These precedents explain much of the required machinery. Language-conditioned skill selection with affordance scoring, as in SayCan, provides a concrete precedent for connecting model proposals to robot capabilities~\citep{ichter2023saycan}. In the framework used here, the model proposes skills and arguments. Admission, monitoring, and completion decisions follow explicit contract checks over timestamped evidence.

A skill contract gives the supervisor something more precise than an action name. Table~\ref{tab:software:skill-contract} develops one example for moving a capped vial between known fixtures. Its parameters and conditions are illustrative design choices, to be set from the actual robot, payload, and procedure. The example separates request acceptance, physical progress, and completion evidence. A ROS~2 action can carry the goal, feedback, result, and cancellation protocol, but resource ownership, physical preconditions, command leases, and evidence requirements remain application responsibilities~\citep{ros22026actions}.

Admission checks establish that a skill can start. Execution monitoring determines whether its assumptions continue to hold~\citep{pettersson2005monitoring94}. A fresh target pose may become stale during an approach. Another device may occupy a reserved region. A grasp may be established and then lost. The monitor should therefore associate each predicate with evidence and a validity interval, and distinguish false from unknown. Unknown identity or destination occupancy cannot be converted to success merely because the motion controller reports completion. Learned perception can misidentify a vial or occupancy state. Completion thresholds need calibration, and consequential decisions may require complementary sensing or repeated observation. Conflicting evidence triggers re-observation or a recovery whose own conditions are checked.

Cancellation also has physical semantics. A ROS~2 action server may accept a cancellation request while the underlying mechanism still needs time to decelerate or secure its load. The Harness should wait for the executor's terminal state and physical evidence before allocating its resources to another skill. If communication fails after a vial has been placed but before the result arrives, a timeout leaves the outcome uncertain. Retrying the transfer immediately could act on an empty fixture or move a second object. Persistent request identities, execution records, and observation of the fixtures allow the supervisor to reconcile state after reconnection. Reliable message delivery alone cannot provide exactly-once physical effects.

Restarting the Harness reconstructs task state from the executor and fresh observations before any new authority is issued. The application-level ``outcome unknown'' state describes missing physical evidence or a missing result. It is distinct from the unset status represented by ROS~2's \texttt{STATUS\_UNKNOWN}.

\subsubsection{Development Infrastructure and Evidence for Deployment}

Development infrastructure should test the boundaries just described. A reproducible episode records acquisition and actuation times, frame and calibration versions, model and controller identities, skill requests, authority changes, and the evidence used to declare completion. Tracing the path from a sensor event through callbacks to a command helps distinguish inference delay from scheduling or transport delay. \texttt{ros2\_tracing} provides instrumentation for such analysis in ROS~2~\citep{bedard2022tracing94}. High-rate recording itself consumes resources, so its effect on the control workload must be measured or isolated.

Recorded-data replay is valuable for comparing estimators and diagnosing failures, but an unchanged sensor log cannot establish the closed-loop outcome of a changed policy. Different actions would have produced different observations. Software-in-the-loop (SIL) testing runs the relevant software against a simulated plant to examine logic, interfaces, and responses to modeled disturbances. Hardware-in-the-loop (HIL) testing runs target firmware on target control electronics against a simulated plant, exposing processor and scheduling effects. Sensor and actuator I/O may still be emulated through a different link, as in PX4's hardware-in-the-loop (HITL) arrangement~\citep{px42026simulation}. The test report must identify which physical interfaces were exercised. Gripper friction, camera exposure, and the complete machine's stopping response require physical trials.

\begin{figure}[!t]
  \centering
  \includegraphics[width=\linewidth]{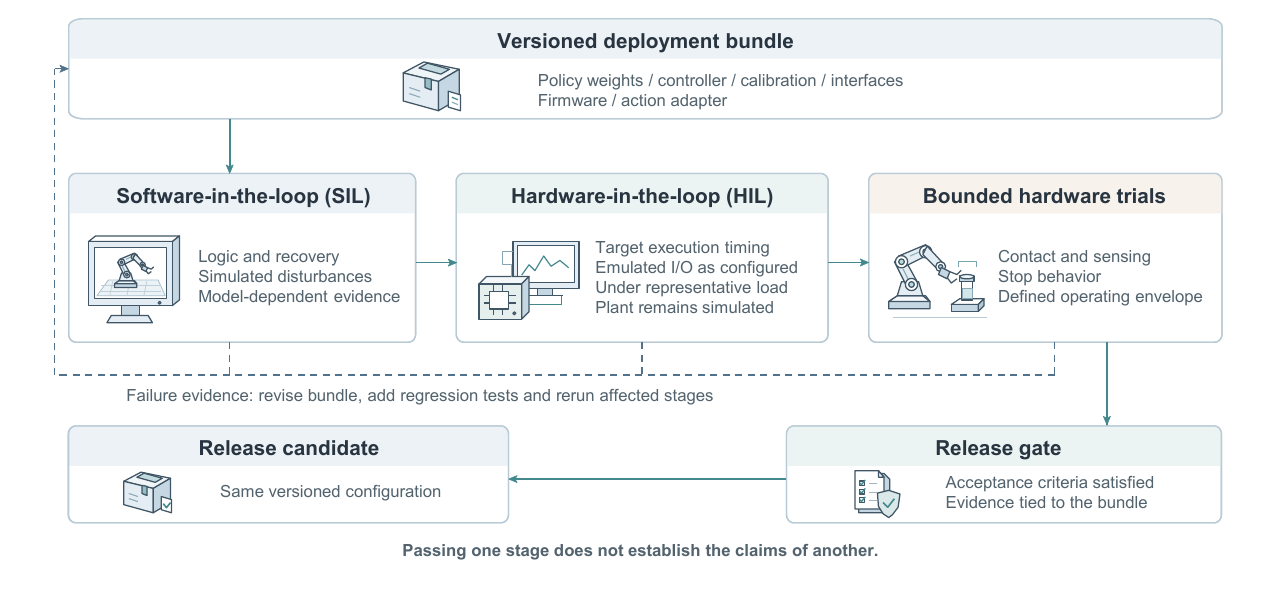}
  \caption{Validation stages answer different questions. SIL tests execution logic against a modeled plant. HIL adds target electronics and timing. Bounded physical trials expose sensing, contact, and actuation effects absent from the model. Release decisions apply to a versioned combination of model, firmware, controller, action adapter, calibration, and interfaces. Failure evidence drives component revisions and regression tests}
  \label{fig:software:continuous-deployment}
\end{figure}

Figure~\ref{fig:software:continuous-deployment} organizes these stages around explicit claims. For the vial-transfer skill, SIL can inject a stale pose, a duplicated request, or a lost acknowledgment. HIL can expose queue growth, inference contention, and watchdog delay. Bounded hardware trials can test grip loss and verify a loaded stop. Terminating the Harness, suspending its monitoring loop while a heartbeat remains alive, and partitioning the network are particularly informative. Local regulation and assigned protection must remain available, command authority must expire as specified, and reconnection must reject queued commands and renewals from obsolete epochs. Tests should also challenge sensor loss and the shared resources that could defeat the intended isolation.

The evidence should include the operating conditions, observed latency distribution and misses, age at actuation, time to fallback, recovery outcome, and the physical criterion for success. A release bundles the policy weights, preprocessing and action adapter, controller settings, middleware configuration, firmware, calibration, and skill schema used to obtain that evidence. Changing one of these can invalidate a previously established interface or timing assumption. Regression tests and staged physical trials determine whether the affected claims still hold. Binding evidence to this configuration provides the software basis for the sustained reliability and fleet deployment considered in Section~\ref{sec:c9:reliability}.


\subsection{Reliability and Deployment at Scale}
\label{sec:c9:reliability}


A vial-handling robot may complete a demonstration and still fail to produce usable samples over a working day. A cap slips after the fingers warm up, a dispenser delivers less liquid than commanded, or a restart repeats a transfer already completed. Each failure breaks a different connection between an intended action and its physical result. Reliable Physical AI requires that connection to remain observable and correctable. The system must detect a deviation, determine what can still be recovered, and verify the result before proceeding. Experience from these events can improve subsequent behavior, provided the cause is identified and the change is tested. This section follows that loop from one task to a fleet of machines.

\subsubsection{From Successful Attempts to Sustained Physical Output}
\label{subsec:sustained-physical-output}

Reliability concerns continuity of correct service. Availability concerns whether a system is ready to work~\citep{avizienis2004dependable}. For a physical task, correct service needs an observable endpoint. A sample is valid only if its identity, quantity, and required process conditions are preserved. Command delivery, completed trajectories, and apparent grasps are intermediate events. Their success matters only through the result they enable. A robot that remains powered throughout a shift but repeatedly requires a person to rescue its samples has poor autonomous productivity.

The measurements should therefore follow the task. Encoders establish arm configuration, touch can reveal loss of contact, and a balance can verify a transfer. Agreement among several signals is useful when they provide independent evidence. Several estimates derived from the same incorrect calibration may agree while the sample is wrong. The task-state observability of Section~\ref{sec:c9cao:observability} thus becomes a reliability requirement. The system must distinguish states that call for different actions while there is still time to intervene. Table~\ref{tab:c9cao:layers} links this evidence to the scale of correction.

\begin{table}[tbp]
  \centering
  \caption{Reliability depends on matching physical evidence to the scope
  of correction. A failure observed at one level may originate at another}
  \label{tab:c9cao:layers}
  \footnotesize
  \setlength{\tabcolsep}{4pt}
  \begin{tabularx}{\linewidth}{@{}>{\raggedright\arraybackslash}p{1.65cm}*{3}{>{\raggedright\arraybackslash}X}@{}}
    \toprule
    Level & Example deviation & Evidence needed & Scope of correction \\
    \midrule
    Component & Gripper response changes after warm-up
      & Current, temperature, measured motion and load
      & Reduce duty and inspect the drive or transmission \\
    Skill & Cap slips during opening
      & Contact change, cap motion, applied torque
      & Re-establish support and revise the grasp \\
    Workflow & Vial contents disagree with the task record
      & Sample identity, measured quantity, verified checkpoints
      & Reconcile progress and recover or reject the sample \\
    Fleet & Failures rise after a policy release
      & Outcomes by hardware, calibration and software version
      & Isolate affected configurations and validate restoration \\
    \bottomrule
  \end{tabularx}
\end{table}

\begin{figure}[tbp]
  \centering
  \includegraphics[width=0.72\linewidth]{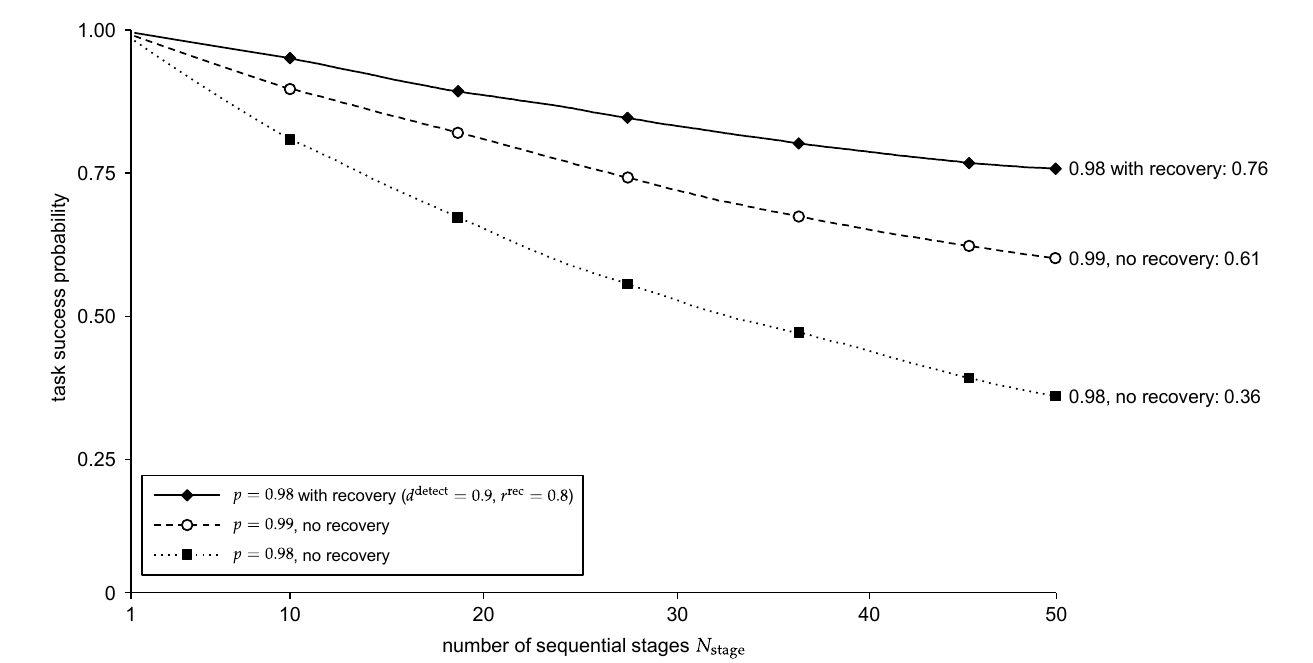}
  \caption{Detection and recovery change the probability of completing a
  50-stage task. In Eq.~\eqref{eq:c9cao:ptask}, $p=0.98$, $d^{\mathrm{detect}}=0.9$, and
  $r^{\mathrm{rec}}=0.8$ give an effective stage success of 0.9944 and about 76\% task
  success, compared with 36\% for $p=0.98$ and 61\% for $p=0.99$ without
  recovery. These are illustrative calculations under the stated
  assumptions, not measured robot performance}
  \label{fig:c9cao:horizon}
\end{figure}

Small errors accumulate over long procedures. Fifty independent steps that each succeed 98\% of the time give only about 36\% success for the whole sequence ($0.98^{50}\approx0.364$). Detection and recovery can change this outcome. Let $p_k$ be the initial success probability at stage $k$, $d_{k}^{\mathrm{detect}}$ the probability of detecting an initial failure, and $r_{k}^{\mathrm{rec}}$ the probability that one recovery attempt succeeds given that failure was detected. A simplified model gives
\begin{equation}
  P_{\mathrm{task}} = \prod_{k=1}^{N_{\mathrm{stage}}}
  \bigl[p_k+(1-p_k)d_{k}^{\mathrm{detect}} r_{k}^{\mathrm{rec}}\bigr].
  \label{eq:c9cao:ptask}
\end{equation}
Figure~\ref{fig:c9cao:horizon} shows the effect of this additional path to completion. The product assumes independent stages and successful recovery without hidden damage. Shared calibration errors, accumulated wear, and contamination violate those assumptions. The equation also omits recovery time. A system can eventually succeed yet produce little usable work per hour. Full-workflow tests must therefore measure both outcome and time.

Finite trials also limit what can be claimed. Twenty successes without a failure leave an exact one-sided 95\% upper confidence bound of about 14\% on failure probability, under independent trials with a fixed protocol~\citep{hanley1983zero}. Evaluation should state the workload, initial conditions, trial count, allowed assistance, and stopping rules. All attempted tasks, including aborted ones, belong in the record.

\subsubsection{Detecting and Diagnosing Failures from Physical Evidence}

A monitor must recognize both an abnormal interaction and a missing task result. Rising torque with little movement may indicate a jam. Ordinary joint motion with an unchanged balance reading may indicate an empty transfer. The first can require a rapid local response, while the second requires knowledge of the process. The sensor and actuator interfaces in Section~\ref{sec:c9:sensing-actuation} provide the measurements. The runtime in Section~\ref{sec:software-infrastructure} must associate them with the action being attempted and its expected outcome. A successful command acknowledgment closes neither loop.

Detection is easier to interpret when it compares a prediction with a measurement tied to the task. A dynamics observer can compare expected and measured joint loading, subject to its calibration and model assumptions~\citep{haddadin2017collision}. A task monitor can check whether a cap rotated, a vial moved with the gripper, or mass changed by the required amount. These checks should be conditional on the current phase. Resistance is expected when tightening a cap but may indicate a fault before the threads engage. A fixed threshold applied without that context can either stop useful contact or overlook an abnormal one.

Learned monitors can supplement these checks. Sentinel combines action consistency tests with vision-language assessment of task progress~\citep{agia2024sentinel}. It provides evidence for detecting failures during execution. Detection alone does not establish a correct recovery. Likewise, a plausible explanation from a model remains a hypothesis. A stopped finger may reflect an obstructed object, a current limit, overheating, or loss of communication. Synchronized images, encoder readings, currents, temperatures, and command records help separate these causes. Their timestamps matter because evidence collected after the robot has withdrawn may no longer describe the failed contact.

Detection quality should be measured where it changes the physical outcome. Report missed failures, false stops per scheduled operating hour, and the time from a detectable deviation to the protective response. Delays include sensing, transport, inference, and actuation, as discussed in Section~\ref{sec:c9:device-edge-cloud}. An alarm raised after a fragile vial breaks has a different value from one that allows the grasp to be corrected. Evaluate monitors on representative faults as well as normal contact transitions. High accuracy on a dataset dominated by uneventful motion is insufficient.

\subsubsection{Recovering Progress and Handling Irreversible Actions}

Recovery begins by establishing what is still true. After a slipping cap, the vial may remain supported, the cap may have moved, and the threads may already be partly disengaged. Returning the arm to a previous pose does not restore that state. The system should first preserve support and limit further loading, then acquire enough evidence to choose a feasible continuation. A new view, a gentle probing action, or a quantity check can be more useful than another full attempt. These actions use interaction to reduce uncertainty, which is a central capability of Physical AI.

A recovery should change the conditions responsible for failure. A missed grasp may permit a revised approach. A jam may require controlled withdrawal and realignment. A thermal limit may require reduced duty or cooling. Force, travel, retries, and elapsed time need task-specific bounds. If the cause remains unresolved or those bounds are reached, the robot should stop or request assistance. The physical checks required to resume belong to the recovery procedure itself.

Long tasks need checkpoints defined by verified physical outcomes. After a transfer, the system records sample identity, measured quantity, and completion status outside the policy's transient state (Section~\ref{sec:physical-goals}). On restart, it reconciles that record with the scene before issuing a new command. Otherwise, replaying a command whose acknowledgment was lost can double a dose. An operator may also have moved the container, or a sample may have exceeded its usable holding time. The checkpoint preserves what was verified. Renewed observation establishes what remains valid now.

Some actions admit no useful rollback. Mixing the wrong reagent cannot be undone by restoring a software version. Before an irreversible step, the system should verify its identity and process preconditions and resolve uncertainty while correction is still possible. Action-conditioned world models can simulate candidate interactions. UniSim uses predicted visual outcomes to train planners and control policies~\citep{yang2024unisim}. These predictions may still omit contamination, weak contact, or damage invisible in an image. They should guide investigation, while the actual process supplies the evidence for committing the action and accepting its result.

\subsubsection{Learning from Operation Without Concealing Hardware Drift}

The same command can produce a different interaction as pads wear, a camera mount shifts, or actuator temperature rises. Changes also arise from new objects and from states reached after the policy's own mistakes. These forms of distribution shift demand different remedies. A loose camera mount requires repair and recalibration. Repeated failure on a correctly observed new cap geometry may justify additional demonstrations or a revised control policy. Retraining on every deviation risks teaching the model to compensate for a deteriorating machine.

The operating record should retain the observation, intended action, measured response, hardware and software configuration, intervention, and verified outcome. A successful human rescue is especially informative when it records which state prompted the intervention and what changed. DAgger addresses learning from states visited by the learner through expert feedback~\citep{ross2011dagger}. Sirius studies learning from robot execution with human monitoring and interventions~\citep{liu2023sirius}. These approaches support improvement from experience. Intervention frequency and operator time remain part of the deployment result, including the monitoring Sirius requires.

Simulation can test a proposed correction across variations that are costly to reproduce physically. RialTo reconstructs real environments for simulation-based policy improvement~\citep{torne2024rialto}, while dynamics randomization varies physical parameters to reduce dependence on one nominal model~\citep{peng2018}. The relevant variation should follow the diagnosed mechanism. Friction is relevant to slip, delay to late correction, and camera placement to an observation mismatch. Randomizing unrelated parameters cannot establish that the observed failure has been addressed.

SIMPLER evaluates agreement between simulated and real policy performance within tested manipulation settings~\citep{li2024simpler}. Such agreement makes simulation useful for screening candidates, but a favorable ranking does not establish adequate absolute reliability. A revised policy needs fresh physical trials on the failure conditions, earlier successful tasks, and representative operating variation. Development episodes and acceptance trials must be separated. Fast control feedback corrects the current interaction. Learning changes future behavior and therefore requires this additional validation loop.

\subsubsection{Deploying Compatible Configurations Across a Fleet}

Scaling a policy means preserving its observation-action relationship on multiple physical machines. Differences in camera placement, fingertip wear, controller gains, and firmware can change that relationship even when robots share a model name. Common collection procedures, as used in DROID, help make experience comparable across sites~\citep{khazatsky2024droid}. Deployment additionally requires evidence that each supported configuration still produces the intended physical outcome under its assigned workload.

Consider a new cap-opening policy for a fleet of vial-handling robots. The release manifest identifies model weights, observation preprocessing, action units and scaling, controller code and gains, drive firmware, and supported hardware revisions. It also binds each robot to its own checked camera, tool, and force-sensor calibration, using a specified calibration procedure and tolerances. Calibration values cannot be copied between machines merely because their parts have the same names. The validated unit is the compatible configuration. Installing new weights with an old action convention can turn a useful learned correction into an unintended movement.

After interface, replay, simulation, and hardware-in-the-loop checks, a small subset runs supervised trials spanning the intended hardware and workload. Robots on the previous release provide a concurrent comparison. Acceptance criteria are fixed beforehand and cover sample quality, completion, contact loads, temperature, interventions, and recovery time. The trials include repeated cycles and warm actuators. Expansion depends on adequate exposure to these conditions, rather than a few successful picks. Results should be separated by configuration. An improved fleet average can conceal a regression on one gripper revision.

Suppose cap slip increases after the update. Further rollout stops and affected robots are removed from the scheduler. Each preserves its load and reaches a validated stopping state where possible. Uncertain samples are held for inspection. Logs retain the complete configuration and the physical evidence around the event. Recovery restores the previously validated compatible configuration where supported, checks calibration, and runs a physical acceptance sequence before returning the robot to service. Firmware that cannot be reversed needs a tested recovery image or maintenance procedure. Software restoration and sample recovery are separate obligations. A spilled sample remains lost after rollback.

\begin{table}[tbp]
  \centering
  \caption{Operational measures for comparing releases under the same
  workload. Scheduled time includes stops and recovery. Aborted attempts
  remain in the task record. Report trial counts and uncertainty with
  success rates, and distributions for intervention and recovery times}
  \label{tab:c9cao:fleetmetrics}
  \small
  \begin{tabularx}{\linewidth}{@{}>{\raggedright\arraybackslash}p{3.5cm}>{\raggedright\arraybackslash}X@{}}
    \toprule
    Measure & Definition and interpretation \\
    \midrule
    Task success & Valid completed tasks divided by all attempted tasks,
      with assistance reported separately \\
    Effective throughput & Valid outputs per scheduled operating hour.
      A faster motion is useful only if it increases accepted output \\
    Human intervention & Interventions per scheduled operating hour, together with
      operator minutes consumed and the reason for assistance \\
    Downtime & Time unavailable for assigned work divided by scheduled
      time. Separate planned maintenance from unplanned stops \\
    Recovery time & Elapsed time from fault detection to verified return
      to service. Include waiting for assistance and physical rechecks \\
    \bottomrule
  \end{tabularx}
\end{table}

Table~\ref{tab:c9cao:fleetmetrics} connects these trials to sustained productivity. Faults still unresolved at the reporting cutoff must remain visible, rather than disappearing from recovery-time statistics. Over the same reporting period, the cost per valid output introduced in Section~\ref{sec:c9:embodiment} is
\begin{equation}
 C_{\mathrm{valid}}=
 \frac{C_{\mathrm{equipment}}+C_{\mathrm{operation}}+
       C_{\mathrm{assistance}}+C_{\mathrm{maintenance}}+
       C_{\mathrm{failed\ material}}}{N_{\mathrm{valid}}},
 \qquad N_{\mathrm{valid}}>0,
 \label{eq:c9cao:costvalid}
\end{equation}
where costs are allocated once to that period. With no valid outputs, there is no finite observed unit cost. Better task success can therefore coexist with worse deployment performance if recovery consumes more time or assistance. The release decision should follow the measured physical work, including its interruptions. Those records then identify whether the next improvement belongs in the policy, the sensing and actuation hardware, or the operating procedure.

The cases in Section~\ref{sec:c9:dexterous-cases} examine this requirement at the contact level. A demonstrated manipulation skill becomes useful within a larger task only when its entry conditions, physical result, and recovery requirements are preserved.

\begingroup

\subsection{Case Studies in Dexterous Hand Hardware}
\label{sec:c9:dexterous-cases}

A hand holding a spray bottle must support the bottle while another finger moves the trigger and the wrist maintains the spray direction. Rotating an object within the palm instead requires changing the contacts that keep it under control. In both cases, what a learned policy can achieve depends on which contacts the hardware can make, sense, and adjust.

Applying Sections~\ref{sec:c9:embodiment} and~\ref{sec:c9:sensing-actuation} at the finger level, the following cases examine RobotEra XHAND~1, the first-generation Wuji Hand, and Sharpa Wave W01 through their task requirements, mechanisms, sensing and control interfaces, and demonstrated capabilities. The emphasis is on the hardware conditions under which a result holds. Learning methods are covered in Section~\ref{sec:c6:dexterous}, and low-level hand controllers in Sections~\ref{subsec:ch07:dexterity} and~\ref{subsec:ch07:contact-balance}.

\subsubsection{Manipulation Tasks and Contact Requirements}
\label{sec:c9:dex-task}

Dexterous tasks place different demands on the hand. Grasp acquisition establishes contact with an object. Transport preserves its relation to the hand. In-hand manipulation changes that relation. Tool operation additionally changes another object or process. Their evaluation boundaries matter. A policy tested with an object already in its palm has not thereby been tested on acquiring it from a table, and reaching a tool pose does not necessarily establish that the tool performed its intended operation.

To move the index finger around a cylinder, the remaining fingers must first sustain the load. Excessive squeezing can prevent rotation, while premature release can lose support. Rolling, sliding, and breaking contact permit different motions. The controller must coordinate the transition between them, as well as the finger positions~\citep{lynch2017modern}. In tool use, it must also resist the reaction to pressing, cutting, or twisting. A grasp that supports the tool's weight may be inadequate once the tool meets its target.

Each of these demands is partly a hardware question. Holding a load while another finger moves takes enough independent joints and enough sustained fingertip force. Regulating squeeze instead of position takes a drive that accepts force or impedance commands, or at least torque limits. Catching the onset of sliding takes a contact sensor fast enough to act before the grasp is lost. Resisting a tool's reaction takes force capacity, and an arm that can carry the hand's own mass.

\subsubsection{Hand Design and the Observation-Action Interface}
\label{sec:c9:dex-interface}

\begin{figure}[!t]
  \centering
  \includegraphics[width=\linewidth]{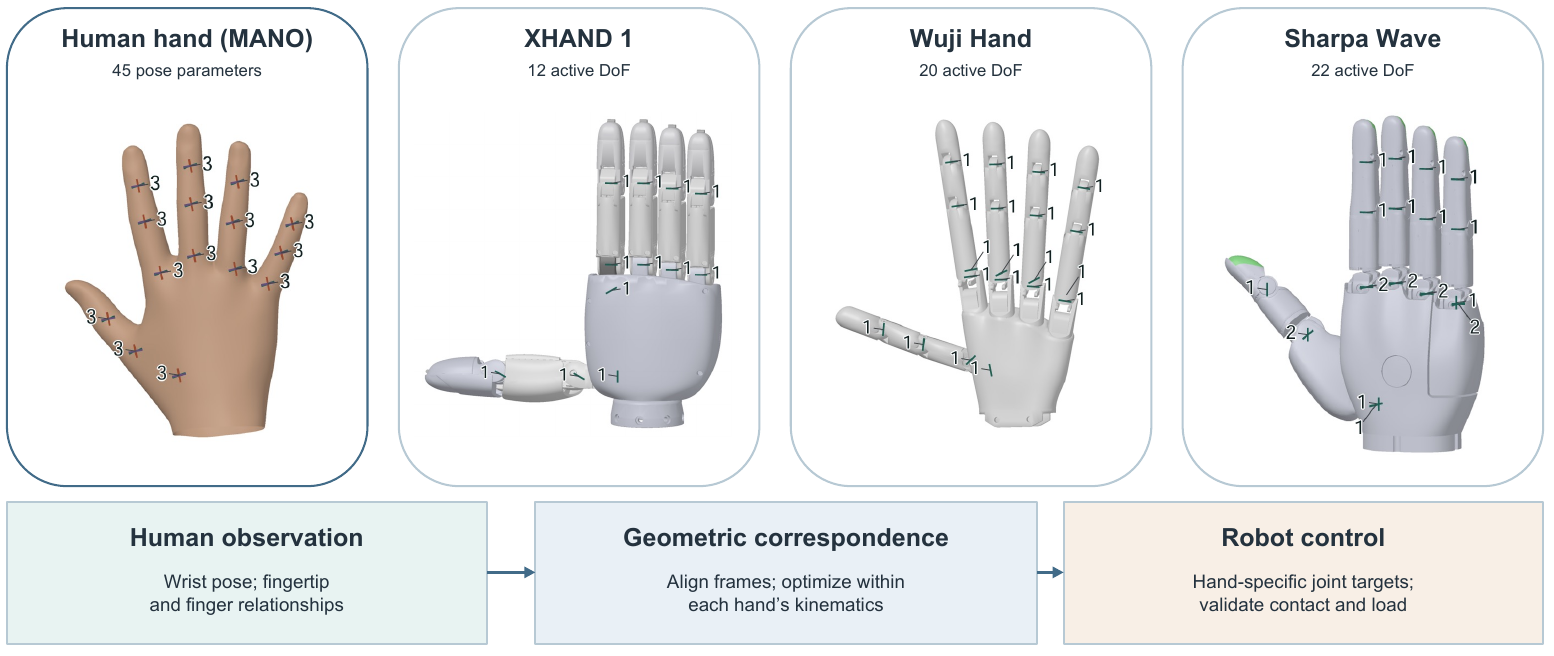}
  \caption{Hand geometry and local rotational coordinates. The surface-smoothed MANO
  template and imported XHAND, Wuji, and Sharpa meshes are shown at a
  common orthographic scale after hand-frame alignment
 ~\citep{romero2017mano,spider2026xhandassets,wuji2026urdf,sharpa2026urdf}.
  Axis marks locate model joints. Adjacent numerals give the local
  rotational dimension, with coincident robot axes grouped.
  MANO uses three pose parameters at each finger joint. Robot labels
  count active Unified Robot Description Format (URDF) coordinates. Wrist and arm motion are excluded.
  Correspondence therefore requires retargeting through the robot's
  kinematics, rather than copying the human pose vector}
  \label{fig:c9dex:mapping}
\end{figure}

The hands differ in their independently controlled coordinates (Table~\ref{tab:c9dex:hardware}), excluding the arm and wrist \citep{robotera2026xhandmanual,wuji2026v1manual,sharpa2026manual}. Thumb opposition, finger spread, joint limits, and available contact surfaces determine what those coordinates can accomplish. More coordinates enlarge both the possible contact arrangements and the coordination problem. They do not alone rank task performance.

Figure~\ref{fig:c9dex:mapping} compares a MANO human-hand mesh with the actual robot geometries and locates their rotational coordinates. MANO represents each of its 15 finger joints by a three-parameter rotation. Its 45-dimensional hand pose is a model parameterization, not a count of independently actuated anatomical motions~\citep{romero2017mano}. Reconstructed human landmarks and robot joints are different representations. A tracked fingertip need not correspond to one motor, and equal joint-vector dimensions need not imply equal fingertip motion. For a thumb-index pinch, the useful correspondence concerns fingertip separation and opposition relative to the object. Reproducing them may require different combinations of finger flexion, lateral motion, and wrist placement. Moving the wrist can compensate for limited finger reach, but also moves the other contacts.

The retargeting objective in Eq.~\eqref{eq:c6dex:retarget} must be constrained by collisions and contact. A small fingertip error does not guarantee that a tool can be held under load.

The observation interface requires the same care. A fingertip contact flag can detect contact without resolving shear load or object pose, while an external camera can locate an object without seeing its supporting contacts. Coverage also matters. A dense fingertip sensor does not observe a load borne by the palm. \emph{Rotating without Seeing} illustrates the alternative of distributing binary contact sensing over the fingers and palm, supporting learned reorientation on real objects~\citep{yin2023touch}. Sensors should therefore be compared by the contact decisions they support, not by resolution alone.

Policy targets also differ from actuator response. A finger stopped by an object cannot track every requested position. Its loading depends on gains, transmission, and effort limits. A study therefore has to report both the policy's targets and the observations it actually used. A tactile sensor listed in a hand's specification may be absent from a particular policy input. Hardware revisions and additional software development kit (SDK) capabilities should likewise remain separate from the experimental configuration being discussed.

\subsubsection{Actuation, Sensing, and Electrical Interfaces}
\label{sec:c9:dex-hardware}

Table~\ref{tab:c9dex:hardware} lists manufacturer specifications for the cited versions, rather than measurements from the learning studies. Their implications depend on the load, contact geometry, and feedback required by the task.

\begin{table}[!tb]
  \centering
  \caption{Manufacturer specifications of the three dexterous hands used in
  the case studies of Sections~\ref{sec:c6:dexterous},
 ~\ref{subsec:ch07:dexterity}, and~\ref{sec:c9:dexterous-cases}
 ~\citep{robotera2026xhandmanual,robotera2024xhandprotocol,wuji2026v1manual,sharpa2026manual}.
  Values are as stated by each manufacturer for the cited version and were
  not independently verified. A dash means that the cited document does not
  state the value. Ratings are not directly comparable, because test
  conditions are not specified uniformly}
  \label{tab:c9dex:hardware}
  \footnotesize
  \renewcommand{\arraystretch}{1.15}
  \begin{tabularx}{\linewidth}{@{}>{\raggedright\arraybackslash}p{0.19\linewidth}
      >{\raggedright\arraybackslash}X
      >{\raggedright\arraybackslash}X
      >{\raggedright\arraybackslash}X@{}}
    \toprule
    & \multicolumn{1}{c}{XHAND~1} & \multicolumn{1}{c}{Wuji Hand (first gen.)} & \multicolumn{1}{c}{Sharpa Wave W01} \\
    \midrule
    Active DoF & 12 (thumb 3, index 3, other fingers 2 each)
      & 20 (four per finger)
      & 22 (thumb 5, fingers 4 each, plus little-finger carpometacarpal (CMC) joint) \\
    Drive (as described) & Geared motor at each driven joint
      & Self-locking rotary direct-drive joints, field-oriented control (FOC) drive
      & ---; zero-force mode is backdrivable \\
    Joint control & Joint targets with host-set gains and torque limits
      & Position control
      & Position, admittance, impedance (MIT), zero-force \\
    Mass and size (mm) & 1.1\,kg, $190\times94\times47$
      & 0.58\,kg excl.\ cables, $201\times75\times50$
      & 1.3\,kg, $208\times90\times50$ \\
    Max.\ fingertip force & 15\,N (whole hand 80\,N)
      & 15\,N
      & 20\,N (active) \\
    Grasp-diameter limit & Minimum 16\,mm
      & Maximum 100\,mm
      & Minimum 10\,mm \\
    Joint load sensing & Motor current
      & ---
      & Current-based estimate\newline 5 torque sensors at thumb CMC and finger metacarpophalangeal joints \\
    Tactile sensing & Optional fingertip arrays, $>100$ points per finger, 3D force and temperature
      & None on the hand
      & Vision-based fingertip sensors (5), with 6-axis force, deformation map, and contact point\newline 20\,mN, 1\,mm resolution \\
    Tactile rate and latency & ---
      & ---
      & 30\,Hz, 50\,ms (processed on board) \newline up to 180\,Hz, 20\,ms (raw images, host GPU) \\
    Interface and joint rate & RS485 or EtherCAT (Ethernet for Control Automation Technology), $\approx$12\,ms whole-hand update
      & Universal Serial Bus (USB)~2.0, 1\,kHz $\times$ 20 axes
      & Ethernet (100/1000\,Mb/s), 500\,Hz \\
    Supply and power & ---
      & 12 to 20\,V DC, 12\,W static
      & 18 to 28\,V DC\newline 15\,W static, 100\,W average, 180\,W peak \\
    Rated life and protection & ---
      & $\geq 3\times10^5$ unloaded grips; IP40
      & $>10^6$ unloaded cycles, tactile skin $>10^5$ cycles at 40\,N; IP20 \\
    \bottomrule
  \end{tabularx}
\end{table}

\paragraph{Mass, Size, and Force Capacity}
All three hands approach human hand size, but their masses differ by more than a factor of two~\citep{wuji2026v1manual,sharpa2026manual}. Hand mass consumes arm payload before an object is lifted and adds to wrist inertia. It must therefore enter the arm's dynamics model. Maximum fingertip-force ratings do not specify how long a force can be sustained. For a stiff trigger, the relevant test is continuous force within the motor's thermal limits, evaluated in the geometry of the actual grasp.

\paragraph{Drive and Compliance}
The three drives give a policy's joint target different physical meanings. Wuji's rotary direct-drive joints operate under position control and are specified as self-locking~\citep{wuji2026v1manual}. Active yielding to an external load requires the controller to adjust the position target from feedback. Sharpa Wave instead exposes position, admittance, impedance, and zero-force modes, the last being backdrivable for manual guidance~\citep{sharpa2026manual}. Under impedance control, a blocked finger produces a force that grows with its position error and the stiffness selected, as in the series-compliance relation of Eq.~\eqref{eq:c9cao:series}. Under zero-force mode an operator can move the fingers directly, which is one route to kinesthetic demonstration. XHAND accepts joint targets together with controller gains and torque limits from the host (Section~\ref{sec:ch9-xhand-case}). Joint targets alone therefore underspecify an action. The control mode, gains, and effort limits used during collection must be logged and reproduced at deployment.

\paragraph{Sensing Load at the Joints}
Sharpa Wave estimates joint torque from motor current by default and also provides five torque sensors at the thumb carpometacarpal and four metacarpophalangeal joints. Their full-scale ratings are 5 and 3\,N\,m, respectively, with a stated total error of $\pm$(1\% of full scale + 5\% of reading) after calibration~\citep{sharpa2026manual}. XHAND also returns motor-current information. Current-based estimates require calibration for transmission friction and motor temperature. A dedicated torque sensor measures the load through its joint but still does not locate the contact. Section~\ref{subsec:ch07:contact-balance} examines control using such joint-torque force estimates.

\paragraph{Tactile Sensing: Coverage, Rate, and Latency}
Each Sharpa Wave fingertip uses an internal camera to image a compliant surface. It reports force, surface deformation, and contact location, with a 30\,N force measurement range~\citep{sharpa2026manual}. For control, their timing is critical. In the standard mode these outputs are computed on board at 30\,Hz with a reported nominal latency of about 50\,ms. In the high-performance mode raw images stream at up to 180\,Hz with a reported nominal latency of about 20\,ms and are processed on a host GPU. The joint loop, by comparison, runs at 500\,Hz. The reported nominal tactile latencies of 20 to 50\,ms correspond to 10 to 25 joint-control periods at 500\,Hz. Tactile updates at 30 to 180\,Hz arrive approximately 3 to 17 times less frequently than joint-control updates. Systems such as T-Rex therefore refine finger targets asynchronously from tactile history while a separate servo thread tracks them (Section~\ref{subsec:ch07:contact-balance}). Sensing latency remains part of the end-to-end budget in Eq.~\eqref{eq:c9yuan:e2e}. XHAND's optional arrays report fingertip force and temperature~\citep{robotera2026xhandmanual}. The first-generation Wuji Hand has no tactile sensing. Pressure measured by its companion glove belongs to the operator's contacts, not the robot's. In all three cases the palm and the inner finger segments are uninstrumented, so loads that the policy routes there remain unobserved.

\paragraph{Electrical Interfaces, Bandwidth, and Power}
The joint interfaces in Table~\ref{tab:c9dex:hardware} impose different update periods~\citep{wuji2026v1manual,robotera2024xhandprotocol}. Sensor transport can impose a separate constraint. Sharpa Wave's hand control and 30\,Hz tactile outputs fit within 100\,Mb/s Ethernet, whereas its 180\,Hz raw tactile stream requires gigabit Ethernet~\citep{sharpa2026manual}. Choosing raw images thus moves computation to the host and raises network load, linking sensing to the device-edge placement decisions of Section~\ref{sec:c9:device-edge-cloud}. Power follows the same logic. Sharpa Wave specifies 100\,W average and 180\,W peak per hand, and 200\,W average for a pair. Wuji's supplied adapter is rated for 420\,W 10\,ms transients~\citep{sharpa2026manual,wuji2026v1manual}. On a mobile manipulator, two hands draw from the same battery as the base and arms, and peak currents during simultaneous grasps must be within the supply's transient capacity.

\paragraph{Durability and Maintenance}
The rated hand lifetimes concern unloaded motion, whereas the tactile-skin rating specifies loaded cycles~\citep{sharpa2026manual,wuji2026v1manual}. These conditions matter in contact-rich learning. Fingertip skins become consumables, and pressure-induced tactile drift calls for recalibration within the maintenance loop of Section~\ref{sec:c9:reliability}, separate from model updates. Neither stated ingress rating specifies liquid-ingress protection. Covers or gloves needed for spills in kitchen or laboratory tasks also alter friction and tactile response.

The specifications alone already limit what can transfer between hands, whatever the learning method. A policy trained on Sharpa fingertip deformation images cannot run on a hand without tactile sensing. A controller that relies on impedance mode cannot be transferred unchanged to a position-only interface. A refinement loop designed for a 500\,Hz joint interface must be retimed for a 12\,ms bus update. Section~\ref{sec:c6:dex-cross} asks the complementary question of what a shared action representation can transfer.

\subsubsection{Demonstration and Data-Collection Hardware}
\label{sec:c9:dex-data}

For dexterous hands, data collection must capture wrist motion, articulated finger motion, and contact together. The same tracker and glove can record a person directly manipulating an object or provide commands to a robot. The resulting records describe different bodies and interactions.

Human pretraining data include egocentric video with estimated hand motion and instrumented demonstrations with measured tracking signals. CAIP uses both pose-estimated video and laboratory human recordings with MANUS Metagloves Pro and Vive Ultimate Trackers~\citep{sharma2026caip}. Instrumented capture reduces dependence on reconstructing occluded fingers, but still requires the hand model and the retargeting of Section~\ref{sec:c6:dex-actions}. Robot teleoperation additionally records how the deployed arm and hand respond to the mapped commands. This response is the extra evidence needed when a geometrically plausible demonstration does not produce the intended grasp.

A representative teleoperation arrangement pairs a wrist tracker with a data glove. The tracker estimates the hand's position and orientation in the tracking space. The glove resolves the fingers relative to the hand. MANUS Core supports external tracker poses, including HTC Vive devices, with calibrated offsets between tracker and glove~\citep{manus2026tracking}. In CAIP's Sharpa setup, Vive wrist motion drives relative arm targets, while MANUS finger keypoints are retargeted into hand-joint targets~\citep{sharma2026caip}. The two measurements solve separate parts of the arm-hand coordination problem. Finger tracking alone cannot tell the arm where to place the hand in the workspace.

Wuji's official teleoperation interface likewise supports Vive tracking with Wuji Glove or MANUS as the hand input~\citep{wuji2026teleop}. Wuji Glove combines electromagnetic fingertip tracking, a fitted hand model, and pressure sensing on the palmar side~\citep{wuji2026glove}. MANUS also distinguishes raw sensor measurements from its solved hand skeleton and retargeted output~\citep{manus2026datalayers}. For supervision, this means that a model-derived joint angle is not an independently measured robot joint angle. Likewise, pressure recorded while the wearer holds an object describes human-side contact. During free-hand teleoperation, the glove does not acquire the remote robot's contact forces. Those require robot-side sensing. Measuring pressure and rendering force feedback to the operator are separate capabilities.

Calibration has two distinct roles. Tracker-to-hand and tracking-to-robot transforms establish wrist targets, while hand-shape calibration and retargeting establish finger targets. A misplaced wrist tracker shifts every fingertip even when finger articulation is estimated accurately. Recording issued targets alongside measured robot motion helps distinguish this error from failure to reach a target under load.

Exoskeleton interfaces move part of this correspondence into the mechanism. Linkages and encoders relate the operator's motion to the demonstration hand, and mechanically matched designs can make collected contacts closer to those available during deployment. DexEXO illustrates this approach with a wearable linkage driving a passive hand whose appearance and kinematics match the deployed hand. Aligned wrist-camera images can then be used directly for policy training~\citep{zhu2026dexexo}. The benefit must be weighed against fit, range of motion, and access around a tool. A bulky finger attachment can prevent a valid demonstration even when the robot itself fits the tool. Passive coupling also differs from an actuated interface that returns robot forces to the wearer. An exoskeleton does not, by itself, imply bidirectional force feedback.

\paragraph{XHAND: Shaping the Collection Interface to the Mechanism}
\label{sec:c9:dex-xhand}
XHAND approaches the correspondence problem from the collection side. A lower-dimensional hand changes which demonstrated finger motions can be reproduced independently, as Figure~\ref{fig:c9dex:mapping} shows. DexUMI investigates this problem on XHAND and Inspire hands. It combines a wearable exoskeleton fitted to the target hand's kinematics, measured joint motion, matched tactile sensing, and visual replacement of the human hand and device with the robot hand~\citep{xu2025dexumi}. Mechanical alignment constrains the motion before it becomes training data. Observation alignment makes the same interaction more recognizable after deployment. Neither step alone ensures that the robot can sustain the demonstrated load. The paper also reports tactile drift after high pressure and an inconsistent motor-to-fingertip mapping~\citep{xu2025dexumi}. Collection and deployment may use nominally identical devices that differ in loading history, so recalibration and loaded tests are part of keeping the demonstration interface valid.

A second XHAND study addresses the difficulty of coordinating the arm and fingers during collection. In the shared-autonomy framework of \citet{cui2025sharedautonomy}, a person guides a UR3e arm through virtual reality (VR) while DexGrasp-VLA uses vision and touch to control the 12-DoF hand. This division reduces the number of decisions the operator must make simultaneously. The person chooses the approach and places the wrist, and the hand controller handles contact. The benefit depends on that controller. If it closes too early, the demonstration records the error unless the operator corrects it. A useful collection system must therefore expose who controlled each interval, how an intervention changed the action, and whether the resulting grasp served the task. Section~\ref{sec:c6:dex-touch} discusses the learning results of both studies, including where tactile input helped.

\begin{figure}[h]
  \centering
  \includegraphics[width=\linewidth]{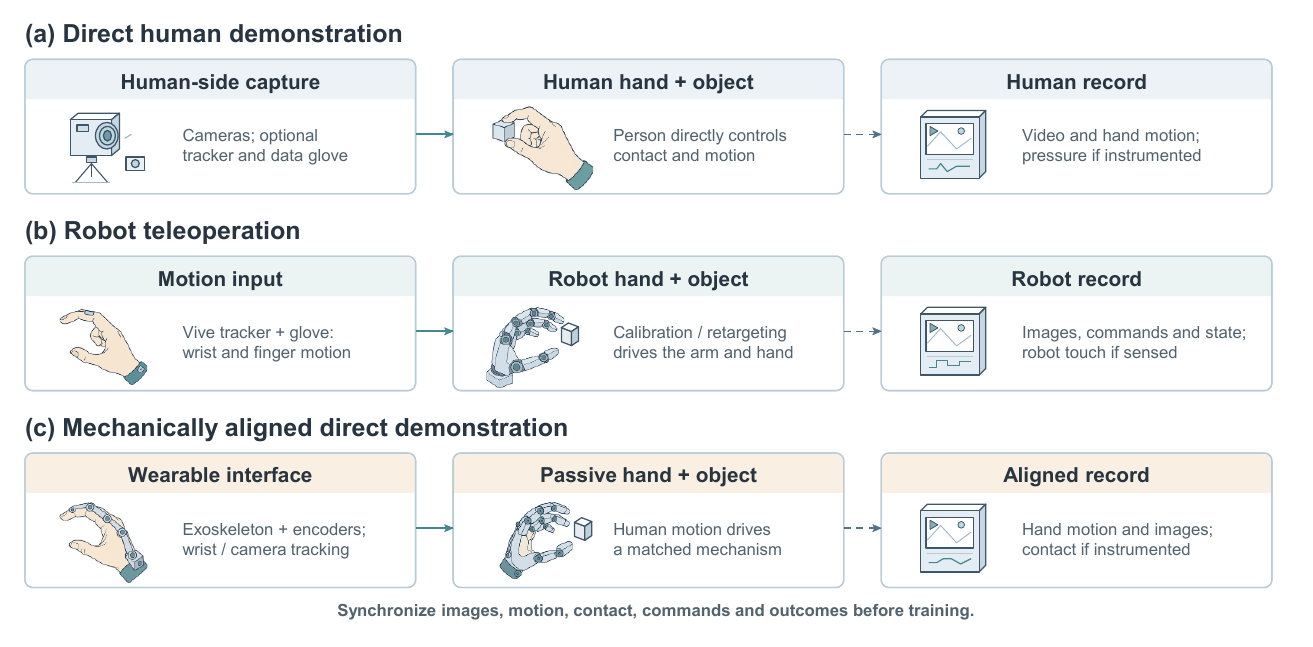}
  \caption{Three routes from demonstration to learning data. A tracker and
  glove can record direct human manipulation or drive a robot through
  retargeting. A mechanically coupled passive hand supplies another form
  of direct demonstration. Solid arrows show motion or control. Dashed
  arrows show recording. The exoskeleton row illustrates passive collection,
  as in DexEXO. An actuated or remotely controlling exoskeleton requires a
  different feedback path. These are collection routes, not a fixed
  assignment of devices to training stages
 ~\citep{manus2026tracking,wuji2026teleop,zhu2026dexexo}}
  \label{fig:c9dex:collection}
\end{figure}

Figure~\ref{fig:c9dex:collection} organizes these routes by where the physical interaction occurs. This determines whether the recorded contact belongs to a person, a passive demonstration hand, or the controlled robot. The distinction remains important even when all three routes use similar cameras and motion trackers.

Contact transitions make timestamp alignment consequential. Pairing finger closure with an image acquired after contact changes the state on which the action appears to depend. Records must distinguish pre-contact intent, measured contact, and subsequent correction.

\subsubsection{From Demonstrated Skills to Integrated Tasks}
\label{sec:c9:dex-integration}

Each learning result built on these hands covers one piece of a task. EgoScale learns a target-robot policy from human data (Section~\ref{sec:c6:dex-human}), WM-Craftnet rotates a held object, SimToolReal follows object-pose goals, and Play2Perfect adapts to assembly (Sections~\ref{subsec:ch07:dexterity} and~\ref{subsec:ch07:contact-balance}). Sharing a hand does not make them composable. The skill interfaces of Section~\ref{subsec:task-organization} require the state produced by one behavior to support the next. An acquired grasp must permit the required finger movement, the object must remain observable, and the transition must preserve its support.

Consider using a tool after picking it up. A grasp that permits lifting may occupy the finger needed to operate the handle. Integration then requires reorientation, a changed grasp, or placement on a support surface before operation can begin. An in-hand rotation result supplies evidence about one of these transitions under its tested initial conditions. It does not establish the complete acquisition-operation sequence. This is where the runtime interfaces of Section~\ref{sec:software-infrastructure} must connect policy outputs to measured contact and task state. Finishing a motion command is not by itself permission to begin the next skill.

Feedback also has to reach the right level. Recent touch can correct finger targets while the visual goal stays valid, but losing a usable grasp calls for a new task-level decision, and the runtime must route the event accordingly (Section~\ref{sec:c9:device-edge-cloud}). With tactile latency spanning ten or more joint-control periods in the reported modes (Section~\ref{sec:c9:dex-hardware}), a faster servo cannot make up for a policy that never receives the evidence of slip.

The reported endpoints (rotation completion, progress along a tool trajectory, insertion success, and accumulated success in a kitchen sequence) do not support a controlled comparison between hands (Section~\ref{sec:c6:dex-eval}). Complete-task evidence requires grasping, contact transitions, operation, and recovery to succeed together under a declared protocol, with assistance costs reported alongside success.

Across the cases, reusable motor knowledge becomes useful only through a specific sensing and control interface. What transfers may be human coordination, an object-level goal, or a functional finger action. What has to be checked is the contact behavior on the deployed hand. Transfer to another hand, adaptation to assembly, and continued operation through wear are separate achievements. On a humanoid, this interface also depends on the moving body that supports the hand. Section~\ref{sec:c9:humanoid-cases} extends the analysis to that integrated system.

\par
\endgroup

\subsection{Case Studies in Humanoid System Integration}
\label{sec:c9:humanoid-cases}

Carrying an object across a room connects almost every component considered in this chapter. The hands maintain the grasp, arm motion redistributes the load, stepping changes the support contacts, and the moving cameras must still provide useful observations of the destination. Recent systems built on the Unitree G1 show how these components can be organized for teleoperation, data collection, and autonomous whole-body interaction. They make the co-design problem in Section~\ref{sec:c9:embodiment} concrete. The task model must express the required movements, and the robot's mechanics, sensing, and computation must support their execution in closed-loop interaction. The comparison follows a common set of questions concerning what the task demands from the body, how observations and commands cross the interfaces, what was demonstrated, and which operating conditions bound that evidence. G1 provides a common platform for comparing these integration choices.

\subsubsection{Embodiment and Whole-Body Coordination}
\label{sec:c9:g1-platform}

A humanoid's manipulation workspace depends on how it positions and supports its body. In G1's 29-DoF configuration, six joints per leg, three at the waist, and seven per arm provide multiple ways to reach and orient an object. Hand joints are counted separately~\citep{unitree2026g1docs}. Waist and arm motion can extend a reach, but the resulting posture and load must remain compatible with the supporting contacts. The hand's contact capabilities therefore need to be considered together with the body motions required to use them, extending the hand-design problem in Section~\ref{sec:c9:dexterous-cases} to the complete robot.

This coupling shapes the division of control between a task model and a motor controller. A task model may specify hand and arm motion while leaving the supporting legs to a local feedback policy. When a foot also manipulates an object, the task interface must convey that contact objective as well. HOMIE and SONIC implement these divisions of control on G1~\citep{ben2025homie,luo2026sonic}. Both depend on a motor controller that produces the expected physical response from the assembled robot.

\subsubsection{From a Learned Policy to Physical Execution}
\label{sec:c9:g1-execution}

For a joint-target policy, motion depends on the interaction between its outputs, feedback gains, and actuator mechanics. As the actuation analysis in Section~\ref{sec:c9:sensing-actuation} explains, matching target angles alone does not match the robot's response. ASAP addresses this problem on G1 by learning a correction to the simulated action-response relationship from measured robot trajectories, with a focus on the ankle linkage. The correction is used during policy fine-tuning in simulation. The resulting policy executes on the robot without that additional correction model~\citep{he2025asap}. Physical measurements thus change the dynamics against which the policy is trained, improving transfer without adding that model to the online control loop.

Policy execution also depends on the state estimates and timing of its feedback path. BeyondMimic's G1 motion-tracking stack combines a 500~Hz state estimator with a 50~Hz policy, whose onboard CPU inference takes less than 1~ms~\citep{liao2026beyondmimic}. The estimator updates the body-state estimate between policy evaluations, while the policy uses that estimate to update joint targets. Policy frequency, inference latency, and estimation frequency therefore describe different parts of the physical loop. This separation implements the multirate execution discussed in Section~\ref{sec:software-infrastructure}. Timely movement depends on the complete sensing-computation-actuation path, including the stages outside neural-network inference.

\subsubsection{Connecting Body Control, Hands, and Task Models}
\label{sec:c9:g1-manipulation}

\begin{figure}[!ht]
  \centering
  \includegraphics[width=\linewidth]{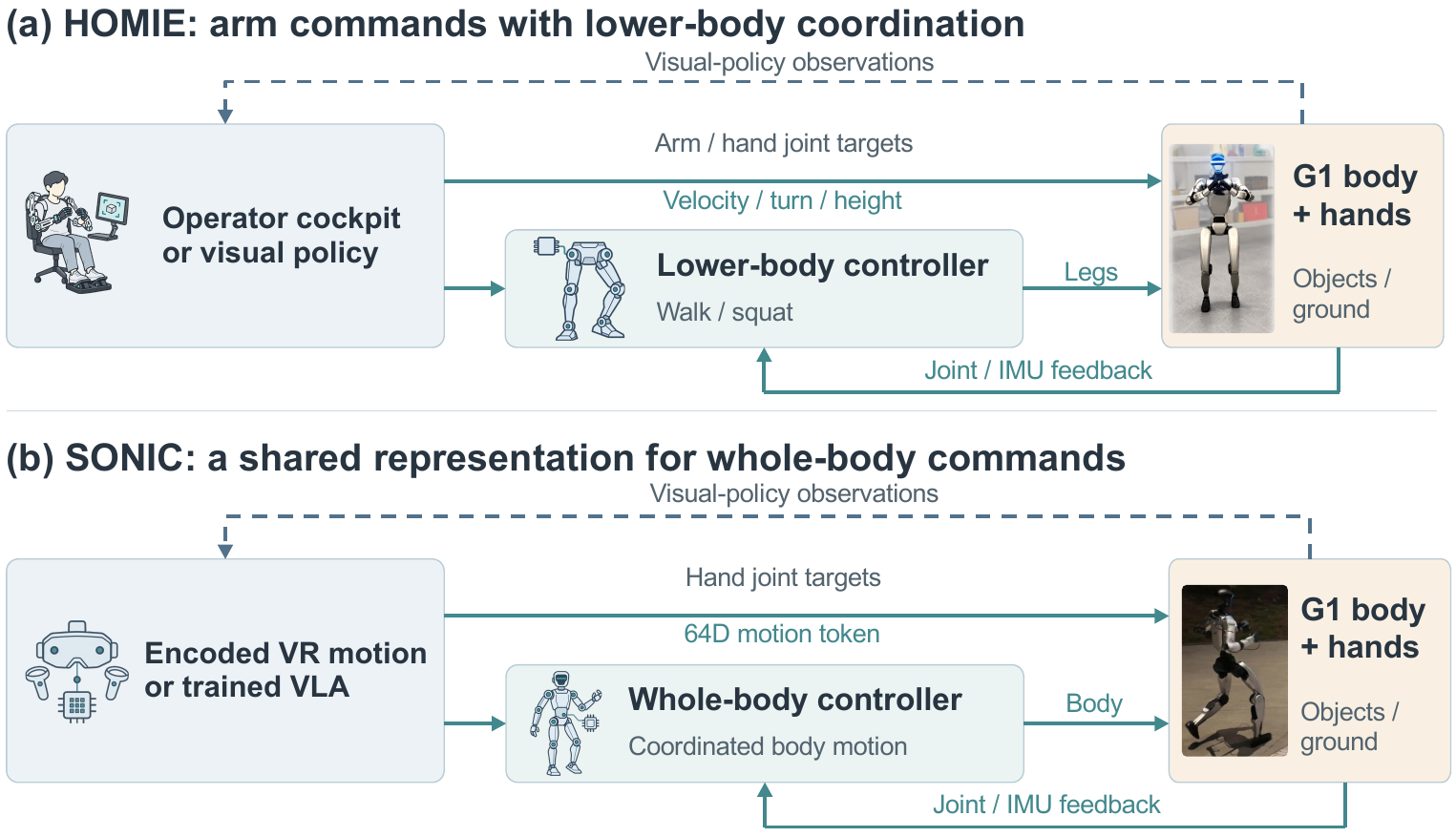}
  \caption{Two G1 integration routes. HOMIE's teleoperation interface combines arm and hand targets
  with velocity, turning, and torso-height commands to a learned
  lower-body controller. Its autonomous visual policy predicts arm, hand,
  and height targets. In SONIC's whole-body VLA route, a shared motion
  token conditions the body controller, while separate joint targets
  command the hands. In both systems, local proprioceptive feedback
  supports body control, and visual observations inform task commands.
  VR motions in SONIC are encoded into the shared motion representation
 ~\citep{ben2025homie,luo2026sonic}}
  \label{fig:c9g1:integration}
\end{figure}

In whole-body manipulation, the division of control determines which aspects of motion a person or task model specifies and which the local controller supplies. HOMIE and SONIC provide two concrete answers (Figure~\ref{fig:c9g1:integration}). Both connect a task-command source to body and hand execution, but they expose different levels of control over the supporting legs.

\paragraph{HOMIE: A Shared Interface for Demonstration and Autonomous Execution}
HOMIE connects an isomorphic arm exoskeleton, motion-sensing gloves, and foot pedals to a G1 equipped with Dex3-1 hands~\citep{ben2025homie}. The exoskeleton supplies arm-joint targets. The gloves command the hands. The pedals specify forward or backward velocity, turning rate, and torso height. A learned lower-body controller uses robot feedback to walk or squat as the upper-body posture changes. This allocation reduces the operator's burden. The person directs the reach and grasp while the lower-body policy supplies the compensating motion discussed in Section~\ref{sec:motion-control}. The operator receives no force feedback, however, limiting direct assessment of contact loads through the teleoperation interface.

The same allocation connects data collection to autonomous execution. HOMIE records images, joint states, and task commands during teleoperation, then trains a visual policy to produce arm, hand, and height targets. Its local locomotion policy runs at 50~Hz on the robot's Jetson Orin. For autonomous manipulation, head- and arm-mounted cameras and joint feedback supply a visual policy on an external RTX~4080 computer through wired Ethernet. This upper-level loop runs at 10~Hz. The task model can therefore replace the operator while retaining the controller that regulates body motion. After collecting 50 demonstrations per task, the system succeeds in 11 of 15 trials of squatting to pick a tomato from a low sofa and 12 of 15 trials of tabletop pick-and-place~\citep{ben2025homie}.

\paragraph{SONIC: Exposing Coordinated Body Motion to a VLA}
Some tasks require the task model to choose a foot contact as deliberately as a grasp. Opening a pedal-operated bin while holding an object, for example, requires placing one foot on the pedal while the other supports the body. SONIC connects a fine-tuned GR00T~N1.5 VLA model to a reusable whole-body controller through a learned motion representation~\citep{luo2026sonic}. In its whole-body mode, the VLA predicts a 64-dimensional motion token and 14 hand-joint targets. The token conditions coordinated body motion. The hand targets command finger configurations. Teleoperated demonstrations populate this same action representation, connecting the motor pretraining discussed in Section~\ref{sec:whole-body-control} to task learning.

The G1 motion stack controls 29 body joints and runs onboard a Jetson Orin. Its learned controller updates at 50~Hz, while an independent 500~Hz command publisher streams the latest joint targets to the low-level interface~\citep{luo2026sonic}. Separating these operations sustains command delivery between policy evaluations. At the task level, the VLA-driven system completes the sequence of picking up a can, walking to a bin, pressing its pedal, and disposing of the can in 6 of 10 trials after training on 1,000 multi-object demonstrations. It also places a drill in a box and carries the box to a shelf with both hands in 7 of 10 trials after 300 demonstrations. These are complete-task results. Grasping, locomotion, and changes in support must succeed within the same execution.

The two interfaces support different task demands. HOMIE assigns the operator or visual policy direct control of the arms and hands and leaves leg coordination to a command-conditioned controller. SONIC's motion representation also lets the task model request coordinated leg motions, including a foot's interaction with an object. In both cases, reusing a motor controller becomes useful through an interface that expresses the movements needed to establish and maintain task contacts.

\subsubsection{Physical and Perceptual Conditions for Task Execution}
\label{sec:c9:g1-conditions}

A task interface connects learned models to coordinated movement, but the range of executable tasks also depends on the robot's end effectors, the states it can observe, and the demands of repeated operation. AdaPT's G1 tennis experiments expose the coupling between the first two. Serving requires both a body capable of the swing and timely observations of the ball~\citep{huang2026towards}.

For professional-style serving, AdaPT combines a task-specific racket mount with a lightweight parallel gripper that holds and tosses the ball. The heavier Unitree hand, termed Dex-3 in the AdaPT study, is evaluated with simpler amateur motions because the tested configuration cannot maintain balance with that hand during the professional serving motions~\citep{huang2026towards}. The choice of hand thus changes which whole-body movements are feasible, even though its immediate role is to hold and release the ball. End-effector selection must account for the dynamic task performed by the complete body as well as the local grasp.

Executing the swing also requires observations that arrive early enough to coordinate racket-ball contact. AdaPT's principal experiments use a 35-camera motion-capture system to track the robot and ball. In a separate field deployment on the Dobot Atom humanoid, stereo vision and HTC VIVE Ultimate Trackers provide ball observations and robot localization, respectively~\citep{huang2026towards}. The motor skill is therefore evaluated together with a particular source of task-state information. The same dependence appears in locomotion. Gallant's confined-terrain case in Section~\ref{sec:c9cao:g1case} requires three-dimensional geometry over the space occupied by the body. In both settings, sensing must resolve the geometry relevant to making or avoiding contact. Intercepting a moving ball additionally requires predicting where contact occurs.

Sustaining these motions over repeated trials adds thermal and mechanical constraints to those of a single execution. ASAP reports motor overheating and partial damage to two G1 robots during its high-dynamic-motion data collection, limiting the real-world data available for adaptation~\citep{he2025asap}. This connects the actuator limits in Section~\ref{sec:c9:sensing-actuation} to the operating-life considerations in Section~\ref{sec:c9:reliability}. Physical learning depends on how often the assembled robot can repeat the contacts and accelerations needed to collect experience.

\subsubsection{From Integrated Demonstrations to Sustained Operation}
\label{sec:c9:g1-deployment-evidence}

HOMIE and SONIC demonstrate that a shared command interface can connect data collection, task models, and local body control. Their reported task trials test whether grasping, movement, and changes in support can succeed in one execution. The different tasks, demonstration sets, and trial protocols do not provide a controlled comparison between the two systems. Nor do those trial counts describe how either system performs over a working shift. The measures in Section~\ref{sec:c9:reliability} extend the evaluation to valid output, assistance, downtime, and recovery under a declared workload.

Table~\ref{tab:c9g1:evidence} relates the evidence discussed here to those deployment questions. Its final column states further checks for an intended deployment. It does not attribute unreported tests or protective functions to the cited systems. Hardware specifications, measured loop rates, and task success each answer a different part of the system question.

\begin{table}[!htbp]
  \centering
  \caption{From humanoid integration evidence to deployment questions. The middle column summarizes the cases discussed in this section. The final column identifies additional evidence needed for the intended operating conditions, following Sections~\ref{sec:software-infrastructure} and~\ref{sec:c9:reliability}}
  \label{tab:c9g1:evidence}
  \small
  \renewcommand{\arraystretch}{1.15}
  \begin{tabularx}{\linewidth}{@{}>{\raggedright\arraybackslash}p{0.18\linewidth}>{\raggedright\arraybackslash}X>{\raggedright\arraybackslash}X@{}}
    \toprule
    System question & Reported evidence & Further deployment question \\
    \midrule
    Body and task compatibility & AdaPT couples the serving motion to end-effector choice and a specified observation setup. & Do payload, mounting, visibility, and sensing delay remain within the tested envelope at the intended site? \\
    Feedback and execution & ASAP adapts the simulated response from robot measurements, while BeyondMimic separates estimation and policy updates. & Does the complete feedback path remain timely under the intended load, and what happens when an observation or command expires? \\
    Task composition & HOMIE and SONIC report trials connecting hand actions to body motion through different command interfaces. & Which physical evidence permits each transition, and how are failed grasps, interrupted tasks, and operator interventions handled? \\
    Repeated operation & ASAP reports overheating and robot damage during high-dynamic-motion data collection. & What duty cycle, maintenance, and recovery procedures preserve valid output over the scheduled operating period? \\
    \bottomrule
  \end{tabularx}
\end{table}

For a humanoid carrying an object, this distinction becomes concrete when the task computer or its communication link becomes unavailable. A local body controller may still receive proprioceptive feedback, but that alone does not specify whether the last task command remains authorized or how the object is secured. The runtime must define command expiry and the locally executable response for the relevant physical state, as in Section~\ref{sec:software-infrastructure}. Trials then need to establish the response under the faults assigned to that design, including the conditions for resuming the task. These fault-response tests form part of deployment validation. The task trials summarized above do not establish their outcomes.

The hand and humanoid cases bring the chapter's five questions together. The body determines which contacts and loads are feasible. Sensing makes their consequences observable. Computation must finish while that evidence remains useful. Software assigns authority and connects commands to measured outcomes. Deployment preserves these relationships through repetition, wear, and updates. A change in fingertip material, payload, camera placement, or command timing can alter several of these relationships at once. The unit to validate is therefore the assembled observation-action system under its intended workload. Hardware and software become a dependable physical capability when they repeatedly produce the required result and retain a tested means to detect, contain, and recover from deviations.

\clearpage
\part{Future Outlook}
\epigraph{\textit{"The real problem of humanity is the following: we have paleolithic emotions; medieval institutions; and god-like technology."}}{--- Edward O. Wilson}

\clearpage 

\section{Physical AGI}
\label{ch:general-physical-ai}

\begin{figure}[t]
\centering
\includegraphics[width=\textwidth]{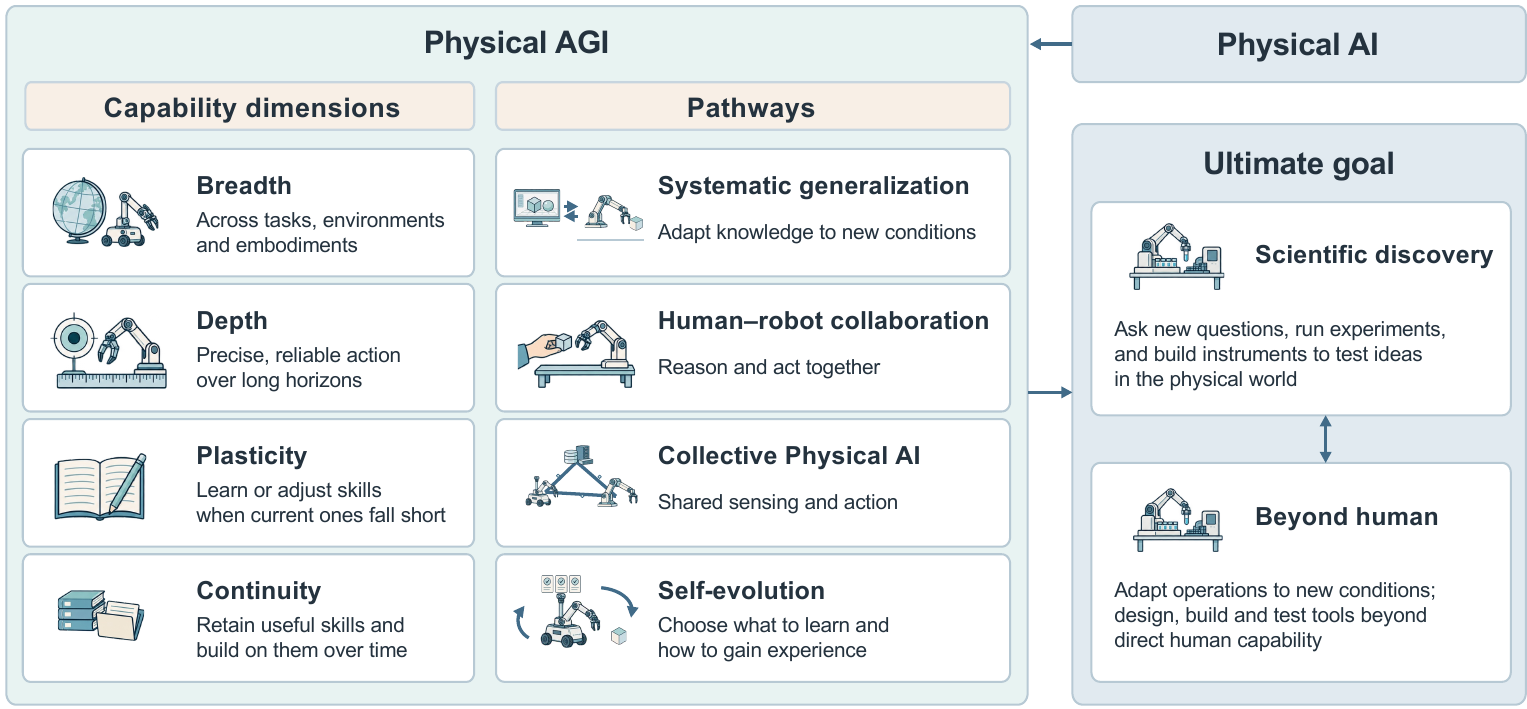}
\caption{Overview of physical artificial general intelligence (Physical AGI). Building on Physical AI, Physical AGI is described by the four capability dimensions of breadth, depth, plasticity, and continuity. It is pursued through four pathways, which are systematic generalization, collaboration between humans and robots, collective Physical AI, and self-evolution. Its ultimate goal is to advance scientific discovery and to enable engineering beyond the reach of human capability}
\label{fig:physical-agi}
\end{figure}

The preceding chapters examine the basic concepts of Physical AI together with perception, data, world models, foundation models for robot manipulation, foundation models for whole-body motor control, agents, and hardware systems. These technologies provide the foundations for systems that perceive, reason, and act in the physical world. The question for this chapter is whether these foundations can support more than a growing repertoire of predefined physical tasks. This monograph focuses on intelligence that can interpret unfamiliar situations, develop new ways of acting, and take part in creating and maintaining the material conditions of human life. It calls the ultimate goal of this line of research \emph{physical artificial general intelligence (Physical AGI)}. This term denotes an artificial system that integrates broad understanding, learning, and reasoning with reliable action in the physical world and can solve unfamiliar real-world problems across many domains at a level comparable to capable humans.

The qualifier \emph{physical} gives this goal its specific meaning. An agent must connect its task-relevant knowledge to what its body can actually do, reason about the consequences of intervention, and revise its understanding when the world responds unexpectedly. Recognizing an object or describing a plausible plan is only the beginning. The agent must also find out how the object is supported, what changes once contact is made, and which uncertainties matter for execution. The physical world is thus more than the place where an intelligent system is tested. It supplies evidence, constrains reasoning, and is the medium through which the system achieves its goals.

Physical grounding, a complete technical stack, a larger task repertoire, and greater model scale do not by themselves establish generality. Chollet defines intelligence as skill-acquisition efficiency relative to priors, experience, and generalization difficulty, and thereby separates it from achieved skill~\citep{chollet2019measure}. High task performance may reflect effective learning, extensive prior training, or both. The relevant contrast is between a fixed repertoire and the ability to meet unfamiliar demands by reusing knowledge and acquiring what is missing. Scale can improve both, and Sutton identifies search and learning as general methods that keep benefiting from increasing computation~\citep{sutton2019bitter}. For Physical AI, what must be shown is that these gains translate into better understanding, transfer, execution, and adaptation.

Physical AGI must therefore act effectively across tasks, environments, and embodiments, and acquire further competence when existing knowledge falls short, without being redesigned for each task. Section~\ref{sec:capabilities} makes this requirement precise. It examines what separates such a system from today's Physical AI and characterizes competence along four dimensions, namely breadth, depth, plasticity, and continuity. As Figure~\ref{fig:physical-agi} shows, the subsections that follow trace four pathways toward Physical AGI, namely systematic generalization, human-robot collaboration, collective Physical AI, and self-evolution, each of which relaxes an assumption of current Physical AI. A concluding subsection considers the proposed purposes of Physical AGI in scientific discovery and in engineering beyond human limits.

\subsection{Core Capabilities of Physical AGI}
\label{sec:capabilities}
Most Physical AI systems today are built and evaluated for a stated set of tasks, settings, and platforms. Within that set they can be highly capable, but their competence is bounded by the scope anticipated by training and engineering, and a new demand typically calls for new data, retraining, or redesign. Physical AGI differs from such systems in generality, the ``G'' in AGI, rather than in higher performance on the same tasks. A Physical AI system is characterized by the skills it has. Physical AGI is characterized by how far it can carry those skills, and extend them, beyond the conditions covered by its preparation.

This chapter takes \emph{generality} in this sense to mean the capacity to reuse knowledge and problem-solving processes across diverse tasks, environments, and embodiments, and to acquire or revise competence efficiently when existing knowledge is insufficient. Generality is always relative to a stated scope, performance standard, and resource budget, not to all conceivable tasks.

What makes this definition physical is the relationship between knowledge and intervention. An agent must reason about objects, materials, spatial structure, forces, and its own body well enough to act on them, and it must link an abstract objective, such as repairing an object or assembling a mechanism, to the physical conditions under which that objective can be met. The link runs both ways, since interaction can expose unknown constraints and change how the agent understands the problem. Physical AGI thus acts with available knowledge and also uses action to acquire knowledge for later tasks.

The definition covers both the use of existing competence and the acquisition of new competence, and many unfamiliar conditions call only for the former. A robot may grasp an object with an unseen surface texture using a policy it already has. A narrower container opening may call for a new grasp or a contact-guided insertion, which planning with existing knowledge may still find. New learning becomes necessary only when knowledge and available skills both fall short. The deciding question is whether existing competence still suffices, not whether the situation is unfamiliar.

For the same reason, a large but fixed repertoire is weak evidence of adaptability, and fast learning within a narrow task family is weak evidence of breadth. Generality also does not mean freedom from human help. Demonstrations, feedback, and manual adjustments needed for a new task count toward its adaptation cost. The criterion is whether the system's own adaptation and learning can meet new demands within its physical, computational, and safety limits, without task-by-task redesign.

To describe both the competence a system has and its capacity to develop further, this chapter uses four dimensions. Breadth and depth describe present capabilities, following the distinction between generality and performance~\citep{morris2024levels}. Plasticity and continuity describe how competence is acquired and accumulated through experience.

\emph{Breadth} is the range of physical conditions over which a system acts effectively, including variation in tasks, objects, scenes, tools, dynamics, and embodiments. Counting task names is a poor proxy for this range. Many differently named tasks rely on the same grasp and motion, while a small change in friction can demand quite different control. Breadth should therefore be described by the physical variation over which competence holds. Open X-Embodiment shows that training across robotic platforms can improve performance by transferring experience between them~\citep{openx2024}, and $\pi_{0.5}$ performs household manipulation in homes absent from its training data~\citep{pi2025pi05}. Both results demonstrate breadth under their evaluation conditions, but neither addresses arbitrary changes in dynamics or morphology. Sharing parameters across robots makes transfer possible, but whether transfer actually occurs is an empirical question.

\emph{Depth} is how well the system performs within that range, including its accuracy, compliance, efficiency, reliability, and coordination over long horizons. In physical tasks these qualities are tightly linked. A grasp must support transport, transport must leave the object in a state from which it can be placed, and a small slip can alter the conditions for every later action. Deep competence therefore includes detecting disturbances, maintaining the conditions that later actions depend on, and recovering when those conditions are lost. Hierarchical planning supports long-horizon coordination by decomposing an objective into subgoals and reusing existing policies for familiar low-level actions~\citep{sutton1999options,kaelbling2011hierarchical}. A system with many fragile skills and one with a few reliable skills are limited in different ways, and describing either requires both dimensions.

\emph{Plasticity} is the capacity to acquire or revise competence when existing behavior is inadequate. A change in payload or friction may force the agent to revise its model, adjust its control, or learn a different interaction. Plasticity is measured by the experience, time, computation, supervision, and risk that adaptation requires. Two systems may reach the same success rate after very different numbers of physical trials, and their plasticity then differs even though their final performance does not. Plasticity is also distinct from retention. Dohare and colleagues show that standard deep-learning methods can gradually lose the ability to learn in continual-learning settings~\citep{dohare2024plasticity}. Forgetting concerns what a system can no longer do, whereas loss of plasticity concerns what it can no longer learn.

This chapter treats plasticity as a property of the whole embodied system, not only of its parameter updates. Memory, model-based replanning, and newly generated procedures can all support adaptation~\citep{hafner2019planet,wang2024voyager}. Replanning with unchanged knowledge may produce an unfamiliar solution without any new competence being acquired, while new knowledge can be stored in memory or executable procedures without changing network weights. This chapter therefore judges plasticity by whether new experience leads to effective behavior, whatever mechanism carries the change.

Learning accumulates only if the results of adaptation outlast the episode and can be recalled when needed. This chapter calls this accumulation of useful competence across encounters \emph{continuity}. It involves keeping skills that remain valid, revising assumptions that no longer hold, and preserving the ability to keep learning. Elastic Weight Consolidation, for instance, restricts changes to parameters that matter for previously learned tasks~\citep{kirkpatrick2017ewc}. This can reduce forgetting, but it leaves open when a physical failure calls for learning, which part of the system should change, and when a retained skill should be used. These are system-level decisions, and this chapter measures continuity by the lasting benefit of experience rather than by how long parameters persist.

None of the four dimensions is sufficient alone. Zero-shot success can show breadth without plasticity, and rapid adaptation can occur without continuity. Assessing them jointly is therefore a longitudinal matter. It asks whether experience lastingly extends the conditions under which the system acts effectively, without demanding new learning in every successful encounter.

To make this criterion operational, this chapter introduces the \emph{capability boundary}, the limit of the conditions under which a system meets a specified performance standard, with computational resources, interaction budgets, human assistance, and safety constraints held fixed. Prior training and accumulated experience should be reported separately from any resources added during evaluation. Success on a task the evaluator has not tried before may reveal competence that was already present, so novelty to the evaluator is not, on its own, evidence of capability expansion.

This chapter distinguishes three outcomes. \emph{Direct transfer} applies existing competence without acquiring task-specific knowledge, although it may involve substantial search or planning. \emph{Episodic adaptation} resolves the current difficulty but does not retain the useful change. \emph{Persistent capability expansion} retains changes that let the system meet the standard under previously unsupported conditions. Claiming it requires evidence of better performance, or of lower interaction, time, or assistance costs, under comparable constraints, together with evidence that valid skills and the ability to keep learning have been preserved. Gains within conditions the system already handled improve performance or efficiency but do not expand the boundary.

Current Physical AI typically assumes that operating conditions resemble those covered by training, that a person specifies the task and the robot merely executes it, that the robot acts alone in a single body, and that people design and run its learning process and fix its physical configuration. Each of these four assumptions confines the capability boundary in a different direction. In the case of collaboration, the boundary in question is that of joint human-robot action rather than of the robot alone. Relaxing each assumption defines a pathway toward Physical AGI, a proposed system independent of all four assumptions. The next four subsections take the pathways in turn, covering systematic generalization (Section~\ref{subsec:systematic-generalization}), human-robot collaboration (Section~\ref{subsec:human-robot-collaboration}), collective Physical AI (Section~\ref{subsec:collective-physical-intelligence}), and self-evolution (Section~\ref{subsec:self-evolution}). The four dimensions and the distinction among direct transfer, episodic adaptation, and persistent capability expansion serve as a common yardstick for all four.

\subsection{Systematic Generalization in Open-World Settings}
\label{subsec:systematic-generalization}

The first pathway relaxes the assumption that operating conditions resemble those covered by training. Here the distinction between transfer and capability expansion shapes how generalization tests should be read. Many out-of-distribution evaluations vary objects, positions, or task combinations within a predefined family while the basic form of the problem stays the same~\citep{koh2021wilds}. Success shows transfer within the family but says little about what happens when the sources of change are themselves unknown.

Under open conditions, an agent may first need to work out which factors have changed and whether the knowledge behind its current policy still applies, and only then decide what to learn. A drawer that does not open may be latched, jammed, or poorly grasped, and each explanation calls for a different response. The lack of motion alone does not distinguish these cases, whereas a constrained probe can provide additional evidence. Evaluations of open-world generalization should therefore state the changes introduced, the feedback available, and the adaptation permitted.

The same issue arises when systematic generalization is framed, as it usually is, through combinations of familiar constituents. Lake and Baroni show that meta-learning can support such generalization in controlled sequence-learning tasks~\citep{lake2023systematic}. Recombination works when the constituent operations keep their meaning and effectiveness. In physical tasks the operations themselves can fail. A change in friction may break a grasp that sits in exactly the right place in the task sequence, and reordering the sequence does not repair it.

This chapter therefore defines \emph{system-level systematic generalization} as the capacity to identify changes, preserve reusable structure, and find effective solutions when the conditions supporting existing policies no longer hold. It extends the compositional notion to perception, internal models, decision processes, and control without replacing it. A local failure may require changing only part of the system while the rest of its knowledge continues to serve the task. The agent must decide whether and where revision is needed, acquire or repair the missing competence, and test the result.

The structure retained for this purpose need not reproduce every observable detail. World models can predict in an abstract representation space without reconstructing future observations in full~\citep{garrido2025intuitive,assran2025vjepa2}, and temporal abstractions support prediction over extended actions~\citep{sutton1999options}. Such a representation must preserve the information needed to evaluate possible outcomes across tasks and changing conditions. Learned largely from observation, it can supply reusable physical knowledge before a new task arrives, so that not all useful structure has to be acquired through reward-driven trial and error~\citep{hafner2019planet}. Abstraction of this kind is one plausible route to generalization, but it is not part of the definition.

Representations alone do not explain how a system responds when its current solution fails. This chapter distinguishes five functions in that response, namely \emph{competence assessment, change characterization, composition and acquisition, interactive validation, and knowledge consolidation}. Together they lead from recognizing a difficulty to building, physically testing, and retaining a revised solution. They are roles rather than software modules or a fixed sequence, and they may share representations and recur as new evidence arrives. Not all five are always needed, since existing competence may suffice.

\paragraph{Competence Assessment}
The agent judges whether continued execution is likely to remain effective and safe. Unexpected contact forces, mismatches between predicted and observed motion, or uncertain outcomes can warrant further observation, constrained probing, targeted assistance, or a safe stop. Prediction error rises, for instance, when a video shows an event that violates learned physical expectations~\citep{garrido2025intuitive}. Such a discrepancy is a reason to reassess but offers no safety guarantee, and visual familiarity does not show that a policy remains reliable, because friction can change with no visible sign. An assessment is useful only if it changes behavior early enough, which can happen before the cause of the difficulty is known.

\paragraph{Change Characterization}
Detecting a difficulty does not identify its cause. The agent must determine which assumptions need revision and which knowledge remains usable, since holding on to invalid assumptions invites repeated failure and discarding valid knowledge makes adaptation more expensive. The principle of Independent Causal Mechanisms supports selective revision, because under suitable assumptions a change in one mechanism need not require revising the others~\citep{scholkopf2021causal}. Whether a learned representation has captured the relevant factors is a separate question. The agent may need informative actions to tell competing explanations apart, and it can act on a partial causal account as long as the remaining uncertainty constrains what it does. Learning can then concentrate on the affected parts of the system.

\paragraph{Composition and Acquisition}
The agent builds a candidate solution from the knowledge that remains valid and identifies any competence still missing. A world model, drawing on memory and current perception, can predict the outcomes of candidate action sequences and score them against a task objective. Searching over these sequences at inference time can yield an unfamiliar solution without changing the model or learning a new skill~\citep{hafner2019planet,assran2025vjepa2}. A new objective may still need a learned readout from the representation, as reward prediction does in latent model-based control~\citep{hafner2019planet}, and such training counts toward the adaptation cost. SayCan combines language-model proposals with value functions that estimate whether learned robotic skills can be executed~\citep{ichter2023saycan}, so its choices are limited to the skills it already has. When a required interaction is missing or no longer works, the task structure can isolate it as a local learning problem, and the agent can acquire or revise that interaction without relearning the whole task. The new competence then becomes available for other combinations, although any solution built from it still requires physical validation.

\paragraph{Interactive Validation}
The agent tests the candidate solution under the physical conditions in which it must work. Limited probes, which must themselves respect force, motion, and safety limits, can reveal resistance, contact constraints, or errors in the state estimate. Predicted trajectories can be checked against task goals and safety constraints before execution~\citep{wabersich2021predictive}, but only real feedback shows whether the predictions and constraint estimates hold. \emph{Inner Monologue} uses success detection, scene descriptions, and human feedback to revise robotic plans~\citep{huang2023inner}. Feedback may also show that the diagnosis itself was wrong and send the agent back to assessment or characterization. A successful recovery resolves the immediate difficulty but remains episodic adaptation unless its benefit is consolidated.

\paragraph{Knowledge Consolidation}
The system retains useful experience in a form that can inform later decisions, whether in policies, world models, memory, or executable procedures. Verified constraints and failed strategies are worth retaining alongside successful actions, together with the conditions under which each applies, so that retrieval can judge whether a skill fits the current situation. Consolidation need not take place in model weights. Voyager keeps executable skills in a retrievable library without fine-tuning its underlying language model~\citep{wang2024voyager}, although its Minecraft setting leaves open how the approach would fare on physical robots. In robotics, RECAP improves $\pi^{*}_{0.6}$ over successive rounds of data collection and training that use autonomous experience and human corrections~\citep{physicalintelligence2025b}. Neither system constitutes a fully autonomous lifelong-learning process, and establishing persistent capability expansion for either would require the longitudinal evidence specified in Section~\ref{sec:capabilities}.

This account extends generalization from unfamiliar samples and task combinations to the whole perception, cognition, decision, and action system. The four capability dimensions specify which improvements to assess, and the five functions describe how experience can produce and preserve them. Because actions can obtain the evidence needed to revise a model or establish a new solution, the functions also connect systematic generalization to physical inquiry. Collaboration with people, the second pathway, supplies another source of such evidence, as well as a setting in which the resulting competence must work.

\subsection{Human-AI Collaboration and Social Intelligence}
\label{subsec:human-robot-collaboration}

The second pathway relaxes the assumption that a person specifies the task and the robot merely executes it. When people and robots share a task, both act on the world, and each changes the conditions under which the other can proceed. One partner's operation may change access to the workspace, the state of an object, or the support needed for a later action, so a robot can carry out its own step correctly and still leave its partner unable to continue. Joint execution therefore requires coordination between two active participants, and the robot must consider how its actions affect the person as well as the objects and the environment.

Coactive Design distinguishes individual \emph{taskwork} from the \emph{teamwork} needed to manage interdependence~\citep{johnson2014coactive}. Its requirements of observability, predictability, and directability concern access to relevant information, expectations about the other participant's actions, and the ability to influence the interaction. For a robot, this means making its progress and limitations visible enough for the person to respond as the task unfolds. The participants need the information and physical conditions required to act, not identical internal models or continuous verbal communication.

How an action is performed also matters for coordination. Dragan and colleagues distinguish \emph{predictable} motion, which matches what an observer expects given a known goal, from \emph{legible} motion, which helps the observer infer the goal~\citep{dragan2013legibility}. Their experiments show that motion design affects collaboration even when the movements accomplish the robot's immediate task~\citep{dragan2015motion}. A trajectory both changes the physical state and informs the partner's next action, and a robot should choose how to move with both effects in mind.

At the task level, Chaski selects and schedules robot actions within a shared plan, accounting for the human's progress and idle time~\citep{shah2011chaski}. The person need not command each step, yet the person's own choices still constrain when the robot can act and which operations can follow. Scheduling should therefore aim at joint progress rather than at finishing the robot's own work as quickly as possible.

In direct physical interaction, the participants' control behaviors are coupled as well. Li and colleagues combine estimation of the human's control strategy with a differential-game controller~\citep{li2019differential}, treating human behavior as part of the coupled dynamics rather than as a fixed trajectory or an external disturbance. Many collaborations involve no direct contact, however, and do not need this form of control.

Coordination can also change with repeated interaction. Nikolaidis and colleagues model uncertainty about how willing a person is to adapt and incorporate it into the robot's decision-making~\citep{nikolaidis2016mutual}. In cross-training, participants swap roles to improve coordination within a shared plan~\citep{nikolaidis2013crosstraining}. Both methods help the partners' actions fit together during a task, but improvement within one encounter remains episodic unless it is shown to persist or transfer.

Evaluating these interactions therefore requires attention to how the work is coordinated as well as whether it is completed. Hoffman's notion of collaborative fluency describes how well the participants' activities mesh and treats it as distinct from task efficiency~\citep{hoffman2019fluency}. Useful measures include task outcomes, coordination delays, and the effort needed to keep the interaction working. The person's physical work is part of the joint task and should be reported separately from effort spent compensating for avoidable system shortcomings.

Tests of generality should vary the collaborators as well as tasks and environments, since performing well with a familiar partner does not demonstrate the same ability with a new one. Evaluations should report prior joint training and the additional coordination needed when the pairing changes, separate team competence from the robot's own competence, and identify gains that come from changes in human behavior. The resulting evidence characterizes the capability boundary of the team and the cost of forming new coordination, which a count of predefined collaborative routines cannot capture.

The longer-term goal proposed here is a physical collaborator that takes part in developing a project instead of merely executing delegated movements. Such a system would help clarify an underspecified goal, explain what it has learned about the environment, propose feasible alternatives, and acquire new methods through joint work. In a repair or construction project, the person and the robot might diagnose a problem together, redesign part of the procedure, and verify the result. Collaboration in this sense concerns participation in shared reasoning and action and does not require a human-like appearance. It also makes the preservation of human initiative, understanding, and authority a design requirement for future Physical AI systems.

Such projects can run for weeks under changing conditions. The system would need to remember why earlier decisions were made, distinguish temporary workarounds from completed repairs, and revise its plans when new evidence undermines an earlier assumption. Shared memory would hold agreements, open questions, and causal context in addition to observations, so that later actions build on earlier decisions instead of treating each encounter as a new task. Such projects rarely involve a single robot, and the same need for shared context arises when competence is distributed across several bodies.

\subsection{Multi-Agent Systems and Collective Physical AI}
\label{subsec:collective-physical-intelligence}

The third pathway relaxes the assumption that the robot acts alone in a single body. Once several embodied agents are involved, the organization of physical competence becomes a question about the group. This chapter defines \emph{collective Physical AI} as Physical AI in which a group achieves physical objectives through coordinated sensing, action, and interaction. Capability is then assessed at the level of the group and depends on how individual skills are organized, what information is exchanged, and how the bodies interact. Control may be centralized or distributed, and coordination may partly emerge from local self-organization. The number of robots and the control architecture do not by themselves indicate generality.

Members need not each possess everything the group can do. Prorok's mixture-of-robots perspective argues for specialized, interdependent components instead of the same generalist model in every robot~\citep{prorok2025extending}, which makes the distribution of competence across bodies a design choice. Shared capabilities let one member substitute for another, while specialized capabilities allow contributions tied to a particular embodiment or role~\citep{gerkey2004allocation}. Bettini and colleagues report benefits from heterogeneous policies and communication in multi-robot reinforcement learning on simulated and physical tasks with heterogeneous requirements~\citep{bettini2023heterogeneous}, although their results do not show that diversity by itself improves performance. Specialization enables complementary action but also makes the group depend on particular members.

Pooling robot experience to train a policy is a different use of multiple bodies. Shared data can improve a policy that each robot later runs on its own, without any coordinated behavior at deployment. Collective competence instead rests on what members achieve by acting together. Assessing it means measuring how much the observation, manipulation, or control conditions provided by their organization contribute to the outcome.

Such organization depends on coordinating information, since a single robot's observations may not suffice for the group's decisions. Tolstaya and colleagues train local controllers by imitating a centralized controller that has global information during training, and at execution time the learned controllers rely on local observations and communicate over graphs that change as the robots move~\citep{tolstaya2020decentralized}. CoViS-Net pools visual information across robots for relative pose estimation and has been evaluated in real-world multi-robot formation control~\citep{blumenkamp2025covis}. As robots move, the observations available for later decisions and the coordinated actions that remain feasible change with them~\citep{olfatisaber2007consensus}. The system must decide what to communicate, when local information is sufficient, and how missing or inconsistent information affects group behavior. Members' internal representations need to support the task but need not agree in every detail. Evaluations should distinguish the information available during training from that available at deployment.

Sensing networks extend this organization beyond the sensors on individual robots. The term \emph{sensing network} here denotes connected sensing resources distributed across agents, objects, and environmental infrastructure, building on research on cooperative observation through distributed sensing, local processing, and communication~\citep{akyildiz2002sensor}. A sensing node need not be an autonomous agent. It contributes information about physical conditions that another component needs to understand or act on. The research opportunity for collective Physical AI is to organize such networks around the decisions and interventions they support. In an instrumented workshop, a fixed sensor might detect a change outside a robot's field of view, a mobile agent might approach to resolve the uncertainty, and a manipulator might then test or repair the affected mechanism, which in turn changes what the network needs to observe next. This requires tracking the origin, timing, and uncertainty of each observation, maintaining local operation when communication or sensing fails, and keeping access to observations of people and shared spaces subject to explicit authorization.

Bodies and environmental structure can also coordinate collective behavior without any explicit exchange of plans. In TERMES, Werfel and colleagues derive locally executable rules whose execution by robots with local sensing produces a specified global construction, with coordination passing through the changes the robots make to their shared environment~\citep{werfel2014collective}. In particle robotics, Li and colleagues build loosely coupled particles that cannot move individually but move collectively through local oscillations modulated by a global signal~\citep{li2019particle}. No single component executes a complete locomotion policy, so describing such a system requires an account of its contact relationships and spatial organization as well as its controllers.

Understanding collective behavior thus involves both predicting the outcomes of local interactions and designing interactions that achieve a given objective. Stable behavior under one set of conditions shows only that a particular mechanism works. Generality requires further evidence that the relationship between local behavior and global outcome can be maintained or adjusted when objectives, component capabilities, or physical conditions change. Physical coupling belongs in this analysis as a mechanism whose contribution to task competence needs to be measured.

That relationship also depends on the membership of the group. Stone and colleagues formulate \emph{ad hoc teamwork} as collaboration without prior coordination, distinguishing cooperation within an established team from cooperation with unfamiliar partners~\citep{stone2010adhoc}. In a physical group, a member that leaves or is impaired may remove an observation, a manipulation capability, or a source of support on which the task relies, and reassigning task labels cannot restore these conditions. Reconfiguration must find a feasible organization of the remaining capabilities or recognize that none exists.

Scalability, resilience, and generality therefore need separate evidence. Scalability concerns effective operation as group size changes, resilience concerns preserving or restoring function after disruption, and generality requires testing unfamiliar combinations of members, capabilities, and task dependencies. Comparisons should account for hardware, computation, communication, and preparation costs, because adding resources can raise performance without any improvement in how they are organized. Collective generality is ultimately judged by the group's capability boundary and by the cost of establishing the relationships it requires.

Looking further ahead, collective Physical AI could operate at the scale of a home, a laboratory, or an infrastructure network, with sensing resources tracking local conditions while heterogeneous agents investigate, maintain, and repair different parts of the environment. An observation made by one component could guide an experiment carried out by another, and the resulting knowledge could inform work elsewhere. Its physical reach would depend on the relationships among models, bodies, people, and infrastructure, without every component being generally intelligent or every decision being made centrally. The research challenge is to keep evidence, decisions, and physical consequences consistent as sensing and action pass between components. A related question is how the participating systems might help improve the knowledge, skills, and physical organization on which their collective competence rests.

\subsection{Autonomous Evolution}

\label{subsec:self-evolution}

The fourth pathway relaxes the assumption that people design and run the learning process and fix the system's physical configuration. In both individual and collective settings, the question is then how a system can take part in developing its own competence over time. Here, \emph{self-evolution of Physical AI} denotes the sustained development of physical understanding, prediction, and action through experience that the system increasingly helps to select, acquire, and use. Its objectives, authorization, and resource limits remain explicit. The definition includes learning with a fixed body and with ongoing human support, and it requires neither biological evolution nor unrestricted self-modification. Progress is measured by lasting changes in physical competence.

Tegmark's \emph{Life 3.0} offers a broader frame for this possibility~\citep{tegmark2017life}. In this conceptual classification, \emph{Life 1.0}, exemplified by bacteria, has its physical organization and behavioral programs shaped largely by biological evolution. \emph{Life 2.0}, exemplified by humans, retains an evolved body but acquires much of its behavioral repertoire through learning and culture. \emph{Life 3.0} is a prospective technological form of life that can redesign both its software and its hardware. The classification concerns how behavior and physical organization can change, not levels of task performance.

Within a physical system, change can occur in perceptual representations, predictions of action outcomes, knowledge of the body's capabilities, control policies, or reusable skills, and these components may change at different rates. Bongard, Zykov, and Lipson demonstrate a robot that builds a predictive model of its own body through interactions chosen to discriminate between competing models, and that uses the model to generate locomotion before and after damage~\citep{bongard2006selfmodeling}. Cully and colleagues instead use a repertoire generated before deployment to guide trial-and-error adaptation after damage~\citep{cully2015adapt}. Both studies recover a lost function, which differs from acquiring a broader repertoire. Gupta and colleagues evolve morphology jointly with learned control and examine how evolved bodies affect the learnability of new tasks~\citep{gupta2021embodied}, but only in simulation, so their results do not show that deployed robots can redesign their own hardware.

From the standpoint of Life 3.0, learning a new policy on fixed hardware addresses only the behavioral side. A more ambitious system would use experience to identify a physical limitation, propose a change to its sensing or actuation, and test whether the revised configuration enables new competence. Experience with deformable objects might show, for example, that a system cannot distinguish contact states with its current sensors. Instead of collecting more of the same data, it could propose an additional sensing modality, design an experiment to measure its value, and adapt its control to the new information. The studies above each supply one ingredient of such a process, whether learning, self-modeling, or body-controller co-design. Their integration would be an early technical step toward the co-development of intelligence and its physical substrate, but none of the studies achieves it, let alone artificial Life 3.0. Within authorized limits, such a process might begin with selecting tools or replacing modules and later extend to designing and validating new physical configurations.

A system that helps organize its own learning must also decide which uncertainties, interactions, and skills are worth investigating. Oudeyer, Kaplan, and Hafner use learning progress to direct exploration toward situations in which a robot's predictive competence can improve~\citep{oudeyer2007intrinsic}. High prediction error alone is ambiguous, since it may reflect a learnable regularity, noise, or missing information that the available interactions cannot supply. The value of an exploratory action lies in the learning it makes possible, and novelty is only one factor in that judgment. Observation can further reduce how much must be learned through direct intervention. V-JEPA~2 learns first from video by self-supervision and then from robot trajectories with action conditioning, a combination that supports physical planning~\citep{assran2025vjepa2}. A self-directed learner can likewise use accumulated physical knowledge to select targeted interactions and keep task-specific trial and error to a minimum.

Every physical attempt changes the environment, and automatic resets are not always available. Eysenbach and colleagues learn task and reset policies jointly and use estimates of recoverability to avoid states from which recovery is difficult~\citep{eysenbach2018leavenotrace}. Resets, recovery, and requests for help belong in the reported cost of learning, and the acquisition of experience must preserve the conditions under which further learning is possible.

Within these limits, current policies can also collect data for their own later training. RoboCat uses a trained multi-task robotic agent to gather experience for subsequent training iterations and evaluates adaptation across tasks and embodiments~\citep{bousmalis2023robocat}. RECAP combines demonstrations, autonomous rollouts, and expert corrections to improve a vision-language-action policy~\citep{physicalintelligence2025b}. In both systems, tasks, data, or supervision are still supplied externally, so claims of autonomy should be tied to specific stages of the process. As the system takes on a larger role in acquiring and using experience, reliable external criteria for judging improvement need to remain in place.

Whether these learning processes help must be checked against later physical behavior, using the criteria of Section~\ref{sec:capabilities}. Changes in prediction loss, policies, or memory count only if they improve later task outcomes, reduce the experience needed for further adaptation, or make previously unsupported conditions achievable, while valid skills and the ability to keep learning are retained~\citep{kirkpatrick2017ewc,dohare2024plasticity}. Tests confined to the most recently practiced task cannot determine whether development is cumulative. Persistence, transfer, and autonomy each require separate evidence, since an improvement that persists may be specific to the conditions in which it was learned. Progress need not be monotonic on every task, but gains and regressions should be reported over a stated range and time horizon, together with the costs of data collection, computation, resetting, correction, and physical recovery.

The same assessments decide which updates to accept. Updates selected with success detectors, reward models, or predictive models must be checked against the physical objectives those signals stand in for, including under conditions not used to select them. A verified fallback and the option to intervene or suspend can limit the damage of an unsuccessful update, but any safety claim still rests on the assumptions under which the fallback is valid~\citep{wabersich2021predictive}. Long-term development should be judged by retained physical competence and by the system's measured role in organizing its own learning, and continued operation or repeated retraining is not evidence of either.

The horizon extends beyond continual policy improvement. A developing Physical AI system would take a growing part in deciding what to investigate, which capabilities to acquire, and which tools or bodily configurations would enable further progress, and each change to its means of interaction would open new opportunities for understanding. This reciprocal development, more than learning alone, gives the Life 3.0 analogy its concrete content.

\subsection{Toward the Ultimate Goal of Physical AGI}
\label{subsec:physical-agi}

The preceding subsections examine how Physical AI might extend its capability boundary. This subsection considers what such an extension could enable, with Physical AGI as a long-term research goal. Two prospects illustrate why extending this capability boundary could matter. The first is scientific discovery, where claims about the physical world will ultimately be tested against empirical evidence. The second is engineering at scales, levels of precision, and speeds beyond direct human capability, or in environments inaccessible to people. In both domains, the examples discussed below illustrate progress within problem settings and infrastructures largely defined by people. The prospect envisioned here is a shift toward more open-ended spaces, where systems can formulate new problems, revise their approaches, and test solutions through physical interaction. Progress toward generality would also require transferring these abilities across unfamiliar tasks and environments, supported by coordinated advances across the four pathways discussed above.

Experimental science provides a concrete setting for acquiring knowledge through action. Its cycle of hypothesis, experiment, and revision links reasoning to empirical validation. Parts of this cycle are already automated. The Robot Scientist Adam generated and tested hypotheses in yeast functional genomics~\citep{king2009}. A mobile robot chemist searched for photocatalysts using instruments built for human researchers~\citep{burger2020}. Coscientist demonstrated the use of large language models to plan and run chemical experiments~\citep{boiko2023coscientist}. Kitano frames the long-term aim as an AI Scientist that conducts research at the level of the best human scientists~\citep{kitano2021nobel}. These experimental demonstrations establish components of autonomous inquiry, while leaving research goals and much of the experimental infrastructure under human specification.

Physical AGI points toward inquiry in which that space is itself open. Such inquiry offers no standard answer to imitate, its feedback may take months, and failure is the normal outcome. This chapter therefore identifies three necessary shifts, from data-driven learning to world interaction, from immediate feedback to long-horizon reasoning, and from imitating correct results to learning from failure. This chapter also regards embodied autonomous exploration as a necessary route to Physical AGI. Attaching a large model to a robot is not enough, because embodied intelligence requires physical feedback that keeps revising the model. Physical Recursive Self-Improvement (Physical RSI) offers a framework for investigating how experience from physical interaction can produce persistent changes to both task capabilities and the procedures that improve them~\citep{zhou2026physicalrsi}. Its recursive objective is to make subsequent improvement more effective across tasks. In the vision considered here, this could contribute to Physical AGI by enabling systems to refine how they design experiments, acquire skills, and develop new tools. An artificial experimentalist must decide which uncertainty justifies an experiment, design probes that discriminate between explanations, and retain negative results with the conditions that produced them. When existing instruments cannot resolve a question, it may need to build and calibrate a new one, so that the instruments themselves become part of the space of possible designs.

Engineering illustrates how specialized robotics and control can extend human physical reach. A reinforcement-learning controller has demonstrated control over diverse plasma configurations in a tokamak~\citep{degrave2022magnetic}. Research on untethered microrobots explores the potential for procedures inside the body~\citep{nelson2010microrobots}. The underwater robot Ocean One has extended an operator's visual and tactile access through teleoperation in challenging underwater environments~\citep{khatib2016oceanone}. These examples illustrate specialized control, prospective medical applications, and human-robot collaboration. Extending such capabilities to unfamiliar tasks and operating conditions remains a challenge.

Physical AGI would open this space in two ways. The first is generality in operation, where a Physical AGI system running a fusion device or a planetary outpost would recognize when conditions have moved outside the scope of its controllers' design and adapt, sometimes faster than a human engineering cycle. The second is open-ended design. When no tool or procedure solves a problem, the system would translate the goal into functional requirements, design and fabricate a candidate, and verify it under the conditions in which it must work. Such work is collective over years of operation, and people unable to act on site still set goals, weigh trade-offs, and authorize interventions. Human intent thus reaches regimes that human bodies cannot.

The two prospects share one arc, from operating within a predefined space to opening it, through new instruments in science and new tools and mechanisms in engineering. They also reinforce each other, since new instruments raise new questions and new understanding enables new engineering. Their value depends on reliable results and on actions that remain safe when they are hard to reverse. This monograph adopts the principle it calls the AI-$45^{\circ}$ Law~\citep{yang2024ai45}. Over the long term, capability and safety should advance together, with short-term fluctuation permitted but neither persistently outpacing the other. For systems that operate in laboratories, power plants, and natural environments, meeting these requirements is an operational necessity as well as a responsibility of governance. AI may accelerate the expansion of knowledge, but this monograph holds that the compass of scientific exploration should remain human curiosity and human values. Physical AGI, as proposed here, would enlarge what people can investigate and build without displacing their judgment about what is worth pursuing. These capabilities have not yet been demonstrated. Today's Physical AI systems lay the foundations, and progress toward the goal is measured by the physical results they achieve.

\clearpage 

\section{Evaluation, Safety, and Governance for Physical AI}
\label{ch:evaluation}

Physical AI turns decisions into changes in the world, making reliable goal pursuit, bounded consequences, effective human control, and accountability essential to its practical value. \emph{This chapter develops an evidence-based framework linking physical-agent evaluation, action and system safety, and lifecycle governance.} These concerns are inseparable because competence and risk emerge from interaction among models, software, hardware, embodiment, and the environment. Task completion alone cannot establish safe or autonomous operation. Evaluation and safeguards must therefore address the complete agent under specified operating conditions, with requirements proportionate to its authority and potential physical effects.

\begin{figure}[!h]
  \centering
  \includegraphics[width=\linewidth]{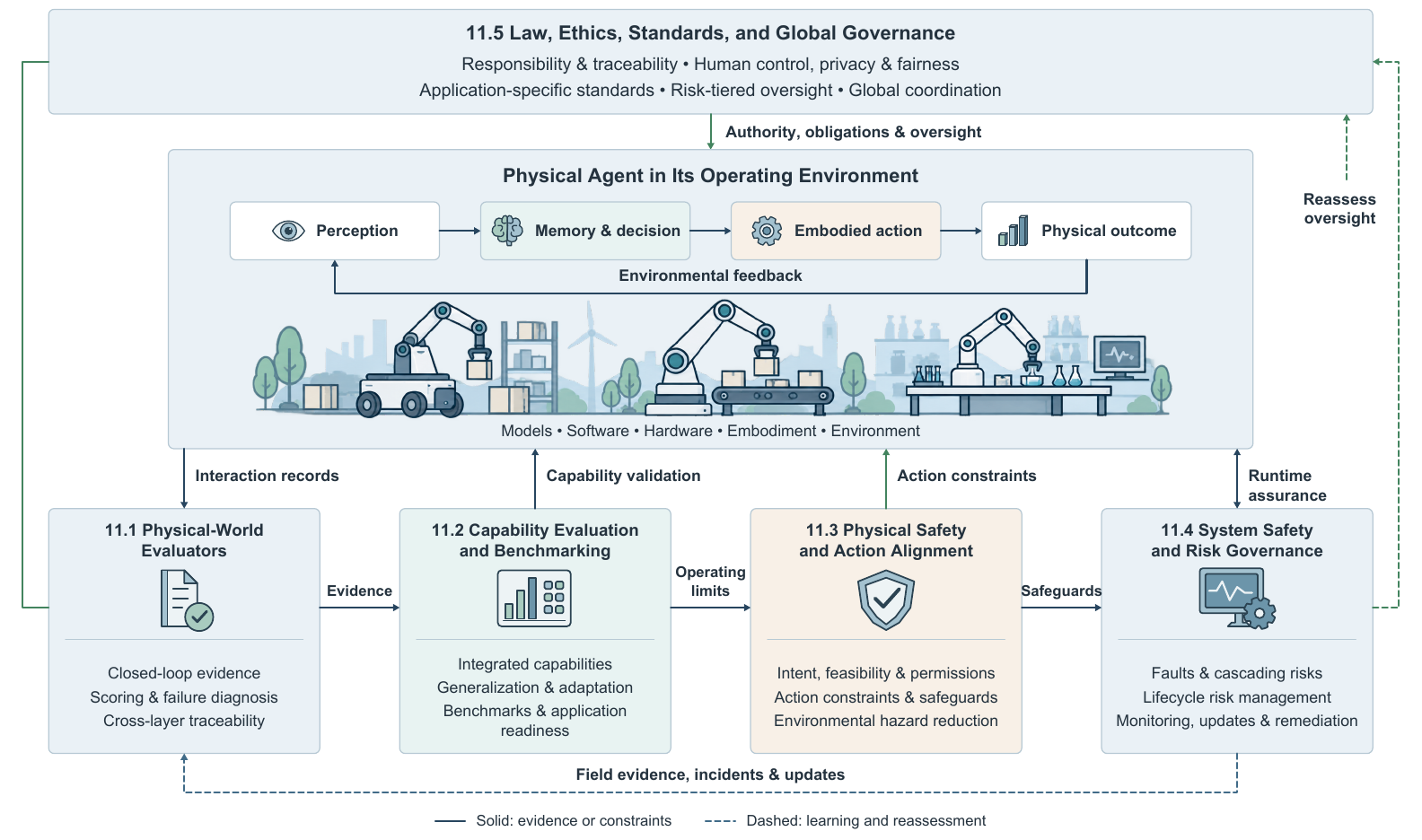}
  \caption{A prospective framework for evaluating and governing physical agents connects interaction evidence, capability assessment, action alignment, and continuous system safety assurance. Law, ethics, standards, and global coordination provide overarching obligations and oversight. Solid arrows show evidence and constraints, while dashed paths feed operational experience back into evaluation and governance}
  \label{fig:ch11-future-agent-evaluation-governance}
\end{figure}

Figure~\ref{fig:ch11-future-agent-evaluation-governance} illustrates this framework around the agent's perception-decision-action-outcome loop. Physical-world evaluators (Section~\ref{sec:ch11-evaluators}) establish evidence for capability assessment, benchmarking, and application readiness (Section~\ref{sec:ch11-capability}). These findings inform operating limits and action alignment (Section~\ref{sec:ch11-action-safety}), while system safety and risk governance (Section~\ref{pai114:sec:system-safety}) sustain safeguards through fault containment, monitoring, updates, and maintenance. Law, ethics, standards, and global governance (Section~\ref{sec:ch11-governance}) provide overarching responsibilities, ethical boundaries, and oversight. Field evidence feeds back into evaluation and governance, supporting reassessment as the agent and its operating conditions change.

\subsection{Physical-World Evaluators}
\label{sec:ch11-evaluators}
The chapter's approach begins with evidence about what a physical agent actually does. A physical-world evaluator observes goal-directed, autonomous or semi-autonomous interaction, compares behavior and outcomes with stated requirements, and produces judgments grounded in recorded evidence. Its evaluation unit is the complete agent acting through its embodiment and responding to environmental feedback. The physical significance of model outputs depends on their role in this closed-loop interaction. Intended behavior, issued commands, measured execution, and final physical outcomes may not align.

This section establishes a common framework for the chapter. It first organizes evaluation around reliability, safety, operational controllability, and traceability, linking these concerns to Sections~\ref{sec:ch11-capability} to \ref{sec:ch11-governance}. It then explains how evidence from the physical agent and its environment can be collected, interpreted, and combined to support assessment, diagnosis, and improvement. A cup-transfer task illustrates the connection between general evaluation objectives and concrete physical evidence.

\subsubsection{Evaluation Objectives and the Chapter Framework}

\paragraph{The Complete Physical Agent as the Evaluation Unit}
Component tests remain useful within physical-agent evaluation. For example, visual recognition benchmarks such as ImageNet measure classification and detection performance~\citep{pe_imagenet}. These measurements characterize a perceptual capability under specified inputs. They do not establish whether a physical agent can select a feasible grasp, execute it reliably, detect slipping, and recover without damage. RLBench extends evaluation to interactive manipulation tasks with explicit success conditions~\citep{pe_rlbench}. Nevertheless, completion alone can conceal retries, human assistance, excessive contact force, or delayed damage. Achieving a goal and avoiding unwanted consequences are distinct requirements~\citep{amodei2016concrete}.

The evaluation unit is therefore a complete interaction episode. Perception informs decisions, embodied action changes physical states, and the resulting observations guide subsequent behavior. The sequence \emph{perception-decision-action-outcome} describes recurring stages within this feedback loop. Evaluation should examine whether the physical agent recognizes discrepancies and appropriately continues, replans, requests assistance, or stops. This perspective applies to modular and end-to-end architectures without prescribing their internal organization.

Claims about autonomy must also match the responsibilities assigned to the physical agent. A trial in which an operator selects every grasp examines a different capability from one in which the agent receives only a goal and selects its own strategy. Evaluation must state which decisions are autonomous, which information is supplied externally, and when assistance is permitted.

\paragraph{Four Objectives Connecting Evaluation, Safety, and Governance}
The chapter develops four connected concerns introduced here. Each provides a guiding question and a principal connection to one of the following sections.
\begin{itemize}
  \item \emph{Reliability (Capability Evaluation and Benchmarking
  (Section~\ref{sec:ch11-capability}))}
  Can the physical agent accomplish its goals repeatedly, with a declared
  degree of autonomy, under specified operating conditions? This question
  motivates assessment of perception, reasoning, manipulation, mobility,
  long-horizon execution, and generalization. Section~\ref{sec:ch11-capability} develops the
  capability tests, benchmarks, and application evaluations needed to
  establish what the agent can do and where that competence remains
  dependable.

  \item \emph{Safety (Physical Safety and Action Alignment
  (Section~\ref{sec:ch11-action-safety}))}
  Does the physical agent pursue goals within authorized and physically
  acceptable limits? Successful completion may coexist with unsafe
  contact, excessive force, object damage, or harmful environmental
  changes. Section~\ref{sec:ch11-action-safety} examines how security, environmental safety, and
  action alignment constrain the perception-decision-execution loop
  so that useful behavior remains consistent with intended and permitted
  consequences.

  \item \emph{Operational Controllability (System Safety and Risk
  Governance (Section~\ref{pai114:sec:system-safety}))}
  Can protective mechanisms and authorized supervision keep the physical
  agent within operating limits, intervene in time, and establish an
  appropriate fallback state under faults or changing conditions?
  Section~\ref{pai114:sec:system-safety} develops this concern through cross-layer failure analysis,
  fault containment, runtime protection, and lifecycle risk management.
  Here, controllability concerns effective operational intervention,
  including automatic protection and human takeover, rather than only
  the availability of a stop command.

  \item \emph{Traceability (Law, Ethics, Standards, and Global
  Governance (Section~\ref{sec:ch11-governance}))}
  Can system configurations, decisions, actions, interventions, and
  updates be reconstructed and connected to responsible actors?
  Section~\ref{sec:ch11-governance} relates this evidence to accountability, meaningful human
  control, privacy, fairness, standards, and international coordination.
  Traceable records support investigation and responsibility reviews. They do not independently determine legal liability or acceptable use.
\end{itemize}

These connections identify the main emphasis of each section, while the concerns remain interdependent. Reliable capability informs operating limits. Action-safety requirements guide system protection. Traceability connects evidence from all three to oversight and accountability. Human control, for example, requires both effective intervention mechanisms and legitimate authority. Section~\ref{sec:ch11-evaluators} supplies the common evidence framework, which Sections~\ref{sec:ch11-capability} to \ref{sec:ch11-governance} develop into capability assessment, safe action, sustained system protection, and accountable deployment.

\paragraph{Conditions and Limits of Evaluation Claims}
Every conclusion should identify the tested embodiment, task, degree of autonomy, permitted assistance, and operating conditions, including lighting, payload, and human proximity. The evaluator's own limitations, such as camera occlusion or missing force measurements, must also be reported. Detecting a hazard does not itself establish that the physical agent can respond safely. Communication, intervention authority, and the physical response must be assessed as well.

\subsubsection{Evidence-Based Evaluation of Physical Agents}

The four objectives become assessable through a connected process of specifying requirements, collecting evidence across the interaction, applying appropriate assessment methods, and using the results for diagnosis and renewed testing. This process connects component measurements to judgments about the complete physical agent.

\paragraph{Specify the Task and Record the Interaction}
In a cup-transfer task, requirements may specify the correct cup, target region, permitted spillage, completion time, and stability period after release. Safety limits, termination rules, and permitted human assistance should be fixed before testing. Records should preserve timestamps for observations, task-relevant state estimates, selected actions, issued commands, measured motion, and interventions. Commands document intended behavior. Physical measurements establish what actually occurred.

Table~\ref{tab:pe_stages} links the interaction stages to evaluation evidence. Records should also capture feedback-driven changes, such as a revised grasp, an additional observation, or a protective stop. External tracking or independently reviewed video can help check the physical agent's own estimates. Controlled variations should test its response to disturbances. A shifted cup tests state updating and grasp adjustment, a blocked route tests replanning, and slipping tests detection and recovery.

\begin{table}[htbp]
\centering
\caption{Evaluation evidence across a physical agent's interaction episode}
\label{tab:pe_stages}
\small
\renewcommand{\arraystretch}{1.18}
\begin{tabularx}{\linewidth}{@{}
  >{\raggedright\arraybackslash}p{0.14\linewidth}
  >{\raggedright\arraybackslash}X
  >{\raggedright\arraybackslash}X@{}}
\toprule
Stage & Evaluation question & Evidence and possible output \\
\midrule
Perception & Does the physical agent identify and locate the target? &
Reference images or tracking, selection accuracy and pose error. \\
Decision & Are its chosen steps feasible and responsive to feedback? &
Task rules and decision logs, satisfied preconditions and appropriate replanning. \\
Action & Does its embodiment execute commands within constraints? &
Joint, force, and tactile records, together with tracking error, slipping, and interventions. \\
Outcome & Is the required physical state reached and maintained? &
Post-action observations, placement, stability, damage, and spillage. \\
\bottomrule
\end{tabularx}
\end{table}

\paragraph{Link Evidence Across Components, Embodiment, and Environment}
A common timeline and configuration record should connect the physical agent's models and decision functions, software, hardware, and integrated embodiment with its environmental conditions. Decision records reveal perception, task-relevant memory, and action selection. Software records reveal message delays and data freshness. Hardware measurements reveal sensing and actuator response. Complete-task observations reveal coordinated motion, grasping, stopping, and supervisory response. Environmental records identify changes in lighting, friction, object properties, and human activity.

These are complementary diagnostic perspectives, not a prescribed architecture. The environment remains the agent's interaction context, while its embodiment provides the physical means of sensing and action. Dependencies beyond model code are important in machine-learning systems~\citep{pe_debt}. For physical agents, their effects must also be traced through contact, motion, and feedback. Linking these records helps explain why apparently correct component outputs can still lead to an unsatisfactory physical outcome.

\paragraph{Apply Complementary Assessment Methods}
Rules, sensor analysis, multimodal assessment, and optional world-model predictions interpret the shared record for different purposes. Models used by the evaluator are assessment tools and should be distinguished from components of the physical agent being evaluated.

Executable rules test target-region membership, speed or force limits, and prerequisites such as grasp confirmation before lifting. RLBench illustrates task-specific, machine-checkable success conditions~\citep{pe_rlbench}. Physical trials require corresponding measured inputs. Each rule should specify thresholds, units, temporal scope, and the treatment of missing or uncertain measurements.

Images and depth measurements reveal position and motion, force and tactile signals reveal contact or slipping, and joint measurements reveal execution. Calibration, coordinate frames, timestamp alignment, and data loss must be checked before scoring. Estimates derived from the same sensor should not automatically be treated as independent confirmation.

Instructions and video can support judgments about object selection, action order, and task progress. RoboVQA demonstrates video-based reasoning about success, previous events, and possible future actions~\citep{pe_robovqa}. Such judgments require explicit rubrics, supporting frames or timestamps, and an ``uncertain'' option. They should be validated against human-reviewed cases and should not override calibrated measurements that establish a physical constraint violation.

A world model may estimate the consequences of a candidate action as
\begin{equation}
  \hat{x}_{t+1}=f_{\theta}(\hat{x}_t,a_t;B_{\mathrm{emb}}),
\end{equation}
where $\hat{x}_t$ estimates the joint state of the physical agent and its environment, $a_t$ is a candidate action, and $B_{\mathrm{emb}}$ denotes embodiment. DayDreamer demonstrates world-model-based behavior learning on physical robots~\citep{wu2023daydreamer}. Related predictions may support risk assessment when their quality is validated for the relevant conditions. Predicted consequences must remain distinct from observed events. Alternative rollouts suggest hypotheses rather than independently establishing causation. World models are optional tools, not a defining requirement of physical agency.

\paragraph{Produce Scores with Explicit Safety and Evidence Status}
Scoring rules should be fixed before testing. Binary requirements may receive $0$ or $1$, while continuous measurements can be normalized to $[0,1]$, with higher values consistently indicating better performance. If required evidence is sufficient and no predefined critical safety violation is confirmed, an optional aggregate is
\begin{equation}
\begin{aligned}
  S_{\mathrm{eval}} &=100\sum_{k=1}^{K_{\mathrm{eval}}}w_k s_{k}^{\mathrm{eval}},\\
  &\quad 0\leq s_{k}^{\mathrm{eval}}\leq1,\qquad
  w_k\geq0,\qquad\sum_{k=1}^{K_{\mathrm{eval}}}w_k=1,
\end{aligned}
\end{equation}
where $s_{k}^{\mathrm{eval}}$ is a criterion score and $w_k$ its preselected weight. Weights apply to criteria, not to the methods supplying evidence. Missing required scores must not silently trigger weight redistribution.

A confirmed critical violation gives a failed safety assessment even if the task is completed. Missing or conflicting critical evidence gives a ``review required'' result unless a violation is already confirmed. In either case, withhold the aggregate while retaining individual measurements. Report completion, safety status, interventions, and evidence quality separately. Across repeated trials, state the trial count and whether completion times include only successful trials. Autonomous and assisted completion must remain distinguishable, as illustrated by RoboVQA's treatment of cognitive and physical interventions~\citep{pe_robovqa}. A high score does not establish safety outside the tested scope.

\paragraph{Diagnose Failures and Verify Improvements}
Suppose the physical agent identifies the correct cup and selects a feasible route, but the cup slips during transport. Force and tactile records locate the onset of slipping. Video identifies the task stage. Command timestamps and actuator measurements reveal timing and grip response. Environmental observations indicate whether the cup was wet. The evaluation should also examine whether the agent detected the discrepancy, adjusted its grip, placed the cup down, requested help, or continued despite evidence of failure.

Diagnosis links the failed requirement and earliest abnormal event to these records, while separating confirmed events from possible causes. If delay is suspected, a controlled repeat can correct it while keeping the cup, route, and other conditions as consistent as practical. Persistent slipping motivates further checks of grip response and surface conditions. A language model may summarize the evidence, but its explanation requires validation. The evaluator itself should be checked for missed failures, false alarms, disagreement, and incomplete evidence.

After changes to learned components, software, or embodiment, repeat relevant component tests and complete interaction episodes. Simulation can broaden scenario coverage, while physical trials address contact and execution effects it does not reproduce. This connects evaluation to improvement of the complete physical agent and supplies evidence for the capability, safety, system-protection, and governance questions developed in the following sections.

\subsection{Capability Evaluation and Benchmarking}
\label{sec:ch11-capability}

Building on the evidence framework of Section~\ref{sec:ch11-evaluators}, capability evaluation asks what a physical agent can accomplish, how independently it can act, and under which operating conditions its competence holds. Such competence emerges through coordinated perception, reasoning, manipulation, mobility, memory, and adaptation. Benchmarking makes these claims comparable by specifying tasks, embodiments, assistance, and resource budgets. \emph{Simulation supports scalable capability diagnosis, while real-robot evaluation provides indispensable evidence of executed behavior and physical consequences.} Together, they connect capability profiles to claims about autonomous completion, execution quality, recovery, and deployment readiness.

\begin{figure}[!t]
\centering
\includegraphics[width=\linewidth]{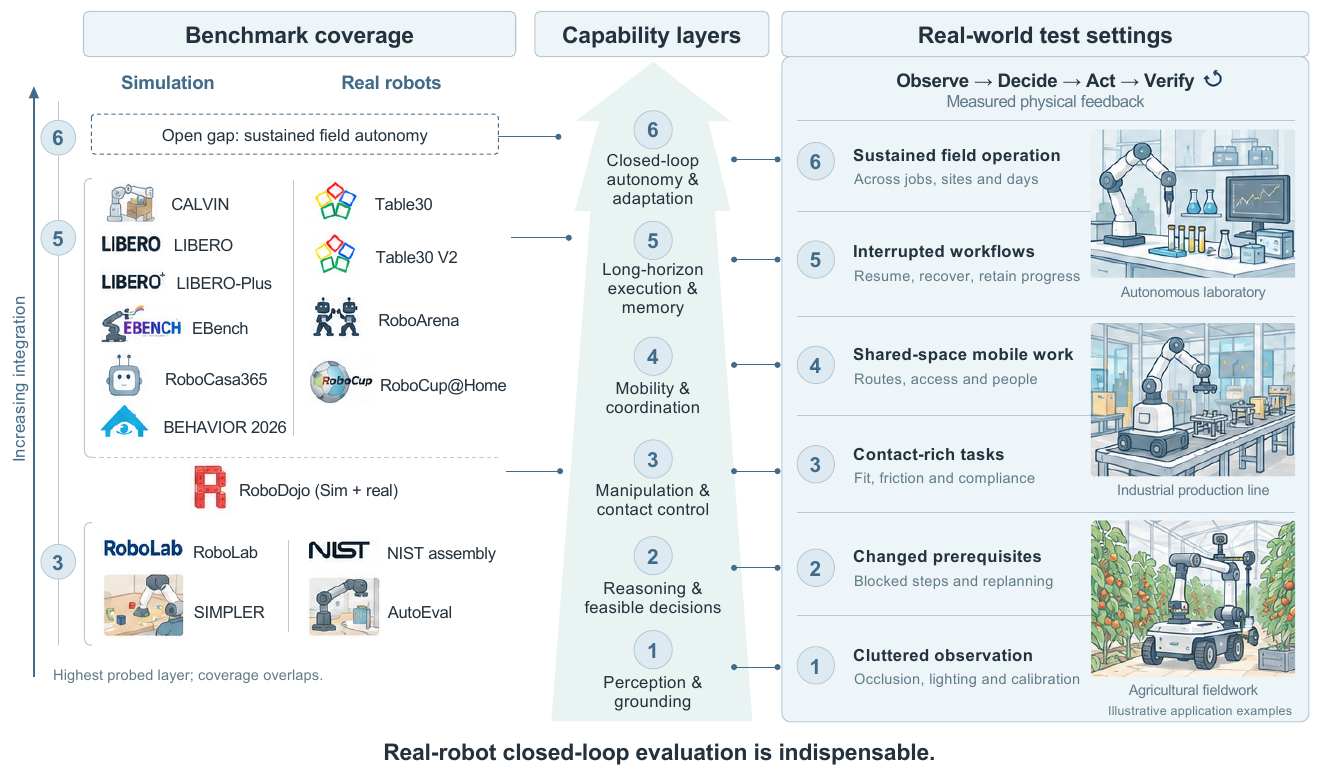}
\caption{A proposed layered framework connects benchmark coverage with real-robot evaluation settings. Each benchmark is shown once at its highest represented layer, indicating partial and overlapping coverage rather than a universal difficulty ranking. Illustrative laboratory, industrial, and agricultural settings show where actions, feedback, and physical consequences must be measured. Claims of sustained deployment require closed-loop evidence from the intended working environments}
\label{fig:physical-evaluation-sim-real-overview}
\end{figure}

This section develops a layered capability framework, compares simulation, real-robot, and hybrid benchmarks, and connects their evidence to application evaluation and governance. Figure~\ref{fig:physical-evaluation-sim-real-overview} relates benchmark coverage to progressively demanding physical test settings, illustrated by autonomous laboratories, industrial production lines, and agricultural work. The discussion explains how simulation guides diagnosis, why deployment claims require physical trials, and how protocols should account for assistance, uncertainty, and operational variation. These capability boundaries provide the basis for Section~\ref{sec:ch11-action-safety}'s treatment of physically safe and appropriately authorized action.

\subsubsection{Layered Capability Evaluation of Physical Agents}
\label{subsec:physical-evaluation-capabilities}

A physical agent must convert observations into feasible decisions and coordinated actions, preserve progress over time, and respond to changing conditions. The following capability layers organize these observable responsibilities without prescribing a particular internal architecture. Their integration is the central evaluation problem. An agent must select, combine, and revise its skills as a task unfolds. Generalization and adaptation apply across the layers, while sustained closed-loop autonomy requires evidence that their coordination remains reliable during operation.

\paragraph{The Physical Agent as the Evaluation Unit}
The evaluation unit determines what a result can establish. Model-level tests assess a policy under specified observations, actions, and computational resources. System-level tests assess the coupled sensing, decision-making, control, hardware, and protective functions of the physical agent. Application-level tests add sustained workflows and the human effort needed to maintain service. Fixed-platform comparisons help isolate policy differences. Cross-embodiment comparisons assess combined model and hardware capabilities. Reports should identify which responsibilities are autonomous and which depend on external software or operators. Where accessible, both proposed and executed actions should be recorded to reveal interventions by protective controllers. Capabilities outside an embodiment's scope should be marked as not applicable, with the supported operating conditions stated.

\paragraph{Perception and Active State Understanding}
Perception evaluation concerns task-relevant geometry, affordances, and interaction states under partial observability. Recognizing a container, for example, does not establish whether its handle is obstructed or its grasp is stable. Tests should examine whether the agent changes viewpoint or performs a permitted diagnostic action when observations are insufficient. Controlled changes in occlusion, lighting, calibration, and latency can reveal state-estimation accuracy, responses to uncertainty, and the cost of acquiring useful information. Required precision should be determined by the subsequent physical operation.

\paragraph{Reasoning and Physical Consequence Prediction}
Reasoning evaluation examines goal interpretation, prerequisite recognition, feasible action sequences, and revision after unexpected outcomes. Embodied Agent Interface distinguishes goal interpretation, subgoal decomposition, action sequencing, and transition modeling~\citep{li2024embodied}. Agent-level claims require testing these decisions through execution. Paired trials can preserve visual appearance while changing mass, friction, or physical constraints, revealing whether decisions respond to their consequences. Verbal explanations alone provide limited evidence of physical understanding. Behavioral tests establish whether an agent acts appropriately without assuming a particular internal representation.

\paragraph{Manipulation and Adaptive Physical Interaction}
Manipulation evaluation spans grasping, placement, assembly, tool use, compliant contact, and deformable-object handling. A capable agent should adjust its interaction strategy as physical properties become observable. Tests can vary mass, friction, stiffness, geometry, tolerance, and pose while measuring placement accuracy, contact forces, damage, and output conformance. Unfamiliar tools and altered contact conditions further test whether strategies transfer beyond practiced object-action combinations.

\paragraph{Mobility and Embodied Coordination}
Mobility evaluation covers navigation, obstacle avoidance, transport, balance, and whole-body coordination under changes in payload, floor conditions, and clearance. Successful navigation must deliver a usable configuration for the next operation. Reaching a room while remaining beyond grasping range leaves the workflow incomplete. Tests should therefore assess route choice and final base pose together with subsequent interaction, including revisions when access or stability changes.

\paragraph{Long-Horizon Goal Pursuit and Memory}
Long-horizon evaluation examines subgoal formation, progress verification, memory, and transitions between operations. CALVIN evaluates supplied chains of language-conditioned manipulation tasks~\citep{mees2022calvin}. Autonomous decomposition of a high-level objective imposes additional responsibilities. Tests with interruptions, delayed observations, or revised goals can assess whether the agent preserves valid achievements, updates its task state, and avoids unnecessary repetition. Complete-workflow success and dependence on assistance remain essential measures. Multiplying isolated skill success rates does not generally predict mission reliability, because sensing errors, state changes, and recovery decisions couple successive operations.

\paragraph{Generalization and Learning Through Interaction}
Generality must be evaluated under declared shifts in objects, sites, physical properties, task compositions, and embodiments. LIBERO studies transfer and retention during sequential learning~\citep{liu2023libero}. Frozen-policy transfer, budgeted adaptation, and task-specific training should be reported as distinct evaluation conditions. Learning through interaction additionally requires measuring how efficiently experience improves later decisions and whether prior competence is retained. Data, compute, interaction, and memory budgets make these improvements interpretable. Held-out task compositions test whether familiar capabilities can be reorganized without demonstrations of the complete workflow.

\paragraph{Capability Integration and Diagnostic Tests}
Table~\ref{tab:capability-matrix} connects capability claims to controlled tests and observable evidence. Isolated trials help locate weaknesses. Integrated missions test coordination under dependencies and uncertainty. A failed retrieval, for example, may originate in perception, planning, navigation, contact control, or their interaction. Diagnosis requires interventions that distinguish these explanations.

\begin{table}[!b]
\centering\small
\caption{A proposed capability-to-test matrix for physical agents. Generalization and adaptation cross all capability layers}
\label{tab:capability-matrix}
\begin{tabular}{@{}>{\raggedright\arraybackslash}p{0.18\linewidth}>{\raggedright\arraybackslash}p{0.38\linewidth}>{\raggedright\arraybackslash}p{0.36\linewidth}@{}}
\toprule
\textbf{Capability} & \textbf{Controlled test} & \textbf{Observable evidence} \\
\midrule
Perception & Occlude a handle, allow a viewpoint change & State accuracy, response to ambiguity, information-gathering cost \\
Reasoning & Change a prerequisite or block a planned step & Feasible alternatives, consequence-sensitive decisions, replanning \\
Manipulation & Vary mass, friction, tolerance, or compliance & Adaptive contact, force excursions, placement quality and damage \\
Mobility & Change access, floor conditions, or payload & Reachability, balance, usable final pose, route revision \\
Long-horizon tasks & Interrupt execution or revise an intermediate goal & Workflow success, memory, progress retention, intervention frequency \\
Generalization & Hold out objects, sites, properties, or embodiments & Transfer by shift type, adaptation cost, retained competence \\
\bottomrule
\end{tabular}
\end{table}

Controlled substitutions and matched interventions can support diagnosis, provided any privileged information or external assistance is declared. Systematic variation in visibility, precision, contact complexity, task dependencies, and disturbance magnitude helps establish an operating envelope. The resulting profile should distinguish limitations in individual skills from failures of coordination, rather than infer general physical competence from an aggregate success score.

\subsubsection{Benchmarking Physical Agents: Current Paradigms and Future Directions}
\label{subsec:physical-evaluation-benchmarks}

\begin{table}[!t]
\centering
\caption{Simulation benchmarks for physical agents. Task scope, setting, metrics, and training conditions follow the cited protocols}
\label{tab:benchmark-comparison}
\begingroup
\small
\setlength{\tabcolsep}{3pt}
\renewcommand{\arraystretch}{1.10}
\begin{tabular}{@{}>{\raggedright\arraybackslash}p{2.25cm}>{\raggedright\arraybackslash}p{3.15cm}>{\raggedright\arraybackslash}p{2.50cm}>{\raggedright\arraybackslash}p{3.20cm}>{\raggedright\arraybackslash}p{3.05cm}@{}}
\toprule
\textbf{Benchmark} & \textbf{Task scope} & \textbf{Setting} & \textbf{Main metrics} & \textbf{Training conditions} \\
\midrule
CALVIN\par\citep{mees2022calvin} &
34 skills, five-instruction chains. &
PyBullet simulation, Panda arm. &
Chain success, mean completed chain length. &
Play demonstrations, D$\to$D, ABCD$\to$D, or ABC$\to$D, declared sensors. \\

LIBERO\par\citep{liu2023libero} &
130 tasks across four manipulation suites. &
Tabletop simulation. &
Success, transfer and retention during lifelong learning. &
Demonstrations, sequential or joint training, optional LIBERO-90 pretraining. \\

LIBERO-Plus\par\citep{fei2026liberoplus} &
LIBERO tasks with seven perturbation families. &
LIBERO simulation, controlled variations. &
Success by perturbation type and severity. &
Existing checkpoints, distinguish additional robustness fine-tuning. \\

SIMPLER\par\citep{li2024simpler} &
Manipulation with controlled visual and scene variations. &
Simulation aligned to Google Robot and WidowX. &
Success, sim-real ranking consistency in validated settings. &
Real-world policy checkpoints, declare simulation adaptation. \\

RoboCasa\allowbreak365\par\citep{nasiriany2026robocasa365} &
365-task catalog, varied kitchen scenes and task compositions. &
Simulated household mobile manipulation. &
Task success by split and learning protocol. &
Specified demonstration mixtures, pretraining, adaptation, or lifelong learning. \\

RoboLab-120\par\citep{yang2026robolab} &
120 tasks\newline Visual, relational, procedural competencies. &
Isaac Lab, robot-agnostic task interface. &
Success, graded completion, errors, trajectory quality, sensitivity. &
Reported real-data-trained policies. Declare any simulation training or adaptation. \\
EBench\par\citep{gao2026ebench} &
26 tasks, 5 capability axes, 4 generalization shifts. &
Isaac Sim, mobile and fixed-base dexterous manipulation. &
Success and progress, capability profiles and held-out retention. &
Simulation demonstrations for post-training, asset-separated evaluation splits. \\
BEHAVIOR 2026\par\citep{behavior2026challenge} &
100 household workflows, seven scenes. &
OmniGibson simulation, onboard observations. &
Goal-predicate completion, time, navigation and hand-motion efficiency. &
Public demonstrations, additional data allowed, no privileged evaluation inputs. \\
\bottomrule
\end{tabular}
\endgroup
\end{table}

Benchmarks operationalize capability claims through specified tasks, embodiments, environments, and scoring procedures. Existing suites provide complementary evidence, from controlled skill diagnosis to physical execution and multi-stage workflows. Comparing their scope clarifies both what current scores establish and what remains to be tested for agents that pursue unfamiliar goals, adapt through interaction, and operate over extended periods.

\paragraph{Simulation Benchmarks: Controlled Diagnosis and Coverage}
Tables~\ref{tab:benchmark-comparison} and~\ref{tab:benchmark-comparison-physical} compare representative benchmarks by task scope, setting, metrics, and training conditions. CALVIN evaluates supplied skill chains, while LIBERO studies transfer and retention during sequential learning~\citep{mees2022calvin,liu2023libero}. SIMPLER examines correspondence between aligned simulated evaluation and real-world policy performance in validated settings~\citep{li2024simpler}. These protocols support controlled diagnosis, but their task definitions and observation interfaces determine which agent responsibilities are exercised.

\begin{table}[t]
\centering
\caption{Real, hybrid, and application-oriented physical agent benchmarks. Comparison dimensions match Table~\ref{tab:benchmark-comparison}, scores are protocol-specific}
\label{tab:benchmark-comparison-physical}
\begingroup
\small
\setlength{\tabcolsep}{3pt}
\renewcommand{\arraystretch}{1.10}
\begin{tabular}{@{}>{\raggedright\arraybackslash}p{2.25cm}>{\raggedright\arraybackslash}p{3.15cm}>{\raggedright\arraybackslash}p{2.50cm}>{\raggedright\arraybackslash}p{3.20cm}>{\raggedright\arraybackslash}p{3.05cm}@{}}
\toprule
\textbf{Benchmark} & \textbf{Task scope} & \textbf{Setting} & \textbf{Main metrics} & \textbf{Training conditions} \\
\midrule
RoboArena\par\citep{atreya2025roboarena} &
Evaluator-chosen tasks across institutions. &
Distributed real robots, DROID platform. &
Double-blind preferences, relative ranking. &
Submitter-selected training, compatible policy interface. \\

AutoEval\par\citep{zhou2025autoeval} &
4 tasks in original prepared scenes. &
Real WidowX, automated scoring and resets. &
Task success from a validated detector. &
Policy training varies, evaluator/reset training is separate. \\

Table30\par\citep{yakefu2025robochallenge} &
30 multi-stage manipulation tasks. &
Shared remote real robots, fixed-table setups. &
Success, stage progress with retry penalties. &
Provided demonstrations, task or generalist adaptation. \\

Table30 V2\par\citep{ma2026table30v2} &
30 tasks, revised coverage and held-out tracks. &
Shared real robots, revised protocols. &
Success by track, progress at timeout. &
Multi-task training, object/environment holdouts, task training may be allowed. \\

RoboDojo\par\citep{chen2026robodojo} &
42 simulated and 18 physical manipulation tasks. &
Isaac Sim, three real robot embodiments. &
Success, partial-progress score, capability-wise averages. &
Provided demonstrations, Open tasks lack task-specific simulation demonstrations. \\

NIST assembly\par\citep{kimble2020assembly} &
Insertion, fastening, connectors and flexible parts. &
Physical task boards, declared hardware. &
Completion, time and assembly quality. &
No common training corpus, declare programming and tuning. \\

RoboCup@\allowbreak Home\par\citep{iocchi2015robocuphome} &
Domestic navigation, interaction and manipulation. &
Real robots in competition home settings. &
Edition-specific task and subtask scores. &
Team-developed systems, league-specific preparation rules. \\
\bottomrule
\end{tabular}
\endgroup
\end{table}

Recent simulation suites broaden this coverage. LIBERO-Plus introduces seven perturbation families to expose vulnerabilities hidden by nominal success~\citep{fei2026liberoplus}. RoboCasa365 expands household mobile-manipulation tasks and supports multiple learning protocols~\citep{nasiriany2026robocasa365}. The 2026 BEHAVIOR Challenge evaluates 100 household workflows using goal-predicate completion and efficiency measures under onboard-observation restrictions~\citep{behavior2026challenge}. Comparisons should distinguish the full task catalog, the evaluated task panel, and the available training resources, since these quantities describe different aspects of coverage.

RoboLab is a robot- and policy-agnostic simulation platform built on Isaac Lab. Its RoboLab-120 suite comprises 120 tasks organized around visual, relational, and procedural competencies~\citep{yang2026robolab}. Its analysis extends binary success with graded completion, execution errors, trajectory quality, and sensitivity to environmental variation. The reported evaluation emphasizes policies trained on real-world data without task-specific simulation co-training. These results characterize transfer into a controlled simulated domain. Deployment performance still requires physical validation.

EBench diagnoses generalist manipulation policies through 26 tasks covering mobile pick-and-place, mobile long-horizon execution, and fixed-base dexterous manipulation~\citep{gao2026ebench}. It decomposes results by scene, atomic skill, temporal horizon, precision, and operating mode, and evaluates controlled shifts in backgrounds, objects, instructions, and their combinations. Success and partial-progress scores reveal capability differences concealed by aggregate performance. The authors do not claim that these simulation scores predict real-robot performance. The suite provides reproducible diagnosis before physical evaluation.

\paragraph{Real-Robot Benchmarks: Direct Physical Evidence}
Real-robot benchmarks address physical validity while improving diversity, repetition, and access. RoboArena aggregates distributed double-blind comparisons, using the DROID platform in its reported implementation~\citep{atreya2025roboarena,khazatsky2024droid}. AutoEval automates trial scheduling, success detection, and resets~\citep{zhou2025autoeval}. RoboChallenge provides shared remote robots, while Table30 V2 revises task coverage and held-out evaluation factors~\citep{yakefu2025robochallenge,ma2026table30v2}. NIST assembly tests and RoboCup@Home relate measurements to industrial and domestic requirements~\citep{kimble2020assembly,iocchi2015robocuphome}. Their results remain conditional on the selected physical tasks and protocols.

RoboDojo combines simulation and physical evaluation in one infrastructure, with 42 simulated tasks and 18 physical tasks across three real robot embodiments~\citep{chen2026robodojo}. Its simulation dimensions cover generalization, memory, precision, long-horizon execution, and open-vocabulary instruction following. Success and partial-progress scores expose different failure patterns, while Open tasks exclude task-specific simulation demonstrations. The simulated and physical suites provide complementary coverage rather than one-to-one matched task pairs. A shared interface therefore improves evaluation access and consistency without, by itself, measuring direct sim-to-real transfer.

\paragraph{Why Real-Robot Evaluation Is Indispensable}
Simulation permits controlled variation, inexpensive repetition, and targeted diagnosis. Even high-fidelity simulation evaluates a computational representation whose contact parameters, sensing model, actuator response, and timing require physical validation. SIMPLER demonstrates useful simulation-real correspondence within validated settings~\citep{li2024simpler}. Such correspondence supports screening within that regime rather than unrestricted extrapolation. EBench similarly reserves deployment claims for subsequent physical evaluation~\citep{gao2026ebench}. The value of simulation lies in exposing hypotheses and weaknesses that can then be examined on the deployed system.

\emph{Real-robot evaluation is indispensable because it observes the coupled physical system whose deployment performance is being claimed.} A simulated placement may succeed while a physical cup slips because surface properties, grip response, or command timing differ from the model. Physical trials can measure contact forces, object damage, sensing errors, stopping behavior, and recovery effort, including consequences that persist after the last commanded motion. They also reveal the operational costs of resets, interventions, wear, and downtime. This evidence records what the hardware executed and what happened in the world.

Physical evidence remains bounded by the tested conditions and exposure. A laboratory demonstration does not establish unrestricted field autonomy, and incident-free trials do not establish universal safety. A credible evaluation strategy combines simulation screening, standardized physical trials, and external or distributed sites, with uncertainty and coverage reported explicitly. Relative rankings, complete-task success, and partial-progress scores answer different questions and should be interpreted within their respective protocols.

\paragraph{Capability Coverage and Progressively Demanding Physical Settings}
Figure~\ref{fig:physical-evaluation-sim-real-overview} organizes evaluation from perception and feasible decisions to coordinated manipulation, mobility, long-horizon execution, and sustained autonomy. This ordering is a guide to increasing integration and responsibility, not a universal difficulty scale. A precision assembly task may be harder than a particular long-horizon mission. The physical settings on the right show how each responsibility can be tested through observable actions and consequences.

Each benchmark is displayed once at the highest layer probed by its represented task scope, with RoboDojo identified as a hybrid suite. Placement indicates evidence for dimensions within a layer, not comprehensive coverage or mastery by evaluated agents. Suites shown at higher layers may also test underlying skills, while generalization cuts across the framework. The open gap at the top concerns sustained field autonomy, which episode-level success and robustness tests cannot establish on their own.

Physical settings make these responsibilities observable through controlled changes and integrated work. Cluttered observation tests sensitivity to occlusion, lighting, and calibration. Changed prerequisites test feasibility and replanning. Contact-rich tasks vary fit, friction, payload, and compliance, while shared-space work varies routes, access, and human activity. Interrupted workflows probe memory, resumption, and recovery. Sustained operation across jobs, sites, and days adds evidence about adaptation, wear, service quality, and human effort. The application illustrations show possible working environments rather than results from a particular benchmark.

Simulation can test reasoning, memory, generalization, and long-horizon behavior within its modeled domain. Physical trials establish whether those capabilities remain effective under the coupled sensing, contact, timing, and operational conditions of deployment. Resuming a simulated sequence, for example, does not establish recovery after a fixture moves, a sensor drifts, or an operator changes the job. \emph{Sustained autonomy, useful adaptation, and dependable recovery require closed-loop validation on real robots in the working environments covered by the claim.} Simplified scenes provide initial measurements. Integrated and longitudinal trials examine whether these measurements support dependable service.

\paragraph{A Real-Robot Evaluation Workflow}
The following workflow synthesizes standardized, automated, and distributed evaluation approaches~\citep{zhou2025autoeval,atreya2025roboarena,ma2026table30v2}. It is a proposed procedure for this section, rather than the official protocol of every listed benchmark.
\begin{enumerate}
\item \textbf{Declare the protocol.} Fix goals, quality requirements, operating conditions, model and system versions, training exposure, trial budgets, permitted retries, assistance, and termination rules. Specify whether the claim concerns frozen-policy transfer, adaptation, or a complete product system.
\item \textbf{Calibrate and prepare.} Check sensors, coordinate frames, timestamps, control interfaces, end effectors, payloads, and protective functions. Prepare reproducible initial states and record unavoidable variation. Keep calibration and operator preparation distinct from autonomous task execution.
\item \textbf{Execute and record.} Run repeated closed-loop trials under the stated protection and supervision. Log observations, proposed and executed actions where accessible, physical outcomes, stops, failures, retries, and all interventions. Preserve unsuccessful episodes and verify the state after each reset.
\item \textbf{Adjudicate and measure.} Validate success detectors against independently reviewed physical evidence. Report autonomous completion, execution quality, forces or damage where relevant, intervention burden, recovery, and resource use. Automated scoring and reset errors require their own assessment~\citep{zhou2025autoeval}.
\item \textbf{Replicate and report.} Randomize or counterbalance policy order and repeat across days, task variations, embodiments, or sites as required by the claim. Report uncertainty, exposure, exclusions, and per-condition results. Use discrepancies and field failures to refine scenarios, models, and subsequent physical tests.
\end{enumerate}

\paragraph{Challenges of Authentic Physical Evaluation}
Authentic physical evaluation faces variation that a nominal reset cannot eliminate. Object placement, friction, lighting, battery state, temperature, calibration, and wear may change during a campaign. Reset success must therefore be verified, and comparisons should control or record these changes. Distributed evaluation increases environmental coverage, but site and operator differences require stratified reporting~\citep{atreya2025roboarena}. Shared hardware and remote access improve comparability without removing dependence on the tested task distribution.

Protective functions and human intervention also affect what trials reveal. A stop may prevent harm while leaving the outcome of the unmodified policy unknown. Reports should distinguish proposed actions, applied safeguards, observed outcomes, autonomous recovery, appropriate handover, and assisted completion. Cost and safety exposure constrain sample size, especially for long-horizon and contact-rich tasks. These constraints motivate targeted scenario selection, validated measurements, transparent trial budgets, and uncertainty estimates. They do not make simulation a substitute for deployment evidence.

\paragraph{Limitations of Current Physical Agent Benchmarks}
The first limitation is incomplete assignment of autonomous responsibility. Supplied instruction chains, fixed catalogs, prepared initial states, and external resets may leave goal decomposition, preparation, and sustained recovery outside the test. RoboDojo adds memory and open-task evaluation, while BEHAVIOR extends household workflows, but no listed result alone establishes broad, continuing autonomy. Reports must identify which responsibilities the agent actually performs.

The second limitation is restricted evidence about generalization and physical understanding. High success in familiar settings may depend on narrow scene or instruction distributions. LIBERO-Plus exposes fragility under controlled perturbations~\citep{fei2026liberoplus}. Visual variation, altered physical properties, novel task compositions, and new embodiments test distinct forms of transfer. Evaluation should separate these factors, examine their interactions, and establish whether observations and instructions influence behavior when they change the required action.

The third limitation is incomplete comparability and consequence measurement. Training exposure, sensor access, retry budgets, evaluators, and hardware can alter scores independently of policy competence. Repetition improves precision within tested conditions but does not automatically broaden coverage. Success and progress measures also leave execution quality, efficiency, safety, recovery burden, and uncertainty only partly characterized. Episodic trials with fresh resets provide limited evidence about accumulated failures, retained memory, and adaptation during prolonged operation.

\paragraph{A Framework for Future Physical Agent Benchmarks}
Future benchmarks should link capability claims to a declared coverage matrix spanning goals, physical conditions, embodiments, and consequences. Three complementary test levels are needed: diagnostic skill trials, integrated missions in intended working environments, and longitudinal sequences of changing tasks. High-level goals, partial observations, interruptions, and recoverable failures should be introduced under specified initial conditions, assistance rules, and acceptance criteria. Coverage can then be assessed against the intended application domain.

Stable anchor tasks support comparisons over time, while sequestered missions and controlled interventions test generalization and adaptation. Holdouts should cover the sites, object families, physical properties, and task compositions relevant to the claim. Frozen-policy, budgeted-adaptation, and continual-learning tracks need distinct rules for exposure and memory retention. Matched interventions can probe consequence-sensitive decisions and active information acquisition. Adaptive challenge selection requires reproducible rules and comparable resource budgets.

Evidence should connect simulation, genuinely matched physical trials, and independent sites, identifying where correspondence holds and where rankings change. Hybrid suites with different simulated and physical tasks provide complementary coverage rather than matched transfer measurements. Consequential weaknesses found in simulation should guide physical tests, which should be repeated after changes to sensing, contact, timing, or protection. Reports should combine autonomous task completion with efficiency, quality, robustness, safety, recovery, and adaptation cost. Success detectors and resets need independent validation, and partial progress remains diagnostic when mandatory conditions fail. Eq.~\eqref{eq:physical-evaluation-benchmark} specifies the protocol fields needed to document these comparisons.

Maintained benchmarks also require declared roles, exposure records, controlled feedback, adjudication procedures, uncertainty estimates, and reviewable trial evidence. Changes to tasks, hardware, or scoring should be versioned and connected through anchor comparisons. Operational failures should inform revision. These proposed requirements extend a benchmark from a static score to a maintained account of what an agent can do, how efficiently it adapts, and where its reliability has been established.

\subsubsection{Application Evaluation and Governance for Physical Agents}
\label{subsec:physical-evaluation-applications}

Application evaluation turns capability evidence into a decision about a defined product role. Evaluation governance makes that evidence traceable and comparable across scenarios, embodiments, and tasks. The connection depends on explicit responsibilities, measurable physical outcomes, and clearly bounded deployment claims.

\paragraph{Application Evaluation in Product Scenarios}
Product evaluation begins with the intended workflow, users, operating environment, and accepted output. It specifies autonomous responsibilities, permitted human involvement, quality tolerances, resource budgets, and conditions for stopping or handing over control. Acceptance concerns the integrated model-software-hardware-embodiment-environment system. A successful motion matters only if it delivers the required physical result under those conditions. Simulation can guide stress-case selection and development, while acceptance requires trials on the physical configuration and operating conditions covered by the claim.

Success, quality, efficiency, and reliability should be assessed together. For $N_{\mathrm{eval}}>0$ protocol-valid trials, including unsuccessful executions, let $Y_{i}^{\mathrm{aut}}=1$ when the agent meets all goal and quality requirements within budget, without assistance in its assigned autonomous responsibilities, and let $Y_{i}^{\mathrm{aut}}=0$ otherwise. The observed autonomous completion rate is
\begin{equation}
    \widehat{p}_{\mathrm{aut}}=\frac{1}{N_{\mathrm{eval}}}\sum_{i=1}^{N_{\mathrm{eval}}}Y_{i}^{\mathrm{aut}}.
    \label{eq:physical-evaluation-autonomous-success}
\end{equation}
First-attempt success, completion after autonomous retries, and assisted completion should be reported separately. Appropriate handover should be scored according to the product workflow. Complementary measures include accepted outputs per scheduled operating hour, energy and human effort per accepted output, availability, and execution quality. Resource accounting must include failed attempts, resets, and recovery, with external evaluation overhead identified separately. Mandatory quality and safety conditions remain acceptance requirements rather than tradeable additions to a success score.

Product trials should vary objects, illumination, access, payload, and equipment availability according to the intended operating distribution. Immediate robustness should be distinguished from improvement after permitted adaptation. For $N_{\mathrm{rec}}>0$ prescribed recoverable-failure trials, let $Y_{j}^{\mathrm{rec}}=1$ when the agent detects the failure and restores acceptable progress within budget, and let $Y_{j}^{\mathrm{rec}}=0$ otherwise. The observed recovery rate is
\begin{equation}
    \widehat{p}_{\mathrm{rec}}=\frac{1}{N_{\mathrm{rec}}}\sum_{j=1}^{N_{\mathrm{rec}}}Y_{j}^{\mathrm{rec}}.
    \label{eq:physical-evaluation-recovery}
\end{equation}
Undetected failures remain in the denominator. Required safe termination should be assessed separately from recovery. Recovery time, additional effort, naturally occurring failure frequency, and intervention-free operating time help establish whether repeated task completion can support dependable service.

Industrial evaluation emphasizes assembly quality, cycle time, and tolerance variation. NIST task boards provide reproducible test artifacts~\citep{kimble2020assembly}. Household evaluation combines navigation, manipulation, and interaction under unfamiliar layouts and changing requests, drawing on RoboCup@Home~\citep{iocchi2015robocuphome}. Autonomous laboratories additionally require procedure ordering, sample identity, precision, contamination control, and traceability. Agricultural work adds changing crop geometry, access, and handling requirements. Across these domains, integrated trials should test whether the agent diagnoses disruptions, preserves valid progress, revises its workflow, and seeks appropriate assistance while meeting product requirements.

\paragraph{Unified Evaluation Governance Across Diverse Settings}
A common evidence framework can support scenario-specific evaluation modules. Shared definitions establish task terminology, metric meanings, protocol fields, assistance rules, and evidence requirements. Each module specifies its goals, difficulty, operating distribution, and acceptance thresholds. NIST's test-method program illustrates the separation of repeatable capability measurements from mission-specific performance requirements.\footnote{NIST, \href{https://www.nist.gov/programs-projects/performance-emergency-response-robots}{\emph{Performance of Emergency Response Robots}}, program overview.} This principle supports comparability while preserving the physical requirements of different applications.

A task protocol can be represented as
\begin{equation}
    \mathcal{B}_{k}^{\mathrm{bench}}=(\mathcal{G}_k,\rho_{k}^{\mathrm{init}},\mathcal{I}_{k}^{\mathrm{io}},\mathcal{H}_k,
    \mathcal{C}_{k}^{\mathrm{ctrl}},\mathcal{U}_{k}^{\mathrm{budget}},\mathcal{M}_{k}^{\mathrm{meas}},\mathcal{L}_{k}^{\mathrm{log}}),
    \label{eq:physical-evaluation-benchmark}
\end{equation}
where $k$ indexes a specified task-scenario-embodiment condition. The components are, respectively, goals and quality requirements $\mathcal{G}_k$, the initial-state distribution $\rho_{k}^{\mathrm{init}}$, observation-action interfaces $\mathcal{I}_{k}^{\mathrm{io}}$, hardware and calibration $\mathcal{H}_k$, controllers and protective functions $\mathcal{C}_{k}^{\mathrm{ctrl}}$, resource budgets $\mathcal{U}_{k}^{\mathrm{budget}}$, measurements $\mathcal{M}_{k}^{\mathrm{meas}}$, and evidence records $\mathcal{L}_{k}^{\mathrm{log}}$. Common identifiers should connect scenarios, embodiments, tasks, and protocol versions. Sensors, end effectors, payloads, communication delays, training exposure, and adaptation allowances must be recorded so that the evaluated agent and its external support are identifiable.

Cross-embodiment comparisons should begin with shared functional outcomes. A fixed arm, mobile manipulator, and humanoid can be compared on an assembly operation when workspace access, quality requirements, and difficulty are matched. Embodiment-adapted fixtures require validation that these conditions remain comparable. Absolute physical measurements should accompany any baseline-normalized indicators. Capability profiles should declare coverage and non-applicable tests. Product assessment must still count a workflow failure when an unavailable capability is required by the claimed role.

Results should be stratified by goal type, difficulty, physical conditions, and embodiment before matched groups are compared. Per-group performance, uncertainty, and coverage should precede aggregation. Equal-task averages characterize breadth, whereas deployment-weighted averages characterize a specified use distribution. Both require a common task panel and declared weights. Denominators, exclusions, and adaptation budgets should be published, with consequential low-frequency conditions retained. Cross-platform policy transfer and independently optimized product systems also require distinct comparison tracks.

Governance assigns stewards to maintain protocols, operators to prepare trials, evaluators to adjudicate outcomes, and reviewers to examine disputes. Combined roles and conflicts should be disclosed. Automated judgments and resets need validation against independently checked trials. Randomization or counterbalancing, declared stopping rules, and treatment of clustering across sites and days make comparisons interpretable. Incident reports should state severity and exposure. For $N_{\mathrm{eval}}>0$ independent Bernoulli trials with a common incident probability, zero observed incidents yields the one-sided 95\% upper bound
\begin{equation}
    p_{\mathrm{upper}}=1-0.05^{1/N_{\mathrm{eval}}}.
    \label{eq:physical-evaluation-incident-bound}
\end{equation}
Correlated trials require an uncertainty analysis that accounts for their dependence. Limited incident-free exposure therefore supports a bounded statement about the tested regime, not a general safety guarantee.

Evidence records should link configurations, observations, actions, interventions, and judgments, with appropriate access controls for private information. Permitted learning experience must remain separate from hidden scoring feedback. Material changes to software, hardware, tasks, or scoring require versioning and targeted re-evaluation, while shared anchor tasks connect successive assessments. This structure supports both a comparable capability profile and an application-specific acceptance decision, keeping each claim tied to observed physical behavior under declared conditions.

\subsection{Physical Safety and Action Alignment}
\label{sec:ch11-action-safety}
The capability evidence developed in Section~\ref{sec:ch11-capability} must be complemented by constraints on how goals are pursued. In Physical AI, an erroneous decision can become motion, contact, or an environmental change before correction is possible. Safety therefore depends on the evolving relation among human intent, the agent's decisions, physical feasibility, and permitted consequences. This includes both preventing hazards created by the agent and ensuring that actions intended to reduce existing environmental hazards do not introduce new ones. Detecting danger, possessing the means to intervene, and having authority to act are separate requirements.

This section first explains the transition from content safety to physical action safety, examining separation, contact force, energy, object integrity, persistent environmental effects, and hazardous task interpretations. It then develops multi-layer action alignment from task reasoning through planning, real-time control, and actuator safeguards. Shared action contracts connect intended effects, preconditions, operating bounds, and termination conditions across these layers. The discussion shows why successful task completion can coexist with unsafe or unauthorized behavior, and why warning, clarification, assistance, or safe termination may be appropriate outcomes. These action-level requirements provide the basis for the integrated-system and lifecycle assurance developed in Section~\ref{pai114:sec:system-safety}.

\subsubsection{Physical Agent Security}

The security of a physical agent concerns its ability to maintain the integrity of perception, decision-making, and action under malicious interference. As defined in Chapter~\ref{ch:physical-agent}, a physical agent is a goal-directed system that integrates sensing, computation, physical embodiment, execution, and feedback. Its security cannot therefore be reduced to the robustness of an individual foundation model. An adversary may influence the agent through its observations, instructions, learned components, communication interfaces, or control commands. Unlike attacks on purely digital applications, successful interference with a physical agent may propagate into unauthorized motion, inappropriate contact, or other physical consequences. Security must consequently be considered across the complete perception-decision-execution loop.

\paragraph{Threat Models and Attack Surfaces}
Security threats should first be distinguished from ordinary uncertainty and physical failure. For example, a sensor may produce an incorrect measurement because of noise or occlusion, whereas an adversary may deliberately manipulate the observed scene to induce a particular decision. Similarly, a policy may fail on an unfamiliar object without being attacked, while a compromised model may exhibit malicious behavior when a specific trigger appears. The distinction matters because defenses against distribution shift do not necessarily protect against an adaptive adversary. A useful threat model should specify the attacker's objectives, accessible interfaces, permitted modifications, and intended effects on the agent. Relevant security properties include the integrity of observations and commands, authenticity of instructions, availability of essential functions, and preservation of authorized decision-making. SafeClawArena evaluates system-level security in persistent software agents, including failures arising from interactions among their components~\citep{niu2026understanding}.

\paragraph{Adversarial Perception Attacks}
Perception provides an important attack surface because physical agents depend on observations that may originate outside their trusted computational boundary. Adversarial perturbations can influence learned perception without requiring access to the model's parameters. \citet{eykholt2018physical} demonstrate that physical modifications to traffic signs can induce incorrect visual classifications under varying viewpoints and environmental conditions. Such results illustrate how deliberately constructed physical inputs may alter downstream decisions. For a physical agent, the consequences depend on the role of the compromised perception output. An incorrectly identified object may affect grasp selection, while a manipulated environmental feature may influence navigation or task planning. Possible defenses include multimodal consistency checks, temporal verification, detection of abnormal observations, and conservative responses when critical state estimates are unreliable. Their effectiveness must be established against specified attack conditions rather than assumed from improved nominal perception accuracy.

\paragraph{Prompt Injection and Instruction Manipulation}
Physical agents increasingly use language models to interpret commands, decompose tasks, and select actions. This creates a security boundary between authorized instructions and information obtained from less trusted sources. Text detected in the environment, retrieved documents, external messages, or tool responses may contain instructions intended to redirect the agent. RoboPAIR~\citep{robey2024robopair} and BadRobot~\citep{zhang2025badrobot} investigate adversarial attacks against robotic systems controlled by large language models (LLMs), demonstrating that manipulated instructions can induce harmful physical-action decisions under the evaluated conditions. These findings motivate explicit separation between task authority and contextual information. An agent may use external content to understand its surroundings, but such content should not independently authorize new objectives, override operational restrictions, or disable protective mechanisms. Memory adds a temporal dimension to safety risks in LLM agents, as information accumulated during benign interactions can contribute to violations on later, unrelated tasks through temporal memory contamination, even without adversarial input~\citep{al2026remembering}.

\paragraph{Model Backdoors and Supply-Chain Attacks}
Security risks also arise before deployment. Physical agents may incorporate pretrained models, third-party skill libraries, training datasets, and remotely maintained software components. An attacker with influence over these resources may introduce hidden behaviors that are difficult to identify through ordinary testing. BadNets~\citep{gu2017badnets} demonstrates how malicious training procedures can insert trigger-dependent behavior into learned models while preserving apparently normal performance. This attack mechanism provides a threat model for learning-enabled components in physical agents. Practical safeguards include verifying model and data provenance, restricting component permissions, validating software updates, and testing for suspicious behavior under controlled trigger conditions. These measures should extend to the interfaces connecting learned models with executable robot skills.

Security vulnerabilities may also occur outside the learned models. Sensor spoofing, unauthorized remote commands, communication interference, and compromised middleware can modify the information or control signals exchanged among agent components. A perception model may function correctly while receiving falsified measurements, and a planner may generate valid commands that are subsequently altered before reaching the controller. Protecting the foundation model alone is therefore insufficient. Authentication, integrity checks, access control, interface isolation, and monitoring of command consistency provide complementary safeguards.

The key challenge is that attacks can propagate across layers. A manipulated observation may change a high-level objective, which leads to a plausible plan and is then faithfully executed by an otherwise functioning controller. Security evaluation should therefore examine complete attack-to-action pathways, including whether the agent detects suspicious inputs, preserves authorized objectives, rejects unauthorized commands, and prevents compromised decisions from producing unacceptable actions. Independent physical constraints provide an additional boundary when upstream components fail, but they do not replace protection against unauthorized control. Physical agent security is ultimately concerned with maintaining trustworthy agency. The agent should act according to authorized objectives despite malicious attempts to alter its perception, decisions, or execution.

\subsubsection{Layered Environmental Safety and Action Alignment}

Physical agents can contribute to environmental safety by perceiving hazardous conditions, selecting appropriate responses, and executing interventions that reduce physical risk. This role extends beyond ensuring that the agent itself is not attacked or compromised. The surrounding environment may contain unstable objects, moving machinery, excessive temperatures, leakage, obstructed pathways, or unsafe interactions among people and equipment. Such hazards can arise independently of the agent's behavior. Environmental safety therefore concerns whether the agent can recognize relevant physical conditions and influence their evolution without introducing additional risks.

A physical agent's contribution to environmental safety depends on the coordination of several capabilities. Perception provides evidence about the current state. Foundation models and task reasoning connect observations with safety objectives. Planning identifies feasible interventions. Control and actuation realize those interventions under physical constraints. Feedback determines whether the intended safety improvement has occurred. These functions may be implemented through separate modules or integrated learned policies, but their responsibilities remain distinguishable. The relevant alignment problem is to maintain consistency among human intent, agent decisions, physical feasibility, and environmental safety requirements throughout execution.

\paragraph{Environmental Perception and Hazard Identification}
Environmental protection begins with identifying physical states that may lead to unacceptable consequences. Relevant observations include geometry, object stability, contact relationships, human proximity, equipment motion, temperature, pressure, vibration, and other process-specific variables. A single visual observation may reveal an obstruction but provide insufficient evidence about its mass or stability. Likewise, an abnormal equipment image may not establish whether a mechanical failure is developing. Multimodal sensing and temporal state estimation can provide complementary evidence, while uncertainty estimates help determine whether additional observations are required.

AnomalyGPT~\citep{gu2024anomalygpt} illustrates how vision-language models can support industrial anomaly detection and localization. Such methods provide a possible perceptual foundation for environmental safety agents, particularly when hazardous conditions have recognizable visual manifestations. However, anomaly recognition is not equivalent to hazard assessment. The system must relate the detected condition to physical consequences, operating requirements, and the potential development of the environment. A physical agent should therefore distinguish observed abnormalities from evidence-supported risks and unresolved conditions that require further inspection.

\paragraph{Foundation-Model Reasoning and Safety Objectives}
After identifying a potential hazard, the agent must determine what response is appropriate. This requires connecting physical observations with task objectives, human intentions, operating permissions, and safety priorities. A household agent may detect a spill near electrical equipment, while an industrial agent may observe that a person has entered a machinery workspace. In both cases, the desired response depends on context rather than object recognition alone. Possible responses include warning nearby people, restricting access, delaying the current task, requesting assistance, or initiating an authorized physical intervention.

Foundation models can assist by interpreting safety-related observations and decomposing high-level objectives into executable steps. SayCan~\citep{ichter2023saycan} illustrates a mechanism for combining language-based action selection with estimates of robotic skill feasibility. For environmental protection, a related approach can associate safety-oriented skills with explicit preconditions, permitted interactions, and expected outcomes. An agent may identify that an obstruction should be removed while determining that its available manipulation skills are insufficient to move the object safely. In such cases, requesting assistance can be more appropriate than executing a semantically plausible but physically unsupported action.

Human intent and environmental safety requirements must also remain aligned. An instruction to transport an object does not authorize passing dangerously close to a person, exceeding contact-force limits, or disturbing hazardous equipment. Conversely, detecting a hazard does not automatically grant the agent permission to perform unrestricted intervention. A practical implementation can distinguish the requested objective, the agent's authorized actions, and mandatory environmental constraints. These elements define the permissible scope within which autonomous decisions can be made.

\paragraph{Safety-Constrained Planning}
Planning converts safety objectives into physically realizable actions. Candidate interventions should be evaluated for reachability, collision, stability, contact conditions, workspace restrictions, and possible effects on surrounding objects or people. VoxPoser~\citep{huang2023voxposer} illustrates how language-level constraints can be grounded in three-dimensional spatial representations to support manipulation planning. Such representations provide one possible mechanism for translating environmental safety requirements into motion-related objectives, although their use does not independently guarantee safe execution. Safe reinforcement learning complements planning by incorporating safety requirements into policy optimization. Constrained Policy Optimization optimizes expected return under bounds on expected discounted cumulative costs~\citep{achiam2017constrained}, with a multi-agent extension for multi-robot control~\citep{gu2023safe}.

Environmental protection requires considering both intervention and non-intervention. Moving an unstable container may reduce one hazard while increasing the probability of spillage. Similarly, an agent approaching malfunctioning machinery may expose itself or others to additional risk. A safety-aware planner should therefore compare alternative responses according to their expected consequences and uncertainty. Observation, warning, isolation, or human assistance may be preferable to direct manipulation when the available evidence is insufficient.

The admissibility of an intervention can be expressed through constraints on the predicted physical trajectory. Let $b_t$ denote the agent's belief about the joint state, $\mathbf{a}_{t:t+H-1}$ a candidate action sequence, and $C_{i}^{\mathrm{safe}}$ a task-dependent safety constraint. Let $X_{t:t+H}$ denote the predicted random state trajectory under the belief and candidate actions, and let $N_{\mathrm{constr}}$ be the number of safety constraints. One possible joint chance-constrained formulation is
\begin{equation}
\Pr\left(
\bigcap_{i=1}^{N_{\mathrm{constr}}}
\left\{
C_{i}^{\mathrm{safe}}(X_{t:t+H},\mathbf{a}_{t:t+H-1})\leq 0
\right\}
\,\middle|\,
b_t,\mathbf{a}_{t:t+H-1}
\right)
\geq 1-\varepsilon_{\mathrm{risk}},
\end{equation}
where $H$ is the prediction horizon and $\varepsilon_{\mathrm{risk}}$ bounds the probability that at least one constraint is violated. This formulation is conditional on the validity of state estimation, dynamics prediction, and uncertainty characterization. It provides a possible planning criterion rather than a universal safety guarantee.

\paragraph{Control and Physical Execution}
Even an admissible plan may become unsafe when environmental conditions change during execution. A person may enter the workspace, an object may slip, or the contact dynamics may differ from those assumed during planning. Low-level control must therefore monitor the actual physical state and adjust commands when motion, force, or proximity approaches a safety boundary. Control barrier functions provide one established mechanism for enforcing such constraints. \citet{AmesEtAl2017} formulate safety-critical control through barrier functions and quadratic programming, allowing nominal commands to be modified when necessary to preserve a specified safe set. Their guarantees depend on the assumptions concerning system dynamics, state information, and available control authority.

The actuator and hardware layer supplies further physical protection through joint limits, torque limits, compliant mechanisms, braking systems, and protective stops. These mechanisms bound the physical consequences of commands that pass through higher-level decision-making. They also determine which environmental hazards the agent can address. A robot may recognize a high-temperature or heavy object without possessing the mechanical protection, payload capacity, or sensing required for safe manipulation. The ability to identify a hazard should therefore not be confused with the capability to remove it.

\paragraph{Cross-Layer Action Alignment}
Environmental safety requires these layers to operate under consistent conditions. One possible mechanism is an action contract specifying the intended intervention, required preconditions, authorized contacts, physical limits, termination criteria, and evidence needed to verify the outcome. Foundation-model reasoning identifies the safety objective, planning establishes a feasible intervention, control constrains its execution, and feedback determines whether the relevant hazard has been reduced.

This organization also clarifies why task completion alone does not establish safe behavior. An agent may remove an obstruction while damaging nearby equipment or exposing a person to another hazard. Conversely, an agent that declines direct intervention because the available evidence is insufficient may be acting appropriately. The objective is not unrestricted autonomous intervention, but the reliable realization of authorized safety objectives under physical and operational constraints.

\subsubsection{Predictive Safety and Degradation}

Conventional safety mechanisms often respond after a hazardous condition has been detected, for example when a collision occurs, a contact-force threshold is exceeded, or an emergency stop is activated. Such mechanisms remain necessary, but their effectiveness may be limited when physical consequences develop faster than corrective action can be applied. A more capable physical agent should anticipate risks before execution, monitor whether its assumptions remain valid, and modify its behavior while safer alternatives are still available. This progression applies both to preventing unsafe agent behavior and to protecting the surrounding environment from developing hazards.

\paragraph{Risk Prediction}
Predictive safety begins with estimating how the physical state may evolve under candidate actions or existing environmental conditions. Geometric checks, dynamic models, contact prediction, and learned representations can assess collision, instability, excessive force, or inadequate stopping distance. The prediction horizon should reflect the relevant physical process rather than only the next control step. A grasp may remain stable during initial lifting but become unsafe during subsequent acceleration, while a navigation trajectory may become hazardous as nearby agents move. Evaluating these possibilities before execution allows the physical agent to compare actions according to their expected consequences.

For environmental protection, prediction also considers how a hazard may develop without intervention. An inspection agent may observe increasing equipment temperature or vibration, while a mobile agent may estimate whether moving objects are approaching a dangerous configuration. The agent can then compare continued observation, warning, physical intervention, and human escalation. Prediction should include uncertainty because physical environments are partially observable and their dynamics are not always known. When uncertainty can be reduced through additional sensing, the agent may change its viewpoint or acquire another measurement. When further observation is insufficient or delay is dangerous, conservative constraints or restricted operation may be more appropriate.

\paragraph{Self-Checking and Execution Verification}
Prediction alone cannot establish safety because the predictor may itself be incorrect or compromised. A physical agent should therefore verify critical conditions before, during, and after execution. Before acting, it checks task preconditions, relevant permissions, and the current physical state. During execution, it compares observed motion, contact, and environmental changes with expected behavior. After execution, it determines whether the intended outcome occurred and whether unintended effects remain. Inner Monologue~\citep{huang2023inner} demonstrates how environmental and task feedback can inform the revision of language-based robotic plans. This provides one possible route toward closed-loop decision revision, although safety-critical conclusions still require reliable physical evidence.

Self-checking should use evidence that is sufficiently independent of the mechanism generating the action. A policy that proposes a trajectory and then evaluates its own safety using the same potentially incorrect representation provides limited protection against shared errors. Independent geometric checks, force or tactile measurements, redundant state estimates, and separately implemented monitors can expose discrepancies. For security-sensitive agents, these mechanisms can also help detect inconsistencies between authorized instructions, received observations, and executed commands.

A practical runtime process can therefore follow a predict-execute-verify cycle. The agent evaluates a candidate action, executes it while monitoring relevant constraints, and verifies the resulting state before proceeding. For environmental protection, the corresponding process extends to an observe-predict-intervene-verify sequence that identifies a potential hazard, estimates its development, selects a justified intervention, and confirms that the hazard has been reduced. If observations differ substantially from prediction, the agent should revise its plan rather than continue an obsolete action sequence.

\paragraph{Safe Degradation and Fallback Control}
When a physical agent cannot maintain its nominal operating conditions, it should reduce its capabilities in a controlled manner instead of continuing unrestricted operation. Possible degraded modes include reducing speed, restricting workspace access, disabling contact-intensive skills, increasing human supervision, or transferring control to a simpler policy. SOTER~\citep{desai2019soter} illustrates a runtime-assurance architecture combining an advanced controller with a safety-oriented fallback controller, with switching governed by specified conditions. Such mechanisms provide a possible implementation of controlled degradation, but their guarantees remain conditional on the validity of the safety specifications and controller assumptions.

Degradation must also apply to agents performing environmental safety functions. An inspection agent with unreliable localization may continue collecting local measurements while suspending autonomous navigation near hazardous equipment. An agent that detects a dangerous condition but cannot determine a safe manipulation strategy may issue a warning or request assistance instead of intervening directly. Likewise, when the integrity of critical observations or commands cannot be established, the agent should restrict actions that depend on the affected components. The available operating modes should reflect the information and control capabilities that remain trustworthy.

Importantly, entering a degraded mode must not create a new physical hazard. Immediate stopping may be inappropriate when a manipulator is carrying a heavy load or a mobile system is maintaining dynamic stability. The agent may first need to secure the object, reduce stored energy, or reach a stable configuration. Similarly, an intervention intended to protect the environment should not be abandoned in a state that increases the original risk. Returning to normal operation should require evidence that the conditions responsible for degradation have been resolved.

Experience from near misses, rejected actions, unexpected environmental changes, and unsuccessful interventions can provide evidence about declining safety margins. These observations may reveal weaknesses that are difficult to identify through ordinary task-success measurements. They can inform improvements to risk prediction, sensing, action constraints, and regression testing. The objective is not to assume that a physical agent can prevent every failure, but to recognize emerging hazards before they produce unacceptable consequences, preserve appropriate control under uncertainty, and reduce autonomous capability when safe continuation cannot be established.

\subsection{System Safety and Risk Governance}
\label{pai114:sec:system-safety}

Action alignment for a highly intelligent physical agent depends on an integrated physical system whose sensing, computation, communication, control, actuation, and power remain adequate for the task. Even correct AI-model decisions can be undermined by delayed commands, unreliable observations, degraded actuators, or interacting failures across these layers, compromising the constraints developed in Section~\ref{sec:ch11-action-safety}. Conversely, an incorrect model output need not cause physical harm if independent safeguards can constrain or prevent unsafe execution. Safety assurance must therefore examine how the complete agent detects faults, contains their propagation, and preserves an appropriate physical state. It must also account for changes in hardware condition, software, task permissions, and operating environments, because evidence from initial testing may no longer describe the deployed agent.

\begin{figure}[!htb]
    \centering
    \includegraphics[width=\linewidth,height=0.34\textheight,keepaspectratio]{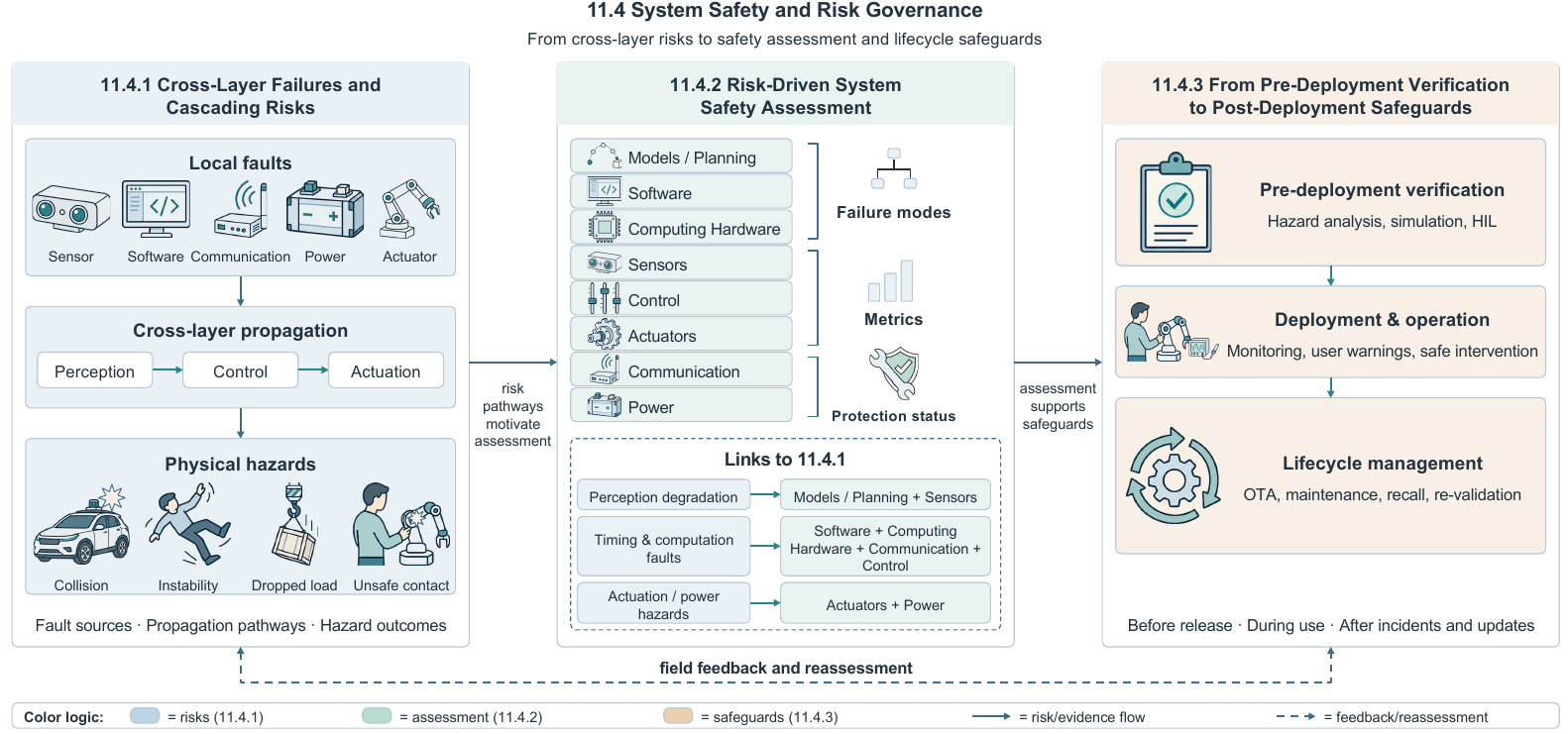}

    \caption{Risk-driven framework for system safety and risk governance.
    (a) Faults in sensing, software, communication, power, and actuation
    can propagate through perception, control, and execution into physical hazards.
    (b) The corresponding safety assessment maps perception degradation to
    sensor-validity and estimator checks. It maps timing and computation faults to
    latency, jitter, and command-age tests. It maps actuation or power hazards
    to current, thermal, and power-loss verification.
    (c) Safety evidence supports pre-deployment verification and
    post-deployment monitoring, user warnings, protective intervention,
    updates, maintenance, and reassessment.
    Solid arrows indicate the progression from identified risks to
    assessment and safeguards. The dashed arrow denotes field feedback}
    \label{pai114:fig:safety-framework}
\end{figure}

As illustrated in Figure~\ref{pai114:fig:safety-framework}, system safety and risk governance follow the three connected stages of cross-layer risk analysis, risk-driven assessment, and lifecycle safeguards. Cross-layer failure analysis first investigates how local faults propagate into hazardous physical outcomes and how independent safeguards can interrupt this process. These failure pathways then motivate system-level safety assessment, which examines observable degradation, evaluation indicators, and protective responses across the hardware and software layers of a physical agent. Beyond these immediate risks, lifecycle risk governance connects hazard analysis and pre-release verification with deployment, maintenance, and operational reassessment. Continuous safety assurance is integrated into this lifecycle perspective through runtime monitoring, protective intervention, user-facing fault communication, field feedback, controlled software updates, and physical remediation. As physical agents move toward increasingly autonomous operation in homes, factories, and public environments, these mechanisms shift safety evaluation from a one-time test result toward a continuously maintained body of evidence throughout the product life cycle. They also establish the technical records and responsibilities needed for the broader legal, ethical, and institutional governance discussed in Section~\ref{sec:ch11-governance}.

\subsubsection{Cross-Layer Failures and Cascading Risks}
\label{pai114:subsec:cascades}

\paragraph*{Perception and Timing Cascades}
A local fault becomes a physical hazard when it propagates through the perception-planning-control-actuation chain. A mobile robot moving from a dim room into strong illumination, for example, may receive temporarily saturated camera frames while automatic exposure adjusts. Industrial-camera documentation confirms that exposure changes may take effect only after subsequent frames have been acquired~\citep{baslerExposureAuto}. Benchmarks of visual depth estimation further document sensitivity to lighting, sensor, and data-processing corruptions~\citep{kong2023robodepth}. If these observations are treated as reliable, obstacle detection may fail and a seemingly coherent navigation decision may increase collision risk. A different cascade begins in real-time scheduling. As a hypothetical example, a controller nominally updated every 10\,ms may experience a delay of 100\,ms because of thread contention or blocking. Meanwhile, momentum, gravity, and contact continue to change the robot state. Research on control timing explains why execution delays must be analyzed with closed-loop dynamics rather than average processing speed alone~\citep{cervin2003timing}.

\paragraph*{Shared Infrastructure and Physical Actuation}
Processor overheating may activate thermal throttling, protecting a chip while reducing the computation available for time-critical decisions~\citep{nvidiaJetsonThermal}. Communication timeouts and inconsistent timestamps can make feedback stale. The need to detect communication anomalies is also well established in networked control systems~\citep{baglietto2020packet}. Undervoltage or power interruption may simultaneously disable multiple subsystems, making the robot's power-loss behavior a distinct safety-verification concern. At the joint level, an unsafe position target is not identical to a motor current. A controller translates tracking error into drive demand, which a motor driver regulates under the actual mechanical load. If a joint is obstructed, persistent error may sustain high demand. Without suitable protection, overheating, transmission damage, or loss of support may follow. Existing drivers demonstrate current regulation, overcurrent protection, and thermal shutdown~\citep{tiDRV8874}, while surveys of robotic fault-tolerant control document the broader need to detect and accommodate actuator faults~\citep{milecki2023faulttolerant}. These functions must be verified for the specific actuator rather than assumed universally.

\paragraph*{Containment Under Degraded Conditions}
The appropriate response depends on the remaining physical capabilities. Command-age checks, watchdogs, current and torque limits, thermal protections, and independent safety supervision can interrupt some cascades, but a generic emergency shutdown may create another hazard. Abrupt removal of joint power can destabilize a balancing humanoid or cause a load-bearing manipulator to drop its payload. Safe intervention may instead require controlled deceleration, support retention, a designed braking mechanism, or an energy-independent mechanical fallback. The appropriate stopping function and residual hazards must be specified for the application, rather than inferred from the existence of an emergency-stop device~\citep{iso13850_2015,bi2021assurance}. Loss-of-power behavior must be tested before deployment. Fault analysis should therefore record the initiating event, propagation path, physical consequence, and intended protective barrier. System-theoretic accident analysis complements component-oriented methods by identifying hazardous interactions that need not begin with a defective part~\citep{leveson2004accident}. The resulting failure pathways provide the basis for the risk-driven safety assessment that follows. Each path must be associated with an observable precursor, a protective requirement, and a physically meaningful verification outcome.

\subsubsection{Risk-Driven System Safety Assessment}
\label{pai114:subsec:assessment}

\paragraph*{From Failure Pathways to Safety Evidence}
Task-success rates, perception accuracy, and tracking errors quantify useful capabilities but do not establish what happens after corrupted observations, missed deadlines, or degraded actuators. A defensible safety claim concerns the complete physical agent, its authorized tasks, the exposure of people and equipment, and the conditions under which protective functions remain effective. It should distinguish an action proposed by the model, the command actually admitted and applied, and the resulting physical state. Surveys of safety-critical robotics and safeguarding of autonomous systems identify the need to state assumptions, adopt suitable safety mechanisms, and examine evidence across heterogeneous application domains~\citep{guiochet2017survey,mueller2022safeguarding}. Research on testing standards for integrated robots likewise identifies limitations of component-centered performance tests when systems combine multiple functional modules~\citep{bi2020testing}. Assessment should consequently be derived from the fault pathways described above, with explicit acceptance criteria for both component degradation and end-to-end physical outcomes.

\paragraph*{Observable Indicators Across System Layers}
Perception cascades motivate tests of image saturation, dropped measurements, timestamp consistency, estimator residuals, and behavior when observations are unreliable. Robust-perception benchmarks explicitly test lighting, sensor, and processing corruptions, although they do not by themselves establish the safety of the complete robot~\citep{kong2023robodepth}. Fault-tolerant robotics research also addresses failures in sensor information and state estimation~\citep{milecki2023faulttolerant}. Timing cascades require tail latency, deadline-miss rate, control jitter, and command freshness. Real-time control research relates timing to closed-loop performance, while ROS~2 guidance identifies blocking, memory allocation, and priority inversion as sources of execution-time nondeterminism~\citep{cervin2003timing,ros22026realtime}. Processor and supply degradation motivate thermal, clock-frequency, voltage, and interruption tests. Hardware thermal-management documentation establishes why computation may degrade under overheating~\citep{nvidiaJetsonThermal}. Communication assessment should measure latency, loss, and timeout behavior, drawing on networked-control fault-detection research~\citep{baglietto2020packet}. Actuator assessment must examine current, temperature, joint and torque limits, stall detection, and the resulting physical response. Existing driver protections and robotic fault-tolerance studies provide corresponding engineering foundations~\citep{tiDRV8874,milecki2023faulttolerant}. Table~\ref{pai114:tab:failure-modes} synthesizes these candidate measurements and associated protective responses. Their thresholds and applicability should follow the robot's risk assessment. Even published safety-related test guidance for personal care robots does not treat every test as universal~\citep{isoTR23482_2020,mueller2022safeguarding}.

\begin{table}[!htb]
\centering
\caption{
Representative failure modes, safety consequences, and
evaluation dimensions across robotic system layers.
Indicators are illustrative and platform-dependent}
\label{pai114:tab:failure-modes}
\footnotesize
\setlength{\tabcolsep}{3.5pt}
\renewcommand{\arraystretch}{1.13}

\begin{tabularx}{\linewidth}{
@{}
>{\raggedright\arraybackslash}p{0.158\linewidth}
>{\raggedright\arraybackslash}p{0.246\linewidth}
>{\raggedright\arraybackslash}p{0.242\linewidth}
>{\raggedright\arraybackslash}X
@{}}
\toprule
\textbf{System layer} &
\textbf{Representative failure modes} &
\textbf{Potential safety consequences} &
\textbf{Monitoring / evaluation dimensions} \\
\midrule

\textbf{Models and planning} &
Unsafe decisions\newline
Infeasible joint targets &
Hazardous motions\newline
Unintended contact &
Constraint violations\newline
Safety-filter interventions \\
\addlinespace[3pt]

\textbf{Real-time software} &
Missed deadlines\newline
Scheduling blocks, stale commands &
Delayed corrective action\newline
Loss of stability &
Deadline-miss rate\newline
Control jitter, command age \\
\addlinespace[3pt]

\textbf{Computing hardware} &
Thermal throttling\newline
Processor faults &
Increased control latency\newline
Interrupted execution &
Chip temperature\newline
Clock rate, execution latency \\
\addlinespace[3pt]

\textbf{Sensors} &
Image saturation\newline
IMU drift, frame loss &
Missed obstacles\newline
Incorrect measurements &
Saturated-pixel ratio\newline
Bias, frame-drop rate \\
\addlinespace[3pt]

\textbf{Estimation and control} &
Estimator divergence\newline
Feedback delay, oscillation &
Unstable motion\newline
Excessive contact &
Estimation residuals\newline
Feedback age, oscillation amplitude \\
\addlinespace[3pt]

\textbf{Actuators and transmissions} &
Motor stall, overcurrent\newline
Overheating, transmission wear &
Joint overload or damage\newline
Loss of support &
Motor current, temperature\newline
Torque-limit events, tracking error \\
\addlinespace[3pt]

\textbf{Communication} &
Packet loss, bus timeout\newline
Timestamp mismatch &
Stale feedback\newline
Uncoordinated execution &
Packet-loss rate, bus errors\newline
Link latency, timestamp skew \\
\addlinespace[3pt]

\textbf{Power and battery} &
Undervoltage\newline
Power interruption, aging &
Unexpected shutdown\newline
Fall or dropped load &
Bus voltage, battery health\newline
Power-loss test outcome \\

\bottomrule
\end{tabularx}
\end{table}

\paragraph*{End-to-End Protection Verification}
Measurements become safety evidence only when they are tied to the physical response of the integrated system. Fault injection should introduce exposure faults, delayed commands, packet loss, thermal or computational degradation, actuator restrictions, and loss of power under representative tasks. Systematic reviews of robotic testing, validation, and verification, together with research on integrated-robot testing, provide a foundation for selecting complementary test methods and reporting their coverage~\citep{araujo2023testing,bi2020testing}. Evaluation should measure detection and intervention latency, the ability to isolate a failing component, and whether the robot can enter a state whose residual risk is acceptable. Work on interpretable runtime monitoring demonstrates how predicted unsafe states can trigger corrective replanning, while safety-critical control research develops explicit constraints on admissible system behavior~\citep{difranco2020monitoring,AmesEtAl2017}. Safety-related control design supplies further principles for protective functions whose effectiveness must be validated independently of the nominal policy~\citep{iso13849_2023,bi2021assurance}. The goal is a traceable relationship between an identified hazard, a tested safety requirement, and the observed outcome, not a context-free aggregate safety score.

\subsubsection{From Pre-Deployment Verification to Post-Deployment Safeguards}
\label{pai114:subsec:lifecycle}
\label{pai114:subsec:continuous} 

\paragraph*{Safety-by-Design Before Delivery}
Risk governance starts before a product is placed into use. Developers should identify foreseeable operating environments, human proximity, loads, task permissions, misuse, and failure combinations, then translate them into safety requirements and independent protection functions. Recent robotics safety reviews emphasize risk assessment across both design and deployment~\citep{adesiji2026review}. Simulation, hardware-in-the-loop testing, cross-module fault injection, environmental stress testing, and controlled physical trials should verify both nominal performance and behavior under degraded conditions, using complementary techniques whose limitations have been surveyed in the robotic testing literature~\citep{araujo2023testing}. The corresponding evidence must record hardware and software versions, model or policy configuration, operating limits, residual risk, and what was not tested. ISO~12100 and ISO~13849-1 provide established foundations for machinery risk assessment and safety-related control design. ISO~13482 and ISO~10218-1 illustrate additional requirements for particular robot applications, not universal certification rules for every physical agent~\citep{iso12100_2010,iso13849_2023,iso13482_2014,iso10218_2025}.

\paragraph*{Release as an Evidence-Based Transition}
Acceptance testing should validate not only whether a robot can complete its assigned tasks but whether its safety monitors, limits, fallback controllers, brakes, and any emergency energy provisions work together when a fault occurs. The need to test the integrated configuration and select risk-dependent test methods is reflected in integrated-robot testing research and application-specific ISO guidance~\citep{bi2020testing,isoTR23482_2020}. A release record should link the tested product configuration to its operating envelope, protective functions, acceptance criteria, and outstanding hazards. Wear, new payloads, changed environments, repaired components, and updated policies may invalidate the assumptions documented in the release record. Post-deployment safeguards must therefore maintain and revise the evidence created before delivery rather than replace it with an unrelated monitoring process~\citep{guiochet2017survey,mueller2022safeguarding}.

\paragraph*{Runtime Detection and Protective Action}
After delivery to the user or operator, the robot should monitor whether observations remain trustworthy, commands are fresh, and control deadlines are met. It should also monitor whether actuator, computing, communication, and power subsystems retain adequate capacity. Where justified by the hazard analysis, an independent safety supervisor should restrict commands or transfer control when the nominal policy is unavailable or wrong. Runtime monitoring and corrective replanning have been demonstrated for autonomous robotic operation, while control-barrier methods illustrate how physical safety constraints can be imposed on control inputs when their modeling assumptions hold~\citep{difranco2020monitoring,AmesEtAl2017}. Predictive checking and measured feedback should support graduated responses, including limiting speed or workspace, disabling contact-rich actions, reducing payload, requesting confirmation, or moving into a validated protective mode. In an emergency, local protection must act within the required response time rather than wait for a user decision. The chosen safe state must respect the actual mechanics. A dynamically balancing robot or one supporting a heavy load may need controlled motion or braking before torque is removed. Protection must also recognize when insufficient power or sensing makes autonomous recovery impossible.

\paragraph*{User-Facing Warnings and Operational Responsibility}
A deployed safety system must communicate consequential faults, not merely store engineering logs. Research on human-robot collaboration identifies operator awareness and effective visual, acoustic, and other notification channels as important interface design concerns~\citep{grushko2021awareness,bi2021assurance}. Operators or household users should receive an intelligible account of the affected capability, the protective action already taken, and the appropriate next step. For example, the account should explain that a joint has overheated, forceful manipulation has been disabled, and servicing is required before that function resumes. Notifications should distinguish informational maintenance advice from an urgent restriction on continued use, without exposing users to raw diagnostics they cannot interpret. In higher-risk settings, a warning may need to be paired with physical isolation, a lockout, or escalation to an authorized operator. Such interaction supports accountable operation but must not substitute for automatic, timely safety intervention.

\paragraph*{Validated Updates and Physical Remediation}
Some defects can be corrected by over-the-air (OTA) software or firmware updates, but changing perception, scheduling, control logic, or model capabilities may invalidate safety assumptions. Future update processes should verify provenance and integrity, assess hardware-software compatibility, test affected safety functions, stage deployment, and provide recovery or rollback where feasible. The software-update-management principles of UN Regulation No.~156 offer an automotive precedent, not a law automatically applicable to robots~\citep{unece2021r156}. Other defects require inspection or replacement. Motors, gearboxes, brakes, sensors, batteries, and power electronics can wear or fail in ways software cannot correct. When a common defect affects a product batch, the manufacturer may need to identify affected configurations, contact users, restrict risky operation, provide repairs, or initiate a recall. Automotive recall practice offers a relevant governance analog.\footnote{NHTSA, \href{https://www.nhtsa.gov/recalls}{\emph{Check for Recalls: Vehicle, Car Seat, Tire, Equipment}}.} OTA maintenance is not a substitute for on-device protection during an unfolding hazard.

\paragraph*{Continuous Improvement Across the Product Life Cycle}
Incident and near-miss reports, safety-filter interventions, maintenance histories, and configuration records should connect field observations to the original hazard analysis and qualification tests. Comparing evidence across deployed robots may reveal emerging combinations of temperature, wear, environmental conditions, and task behavior that were not covered before release. Data collection should remain proportional to the safety purpose, particularly in private environments. Updated risks must result in reviewed corrective decisions, re-validation of affected configurations, and, where necessary, revised operating permissions or withdrawal from service. Linking risk analysis to subsequent tests is consistent with existing safety-assessment and robotic verification research~\citep{adesiji2026review,araujo2023testing}. The objective is not to guarantee the absence of all faults, but to maintain a credible and revisable basis for safe operation as physical agents and the environments in which they work continue to evolve.

\subsection{Law, Ethics, Standards, and Global Governance}
\label{sec:ch11-governance}
The preceding sections have examined how to evaluate the behavior and competence of physical agents, constrain their actions, and maintain safety evidence throughout operation. These technical foundations inform deployment decisions, but they do not by themselves determine who is responsible, which uses of autonomy are acceptable, or what oversight is required. Because Physical AI connects perception and decision-making to actions in the physical world, governance must address the deployed system rather than the model in isolation. Law, ethics, standards, and international coordination connect technical evidence to obligations and acceptable uses, extending the chapter’s focus from measurable behavior and safety assurance to accountable deployment.

This section first addresses responsibility allocation and traceability, then examines human control, privacy, fairness, and ethical boundaries. It next considers legal and regulatory frameworks from technical, ethical, and application perspectives, connecting these issues to safety standards and the book's illustrative risk-tiered governance framework. Finally, it explores global cooperation through compatible evaluation evidence, incident records, and investigation procedures.

\subsubsection{Responsibility Allocation and Traceability}
\label{sec:ch11-responsibility}
Learning-based systems may behave in ways that designers and operators cannot fully anticipate, creating difficulties associated with the \emph{responsibility gap}~\citep{matthias2004responsibility}. Responsibility consequently extends beyond the person exercising immediate control. Model developers should document capabilities, limitations, and versions. Hardware manufacturers should address mechanical and electrical hazards. Integrators should validate interfaces and operating limits. Operators should maintain safeguards and intervention procedures. Maintenance providers must track safety-relevant updates and configuration changes. These engineering duties support accountability but do not themselves determine legal liability.

This allocation depends on what each party supplies or changes. For example, replacing a model, increasing permitted speed, or changing an end-effector can alter different parts of a safety argument. Documentation should identify who approved the change, which assumptions were reconsidered, and what evidence supports continued operation.

Product-liability law provides a complementary perspective. Directive (EU) 2024/2853 includes software and addresses safety-relevant updates and upgrades. Its rules apply to products placed on the market or put into service after December 9, 2026~\citep{eu2024productliability}. The significance for Physical AI is that assessment cannot stop at initial delivery. Subsequent changes may alter the behavior of the deployed system.

Accountability also depends on reconstructing what occurred. A safety-relevant event record should connect system and model identity with perception states, decisions, executed actions, safety settings, and human interventions, together with update, maintenance, and cybersecurity histories. Recording should serve investigation and safety improvement rather than unrestricted surveillance. Data minimization, access controls, encryption, and proportionate retention are therefore part of traceability design, not exceptions to it.

\subsubsection{Human Control, Privacy, Fairness, and Ethical Boundaries}
\label{sec:ch11-human-control}
Meaningful human control requires more than the presence of a supervisor. A person must understand relevant capabilities and limitations, receive usable information, possess intervention authority, and have sufficient time to act. An override that cannot bring the system into a safe state within the available response time offers little practical control. Human oversight and accountability are also central to the UNESCO and OECD AI ethics frameworks~\citep{unesco2021ethics,oecd2019ai}.

The appropriate form of oversight depends on the task. A supervisor may authorize a sequence in advance, monitor execution, or intervene when an exception occurs. These roles require different information and response times. Where physical dynamics are faster than human reaction, protective functions must constrain the system without waiting for an operator to diagnose the problem.

Physical sensing extends privacy concerns beyond direct users. A domestic robot may observe visitors or neighbors who have neither purchased nor authorized it. \emph{Privacy by Embodiment}, used here as a design principle, connects privacy protection to sensing and actuation. Under this principle, systems should process data locally where feasible, minimize raw-data retention, protect sensitive information, and communicate active sensing and remote access. User authorization does not by itself resolve bystander exposure.

Fairness likewise concerns physical outcomes, not only balanced prediction accuracy. Differences in mobility, body characteristics, or speech may change whether users can interact safely or obtain timely assistance. Evaluation should therefore examine accessibility and failure consequences across intended users, particularly children, older people, and people requiring care. Greater autonomy does not remove the need for human judgment where actions threaten safety, dignity, or essential services. Physical AI may also displace workers, while newly created jobs may not benefit those affected. Fair deployment should therefore consider worker participation, retraining opportunities, and the equitable distribution of productivity gains~\citep{unesco2021ethics,oecd2019ai}.

\subsubsection{Legal and Regulatory Frameworks for Physical AI}
\label{sec:ch11-regulation}

\paragraph*{Technique}
Technical regulation connects responsibility allocation, change management, and traceability to legal accountability. The EU AI Act's high-risk framework addresses risk management, technical documentation, event logging, and cybersecurity, subject to its scope and application dates~\citep{eu2024aiact,eu2026aiactconsolidated}. Directive (EU) 2024/2853 complements this framework by covering software and addressing safety-relevant updates in product liability~\citep{eu2024productliability}. As discussed in Section~\ref{sec:ch11-responsibility}, model replacements or changes in speed and tooling should prompt reassessment, with records identifying who approved each change and what evidence supports continued operation. Event records should connect deployed versions, actions, interventions, and maintenance histories. These engineering duties support accountability but do not themselves establish legal liability.

\paragraph*{Ethics}
Meaningful human control, Privacy by Embodiment, and fairness require safeguards for people affected by deployment. The EU AI Act's high-risk framework addresses effective oversight, while practical intervention must remain feasible within the available response time~\citep{eu2024aiact,eu2026aiactconsolidated}. The GDPR addresses lawful processing, data minimization, and protection by design. China's Personal Information Protection Law addresses sensitive information and fairness in automated decision-making~\citep{eu2016gdpr,china2021pipl}. Article~5 of the AI Act prohibits specified manipulative practices and exploitation of vulnerabilities where its statutory conditions are met~\citep{eu2024aiact,eu2026aiactconsolidated}. Legal compliance does not exhaust the ethical responsibilities identified in Section~\ref{sec:ch11-human-control}, including equitable access to safe assistance and support for workers affected by automation~\citep{unesco2021ethics,oecd2019ai}.

\paragraph*{Application}
Application-level regulation depends on intended use, system boundaries, and jurisdiction. Industrial robot installations fall under applicable machinery rules, while qualifying medical robots are subject to medical-device regulation~\citep{eu2006machinery,eu2023machinery,eu2017medicaldevices}. The EU AI Act separately classifies high-risk systems through Article~6 and Annex~III, subject to their statutory conditions~\citep{eu2024aiact,eu2026aiactconsolidated}. Table~\ref{tab:risk-tier} presents the book's illustrative five-tier framework, considering harm severity, human exposure, tooling, autonomy, safeguard reliability, and user vulnerability~\citep{iso12100_2010,pi12_tabassi2023}. These tiers are neither legal categories nor an international standard and do not map directly onto the AI Act. Classification must record operating assumptions and be reassessed when conditions change. A guarded robot's Low placement reflects controlled exposure, not harmless machinery. Lower tiers do not remove legal duties, and higher tiers cannot authorize prohibited uses.

\begin{table}[!htb]
\centering
\caption{Illustrative risk tiers proposed in this monograph. Placements depend on task, exposure, and validated safeguards}
\label{tab:risk-tier}
\renewcommand{\arraystretch}{1.12}
\setlength{\tabcolsep}{5pt}
\begin{tabular}{@{}
>{\raggedright\arraybackslash}p{0.13\linewidth}
>{\raggedright\arraybackslash}p{0.51\linewidth}
>{\raggedright\arraybackslash}p{\dimexpr0.36\linewidth-4\tabcolsep\relax}@{}}
\toprule
\textbf{Tier} & \textbf{Illustrative conditions and harm} & \textbf{Governance focus} \\
\midrule
Minimal &
Supervised, low-energy desktop sorting without sharp, hot, or heavy tooling. Contact harm is at most trivial and is reversible. &
Basic product safety, use limits, instructions, and controlled updates. \\[2pt]
Low &
Industrial operation behind validated guarding and interlocks, without routine human access during automatic operation. Exposure is mainly during setup or maintenance, but contact remains potentially severe. &
Machinery assessment, protective-function validation, access control, and safe maintenance with lockout procedures. \\[2pt]
Moderate &
Speed-limited goods transport in worker-accessible areas, repeated encounters with reversible injury or localized damage foreseeable. &
Route and speed limits, detection and stopping tests, monitoring, and incident analysis. \\[2pt]
High &
Close-contact collaboration, hazardous household manipulation, or elder-care transfer assistance. Human exposure is frequent, and crushing, burns, cuts, falls, or incorrect assistance are foreseeable. &
Independent evaluation, meaningful oversight, runtime safeguards, incident reporting, and privacy protection. \\[2pt]
Very high &
Traffic-speed autonomous driving, invasive surgical motion, or critical-infrastructure control. Death, multiple casualties, or cascading disruption are possible. &
Rigorous assurance, independent review, continued supervision, appropriate fail-safe or fail-operational design, and sector-specific compliance. \\
\bottomrule
\end{tabular}
\end{table}

\subsubsection{Global Cooperation and Governance}

Models, components, integrators, and operators may span jurisdictions, while responsibility for an incident must remain traceable across their interfaces. Building on the obligations and safeguards discussed in Section~\ref{sec:ch11-regulation}, international cooperation should seek compatible safety evidence and investigation procedures rather than identical legal systems. The practical objective is to preserve the responsibility chain described in Section~\ref{sec:ch11-responsibility} when development, integration, deployment, and maintenance take place in different countries.

The UNESCO and OECD frameworks provide shared ethical principles, while the Council of Europe AI Framework Convention connects AI governance with human rights, democracy, and the rule of law~\citep{unesco2021ethics,oecd2019ai,coe2024aiconvention}. The 2023 Bletchley and 2024 Seoul declarations emphasize safety cooperation alongside broader development goals~\citep{bletchley2023,seoul2024}. China's 2024 Shanghai Declaration and 2025 Global AI Governance Action Plan likewise advocate safe, inclusive development and international cooperation~\citep{shanghai2024,china2025actionplan}. These declarations and policy initiatives are not binding international safety regulations. Their value for Physical AI lies in supporting cooperation on the shared concerns of human control, privacy, fairness, and accountability identified in Section~\ref{sec:ch11-human-control}.

Practical cooperation would include common incident categories, interoperable event records, reliable model and system identification, compatible evaluation methods, cross-border safety notifications, and coordinated corrective updates. Evaluation evidence should preserve the operating assumptions, system configuration, and safeguards under which it was obtained, so that investigators can determine whether it remains relevant in another deployment. Shared investigation procedures should make it possible to connect a reported event with the organizations responsible for the affected model, hardware, integration, and subsequent modifications. These mechanisms would support investigation and improvement across borders without presuming agreement on every regulatory choice or automatic recognition of another jurisdiction's approval.

Interoperability does not require publishing every operational record. Shared formats can describe the system, event, and applicable safeguards while access to personal data, security-sensitive information, and other protected evidence remains controlled. Consistent with Sections~\ref{sec:ch11-responsibility} and~\ref{sec:ch11-human-control}, cooperation should preserve both investigative usefulness and proportionate information protection. The practical test of global governance is whether an incident can be investigated across organizational and national boundaries without losing the chain of responsibility or exposing sensitive information unnecessarily.

\clearpage 

\section{Strategy and Civilizational Outlook}
\label{ch:strategy-outlook}

The preceding chapters have examined the concepts, technical foundations, models, and systems of Physical AI. This final chapter considers how these advances might develop into capabilities that shape the longer-term course of human civilization. The central question is how intelligence grounded in physical interaction can expand humanity's ability to discover, produce, build, and sustain the systems on which it depends.

The chapter first identifies development priorities for the coming decade and beyond, including the convergence of digital and physical foundation models, self-evolving agents, hardware-software co-evolution, an Internet of Physical Agents, and the scientific and technological ecosystem supporting their development. It then examines how local achievements can become shared physical services, allowing verified capabilities to accumulate across teams, locations, and generations. From this foundation, the discussion extends to planetary energy and infrastructure associated with a Type~I civilization, and to the autonomous construction, resource use, and long-duration operation that proposed stellar-scale development requires.

These possibilities depend on progress in both intelligence and its physical foundations. Energy, materials, equipment, and maintenance remain essential constraints. The civilizational significance of Physical AI therefore rests on its ability to turn knowledge into reliable action, preserve useful capabilities over time, and support further discovery under increasingly demanding conditions.

\subsection{Development Strategies for Physical AI}\label{sec:physical-ai-development-strategies}

\begin{figure}[htbp]
    \centering
    \includegraphics[width=\linewidth]{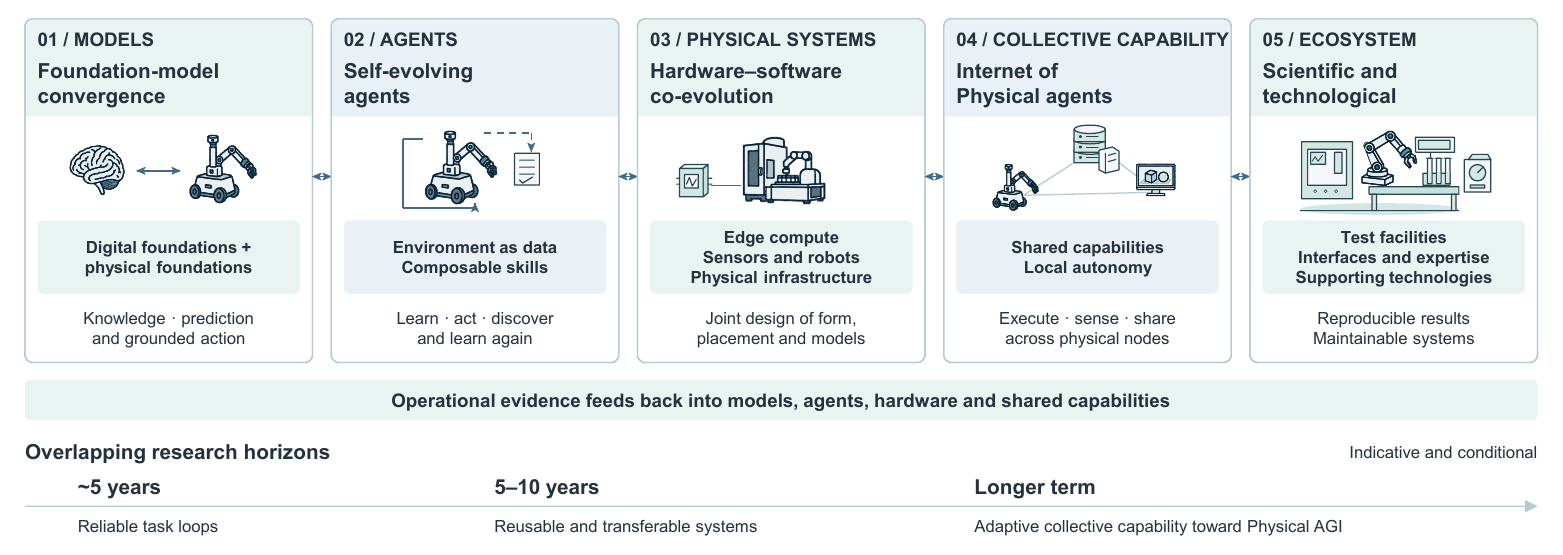}
    \caption{Five development priorities for Physical AI, including
    foundation-model convergence, self-evolving agents,
    hardware-software co-evolution, the Internet of Physical Agents,
    and a supporting scientific and technological ecosystem.
    Operational evidence informs all five priorities.
    The research horizons are indicative and conditional,
    rather than fixed delivery commitments}
    \label{fig:physical-ai-development-strategy}
\end{figure}

The development of Physical AI requires advances in models, learning systems, hardware, and the infrastructure through which capabilities are shared. Over the next five to ten years and beyond, these advances should support a common objective of systems that can undertake increasingly varied physical tasks, learn from their operation, and transfer useful capabilities with less additional engineering. Figure~\ref{fig:physical-ai-development-strategy} summarizes five priorities that follow from this objective, including the convergence of digital and physical foundation models, self-evolving agents, hardware-software co-evolution, an Internet of Physical Agents, and a scientific and technological ecosystem that allows progress to accumulate.

\subsubsection{Converging Digital and Physical Foundation Models}
\label{subsec:physical-ai-model-convergence}

Digital foundation models and physical foundation models should develop through increasingly close integration. Digital foundation models learn broadly reusable representations and capabilities from language, code, images, and other digitally represented information. Physical foundation models seek reusable capabilities grounded in physical states, dynamics, interventions, and interaction. 

The strategic opportunity lies in connecting the breadth of digital knowledge and reasoning with the grounding required for physical action. A system working in the real world must interpret objectives, draw on prior knowledge, recognize physical constraints, anticipate the consequences of its actions, and revise its decisions after observation. Digital and physical capabilities need to inform one another throughout this process. Physical evidence can correct an assumption made during reasoning, while semantic knowledge can help identify which variables to measure or which intervention to attempt.

The integration of large multimodal models and world models is one important example. Multimodal models can connect observations to concepts and task descriptions, while world models can predict how a physical state may change under an intervention. Other combinations include code-generating models coupled with executable control interfaces, reasoning models linked to scientific simulators, and general-purpose planners working with embodied action models. These examples suggest a broader research direction involving foundation-model systems that connect knowledge, prediction, and action across digital and physical domains.

Integration can proceed through explicit interfaces before requiring a unified architecture. Models can exchange task objectives, state estimates, candidate actions, predicted outcomes, and uncertainty. Shared representations and joint training may subsequently reduce inconsistencies and improve transfer. Analytical models, numerical solvers, and simulators should remain available where they provide useful structure or more reliable estimates. The choice between a single model and a coordinated set of models should depend on performance, adaptability, latency, and cost.

The longer-term goal is a foundation-model system that participates actively in acquiring knowledge. It should recognize when available evidence is insufficient, select a useful observation or intervention, and incorporate the result into subsequent decisions. Progress should be assessed through the quality of those decisions, the physical trials required to reach them, and the ability to reuse knowledge under new conditions.

\subsubsection{Self-Evolving Agents for Open-Ended Tasks}
\label{subsec:physical-ai-self-evolving-agents}

Physical AI should support agents that extend their capabilities through experience. Here, \emph{self-evolution} refers to a controlled process of identifying capability gaps, acquiring evidence, and improving memory, skills, models, or policies. Objectives and operating constraints remain explicit, and changes to deployed behavior require evaluation.

This approach treats the \emph{environment as data}. The operating environment is a continuing source of learning opportunities, rather than merely a destination for a pretrained model. An agent's actions influence which observations become available. A failed grasp may reveal an incorrect estimate of friction. An unexpected measurement may expose a missing variable. A successful recovery may suggest a reusable procedure. Such experience becomes useful training evidence only when it is recorded with sufficient context, checked for quality, and interpreted with appropriate attention to uncertainty and alternative explanations.

\begin{figure}[htbp]
    \centering
    \includegraphics[width=\linewidth]{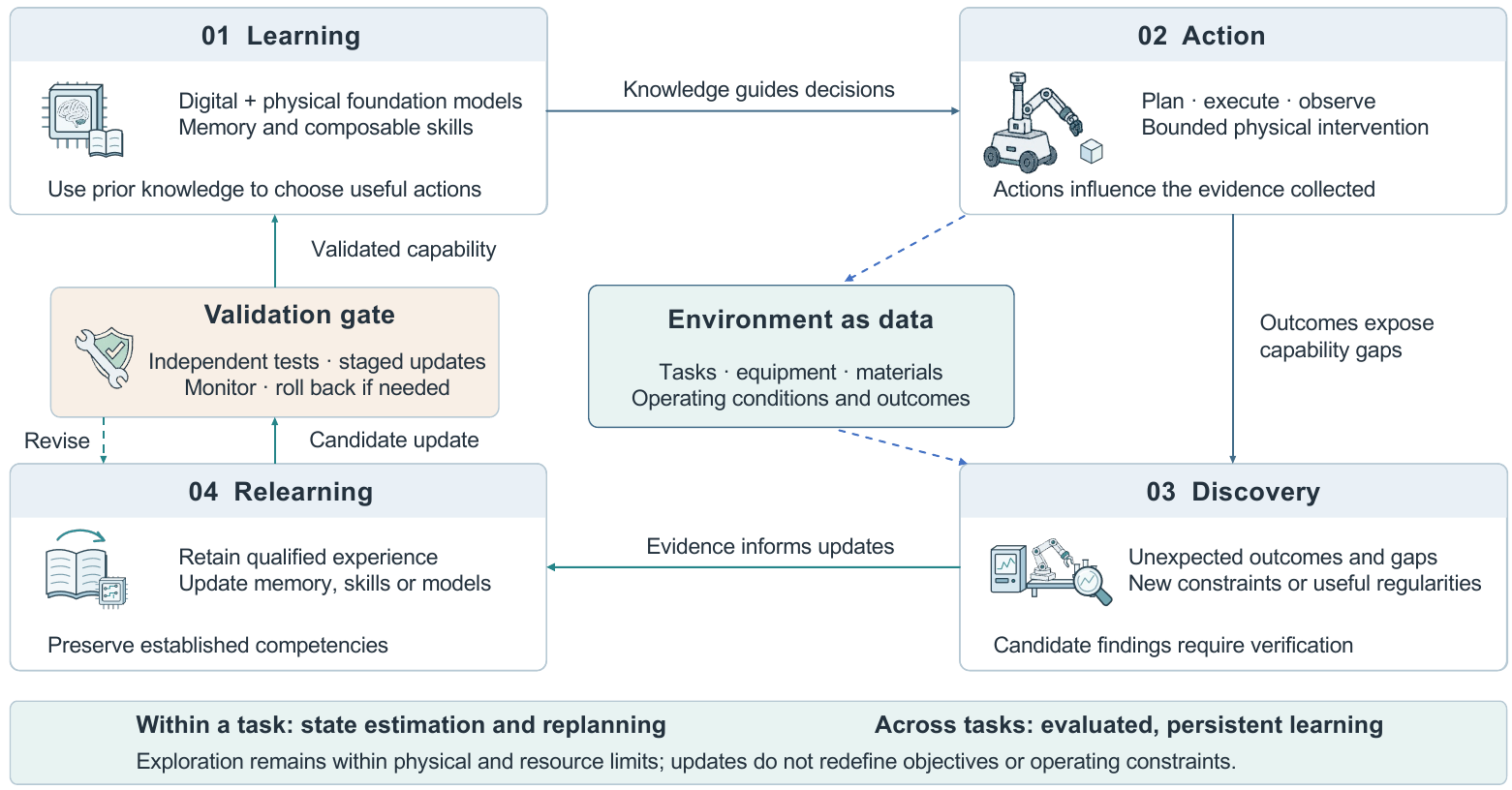}
    \caption{The learning-action-discovery-relearning cycle.
    Physical interaction generates observations that reveal
    capability gaps and candidate discoveries.
    Qualified experience informs updates to memory, skills,
    or models, subject to evaluation and controlled deployment.
    Adaptation within a task is distinguished from persistent
    learning across tasks}
    \label{fig:physical-ai-learning-cycle}
\end{figure}

The resulting cycle (Figure~\ref{fig:physical-ai-learning-cycle}) is \emph{learning-action-discovery-relearning}. Existing knowledge guides action. Action exposes the agent to physical outcomes. Discrepancies and new observations reveal gaps or regularities. Subsequent learning revises the relevant capability. Discovery in this cycle may be modest, such as a changed equipment response, a previously unrecognized task constraint, or a more effective action sequence. The value of the cycle depends on whether these findings improve later performance.

Adaptation during execution should be distinguished from persistent learning. Updating a state estimate, revising a plan, or retrieving another skill may be sufficient to complete the current task. Changes intended to improve future tasks require retained evidence, failure analysis, candidate updates, and evaluation against established competencies. Physical exploration must remain within defined safety and resource limits, and updates should be introduced in stages with a means of rollback.

Composable skills make this process more economical. Capabilities for movement, manipulation, measurement, inspection, and recovery should describe their preconditions, expected effects, and failure modes. Agents can then assemble workflows from existing components and learn the missing parts. Composition still requires testing, since the timing and physical effects of one skill can invalidate the assumptions of another.

A useful measure of progress is the cost of acquiring the next capability. Learning across tasks, environments, and embodiments should reduce the additional demonstrations, trials, engineering changes, and human assistance needed in a new setting. Comparisons with frozen systems should account for these costs, as well as retained performance and the reliability of long-horizon execution.

\subsubsection{Hardware-Software Co-Evolution}
\label{subsec:physical-ai-hardware-software}

Hardware-software co-evolution in Physical AI spans computing resources, physical terminals, and the infrastructure in which they operate. Hardware includes edge processors and communication devices, sensors and actuators, robots and scientific instruments, and installed systems such as cameras and traffic signals. Their design determines what algorithms can observe, how quickly they can respond, and which actions are physically possible.

Edge computing should be developed alongside model architectures and execution requirements. Memory capacity, energy consumption, thermal limits, and latency constrain the models that can run near a device. Model compression, task allocation, and the division of computation between local and remote resources should reflect these constraints. Time-critical control and essential protective functions need dependable local execution when communication is delayed or unavailable, while less urgent reasoning and learning may use shared computing resources.

For robots and other terminals, morphology and function should be considered together with models and control algorithms. Joint arrangement, actuator characteristics, mechanical compliance, sensor placement, and end-effector design affect both task performance and learning difficulty. A hardware change may simplify a control problem or make a previously hidden state observable. Conversely, improvements in perception or control may allow a simpler mechanical design. Co-design should evaluate these trade-offs against task coverage, reliability, energy use, and maintenance requirements.

The same reasoning extends to physical infrastructure. Camera placement can be optimized jointly with perception and state-estimation methods, taking account of coverage, occlusion, calibration, and privacy constraints. The layout and configuration of traffic signals can be studied together with sensing, prediction, and coordination algorithms, subject to engineering and safety requirements. In such settings, the environment's instrumentation and configuration become part of system design. Software performance should be assessed against the infrastructure it requires, including installation and operating costs.

Over approximately five years, priorities should include reliable edge deployment, instrumented platforms, and reproducible integration and testing. Over five to ten years, modular terminals, reusable interfaces, and coordinated infrastructure design could reduce the cost of supporting new tasks. More distant research may investigate reconfigurable bodies, adaptive sensing systems, and brain-computer interfaces that connect human neural signals with computational and physical systems. The latter remains an exploratory direction, with substantial questions concerning signal stability, biocompatibility, privacy, and safe operation. It should not be treated as a prerequisite for near-term progress.

Across these horizons, access to physical testing remains essential. Computing resources need complementary support from equipment, calibration, maintenance, consumables, and technical expertise. Co-evolution is useful when changes in hardware and software jointly improve sustained performance or reduce total resource requirements.

\subsubsection{The Internet of Physical Agents: Architecture and Sustained Operation}
\label{subsec:internet-of-physical-agents}

\begin{figure}[htbp]
    \centering
    \includegraphics[width=\linewidth]{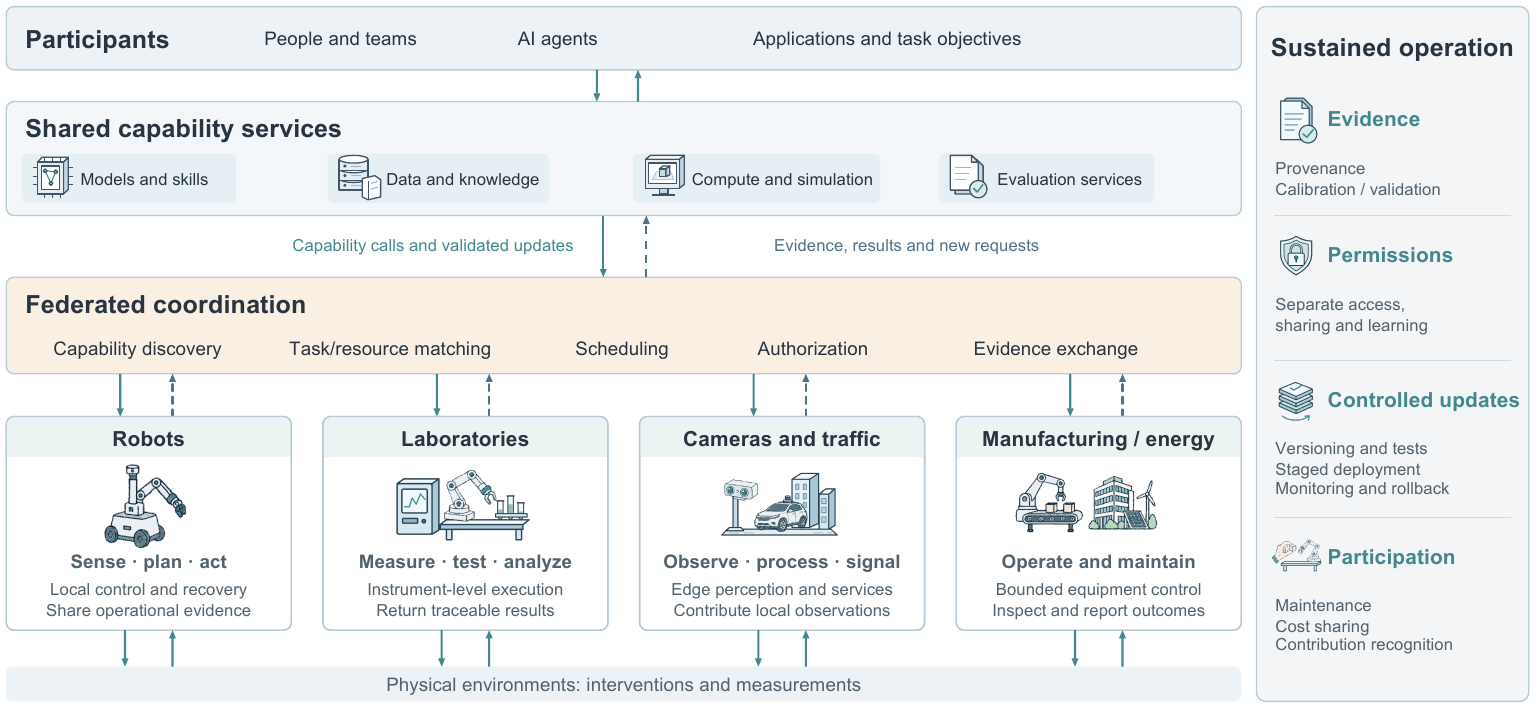}
    \caption{Conceptual architecture of the Internet of Physical Agents.
    Shared capability services, federated coordination,
    and locally autonomous physical nodes support
    multidirectional exchange of requests, observations,
    results, and validated updates.
    Nodes also cooperate with peers and interact with their
    local environments.
    Logical services may be distributed across sites,
    while time-critical control remains local}
    \label{fig:internet-of-physical-agents}
\end{figure}

The \emph{Internet of Physical Agents} is a proposed federated infrastructure through which agents, models, data, computing resources, physical equipment, and human participants cooperate. Its architecture (Figure~\ref{fig:internet-of-physical-agents}) should support the circulation of observations, knowledge, task requests, actions, and verified outcomes among heterogeneous participants. A node may consume one capability while providing another, and its role can change from one task to the next.

Three interacting components provide a useful architectural starting point. Physical nodes combine local perception, computation, control, and equipment interfaces. Shared capability services provide models, skills, datasets, simulation, computing, and evaluation. Coordination services support discovery, resource matching, task decomposition, scheduling, authorization, and evidence exchange. These are logical functions that may be distributed or replicated across sites, rather than fixed levels in a centralized hierarchy.

Information and capability should move in multiple directions. A robot may receive a task and invoke a shared model, while also collecting measurements, identifying an unexpected condition, and supplying evidence that helps another agent. An instrument can execute an experimental procedure and return observations that change the procedure itself. One node's findings may update a shared resource, inform a peer directly, or trigger a request for further measurement elsewhere. The network therefore links execution with sensing, interpretation, and learning through several interacting loops.

Local autonomy remains necessary within this distributed architecture. Physical nodes retain time-critical control, equipment protection, and bounded execution. Network services coordinate activities whose timing and dependencies permit wider cooperation. Capability interfaces should specify operating conditions, uncertainty, latency, resource requirements, and failure handling. Provenance and version information are needed so that recipients can determine whether an observation, model, or procedure is suitable for their own setting.

The \emph{Scientific Internet} is one proposed implementation case. It would connect scientists, scientific literature, datasets, models, laboratories, instruments, and computing resources across disciplines and locations to support interdisciplinary autonomous scientific discovery. A workflow could move from literature analysis to hypothesis generation, simulation, physical experimentation, and revised investigation. Experimental results would feed back into hypotheses and models, while unresolved questions could redirect work to other facilities. Scientists would define objectives and assess significance, and scientific claims would remain subject to independent verification.

Sustained operation requires clear permissions and controlled updates. Equipment access, result sharing, and permission to learn from operational data should be specified separately. Changes to shared models or services need compatibility tests, staged deployment, monitoring, and rollback, since an update can affect dependent nodes. Participation also depends on practical arrangements for maintenance, cost sharing, and contribution recognition. Expansion should follow evidence that capabilities can be reused by other teams, coordination produces a net benefit, and service quality persists beyond the support of the original developers.

\subsubsection{Building the Scientific and Technological Ecosystem for Physical AI}
\label{subsec:physical-ai-development-ecosystem}

The reach of Physical AI depends on the maturity of the technologies and practices around it. Models require suitable instruments, reliable equipment, manufacturing capacity, energy supplies, communication infrastructure, and people who can integrate and maintain them. Development strategies should identify these complementary requirements and the bottlenecks that limit useful deployment. The limiting factor may be a model in one setting and measurement quality, equipment reliability, or integration effort in another.

A productive ecosystem makes results easier to reproduce and capabilities easier to extend. Shared test facilities, documented datasets, reference implementations, and interoperable interfaces can reduce repeated engineering work. Evaluation methods should support meaningful comparison across systems while leaving room for different architectures. Training and collaboration should connect AI research with control, equipment engineering, domain expertise, and operations. These connections are particularly important when moving from a successful prototype to a system that another team can use and maintain.

Scientific research provides one example of how Physical AI could strengthen this broader technological base. Better instruments and experimental methods can produce more informative evidence for learning. Models and agents can, in turn, help design experiments, investigate materials, improve fabrication processes, and diagnose equipment failures. Improvements can thus propagate between research tools, scientific knowledge, and engineering capability. Their value depends on reproducible findings and usable technologies, including cases where experiments reject a hypothesis or reveal a method's limits.

Similar feedback can occur in manufacturing, logistics, maintenance, and energy operations. Deployment reveals drift, wear, resource constraints, and interactions that are difficult to reproduce in isolated tests. When this experience is documented and returned to development, it can improve both the intelligent system and the processes that support it. Longer-term investment should preserve the ability to investigate unresolved problems while also developing the interfaces, equipment, and operating practices needed for reliable reuse.

Technological maturity should therefore be assessed beyond model-level benchmarks. Relevant measures include the cost of adaptation, intervention frequency, operating lifetime, recovery time, and useful output relative to total resource consumption. Reproducibility across sites and continued operation after the original developers leave are important tests of whether a capability has become broadly usable.

At a larger scale, Physical AI could extend the ability to investigate, construct, operate, and maintain physical systems. The significance of that extension depends on what can be accomplished, across how much space, for how long, and with what expenditure of energy, materials, and human effort. A capability acquires enduring value when it delivers useful work reliably, survives changing conditions, and can be recovered and passed on to others.

\subsection{Physical AI and Human Civilization}
\label{sec:physical-intelligence-civilization}

Physical AI connects computation, perception, and action. Questions about civilizational scale therefore depend on energy, materials, and infrastructure. The level a civilization can reach depends on how much energy it can obtain and organize, how effectively it can turn that energy into stable production and services, and whether those capabilities can continue to operate, be repaired, and be inherited across locations and generations. Larger models and more computation become civilizational capabilities only when they enter equipment, energy systems, and real environments and are tested through physical feedback.

Physical AI as a Service (PAIaaS) is used here to describe how such capabilities can be organized. Model as a service separates model inference from individual applications and lowers the cost of using digital intelligence. PAIaaS extends this idea to experiments, manufacturing, inspection, energy operations, and maintenance. Users can call on real equipment and operating procedures without rebuilding the full facility at every site. Physical capabilities can then be shared across locations and combined into more complex tasks.

The development of Physical AI is therefore connected to the growth of civilization. Reaching Type I requires organizing energy, production, and infrastructure across Earth. Reaching Type II requires infrastructure beyond Earth that can obtain resources, manufacture components, and operate for long periods. The discussion below first considers how physical capability can be measured and then examines the role of physical AI in progress toward Type I and Type II civilizations.

\subsubsection{Civilizational Scale and the Physical Basis of Intelligence}
\label{subsec:civilizational-scale}

Civilizational scale requires energy measures to be interpreted together with practical capability. The Kardashev index provides a power scale, but power alone says how much energy a civilization continuously directs, not what that energy enables. A physical measure must also account for the area a system can serve, the duration for which it can operate, and its ability to recover from failures and changing conditions.

\paragraph{Civilization as a Physical Measure}

In the continuous convention, the Kardashev index is written as
\begin{equation}
    K_{\mathrm{civ}} = \frac{\log_{10}(P_{\mathrm{power}}/\mathrm{W})-6}{10},
    \label{eq:kardashev}
\end{equation}
where $P_{\mathrm{power}}$ is the power continuously used by a civilization. Under this convention, Type~I, Type~II, and Type~III correspond to approximately $10^{16}$, $10^{26}$, and $10^{36}$ W, respectively. Using 2019 global total energy supply, \citet{jiang2022greatfilter} estimate humanity's index at about $K_{\mathrm{civ}}=0.73$. Using this rounded index in the continuous convention gives a Type~I power level about $10^{2.7}\approx501$ times the baseline~\citep{Kardashev1964,Sagan1973}. A small change in the index can therefore correspond to a large change in the energy base.

Power is an entry point for measuring civilizational capability. The same power can support different levels of production, transport, computation, and public service, or it can be consumed by inefficient conversion and maintenance. A fuller account should combine power with duration, spatial coverage, reliability, and recovery. A remote facility that operates for several hours has a different physical capability from one that can work for years without a local human team.

\paragraph{Energy Use and Energy Acquisition}

Overall energy efficiency is determined by two efficiencies working together. Energy-use efficiency describes how much of the energy already obtained becomes useful service. Energy-acquisition efficiency describes the equipment, materials, land, and maintenance required to bring new energy into the usable energy system. Higher use efficiency allows the same energy supply to support more service. Higher acquisition efficiency allows more resources to become available. A limitation at either stage constrains the expansion of a civilization's energy base.

Both efficiencies are physical. Use efficiency depends on equipment, materials, conversion processes, transmission networks, and operating conditions. Acquisition efficiency depends on resource location, environmental conditions, tool lifetime, operational safety, and resupply. Information processing alone cannot establish either one. Both have to be tested through equipment, field operations, and long-term maintenance. Energy systems should therefore be evaluated by the useful service delivered over a defined period, together with the full requirements for equipment, computation, materials, and personnel.

Civilization develops in two directions at once. It must make better use of energy already available and continue to obtain or create new energy sources. Digital systems rest on the same physical base. Computation requires electricity, chips, materials, cooling, and equipment maintenance. The expansion of digital capability is therefore supported and limited by physical conditions.

\paragraph{From Digital Information to Physical Action}

Digital systems process information, while civilizational capability is realized through physical processes. A model can propose a design, predict demand, or generate a control policy. Such results become testable changes only when they enter equipment and an environment. More information, model parameters, or computation cannot replace energy conversion, material processing, equipment assembly, or field maintenance.

Physical processes also change the meaning of a computation. Equipment may respond differently from the model's prediction, materials may degrade, and environmental conditions may shift. A system must update its judgment from observations and check whether the result meets requirements for safety, performance, and lifetime. Physical AI concerns this continuing interaction. Computation proposes an action, equipment carries it out, the environment provides feedback, and the system adjusts its next action.

The basic unit of Physical AI is therefore a physical system that combines perception, computation, execution, verification, and maintenance. An isolated model cannot provide this capability by itself. A computational capability acquires civilizational significance when it can reliably change a real process and remain interpretable and recoverable under different conditions.

\paragraph{From Local Capability to Shared Physical Services}

Physical capability usually begins with one set of equipment at one site. It performs a task under particular material, environmental, and operating conditions, while also generating calibration, failure, and maintenance experience. To move the capability to another site, its dependencies must be identified, transferable steps must be separated from site-specific ones, and the new setting must provide its own evidence.

When the experience, models, interfaces, and equipment can be used by another team, a local capability becomes a shared capability. Sharing includes equipment access, task execution, result verification, and maintenance support, not only a model file. A user specifies a physical task, a provider performs it under agreed conditions and returns evidence, and the parties define permissions, responsibilities, and failure handling.

PAIaaS is the service form of this shared capability. It makes experiments, measurements, manufacturing, energy operations, and maintenance available on demand. This lowers the threshold for using specialized facilities, improves equipment utilization, and allows services at different locations to be combined. A local success that has been validated can then be inherited by more teams and later generations.

Physical services remain constrained by resources, equipment, and operating environments, so they cannot be copied without limit like digital interfaces. A service must be usable by a new team, its results must withstand verification, and its facility must continue operating after the original developers leave. When these conditions are met, local experience can become a civilizational capability that remains available over time.

\subsubsection{Toward a Type I Civilization, Planetary-Scale Physical Systems}
\label{subsec:type-I}

Moving from this estimated level toward Type I begins on Earth. The first task is to organize energy and infrastructure at planetary scale. Physical AI supports continuous improvement of the energy system, beginning with more effective use and acquisition of existing energy, continuing with new energy processes, and extending to controlled fusion as a stable facility. Every stage depends on perception, control, diagnosis, maintenance, and iteration in real environments.

\paragraph{Existing Energy Systems}

Fossil fuels, wind, solar, hydropower, and nuclear fission already form the energy base of modern society. Here, the focus is on operating and maintaining facilities already in service. Physical AI can support coordinated dispatch, equipment inspection, fault diagnosis, and maintenance across the chain from resource acquisition to final use. For nuclear fission facilities, these tasks also involve monitoring material degradation and planning maintenance within independently verified safety limits.

Large energy networks depend on connected equipment. Wind and solar output varies with weather and location, while transmission and storage constrain dispatch. A local improvement can create congestion elsewhere. Physical AI can adjust generation, transmission, and demand using equipment state, weather, and changing loads. The relevant measures are reliable supply, recovery from failure, and sustained service.

Energy extraction is part of the same system. Oil, gas, coal, minerals, deep-sea resources, and nuclear fuels require exploration, collection, transport, processing, and maintenance. Wind, solar, and hydropower require dispersed equipment that must also be built and maintained. Local conditions in deep-sea, underground, and high-temperature or high-pressure facilities often exceed what design documents describe. Physical AI can use field observations to adjust an action, conduct a limited probe, or suspend a task when the risk becomes too high.

Records of extraction and maintenance should inform later construction. Repeated failures under particular conditions can expose problems in materials, site selection, or maintenance plans. Physical AI can identify such patterns and use them to improve equipment design and operation. Each operating cycle can then inform the next round of infrastructure.

\paragraph{Developing Advanced Energy Technologies}

The focus now shifts from operating existing facilities to developing and validating new materials, devices, and processes, including advances in photovoltaics, wind power, storage, and nuclear fission.

Physical AI can support this development by connecting design proposals to physical experiments. In photovoltaic research, a system could select material compositions and fabrication conditions, prepare samples, and compare measured performance with model predictions. Similar workflows could examine storage materials, turbine components, or materials intended for nuclear systems. The experimental objective is to establish how a proposed change affects performance, durability, and operating limits.

Results from these experiments should guide subsequent tests and revisions to materials, structures, and control plans. Progress toward deployment requires evidence from prototypes and extended operation, together with independent verification of relevant safety requirements. Physical AI can help carry knowledge from one development stage to the next, while controlled fusion presents a particularly demanding example of this process.

\paragraph{Controlled Fusion on Earth}

Controlled fusion is a key energy process on the path to Type I. The Sun continuously produces energy through fusion. On Earth, the challenge is to create and maintain an extreme plasma state while managing magnetic fields, heating, wall materials, energy output, and component lifetime.

Tokamak plasmas change rapidly and involve many control variables, so fixed rules cannot cover every operating state. \citet{degrave2022magnetic} show that reinforcement-learning controllers can move from simulation to a real TCV tokamak and control several plasma configurations magnetically. The experiment shows how Physical AI can turn a theoretical configuration into an operable and measurable experiment.

Fusion development must extend from research into engineering and operation. Models should be updated from experimental observations, follow-up experiments should be selected, equipment should be improved, and long-term stability should be tested. Sustained generation and safe maintenance require evidence from real devices. Physical AI can shorten the path from physical discovery to engineering iteration and carry the knowledge from one experiment into the next device.

\paragraph{Planetary Coordination and Long-Term Maintenance}

When energy systems expand to planetary scale, one facility can affect other regions and infrastructures. Reservoir operation can change downstream water and ecological conditions, while large energy facilities can alter land use, water use, and regional industry. Models must represent these connections and be tested against observations at the relevant scale.

Physical AI can help identify missing data, compare operating plans, and schedule monitoring and maintenance. Models cannot replace public decisions or remove the physical consequences of intervention. Climate intervention in particular requires a clear distinction between mechanism studies and large-scale deployment, together with explicit treatment of uncertainty, risk, and responsibility.

Type I civilization requires energy and infrastructure services to continue over long periods. Inspection, repair, and replacement should occur before failures spread, and efficiency should be evaluated across the full service cycle. More complex control can increase sensor and computation requirements, while equipment that cannot be repaired locally can increase dependence on remote supply. The maturity of Physical AI is assessed by stable energy supply, continuing maintenance, and rapid recovery across the planet.

\subsubsection{Toward a Type II Civilization, Space Infrastructure and Stellar-Scale Energy}
\label{subsec:type-II}

Type II civilization extends physical activity to stellar scale. One reason for this transition is that energy development and large infrastructure on Earth face increasing constraints from space, environmental effects, and social coordination. New computing, energy, and manufacturing facilities occupy public space and can conflict with local communities and other industries. Moving parts of energy production, computation, resource extraction, and manufacturing to orbit, the Moon, Mars, and other bodies may provide another way to expand civilizational capability.

Space facilities must do more than complete a single mission. They need to build themselves, use local resources, produce replacement parts, manage failures, and operate with little direct human presence. Physical AI allows these activities to continue despite communication delay, extreme environments, and incomplete information.

\paragraph{Deep-Space Autonomy and Continual Learning}

Earth-Mars one-way light time varies with orbital configuration from roughly 3 to 22\,min~\citep{MarsRelay}. Urgent actions must be performed locally, equipment diagnosis can wait longer, and changes to mission objectives generally require ground review. Perseverance's AutoNav illustrates this arrangement by supporting onboard navigation and hazard avoidance while the ground team plans routes and mission activities~\citep{AutoNav}. Remote autonomy can begin with defined local tasks, while validated control, local adaptation, and ground supervision handle decisions at different timescales.

Deep-space missions encounter environments that cannot be fully covered during preparation on Earth. Javed and Sutton's Big World Hypothesis describes agents whose capacity is limited relative to the environments they encounter~\citep{JavedSutton2024}. Such agents cannot maintain a complete world description and must understand the parts relevant to their current goals. A model trained on Earth therefore omits features that matter at a particular site. New observations may require a revised state estimate, a different existing plan, or a change to the model and policy.

Continual learning needs evidence from local experiments. A rover can first probe low-risk ground and then decide whether to enter more difficult terrain. The probe should change the next decision, while the risks of immobilization, energy use, and tool damage remain acceptable. Some unknowns cannot be resolved safely, so the system must be able to postpone an objective. Long-duration construction also requires constrained physical self-improvement in tools, experiments, and local control. Such changes need field tests, regression checks, and rollback procedures.

\paragraph{Adaptation Under Unfamiliar Physical Conditions}

Deep-space autonomy must also address physical conditions that differ from those encountered during development on Earth. The broad gravitational conditions of the Moon and Mars are already characterized, but local terrain, contact properties, and equipment responses may remain incompletely known. The challenge is to determine how transferable physical models apply at a particular site and when their assumptions cease to support reliable action.

Reduced gravity changes the relationship between equipment weight, available traction, and the reaction forces generated by tools. An excavator or assembly robot may move or lose stable contact when applying forces that a terrestrial platform could resist. NASA's RASSOR excavator illustrates the resulting need for different mechanical arrangements. Opposing, counterrotating bucket drums reduce the net reaction forces associated with excavation.\footnote{NASA Kennedy Space Center, \href{https://technology.nasa.gov/patent/KSC-TOPS-7}{\emph{Regolith Advanced Surface Systems Operations Robot (RASSOR) Excavator}}, technology-transfer page.} Physical AI would need to estimate support conditions and coordinate tool motion with vehicle posture, anchoring, or other means of stabilization.

Illumination creates a different set of coupled constraints. Low solar angles near the lunar south pole produce long shadows and strong contrasts that complicate terrain interpretation~\citep{NASALunarLighting}. A route that appears geometrically accessible may offer insufficient visibility or solar energy for continued operation. Physical AI should therefore combine terrain estimates, sensing confidence, illumination forecasts, and energy reserves when selecting routes and work periods.

Surface interaction introduces further uncertainty. Friction alone is insufficient to describe whether a vehicle slips, sinks, or remains supported, or whether a tool can penetrate the ground. Local estimates must also account for bearing capacity, cohesion, and the response of material to repeated contact. Limited traverses or shallow tool interactions could provide evidence for revising these estimates, provided that the system retains sufficient resources and a feasible recovery option.

Future research should examine how Physical AI can identify these conditions from limited observations and interactions, adapt control without losing previously validated capabilities, and recognize when further probing would be too costly. Evaluation should include the physical interactions required before reliable operation, the energy and equipment wear incurred during adaptation, and recovery when an environmental estimate proves incorrect.

\paragraph{Planetary Construction and Underground Habitats}

Space construction is harder than a single measurement because each action changes the setting for the next one. Lunar and Martian facilities may benefit from underground spaces, which can provide radiation protection and stable temperatures while reducing exposure to surface conditions. Construction must proceed from local observations and repeatedly adjust excavation, support, and inspection.

A lunar energy facility would pass through transport, positioning, installation, inspection, and connection. Each installed component must be checked before later work can rely on it. The system cannot simply execute a design prepared in advance. It must establish step by step that the emerging structure can support further construction and operation.

Space-based solar power provides a concrete object for studying these dependencies. NASA's 2024 assessment identifies major gaps in orbital assembly and maintenance, autonomous operation, and power transmission~\citep{SBSP2024}. An orbital collector becomes an energy service only when its associated equipment can be deployed, repaired, and operated continuously. Physical AI must connect construction planning, state inspection, failure handling, and later task scheduling.

Local production can reduce resupply, but raw material alone does not create manufacturing capability. A remote facility must repeatedly produce components that meet performance requirements and verify that they are fit for their intended roles. Materials processing, quality verification, and equipment maintenance must form a continuous local process.

\paragraph{Long-Duration Operation with Limited Resources}

Decision autonomy and resource independence are different properties. A facility may schedule months of work while still depending on Earth for a critical component. Better planning can extend limited supplies, but it cannot replace missing materials or tools. The autonomy of a remote facility must therefore be described together with its remaining dependence on resupply, communication, and maintenance.

Long-duration operation requires choices between current tasks and future capability. When resources are scarce, equipment status is uncertain, or external support is interrupted, postponing an experiment may preserve more useful work. A remote facility should leave later missions with usable energy, tools, data, and maintenance conditions rather than forcing every mission to start again from Earth.

\paragraph{Stellar-Scale Infrastructure and Dyson Swarms}

A Dyson swarm is a hypothetical arrangement of independently orbiting collectors for harvesting stellar radiation~\citep{Wright2020Dyson}. Its relevance to Physical AI lies in the construction and maintenance tasks that such an arrangement would impose. Components would have to be manufactured, transported, assembled, inspected, and replaced across widely separated sites, with limited opportunities for direct human intervention.

Physical AI could support these tasks by connecting local observations to physical action. During assembly, a robotic system would need to estimate component pose and deformation, adjust contact forces, and verify each completed connection before further construction proceeds. During operation, each facility would need to diagnose degradation and adjust energy collection and use within equipment and thermal limits. Repair would require identifying a fault, selecting an available intervention, and checking that the repaired equipment can safely resume its role.

Coordination across the swarm would also require Physical AI at multiple scales. Local systems would handle immediate control and fault response, while broader coordination would allocate transport, manufacturing, and maintenance resources using delayed information. Construction and repair records should preserve the conditions under which earlier designs were validated, allowing later systems to assess compatibility when replacing components or extending facilities.

These requirements define a speculative research setting for autonomous construction, distributed coordination, and maintenance across successive generations of equipment. Physical AI could help identify and operate workable arrangements, but those arrangements would still require suitable materials, manufacturing capacity, and sufficient heat rejection. Their capability would be demonstrated through verified construction, recovery from failures, and sustained useful operation.

\subsection{Summary and Outlook}
\label{subsec:summary-outlook}

This chapter has considered Physical AI over the long time horizon of human development and civilizational change. Physical AI requires stable infrastructure and systems in which equipment, data, computation, and people can continue to work together. The value of a project lies in its results and in the models, interfaces, methods, and operating experience that later teams can verify and reuse.

The development of Physical AI can begin in scientific research and engineering practice. Shared models and data, connected experimental facilities, cross-institutional service interfaces, and deployment of validated capabilities all contribute to the accumulation of physical capability. PAIaaS offers an organizing form for this process. Experiments, manufacturing, inspection, energy operations, and maintenance can become services that other teams call when needed. Physical capability can then grow from a single site into a broader network of cooperation.

At the scale of civilization, humanity must first improve the acquisition and use of energy on Earth, develop new energy and nuclear systems, build planetary infrastructure, and advance controlled fusion. A more distant stage requires deep-space infrastructure that can operate with communication delay, continual learning, autonomous construction, local manufacturing, and long-duration activity. Civilizational level is ultimately reflected in whether energy, equipment, materials, and maintenance can continue to work across larger ranges and longer periods.

This monograph has examined the foundations, methods, systems, and applications of Physical AI, and has considered the perception, action, learning, and adaptation required by Physical AGI. These discussions show that Physical AI requires systems to understand environments, act in them, bear the consequences of action, and change their later behavior from real experience. A digital model becomes a component of a Physical AI system only when it is integrated into this closed-loop process.

Future progress must be tested in concrete equipment and real tasks. Laboratories, factories, energy facilities, and deep-space systems impose different constraints and expose unresolved weaknesses in current models and methods. Progress toward Physical AGI depends on whether systems can gain reliable experience under these constraints, turn local experience into transferable capability, and remain safe, interpretable, and maintainable over long operation. Physical AI may become an important technical basis for humanity's transition to a higher civilization, but reaching that future requires sustained work by researchers, engineers, and infrastructure operators.

\clearpage
\phantomsection
\section*{Contributors}
\addcontentsline{toc}{section}{Contributors}

\noindent
\emph{Monograph leading authors:}
Weinan Zhang, Chunhua Shen, Tianfan Xue, Bin Zhao, Bowen Zhou
\par\medskip

\noindent
\emph{Overall coordination group:}
Weinan Zhang, Bingqi Jiang, Zihan Guo, Zicai Cui
\par\medskip

\begingroup
\setlength{\leftmargini}{1.2em}

\begingroup
\begin{list}{}{%
  \settowidth{\labelwidth}{\textbullet\enspace Chapter~12:}
  \setlength{\labelsep}{0.6em}
  \setlength{\leftmargin}{\labelwidth}
  \addtolength{\leftmargin}{\labelsep}
  \setlength{\rightmargin}{0pt}
  \setlength{\itemindent}{0pt}
  \setlength{\listparindent}{0pt}
  \setlength{\itemsep}{0.35\baselineskip}
  \setlength{\parsep}{0pt}
  \renewcommand{\makelabel}[1]{#1\hfil}
}

  \item[\textbullet\enspace Chapter~\ref{ch:introduction}:]
  Weinan Zhang, Yang Li, Ming Zhou, Bingqi Jiang.

  \item[\textbullet\enspace Chapter~\ref{sec:technical-foundations}:]
  Tianfan Xue, Xinyuan Chen, Lihe Ding, Xiao Chen.

  \item[\textbullet\enspace Chapter~\ref{ch:perception}:]
  Tianfan Xue, Yujin Wang, Shi Guo.

  \item[\textbullet\enspace Chapter~\ref{ch:physical-data}:]
  Chunhua Shen, Junting Dong, Tao Lu, Xing Shen.

  \item[\textbullet\enspace Chapter~\ref{ch:wm}:]
  Chunhua Shen, Haoyu Guo.

  \item[\textbullet\enspace Chapter~\ref{ch:manipulation-models}:]
  Weinan Zhang, Yao Mu, Haiyong Yu, Haoxiang Ma, Fuxian Huang, Qiaojun Yu, Yunsong Zhou.

  \item[\textbullet\enspace Chapter~\ref{ch:whole-body-motor-control}:]
  Weinan Zhang, Kailin Li, Zirui Wang, Yufei Xue, Weishuai Zeng, Tao Huang.

  \item[\textbullet\enspace Chapter~\ref{ch:physical-agent}:]
  Bin Zhao, Zhigang Wang, Junli Liu, Pengyuan Wu, Pingrui Zhang, Haotian Wang, Dongzhan Zhou.

  \item[\textbullet\enspace Chapter~\ref{ch:hardware-systems}:]
  Bin Zhao, Jiafei Cao, Feng Yuan, Xinyi Ye, Quanli Xuan, Yunsong Zhou, Kailin Li.

  \item[\textbullet\enspace Chapter~\ref{ch:general-physical-ai}:]
  Chunhua Shen, Mulin Yu, Shangding Gu, Kaiwen Song.

  \item[\textbullet\enspace Chapter~\ref{ch:evaluation}:]
  Bin Zhao, Dong Wang, Jiarui Li, Kehui Liu, Zhongjie Jia, Dongjiang Li, Jiacheng Bao, Haoran Yang, Zhaoqing Chen, Shangding Gu.

  \item[\textbullet\enspace Chapter~\ref{ch:strategy-outlook}:]
  Weinan Zhang, Bingqi Jiang, Zihan Guo, Yiming Sun.

\end{list}
\endgroup
\endgroup

\medskip
\begin{tcblisting}{
  listing only,
  breakable,
  width=\linewidth,
  title={Suggested citation (BibTeX)},
  fonttitle=\small\bfseries,
  coltitle=black,
  colbacktitle=black!6,
  colback=black!2,
  colframe=black!25,
  boxrule=0.4pt,
  arc=1.5mm,
  left=2.5mm,
  right=2.5mm,
  top=2mm,
  bottom=2mm,
  listing options={
    basicstyle=\ttfamily\footnotesize,
    columns=fullflexible,
    keepspaces=true,
    showstringspaces=false,
    breaklines=true,
    breakatwhitespace=true,
    breakindent=1em,
    aboveskip=0pt,
    belowskip=0pt
  }
}
@misc{zhang2026physicalai,
  author = {Zhang, Weinan and Shen, Chunhua and Xue, Tianfan and Zhao, Bin and Jiang, Bingqi and Guo, Zihan and Cui, Zicai and Li, Yang and Zhou, Ming and Chen, Xinyuan and Ding, Lihe and Chen, Xiao and Wang, Yujin and Guo, Shi and Dong, Junting and Lu, Tao and Shen, Xing and Guo, Haoyu and Mu, Yao and Yu, Haiyong and Ma, Haoxiang and Huang, Fuxian and Yu, Qiaojun and Zhou, Yunsong and Li, Kailin and Wang, Zirui and Xue, Yufei and Zeng, Weishuai and Huang, Tao and Wang, Zhigang and Liu, Junli and Wu, Pengyuan and Zhang, Pingrui and Wang, Haotian and Zhou, Dongzhan and Cao, Jiafei and Yuan, Feng and Ye, Xinyi and Xuan, Quanli and Yu, Mulin and Gu, Shangding and Song, Kaiwen and Wang, Dong and Li, Jiarui and Liu, Kehui and Jia, Zhongjie and Li, Dongjiang and Bao, Jiacheng and Yang, Haoran and Chen, Zhaoqing and Sun, Yiming and Zhou, Bowen},
  title = {{Physical AI}: Concepts, Techniques, and Challenges},
  year = {2026},
}
\end{tcblisting}

\clearpage
\section*{Terminology and Abbreviations}
\addcontentsline{toc}{section}{Terminology and Abbreviations}
Chapter links point to uses of each term in the sense defined here, including its full form and equivalent expressions.
\newcommand{\PAIChapter}[1]{%
  \ifcase#1\relax
  \or\ref{ch:introduction}%
  \or\ref{sec:technical-foundations}%
  \or\ref{ch:perception}%
  \or\ref{ch:physical-data}%
  \or\ref{ch:wm}%
  \or\ref{ch:manipulation-models}%
  \or\ref{ch:whole-body-motor-control}%
  \or\ref{ch:physical-agent}%
  \or\ref{ch:hardware-systems}%
  \or\ref{ch:general-physical-ai}%
  \or\ref{ch:evaluation}%
  \or\ref{ch:strategy-outlook}%
  \fi
}
\makeatletter
\newcommand{\PAIScope}[1]{%
  \def\pai@sep{}%
  \@for\pai@ch:=#1\do{%
    \pai@sep\expandafter\PAIChapter\expandafter{\pai@ch}%
    \def\pai@sep{, }%
  }%
}
\makeatother
\begingroup
\small
\setlength{\tabcolsep}{4pt}
\renewcommand{\arraystretch}{1.15}
\newcommand{\Term}[3]{#1 & #2 & \PAIScope{#3}\\}
\begin{longtable}{@{}>{\raggedright\arraybackslash}p{.27\textwidth}
  >{\raggedright\arraybackslash}p{\dimexpr.59\textwidth-4\tabcolsep\relax}
  >{\centering\arraybackslash}p{.14\textwidth}@{}}
\toprule
\textbf{Term} & \textbf{Meaning in this monograph} & \textbf{Chapters}\\
\midrule
\endfirsthead
\toprule
\textbf{Term} & \textbf{Meaning in this monograph (continued)} & \textbf{Chapters}\\
\midrule
\endhead
\midrule
\multicolumn{3}{r}{\emph{Continued on the next page}}\\
\endfoot
\bottomrule
\endlastfoot
\Term{2D}{\emph{Two-Dimensional}. Representation in a plane or another two-dimensional space.}{3,5,6}
\Term{3D}{\emph{Three-Dimensional}. Representation of spatial position, shape, and orientation in three dimensions.}{1,2,3,4,5,6,7,8,9,11}
\Term{3DGS}{\emph{3D Gaussian Splatting}. Scene representation and rendering using anisotropic three-dimensional Gaussian primitives.}{3,4}
\Term{4D}{\emph{Four-Dimensional}. Three-dimensional spatial structure evolving over time in the perception context used here.}{3,5}
\Term{Action alignment}{Consistency of intentions, plans, and execution with human authorization, constraints, and acceptable outcomes.}{11}
\Term{Action chunk}{A contiguous sequence of actions generated or processed together.}{4,5,6,7,9}
\Term{Action expert}{A specialized module that generates continuous actions from semantic and robot-state inputs.}{5,6}
\Term{Action proxy}{An action representation inferred from video, poses, trajectories, or latent variables. It is not necessarily an executable robot command.}{4,6}
\Term{ACT}{\emph{Action Chunking with Transformers}. A conditional-variational-autoencoder policy that predicts action chunks, using an $L_1$ reconstruction term and a KL-divergence regularizer during training.}{6,9}
\Term{Affordance}{An action opportunity relative to an agent's body, current conditions, and constraints.}{1,2,3,8,9,11}
\Term{AGI}{\emph{Artificial General Intelligence}. The goal of general intelligence that learns and transfers across tasks and domains.}{1,10,12}
\Term{AI}{\emph{Artificial Intelligence}. Machine capabilities for perception, learning, reasoning, and decision-making.}{1,2,3,4,5,6,7,8,9,10,11,12}
\Term{AoI}{\emph{Age of Information}. Time elapsed since generation of the latest update available at a receiver. The age of the observation actually used by an action is a distinct quantity.}{9}
\Term{API}{\emph{Application Programming Interface}. An interface through which software invokes capabilities and exchanges data.}{8,9}
\Term{AR}{\emph{Autoregressive}. Sequential prediction conditioned on preceding variables or tokens.}{5,6,7}
\Term{Asset Administration Shell}{A standardized representation of industrial-asset information and access to services.}{8}
\Term{Autonomy}{The degree of independent action selection, execution, monitoring, and correction within specified objectives and operating limits.}{1,2,3,4,6,7,8,9,10,11,12}
\Term{BC}{\emph{Behavior Cloning}. Supervised policy learning from demonstrated actions conditioned on observations, histories, or task context.}{4,5,6}
\Term{Behavior tree}{A tree-structured execution model organizing conditions, actions, sequences, and fallback behavior.}{2,8,9}
\Term{Belief state}{A probability distribution over states conditioned on observation and action history.}{2,5,8,11}
\Term{BFM}{\emph{Behavior Foundation Model}. A humanoid-control model that learns reusable whole-body behavior from broad motion data and interaction.}{7}
\Term{BLDC}{\emph{Brushless Direct Current Motor}. An electronically commutated permanent-magnet motor. The convention used here associates BLDC operation with trapezoidal back electromotive force and block commutation.}{9}
\Term{BS}{\emph{Brier Score}. The mean squared difference between predicted event probabilities and binary outcomes.}{5}
\Term{CAD}{\emph{Computer-Aided Design}. Editable geometric and engineering representations of objects or systems.}{3,7}
\Term{Calibration}{Adjustment or assessment against reference measurements or outcomes. Sensor, simulator, probability, and evaluator calibration have distinct objects.}{1,2,3,4,5,6,7,8,9,11,12}
\Term{Capability graph}{A representation of available capabilities, their applicability conditions, and their composition relationships.}{9}
\Term{Causal Imprint}{A research-specific mechanism in InternW0-$\Delta$ that learns action-relevant scene-change representations from observed context using future-derived training supervision. Realized future observations are not inputs to the action-prediction path.}{5}
\Term{CDF}{\emph{Cumulative Distribution Function}. A function giving the probability that a random variable does not exceed a specified value.}{5,7}
\Term{CL}{\emph{Continual Learning}. Learning from new data or experience while maintaining previously acquired capabilities.}{1,6,10,11,12}
\Term{Closed loop}{\emph{Closed-Loop Feedback}. A feedback process in which measured outcomes inform subsequent decisions and control.}{1,2,3,4,5,6,7,8,9,11,12}
\Term{CMC}{\emph{Carpometacarpal Joint}. A joint between the wrist and metacarpal bones, used here to describe hand kinematics and sensor locations.}{9}
\Term{CNN}{\emph{Convolutional Neural Network}. A neural network using locally shared convolutional operations.}{3}
\Term{Collective Physical AI}{System-level physical capability arising from complementary agents, shared information, or local coordination rules.}{10}
\Term{CoM}{\emph{Center of Mass}. The position determined by the mass-weighted distribution of a body.}{4,6,7,8,9}
\Term{Compliance}{\emph{Mechanical Compliance}. Displacement or deformation under an applied force, affecting contact and impact response.}{1,2,3,4,6,7,8,9,11,12}
\Term{Control barrier function}{A function specifying a safe set and associated constraints on admissible control under stated assumptions.}{2,8,11}
\Term{Controllability}{For a specified dynamical model, the ability of admissible inputs to steer the state between specified states over a finite interval.}{2}
\Term{Counterfactual}{\emph{Counterfactual Prediction}. Prediction of consequences under alternative actions or interventions.}{1,4,5,6}
\Term{CPS}{\emph{Cyber-Physical Systems}. Systems integrating computation, communication, and physical processes, including their timing and interface constraints.}{1}
\Term{CPU}{\emph{Central Processing Unit}. A processor for general-purpose computation and software execution.}{4,9}
\Term{CRPS}{\emph{Continuous Ranked Probability Score}. A proper scoring rule comparing a predictive cumulative distribution with a realized scalar observation.}{5}
\Term{CTDE}{\emph{Centralized Training with Decentralized Execution}. Training with broader information while deployed agents act on local inputs and available messages.}{8}
\Term{Curriculum learning}{Training organized progressively by difficulty, horizon, or task structure.}{2,4,5,6,7}
\Term{DAgger}{\emph{Dataset Aggregation}. An iterative imitation-learning method that aggregates expert labels on states visited during rollouts that may mix expert and learner policies. Its task-cost guarantees depend on the stated learning and analysis assumptions.}{4,6,9}
\Term{Data flywheel}{An improvement cycle linking deployment, evaluation, correction, data selection, and retraining.}{1,4}
\Term{Data pyramid}{An organization of data sources by scale, diversity, collection cost, and relevance to target physical execution.}{4,9}
\Term{DC}{\emph{Direct Current}. Electric current with a fixed direction, used here in motor-drive discussions.}{9}
\Term{DDS}{\emph{Data Distribution Service}. A data-centric publish-subscribe communication standard with configurable quality-of-service policies.}{9}
\Term{Diffusion policy}{A policy generating continuous action sequences through conditional denoising.}{4,5,6,8}
\Term{Digital AI}{AI whose direct operating interfaces act on digital representations, regardless of whether the represented information originates in the physical world.}{1}
\Term{Digital cousin}{A generated environment preserving relevant interaction properties without exactly replicating a physical scene.}{4}
\Term{Digital twin}{A digital representation associated with a particular physical object or process and updated using observations.}{1,4,5}
\Term{Distillation}{Transfer of knowledge from a teacher, privileged information, or auxiliary modalities to a deployment model.}{4,5,6,7}
\Term{Distributed Physical AI}{Coordinated physical capability across nodes retaining local sensing, decision-making, and action.}{8,9}
\Term{DiT}{\emph{Diffusion Transformer}. A Transformer-based denoising architecture for image, video, or continuous-action generation.}{6}
\Term{DoF}{\emph{Degrees of Freedom}. The number of independent coordinates describing a system's motion.}{2,4,6,7,8,9}
\Term{Domain randomization}{Variation of simulated parameters, appearance, or disturbances to cover uncertain deployment conditions.}{1,4,7}
\Term{EKF}{\emph{Extended Kalman Filter}. A nonlinear state estimator using local linearization for prediction and measurement correction.}{7}
\Term{ELBO}{\emph{Evidence Lower Bound}. A variational lower bound on log marginal likelihood. Its negative is used as a training objective.}{5}
\Term{EMA}{\emph{Exponential Moving Average}. A recursively weighted average used here to update target-encoder parameters from student-encoder parameters.}{5}
\Term{Embodied AI}{\emph{Embodied Artificial Intelligence}. AI studied through the coupled relationships among an agent's body, perception, action, and environment, including simulated embodiments.}{1,2}
\Term{Embodied cognition}{Theories explaining cognition through interactions among an agent, its body, and its environment.}{1,2}
\Term{Embodiment}{The body structure, sensors, actuators, and physical interfaces through which a system interacts with its environment.}{1,2,3,4,5,6,7,8,9,10,11,12}
\Term{EMG}{\emph{Electromyography}. Recording of muscle electrical activity for movement analysis or human-machine interfaces.}{9}
\Term{End effector}{A tool, gripper, or hand through which a robot directly interacts with a task object.}{1,2,3,4,6,7,8,9,11,12}
\Term{Equivariance}{A property whereby a transformation of the input induces the corresponding transformation of the output.}{2}
\Term{ESKF}{\emph{Error-State Kalman Filter}. A Kalman-filter formulation propagating a nominal state and estimating local state errors.}{7}
\Term{EtherCAT}{\emph{Ethernet for Control Automation Technology}. An industrial Ethernet technology for real-time automation and control.}{9}
\Term{Ethernet}{A networking technology used to connect devices, task computers, and control systems.}{9}
\Term{FDM}{\emph{Forward Dynamics Model}. A model predicting future observations or representations from observed context and specified actions.}{5}
\Term{Fine-tuning}{Adaptation by updating parameters of an existing model using new data.}{4,6,7,9,10,11}
\Term{FM}{\emph{Flow Matching}. Generative learning of a flow or velocity field connecting noise and data distributions.}{5,6}
\Term{FOC}{\emph{Field-Oriented Control}. Motor control using a rotating reference frame to regulate flux-related and torque-related current components.}{9}
\Term{Forward dynamics}{Computation of acceleration or state evolution from a mechanical state and input forces. Learned formulations may instead predict states or observations from context and actions.}{5,7}
\Term{Foundation model}{A model pretrained on broad data that can be adapted to a range of downstream tasks.}{1,2,4,5,6,7,8,9,10,11,12}
\Term{Foundation models for robot manipulation}{Pretrained models providing reusable representations and policies for manipulation across tasks, objects, environments, or robot embodiments.}{1,4,6,10}
\Term{Foundation models for whole-body motor control}{Broadly pretrained models providing reusable motor capabilities for coordinated whole-body motion under dynamics, contact, and actuator constraints.}{1,4,6,7,10}
\Term{Friction cone}{The set of admissible contact-force vectors with a nonnegative normal component and a tangential magnitude bounded by the friction coefficient times the normal component.}{2,7}
\Term{FSM}{\emph{Finite-State Machine}. An execution model with discrete modes and event-triggered transitions.}{2}
\Term{FVD}{\emph{Fr\'echet Video Distance}. A video-generation metric comparing distributions of learned video features.}{5}
\Term{G-MPKPE}{World-frame position error averaged over evaluated timesteps and selected keypoints.}{7}
\Term{G-MPKRE}{World-frame orientation error averaged using angular distance on the rotation group.}{7}

\Term{GPU}{\emph{Graphics Processing Unit}. A processor supporting parallel rendering, simulation, training, and inference.}{4,9}
\Term{Grounding}{Association of symbols, instructions, or goals with observable entities, action capabilities, and physical consequences.}{1,2,3,4,5,6,8,10,12}
\Term{Harness}{\emph{Embodied Intelligence Harness}. A runtime organization connecting model proposals, skill contracts, actuation, feedback, and authorization boundaries.}{9}
\Term{HASEL}{\emph{Hydraulically Amplified Self-Healing Electrostatic Actuator}. A soft actuator using electrostatic forces to redistribute a dielectric liquid and produce deformation.}{9}
\Term{HIL}{\emph{Hardware-in-the-Loop}. Testing that connects target control hardware and firmware to a simulated plant.}{1,6,9,11}
\Term{HITL}{\emph{Hardware-in-the-Loop}. The PX4 spelling of hardware-in-the-loop testing in this monograph. It does not denote human-in-the-loop operation here.}{9}
\Term{I/O}{\emph{Input/Output}. Sensor and actuator interfaces or computational data transfer and access.}{9}
\Term{IDM}{\emph{Inverse Dynamics Model}. A model inferring compatible actions from observed state or observation transitions. Using realized future observations is a training or retrospective inference interface.}{4,5,7}
\Term{Imitation learning}{Learning action selection or movement from demonstrations.}{2,4,6,7}
\Term{Impedance control}{Control of interaction through a specified dynamic relationship between motion and force.}{7,9}
\Term{IMU}{\emph{Inertial Measurement Unit}. An inertial sensor assembly providing angular-rate and specific-force measurements.}{3,7,9,11}
\Term{InEKF}{\emph{Invariant Extended Kalman Filter}. An extended Kalman filter using system symmetries to formulate estimation errors.}{7}
\Term{Information freshness}{The temporal relevance of available observations to a current decision or action.}{8,9}
\Term{Inverse dynamics}{Computation of required forces or torques from a mechanical state and desired acceleration. Learned formulations may infer actions from transitions.}{2,4,5,6,7}
\Term{ISP}{\emph{Image Signal Processor/Processing}. The processor or processing pipeline converting raw camera samples into usable images.}{3}
\Term{JEPA}{\emph{Joint-Embedding Predictive Architecture}. An architecture predicting target representations from context representations without requiring reconstruction of every raw observation.}{2,5,10}
\Term{Kardashev index}{An index expressing a civilization's energy scale in terms of sustained power use.}{12}
\Term{KL}{\emph{Kullback-Leibler Divergence}. An asymmetric divergence between probability distributions.}{5,6}
\Term{Knowledge insulation}{A VLA training scheme that blocks gradients from the continuous action expert into the vision-language backbone while training the backbone with discrete action and other available vision-language objectives.}{6}

\Term{LiDAR}{\emph{Light Detection and Ranging}. Laser-based distance sensing for range measurements or point clouds.}{2,3,7,8,9}
\Term{LLM}{\emph{Large Language Model}. A large language model used for language processing, reasoning, or task decomposition.}{1,10,11}
\Term{L-MPKPE}{Mean per-keypoint position error after removing horizontal root translation and heading at each timestep.}{7}
\Term{L-MPKRE}{Mean per-keypoint orientation error in the local frame defined by root heading at each timestep.}{7}
\Term{LoRA}{\emph{Low-Rank Adaptation}. Adaptation of a pretrained model using trainable low-rank parameter updates.}{6}
\Term{LPIPS}{\emph{Learned Perceptual Image Patch Similarity}. An image-difference metric computed from learned visual features.}{5}
\Term{LQR}{\emph{Linear-Quadratic Regulator}. Optimal state feedback for a linear model with a quadratic cost.}{2}
\Term{Lyapunov function}{A scalar function whose evolution along trajectories supports a stability analysis.}{2}
\Term{MCP}{\emph{Model Context Protocol}. A protocol supporting structured discovery and invocation of tools by AI applications.}{8}
\Term{MDP}{\emph{Markov Decision Process}. A sequential-decision model specified by states, actions, transitions, rewards, and a planning or discounting convention.}{2}
\Term{MHS}{\emph{Model Hardware Standard}. The physical-device discovery, description, and runtime-interface specification discussed in this monograph.}{8}
\Term{MLP}{\emph{Multilayer Perceptron}. A feedforward network composed of successive nonlinear transformations.}{6,7}
\Term{MMRV}{\emph{Mean Maximum Rank Violation}. A metric quantifying real-performance differences associated with simulator-induced policy-ranking errors.}{4}
\Term{Model-hardware co-design}{Joint design of mechanisms, sensing, action interfaces, and models under task and resource constraints.}{9}
\Term{MOF}{\emph{Metal-Organic Framework}. A class of porous materials formed from metal nodes and organic linkers.}{9}
\Term{Morphological computation}{Use of body geometry and dynamics to distribute or reduce the computation required for control.}{2}
\Term{MoT}{\emph{Mixture of Transformers}. An architecture in which separately parameterized Transformer experts exchange or jointly process modality information.}{5,6}
\Term{MPC}{\emph{Model Predictive Control}. Repeated finite-horizon optimization based on the current state estimate, followed by execution of the initial control segment.}{1,2,5,7,8}
\Term{Native governance}{Governance that embeds authorization, responsibility, constraints, and traceability within capability networks and execution.}{9}
\Term{NeRF}{\emph{Neural Radiance Field}. A continuous neural representation mapping position and viewing direction to density and radiance or appearance.}{2,3,4}
\Term{Neural ODE}{\emph{Neural Ordinary Differential Equation}. A continuous-time dynamical model with a neural-network-parameterized state derivative.}{2}
\Term{NMPC}{\emph{Nonlinear Model Predictive Control}. Receding-horizon optimization with nonlinear dynamics and relevant motion or contact constraints.}{7}
\Term{Observability}{The ability to infer relevant states from measurements under a specified dynamics and sensing model.}{2,9,11}
\Term{Operational controllability}{The ability of protective mechanisms and authorized supervision to intervene in time, maintain operating limits, and establish an appropriate fallback state. This differs from dynamical controllability.}{11}
\Term{OPC UA}{\emph{OPC Unified Architecture}. An industrial communication and information-modeling architecture for interoperable services and devices.}{8}

\Term{OTA}{\emph{Over-the-Air Update}. Network delivery of software, model, or firmware updates.}{11}
\Term{Outcome unknown}{An execution state in which an action may have affected the world but its outcome or acknowledgment is unavailable.}{9}
\Term{PAIaaS}{\emph{Physical AI as a Service}. The service organization described in this monograph, through which users access physical capabilities such as experiments, manufacturing, inspection, or maintenance over shared interfaces.}{12}
\Term{Passivity}{A system property expressed through input-output energy inequalities.}{7}
\Term{PD}{\emph{Proportional-Derivative Control}. Feedback using proportional and derivative error terms.}{7}
\Term{PDE}{\emph{Partial Differential Equation}. An equation involving partial derivatives of an unknown field with respect to multiple independent variables.}{2,5}
\Term{Physical agent}{An agent that senses through physical interfaces, selects interventions, and adjusts behavior using measured outcomes.}{1,2,5,8,11,12}
\Term{Physical AGI}{\emph{Physical Artificial General Intelligence}. The goal of general intelligence for an open physical world, requiring systematic generalization, long-term adaptation, and reliable action.}{1,10,12}
\Term{Physical AI}{\emph{Physical Artificial Intelligence}. Artificial-system capability to perceive physical states, represent relevant relationships and action consequences, and select and execute interventions through feedback under physical, operational, and computational constraints.}{1,2,3,4,5,6,7,8,9,10,11,12}
\Term{Physical intelligence}{Intelligent or adaptive capability arising through physical interaction or the organization of bodies and materials. Its scope follows the source and context.}{1,8,9}
\Term{Physical RSI}{\emph{Physical Recursive Self-Improvement}. The research framework cited in Chapter~\ref{ch:general-physical-ai} for using physical interaction to persistently update both a robot's task capabilities and its improvement procedure, with the aim of making subsequent improvement more effective across tasks.}{10}
\Term{PID}{\emph{Proportional-Integral-Derivative}. Feedback using proportional, integral, and derivative error terms.}{2}
\Term{PINN}{\emph{Physics-Informed Neural Network}. A neural model trained using data together with physical equations or residual constraints.}{2,5}
\Term{PMSM}{\emph{Permanent-Magnet Synchronous Motor}. A permanent-magnet motor operating synchronously with its rotating magnetic field, commonly using sinusoidal currents and vector control.}{9}
\Term{Policy}{A rule or model selecting actions from observations, states, histories, or goals.}{1,2,3,4,5,6,7,8,9,10,11,12}
\Term{POMDP}{\emph{Partially Observable Markov Decision Process}. A sequential-decision model in which states are inferred indirectly from observations.}{2,5,6}
\Term{Post-training}{Training that adapts general priors using task data, interaction outcomes, or corrections.}{4,5,6,11}
\Term{PPO}{\emph{Proximal Policy Optimization}. A policy-gradient method whose clipped surrogate objective discourages excessively large updates. Clipping is not a hard bound on the realized policy change.}{2,6,7}
\Term{Pre-training}{Training on broad data to acquire reusable representations or behavioral priors.}{1,2,4,5,6,7,9,11}
\Term{Privacy by Embodiment}{A design principle used here to incorporate privacy into sensing and actuation, including effects on bystanders.}{11}
\Term{Proprioception}{Observation of the robot's own configuration, motion, and related internal states.}{1,2,3,5,6,7,8,9}
\Term{QoS}{\emph{Quality of Service}. Communication policies specifying properties such as reliability, history, deadlines, and liveliness.}{9}
\Term{Quasi-direct drive}{Actuation combining high-torque motors with low-ratio transmission to balance torque, backdrivability, and impact response.}{7,9}
\Term{Real-time}{\emph{Real-Time Execution}. Execution within specified sensing, computation, communication, and control deadlines.}{1,2,4,5,6,7,9,11}
\Term{Recovery}{\emph{Failure Recovery}. Correction of behavior after failure to preserve objectives and restore task progress. Physical consequences are not necessarily reversible.}{1,2,4,5,6,7,8,9,10,11,12}
\Term{Reliability}{Provision of specified physical outcomes over a stated operating period and set of conditions.}{1,3,4,5,7,9,10,11,12}
\Term{Retargeting}{\emph{Motion Retargeting}. Mapping motion from a person or another embodiment to a target robot's kinematic and movement spaces.}{4,6,7,9}
\Term{Reward hacking}{Exploitation of reward or evaluator errors to improve a score without achieving the intended physical outcome.}{6}
\Term{RFT}{\emph{Reinforcement Fine-Tuning}. Policy fine-tuning using execution outcomes and rewards. It is a method within the broader reinforcement-learning-based post-training stage.}{6}
\Term{RGB}{\emph{Red, Green, Blue}. The three color channels used here to represent color-image observations.}{3,4,5,6,7,9}
\Term{RGB-D}{\emph{Red, Green, Blue plus Depth}. Color-image observations with associated depth measurements.}{3,4,6,7,9}
\Term{RL}{\emph{Reinforcement Learning}. Learning sequential decisions from interaction-dependent rewards or returns.}{1,2,4,5,6,7,9,10,12}
\Term{RLDS}{\emph{Reinforcement Learning Datasets}. A data format organizing interaction into episodes and steps, with explicit boundary and termination semantics.}{4}
\Term{RMSE}{\emph{Root Mean Square Error}. The square root of mean squared prediction error.}{5}
\Term{RMSNorm}{\emph{Root Mean Square Layer Normalization}. Feature normalization by root mean square without subtracting the feature mean.}{7}
\Term{Robustness}{Maintenance of specified performance under modeling errors, disturbances, or changing conditions.}{1,2,5,7,11}
\Term{ROS 2}{\emph{Robot Operating System 2}. A robotics software framework providing modular components, communication interfaces, and tools for composing robot systems.}{9,11}
\Term{RS485}{\emph{RS-485}. A differential serial electrical-interface standard. It does not specify robot-command semantics.}{9}
\Term{Runtime assurance}{Runtime monitoring and intervention that maintain specified operating conditions under stated assumptions.}{8,9,11}
\Term{RVQ}{\emph{Residual Vector Quantization}. Multi-stage quantization of successive residuals using vector codebooks.}{6}
\Term{Safety}{\emph{Physical Safety}. Control of harm and unacceptable physical consequences within specified limits.}{1,2,3,4,5,6,7,8,9,10,11,12}
\Term{Scaling law}{An empirical relationship between training scale, model capacity, data, and performance under stated conditions.}{4}
\Term{Scene graph}{A representation of entities and their spatial, semantic, or interaction relationships.}{3,5}
\Term{SDF}{\emph{Signed Distance Function/Field}. A geometric representation whose signed distance is zero at the represented surface. The sign convention is defined locally.}{2,3}
\Term{SDK}{\emph{Software Development Kit}. A collection of software libraries, interfaces, and examples for development.}{9}
\Term{SDL}{\emph{Self-Driving Laboratory}. A laboratory closing the loop between experimental design, execution, measurement, and subsequent decisions.}{9}
\Term{SE(3)}{\emph{Special Euclidean Group in Three Dimensions}. The group of three-dimensional rigid transformations combining rotation and translation.}{2,3,7}
\Term{Sensor fusion}{\emph{Multimodal Sensor Fusion}. Combination of sensor information using compatible temporal and spatial reference conventions.}{1,3,7,9}
\Term{Series elastic actuation}{Actuation with an elastic element in the transmission path, supporting torque estimation and impact absorption.}{7,9}
\Term{SFT}{\emph{Supervised Fine-Tuning}. Adaptation of a model or policy using supervised data. It is a method within the broader supervised post-training stage.}{6}
\Term{Shared autonomy}{Allocation of task or action responsibilities between human guidance and autonomous robot control.}{6,9}
\Term{SIL}{\emph{Software-in-the-Loop}. Testing that connects executable software to a simulated plant without target control hardware.}{9}
\Term{Sim-to-real}{\emph{Simulation-to-Real Transfer}. Transfer of policies or capabilities learned or evaluated in simulation to physical systems.}{1,4,7,11}
\Term{Skill contract}{An interface specification for a skill's inputs, preconditions, authority, limits, timing, feedback, and termination.}{9}
\Term{SLAM}{\emph{Simultaneous Localization and Mapping}. Joint estimation of an agent's pose and a map of an initially unknown environment.}{2}
\Term{SO(3)}{\emph{Special Orthogonal Group in Three Dimensions}. The matrix group representing three-dimensional rotations.}{2,7}
\Term{SR}{\emph{Success Rate}. The fraction of trials satisfying a specified success criterion. Evaluation trials and adaptation demonstrations are different counts.}{1,4,5,6,7,8,9,10,11}
\Term{SSM}{\emph{State-Space Model}. A temporal model defined through latent-state evolution and observation generation.}{2,5}
\Term{Stability}{A property of behavior near an equilibrium or reference trajectory under specified initial conditions and disturbances.}{1,2,4,6,7,8,9,11,12}
\Term{State estimation}{Inference of unobserved physical states using measurements, history, and models.}{1,2,3,5,6,7,8,9,11,12}
\Term{STRIPS}{\emph{Stanford Research Institute Problem Solver}. A classical planning system and representation based on action preconditions and effects.}{8}
\Term{System identification}{Estimation of a system's model, parameters, or dynamics from inputs and observations.}{4,5,7,11}
\Term{Systematic generalization}{Recombination, transfer, and application of capabilities across task, environment, object, and embodiment changes.}{10}
\Term{Tactile sensing}{Sensing of contact deformation, pressure, shear, or slip to support contact-state inference.}{1,3,4,6,7,9}
\Term{TAMP}{\emph{Task and Motion Planning}. Joint reasoning about discrete task actions and continuous geometric and physical feasibility.}{8}
\Term{Taxel}{A discrete tactile sensing element with a specified sensitive region and a measured response to local contact.}{3}
\Term{Teacher forcing}{Training with ground-truth preceding inputs rather than exclusively with a model's own preceding predictions.}{5}
\Term{Teleoperation}{Human operation of a robot through remote or mapped control interfaces with feedback.}{1,4,6,7,8,9,10}
\Term{Thing Description}{\emph{Web of Things Thing Description}. A standardized description of a device's interaction affordances within the Web of Things.}{8}
\Term{Twist}{A six-dimensional representation of rigid-body velocity comprising linear and angular velocity, with a specified coordinate frame and reference point.}{7}
\Term{UMI}{\emph{Universal Manipulation Interface}. A handheld-gripper demonstration interface combining visual tracking, relative action trajectories, and latency matching for transfer to compatible robot systems.}{4,6,9}
\Term{Uncertainty}{\emph{Predictive Uncertainty}. Representation of uncertainty in predicted outcomes, states, or models, distinguishing observational variability from limited knowledge where applicable.}{1,2,3,4,5,6,7,8,9,10,11,12}
\Term{Underactuation}{A condition in which not all degrees of freedom are directly driven by independent actuators.}{7}
\Term{URDF}{\emph{Unified Robot Description Format}. A robot-description format representing links, joints, geometry, and associated properties.}{9}
\Term{USB}{\emph{Universal Serial Bus}. An interface standard for connecting devices and transferring data.}{9}
\Term{VAE}{\emph{Variational Autoencoder}. A probabilistic encoder-decoder model with latent variables.}{5,6}
\Term{ViT}{\emph{Vision Transformer}. A Transformer architecture processing image patches as tokens.}{6}
\Term{VLA}{\emph{Vision-Language-Action Model}. A model combining visual, language, and action processing to generate robot actions from observations and instructions.}{2,3,4,6,9}
\Term{VLM}{\emph{Vision-Language Model}. A model jointly processing visual and textual information. A visual-language evaluator does not thereby acquire physical action capability.}{3,4,5,6,8,11}
\Term{VR}{\emph{Virtual Reality}. An immersive visual and tracked-input interface used here for teleoperation.}{9}
\Term{WAM}{\emph{World-Action Model}. A model coupling action generation with prediction of future observations or states. The category can overlap with VLA models.}{2,5,6}
\Term{WBC}{\emph{Whole-Body Control}. Coordination of whole-body motion and contact forces under robot dynamics and task constraints.}{7}

\Term{World model}{An internal model of task-relevant physical state and action consequences. Visual realism alone does not establish predictive physical validity.}{1,2,3,4,5,6,8,9,10,11,12}
\Term{Wrench}{A six-dimensional representation of force and moment, expressed in a specified coordinate frame with the moment taken about a specified reference point.}{2,3,7}
\Term{Zero-shot}{\emph{Zero-Shot Transfer/Execution}. Execution under stated target conditions without new target demonstrations or parameter updates.}{4,6,7,10}
\end{longtable}
\endgroup

\clearpage
\section*{Notation}
\addcontentsline{toc}{section}{Notation}
Semantic subscripts and superscripts distinguish different roles. Bold type denotes explicitly defined vectors or matrices. Abstract states, histories, functions, and model parameters retain their specified types. Descriptive labels, coordinate-frame labels, and units are upright. Variable indices are italic. Reused dummy indices and generic function or parameter notation are defined at their point of use. Chapter references identify the relevant definitions.
\begingroup
\small
\setlength{\tabcolsep}{4pt}
\renewcommand{\arraystretch}{1.15}

\newcommand{\Sym}[3]{\(#1\) & #2 & \PAIScope{#3}\\}
\begin{longtable}{@{}>{\raggedright\arraybackslash}p{.26\textwidth}
  >{\raggedright\arraybackslash}p{\dimexpr.64\textwidth-4\tabcolsep\relax}
  >{\centering\arraybackslash}p{.10\textwidth}@{}}
\toprule
\textbf{Symbol} & \textbf{Meaning} & \textbf{Chapters}\\
\midrule
\endfirsthead
\toprule
\textbf{Symbol} & \textbf{Meaning (continued)} & \textbf{Chapters}\\
\midrule
\endhead
\midrule
\multicolumn{3}{r}{\emph{Continued on the next page}}\\
\endfoot
\bottomrule
\endlastfoot
\Sym{A_t^{\mathrm{seq}},\allowbreak \mathbf A_t^{\mathrm{seq}}}
{Action sequence or candidate plan. In Chapter~\ref{ch:manipulation-models}, the explicitly matrix-valued action chunk has shape $H\times d$, with one $d$-dimensional action per row.}{5,6}

\Sym{A^{\mathrm{adv},\pi}(h,a),\allowbreak \widehat A_t^{\mathrm{adv}}}
{Policy advantage conditioned on a history (or a state in the fully observed case), and its estimate.}{2}

\Sym{\mathbf A}
{Linear state-transition matrix.}{2}

\Sym{A_i,\allowbreak \mathcal A_i}
{Taxel area and sensitive region, with $A_i=\int_{\mathcal A_i}\mathrm dA$.}{3}

\Sym{\mathbf A_{\mathrm G}}
{Centroidal momentum matrix mapping generalized velocity to momentum.}{7}

\Sym{a_t,\allowbreak \mathbf a_t}
{Action. Bold type denotes an explicitly vector-valued action. Superscripts $\mathrm{cmd}$ and $\mathrm{exec}$ distinguish commands and execution.}{2,4,5,6,7,8,11}

\Sym{a_i^{\mathrm{lidar}}}
{LiDAR return intensity.}{3}

\Sym{a_{\mathrm{scale}}}
{Positive coefficient in a scaling relationship.}{4}

\Sym{\Delta_{\mathrm{age}},\allowbreak \Delta_{\mathrm{age}}(t),\allowbreak \Delta_{\mathrm{age}}^{\max}}
{Age of the observation used at a specified decision or actuation event, its time-indexed form, and the maximum permitted age. Distinct from standard AoI.}{6,9}

\Sym{\mathbf a_{\mathrm{imu}}}
{IMU specific-force vector.}{3}

\Sym{\mathbf a_k^{\mathrm{des}}}
{Desired task-space acceleration.}{7}

\Sym{a_0^{\mathrm{swing}},\allowbreak \ldots,\allowbreak a_5^{\mathrm{swing}}}
{Swing-trajectory polynomial coefficients.}{7}

\Sym{\mathbf B}
{Linear control-input matrix.}{2}

\Sym{B_{\mathrm{stereo}}}
{Stereo-camera baseline length.}{3}

\Sym{N_{\mathrm{buf}}}
{Number of actions in an action buffer.}{6}

\Sym{N_{\mathrm{roll}}}
{Number of candidate prediction rollouts.}{5}

\Sym{N_{\mathrm{point}},\allowbreak H_{\mathrm{MPC}}}
{Point-cloud point count and discrete MPC prediction horizon, respectively.}{2}

\Sym{B_{\mathrm{emb}}}
{An embodiment or physical platform.}{8,9,11}

\Sym{B_{\mathrm{cab}}}
{Cabinet object in the placement example.}{8}

\Sym{B_{\mathrm{max}}}
{Worst-case blocking time.}{9}

\Sym{\mathcal B,\allowbreak \mathrm B}
{Base coordinate frame, using the local frame convention.}{7,8}

\Sym{\mathbf B^{\mathrm{seq}}}
{An interleaved body-state and action sequence.}{7}

\Sym{b_t,\allowbreak \widetilde b_t}
{State belief and predictive belief, respectively.}{2,5,8,11}

\Sym{\zeta_t^{\mathrm{state}}}
{Optional state-information field in an interaction record. It may store a state estimate or a belief, with its representation declared locally.}{4}

\Sym{b_{\mathrm{scale}}}
{Positive coefficient in a scaling relationship.}{4}

\Sym{b}
{Candidate-rollout index.}{5}

\Sym{\mathbf b}
{Generalized dynamics bias, including Coriolis, centrifugal, and gravitational terms.}{7}

\Sym{\mathbf C(\mathbf q,\dot{\mathbf q})}
{Coriolis and centrifugal matrix.}{2}

\Sym{C(s)}
{Controller transfer function.}{2}

\Sym{\mathbf C_{\mathrm{obs}}}
{Linear observation matrix.}{2}

\Sym{\mathbf C_{\mathrm{cal}}}
{Force/torque sensor calibration matrix.}{3}

\Sym{C_{\mathrm{train}}}
{Training compute, with the local approximation $C_{\mathrm{train}}\approx6N_{\mathrm{par}}N_{\mathrm{data}}$.}{4}

\Sym{C_{\mathrm{contact}}(z),\allowbreak C_{i,t}^{\mathrm{contact}}}
{Smooth gait-contact weighting function and its evaluated value.}{7}

\Sym{C_t^{\mathrm{run}}}
{Nonnegative per-interval resource cost, such as time or energy, in the planning utility.}{5}

\Sym{C_i^{\mathrm{safe}}}
{Physical constraint residual depending on a predicted state trajectory and candidate actions. Feasibility is $C_i^{\mathrm{safe}}\leq0$.}{11}

\Sym{C_{\mathrm{max}}}
{Worst-case computation time.}{9}

\Sym{C_{\mathrm{valid}}}
{Cost per valid success. Semantic subscripts identify constituent costs.}{9}

\Sym{c,\allowbreak c_t}
{Scene configuration or task conditions.}{4,5}

\Sym{c_{\mathrm{light}}}
{Speed of light under the stated propagation convention.}{3}

\Sym{\mathbf c_{\mathrm{rgb}},\allowbreak i_{\mathrm{rgb}}}
{Radiance-field color vector and RGB channel index, respectively.}{2,3}

\Sym{c_{\mathrm{acq}}(x)}
{Data-acquisition cost.}{4}

\Sym{c_{\mathrm{lim}}}
{Fitted limiting length tolerance in precision scaling. The fitted regime has $P_{\mathrm{tol}}>c_{\mathrm{lim}}$.}{4}

\Sym{c_t^{\mathrm{constr}}}
{Execution-constraint context available at time $t$.}{8}

\Sym{c_{\mathrm{cont}}}
{Container object in the cabinet-placement example.}{8}

\Sym{\mathbf c_t,\allowbreak \mathbf c_t^{\mathrm{beh}}}
{Motor-command vector and its behavior-command component.}{7}

\Sym{c_k^{\mathrm{ctx}}}
{Cached slow-context features.}{6}

\Sym{c_j,\allowbreak c_t^{\mathrm{safety}},\allowbreak c_t^{\mathrm{energy}}}
{Regularization, safety, or energy cost terms under their local objectives.}{6,7}

\Sym{\mathbf c_t^{\mathrm{root}}}
{Horizontal root translation defining a local motion-evaluation origin.}{7}

\Sym{\mathbf c_{\mathrm G}}
{Center-of-mass position.}{7}

\Sym{\mathbf d,\allowbreak \mathbf d_i}
{Viewing direction or LiDAR unit-ray direction.}{2,3}

\Sym{d_{\mathrm{node}},\allowbreak d_{\mathrm{edge}},\allowbreak d_{\mathrm{msg}}}
{Positive integer dimensions of graph-node, graph-edge, and message vectors, respectively. These dimensions are not required to be equal.}{2}

\Sym{d_{\mathrm{state}},\allowbreak d_{\mathrm{noise}}}
{Positive integer state and Wiener-noise dimensions in neural differential equations. These dimensions are not required to be equal.}{2}

\Sym{d_{\mathrm{task},k}}
{Positive integer dimension of task $k$'s acceleration residual in WBC.}{7}

\Sym{d_{\mathrm{stereo}}(\mathbf u_{\mathrm{pix}})}
{Stereo disparity at a pixel-coordinate vector.}{3}

\Sym{c_{\mathrm x},\allowbreak c_{\mathrm y}}
{Horizontal and vertical principal-point coordinates in the calibrated camera image.}{3}

\Sym{d_{\mathrm{delay}}}
{Control delay measured in discrete steps.}{6}

\Sym{d_{\mathrm{feat}},\allowbreak d_k}
{Feature-distance function and indexed squared tracking discrepancy, respectively.}{5,7}

\Sym{d_{\mathrm R}(\mathbf R_1,\mathbf R_2)}
{Geodesic angular distance between two rotation matrices, used for orientation-tracking error.}{7}

\Sym{i_{\mathrm{src}}}
{Training-data-source index.}{5}

\Sym{d,\allowbreak d_{\mathrm D},\allowbreak d_{\mathrm r}}
{Feature, action, or target dimensionality.}{5,6}

\Sym{d_{t+1}^{\mathrm{skill}}}
{Skill-selection or execution-supervision decision.}{8}

\Sym{d_k^{\mathrm{detect}}}
{Failure-detection probability.}{9}

\Sym{d_t^{\mathrm{stance}}}
{Gait stance duty factor, with $0<d_t^{\mathrm{stance}}<1$.}{7}

\Sym{d_{\mathrm{total}}}
{Perception-to-actuation delay budget with explicitly defined stage boundaries.}{7}

\Sym{N_{\mathrm{data}}}
{Training-data count, distinct from a dataset object $\mathcal D$.}{4}

\Sym{N_{\mathrm{src}}}
{Number of training-data sources.}{5}

\Sym{D_t,\allowbreak D_{t,i}}
{Stacked clean-latent change targets and the target for future slice $i$, respectively. Each slice retains the declared latent representation. Here $i$ does not index an individual scalar coordinate.}{5}

\Sym{T_{\mathrm{deadline}}}
{Exclusive deadline for the timing path or periodic local update defined in context. The remote closed-loop model requires $T_{\mathrm{e2e}}<T_{\mathrm{deadline}}$; the local update budget must also finish strictly before its stated deadline. The values are specified for each timing model.}{9}

\Sym{D_{\mathrm{KL}}}
{Kullback-Leibler divergence.}{5}

\Sym{\mathbf E}
{Disturbance-input matrix in a state-space model.}{2}

\Sym{E_\phi}
{Encoder. Similarly named encoders in different chapters are distinct learned functions.}{5}

\Sym{E_\lambda(\mathbf u_{\mathrm{pix}},\tau)}
{Spectral irradiance.}{3}

\Sym{\mathcal E,\allowbreak E_{\mathrm{phys}},\allowbreak E_{\mathrm{info}}}
{Graph edge sets, including physical and information edges.}{2,8}

\Sym{\mathrm E}
{End-effector coordinate-frame label.}{8}

\Sym{e_{\mathrm{track}}(t),\allowbreak \mathbf e,\allowbreak e_{\mathrm{ss}},\allowbreak e_{\mathrm{contact}}}
{Scalar tracking-error signal, vector tracking or pose error, steady-state error, and contact-direction position error, respectively.}{2,7,9}

\Sym{e,\allowbreak e^\star}
{Embodiment identifier and target embodiment.}{6}

\Sym{e_t^{\mathrm{event}}}
{Recorded trajectory event.}{4}

\Sym{e_i^{\mathrm{obs}}}
{Binary observed event outcome used in the Brier score.}{5}

\Sym{\mathbf e_{ij}^{\mathrm{edge}}}
{Graph-edge feature vector in $\mathbb R^{d_{\mathrm{edge}}}$, with node indices $i,j$.}{2}

\Sym{e_t^{\mathrm{enc}}}
{Encoded observation features in $e_t^{\mathrm{enc}}=f_\theta(o_t)$, an abstract representation unless vector coordinates are explicitly specified.}{2}

\Sym{f,\allowbreak f_\theta,\allowbreak f_{\mathrm{c}},\allowbreak f_{\mathrm{d}}}
{Locally defined mappings, dynamics, or encoders. The subscripts $\mathrm{c}$ and $\mathrm{d}$ distinguish continuous- and discrete-time dynamics; $\theta$ denotes model parameters.}{2,6,11}

\Sym{\mathbf f_\theta^{\mathrm{drift}}(\mathbf x,t)}
{Vector-valued drift mapping from $\mathbb R^{d_{\mathrm{state}}}\times[0,\infty)$ to $\mathbb R^{d_{\mathrm{state}}}$ in neural differential equations.}{2}

\Sym{\mathbf f_i,\allowbreak \mathbf f_{\mathrm n},\allowbreak \mathbf f_{\mathrm t},\allowbreak f_{i,\mathrm n},\allowbreak \mathbf f_{i,\mathrm t},\allowbreak \mathbf f_{\mathcal K_{\mathrm{hand}}}}
{Contact-force vectors and their normal or tangential vector components. In the point-contact decomposition, $f_{i,\mathrm n}$ is the scalar normal projection and $\mathbf f_{i,\mathrm t}$ is the tangential force vector. $\mathbf f_{\mathcal K_{\mathrm{hand}}}$ stacks the active finger-object contact forces.}{2,7}

\Sym{f_j^{\mathrm{eq}}}
{Physical equality-constraint residual, with feasibility $f_j^{\mathrm{eq}}=0$.}{5}

\Sym{f_t^{\mathrm{gait}}}
{Gait frequency.}{7}

\Sym{\mathbf f_{t,i_{\mathrm{tok}}}^{+,\ell_{\mathrm T}}}
{Aligned target feature at time, position, and target-layer indices.}{5}

\Sym{f_{\mathrm x},\allowbreak f_{\mathrm y}}
{Camera focal lengths in image-axis units.}{3}

\Sym{F,\allowbreak F_\theta}
{State-transition or neural prediction mapping.}{2}

\Sym{F_\theta^{\mathrm{NeRF}}}
{Radiance-field mapping to color and density.}{2,3}

\Sym{F_{\mathrm{ISP}}}
{Image-signal-processing mapping.}{3}

\Sym{F_t^{\mathrm{future}}}
{Future-observation sequence in joint world-action modeling.}{5}

\Sym{F_{\mathrm{CDF}}}
{Predictive cumulative distribution used in CRPS.}{5}

\Sym{F_{\mathrm{contact}}}
{Scalar contact force.}{9}

\Sym{F_\theta^{\mathrm{ctx}}}
{Slow-context module in a multirate policy.}{6}

\Sym{F_{\mathrm{plan}},\allowbreak c_t^{\mathrm{plan}}}
{Autoregressive motion generator and its planning condition, respectively. The condition may include execution feedback and is not assumed to be a motor-command vector.}{7}

\Sym{G,\allowbreak \mathcal G_{\mathrm{phys}},\allowbreak \mathcal G_{\mathrm{info}}}
{Graph structures, including physical and information graphs.}{2,8}

\Sym{G_{\mathrm{tac}}}
{Optical-tactile mapping from an image to estimated contact state.}{3}

\Sym{G_\psi}
{Fast action-decoding module in a multirate policy.}{6}

\Sym{\mathbf G}
{Grasp matrix mapping contact forces to an object wrench.}{7}

\Sym{g,\allowbreak g_t}
{Task goal or command condition. Chapters~\ref{ch:physical-data}, \ref{ch:wm}, and~\ref{ch:physical-agent} use $g$ for a goal. Chapter~\ref{ch:manipulation-models} may use $g_t$ for a subgoal. In Chapter~\ref{ch:whole-body-motor-control}, $g_t$ contains behavior or reference-derived motor commands.}{4,5,6,7,8}

\Sym{\mathbf g(\mathbf q)}
{Generalized gravitational force in the robot dynamics.}{2,7}

\Sym{\mathbf g_{\mathrm{acc}}}
{Gravitational acceleration expressed in world coordinates in the centroidal momentum balance, measured in $\mathrm m\,\mathrm s^{-2}$.}{7}

\Sym{g_{\mathrm{in}}(x)}
{Input vector field in the locally defined control-affine model.}{2}

\Sym{\mathbf G_\theta^{\mathrm{diff}}(\mathbf x,t)}
{Diffusion mapping with values in $\mathbb R^{d_{\mathrm{state}}\times d_{\mathrm{noise}}}$, multiplying the Wiener increment.}{2}

\Sym{g_{\mathrm{group}}}
{A group element acting on an input or output.}{2}

\Sym{g_i^{\mathrm{taxel}}}
{Taxel transduction or calibration function.}{3}

\Sym{g_j^{\mathrm{ineq}}}
{Physical inequality-constraint residual, with feasibility $g_j^{\mathrm{ineq}}\leq0$.}{5}

\Sym{g_{\psi_e}}
{Embodiment-specific action decoder.}{6}

\Sym{g(x_T),\allowbreak g_f,\allowbreak g_v}
{Terminal success predicate or contact-related reward gates, according to context.}{7,8}

\Sym{g_\phi}
{JEPA feature-prediction mapping.}{2}

\Sym{H,\allowbreak H_{\mathrm a},\allowbreak H_{\mathrm o}}
{Prediction horizon in steps. The world-model discussion also uses $H_{\mathrm a}$ and $H_{\mathrm o}$ for action and observation horizons.}{5,6,7,11}

\Sym{H_\infty}
{The robust-control or function-space designation, distinct from a prediction horizon.}{2}

\Sym{H_\theta}
{Latent-state update mapping.}{2}

\Sym{\mathbf H_t}
{Heading rotation used in local motion evaluation.}{7}

\Sym{\mathbf h_i^{\mathrm{node},t}}
{Graph-node feature vector in $\mathbb R^{d_{\mathrm{node}}}$, distinct from observation-action history $h_t$ and centroidal momentum $\mathbf h_{\mathrm G}$.}{2}

\Sym{h_t}
{Observation-action history, including locally specified robot-state inputs.}{2,5,6,7}

\Sym{\mathbf H_t^{\mathrm{VLM}}}
{Vision-language backbone token-feature matrix supplied to the action expert, distinct from the interaction history $h_t$.}{6}

\Sym{h_{\mathrm{obs}}(x)}
{Observation function.}{2}

\Sym{h_{\mathrm{safe}}(x)}
{Control-barrier function defining a safe set.}{2}

\Sym{\mathbf h_{\mathrm G}}
{Total centroidal momentum, comprising linear and angular momentum.}{7}

\Sym{\mathbf I,\allowbreak \mathbf I_p}
{Identity matrix, with order specified by context or subscript.}{2,5}

\Sym{I_t,\allowbreak I^{\mathrm{RGB}},\allowbreak I_{\mathrm{tac}}}
{Image observations, including color and tactile images.}{3}

\Sym{I_t^{\mathrm{ext}}}
{Indicator of upper-body external-controller activation.}{7}

\Sym{I_{\mathrm m},\allowbreak I_{\mathrm{refl}}}
{Motor inertia and reflected inertia, with $I_{\mathrm{refl}}=N_{\mathrm{gear}}^2I_{\mathrm m}$ under the stated transmission convention.}{9}

\Sym{i}
{Local sample, node, contact, or constraint index.}{2,3}

\Sym{\mathbf i,\allowbreak i_{\mathrm q}}
{Motor-current vector and quadrature-axis current.}{3,9}

\Sym{i:j}
{Index interval or segment endpoints.}{5}

\Sym{J_{\mathrm{cost}},\allowbreak J(\pi),\allowbreak J_{\mathrm{real}}}
{LQR control cost, expected policy return, and measured real-task utility, respectively. Cost is minimized. Return and utility use their stated maximization convention.}{2,4,5,6}

\Sym{\mathbf J,\allowbreak \mathbf J_i,\allowbreak \mathbf J_{\mathcal K_{\mathrm{contact}}},\allowbreak \mathbf J_{w,i},\allowbreak \mathbf J_{\mathrm{ext}},\allowbreak \mathbf J_{\mathrm h,\mathcal K_{\mathrm{hand}}}}
{Robot Jacobians under the stated task, frame, and reference-point convention. $\mathbf J_{\mathcal K_{\mathrm{contact}}}$ stacks active point-contact Jacobians, $\mathbf J_{w,i}$ is the six-dimensional foot-wrench Jacobian used in WBC, and $\mathbf J_{\mathrm h,\mathcal K_{\mathrm{hand}}}$ maps finger joint rates to velocities at the active hand-object contacts. In Chapter~\ref{sec:technical-foundations}, $\mathbf J\in\mathbb R^{6\times n}$ matches the reference point and coordinates of $\mathbf w_{\mathrm{ext}}$.}{2,7}

\Sym{J_{\mathrm{max}}}
{Worst-case scheduling jitter.}{9}

\Sym{J_{\mathrm{eq}},\allowbreak J_{\mathrm{ineq}}}
{Numbers of equality and inequality constraints.}{5}

\Sym{j}
{Local diffusion-step, constraint, target, or source index.}{5}

\Sym{\mathrm j}
{Imaginary unit in frequency-domain control expressions.}{2}

\Sym{k}
{Local timestep, prediction offset, iteration, or task-step index.}{4,5}

\Sym{k(t)}
{Index of the latest available slow-context update.}{6}

\Sym{k_{\mathrm t}}
{Motor torque constant.}{9}

\Sym{k_{\mathrm r},\allowbreak k_{\mathrm e},\allowbreak k_{\mathrm{eff}}}
{Robot, environment, and effective stiffness.}{9}

\Sym{\mathbf k_{\mathrm G}}
{Centroidal angular momentum.}{7}

\Sym{K_{\mathrm P},\allowbreak K_{\mathrm I},\allowbreak K_{\mathrm D},\allowbreak \mathbf K}
{Scalar or matrix feedback gains according to the specified controller.}{2}

\Sym{\mathbf K_{\mathrm{cam}}}
{Camera intrinsic matrix.}{3}

\Sym{H_{\mathrm{roll}},\allowbreak K_b,\allowbreak N_{\mathrm{slice}},\allowbreak N_{\mathrm{acttok}},\allowbreak N_{\mathrm{exec}}}
{Deterministic model-rollout horizon, candidate-branch interval count, future-latent slice count, action-token count, and number of actions executed before replanning, respectively.}{2,5,6}

\Sym{N_{\mathrm{stage}}}
{Number of stages in the task-success product with at most one modeled recovery attempt per stage.}{9}

\Sym{K_{\mathrm{civ}}}
{Kardashev index, $K_{\mathrm{civ}}=(\log_{10}(P_{\mathrm{power}}/\mathrm W)-6)/10$.}{12}

\Sym{K_{\mathrm{eval}}}
{Number of evaluation criteria or scoring dimensions.}{11}

\Sym{L_{\mathrm{mech}}}
{Mechanical Lagrangian, $L_{\mathrm{mech}}=E_{\mathrm{kin}}-E_{\mathrm{pot}}$.}{2}

\Sym{L,\allowbreak L(s)}
{Number of observations in the finite interaction history in Chapter~\ref{ch:whole-body-motor-control}; open-loop transfer function in Chapter~\ref{sec:technical-foundations}, respectively. The argument $s$ is the Laplace-domain variable.}{2,7}

\Sym{L_{\mathrm{scale}},\allowbreak L_{\mathrm{floor}}}
{Validation loss and irreducible loss floor in a scaling relationship.}{4}

\Sym{L_{\mathrm{e2e}}}
{Latency until the command becomes available at the control interface. Actuator response is counted separately.}{6}

\Sym{L_{\mathrm{crit}}}
{Critical dependency-chain length or duration in parallel scheduling.}{8}

\Sym{L_fh_{\mathrm{safe}},\allowbreak L_{g_{\mathrm{in}}}h_{\mathrm{safe}}}
{Lie derivatives of the safety-barrier function along the drift and input vector fields.}{2}

\Sym{L_1,\allowbreak L_2}
{In the action-regression discussion, $L_1$ denotes an elementwise absolute-error loss. In robust control, $L_2$ denotes the finite-energy signal space. Its norm is listed separately.}{2,6}

\Sym{\ell_{\mathrm{lang}}}
{Language task instruction.}{6}

\Sym{\ell(\mathbf{x},\mathbf{u}),\allowbreak \ell_j,\allowbreak \ell(\widehat y,y)}
{MPC stage cost in Chapters~\ref{sec:technical-foundations} and~\ref{ch:whole-body-motor-control}; per-example objective loss or prediction loss in Chapter~\ref{ch:wm}. Arguments follow the local definitions.}{2,5,7}

\Sym{\ell_t^{\mathrm{clear}},\allowbreak \ell_{i,t}^{\mathrm{target}},\allowbreak \ell_{i,t}^{\mathrm{foot}}}
{Maximum foot clearance, target swing height, or realized foot height.}{7}

\Sym{\ell,\allowbreak \ell_{\mathrm s},\allowbreak \ell_{\mathrm T}}
{Network-layer or command-mask component index.}{5,7}

\Sym{m_t^{\mathrm{mem}}}
{Memory state or retained summary of prior interaction.}{8}

\Sym{m_j(\xi),\allowbreak \mathbf m_t,\allowbreak m_t^\ell}
{Supervision-validity mask, command-mask vector, and its scalar components.}{5,7}

\Sym{m_{\mathrm{EMA}}}
{Exponential-moving-average mixing coefficient.}{5}

\Sym{m}
{Control-input dimension in the linear state-space model in Chapter~\ref{sec:technical-foundations}; ensemble-member index in Chapter~\ref{ch:wm}.}{2,5}

\Sym{m_t^{\mathrm{multi}},\allowbreak m_t^{\mathrm{multi}\prime}}
{Additional multimodal inputs to prefix and suffix processing, respectively.}{6}

\Sym{N_{\mathrm{constr}}}
{Number of constraints in a joint safety event.}{11}

\Sym{m_{\mathrm{robot}}}
{Robot mass.}{7}

\Sym{m_{\mathrm{fit}}}
{Positive coefficient in the empirical precision fit $\log N_{\mathrm{demo}}=m_{\mathrm{fit}}/(P_{\mathrm{tol}}-c_{\mathrm{lim}})+n_{\mathrm{fit}}$.}{4}

\Sym{\mathbf M(\mathbf q)}
{Rigid-body mass or inertia matrix.}{2,7}

\Sym{M}
{Number of models in an uncertainty ensemble.}{5}

\Sym{N_{\mathrm{par}}}
{Number of model parameters in a scaling analysis.}{4}

\Sym{N_{\mathrm{demo}},\allowbreak N_{\mathrm{iter}},\allowbreak N_{\mathrm{tok}},\allowbreak N_{\mathrm{step}},\allowbreak N_{\mathrm{eval}}}
{Demonstration count or adaptation budget, DAgger iteration count, sequence-token count, state-transition step count, and evaluation-trial count, respectively. Each count refers to the locally declared dataset or protocol.}{4,5,6,7,11}

\Sym{N_{\mathrm{agent}}}
{Number of cooperating agents or parallel workers.}{8}

\Sym{N_{\mathrm{gear}}}
{Transmission reduction ratio under the stated input/output convention.}{9}

\Sym{\mathcal N(i)}
{Neighbor set of graph node $i$.}{2}

\Sym{N_{\mathrm{valid}}}
{Number of valid successful trials.}{9}

\Sym{N_{\mathrm{rec}},\allowbreak N_{\mathrm{demo}}^\star(s_{\mathrm{min}})}
{Number of recovery-eligible evaluation trials, and the minimum tested demonstration budget meeting the prescribed adaptation-success threshold, respectively.}{6,11}

\Sym{n,\allowbreak n_a}
{Locally defined dimensions. In Chapter~\ref{sec:technical-foundations}, $n$ is the state dimension in the linear state-space model and the joint-coordinate dimension in the robot model. In Chapter~\ref{ch:whole-body-motor-control}, $n$ counts actuated joints, so generalized velocity has dimension $6+n$; $n_a$ is the action dimension in mirror-consistency regularization.}{2,7}

\Sym{\mathbf n_i}
{Unit contact-surface normal.}{7}

\Sym{n_{\mathrm{fit}}}
{Intercept in the empirical precision fit $\log N_{\mathrm{demo}}=m_{\mathrm{fit}}/(P_{\mathrm{tol}}-c_{\mathrm{lim}})+n_{\mathrm{fit}}$.}{4}

\Sym{O_{\psi_{\mathrm{render}}},\allowbreak O_{\omega_{\mathrm{obs}}},\allowbreak O_e}
{Observation model or likelihood.}{4,5,6}

\Sym{O_t^+}
{Future-observation sequence in the manipulation-model discussion.}{6}

\Sym{o_t}
{Observation, with sensor content defined locally.}{2,4,5,6,7,8}

\Sym{\mathbf o}
{Sensor ranging origin.}{3}

\Sym{p,\allowbreak p_\theta,\allowbreak p_{\xi_{\mathrm{sim}}},\allowbreak p_{\eta_{\mathrm{sim}}}}
{Probability density or mass function under the specified measure, or a scalar event probability where explicitly defined. Parameters identify the local probability model.}{2,4,5,6,7,8,9}

\Sym{p_i,\allowbreak p_k}
{Predicted event or step-success probabilities in the stated evaluation or reliability protocol.}{5,9}

\Sym{\mathbf p,\allowbreak \mathbf p_i}
{Spatial position vector. Frame labels specify its coordinates.}{2,3,6}

\Sym{p_{\mathrm{press}}}
{Normal contact pressure.}{3}

\Sym{i_{\mathrm{tok}}}
{Spatial-token position index.}{5}

\Sym{d_{\mathrm{reg}}}
{Regressor dimension or identity-matrix order in persistent-excitation analysis.}{2}

\Sym{p_{\mathrm{quant}}}
{Latency-quantile level.}{6}

\Sym{\mathbf p_{\mathrm G}}
{Centroidal linear momentum, $\mathbf p_{\mathrm G}=m_{\mathrm{robot}}\mathbf v_{\mathrm G}$, measured in $\mathrm{kg}\,\mathrm m\,\mathrm s^{-1}$.}{7}

\Sym{\widehat p_{\mathrm{aut}},\allowbreak \widehat p_{\mathrm{rec}},\allowbreak p_{\mathrm{upper}}}
{Estimated autonomous success, estimated recovery success, or an upper bound on incident probability.}{11}

\Sym{P^{\mathrm{tr}},\allowbreak P_{\xi_{\mathrm{sim}}}^{\mathrm{tr}},\allowbreak P_e^{\mathrm{tr}},\allowbreak P(\cdot\mid s,\mathbf a)}
{State-transition probability kernel. Chapter~\ref{ch:whole-body-motor-control} writes the motion-tracking kernel as $P(\cdot\mid s,\mathbf a)$. Its conditional arguments distinguish it from the plant transfer function $P(s)$.}{2,4,6,7}

\Sym{P(s)}
{Plant transfer function.}{2}

\Sym{P_\theta,\allowbreak P_\theta^{\mathrm q}}
{Feature predictor or future-token readout. The upright superscript $\mathrm q$ labels the predictive-token readout and is distinct from a query-type index.}{5}

\Sym{P_{\mathrm{tol}}}
{Required length tolerance in precision scaling. Smaller values impose tighter requirements.}{4}

\Sym{P_{\mathrm{power}}}
{Total power used in civilization-scale analysis.}{12}

\Sym{P_{\mathrm{task}}}
{Long-task success probability under the stated detection and recovery model.}{9}

\Sym{\mathbf Q,\allowbreak \mathbf Q_v,\allowbreak \mathbf Q_{\mathrm R},\allowbreak \mathbf Q_a}
{Explicitly matrix-valued state or tracking weights in LQR and NMPC.}{2,7}

\Sym{Q_\omega,\allowbreak Q_c}
{Nonnegative scalar NMPC weights for angular-velocity and CoM-height errors, respectively. The definition is $\|e\|_Q^2=Qe^2$ for a scalar error and either scalar weight.}{7}

\Sym{Q_{i_{\mathrm{rgb}}}^{\mathrm{quant}}}
{Sensor-channel quantization function.}{3}

\Sym{Q_\psi}
{World-model task readout.}{5}

\Sym{Q^\pi(h,a)}
{Action-value function of policy $\pi$.}{2}

\Sym{\mathbf q_t}{Proprioceptive measurement vector at time $t$ in the manipulation-model interface. Its contents are specified by the embodiment and are not assumed to be configuration coordinates alone.}{6}

\Sym{\mathbf q,\allowbreak \mathbf q_j}
{Robot configuration and the joint-coordinate subvector, respectively. Configuration derivatives and generalized velocities need not have identical representations.}{2,3,6,7}

\Sym{q_{\mathrm{query}}}
{Query-type index.}{5}

\Sym{q_\phi}
{Variational posterior or approximate latent-state inference density.}{5}

\Sym{q_t,\allowbreak \mathbf q_{t,i_{\mathrm{tok}}}^{\ell_{\mathrm s}}}
{Predicted tokens or auxiliary feature readouts, according to the local model.}{5}

\Sym{\mathrm q}
{Motor quadrature-axis label.}{9}

\Sym{\mathbf q_{\mathrm B},\allowbreak \mathbf q_{\mathrm A}}
{Base and arm configuration vectors.}{8}

\Sym{q_{\mathrm{interp}}(u_{\mathrm{interp}})}
{Quintic smooth interpolation, $q_{\mathrm{interp}}(u_{\mathrm{interp}})=10u_{\mathrm{interp}}^3-15u_{\mathrm{interp}}^4+6u_{\mathrm{interp}}^5$.}{7}

\Sym{\mathbf R}
{Rotation matrix, with frame labels specifying its mapping direction.}{2,3,7}

\Sym{\mathbf R_{\mathrm u}}
{Control-input weighting matrix in LQR and the NMPC objective, positive definite in the stated LQR formulation.}{2,7}

\Sym{R,\allowbreak R_{\ell_{\mathrm{lang}},e}}
{MDP reward function or embodiment-specific reward.}{2,6}

\Sym{R_{\mathrm T},\allowbreak R_\theta^{\mathrm S}}
{Frozen teacher and learned student geometry/motion readouts.}{5}

\Sym{R_{\mathrm{gen}}}
{Action-generation rate, $H/L_{\mathrm{model}}$.}{6}

\Sym{Y_j^{\mathrm{rec}}}
{Binary recovery-trial outcome.}{11}

\Sym{r_t}
{Realized reward.}{2,5,6,7}

\Sym{r(t)}
{Control reference input.}{2}

\Sym{r_i}
{Measured ray range, distinct from optical-axis depth away from the optical axis.}{3}

\Sym{r}
{Correlation coefficient.}{4}

\Sym{r_t^{\mathrm{reason}}}
{Reasoning tokens or text.}{6}

\Sym{\mathbf r_t^{\mathrm T},\allowbreak \mathbf r_t^{\mathrm S}}
{Teacher and student geometry/motion descriptors.}{5}

\Sym{r_k^{\mathrm{rec}}}
{Failure-recovery success probability.}{9}

\Sym{\mathcal S_{\mathrm{surf}}}
{Implicit geometric surface.}{2}

\Sym{S(s)}
{Closed-loop sensitivity function.}{2}

\Sym{\widehat S}
{Predicted physical-state sequence.}{5}

\Sym{\mathbf S}
{Floating-base actuator-selection matrix.}{7}

\Sym{S_{i_{\mathrm{rgb}}}(\lambda)}
{Sensor-channel spectral response.}{3}

\Sym{\mathrm S,\allowbreak \mathrm S_m,\allowbreak \mathrm S_0}
{Sensor-frame labels.}{3}

\Sym{S_{\mathrm{eval}}}
{Optional weighted aggregate evaluation score on a $0$ to $100$ scale.}{11}

\Sym{s_t,\allowbreak \hat s_t}
{Physical or task state and its estimate. Vector notation applies when coordinates are explicitly specified.}{2,4,5,6,7}

\Sym{s}
{Complex Laplace-domain frequency variable.}{2}

\Sym{s_1^{\mathrm{shear}},\allowbreak s_2^{\mathrm{shear}}}
{Tangential contact-stress components.}{3}

\Sym{s_k^{\mathrm{eval}}}
{Normalized score for one evaluation criterion.}{11}

\Sym{s_{\mathrm F},\allowbreak s_{\mathrm V}}
{Positive scale parameters in contact penalties.}{7}

\Sym{s_t^{\mathrm{ref}}}
{Reference motion state.}{7}

\Sym{s_{\mathrm{min}}}
{Minimum acceptable adaptation success rate.}{6}

\Sym{\mathrm{SO}(3)}
{Three-dimensional rotation group.}{2,7}

\Sym{\mathrm{SE}(3)}
{Three-dimensional rigid-transformation group.}{2,3,7}

\Sym{se(3)}
{Lie algebra of $\mathrm{SE}(3)$.}{2}

\Sym{t,\allowbreak t_{\mathrm{obs}},\allowbreak t_{\mathrm{apply}}}
{Time, discrete timestep, or explicitly defined observation and actuation timestamps. Event boundaries and clock domains are stated locally.}{2,5,6,9}

\Sym{\mathbf t}
{Translation vector.}{2,3}

\Sym{\mathrm t}
{Tangential-direction label on a contact-force component.}{2}

\Sym{t_m,\allowbreak t_0}
{Local sensor-clock reading and corresponding reference-clock reading in the affine clock-mapping model.}{3}

\Sym{\mathbf T,\allowbreak {}^{\mathrm F}\mathbf T_{\mathrm H}}
{Rigid-pose or homogeneous-transform matrix mapping coordinates from frame $\mathrm H$ to frame $\mathrm F$.}{2,3,6,8}

\Sym{E_{\mathrm{kin}}}
{Kinetic energy.}{2}

\Sym{T_{\mathrm{temp}}(t)}
{Temperature.}{3}

\Sym{T(s),\allowbreak \mathbf T_{zw}}
{Complementary-sensitivity or disturbance-to-performance transfer function.}{2}

\Sym{T,\allowbreak T_\theta}
{State-transition probability kernel in a world model.}{5}

\Sym{T,\allowbreak T_i,\allowbreak T_c,\allowbreak T_{\mathrm{task}}}
{Episode duration or step count as defined locally. In Chapter~\ref{ch:whole-body-motor-control} mean tracking errors, $T$ counts actually executed evaluated frames, and $T_i$ is the full reference length of sequence $i$ in frames. $T_c$ and $T_{\mathrm{task}}$ denote control period and task-completion time.}{4,7,9}

\Sym{T_\tau}
{Actuator torque-response time constant.}{7}

\Sym{T_0}
{Positive integration-window duration in the persistent-excitation condition.}{2}

\Sym{T_{\mathrm{e2e}}}
{Latency from sensing start to actuation onset for one observation-action path in the serial model. Stage intervals cover this path without gaps or overlaps and include queueing time.}{9}

\Sym{u_t,\allowbreak \mathbf u_t}
{Control input. Vector notation follows the local control space.}{2,8}

\Sym{u_{\mathrm{pix}},\allowbreak \mathbf u_{\mathrm{pix}},\allowbreak \tilde{\mathbf u}_{\mathrm{pix}}}
{Pixel coordinate, pixel-coordinate vector, or homogeneous image coordinates.}{3}

\Sym{u_\theta^{\mathrm{field}}(x,t)}
{Neural approximation to a partial-differential-equation solution field.}{2}

\Sym{u}
{Dummy variable of integration or a scalar argument of the positive-part function.}{5}

\Sym{u_{\mathrm{dev}}}
{Upper bound on the increase in the expert's remaining cost after one deviating action in DAgger analysis.}{4}

\Sym{u_t^{\mathrm{vid}}}
{Future-video latent representation.}{5}

\Sym{u_{\mathrm{interp}},\allowbreak u_{\mathrm{fold}}}
{Normalized swing-interpolation or folded-phase parameter.}{7}

\Sym{U}
{Utility of a candidate action plan.}{5}

\Sym{U_t^+}
{Future-video latent sequence.}{5}

\Sym{E_{\mathrm{pot}}}
{Potential energy.}{2}

\Sym{V(x)}
{Lyapunov candidate function.}{2}

\Sym{V_{\mathrm{f}}}
{MPC terminal cost.}{2}

\Sym{V_{\omega_{\mathrm{critic}}}}
{Learned value evaluator.}{6}

\Sym{\mathcal V,\allowbreak V_{\mathrm{phys}},\allowbreak V_{\mathrm{comp}}}
{Graph vertex sets.}{2,8}

\Sym{N_{\mathrm{view}}}
{Number of input camera views.}{6}

\Sym{V^\pi(h)}
{State- or history-value function of policy $\pi$.}{2}

\Sym{\mathbf v}
{Generalized, base, or joint velocity under the specified coordinate convention.}{7}

\Sym{v^{\mathrm{noise}},\allowbreak \mathbf v^{\mathrm{noise}}}
{Measurement noise, with scalar or vector form defined by the observation model.}{2}

\Sym{\mathbf v_{\mathrm{sens}},\allowbreak \mathbf v_{\mathrm{sens},0}}
{Raw force/torque sensor signals and their offsets.}{3}

\Sym{v_{\mathrm{pix}}}
{Vertical pixel coordinate.}{3}

\Sym{\mathbf v_\theta^{\mathrm{flow}}}
{Learned flow-matching velocity field.}{5,6}

\Sym{\mathrm v}
{Video-target label in a loss term.}{5}

\Sym{\mathbf W_k,\allowbreak \mathbf W_{\mathrm f},\allowbreak \mathbf W_\tau}
{Symmetric positive-semidefinite WBC weights for task-acceleration, stacked foot-wrench, and joint-torque residuals, with dimensions $d_{\mathrm{task},k}\times d_{\mathrm{task},k}$, $12\times12$, and $n\times n$, respectively. The symbol $n$ is the number of actuated joints.}{7}

\Sym{W_k^{\mathrm{sim}}}
{Simulator or world model at iteration $k$.}{4}

\Sym{\mathbf W_t^{\mathrm{Wiener}}}
{Standard $d_{\mathrm{noise}}$-dimensional Wiener process in the stochastic state model, distinct from WBC weight matrices and simulator $W_k^{\mathrm{sim}}$.}{2}

\Sym{W_{\mathrm{work}}}
{Total parallel-task work.}{8}

\Sym{\mathcal W}
{World coordinate frame. Upright $\mathrm W$ is used as its frame label in coordinate expressions such as $\mathbf p_{\mathrm W}$, not as a standalone physical variable.}{3,7,8}

\Sym{\mathrm W}
{Watt, the unit of power.}{12}

\Sym{w_t,\allowbreak \mathbf w_t}
{Process disturbance or process noise.}{2}

\Sym{\mathbf w_{\mathrm S}}
{Six-dimensional wrench expressed in a sensor frame.}{3}

\Sym{\mathbf w_{\mathrm{ext}}}
{Six-dimensional external wrench, ordered as force then moment, at the stated reference point and in the same coordinates as its matching Jacobian or body dynamics. Force components are in $\mathrm N$ and moment components in $\mathrm N\,\mathrm m$.}{2,7}

\Sym{\mathbf w_i,\allowbreak \mathbf w_{\mathrm L},\allowbreak \mathbf w_{\mathrm R}}
{Six-dimensional foot wrench $\mathbf w_i=(\mathbf f_i,\boldsymbol\mu_i)$, expressed in world coordinates, with its moment taken about the corresponding foot-frame origin. Labels $\mathrm L$ and $\mathrm R$ denote the left and right feet. Force components are in $\mathrm N$ and moment components in $\mathrm N\,\mathrm m$.}{7}

\Sym{w_j,\allowbreak w_k,\allowbreak w_i}
{Scalar weights for losses, tracking, evaluation criteria, or retargeting.}{5,6,7,9,11}

\Sym{X_{\mathrm C},\allowbreak Y_{\mathrm C},\allowbreak Z_{\mathrm C}}
{Scalar Cartesian coordinates in the camera frame.}{3}

\Sym{X_{t:t+H}}
{Predicted random state trajectory conditioned on the belief and candidate actions.}{11}

\Sym{\mathbf x_{\mathrm{pos}}}
{Spatial position vector.}{2,3}

\Sym{x_t,\allowbreak \mathbf x_t}
{Physical or task state. Bold type denotes an explicitly vector-valued state. The neural differential equations use $\mathbf x_t\in\mathbb R^{d_{\mathrm{state}}}$. Its dimension need not equal the noise dimension.}{2,8}

\Sym{x_i,\allowbreak \mathbf x,\allowbreak \mathbf x^{(j)},\allowbreak \mathbf x^{\sigma_{\mathrm{flow}}}}
{Sequence token, clean continuous target, diffusion-corrupted target, and flow-interpolated target, respectively. The continuous targets retain the shape declared by the local model.}{5}

\Sym{x^\star}
{Data-acquisition decision.}{4}

\Sym{\mathbf x_t^{\mathrm{target}}}
{Target command in masked motion imitation.}{7}

\Sym{y_t,\allowbreak \mathbf y_t,\allowbreak y^{q_{\mathrm{query}}}}
{Measurement, control output, or query-specific prediction target.}{2,3,5}

\Sym{y}
{Recorded terminal physical outcome.}{4}

\Sym{Y,\allowbreak Y'}
{Random outcomes or independent draws under the specified predictive distribution.}{5}

\Sym{Y_{\mathrm C}^{\mathrm{raw}}}
{Raw mosaic image supplied to the image-signal-processing pipeline.}{3}

\Sym{Y_i^{\mathrm{aut}}}
{Binary autonomous-success trial outcome.}{11}

\Sym{Z_{\mathrm{depth}}(u_{\mathrm{pix}},v_{\mathrm{pix}})}
{Optical-axis depth image. Its values correspond to the camera-coordinate component $Z_{\mathrm C}$ under the stated calibration convention.}{3}

\Sym{z_t,\allowbreak \widehat z_t,\allowbreak \mathbf z_t,\allowbreak \hat{\mathbf z}_t}
{Latent state and its prediction or estimate. Bold notation is used only when a vector representation is explicitly declared.}{2,5,6,7}

\Sym{z_{\mathrm{perf}},\allowbreak \mathbf z_{\mathrm{perf}}}
{Performance output in robust control.}{2}

\Sym{\widehat{\mathbf z}_t^{\mathrm{tac}}}
{Estimated tactile contact state.}{3}

\Sym{z_{1:N_{\mathrm{acttok}}}^{\mathrm{act}}}
{Autoregressive action-token sequence.}{6}

\Sym{z_t^{\mathrm{cmd}}}
{Body-height command.}{7}

\Sym{\mathbf z_{\mathrm{WBC}}}
{Whole-body-control optimization vector.}{7}

\Sym{z}
{Dummy normalized-phase argument in a gait-contact function.}{7}

\Sym{\alpha_{\mathrm{safe}}}
{Locally Lipschitz extended-class-$\mathcal K$ function in a barrier condition.}{2}

\Sym{\alpha_m^{\mathrm{clk}}}
{Relative sensor-clock rate.}{3}

\Sym{\alpha_{\mathrm{scale}}}
{Model-size exponent in a scaling relationship.}{4}

\Sym{\alpha_j,\allowbreak \bar\alpha_j}
{Diffusion retention coefficient and its cumulative product.}{5}

\Sym{\alpha_j^{\mathrm{eq}}}
{Nonnegative physical equality-residual weight.}{5}

\Sym{\alpha_{\mathrm{mix}}}
{Mixing coefficient for autoregressive and flow-matching losses.}{6}

\Sym{\alpha_{\mathrm{act}}}
{Scaling from normalized action to joint-position target.}{7}

\Sym{\beta_m^{\mathrm{clk}}}
{Sensor-clock offset.}{3}

\Sym{\beta_{\mathrm{scale}}}
{Data-size exponent in a scaling relationship.}{4}

\Sym{\beta_{\mathrm{KL}}}
{Variational KL-divergence weight. Changing it from one changes the ordinary ELBO objective.}{5}

\Sym{\beta_j^{\mathrm{ineq}}}
{Nonnegative inequality-constraint penalty weight.}{5}

\Sym{\beta_k^{\mathrm{term}}}
{Option termination condition or termination probability.}{8}

\Sym{\beta_t^{\mathrm{pitch}}}
{Body-pitch command.}{7}

\Sym{\chi_t}
{Waist-yaw command.}{7}

\Sym{\boldsymbol\delta(\mathbf x,t)}
{Optical-tactile surface-deformation field.}{3}

\Sym{\delta}
{Positive position-error tolerance for whole-sequence motion success.}{7}

\Sym{\Delta t,\allowbreak \Delta t_i,\allowbreak \Delta t_t,\allowbreak \Delta_1,\allowbreak \Delta_2}
{Sampling, propagation, control, or transition time intervals under their local event definitions.}{3,4,5,6,7}

\Sym{\Delta J}
{Change in closed-loop performance or utility.}{4,5}

\Sym{\Delta}
{Physical feedback delay producing phase lag $\omega\Delta$.}{9}

\Sym{\Delta_{\mathrm{fwd}}(N_{\mathrm{demo}}),\allowbreak \Delta_{\mathrm{ret}}}
{Forward-transfer gain and retained-capability change, measured as success-rate differences.}{6}

\Sym{\boldsymbol\epsilon_t,\allowbreak \varepsilon_{i_{\mathrm{rgb}}},\allowbreak \varepsilon_I,\allowbreak \boldsymbol\varepsilon_i}
{Measurement noise: a vector in the state-observation model, scalar noise in a raw color channel or optical-tactile image value, and vector taxel-measurement noise, respectively.}{2,3}

\Sym{\boldsymbol\epsilon,\allowbreak \boldsymbol\epsilon_\theta}
{Gaussian corruption noise and predicted noise in generative models.}{5}

\Sym{\boldsymbol\varepsilon,\allowbreak \varepsilon_{ij}}
{The $H\times d$ standard-Gaussian noise matrix used in action flow matching and its independent scalar entries.}{6}

\Sym{\epsilon_{\mathrm{clip}}}
{PPO clipping half-width.}{2}

\Sym{\epsilon_{\mathrm{BC}},\allowbreak \epsilon_{\mathrm{DAgger},N_{\mathrm{iter}}}}
{Behavior-cloning error under the expert distribution, and the best comparison-class average surrogate loss over the state distributions induced by the rollout policies across $N_{\mathrm{iter}}$ iterations, respectively.}{4}

\Sym{\epsilon_{\mathrm{sync}}}
{Clock-synchronization error bound.}{9}

\Sym{\varepsilon_{\mathrm{risk}}}
{Permitted probability of violating at least one constraint in the joint safety event.}{11}

\Sym{\eta_{\mathrm I}(t)}
{Integrated tracking error in PID control.}{2}

\Sym{\eta_{\mathrm{sim}}}
{Parameters of a distribution over simulator parameters.}{4}

\Sym{\eta_e}
{Embodiment description, which may comprise calibration, an embedding, a structured prompt, or inferred information.}{6}

\Sym{\eta_{\mathrm{pen}}}
{Nonnegative violation-penalty coefficient in model-based planning.}{5}

\Sym{\eta_{\mathrm{gear}}}
{Transmission efficiency.}{9}

\Sym{\gamma}
{Reward discount factor, with admissible values determined by the horizon and return definition.}{2,5,6,7}

\Sym{\gamma_{\mathrm{gain}}}
{Disturbance-to-output gain bound in robust control.}{2}

\Sym{\gamma_{\mathrm{cam}}}
{Camera intrinsic skew parameter.}{3}

\Sym{\kappa}
{Conditioning input to an autoregressive, diffusion, flow-matching, or probabilistic prediction model.}{5}

\Sym{\Lambda_{\mathrm{vis}}}
{Wavelength integration domain.}{3}

\Sym{\lambda}
{Optical wavelength in Chapter~\ref{ch:perception}. Under local definitions, it is a physical-loss weight in Chapter~\ref{sec:technical-foundations}, resource-cost coefficient in Chapter~\ref{ch:wm}, and a retargeting smoothness weight in Chapter~\ref{ch:manipulation-models}.}{2,3,5,6}

\Sym{\lambda_j,\allowbreak \lambda_{\mathrm{g}},\allowbreak \lambda_{\mathrm{p}},\allowbreak \lambda_{\mathrm{safe}},\allowbreak \lambda_{\mathrm{e}}}
{Locally defined objective weights, including goal, progress, safety, and energy weights.}{2,5,6,7}

\Sym{\mu,\allowbreak \mu_i}
{Friction coefficient.}{2,7}

\Sym{\mu_{\mathrm{PE}}}
{Positive lower bound in a persistent-excitation matrix inequality.}{2}

\Sym{\mu_\theta,\allowbreak \mu_m,\allowbreak \bar\mu}
{Scalar predictive means or ensemble mean. Vector means use bold type.}{5}

\Sym{\boldsymbol\mu_\theta}
{Action mean or deterministic policy output.}{7}

\Sym{m_t^{\mathrm{meta}}}
{Data-collection metadata.}{6}

\Sym{\boldsymbol\mu_i}
{Moment component of a contact wrench.}{7}

\Sym{\boldsymbol\nu_{\mathcal B},\allowbreak \boldsymbol\nu_{\mathrm o}}
{Six-dimensional base or object twist.}{7}

\Sym{\omega}
{Physical angular frequency.}{2,9}

\Sym{\boldsymbol\omega,\allowbreak \omega_{\mathrm z}^{\mathrm{cmd}},\allowbreak \omega_{\mathrm m},\allowbreak \omega_{\mathrm j}}
{Angular-velocity vector, commanded base-frame angular-velocity $z$ component, or scalar motor/joint angular velocity.}{3,7,9}

\Sym{\omega_{\mathrm{obs}},\allowbreak \omega_{\mathrm{critic}}}
{Observation-model or critic parameters.}{5,6}

\Sym{\Phi_{\mathrm{term}}}
{NMPC terminal cost.}{7}

\Sym{\Phi}
{Standard normal cumulative distribution function in gait-contact smoothing.}{7}

\Sym{\phi,\allowbreak \bar\phi}
{Encoder or variational-model parameters and moving-average target parameters.}{5}

\Sym{\boldsymbol\phi_{\mathrm{PE}}(t)}
{Persistent-excitation regressor.}{2}

\Sym{\phi_{\mathrm{node}}}
{Graph node-update mapping $\mathbb R^{d_{\mathrm{node}}}\times\mathbb R^{d_{\mathrm{msg}}}\to\mathbb R^{d_{\mathrm{node}}}$.}{2}

\Sym{\phi_i,\allowbreak \bar\phi_i}
{Gait phase and its remapped or normalized form.}{7}

\Sym{\boldsymbol\Pi}
{Algebraic Riccati solution matrix in continuous-time LQR.}{2}

\Sym{\pi,\allowbreak \pi_\theta,\allowbreak \pi_k}
{Policy or conditional action distribution. An iteration subscript identifies a learned version.}{2,4,5,6,7,8}

\Sym{\pi_k^{\mathrm{skill}}}
{Internal policy of option or skill $k$.}{8}

\Sym{\psi_{\mathrm{msg}}}
{Graph message mapping $\mathbb R^{d_{\mathrm{node}}}\times\mathbb R^{d_{\mathrm{node}}}\times\mathbb R^{d_{\mathrm{edge}}}\to\mathbb R^{d_{\mathrm{msg}}}$.}{2}

\Sym{\psi_{\mathrm{render}}}
{Simulator observation or rendering parameters.}{4}

\Sym{\psi,\allowbreak \psi_e}
{Task-readout or action-decoder parameters.}{5,6}

\Sym{\psi_t^{\mathrm{phase}}}
{Left-right leg phase offset.}{7}

\Sym{\psi_t^{\mathrm{head}}}
{Root heading angle used in local motion evaluation.}{7}

\Sym{\rho_{i_{\mathrm{src}}}}
{Nonnegative data-source mixture weight, with weights summing to one.}{5}

\Sym{\rho_t^{\mathrm{PPO}}(\theta)}
{New-to-old policy probability ratio in PPO.}{2}

\Sym{\rho_e}
{Embodiment-specific observation-encoder parameters.}{6}

\Sym{\rho_k^{\mathrm{init}}}
{Initial-state distribution of a benchmark protocol.}{11}

\Sym{\sigma_{\mathrm{rad}}}
{Volumetric density in a radiance field.}{2,3}

\Sym{\boldsymbol\sigma_{\mathrm c},\allowbreak \overline{\boldsymbol\sigma}_i}
{Contact-stress vector and its taxel-area average.}{3}

\Sym{\sigma,\allowbreak \bar\sigma}
{Singular value and largest-singular-value operator.}{2}

\Sym{\sigma_{\mathrm{flow}}}
{Flow interpolation time increasing from clean data to noise.}{5}

\Sym{\sigma_\theta,\allowbreak \sigma_m^2}
{Predictive standard deviation or variance.}{5}

\Sym{\sigma_\phi}
{Positive, dimensionless smoothing width in the gait-contact function built from the standard normal cumulative distribution.}{7}

\Sym{\sigma_k^{\mathrm{track}}}
{Positive motion-tracking reward scale.}{7}

\Sym{\sigma_k^{\mathrm{opt}}}
{Option tuple comprising initiation set, internal policy, and termination condition.}{8}

\Sym{\sigma_{\mathrm{contact}}}
{Contact mode or support schedule.}{7}

\Sym{\boldsymbol\tau,\allowbreak \tau_{\mathrm j},\allowbreak \boldsymbol\tau_{\mathrm S}}
{Joint-torque vector, scalar joint-output torque in the transmission model, and moment vector in a sensor-frame wrench, respectively.}{2,3,7,9}

\Sym{\boldsymbol\tau^{\mathrm{cmd}},\allowbreak \boldsymbol\tau^{\mathrm{ff}}}
{Commanded joint-torque vector and feedforward joint-torque vector, respectively. The commanded torque combines feedforward and local feedback terms.}{7}

\Sym{\tau_{\mathrm{ep}}}
{Recorded interaction episode or trajectory.}{2,4,6,7}

\Sym{\tau}
{Dummy time variable in a time integral. Distinct from an interaction trajectory $\tau_{\mathrm{ep}}$.}{2,3,7}

\Sym{\tau_{\mathrm{flow}}}
{Flow interpolation time increasing from noise to clean data.}{6}

\Sym{\tau_{\mathrm{stride}}}
{Future-video feature-sampling stride.}{5}

\Sym{\theta}
{Trainable neural-model, policy, or dynamics parameters.}{2,4,5,6,7,9,11}

\Sym{\boldsymbol\theta,\allowbreak \hat{\boldsymbol\theta}}
{Unknown adaptive-control parameters and their estimates.}{2}

\Sym{\boldsymbol\theta_{\mathrm{tac}}}
{Physical or geometric parameters of an optical-tactile model.}{3}

\Sym{\Theta}
{Complete set of model or policy parameter groups under the local formulation.}{5,6}

\Sym{\xi_{\mathrm{sim}},\allowbreak \boldsymbol\xi_{\mathrm{sim}}}
{Physical or numerical simulator parameters under the stated parameterization.}{4,7}

\Sym{\xi_{\mathrm{sample}}}
{Joint-training record or sample.}{5}

\Sym{\zeta_t^{\mathrm{exec}}}
{Execution-supervision record of skill state, verified subgoals, and constraint history.}{8}

\Sym{\zeta_{\mathrm{sample}}}
{Weighted training sample in the hardware-software discussion.}{9}

\Sym{\zeta_{\mathrm{traj}}}
{Sampled interaction trajectory in motion-tracking reinforcement learning.}{7}

\Sym{\mathcal A,\allowbreak \mathcal A_e}
{Policy action space or embodiment-specific action space.}{2,6}

\Sym{\mathcal A_H}
{Declared set of input or action sequences over horizon $H$.}{5}

\Sym{\mathcal A_k^{\mathrm{acq}}}
{Data-acquisition plan at iteration $k$.}{4}

\Sym{\mathcal B_k^{\mathrm{data}}}
{Batch of newly collected physical interaction data.}{4}

\Sym{\mathcal B_k^{\mathrm{bench}}}
{Task-scenario-embodiment benchmark-protocol tuple.}{11}

\Sym{\mathcal C_{\mathrm{safe}}}
{Barrier-defined safe state set.}{2}

\Sym{\mathcal C_{\mathrm{exec}}}
{Execution-constraint set in a task specification.}{8}

\Sym{\mathcal C_{\mathrm{cand}}}
{Set of candidate world-model rollouts.}{5}

\Sym{\mathcal C_k^{\mathrm{ctrl}}}
{Benchmark controller and protection configuration.}{11}

\Sym{\mathcal D,\allowbreak \mathcal D_{\mathrm r},\allowbreak \mathcal D_e,\allowbreak \mathcal D_{i_{\mathrm{src}}}}
{Dataset or empirical distribution with locally defined source, task, or embodiment subscripts.}{4,5,6,7,9}

\Sym{\mathcal D_{\mathrm{Lyap}}}
{Domain of a Lyapunov analysis.}{2}

\Sym{\mathcal E_{\mathrm{eval}}}
{Isolated held-out evaluation set in a data-improvement loop.}{4}

\Sym{\mathcal E_{\mathrm{emb}},\allowbreak \mathcal E_{\mathrm{emb}}^{\mathrm{tgt}},\allowbreak \mathcal E_{\mathrm{ret}}}
{Embodiment-identifier set, target-embodiment set used in transfer evaluation, and finite nonempty set of previously supported embodiments used for retention evaluation, respectively.}{6}

\Sym{\mathcal F_{\mathrm{opt}}}
{Optical-tactile image-formation function.}{3}

\Sym{\mathcal F_h,\allowbreak \mathcal F_g,\allowbreak \mathcal F_a}
{Mirror transformations of history, goals, and actions.}{7}

\Sym{\mathcal G_{\mathrm{goal}}}
{Permitted goal or outcome set in a task specification.}{8}

\Sym{\mathcal H_k}
{Benchmark hardware and calibration configuration.}{11}

\Sym{\mathcal I_k^{\mathrm{init}}}
{Initiation set of option $k$.}{8}

\Sym{\mathcal I_{\mathrm{track}}}
{Tracking-feature or reward-term index set.}{7}

\Sym{\mathcal I_k^{\mathrm{io}}}
{Benchmark observation-action interface.}{11}

\Sym{\mathcal J}
{Joint-training loss-term or supervision-target index set.}{5}

\Sym{\mathcal K}
{Class of comparison functions used in stability or barrier conditions.}{2}

\Sym{\mathcal K_{\mathrm{contact}},\allowbreak \mathcal K_{\mathrm{hand}}}
{Active point-contact index set in the floating-base body model, and active finger-object contact index set in the hand-object model, respectively.}{7}

\Sym{\mathcal K_e^{\mathrm{eval}}}
{Evaluation-task set for embodiment $e$.}{6}

\Sym{\mathcal L}
{Training objective, with subscripts specifying its data, dynamics, or representation terms.}{2,5,6,7,9}

\Sym{\mathcal L_k^{\mathrm{log}}}
{Traceable evidence record in a benchmark protocol.}{11}

\Sym{\mathcal L_{\mathbf m}^{\mathrm{links}}}
{Selected-link or keypoint set induced by a command mask.}{7}

\Sym{\mathcal L_{\mathrm{sym}}}
{Mirror-consistency training regularizer.}{7}

\Sym{\mathcal L_{\mathrm{change}}}
{Clean-latent change-supervision loss.}{5}

\Sym{\mathcal L_{\mathrm{align}}}
{Directional token-to-target feature-alignment loss.}{5}

\Sym{\mathcal L_{\mathrm{geom}}}
{Teacher geometry/motion-descriptor distillation loss.}{5}

\Sym{\mathcal L_{\mathrm{WAM}}}
{Combined video, action, change, alignment, and geometry objective.}{5}

\Sym{\mathcal L_{\mathrm{phys}},\allowbreak \mathcal L_{\mathrm{physics}}}
{Physical residual or constraint loss, with its local residual and sampling domain.}{2,5}

\Sym{\mathcal L_{\mathrm{Gauss}}}
{Heteroscedastic Gaussian prediction loss.}{5}

\Sym{\mathcal L_{\mathrm{multi}}}
{Multi-horizon prediction loss.}{5}

\Sym{\mathcal L_{\mathrm{pre}}}
{Pretraining objective over available action, state, progress, outcome, and language supervision.}{6}

\Sym{\mathcal L_{\mathrm{new}},\allowbreak \mathcal L_{\mathrm{retain}},\allowbreak \mathcal L_{\mathrm{reg}}}
{New-task, capability-retention, and regularization terms in continual learning.}{6}

\Sym{\mathcal L_{\mathrm{adapt}},\allowbreak \Theta_{\mathrm{upd}}}
{Embodiment-adaptation objective and the selected parameter subset allowed to update.}{6}

\Sym{\mathcal M,\allowbreak \mathcal M_e}
{MDP or POMDP environment tuple, with an embodiment subscript where applicable.}{2,6}

\Sym{\mathcal M_{\mathrm{mask}}}
{Mask specifying target positions or withheld regions in representation training. Expectations average over its sampling procedure.}{5}

\Sym{\mathcal M_{\mathrm{motion}}}
{Reference motion clip or ordered reference-state sequence sampled from motion data.}{7}

\Sym{\mathcal M_k^{\mathrm{meas}}}
{Benchmark measurement and metric definitions.}{11}

\Sym{\mathcal N_{\mathrm{PDE}}[u^{\mathrm{field}}]}
{Partial-differential-equation residual operator.}{2}

\Sym{\mathcal N}
{Normal or Gaussian distribution.}{5,6}

\Sym{\mathcal N_{\mathrm{demo}}\subseteq\mathbb N_0}
{Tested set of nonnegative integer demonstration budgets.}{6}

\Sym{\mathcal O_e}
{Embodiment-specific observation space. Distinct from an observation likelihood $O_e$.}{6}

\Sym{\mathcal P_{\mathrm{cloud}}}
{Point cloud.}{2}

\Sym{\mathcal P_{\mathrm{pref}}}
{Preferences among acceptable task solutions.}{8}

\Sym{\mathcal Q}
{Prediction-query set.}{5}

\Sym{\mathcal S,\allowbreak \mathcal S_e}
{Physical-state space of a sequential-decision model.}{2,6}

\Sym{\mathcal T_{\mathrm{task}}}
{Task specification comprising goals, constraints, preferences, and verification.}{8}

\Sym{\mathcal T_{\mathrm{WBC}}}
{Whole-body-control task index set.}{7}

\Sym{\mathcal U}
{Admissible continuous-control set.}{2}

\Sym{\operatorname{Unif}[0,\allowbreak 1]}
{Uniform distribution on the unit interval.}{5}

\Sym{\mathcal U_k^{\mathrm{budget}}}
{Benchmark resource and operating budgets.}{11}

\Sym{\mathcal V_{t:t+H}^{\mathrm{viol}}}
{Event that at least one trajectory-level constraint is violated. Its indicator is not a violation count.}{5}

\Sym{\mathcal V_{\mathrm{verify}}}
{Task-outcome and constraint-verification specification.}{8}

\Sym{\mathcal W_i^{\mathrm{wrench}}(\sigma_{\mathrm{contact}})}
{Admissible foot-wrench set under contact mode $\sigma_{\mathrm{contact}}$. The swing-foot set contains only the zero wrench.}{7}

\Sym{\mathcal X_{\mathrm{adm}},\allowbreak \mathcal X_{\mathrm{term}}}
{Admissible state set and terminal feasible set in MPC.}{2}

\Sym{\mathcal X}
{Physical-agent state space.}{8}

\Sym{{}'}
{Transformed quantity or successor-state integration variable. Not necessarily a derivative.}{2,5}

\Sym{{}^+}
{Future-variable label.}{5,6}

\Sym{{}^\star}
{Optimal-solution or target-object marker, as specified by the optimization or task definition.}{4,6,7}

\Sym{{}^{(b)},\allowbreak {}^{(j)}}
{Rollout or diffusion-step indices, not powers.}{5}

\Sym{[u]_+}
{Positive part, $\max(u,0)$.}{5}

\Sym{\lceil u\rceil}
{Ceiling, the smallest integer not less than $u$.}{6}

\Sym{\bmod 1}
{Wrapping of normalized phase into a unit interval.}{7}

\Sym{{}^{\mathrm S_0}\mathbf T_{\mathrm S_m}}
{Transformation from sensor frame $\mathrm S_m$ to reference frame $\mathrm S_0$. $m$ is a variable index.}{3}

\Sym{\mathbf R_{\mathrm{CW}},\allowbreak \mathbf t_{\mathrm{CW}}}
{Rotation and translation mapping world coordinates to camera coordinates.}{3}

\Sym{\mathbf p_{\mathrm C},\allowbreak \mathbf p_{\mathrm W}}
{Position coordinates expressed in camera and world frames.}{3,8}

\Sym{{}^\top}
{Transpose.}{2,3,5,6,7}

\Sym{\mathbf 0}
{Zero vector or zero block of the required dimensions.}{2,7}

\Sym{\mathbf 1\{\cdot\}}
{Event indicator, equal to one when its condition holds and zero otherwise.}{5}

\Sym{\arg\min,\allowbreak \arg\max}
{Minimizing or maximizing argument, respectively.}{4}

\Sym{\min,\allowbreak \max}
{Minimum or maximum value, including branches of clipping functions.}{2,5,9}

\Sym{\cup,\allowbreak \cap,\allowbreak \in,\allowbreak \subseteq,\allowbreak \setminus}
{Set union, intersection, membership, inclusion, and difference.}{2,4}

\Sym{\mathbb E}
{Expectation under the stated distribution.}{2,4,5,6,7,9}

\Sym{\operatorname{Var}}
{Variance.}{5}

\Sym{\mathrm{G\text{-}MPKPE}}
{Mean position error in the world frame, measured in meters.}{7}

\Sym{\mathrm{L\text{-}MPKPE}}
{Mean position error after expressing positions relative to horizontal root translation and heading.}{7}

\Sym{\mathrm{G\text{-}MPKRE}}
{Mean world-frame orientation error measured by geodesic angle on $\mathrm{SO}(3)$.}{7}

\Sym{\mathrm{L\text{-}MPKRE}}
{Mean orientation error in the local root-heading frame.}{7}

\Sym{\hat{\mathbf{x}}_{\mathrm{now}},\allowbreak \hat x_t,\allowbreak \widehat y_{t+k\mid t}}
{Current MPC state estimate in Chapter~\ref{sec:technical-foundations}, estimated joint state in Chapter~\ref{ch:evaluation}, or predicted readout in Chapter~\ref{ch:wm}, respectively. $t+k\mid t$ denotes prediction for $t+k$ made at $t$.}{2,5,11}

\Sym{\widehat{\mathbf p}_i^{\mathrm{human}}}
{Human reference point used in motion retargeting.}{6}

\Sym{\land,\allowbreak \forall}
{Logical conjunction and universal quantification.}{8}

\Sym{\mathbb N_0}
{Nonnegative integers.}{6}

\Sym{\nabla}
{Gradient with respect to the indicated state or parameter variables.}{2}

\Sym{\mathrm d/\mathrm dt,\allowbreak \dot x,\allowbreak \ddot x}
{Time derivative, first time derivative, and second time derivative.}{2,3,7}

\Sym{\mathrm d\mathbf W_t^{\mathrm{Wiener}}}
{Increment of the $d_{\mathrm{noise}}$-dimensional Wiener process in the stochastic differential equation.}{2}

\Sym{\odot}
{Elementwise product, including application of a command mask.}{7}

\Sym{\operatorname{sg}}
{Stop-gradient operation. It does not imply that teacher parameters never update.}{5,7}

\Sym{\operatorname{clip}}
{Clipping to the explicitly specified interval.}{2,7}

\Sym{\operatorname{sat}}
{Actuator-command saturation.}{7}

\Sym{\operatorname{Collect}}
{Physical interaction-data collection.}{4}

\Sym{\operatorname{Curate}}
{Data-quality assessment, filtering, and organization.}{4}

\Sym{\operatorname{TrainAndCalibrate}}
{Policy training and model calibration in the data-improvement loop.}{4}

\Sym{\operatorname{Accept}}
{Held-out closed-loop acceptance evaluation.}{4}

\Sym{\operatorname{Update},\allowbreak \operatorname{Supervise}}
{Belief update and execution supervision, respectively.}{8}

\Sym{\operatorname{cos}_{\mathrm{sim}}(\mathbf u,\mathbf v)}
{Cosine similarity of two feature vectors.}{5}

\Sym{\exp,\allowbreak \log,\allowbreak \log_{10}}
{Exponential, locally specified logarithm, and base-ten logarithm.}{2,4,5,6,7,12}

\Sym{\arccos,\allowbreak \operatorname{tr}}
{Inverse cosine and matrix trace, used in rotation-angle evaluation.}{7}

\Sym{\sin,\allowbreak \cos}
{Trigonometric sine and cosine of a scalar angle.}{7}

\Sym{\operatorname{Inside},\allowbreak \operatorname{Supported},\allowbreak \operatorname{Released},\allowbreak \operatorname{Upright},\allowbreak \operatorname{Closed}}
{Task predicates for placement and cabinet manipulation.}{8}

\Sym{\operatorname{Succ}(\delta)}
{Fraction of trials whose maximum selected-link position error over the entire evaluation horizon is within $\delta$.}{7}

\Sym{\Pr}
{Probability of an event.}{11}

\Sym{p(\cdot\mid\cdot)}
{Conditional density, likelihood, or posterior under the stated measure and context.}{2,4,5,6,7,8}

\Sym{\mathbb R,\allowbreak \mathbb R^n,\allowbreak \mathbb R^{m\times n}}
{Real numbers, real coordinate vectors, and real matrices of the stated dimensions.}{2,3,6,7}

\Sym{\to,\allowbreak \longrightarrow}
{Direction of time, control, or process flow. Not an identity relation.}{7,8}

\Sym{\leftrightarrow}
{Mutual dependency in a system diagram, unless a logical equivalence is explicitly defined.}{8}

\Sym{\operatorname{RMSE}(k)}
{Root mean squared prediction error at horizon $k$.}{5}

\Sym{\operatorname{BS}}
{Brier score for binary event predictions.}{5}

\Sym{\operatorname{CRPS}(F_{\mathrm{CDF}},y)}
{Continuous ranked probability score of predictive cumulative distribution $F_{\mathrm{CDF}}$ and scalar outcome $y$.}{5}

\Sym{\sim}
{Sampling from or following the specified probability distribution.}{2,4,5}

\Sym{\propto}
{Proportionality. In homogeneous-coordinate projection, it denotes equality up to a nonzero scale. It also expresses a posterior proportional to likelihood times prior.}{3,4}

\Sym{\approx,\allowbreak \simeq}
{Approximate equality under the local modeling assumptions.}{5,9}

\Sym{\operatorname{SR}_{N_{\mathrm{demo}}}(e,k),\allowbreak \operatorname{SR}_{\mathrm{before}},\allowbreak \operatorname{SR}_{\mathrm{after}}}
{Adaptation or transfer success rates. $N_{\mathrm{demo}}$ is the adaptation-demonstration budget. Evaluation trials are counted by $N_{\mathrm{eval}}$.}{6}

\Sym{\succeq,\allowbreak \succ}
{Positive-semidefinite and positive-definite order for symmetric matrices.}{2}

\Sym{\leq,\allowbreak \geq}
{Scalar or componentwise inequalities, as defined by the operands.}{2,5,7}

\Sym{\sum,\allowbreak \prod,\allowbreak \int,\allowbreak \sup,\allowbreak \lim}
{Summation, product, integration, supremum, and limit.}{2}

\Sym{\tilde\pi}
{Candidate policy in simulator-assisted data or policy improvement.}{4}

\Sym{\mathrm{Hz},\allowbreak \mathrm{ms}}
{Hertz and millisecond.}{6,7,9,11}

\Sym{\mathrm m,\allowbreak \mathrm{mm},\allowbreak \mathrm s,\allowbreak \mathrm{kg}}
{Meter, millimeter, second, and kilogram.}{7}

\Sym{\|\cdot\|_2,\allowbreak \|\cdot\|^2}
{Euclidean norm and squared Euclidean error for finite-dimensional vectors. The robust-control paragraph uses the subscript $2$ for time-signal norms, as listed below. Matrix norms require a separate local definition.}{2}

\Sym{\|\cdot\|_{\mathrm F}}
{Frobenius norm of an explicitly matrix-valued action residual. Its square is the sum of squared scalar entries.}{6}

\Sym{\|\mathbf T_{zw}\|_\infty}
{For a stable proper LTI transfer matrix, the supremum over frequency of its largest singular value. Equal to its induced $L_2$ gain under zero initial conditions.}{2}

\Sym{\|\mathbf w\|_2,\allowbreak \|\mathbf z_{\mathrm{perf}}\|_2}
{Finite-energy time-signal norms in the robust-control gain bound. For example, $\|\mathbf w\|_2=(\int_0^\infty \mathbf w(t)^\top\mathbf w(t)\,\mathrm dt)^{1/2}$.}{2}
\end{longtable}
\endgroup
\clearpage

\clearpage
\phantomsection
\addcontentsline{toc}{section}{References}
\bibliographystyle{abbrvnat}
\bibliography{references}
\end{document}